\documentclass[12pt,oneside,letterpaper]{memoir}

\setsecnumdepth{subsubsection}
\maxtocdepth{subsection}

\usepackage{subcaption}

\usepackage[utf8]{inputenc}
\usepackage[T1]{fontenc}

\usepackage{natbib}

\IfFileExists{appendix.sty}{%
\usepackage{appendix}%
}{}

\DisemulatePackage{setspace}
\usepackage[onehalfspacing]{setspace}

\usepackage{geometry}

\IfFileExists{enumitem.sty}{%
\usepackage[loadonly]{enumitem}[2007/06/30]%
\newlist{hanglist}{enumerate}{1}%
\setlist[hanglist]{label=\arabic*.}%
\setlist[hanglist,1]{leftmargin=0pt}%
}{%
}

\usepackage{amssymb}
\usepackage{amsmath}
\usepackage{amsthm}
\usepackage[colorlinks,citecolor=blue!70!black,linkcolor=blue!70!black,
urlcolor=blue!70!black,breaklinks=true]{hyperref}

\IfFileExists{mathpartir.sty}{%
\usepackage{mathpartir}%
}{}

\IfFileExists{listings.sty}{%
\usepackage{listings}%
}{}

\hypersetup{pdfborder = 0 0 0}

\IfFileExists{microtype.sty}{%
\usepackage[protrusion=true,expansion=true]{microtype}%
}{}

\usepackage{boxedminipage}

\IfFileExists{booktabs.sty}{%
\usepackage{booktabs}%
}{}

\usepackage{ifpdf}

\usepackage{lmodern}
\usepackage[LY1]{fontenc}

\IfFileExists{tgpagella.sty}{%
\usepackage{tgpagella}%
}{}

\usepackage[LY1]{eulervm}

\ifpdf
\usepackage[pdftex]{graphicx}
\else
\usepackage{graphicx}
\fi

\usepackage{makeidx}
\makeindex

{\theoremstyle{plain}

}
{\theoremstyle{remark}

}

\newtheoremstyle{customsty1}
{3pt}%
{3pt}%
{}%
{}%
{\bfseries}%
{:}%
{.5em}%
{}%

{\theoremstyle{customsty1} 
}

\makechapterstyle{deposit}{%

}

\makepagestyle{deposit}
 
\makeatletter

\makeevenfoot{deposit}{}{}{}
\makeoddfoot{deposit}{}{}{}
\makeevenhead{deposit}{\thepage}{}{}
\makeoddhead{deposit}{}{}{\thepage}
\makeatother

\copypagestyle{chapter}{plain}
\makeoddfoot{chapter}{}{}{}
\makeevenhead{chapter}{\thepage}{}{}
\makeoddhead{chapter}{}{}{\thepage}

\makeatletter
\newenvironment{wb-bib}[1]{%
  \chapter*{references}
\ifnobibintoc\else 
\phantomsection 
\addcontentsline{toc}{chapter}{References} 
\fi 
\prebibhook
  \begin{bibitemlist}{#1}}{\end{bibitemlist}\postbibhook}

\AtBeginDocument{%
  \@ifpackageloaded{natbib}{%
    \addtodef{\endthebibliography}{}{\vskip-\lastskip\postbibhook}
    \@ifpackagewith{natbib}{sectionbib}{%
      \renewcommand{\bibsection}{\@memb@bsec}}%
      {\renewcommand{\bibsection}{\@memb@bchap}}}%
  {}
  \@ifpackagewith{chapterbib}{sectionbib}{%
    \renewcommand{\sectionbib}[2]{}
    \renewcommand{\bibsection}{\@memb@bsec}}{}
}
\makeatother

\makeatletter
\newcommand{\wb@episource}{}
\newenvironment{wbepi}[1]{\begin{quote}\renewcommand{\wb@episource}{#1}\itshape}{\par\upshape \raggedleft --- \textsc{\wb@episource}\\ \end{quote}}
\makeatother

\IfFileExists{svn-multi.sty}{%
\usepackage{svn-multi}%
\newcommand{\vcinfo}{}%
}{%
\newcommand{\svnidlong}[4]{}%
\newcommand{\vcinfo}{}%
}

\usepackage{parskip}

\usepackage{booktabs}       %
\usepackage{microtype}      %

\usepackage{amsmath, amsthm, amssymb}
\usepackage{dsfont} 
\usepackage{url}            %
\usepackage{algorithm}
\usepackage[noend]{algorithmic}
\usepackage{listings}
\usepackage{bm}
\usepackage{bbm}
\usepackage{enumitem}
\usepackage{nicefrac}       %
\usepackage{mathrsfs} 
\usepackage{inconsolata}
\usepackage{comment}
\usepackage{xspace}
\usepackage{transparent}
\usepackage[nameinlink, capitalize]{cleveref}
\usepackage{thmtools}
\usepackage{thm-restate}
\usepackage{arydshln}
\usepackage{mathtools}
\usepackage{lineno}
\usepackage{xspace}
\usepackage{graphicx}
\usepackage{bigints}
\usepackage{seqsplit}

\usepackage{enumitem}
\setlist[enumerate]{leftmargin=.2in}
\setlist[itemize]{leftmargin=.2in}

\makeatletter

\newcommand\@thesistitlemedskip{0.25in}
\newcommand\@thesistitlebigskip{0.5in}
\newcommand{\degree}[1]{\gdef\@degree{#1}}
\newcommand{\project}{\gdef\@doctype{A masters project report}}
\newcommand{\prelim}{\gdef\@doctype{A preliminary report}}
\newcommand{\thesis}{\gdef\@doctype{A thesis}}
\newcommand{\dissertation}{\gdef\@doctype{A dissertation}}
\newcommand{\department}[1]{\gdef\@department{(#1)}}
\newcommand{\oralexamdate}[1]{\gdef\@oralexamdate{#1}} 
\newcommand{\committeeone}[1]{\gdef\@committeeone{#1}}
\newcommand{\committeetwo}[1]{\gdef\@committeetwo{#1}}
\newcommand{\committeethree}[1]{\gdef\@committeethree{#1}}
\newcommand{\committeefour}[1]{\gdef\@committeefour{#1}}
\newcommand{\committeefive}[1]{\gdef\@committeefive{#1}}
\newcommand{\committeesix}[1]{\gdef\@committeesix{#1}}
\newcommand{\committeeseven}[1]{\gdef\@committeeseven{#1}}

\newenvironment{titlepage}
 {\@restonecolfalse\if@twocolumn\@restonecoltrue\onecolumn
  \else \newpage \fi \thispagestyle{empty}
}{\if@restonecol\twocolumn \else \newpage \fi}

\gdef\@degree{Doctor of Philosophy}    %
\gdef\@doctype{A dissertation}         %

\gdef\@department{(Electrical Engineering)} %
\gdef\@oralexamdate{}
\gdef\@committeeone{}
\gdef\@committeetwo{}
\gdef\@committeethree{}
\gdef\@committeefour{}
\gdef\@committeefive{}
\gdef\@committeesix{}
\gdef\@committeeseven{}

\renewcommand{\maketitle}{%
  \begin{titlepage}
    \def\thanks##1{\typeout{Warning: `thanks' deleted from thesis titlepage.}}
    \let\footnotesize\small \let\footnoterule\relax \setcounter{page}{1}

    \begin{center}
      {\textbf{\expandafter\uppercase\expandafter{\@title}}} \\[\@thesistitlebigskip]
       by \\[\@thesistitlemedskip]
      \@author \\[\@thesistitlebigskip]
      \@doctype\ submitted in partial fulfillment of \\
      the requirements for the degree of\\[\@thesistitlebigskip]
      \@degree \\[\@thesistitlemedskip]
      \@department \\[\@thesistitlebigskip]
      at the \\[\@thesistitlebigskip]
      UNIVERSITY OF WISCONSIN--MADISON\\[\@thesistitlebigskip]
      \@date \\[\@thesistitlebigskip]
    \end{center}

\noindent Date of final oral examination: \@oralexamdate \hspace*{\fill} \\[\@thesistitlemedskip]
\noindent The dissertation is approved by the following members of the Final Oral Committee:\\*
\indent \@committeeone\\*
\indent \@committeetwo\\*
\indent \@committeethree\\*
\indent \@committeefour\\*
\indent \@committeefive\\*
\indent \@committeesix %

  \end{titlepage}

  \setcounter{footnote}{0}
  \setcounter{page}{1} %
  \let\thanks\relax
  \let\maketitle\relax \let\degree\relax \let\project\relax \let\prelim\relax
  \let\department\relax
  \gdef\@thanks{}\gdef\@degree{}\gdef\@doctype{}
  \gdef\@department{}
}

\def\abstract{
  \chapter*{Abstract}
  \addcontentsline{toc}{chapter}{Abstract}
  \relax\markboth{Abstract}{Abstract}}

\def\advisortitle#1{\gdef\@advisortitle{#1}}
\def\advisorname#1{\gdef\@advisorname{#1}}
\gdef\@advisortitle{Professor}
\gdef\@advisorname{Cheer E.\ Place}

\def\umiabstract{
             \thispagestyle{empty}
                  \addtocounter{page}{-1}
                \begin{center}
                  {\textbf{\expandafter\uppercase\expandafter{\@title}}}\\
                  \vspace{12pt}
                  \@author \\
                  \vspace{12pt}
                  Under the supervision of \@advisortitle\ \@advisorname\\
                  At the University of Wisconsin-Madison
                \end{center}
}

\def\endumiabstract{\vfill \hfill\@advisorname\par\newpage}

\def\verbatimfile#1{\begingroup \singlespace
                    \@verbatim \frenchspacing \@vobeyspaces
                    \input#1 \endgroup
}

\def\copyrightpage{
  \newpage
  \thispagestyle{empty}    %
  \addtocounter{page}{-1}
  \chapter*{}            %
  \begin{center}
   \vfill
   \copyright\ Copyright by \@author\ \@date\\
   All Rights Reserved
  \end{center}}

\def\dedication{
  \newpage
  \null\vfil
  \begin{center}}
\def\enddedication{\end{center}\par\vfil\newpage}

\newcount\@testday
\def\today{\@testday=\day
  \ifnum\@testday>30 \advance\@testday by -30
  \else\ifnum\@testday>20 \advance\@testday by -20
  \fi\fi
  \number\day\ \
  \ifcase\month\or
    January \or February \or March \or April \or May \or June \or
    July \or August \or September \or October \or November \or December
    \fi\ \number\year
}

\newtheorem{theorem}{Theorem}[chapter]

\newtheorem{proposition}[theorem]{Proposition}
\newtheorem{corollary}[theorem]{Corollary}
\newtheorem{definition}[theorem]{Definition}
\newtheorem{example}[theorem]{Example}

\newtheorem{lemma}[theorem]{Lemma}
\newtheorem{assumption}[theorem]{Assumption}

\newtheorem{remark}[theorem]{Remark}

\def\@bibchaptitle{Bibliography}
\def\altbibtitle{\def\@bibchaptitle{Bibliography}}
\def\thebibliography#1{
  \global\@bibpresenttrue
  \chapter*{\@bibchaptitle\markboth{\@bibchaptitle}{\@bibchaptitle}}
  \addcontentsline{toc}{chapter}{\@bibchaptitle}
  \vspace{0.375in}    %
  \interlinepenalty=10000 %
  \singlespace\list
  {[\arabic{enumi}]}{\settowidth\labelwidth{[#1]}\leftmargin\labelwidth
    \advance\leftmargin\labelsep \usecounter{enumi}}
  \def\newblock{\hskip .11em plus .33em minus -.07em}
  \sloppy
  \sfcode`\.=1000\relax}
\let\endthebibliography=\endlist

\makeatother

\renewcommand{\cite}[1]{\citet{#1}}

\def\E{{\mathbb E}}
\def\V{{\mathbb V}}
\def\P{{\mathbb P}}
\def\Prob{\P}
\def\Q{{\mathbb Q}}
\def\R{{\mathbb R}}
\def\N{{\mathbb N}}
\def\B{{\mathbb B}}
\def\S{{\mathbb S}}

\newcommand{\mfe}{\mathfrak{e}}
\newcommand{\mfs}{\mathfrak{s}}
\newcommand{\mfF}{\mathfrak{F}}

\newcommand{\alg}{\mathscr{A}}
\newcommand{\simp}{\Delta}

\DeclareMathOperator*{\argmax}{arg\,max}
\DeclareMathOperator*{\argmin}{arg\,min}
\DeclareMathOperator{\esup}{ess\,sup}

\DeclareMathOperator{\spn}{span}

\DeclareMathOperator{\sgn}{sign}

\DeclareMathOperator{\err}{err}
\DeclareMathOperator{\poly}{poly}
\DeclareMathOperator{\supp}{supp}

\DeclareMathOperator{\opt}{{opt}}
\DeclareMathOperator{\unif}{{unif}}
\DeclareMathOperator{\conv}{{conv}}
\DeclareMathOperator{\hard}{{hard}}
\DeclareMathOperator{\easy}{{easy}}
\DeclareMathOperator{\flex}{{flexible}}

\DeclareMathOperator{\stan}{{standard}}
\DeclareMathOperator{\chow}{{Chow}}

\DeclareMathOperator{\1}{{\mathds{1}}}

\DeclareMathOperator{\val}{{val}}

\DeclareMathOperator{\dom}{{dom}}
\DeclareMathOperator{\kl}{{KL}}

\newcommand{\astar}{a^{\star}}
\newcommand{\fstar}{f^{\star}}
\newcommand{\gstar}{g^{\star}}

\newcommand{\fhat}{\wh{f}}
\newcommand{\ghat}{\wh{g}}
\newcommand{\ahat}{\wh{a}}
\newcommand{\phibar}{\wb{\phi}}
\newcommand{\abar}{\wb{a}}
\newcommand{\vepsbar}{\wb{\veps}}
\newcommand{\acheck}{\check{a}}

\newcommand{\trn}{\top}

\newcommand{\approxleq}{\lesssim}

\newcommand{\bigoh}{O}
\newcommand{\bigoht}{\wt{O}}
\newcommand{\bigom}{\Omega}
\newcommand{\bigomt}{\wt{\Omega}}

\newcommand{\polylog}{\mathrm{polylog}}
\renewcommand{\epsilon}{\varepsilon}

\newcommand{\indic}{\mathbb{I}}

\newcommand{\algcommentlight}[1]{\textcolor{blue!70!black}{\transparent{0.5}\small{\texttt{\textbf{//\hspace{2pt}#1}}}}}

\DeclarePairedDelimiter{\abs}{\lvert}{\rvert} %
\DeclarePairedDelimiter{\brk}{[}{]}
\DeclarePairedDelimiter{\crl}{\{}{\}}
\DeclarePairedDelimiter{\prn}{(}{)}
\DeclarePairedDelimiter{\nrm}{\|}{\|}
\DeclarePairedDelimiter{\ang}{\langle}{\rangle}

\DeclarePairedDelimiter{\ceil}{\lceil}{\rceil}
\DeclarePairedDelimiter{\floor}{\lfloor}{\rfloor}
\DeclarePairedDelimiter{\tri}{\langle}{\rangle}

\DeclarePairedDelimiterX{\infdiv}[2]{(}{)}{%
  #1\;\delimsize\|\;#2%
}

\newcommand{\curly}{\crl}
\newcommand{\paren}{\prn}
\newcommand{\norm}{\nrm}
\newcommand{\sq}{\brk}

\newcommand{\wt}[1]{\widetilde{#1}}
\newcommand{\wh}[1]{\widehat{#1}}
\newcommand{\wb}[1]{\widebar{#1}}

\newcommand{\by}{\bm{y}}
\newcommand{\bpsi}{\bm{\psi}}

\def\ddefloop#1{\ifx\ddefloop#1\else\ddef{#1}\expandafter\ddefloop\fi}
\def\ddef#1{\expandafter\def\csname bb#1\endcsname{\ensuremath{\mathbb{#1}}}}
\ddefloop ABCDEFGHIJKLMNOPQRSTUVWXYZ\ddefloop
\def\ddefloop#1{\ifx\ddefloop#1\else\ddef{#1}\expandafter\ddefloop\fi}
\def\ddef#1{\expandafter\def\csname b#1\endcsname{\ensuremath{\mathbf{#1}}}}
\ddefloop ABCDEFGHIJKLMNOPQRSTUVWXYZ\ddefloop
\def\ddef#1{\expandafter\def\csname sf#1\endcsname{\ensuremath{\mathsf{#1}}}}
\ddefloop ABCDEFGHIJKLMNOPQRSTUVWXYZ\ddefloop
\def\ddef#1{\expandafter\def\csname c#1\endcsname{\ensuremath{\mathcal{#1}}}}
\ddefloop ABCDEFGHIJKLMNOPQRSTUVWXYZ\ddefloop
\def\ddef#1{\expandafter\def\csname h#1\endcsname{\ensuremath{\widehat{#1}}}}
\ddefloop ABCDEFGHIJKLMNOPQRSTUVWXYZ\ddefloop
\def\ddef#1{\expandafter\def\csname hc#1\endcsname{\ensuremath{\widehat{\mathcal{#1}}}}}
\ddefloop ABCDEFGHIJKLMNOPQRSTUVWXYZ\ddefloop
\def\ddef#1{\expandafter\def\csname t#1\endcsname{\ensuremath{\widetilde{#1}}}}
\ddefloop ABCDEFGHIJKLMNOPQRSTUVWXYZ\ddefloop
\def\ddef#1{\expandafter\def\csname tc#1\endcsname{\ensuremath{\widetilde{\mathcal{#1}}}}}
\ddefloop ABCDEFGHIJKLMNOPQRSTUVWXYZ\ddefloop
\def\ddefloop#1{\ifx\ddefloop#1\else\ddef{#1}\expandafter\ddefloop\fi}
\def\ddef#1{\expandafter\def\csname scr#1\endcsname{\ensuremath{\mathscr{#1}}}}
\ddefloop ABCDEFGHIJKLMNOPQRSTUVWXYZ\ddefloop

\let\oldparagraph\paragraph
\renewcommand{\paragraph}[1]{\oldparagraph{#1.}}

\newcommand{\pfref}[1]{Proof of \cref{#1}}

\renewcommand{\epsilon}{\varepsilon}

\newcommand{\ind}{\mathbbm{1}}    %

\newcommand{\eps}{\epsilon}
\newcommand{\veps}{\varepsilon}

\newcommand{\ldef}{\vcentcolon=}
\newcommand{\rdef}{=\vcentcolon}

\renewcommand{\bigm}[1]{%
  \ifcsname fenced@\string#1\endcsname
    \expandafter\@firstoftwo
  \else
    \expandafter\@secondoftwo
  \fi
  {\expandafter\amsmath@bigm\csname fenced@\string#1\endcsname}%
  {\amsmath@bigm#1}%
}

\newcommand{\DeclareFence}[2]{\@namedef{fenced@\string#1}{#2}}
\makeatother

\makeatletter
\let\save@mathaccent\mathaccent
\newcommand*\if@single[3]{%
  \setbox0\hbox{${\mathaccent"0362{#1}}^H$}%
  \setbox2\hbox{${\mathaccent"0362{\kern0pt#1}}^H$}%
  \ifdim\ht0=\ht2 #3\else #2\fi
  }
\newcommand*\rel@kern[1]{\kern#1\dimexpr\macc@kerna}
\newcommand*\widebar[1]{\@ifnextchar^{{\wide@bar{#1}{0}}}{\wide@bar{#1}{1}}}
\newcommand*\wide@bar[2]{\if@single{#1}{\wide@bar@{#1}{#2}{1}}{\wide@bar@{#1}{#2}{2}}}
\newcommand*\wide@bar@[3]{%
  \begingroup
  \def\mathaccent##1##2{%
    \let\mathaccent\save@mathaccent
    \if#32 \let\macc@nucleus\first@char \fi
    \setbox\z@\hbox{$\macc@style{\macc@nucleus}_{}$}%
    \setbox\tw@\hbox{$\macc@style{\macc@nucleus}{}_{}$}%
    \dimen@\wd\tw@
    \advance\dimen@-\wd\z@
    \divide\dimen@ 3
    \@tempdima\wd\tw@
    \advance\@tempdima-\scriptspace
    \divide\@tempdima 10
    \advance\dimen@-\@tempdima
    \ifdim\dimen@>\z@ \dimen@0pt\fi
    \rel@kern{0.6}\kern-\dimen@
    \if#31
      \overline{\rel@kern{-0.6}\kern\dimen@\macc@nucleus\rel@kern{0.4}\kern\dimen@}%
      \advance\dimen@0.4\dimexpr\macc@kerna
      \let\final@kern#2%
      \ifdim\dimen@<\z@ \let\final@kern1\fi
      \if\final@kern1 \kern-\dimen@\fi
    \else
      \overline{\rel@kern{-0.6}\kern\dimen@#1}%
    \fi
  }%
  \macc@depth\@ne
  \let\math@bgroup\@empty \let\math@egroup\macc@set@skewchar
  \mathsurround\z@ \frozen@everymath{\mathgroup\macc@group\relax}%
  \macc@set@skewchar\relax
  \let\mathaccentV\macc@nested@a
  \if#31
    \macc@nested@a\relax111{#1}%
  \else
    \def\gobble@till@marker##1\endmarker{}%
    \futurelet\first@char\gobble@till@marker#1\endmarker
    \ifcat\noexpand\first@char A\else
      \def\first@char{}%
    \fi
    \macc@nested@a\relax111{\first@char}%
  \fi
  \endgroup
}
\makeatother

\newcommand{\paranewline}{\, \\}

\newcommand{\DIS}{\mathsf{{DIS}}}

\newcommand{\AlgLcb}{\mathrm{\mathbf{Alg}}_{\mathsf{lcb}}}
\newcommand{\AlgUcb}{\mathrm{\mathbf{Alg}}_{\mathsf{ucb}}}
\newcommand{\ucb}{\mathsf{ucb}}
\newcommand{\lcb}{\mathsf{lcb}}
\newcommand{\dnn}{\mathsf{dnn}}
\newcommand{\pseud}{\mathrm{Pdim}}
\newcommand{\vcd}{\mathrm{VCdim}}

\newcommand{\reg}{\Reg}

\newcommand{\exc}{\mathsf{excess}}
\newcommand{\BV}{\mathsf{BV}}
\newcommand{\TV}{\mathsf{TV}}
\newcommand{\wderi}{\mathsf{D}}

\newcommand{\textCAL}{\textsf{CAL}\xspace}

\newcommand{\textRCAL}{\textsf{RobustCAL}\xspace}

\newcommand{\textNCAL}{\textsf{NeuralCAL}\xspace}

\newcommand{\textNCALP}{\textsf{NeuralCAL}\texttt{++}\xspace}

\newcommand{\relu}{\mathsf{ReLU}}
\newcommand{\RBV}{\mathscr{R}\, \BV}
\newcommand{\RTV}{\mathscr{R}\, \TV}
\newcommand{\pdim}{\pseud}

\newcommand{\Reg}{\mathrm{\mathbf{Reg}}}
\newcommand{\RegCB}{\mathrm{\mathbf{Reg}}_{\mathsf{CB}}}
\newcommand{\RegSq}{\mathrm{\mathbf{Reg}}_{\mathsf{Sq}}}

\newcommand{\AlgSq}{\mathrm{\mathbf{Alg}}_{\mathsf{Sq}}}
\newcommand{\AlgOpt}{\mathrm{\mathbf{Alg}}_{\mathsf{Opt}}}
\newcommand{\dec}{\mathsf{dec}}
\newcommand{\SqAlg}{\AlgSq}
\newcommand{\AlgSample}{\mathrm{\mathbf{Alg}}_{\mathsf{Sample}}}

\newcommand{\regcbh}{\mathrm{\mathbf{Reg}}_{\mathsf{CB},h}}
\newcommand{\regcbhb}{\mathrm{\mathbf{Reg}}_{\mathsf{CB},h_b}}
\newcommand{\regimphb}{\mathrm{\mathbf{Reg}}_{\mathsf{Imp},h_b}}
\newcommand{\regimph}{\mathrm{\mathbf{Reg}}_{\mathsf{Imp},h}}
\newcommand{\regcb}{\RegCB}
\newcommand{\regsq}{\RegSq}

\newcommand{\sqalgtext}{$\SqAlg$\xspace}
\newcommand{\samplealg}{\AlgSample}
\newcommand{\samplealgtext}{$\samplealg$\xspace}
\newcommand{\smthigw}{\textsf{SmoothIGW}\xspace}

\newcommand{\optalg}{\AlgOpt}
\newcommand{\optalgtext}{$\optalg$\xspace}

\newcommand{\Topt}{\cT_{\mathsf{Opt}}}
\newcommand{\Tsq}{\cT_{\mathsf{Sq}}}

\newcommand{\Msq}{\cM_{\mathsf{Sq}}}
\newcommand{\Mopt}{\cM_{\mathsf{Opt}}}
\newcommand{\Tsample}{\cT_{\mathsf{Sample}}}
\newcommand{\Msample}{\cM_{\mathsf{Sample}}}

\newcommand{\dectext}{Decision-Estimation Coefficient\xspace}

\newcommand{\des}{q^{\opt}}
\newcommand{\dess}{{\check q}^{\opt}}

\newcommand{\rwalgtext}{\textsf{ReweightedSpanner}\xspace}

\newcommand{\squarecb}{\textsf{SquareCB}\xspace}
\newcommand{\greedy}{$\epsilon$-\textsf{Greedy}\xspace}
\newcommand{\spannerGreedy}{\textsf{SpannerGreedy}\xspace}
\newcommand{\spannerIGW}{\textsf{SpannerIGW}\xspace}
\newcommand{\argmaxIGW}{\textsf{IGW-ArgMax}\xspace}
\newcommand{\ball}{\mathsf{Ball}\xspace}
\newcommand{\sgm}{\Omega}
\newcommand{\smthh}{\mathsf{Smooth}_h\xspace}
\newcommand{\smthhb}{\mathsf{Smooth}_{h_b}\xspace}
\newcommand{\smth}{\text{smooth}\xspace}

\newcommand{\cats}{\textsf{CATS}\xspace}
\newcommand{\corral}{\textsf{CORRAL}\xspace}

\newcommand{\oneshotwiki}{\textsf{oneshotwiki}\xspace}
\newcommand{\amazon}{\textsf{amazon}\xspace}
\newcommand{\sentencetransformers}{\textsf{sentence transformers}\xspace}
\newcommand{\supervised}{\textsf{supervised}\xspace}

\newcommand{\mainalg}{\spannerIGW}
\newcommand{\greedyalg}{\spannerGreedy}

\newcommand{\sr}{{\textsf SR}}
\newcommand{\moss}{{\textsf MOSS}\xspace}
\newcommand{\mossPlus}{{\textsf MOSS++}\xspace}
\newcommand{\mossPlusEmp}{{\textsf empMOSS++}\xspace}
\newcommand{\algParallel}{{\textsf Parallel}\xspace}
\newcommand{\mossOracle}{{\textsf MOSS Oracle}\xspace}
\newcommand{\quantile}{{\textsf Quantile}\xspace}

\newcommand{\ucbalg}{\textsf{UCB}\xspace}

\newcommand{\smoothCorral}{\textsf{Smooth Corral}\xspace}

\newcommand{\linucb}{\textsf{LinUCB}\xspace}
\newcommand{\suplinucb}{\textsf{SupLinUCB}\xspace}
\newcommand{\linucbOracle}{\textsf{LinUCB Oracle}\xspace}
\newcommand{\linucbPlus}{\textsf{LinUCB}{\texttt++}\xspace}

\newcommand{\linucbPlusCorral}{\textsf{LinUCB}{\texttt++} \textsf{with Corral}\xspace}
\newcommand{\dynamicBalancing}{\textsf{Dynamic Balancing}\xspace}

\newcommand{\rage}{\textsf{RAGE}\xspace}
\newcommand{\round}{\textsf{ROUND}\xspace}
\newcommand{\OPT}{\textsf{OPT}\xspace}
\newcommand{\gemsc}{\textsf{GEMS-c}\xspace}
\newcommand{\gemsm}{\textsf{GEMS-m}\xspace}
\newcommand{\gemsb}{\textsf{GEMS-b}\xspace}

\usepackage{pgfplots}
\usepackage{tikz-cd}

\allowdisplaybreaks

\clearpage

\title{Interactive Machine Learning: From Theory to Scale}
\author{Yinglun Zhu}
\department{Computer Sciences}
\oralexamdate{June 06, 2023}
\committeeone{Robert D. Nowak, Professor, Electrical \& Computer Engineering, UW--Madison}
\committeetwo{Kevin Jamieson, Assistant Professor, \mbox{Computer Science \& Engineering, University of Washington}}
\committeethree{Kangwook Lee, Assistant Professor, Electrical \& Computer Engineering, UW--Madison}
\committeefour{Rebecca Willett, Professor, Statistics and Computer Science, University of Chicago}
\committeefive{Stephen J. Wright, Professor, Computer Sciences, UW--Madison}
\committeesix{Xiaojin (Jerry) Zhu, Professor, Computer Sciences, UW--Madison}

\date{2023}

\begin{document}

\ifpdf
\DeclareGraphicsExtensions{.pdf, .jpg, .tif}
\else
\DeclareGraphicsExtensions{.eps, .jpg}
\fi

\newgeometry{left=.8in,right=.8in,bottom=1in,top=1in}
\maketitle
\restoregeometry %

\svnidlong{$LastChangedBy$}{$LastChangedRevision$}{$LastChangedDate$}{$HeadURL: http://freevariable.com/dissertation/branches/diss-template/frontmatter/frontmatter.tex $}
\vcinfo{}

\copyrightpage

\begin{dedication}
\hspace{0pt}
\vfill
	\emph{To mom and dad.}
\vfill
\hspace{0pt}
\end{dedication}

\begin{quote}
\hspace{0pt}
\vfill
\begin{wbepi}{George E. P. Box, University of Wisconsin--Madison}
	Essentially, all models are wrong, but some are useful.
\end{wbepi}
\vfill
\hspace{0pt}
\end{quote}

\chapterstyle{deposit}
\pagestyle{deposit}

\begin{acks}

First and foremost, I would like to express my deepest gratitude to my Ph.D. advisor Robert D. Nowak, for his continuous mentoring, support, and encouragement.
Rob gives me great freedom to explore topics I am interested in, but at the same time, he is always passionate about discussing research problems with me and helping me get out of trouble.
Rob has been everything I could ask for as an advisor. 
Rob is also a role model to me, as a researcher and mentor, who will keep motivating me in my future academic career.

I had a wonderful summer intern at Microsoft Research NYC in 2021, where I was fortunate to be mentored by Dylan J. Foster, John Langford, and Paul Mineiro. 
I am extremely grateful to their mentoring; they not only taught me how to approach difficult research problems, but also spent their time helping me sharpen my presentation and communication skills.

I am thankful to my committee members: Kevin Jamieson, Kangwook Lee,  Rebecca Willett, Stephen J. Wright, and Xiaojin (Jerry) Zhu. They have been great sources of guidance for me during my Ph.D. journey, and have consistently provided me with invaluable advice and insightful feedback.

I was fortunate to have collaborated with many outstanding researchers over the past six years: Gregory Canal, Yifang Chen, Simon S. Du, Dylan J. Foster, Quanquan Gu, Kevin Jamieson, Ruoxi Jiang, Sumeet Katariya, Julian Katz-Samuels, John Langford, Paul Mineiro, Stephen Mussmann, Robert D. Nowak, Mark Rucker, Rebecca Willett, Jifan Zhang, and Dongruo Zhou.
I am grateful to all of them for their patience, encouragement, and friendship. The contents of this dissertation have benefited especially from collaborations with Dylan J. Foster, Julian Katz-Samuels, John Langford, Paul Mineiro, and Robert D. Nowak.

Studying at UW--Madison has been a great experience. I would like to thank my labmates for their support and encouragement: Gregory Canal, Danica Fliss, Mina Karzand, Sumeet Katariya, Julian Katz-Samuels, Jeongyeol Kwon, Blake Mason, Haley Massa, Subhojyoti Mukherjee, Julia Nakhleh, Rahul Parhi, Joseph Shenouda, Scott Sievert, Gokcan Tatli, Ardhendu Tripathy, Liu Yang, and Jifan Zhang. 
I would also like to extend my sincere thanks to all my friends who have supported me over the past six years. 
This acknowledgement is way too short to list all the names, but you know who you are.

Finally, I would like to thank my parents, Meirong and Xiangzhong. 
None of this would have been possible without their unconditional love and support.

\end{acks}

\tableofcontents

\clearpage
\listoftables

\clearpage
\listoffigures

\clearpage

\begin{abstract}

Machine learning has achieved remarkable success across a wide range of applications, yet many of its most effective methods rely on access to large amounts of labeled data or extensive online interaction. In practice, both acquiring high-quality labels and making decisions through trial-and-error can be expensive, time-consuming, or risky, particularly in large-scale or high-stakes settings. These challenges motivate the study of interactive machine learning, in which the learner actively influences how information is collected or which actions are taken, using past observations to guide future interactions.

This dissertation develops new algorithmic principles and establishes fundamental limits for interactive machine learning. Rather than passively training on a fixed dataset, the learner adaptively selects what information to request next---such as which data points to label or which actions to take---and updates its model based on the resulting feedback. By closing this interaction loop, interactive learning aims to achieve substantially greater efficiency, for example by learning accurate predictors using far fewer labels. This dissertation focuses on three core challenges that arise when scaling interactive learning to real-world settings: active learning with noisy data and rich model classes, sequential decision making with large action spaces, and model selection under partial feedback.

In the first part, we study active learning with noisy data and rich model classes. While active learning can offer dramatic reductions in labeling cost, most existing theoretical guarantees rely on restrictive low-noise assumptions and simple model classes. We overcome these limitations by introducing an abstention mechanism that allows the learner to defer uncertain predictions at a controlled cost. Leveraging supervised convex loss regression oracles, we develop the first computationally efficient active learning algorithm that achieves exponential label savings without any low-noise assumptions. We further extend this framework to neural networks, providing the first deep active learning algorithms with nearly minimax-optimal label complexity, and exponential label savings when combined with abstention.

In the second part, we study sequential decision making with large action spaces, focusing on contextual bandit problems. Classical exploration strategies scale poorly with the number of actions, rendering them impractical when the action space is large or continuous. We develop the first efficient, general-purpose algorithms whose statistical guarantees and computational complexity are independent of the size of the action space. Our results apply to both structured settings, where actions admit linear structure, and unstructured settings, where we introduce smoothed benchmarks to circumvent inherent intractability. The proposed algorithms achieve near-optimal regret guarantees and demonstrate strong empirical performance on real-world datasets with millions of actions.

In the third part, we investigate model selection in sequential decision making, where the learner must adapt to unknown problem complexity under partial feedback. We establish the first fundamental lower bounds showing that model selection in regret minimization is strictly harder than in supervised learning, requiring a polynomial rather than logarithmic overhead. Despite this hardness, we develop Pareto optimal algorithms for regret minimization that match these limits up to logarithmic factors. We also study model selection in best action identification, showing that near instance-optimal adaptation can be achieved with only modest additional cost.

Overall, this dissertation advances the theoretical foundations of interactive machine learning by developing algorithms that are statistically optimal and computationally efficient, while also providing principled guidance for deploying interactive learning methods in large-scale, real-world settings.

\end{abstract}

\clearpage\pagenumbering{arabic}

\chapter{Overview }
\label{chapter:intro}
\section{Introduction}

Over the past decade, machine learning has achieved remarkable successes across a wide
range of domains, including image recognition \citep{krizhevsky2012imagenet,lecun2015deep},
natural language processing \citep{bahdanau2014neural,brown2020language}, and game playing
\citep{silver2016mastering,berner2019dota}.
At a high level, the learning paradigm is conceptually simple:
given a dataset of labeled examples, the learner fits a model that generalizes to new
inputs.
In many of the most celebrated successes, the key enabler has been scale---in particular,
the availability of massive labeled datasets together with models and optimization
methods capable of exploiting them.

A concrete example is ImageNet.
In 2009, Dr.\ Fei-Fei Li and collaborators curated a large-scale image classification
dataset with roughly $15{,}000{,}000$ labeled images spanning $22{,}000$ categories
\citep{russakovsky2015imagenet}.
With access to this dataset, and following several years of progress in model architectures
and optimization techniques, learned classifiers ultimately achieved superhuman image
classification performance \citep{russakovsky2015imagenet,krizhevsky2012imagenet,he2016deep}.
However, this success also highlights a fundamental bottleneck: obtaining labels at
scale is expensive.
The ImageNet labeling effort relied on $48{,}940$ annotators from Amazon Mechanical Turk
across $167$ countries and took more than two years.
Such costs are difficult to sustain in many real-world applications, especially in
high-stakes domains such as medicine and robotics, where labels may require expert
time, specialized equipment, or physical experimentation.
As a result, the ability to \emph{efficiently acquire information}---rather than merely
fit a model to a fixed dataset---has become a central challenge for modern machine
learning deployments.

This dissertation studies \emph{interactive machine learning}, where the learner is
not a passive recipient of data but instead uses past observations to guide future
data acquisition and decision making.
Rather than collecting a dataset upfront, the learner adaptively chooses what
information to request next (e.g., which examples to label or which actions to take),
and then updates its model based on the feedback it receives.
By iteratively closing this loop, interactive learning aims to achieve substantially
greater efficiency---for example, learning accurate predictors using far fewer labels,
or making near-optimal decisions with far fewer interactions.

We focus on two complementary paradigms of interaction.
On the \emph{prediction} side, interactive learning specializes to \emph{active learning},
where the learner adaptively selects which unlabeled points to query so as to learn an
accurate classifier or regressor with minimal labeling cost.
On the \emph{decision making} side, interaction gives rise to \emph{sequential decision
making}, where the learner repeatedly selects actions and observes feedback, with goals
such as minimizing regret or efficiently identifying the best action.
Although these settings have been studied extensively, many classical guarantees are
derived under idealized assumptions that break down in the regimes most relevant to
practice.

Accordingly, this dissertation develops new algorithmic principles and establishes
fundamental limits for interactive learning in three broad directions:
\begin{itemize}
	\item \textbf{Active learning with noisy data and rich model classes.}
	Most favorable guarantees for active learning were developed under low-noise
	assumptions and for simple model classes.
  This dissertation develops general algorithmic principles that remain effective
		with noisy data and rich model classes, including models motivated by modern deep
		learning practice.

	\item \textbf{Sequential decision making with large action spaces.}
	Many sequential decision making methods rely on exploration strategies whose cost
	scales with the number of actions.
	This becomes infeasible when the action space is large or continuous.
  This dissertation develops algorithms whose statistical guarantees and
		computational complexity do not deteriorate with the size of the action space.

	\item \textbf{Model selection in sequential decision making.}
	Model selection is fundamental in supervised learning, yet it is substantially less
	understood in sequential decision making where feedback is partial and data are collected
	adaptively.
  This dissertation characterizes the fundamental limits of model selection in
		sequential decision making and designs procedures that automatically adapt to the
		underlying problem structure and complexity.
\end{itemize}

Throughout, our goal is to make interactive learning \emph{statistically
optimal and computationally efficient}.
On the statistical side, we aim to establish guarantees that match fundamental lower
bounds whenever possible.
On the computational side, we emphasize algorithms that can be implemented via
efficient primitives (e.g., standard supervised learning and optimization oracles),
so that the resulting methods can plausibly be deployed in large-scale systems.

The remainder of this chapter introduces the basic learning paradigms that appear
throughout the dissertation.
We summarize the organization of the dissertation in \cref{intro:sec:organization},
provide bibliographic details in \cref{intro:sec:bibliographic}, and introduce the 
general notation used throughout the dissertation in \cref{intro:sec:notation}.

\section{Passive and Active Learning}

Machine learning focuses on using data and algorithms to imitate the way humans learn.
In prediction tasks, the learner aims to learn a \emph{classifier} $h:\cX\rightarrow\cY$, where $\cX$ denotes the \emph{instance space} and $\cY$ denotes the \emph{label space}.
We primarily consider classical binary classification tasks, where the label space is $\cY \ldef \crl{0,1}$.
The joint distribution over $\cX\times\cY$ is denoted by $\cD_{\cX\cY}$.
We use $\cD_{\cX}$ to denote the marginal distribution over the input space $\cX$, and $\cD_{\cY\mid x}$ to denote the conditional distribution of $\cY$ given any $x\in\cX$.

For any classifier $h:\cX\rightarrow\cY$, its classification error is defined as
$\err(h) \ldef \P_{(x,y)\sim\cD_{\cX\cY}}(h(x)\neq y)$.
Given a \emph{hypothesis class} $\cH:\cX\rightarrow\cY$, we use $h^{\star}\in\cH$ to denote the classifier that achieves the smallest error within $\cH$, i.e., $h^\star \ldef \argmin_{h\in\cH}\err(h)$.
For any classifier $h:\cX\rightarrow\cY$, we define its \emph{excess error} as $\exc(h) \ldef \err(h)-\err(h^{\star})$.

The learner's goal is to learn a classifier with small excess error.
Learning is commonly studied in the \emph{Probably Approximately Correct} (PAC) framework \citep{vapnik1971uniform,vapnik1995nature,valiant1984theory,haussler1992decision}: given parameters $\eps>0$ and $\delta\in(0,1)$, the learner aims to, with probability at least $1-\delta$, identify a classifier $\wh h$ such that
\begin{align}
	\label{intro:eq:learning}
	\err(\wh h) \le \err(h^{\star}) + \eps.
\end{align}

\paragraph{Passive learning}
We use the term \emph{passive learning} to refer to the classical \emph{supervised} (or statistical) learning setting, in order to distinguish it from the \emph{active learning} setting discussed below.
In passive learning, the learner collects a dataset $\crl{\prn{x_i,y_i}}_{i=1}^{n}$ consisting of i.i.d.\ samples drawn from the joint distribution $\cD_{\cX\cY}$, and then learns a classifier $\wh h:\cX\rightarrow\cY$.
The number of labeled examples $n$ required to satisfy \cref{intro:eq:learning} is referred to as the \emph{sample complexity}.

Passive learning has been extensively studied and is now well understood.
A hypothesis class $\cH$ is PAC learnable if and only if it has finite VC dimension $\vcd(\cH)$, a complexity measure that characterizes the richness of the hypothesis class \citep{vapnik1971uniform,shalev2014understanding}.
When $\vcd(\cH)<\infty$, the sample complexity of passive learning scales as
\linebreak
$\wt\Theta\prn{\vcd(\cH)\cdot\poly(\frac{1}{\eps})}$, that is, polynomially in $\frac{1}{\eps}$.

\paragraph{Active learning}
In contrast to passive learning, \emph{active learning} allows the learner to interactively collect labeled data.
Specifically, the learner has access to a labeling oracle: given any \emph{unlabeled} data point $x$ as input, the oracle returns a label $y\sim\cD_{\cY\mid x}$.
Rather than labeling all available data points, the active learner adaptively selects which data points to query based on previously collected information.

The hope of active learning is that, compared to passive learning, the learner can identify a classifier satisfying \cref{intro:eq:learning} using significantly fewer labeled examples.
We define the \emph{label complexity} as the number of calls made to the labeling oracle, and we evaluate the performance of active learning algorithms primarily in terms of this quantity.

A canonical example illustrating the advantage of active learning is learning a one-dimensional threshold function in the noiseless setting.
In this case, passive learning requires $\Omega(\frac{1}{\eps})$ labeled examples, whereas active learning---instantiated via binary search---can identify a classifier with error at most $\eps$ using only $O\prn{\log\frac{1}{\eps}}$ labels, yielding an \emph{exponential speedup} over passive learning.
Beyond threshold functions, a substantial body of work has established positive active learning results for other hypothesis classes
\citep{balcan2007margin,hanneke2007bound,dasgupta2009analysis,hsu2010algorithms,dekel2012selective,hanneke2014theory,zhang2014beyond,krishnamurthy2019active,katz2021improved}.
However, exponential gains over passive learning are typically observed only for relatively simple hypothesis classes (e.g., linear classifiers) and under favorable noise conditions, such as Massart noise \citep{massart2006risk}.

\section{Sequential Decision Making}

Beyond prediction tasks, another central problem in machine learning is
\emph{sequential decision making}, where the learner makes decisions online and
sequentially observes feedback.
In this setting, the learner is given an action set $\cA$, and the decision making
process unfolds over a sequence of rounds.
At each round $t$, the learner observes a context $x_t$, selects an action
$a_t\in\cA$, and then observes a reward $r_t$.

In this dissertation, we primarily focus on the \emph{bandit} setting, where the
learner receives only \emph{partial feedback}: the observed reward
$r_t=r_t(a_t)$ corresponds to the action $a_t$ taken at round $t$, and no
information is revealed about the rewards of unchosen actions.
This limited feedback structure fundamentally distinguishes bandit problems from
supervised learning and gives rise to the exploration-exploitation trade-off
\citep{bubeck2012regret}.
An important variant is the \emph{contextual bandit} problem, where the context
$x_t$ captures side information about the current decision and the learner selects
the action $a_t$ based on the observed context.
When the context $x_t$ remains fixed across rounds, the contextual bandit problem
reduces to the \emph{non-contextual} (or classical) bandit setting.

We study two distinct objectives in bandit learning:
\emph{regret minimization} and \emph{best action identification}.
We describe these two settings separately below.

\paragraph{Regret minimization}
In regret minimization, the decision making process proceeds for a fixed horizon of
$T$ rounds.
The learner is given a policy class $\Pi$, where each policy $\pi\in\Pi$ is a
mapping from the context space $\cX$ to the action space $\cA$, i.e.,
$\pi:\cX\rightarrow\cA$.
Let
$\pi^{\star} \ldef \argmax_{\pi\in\Pi}\sum_{t=1}^{T} r_t\prn{\pi\prn{x_t}}$
denote the optimal policy in hindsight.
The goal of the learner is to minimize the cumulative regret (or its expectation),
defined as
\begin{align}
	\label{intro:eq:regret}
	\reg(T) \ldef \sum_{t=1}^{T} r_t(\pi^{\star}(x_t)) - r_t(a_t).
\end{align}

The regret in \cref{intro:eq:regret} measures the performance gap between the
learner and the optimal policy $\pi^{\star}$.
In other words, it quantifies how much reward is lost due to not acting optimally
\citep{bubeck2012regret}.
The regret minimization framework has been widely deployed in practice, particularly
in online personalization, recommendation systems, and advertising
\citep{li2010contextual,agarwal2016making,tewari2017ads,cai2021bandit}.
If the regret grows sublinearly in $T$, i.e., $\reg(T)=o(T)$, then the learner's
average performance converges to that of the optimal policy.
A widely accepted benchmark is to achieve regret scaling as
$\reg(T)=\wt\Theta(\sqrt{T})$, which is known to be information-theoretically optimal
in a broad range of settings
\citep{agarwal2012contextual,agarwal2014taming,foster2020beyond,simchi2021bypassing}.

\paragraph{Best action identification}
In best action identification (also known as pure exploration), the learner's goal is to efficiently identify an
action that (approximately) achieves the highest reward.
Focusing on the non-contextual setting with stochastic rewards, the optimal action
$a^{\star}\in\cA$ is defined as
\begin{align}
	\label{intro:eq:best_action}
	a^{\star} \ldef \argmax_{a\in\cA} \E_r\brk{r(a)}.
\end{align}
Best action identification has been widely studied in applications such as online
crowdsourcing and biomedical experimentation
\citep{zhou2014optimal,tanczos2017kl,reda2020machine,aziz2021multi}.

Two settings are studied in best action identification: the \emph{fixed confidence}
setting and the \emph{fixed budget} setting.
In the fixed confidence setting, given a confidence parameter $\delta\in(0,1)$, the
learner aims to identify the best action $a^{\star}$ (or a near-optimal action) with
probability at least $1-\delta$, while minimizing the number of samples
\citep{mannor2004sample,even2006action}.
In the fixed budget setting, given a sampling budget $T$, the learner outputs an
action $\wh a$ and seeks to minimize the probability of error,
$\P\prn{\wh a\neq a^{\star}}$ \citep{hoffman2014correlation,katz2020empirical}.
In both settings, the learner aims to adapt to the fundamental
instance-dependent complexity of the problem, rather than incurring guarantees based
solely on worst-case complexity.

\section{Highlights and Organization}
\label{intro:sec:organization}

\cref{chapter:intro} introduced the interactive learning settings studied in this
dissertation and outlined the central challenges and contributions.
The remainder of the dissertation is organized into three parts, each focusing on a
distinct aspect of interactive machine learning.

\paragraph{\cref{part:active}: Active Learning with Noisy Data and Rich Model Classes}
Active learning has become increasingly important in modern applications, where
unlabeled data are abundant but the labeling process is expensive and time-consuming.
Despite this practical relevance, most existing theoretical guarantees for active
learning were developed under restrictive assumptions, namely (i) noiseless or
low-noise settings, and (ii) simple hypothesis classes such as threshold functions
and linear classifiers.
In \cref{part:active}, we develop efficient algorithms that overcome these two
fundamental limitations, making a significant step toward deploying active learning
in realistic settings involving noisy data and rich model classes.

A central reason for focusing on low-noise assumptions (e.g., Massart or Tsybakov
noise) is a classical lower bound showing that, in high-noise regimes, active learning
offers no improvement over passive learning.
To move beyond this barrier, in \cref{chapter:active:efficient} we study active
learning with an additional \emph{abstention} option: when the classifier abstains,
it incurs a cost marginally smaller than random guessing, formalized through
Chow's error.
With access to a supervised convex loss regression oracle (e.g., least squares for
linear models), we develop the first computationally efficient active learning
algorithm that achieves exponential label savings without imposing any low-noise
assumptions.
These results are not only theoretically appealing but also practically motivated;
for example, in medical applications, it is often preferable to defer high-risk
decisions to human experts when the classifier is uncertain.
We further extend this framework to recover minimax-optimal guarantees in the
standard setting and to achieve \emph{constant} label complexity for finite
hypothesis classes.

To move closer to real-world deployments, in \cref{chapter:active:deep} we study
active learning with neural networks, also known as deep active learning.
While deep active learning has been extensively explored empirically, its theoretical
foundations have remained largely unresolved.
By carefully balancing approximation error and learning error, we develop the first
deep active learning algorithm that achieves nearly minimax-optimal label complexity
guarantees.
When combined with the abstention option, our approach further yields exponential
savings in label complexity.
These results
provide theoretical justification for many empirically successful deep active
learning methods.
Our results are obtained by establishing a general connection between approximation
theory and active learning guarantees, which is of independent interest.

\paragraph{\cref{part:large}: Sequential Decision Making with Large Action Spaces}
While sequential decision making has been extensively studied in settings with a
small number of actions, theoretical guarantees for large or continuous action spaces
have remained limited, creating a substantial gap between theory and practice.
In \cref{part:large}, we address this challenge by developing efficient algorithms
for large-scale sequential decision making in both \emph{structured} and
\emph{unstructured} settings.

In \cref{chapter:large:linear}, we focus on the structured case and develop the first
\emph{efficient, general-purpose} algorithm for contextual bandits with continuous,
linearly structured action spaces.
Our algorithm leverages standard computational oracles for (i) supervised learning
and (ii) \emph{linear} optimization over the action space, achieving nearly optimal
regret guarantees with runtime and memory requirements independent of the size of
the action space.
Beyond its theoretical guarantees, the algorithm is highly practical: it attains
state-of-the-art performance on an Amazon dataset with nearly three million
categories.

Unstructured decision making problems are generally intractable, as unstructured
function classes allow adversarial instances in which the learner must effectively
``identify a needle in a haystack.''
To address such pathological cases, in \cref{chapter:large:smooth} we study
unstructured decision making under \emph{smoothed} benchmarks, where performance is
measured against a smoothed distribution rather than a delta distribution concentrating on
a single optimal action.
Focusing on contextual bandits, we develop the first \emph{efficient, general-purpose}
algorithm that applies to \emph{any} unstructured regression function class (as long as they are measurable).
When additional structural assumptions exist  (e.g., Lipschitz or H\"older continuity), our algorithm further recovers the optimal guarantees when competing against the standard, non-smoothed benchmark.
\looseness=-1

\paragraph{\cref{part:model_selection}: Model Selection in Sequential Decision Making}
Model selection is a fundamental statistical problem, playing a central role in
virtually every machine learning pipeline.
However, model selection in sequential decision making poses unique challenges,
since decisions are made online and feedback is inherently partial.
In \cref{part:model_selection}, we characterize the fundamental limits of model
selection in sequential decision making and develop efficient algorithms that achieve
near-optimal performance.

We first consider model selection in regret minimization.
In \cref{chapter:model:multiple}, we study the unstructured case, where multiple
actions may be optimal and the goal is to scale regret with the effective number of
actions rather than the total number of actions.
In \cref{chapter:model:linear}, we study the structured case, where there is a nested
sequence of linear hypothesis classes and the learner seeks to adapt to the smallest
class containing the true model.
In both settings, we establish the first lower bounds showing that model selection in sequential
decision making is strictly harder than in supervised learning: whereas supervised
learning incurs only an additional logarithmic cost, sequential decision making
requires paying an additional \emph{polynomial} cost.
Despite this hardness, we develop \emph{Pareto optimal} algorithms whose guarantees
match the lower bounds up to logarithmic factors.
A different Pareto optimal model selection algorithm is also provided and analyzed in
\cref{chapter:large:smooth}.

Finally, in \cref{chapter:model:bai}, we study model selection in best action
identification setting, considering both fixed confidence and fixed budget
settings.
Given a nested sequence of hypothesis classes with increasing complexity, our goal is
to adapt to the instance-dependent complexity of the smallest class containing the
true model, rather than incurring the cost associated with the largest class.
We develop algorithms based on a novel experimental design that leverages 
the geometry of the action set to efficiently identify a near-optimal hypothesis
class.
In contrast to regret minimization, we show that model selection in best action
identification can be achieved with only modest additional cost.

\section{Bibliographic Notes}
\label{intro:sec:bibliographic}

Results in \cref{part:active} are based on joint work with Robert D. Nowak:

\begin{itemize}
  \item Yinglun Zhu and Robert D. Nowak. 2022. 
	  Efficient active learning with abstention.
    \emph{Advances in Neural Information Processing Systems.}
    \nocite{zhu2022efficient}

  \item Yinglun Zhu and Robert D. Nowak. 2022. 
Active learning with neural networks: Insights from nonparametric statistics.
    \emph{Advances in Neural Information Processing Systems.}
    \nocite{zhu2022active}
\end{itemize}

Results in \cref{part:large} are based on joint work with Dylan J. Foster, John Langford, and Paul Mineiro:

\begin{itemize}
  \item Yinglun Zhu and Dylan J. Foster, John Langford, and Paul Mineiro. 2022. 
	  Contextual bandits with large action spaces: Made practical.
    \emph{International Conference on Machine Learning.}
    \nocite{zhu2022contextual}

  \item Yinglun Zhu and Paul Mineiro. 2022. 
	  Contextual bandits with smooth regret: Efficient learning in continuous action spaces.
    \emph{International Conference on Machine Learning.}
    \nocite{zhu2022smooth}
\end{itemize}

Results in \cref{part:model_selection} are based on joint work with Julian Katz-Samuels and Robert D. Nowak:

\begin{itemize}
  \item Yinglun Zhu and Robert D. Nowak. 2020. 
	  On regret with multiple best arms.
    \emph{Advances in Neural Information Processing Systems.}
    \nocite{zhu2020regret}

  \item Yinglun Zhu and Robert D. Nowak. 2022. 
	  Pareto optimal model selection in linear bandits.
    \emph{International Conference on Artificial Intelligence and Statistics.}
    \nocite{zhu2022pareto}

  \item Yinglun Zhu, Julian Katz-Samuels, and Robert D. Nowak. 2022. 
	  Near instance optimal model selection for pure exploration linear bandits.
    \emph{International Conference on Artificial Intelligence and Statistics.}
    \nocite{zhu2022near}

\end{itemize}

Additional work completed during my Ph.D. that is not included in this dissertation includes \citet{zhu2020robust, zhu2021pure, rucker2023infinite, zhang2024labelbench}. This dissertation is an updated version of \citet{zhu2023interactive}, in which we correct typographical errors and make minor technical and structural revisions.

\section{Notation}
\label{intro:sec:notation}

We define general notation that will be used throughout this dissertation.
Additional notation specific to individual problems is introduced in later chapters.

    We adopt non-asymptotic big-oh notation: For functions
	$f,g:\cZ\to\bbR_{+}$, we write $f=\bigoh(g)$ (resp. $f=\bigom(g)$) if there exists a constant
	$C>0$ such that $f(z)\leq{}Cg(z)$ (resp. $f(z)\geq{}Cg(z)$)
        for all $z\in\cZ$. We write $f=\bigoht(g)$ if
        $f=\bigoh(g\cdot\mathrm{polylog}(T))$, $f=\bigomt(g)$ if $f=\bigom(g/\polylog(T))$.
         We use $\lesssim$ only in informal statements to
highlight salient elements of an inequality.

	For a vector $z\in\bbR^{d}$, we let $\nrm*{z}$ denote the euclidean
	norm. We define $\nrm{z}_{W}^{2}\ldef{}\tri{z,Wz}$ for a positive definite matrix $W \in \R^{d \times d}$.
  For an integer $n\in\bbN$, we let $[n]$ denote the set
        $\{1,\dots,n\}$.  
        For a set $\cZ$, we let
        $\Delta(\cZ)$ denote the set of all Radon probability measures
        over $\cZ$. We let $\conv(\cZ)$ denote the set of all finitely
        supported convex combinations of elements in $\cZ$.
        When $\cZ$ is finite, we let $\unif(\cZ)$ denote the uniform distribution over all the
        elements in $\cZ$.
        We let $\indic_{z}\in\Delta(\cZ)$ denote the delta distribution on $z$.
         We use the convention
         $a\wedge{}b=\min\crl{a,b}$ and $a\vee{}b=\max\crl{a,b}$.

\part{Active Learning with Noisy Data and Rich Model Classes}
\label{part:active}

\chapter{Efficient Active Learning with Abstention}
\label{chapter:active:efficient}

The goal of active learning is to achieve the same accuracy achievable by passive learning, while using much fewer labels. Exponential savings in terms of label complexity have been proved in very special cases, but fundamental lower bounds show that such improvements are impossible in general. This suggests a need to explore alternative goals for active learning. Learning with abstention is one such alternative.  In this setting, the active learning algorithm may abstain from prediction and incur an error that is marginally smaller than random guessing. We develop the first computationally efficient active learning algorithm with abstention. Our algorithm provably achieves $\mathsf{polylog}(\frac{1}{\varepsilon})$ label complexity, without any low noise conditions. Such performance guarantee reduces the label complexity by an exponential factor, relative to passive learning and active learning that is not allowed to abstain. Furthermore, our algorithm is guaranteed to only abstain on hard examples (where the true label distribution is close to a fair coin), a novel property we term \emph{proper abstention} that also leads to a host of other desirable characteristics (e.g., recovering minimax guarantees in the standard setting, and avoiding the undesirable ``noise-seeking'' behavior often seen in active learning). We also provide novel extensions of our algorithm that achieve \emph{constant} label complexity and deal with model misspecification.

\section{Introduction}

Active learning aims at learning an accurate classifier with a small number of labeled data points \citep{settles2009active, hanneke2014theory}. Active learning has become increasingly important in modern application of machine learning, where unlabeled data points are abundant yet the labeling process requires expensive time and effort. Empirical successes of active learning have been observed in many areas \citep{tong2001support, gal2017deep, sener2018active}.
In noise-free or certain low-noise cases (i.e., under Massart noise \citep{massart2006risk}), active learning algorithms with \emph{provable} exponential savings over the passive counterpart have been developed  \citep{balcan2007margin, hanneke2007bound,  dasgupta2009analysis, hsu2010algorithms, dekel2012selective, hanneke2014theory, zhang2014beyond, krishnamurthy2019active, katz2021improved}.
On the other hand, however, not much can be said in the general case. 
In fact, \citet{kaariainen2006active} provides a $\Omega(\frac{1}{\epsilon^2})$ lower bound by reducing active learning to a simple mean estimation problem: It takes $\Omega(\frac{1}{\epsilon^2})$ samples to distinguish $\eta(x) = \frac{1}{2} + \epsilon$ and $\eta(x) = \frac{1}{2} - \epsilon$. Even with the relatively benign Tsybakov noise \citep{tsybakov2004optimal}, \citet{castro2006upper, castro2008minimax} derive a $\Omega( \poly(\frac{1}{\epsilon}))$ lower bound, again, indicating that exponential speedup over passive learning is not possible in general. These fundamental lower bounds lay out statistical barriers to active learning, and suggests considering a refinement of the label complexity goals in active learning \citep{kaariainen2006active}.

Inspecting  these lower bounds, one can see that active learning suffers from classifying hard examples that are close to the decision boundary. However, \emph{do we really require a trained classifier to do well on those hard examples?} In high-risk domains such as medical imaging, it makes more sense for the classifier to abstain from making the decision and leave the problem to a human expert. Such idea is formalized under Chow's error \citep{chow1970optimum}: Whenever the classifier chooses to abstain, a loss that is barely smaller than random guessing, i.e., $\frac{1}{2} - \gamma$, is incurred. 
The parameter $\gamma$ should be thought as a small positive quantity, e.g., $\gamma = 0.01$.
The inclusion of abstention is not only practically interesting, but also provides a statistical refinement of the label complexity goal of active learning: Achieving exponential improvement under Chow's excess error.
When abstention is allowed as an action, \citet{puchkin2021exponential} shows, for the first time, that exponential improvement in label complexity can be achieved by active learning in the general setting.  
However, the approach provided in \citet{puchkin2021exponential} can not be efficiently implemented. 
Their algorithm follows the disagreement-based approach and requires maintaining a version space and checking whether or not an example lies in the region of disagreement.
It is not clear how to generally implement these operations besides enumeration \citep{beygelzimer2010agnostic}.
Moreover, their algorithm relies on an Empirical Risk Minimization (ERM) oracle, which is known to be NP-Hard even for a simple linear hypothesis class \citep{guruswami2009hardness}.

In this chapter, we break the computational barrier and design an efficient active learning algorithm with exponential improvement in label complexity relative to conventional passive learning.
The algorithm relies on weighted square loss regression oracle, which can be efficiently implemented in many cases \citep{krishnamurthy2017active, krishnamurthy2019active, foster2018practical, foster2020instance}. The  algorithm  also abstains properly, i.e., abstain only when it is the optimal choice, which allows us to easily translate the guarantees to the \emph{standard} excess error.
Along the way, we propose new noise-seeking noise conditions and show that: ``uncertainty-based'' active learners can be easily trapped, yet our algorithm provably overcome these noise-seeking conditions.
As an extension, we also provide the first algorithm that enjoys \emph{constant} label complexity for a \emph{general} set of regression functions.

\subsection{Problem Setting}
\label{al_abs:sec:setting}

\vspace{+2 pt}

Let $\cX$ denote the input space and $\cY$ denote the label space. We focus on the binary classification problem where $\cY = \curly*{0, 1}$. The joint distribution over $\cX \times \cY$ is denoted as $\cD_{\cX \cY}$. We use $\cD_{\cX}$ to denote the marginal distribution over the input space $\cX$, and use $\cD_{\cY \vert x}$ to denote the conditional distribution of $\cY$ with respect to any $x \in \cX$.
We define $\eta(x) \ldef \P_{y \sim \cD_{\cY \vert x}} (y = 1)$ as the conditional probability of taking the label of $1$.
We consider the standard active learning setup where $(x,y) \sim \cD_{\cX \cY}$ but $y$ is observed only after a label querying.
We consider hypothesis class $\cH: \cX \rightarrow \cY$. For any classifier $h \in \cH$, its (standard) error is defined as $\err(h) \ldef \P_{(x,y) \sim \cD_{\cX \cY}} (h(x) \neq y)$.

\paragraph{Function approximation}
We focus on the case where the hypothesis class $\cH$ is induced from a set of regression functions $\cF: \cX \rightarrow [0,1]$ that predicts the conditional probability $\eta(x)$. We write $\cH = \cH_{\cF} \coloneqq \curly*{h_f : f \in \cF}$ where $h_f(x) \ldef \ind(f(x) \geq 1/2)$. 
The complexity of $\cF$ is measured by the well-known complexity measure: the \emph{Pseudo dimension} $\pseud(\cF)$ \citep{pollard1984convergence, haussler1989decision,haussler1995sphere}; 
we assume $\pseud(\cF) < \infty$ throughout the paper.\footnote{See \cref{al_abs:app:concentration} for formal definition of the Pseudo dimension. Many function classes of practical interests have finite Pseudo dimension: (1) when $\cF$ is finite, we have $\pseud(\cF) = O(\log \abs{\cF})$; (2) when $\cF$ is a set of linear functions/generalized linear function with non-decreasing link function, we have  $\cF = O(d)$; (3) when $\cF$ is a set of degree-$r$ polynomial in $\R^{d}$, we have $\pseud(\cF) = O({d+r \choose r})$.
}
Following existing works in active learning \citep{dekel2012selective,krishnamurthy2017active, krishnamurthy2019active} and contextual bandits \citep{agarwal2012contextual, foster2018practical, foster2020beyond, simchi2020bypassing}, we make the following \emph{realizability} assumption.

\begin{assumption}[Realizability]
\label{al_abs:asmp:predictable}
The learner is given a set of regressors $\cF: \cX \to [0, 1]$ such that there exists a $f^\star \in \cF$ characterize the true conditional probability, i.e., $f^\star = \eta$.
\end{assumption}

The realizability assumption allows \emph{rich function approximation}, which strictly generalizes the setting with linear function approximation studied in active learning (e.g., in \citep{dekel2012selective}).
We relax \cref{al_abs:asmp:predictable} in \cref{al_abs:sec:misspecified} to deal with model misspecification.

\paragraph{Regression oracle}
We consider a regression oracle over $\cF$, which is extensively studied in the literature in active learning and contextual bandits \citep{krishnamurthy2017active, krishnamurthy2019active, foster2018practical, foster2020instance}.
Given any set $\cS$ of weighted examples $(w, x, y) \in \R_+ \times \cX \times \cY$ as input, the regression oracle outputs 
\begin{align}
\label{al_abs:eq:regression_oracle}	
    \widehat f = \argmin_{f \in \cF} \sum_{(w, x, y) \in \cS} w \paren*{f(x) - y}^2.
\end{align}
The regression oracle solves a convex optimization problem with respect to the regression function, and admits closed-form solutions in many cases, e.g., it is reduced to least squares when $f$ is linear.
We view the implementation of the regression oracle as an efficient operation and quantify the computational complexity in terms of the number of calls to the regression oracle.

\paragraph{Chow's excess error \citep{chow1970optimum}} 
Let $h^\star \ldef h_{f^{\star}} \in \cH$ denote the Bayes classifier.
The \emph{standard excess error} of classifier $h \in \cH$ is defined as $\err(h) - \err(h^\star)$. 
Since achieving exponential improvement (of active over passive learning) with respect to the standard excess error is impossible in general \citep{kaariainen2006active}, we introduce Chow's excess error next.
We consider classifier of the form $\widehat h: \cX \rightarrow \cY \cup \curly*{\bot}$ where $\bot$ denotes the action of abstention. For any fixed $0 < \gamma < \frac{1}{2}$, the Chow's error is defined as 
\begin{align}
    \err_{\gamma}(\widehat h)  \ldef \P_{(x,y) \sim \cD_{\cX \cY}} \prn{\widehat h (x) \neq y,  \widehat h(x) \neq \bot} + \prn*{{1}/{2} - \gamma} \cdot \P_{(x,y) \sim \cD_{\cX \cY}} \prn{\widehat h(x) = \bot}.
    \label{al_abs:eq:chow_error}
\end{align}

The parameter $\gamma$ can be chosen as a small constant, e.g., $\gamma = 0.01$, to avoid excessive abstention: The price of abstention is only marginally smaller than random guess.
The \emph{Chow's excess error} is then defined as $\err_{\gamma}(\wh h) - \err(h^\star)$ \citep{puchkin2021exponential}.
For any fixed accuracy level $\epsilon >0$, we aim at constructing a classifier $\widehat h: \cX \rightarrow \cY \cup \curly*{\bot}$ with $\eps$ Chow's excess error and $\polylog(\frac{1}{\eps})$ label complexity.   
We also relate Chow's excess error to standard excess error in \cref{al_abs:sec:standard_excess_error}.

\subsection{Why Chow's Excess Error Helps Learning?}
\label{al_abs:sec:chow}

We study the simple case where $\cX = \curly*{x}$ to illustrate the benefits of learning under Chow's excess error.
In this setting, the active learning problem reduces to mean estimation of the conditional probability $\eta(x) \in [0,1]$.
In the following, we compare learning behavior under standard excess error, Chow's excess error, and Chow's excess error relative to the optimal abstaining classifier.

\begin{figure}[t]
    \centering
    \includegraphics[width=\linewidth]{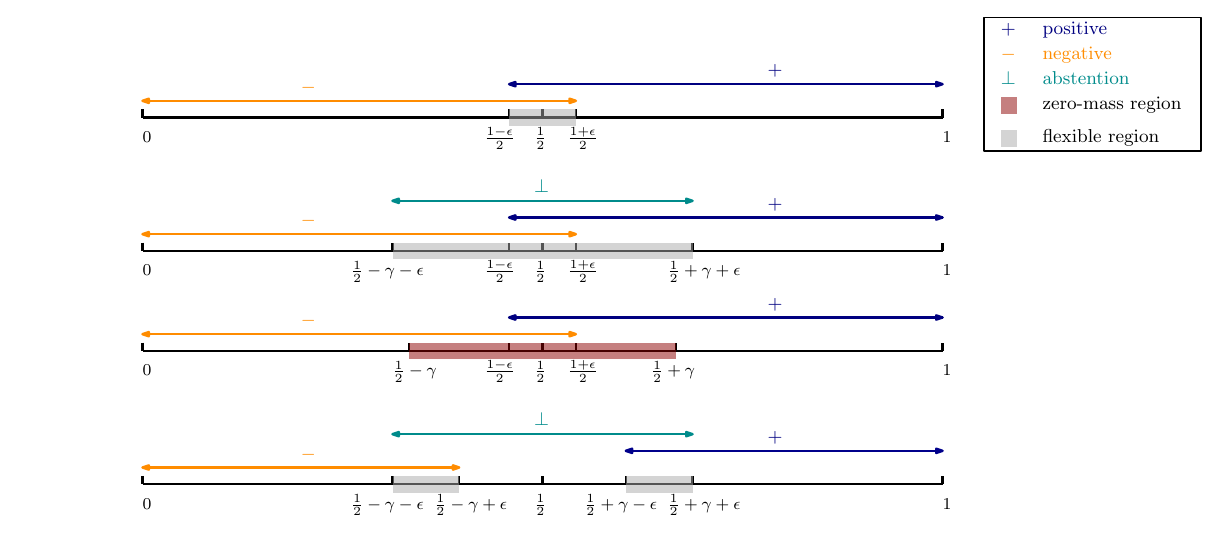}
    \caption{
    Illustration of decision regions under different error criteria.
    \emph{Top:} standard excess error $\err(\widehat h) - \err(h^\star)$. 
    \emph{Second:} Chow's excess error $\err_{\gamma}(\widehat h) - \err(h^\star)$. 
    \emph{Third:} standard excess error $\err(\widehat h) - \err(h^\star)$ under Massart noise condition with parameter $\gamma$. 
    \emph{Bottom:} Chow's excess error relative to the optimal abstaining classifier, i.e., $\err_{\gamma}(\wh h) -\inf_{h: \cX \rightarrow \crl{0, 1,\bot}} \err_{\gamma}(h)$. 
    In this figure, positive corresponds to predicting label 1 and negative to predicting label 0.
    }
    \label{al_abs:fig:flexible_region}
\end{figure}

\paragraph{Learning under standard excess error}
Fix any $\epsilon > 0$. With respect to the conditional probability $ \eta(x)$, we define the \emph{positive region} $\cS_{+,\epsilon} \coloneqq \sq{\frac{1-\epsilon}{2}, 1}$ and the \emph{negative region} $\cS_{-,\epsilon} \coloneqq [0, \frac{1+\epsilon}{2}]$;
here, positive (resp. negative) refers to predicting label 1 (resp. 0).
These regions have the following interpretation:
if $\eta(x) \in \cS_{+,\epsilon}$ (resp. $\eta(x) \in \cS_{-,\eps}$), then labeling $x$ as 1 (resp. 0) incurs no more than $\epsilon$ standard excess error.
Under standard excess error, 
we define the \emph{flexible region} as $\cS_{\flex, \epsilon}^{\stan} \coloneqq \cS_{+, \epsilon} \cap \cS_{-, \epsilon} = [\frac{1-\epsilon}{2}, \frac{1+\epsilon}{2}]$, corresponding to the overlap of $\cS_{+, \eps}$ and $\cS_{-, \eps}$ (highlighted as the grey region in the top plot in \cref{al_abs:fig:flexible_region}).
We have two key observations: (1) if $\eta(x) \in \cS_{\flex, \epsilon}^{\stan}$, then labeling $x$ as either 0 or 1 guarantees excess error at most $\epsilon$; and (2) if $\eta(x) \notin \cS_{\flex, \epsilon}^{\stan}$, achieving excess error at most $\eps$ requires correctly labeling $x$ as 0 or 1.
Since the flexible region has length $\eps$, it is possible to construct two learning scenarios where their $\eta(x)$ values differ by $O(\eps)$ yet require different labels. For instance, distinguishing between $\eta(x) = \frac{1}{2} - \eps$ and $\eta(x) = \frac{1}{2} + \eps$ yields a label complexity lower bound of $\Omega({1}/{\eps^2})$.

\paragraph{Learning under Chow's excess error}
We now consider learning under Chow's excess error.
As before, we define the positive and negative regions $\cS_{+,\epsilon} \coloneqq \sq{\frac{1-\epsilon}{2}, 1}$ and $\cS_{-,\epsilon} \coloneqq \sq{0, \frac{1+\epsilon}{2}}$.
Additionally, we introduce the \emph{abstention region}:
$\cS_{\bot, \epsilon} \ldef [\frac{1}{2}-\gamma-\epsilon, \frac{1}{2}+\gamma+\epsilon]$,
where abstaining on $x$ when $\eta(x) \in \cS_{\bot,\epsilon}$ incurs at most $\epsilon$ Chow's excess error.
Under Chow's excess error, the flexible region is enlarged thanks to the added abstention choice. We now have \emph{positive flexible region} $\cS_{\flex,+,\epsilon}^{\chow} \coloneqq \cS_{+,\epsilon} \cap \cS_{\bot,\epsilon} = \sq{\frac{1-\epsilon}{2}, \frac{1}{2} + \gamma + \epsilon}$, and \emph{negative flexible region} $\cS_{\flex,-,\epsilon}^{\chow} \coloneqq \cS_{-,\epsilon} \cap \cS_{\bot,\epsilon} = \sq{\frac{1}{2} - \gamma - \epsilon, \frac{1+\epsilon}{2}}$, both have length $\gamma + \frac{3 \eps}{2}$ (see the second plot in \cref{al_abs:fig:flexible_region}).
These enlarged flexible regions imply that Chow's excess error can be controlled with fewer samples.
Specifically, it suffices to identify whether $\eta(x)$ lies within $\cS_{\bot,\epsilon}$ or confidently predicts $0/1$.
Constructing a confidence interval of length at most $\gamma/2$ requires $\wt O(1/\gamma^2)$ samples.
If $\eta(x) \in \sq{\frac{1 - \gamma}{2}, \frac{1 + \gamma}{2}}$, the confidence interval lies entirely within $\cS_{\bot,\epsilon}$, certifying the abstention acheives at most $\eps$ Chow's excess error.
If $\eta(x) < \frac{1 - \gamma}{2}$, the upper bound of the interval satisfies $\ucb(x) \le \frac{1}{2}$, certifying that labeling $x$ as $0$ achieves at most $\epsilon$ excess error.
Similarly, if $\eta(x) > \frac{1 + \gamma}{2}$, labeling $x$ as $1$ achieves at most $\epsilon$ excess error.
In summary, learning under Chow's excess error behaves similarly to learning under Massart noise (see the third plot in \cref{al_abs:fig:flexible_region}).
Examples near the decision boundary are effectively filtered out by abstention, and reliable learning is achievable with $\wt O(1/\gamma^2)$ samples.

\oldparagraph{Why not compete against the optimal abstaining classifier?} 
We use $\err_{\gamma}(\wh{h}) - \inf_{h : \cX \to \{0, 1, \bot\}} \err_{\gamma}(h)$ 
to denote the excess error relative to the optimal classifier that is allowed to abstain.
As shown in the bottom plot of \cref{al_abs:fig:flexible_region}, when competing against the optimal abstaining classifier, the flexible regions shrink back to length $O(\epsilon)$.
This occurs because abstention is the only action that guarantees at most $\epsilon$ excess error over the region $\prn{\frac{1}{2} - \gamma + \epsilon, \frac{1}{2} + \gamma - \epsilon}$.
Consequently, the learner must distinguish between cases like $\eta(x) = \frac{1}{2} + \gamma - 2 \epsilon$ and $\eta(x) = \frac{1}{2} + \gamma + 2 \epsilon$, which requires $\Omega(1 / \epsilon^2)$ samples.
Competing against the optimal abstaining classifier is also unreasonable.
For example, when $\eta(x) = \frac{1}{2} + \gamma - 2 \epsilon$, deciding whether to label $x$ as $1$ or abstain demands $\Omega(1 / \epsilon^2)$ samples.
Yet with only $\widetilde O(1 / \gamma^2)$ samples, the learner can already confidently determine that $\eta(x) > \frac{1}{2}$ and safely predict label $1$.

\subsection{Contributions and Organization}

We provide informal statements of our main results in this section.
Our results depend on complexity measures such as \emph{value function} disagreement coefficient $\theta$ and eluder dimension $\mfe$ (formally defined in \cref{al_abs:sec:epoch} and \cref{al_abs:app:star_eluder}). 
These complexity measures are previously analyzed in contextual bandits \citep{russo2013eluder, foster2020instance} and we import them to the active learning setup.
These complexity measures are well-bounded for many function classes of practical interests, e.g., 
we have $\theta, \mfe = \wt O(d)$ for linear and generalized linear functions in $\R^d$.
\looseness=-1

Our first main contribution is that we design the first \emph{computationally efficient} active learning algorithm (\cref{al_abs:alg:epoch}) that achieves exponential labeling savings, \emph{without any low noise assumptions}.

\begin{theorem}[Informal]
\label{al_abs:thm:epoch_informal}
    There exists an algorithm that constructs a classifier $\wh h: \cX \rightarrow \crl*{0, 1, \bot}$ with Chow's excess error at most $\eps$ and label complexity $\wt O \prn{\frac{\theta \, \pseud(\cF)}{\gamma^2} \cdot \polylog(\frac{1}{\eps})}$, without any low noise assumptions. The algorithm can be efficiently implemented via a regression oracle: It takes $\wt O \prn{\frac{\theta \, \pseud(\cF)}{\eps \, \gamma^3}}$ oracle calls for general $\cF$, and $\wt O \prn{ \frac{\theta \, \pseud(\cF)}{\eps \, \gamma} }$ oracle calls for convex $\cF$.
\end{theorem}

The formal statements are provided in \cref{al_abs:sec:epoch}.
The \emph{statistical} guarantees (i.e., label complexity) in \cref{al_abs:thm:epoch_informal} is similar to the one achieved in \citet{puchkin2021exponential}, with one critical difference: The label complexity provided in \citet{puchkin2021exponential} is in terms of the \emph{classifier-based} disagreement coefficient $\check \theta$ \citep{hanneke2014theory}. 
Even for a set of linear classifier, $\check \theta$ is only known to be bounded in special cases, e.g., when $\cD_\cX$ is uniform over the unit sphere \citep{hanneke2007bound}.
On the other hand, we have  $\theta \leq d$ for any  $\cD_\cX$ \citep{foster2020instance}.

We say that a classifier $\wh h : \cX \rightarrow \crl{0, 1,\bot}$ enjoys proper abstention if it abstains only if abstention is indeed the optimal choice (based on \cref{al_abs:eq:chow_error}).
For any classifier that enjoys proper abstention, one can easily relate its \emph{standard} excess error to the Chow's excess error, under commonly studied Massart/Tsybakov noises \citep{massart2006risk, tsybakov2004optimal}.
The classifier obtained in \cref{al_abs:thm:epoch_informal} enjoys proper abstention, and achieves the following guarantees (formally stated in \cref{al_abs:sec:minimax}).

\begin{theorem}[Informal]
\label{al_abs:thm:proper_abs_informal}
Under Massart/Tsybakov noise, with appropriate adjustments,
the classifier learned in \cref{al_abs:thm:epoch_informal} achieves the minimax optimal label complexity under standard excess error.
\end{theorem}

We also propose new noise conditions that \emph{strictly} generalize the usual Massart/Tsybakov noises, which we call noise-seeking conditions.
At a high-level, the noise-seeking conditions allow abundant data points with $\eta(x)$ equal/close to $\frac{1}{2}$. These points are somewhat ``harmless'' since it hardly matters what label is predicted at that point (in terms of excess error).
These seemingly ``harmless'' data points can, however, cause troubles for any active learning algorithm that requests the label for any point that is uncertain, i.e., the algorithm cannot decide if $\abs{\eta(x)-\frac{1}{2}}$ is strictly greater than $0$. We call such algorithms ``uncertainty-based'' active learners.
These algorithms could wastefully sample in these ``harmless'' regions, ignoring other regions where erring could be much more harmful.
We derive the following 
proposition (formally stated in \cref{al_abs:sec:noise_seeking}) under these noise-seeking conditions.

\begin{proposition}[Informal]
\label{al_abs:prop:budget_informal}
For any labeling budget $B \gtrsim \frac{1}{\gamma^2} \cdot \polylog(\frac{1}{\eps}) $, there exists a learning problem such that (1) any uncertainty-based active learner suffers standard excess error $\Omega(B^{-1})$; yet (2) the classifier $\wh h$ learned in \cref{al_abs:thm:epoch_informal} achieves standard excess error at most $\eps$.
\end{proposition}

The above result demonstrates the superiority of our algorithm over any
\linebreak
``uncertainty-based'' active learner.
Moreover, we show that, under these strictly harder noise-seeking conditions, our algorithm still achieve guarantees similar to the ones stated in \cref{al_abs:thm:proper_abs_informal}.

Before presenting our next main result, we first consider a simple active learning problem with $\cX=\crl{x}$.
Under Massart noise, we have $\abs{\eta(x) - \frac{1}{2}} \geq \tau_0$ for some constant $\tau_0 >0$. Thus, it takes no more than $O(\tau_0^{-2} \log\frac{1}{\delta})$ labels to achieve $\eps$ standard excess error, no matter how small $\eps$ is. 
This example shows that, at least in simple cases, we can expect to achieve a \emph{constant} label complexity for active learning, with no dependence on $\frac{1}{\eps}$ at all.
To the best of our knowledge,
our next result provides the first generalization of such phenomenon to a \emph{general} set of (finite) regression functions, as long as its eluder dimension $\mfe$ is bounded.
\begin{theorem}[Informal]
\label{al_abs:thm:eluder_informal}
    Under Massart noise with parameter $\tau_0$ and a general (finite) set of regression function $\cF$. There exists an algorithm that returns a classifier with standard excess error at most $\eps$ and label complexity $O\paren{ \frac{\mfe \cdot \log ({\abs*{\cF}}/{\delta})}{\tau_0^2}}$, which is independent of $\frac{1}{\eps}$.
\end{theorem}

A similar constant label complexity holds with Chow's excess error, without any low noise assumptions.
We also provide discussion on why previous algorithms do not achieve such constant label complexity, even in the case with linear functions.
We defer formal statements and discussion to \cref{al_abs:sec:constant}.
In \cref{al_abs:sec:misspecified}, we relax \cref{al_abs:asmp:predictable} and propose an algorithm that can deal with model misspecification.

\paragraph{Organization}
The rest of this chapter is organized as follows.
We discuss additional related work in \cref{al_abs:sec:related_work}.
We present our main algorithm and its guarantees in \cref{al_abs:sec:epoch}. 
In \cref{al_abs:sec:standard_excess_error}, we analyze our algorithm under standard excess error and discuss other key properties.
Extensions of the algorithm, including achieving \emph{constant} label complexity and handling model misspecification, are presented in \cref{al_abs:sec:extension}.
Additional definitions and all proofs are deferred to \cref{al_abs:al_abs:sec:proofs}.

\subsection{Additional Related Work}
\label{al_abs:sec:related_work}

Learning under Chow's excess error is closely related to learning under Massart noise \citep{massart2006risk}, which assumes that no data point has conditional expectation close to the decision boundary,
i.e., $\P \prn*{ \abs{\eta(x) - 1 / 2} \leq \tau_{0} } = 0$ for some constant $\tau_0> 0$.
Learning under Massart noise is commonly studied in active learning \citep{balcan2007margin, hanneke2014theory, zhang2014beyond, krishnamurthy2019active}, where $\wt O(\tau_0^{-2})$ type of guarantees are achieved. 
Instead of making explicit assumptions on the underlying distribution, learning with Chow's excess error empowers the learner with the ability to abstain:
There is no need to make predictions on hard data points that are close to the decision boundary, 
i.e., $\crl{x: \abs{ \eta(x) - 1 / 2} \leq \gamma }$. 
Learning under Chow's excess error thus works on more general settings and still enjoys the 
$\wt O(\gamma^{-2})$ type of guarantee as learning under Massart noise \citep{puchkin2021exponential}.\footnote{However, passive learning with abstention only achieves error rate $\frac{1}{n \gamma}$ with $n$ samples \citep{bousquet2021fast}.}
We show in \cref{al_abs:sec:standard_excess_error} that statistical guarantees achieved under Chow's excess error can be directly translated to guarantees under (usual and more challenging versions of) Massart/Tsybakov noise \citep{massart2006risk, tsybakov2004optimal}.

Active learning at aim competing the best in-class classifier with few labels. 
A long line of work directly works with the set of classifiers \citep{balcan2007margin, hanneke2007bound, hanneke2014theory, huang2015efficient, puchkin2021exponential}, where the algorithms are developed with (in general) hard-to-implement ERM oracles \citep{guruswami2009hardness} and the the guarantees dependence on the so-called disagreement coefficient \citep{hanneke2014theory}.
More recently, learning with function approximation have been studied inactive learning and contextual bandits \citep{dekel2012selective, agarwal2012contextual, foster2018practical, krishnamurthy2019active}.
The function approximation scheme permits efficient regression oracles, which solve convex optimization problems with respect to regression functions \citep{krishnamurthy2017active,krishnamurthy2019active,foster2018practical}.
It can also be analyzed with the scale-sensitive version of disagreement coefficient, which is usually tighter than the original one \citep{foster2020instance, russo2013eluder}.
Our algorithms are inspired \citet{krishnamurthy2019active}, where the authors study active learning under the standard excess error. 
The main deviation from \citet{krishnamurthy2019active} is that we need to \emph{manually} construct a classifier $\wh h$ with an abstention option and $\wh h \notin \cH$, which leads to differences in the analysis of excess error and label complexity.
We borrow techniques developed in contextual bandits \citet{russo2013eluder, foster2020instance} to analyze our algorithm.

Although one can also apply our algorithms in the nonparametric regime with proper pre-processing schemes such discretizations, our algorithm primarily works in the parametric setting with finite pseudo dimension \citep{haussler1995sphere} and finite (value function) disagreement coefficient \citep{foster2020instance}.
Active learning has also been studied in the nonparametric regime \citep{castro2008minimax,koltchinskii2010rademacher, minsker2012plug, locatelli2017adaptivity}.
Notably, \citet{shekhar2021active} studies Chow's excess error with margin-type of assumptions. 
Their setting is different to ours and $\poly(\frac{1}{\eps})$ label complexities are achieved.
If abundant amounts of data points are allowed to be exactly at the decision boundary, i.e., $\eta(x) = \frac{1}{2}$,
\citet{kpotufe2021nuances} recently shows that, in the nonparametric regime, no active learner can outperform the passive counterpart.

\section{Efficient Active Learning with Abstention}
\label{al_abs:sec:epoch}
We provide our main algorithm (\cref{al_abs:alg:epoch}) in this section. \cref{al_abs:alg:epoch} is an adaptation of the algorithm developed in \citet{krishnamurthy2017active, krishnamurthy2019active}, which studies active learning under the standard excess error (and Massart/Tsybakov noises).
We additionally take the abstention option into consideration, and \emph{manually construct} classifiers using the active set of (uneliminated) regression functions (which do not belong to the original hypothesis class).
These new elements allow us to achieve $\eps$ Chow's excess error with $\polylog(\frac{1}{\eps})$ label complexity, without any low noise assumptions.

\begin{algorithm}[]
	\caption{Efficient Active Learning with Abstention}
	\label{al_abs:alg:epoch} 
	\renewcommand{\algorithmicrequire}{\textbf{Input:}}
	\renewcommand{\algorithmicensure}{\textbf{Output:}}
	\newcommand{\algorithmicbreak}{\textbf{break}}
    \newcommand{\BREAK}{\STATE \algorithmicbreak}
	\begin{algorithmic}[1]
		\REQUIRE Accuracy level $\eps >0 $, abstention parameter $\gamma \in (0, 1/2)$ and confidence level $\delta \in (0, 1)$.
		\STATE 
Define $T \ldef \wt O \prn{\frac{\theta \, \pseud(\cF)}{\eps \, \gamma}}$,
		$M \ldef \ceil{\log_2 T}$ and $C_\delta \ldef O \prn*{\pseud(\cF) \cdot \log( T/\delta)}$.
		
		\STATE Define $\tau_m \ldef 2^m$ for $m\geq1$, $\tau_0 \ldef 0$ and $\beta_m \ldef (M-m+1) \cdot C_\delta$. 
		\FOR{epoch $m = 1, 2, \dots, M$}
		\STATE Get $\widehat f_m \ldef \argmin_{f \in \cF} \sum_{t=1}^{\tau_{m-1}} Q_t \paren{f(x_t) - y_t}^2 $.\\
		\hfill \algcommentlight{We use $Q_t \in \crl{0,1}$ to indicate whether the label of  $x_t$ is queried.}
		\STATE (Implicitly) Construct active set of regression functions $\cF_m \subseteq \cF$ as
		\begin{align*}
		    \cF_m \ldef \crl*{ f \in \cF:  \sum_{t = 1}^{\tau_{m-1}} Q_t \prn*{f(x_t) - y_t}^2 \leq \sum_{t = 1}^{\tau_{m-1}} Q_t \paren{\widehat f_m(x_t) - y_t}^2 + \beta_m }. 
\end{align*}
		\STATE Construct classifier $\wh h_m: \cX \rightarrow \crl{0, 1,\bot}$ as 
		\begin{align*}
			\wh h_m (x) \ldef 
			\begin{cases}
				\bot, & \text{ if } \brk { \lcb(x;\cF_m), \ucb(x;\cF_m)} \subseteq 
				\brk*{ \frac{1}{2} - \gamma, \frac{1}{2} + \gamma}; \\
        \ind(\wh f_m(x) \geq \frac{1}{2} ) , & \text{o.w.}
			\end{cases}
\end{align*}
and construct query function $g_m(x)\ldef \ind \prn*{ \frac{1}{2} \in \prn{\lcb(x;\cF_m), \ucb(x;\cF_m)} } \cdot 
		\ind \prn{\wh h_m(x) \neq \bot}$.
		\IF{epoch $m=M$}
		\STATE \textbf{Return} classifier $\wh h_M$.
		\ENDIF
		\FOR{time $t = \tau_{m-1} + 1 ,\ldots , \tau_{m} $} 
		\STATE Observe $x_t \sim \cD_{\cX}$. Set $Q_t \ldef g_m(x_t)$.
		\IF{$Q_t = 1$}
		\STATE Query the label $y_t$ of $x_t$.
		\ENDIF
		\ENDFOR
		\ENDFOR

	\end{algorithmic}
\end{algorithm}

\cref{al_abs:alg:epoch} runs in epochs of geometrically increasing lengths.
At the beginning of epoch $m \in [M]$, \cref{al_abs:alg:epoch} first computes the empirical best regression function $\wh f_m$ that achieves the smallest cumulative square loss over previously labeled data points ($\wh f_1$ can be selected arbitrarily); it then (implicitly) constructs an active set of regression functions $\cF_m$, where the cumulative square loss of each $f \in \cF_m$ is not too much larger than the cumulative square loss of empirical best regression function $\wh f_m$. 
For any $x \in \cX$, based on the active set of regression functions, \cref{al_abs:alg:epoch} constructs a lower bound $\lcb(x;\cF_m) \ldef \inf_{f \in \cF_m} f(x)$ and an upper bound $\ucb(x;\cF_m) \ldef \sup_{f \in \cF_m} f(x)$ for the true conditional probability $\eta(x)$.
An empirical classifier $\wh h_m: \cX \rightarrow \crl{0, 1, \bot}$ and a query function $g_m:\cX \rightarrow \crl{0, 1}$ are then constructed based on these confidence ranges and the abstention parameter $\gamma$.
For any time step $t$ within epoch $m$, \cref{al_abs:alg:epoch} queries the label of the observed data point $x_t$ if and only if $Q_t \ldef g_m(x_t) = 1$.
\cref{al_abs:alg:epoch} returns $\wh h_M$ as the learned classifier.

We now discuss the empirical classifier $\wh h_m$ and the query function $g_m$ in more detail.
Consider the event where $f^\star \in \cF_m$ for all $m \in [M]$, which can be shown to hold with high probability.
The constructed confidence intervals are valid under this event, i.e., $\eta(x) \in [\lcb(x;\cF_m), \ucb(x;\cF_m)]$.
First, let us examine the conditions that determine a label query.  The label of $x$ is \emph{not} queried if
\begin{itemize}
	\item \textbf{Case 1: $\wh h_m(x) = \bot$.} We have $\eta(x) \in \brk{ \lcb(x;\cF_m) , \ucb(x;\cF_m)} \subseteq \brk{ \frac{1}{2} - \gamma, \frac{1}{2}+\gamma}$.
Abstention leads to the smallest error \citep{herbei2006classification}, and no query is needed.
  \item \textbf{Case 2: $\frac{1}{2} \notin ( \lcb(x;\cF_m) , \ucb(x;\cF_m))$.} We have $\ind(\wh f_m(x)\geq \frac{1}{2}) = \ind(f^{\star}(x)\geq \frac{1}{2})$. Thus, no excess error is incurred and there is no need to query.
\end{itemize}
The only case when label query \emph{is} issued, and thus when the classifier $\wh h_m$ may suffer from excess error, is when 
\begin{align}
\frac{1}{2} \in ( \lcb(x;\cF_m) , \ucb(x;\cF_m)) \quad \text{and} \quad 
\brk*{ \lcb(x;\cF_m) , \ucb(x;\cF_m)} \nsubseteq \brk*{ \frac{1}{2} - \gamma, \frac{1}{2}+\gamma} 
\label{al_abs:eq:epoch_query}
\end{align}
hold simultaneously. 
\cref{al_abs:eq:epoch_query} necessarily leads to the condition $w(x;\cF_m) \ldef \ucb(x;\cF_m) - \lcb(x;\cF_m) > {\gamma}  $.
Our theoretical analysis shows that the event must $\ind( w(x;\cF_m) > \gamma)$ happens infrequently, and its frequency is closely related to the so-called \emph{value function disagreement coefficient} \citep{foster2020instance}, which we introduce as follows.\footnote{Compared to the original definition studied in contextual bandits \citep{foster2020instance}, our definition takes an additional ``sup'' over all possible marginal distributions $\cD_\cX$ to account for \emph{distributional shifts} incurred by selective querying (which do not occur in contextual bandits). Nevertheless, as we show below, our disagreement coefficient is still well-bounded for many important function classes.}
\begin{definition}[Value function disagreement coefficient]
    \label{al_abs:def:dis_coeff}
    For any $f^{\star} \in \cF$ and $\gamma_0, \eps_0 > 0$,
    the value function disagreement coefficient $	 \theta^{\val}_{f^{\star}}(\cF, \gamma_0, \eps_0)$ is defined as 
    \begin{align*}
 \sup_{\cD_\cX}\sup_{\gamma> \gamma_0, \eps> \eps_0} 
	 \crl*{ \frac{\gamma^2}{\epsilon^2} \cdot 
	\P_{\cD_\cX} \prn*{ \exists f \in \cF: \abs{f(x) - f^{\star}(x)} > \gamma,
	\nrm*{ f - f^{\star}}_{\cD_\cX} \leq \eps} } \vee 1,
    \end{align*}
    where $\nrm{f}^2_{\cD_\cX} \ldef \E_{x \sim \cD_\cX} \brk{f^2(x)}$.
\end{definition}

Combining the insights discussed above, we derive the following label complexity guarantee for \cref{al_abs:alg:epoch} (we use $\theta \ldef \sup_{f^{\star} \in \cF, \iota > 0} \theta^{\val}_{f^{\star}}(\cF,\gamma / 2, \iota)$ and discuss its boundedness below).
\footnote{It suffices to take $\theta \ldef \theta^{\val}_{f^{\star}}(\cF, \gamma /2, \iota)$ with $\iota \propto \sqrt{\gamma \eps}$ to derive a slightly different guarantee. See \cref{al_abs:app:epoch}.}
\begin{restatable}{theorem}{thmEpoch}
\label{al_abs:thm:epoch}
With probability at least $1-2\delta$, \cref{al_abs:alg:epoch} returns a classifier with Chow's excess error at most $\epsilon$ and label complexity 
$O\prn{ \frac{\theta \, \pseud(\cF)}{\gamma^2} \cdot \log^2 \prn{\frac{\theta \, \pseud(\cF)}{\eps \, \gamma}}  \cdot  {\log\prn{\frac{\theta \, \pseud(\cF)}{\eps \, \gamma \,\delta}}}}$.
\end{restatable}

\cref{al_abs:thm:epoch} shows that \cref{al_abs:alg:epoch} achieves exponential label savings (i.e., $\polylog(\frac{1}{\eps})$) without any low noise assumptions. 
We discuss the result in more detail next.
\vspace{-3 pt}
\begin{itemize}
	\item \textbf{Boundedness of $\theta$.}
	The value function disagreement coefficient is well-bounded for many function classes of practical interests.
For instance, we have $\theta \leq d$ for linear functions on $\R^d$ and $\theta \leq C_{\textsf{link}} \cdot d$ for generalized linear functions (where $C_{\textsf{link}}$ is a quantity related to the link function).
Moreover, $\theta$ is \emph{always} upper bounded by complexity measures such as (squared) star number and eluder dimension \citep{foster2020instance}. See \cref{al_abs:app:star_eluder} for the detailed definitions/bounds. 
\vspace{-3 pt}
\item \textbf{Comparison to \citet{puchkin2021exponential}.}
The label complexity bound derived in \cref{al_abs:thm:epoch} is similar to the one derived in \citet{puchkin2021exponential}, with one critical difference: The bound derived in \citet{puchkin2021exponential} is in terms of \emph{classifier-based} disagreement coefficient $\check \theta$ \citep{hanneke2014theory}. Even in the case with linear classifiers, $\check \theta$ is only known to be bounded under additional assumptions, e.g., when $\cD_\cX$ is uniform over the unit sphere.
\vspace{-5 pt}
\end{itemize}

\paragraph{Computational efficiency} 
We discuss how to efficiently implement \cref{al_abs:alg:epoch} with the regression oracle defined in \cref{al_abs:eq:regression_oracle}.
\footnote{Recall that the implementation of the regression oracle should be viewed as an efficient operation since it solves a convex optimization problem with respect to the regression function, and it even admits closed-form solutions in many cases, e.g., it is reduced to least squares when $f$ is linear. On the other hand, the ERM oracle used in \citet{puchkin2021exponential} is NP-hard even for a set of linear classifiers \citep{guruswami2009hardness}.}
Our implementation relies on subroutines developed in \citet{krishnamurthy2017active, foster2018practical}, which allow us to approximate confidence bounds $\ucb(x;\cF_m)$ and $\lcb(x;\cF_m)$ up to $\alpha$ approximation error with $O(\frac{1}{\alpha^{2}} \log \frac{1}{\alpha})$ (or $O(\log \frac{1}{\alpha})$ when $\cF$ is convex and closed under pointwise convergence) calls to the regression oracle.
To achieve the same theoretical guarantees shown in \cref{al_abs:thm:epoch} (up to changes in constant terms),
we show that it suffices to (i) control the approximation error at level $O(\frac{\gamma}{\log T})$, (ii) construct the approximated confidence bounds $\wh \lcb(x;\cF_m)$ and  $\wh \ucb(x;\cF_m)$ in a way such that the confidence region is  non-increasing with respect to the epoch $m$, i.e., $(\wh \lcb(x; \cF_m), \wh \ucb(x;\cF_m)) \subseteq (\wh \lcb(x; \cF_{m-1}), \wh \ucb(x;\cF_{m-1}))$ (this ensures that the sampling region is non-increasing even with \emph{approximated} confidence bounds, which is important to our theoretical analysis), and (iii) use the approximated confidence bounds $\wh \lcb(x;\cF_m)$ and  $\wh \ucb(x;\cF_m)$ to construct the classifier $\wh h_m$ and the query function $g_m$.
We provide our guarantees as follows, and leave details to \cref{al_abs:app:epoch}
(we redefine $\theta \ldef \sup_{f^{\star} \in \cF, \iota > 0} \theta^{\val}_{f^{\star}}(\cF,\gamma / 4, \iota)$ in the \cref{al_abs:thm:epoch_efficient} to account to approximation error).
\begin{restatable}{theorem}{thmEpochEfficient}
	\label{al_abs:thm:epoch_efficient}
	\cref{al_abs:alg:epoch} can be efficiently implemented via the regression oracle and enjoys the same theoretical guarantees stated in \cref{al_abs:thm:epoch}.
	The number of oracle calls needed is $\wt O(\frac{\theta \, \pseud(\cF)}{\eps \, \gamma^{3}})$ for a general set of regression functions $\cF$, and $\wt O(\frac{\theta \, \pseud(\cF)}{\eps \, \gamma})$ when $\cF$ is convex and closed under pointwise convergence.
	The per-example inference time of the learned $\wh h_{M}$ is $\wt O ( \frac{1}{\gamma^2} \log^2 \prn{\frac{\theta \, \pseud(\cF)}{\eps }})$ for general $\cF$, and $\wt O ( \log \frac{1}{\gamma}) $ when $\cF$ is convex and closed under pointwise convergence.
\end{restatable}

With \cref{al_abs:thm:epoch_efficient}, we provide the first computationally efficient active learning algorithm that achieves exponential label savings, without any low noise assumptions.

\section{Guarantees under Standard Excess Error}
\label{al_abs:sec:standard_excess_error}
We provide guarantees for \cref{al_abs:alg:epoch} under \emph{standard} excess error. 
In \cref{al_abs:sec:minimax}, we show that \cref{al_abs:alg:epoch} can be used to recover the usual minimax label complexity under Massart/Tsybakov noise; we also provide a new learning paradigm based on \cref{al_abs:alg:epoch} under limited budget.
In \cref{al_abs:sec:noise_seeking}, we show that \cref{al_abs:alg:epoch} provably avoid the undesired \emph{noise-seeking} behavior often seen in active learning.

\subsection{Recovering Minimax Optimal Label Complexity}
\label{al_abs:sec:minimax}

One way to convert an abstaining classifier $\widehat h: \cX \rightarrow \cY \cup \curly*{\bot}$ into a standard classifier $\check h: \cX \rightarrow \cY$ is by randomizing the prediction in its abstention region, i.e., if $\wh h(x) = \bot$, then its randomized version $\check h(x)$ predicts $0$ and $1$ with equal probability \citep{puchkin2021exponential}. With such randomization, the \emph{standard excess error} of $\check h$ can be characterized as   
\begin{align}
	\label{al_abs:eq:randomization}
    \err(\check h) - \err(h^\star) = \err_{\gamma}(\widehat h) - \err(h^\star) + \gamma \cdot \P_{x \sim \cD_{\cX }} (\widehat h(x) = \bot).
\end{align}
The standard excess error depends on the (random) abstention region of $\wh h$, which is difficult to quantify in general. 
To give a more practical characterization of the standard excess error, we introduce the concept of proper abstention in the following. 

\begin{definition}[Proper abstention]
\label{al_abs:def:proper_abstention}
A classifier $\widehat h : \cX \rightarrow \cY \cup \curly*{\bot}$ enjoys proper abstention if and only if it abstains in regions where abstention is indeed the optimal choice, i.e., 
$\crl[\big]{x \in \cX: \widehat h(x) = \bot} \subseteq \crl*{x \in \cX: \eta(x) \in \brk*{\frac{1}{2} - \gamma , \frac{1}{2} + \gamma  } } \rdef \cX_\gamma$.
\end{definition}

\begin{restatable}{proposition}{propProperAbs}
\label{al_abs:prop:proper_abstention}
The classifier $\wh h$ returned by \cref{al_abs:alg:epoch} enjoys proper abstention. With randomization over the abstention region, we have the following upper bound on its standard excess error
\begin{align}
	\label{al_abs:eq:prop_abstention}
    \err(\check h) - \err(h^\star)  
     \leq \err_{\gamma}(\widehat h) - \err(h^\star) + \gamma \cdot \P_{x \sim \cD_{\cX}} (x \in \cX_{\gamma}).
\end{align}
\end{restatable}

The proper abstention property of $\wh h$ returned by \cref{al_abs:alg:epoch} is achieved via conservation: $\wh h$ will avoid abstention unless it is absolutely sure that abstention is the optimal choice.\footnote{On the other hand, however, the algorithm provided in \citet{puchkin2021exponential} is very unlikely to have such property. In fact, only a small but \emph{nonzero} upper bound of abstention rate is provided (Proposition 3.6 therein) under the Massart noise with $\gamma \leq \frac{\tau_0}{2}$; yet any classifier that enjoys proper abstention should have exactly zero abstention rate.}
To characterize the standard excess error of classifier with proper abstention, we only need to upper bound the term $ \P_{x \sim \cD_{\cX}} (x \in \cX_{\gamma})$, which does \emph{not} depends on the (random) classifier $\wh h$. Instead, it only depends on the marginal distribution. 
We next introduce the common Massart/Tsybakov noise conditions.
\begin{definition}[Massart noise, \citet{massart2006risk}]
	\label{al_abs:def:massart}
  A distribution $\cD_{\cX \cY}$ satisfies the Massart noise condition with parameter $\tau_0> 0$ if
  $\P_{x \sim \cD_\cX} \paren*{ \abs*{\eta(x) - 1 / 2} \leq \tau_0} = 0$.
\end{definition}
\begin{definition}[Tsybakov noise, \citet{tsybakov2004optimal}]
	\label{al_abs:def:tsybakov}
  A distribution $\cD_{\cX \cY}$ satisfies the Tsybakov noise condition with parameter $\beta \geq 0$ and a universal constant $c>0$ if
  $\P_{x \sim \cD_\cX} \paren*{\abs*{\eta(x) - 1 / 2} \leq  \tau} \leq c \, \tau^{\beta}$ for any $\tau > 0$.
\end{definition}

As in \citet{balcan2007margin, hanneke2014theory}, we assume knowledge of noise parameters (e.g., $\tau_0, \beta$).
Together with the active learning lower established in \citet{castro2006upper, castro2008minimax}, and focusing on the dependence of $\eps$, our next theorem shows that \cref{al_abs:alg:epoch} can be used to recover the minimax label complexity in active learning, under the \emph{standard} excess error.

\begin{restatable}{theorem}{thmStandardExcessError}
\label{al_abs:thm:standard_excess_error}
With an appropriate choice of the abstention parameter $\gamma$ in \cref{al_abs:alg:epoch} and randomization over the abstention region, \cref{al_abs:alg:epoch} learns a classifier $\check h$ at the minimax optimal rates: To achieve $\eps$ standard excess error, it takes $\wt \Theta(\tau_0^{-2})$ labels under Massart noise and takes $\wt \Theta \prn{ {\eps}^{ - 2 / (1 + \beta)} }$ labels under Tsybakov noise.
\end{restatable}

\begin{remark}
    In addition to recovering the minimax rates, the proper abstention property is desirable in practice:
It guarantees that $\wh h$ will not abstain on easy examples, i.e., it will not mistakenly flag easy examples as ``hard-to-classify'', thus eliminating unnecessary human labeling efforts.
\end{remark}

\cref{al_abs:alg:epoch} can also be used to provide new learning paradigms in the limited budget setting, which we introduce below.
No prior knowledge of noise parameters are required in this setup.

\paragraph{New learning paradigm under limited budget} 
Given any labeling budget $B>0$, we can then choose $\gamma \approx {B}^{-1/2}$ in \cref{al_abs:alg:epoch} to make sure the label complexity is never greater than $B$ (with high probability).
The learned classifier enjoys Chow's excess error (with parameter $\gamma $) at most $\eps$; its standard excess error (with randomization over the abstention region) can be analyzed by relating the $\gamma \cdot \P_{x \sim \cD_\cX} \prn{ x \in \cX_{\gamma}}$ term in \cref{al_abs:eq:prop_abstention} to the Massart/Tsybakov noise conditions, as discussed above.

\subsection{Abstention to Avoid Noise-Seeking}
\label{al_abs:sec:noise_seeking}

Active learning algorithms sometimes exhibit \emph{noise-seeking} behaviors,
i.e., oversampling in regions where $\eta(x)$ is close to the $\frac{1}{2}$ level.
Such noise-seeking behavior is known to be a fundamental barrier to achieve low label complexity (under standard excess error), e.g., see \citet{kaariainen2006active}.
We show in this section that abstention naturally helps avoiding noise-seeking behaviors and speeding up active learning.

To better illustrate how properly abstaining classifiers avoid noise-seeking behavior, we first introduce new noise conditions below, which strictly generalize the usual Massart/Tsybakov noises.
\begin{definition}[Noise-seeking Massart noise]
	\label{al_abs:def:noise_seeking_Massart}
  A distribution $\cD_{\cX \cY}$ satisfies the noise-seeking Massart noise condition with parameters $0 \leq \zeta_0 < \tau_0 \leq 1 /2  $ if $\P_{x \sim \cD_\cX} \prn{\zeta_0 < \abs{ \eta(x) - 1 / 2} \leq \tau_0} = 0$.	
\end{definition}

\begin{definition}[Noise-seeking Tsybakov noise]
	\label{al_abs:def:noise_seeking_Tsybakov}
  A distribution $\cD_{\cX \cY}$ satisfies the noise-seeking Tsybakov noise condition with parameters $0 \leq \zeta_0 < 1 /2  $, $\beta \geq 0$ and a universal constant $c>0$ if $\P_{x \sim \cD_\cX} \prn{ \zeta_0 < \abs{ \eta(x) - 1 / 2} \leq \tau} \leq  c\, \tau^\beta$ for any $\tau > \zeta_0$.	
\end{definition}

Compared to the standard Massart/Tsybakov noises, these newly introduced noise-seeking conditions allow arbitrary probability mass of data points whose conditional probability $\eta(x)$ is equal/close to $1/2$. 
As a result, they can trick standard active learning algorithms into exhibiting the noise-seeking bahaviors (and hence their names).
We also mention that the parameter $\zeta_0$ should be considered as an \emph{extremely small quantity} (e.g., $\zeta_0 \ll \eps$), with the extreme case corresponding to $\zeta_0 = 0$ (which still allow arbitrary probability for region $\crl{x \in \cX: \eta(x) = 1 /2 }$).

Ideally, any active learning algorithm should not be heavily affected by these noise conditions since it hardly matters (in terms of excess error) what label is predicted over region $\crl{x \in \cX: \abs{\eta(x) - 1 /2} \leq \zeta_0}$.
However, these seemingly benign noise-seeking conditions can cause troubles for any ``uncertainty-based'' active learner, i.e., any active learning algorithm that requests the label for any point that is uncertain (see \cref{al_abs:def:proper_learner} in \cref{al_abs:app:standard} for formal definition).
In particular, under limited budget, we derive the following result.

\begin{restatable}{proposition}{propBudget}
	\label{al_abs:prop:budget}
Fix $\eps, \delta, \gamma > 0$.
    For any labeling budget $B \gtrsim \frac{1}{\gamma^2} \cdot \log^{2}\prn{\frac{1}{\eps \, \gamma}} \cdot \log \prn{\frac{1}{\eps \, \gamma \, \delta}}$, there exists a learning problem (with a set of linear regression functions) satisfying \cref{al_abs:def:noise_seeking_Massart}/\cref{al_abs:def:noise_seeking_Tsybakov} such that 
    (1) any ``uncertainty-based'' active learner suffers expected standard excess error $\Omega(B^{-1})$;
    yet (2) with probability at least $1-\delta$, \cref{al_abs:alg:epoch} returns a classifier with standard excess error at most $\eps$.
\end{restatable}

The above result demonstrates the superiority of our \cref{al_abs:alg:epoch} over any 
\linebreak
``uncertainty-based'' active learner.
Moreover, we show that \cref{al_abs:alg:epoch} achieves similar guarantees as in \cref{al_abs:thm:standard_excess_error} under the strictly harder noise-seeking conditions.
Specifically, we have the following guarantees.

\begin{restatable}{theorem}{thmStandardExcessErrorNoise}
\label{al_abs:thm:standard_excess_error_noise}
With an appropriate choice of the abstention parameter $\gamma$ in \cref{al_abs:alg:epoch} and randomization over the abstention region, \cref{al_abs:alg:epoch} learns a classifier $\check h$ with $\eps + \zeta_0$ standard excess error after querying $\wt \Theta(\tau_0^{-2})$ labels under \cref{al_abs:def:noise_seeking_Massart} or querying $\wt \Theta \prn{ {\eps}^{ - 2 / (1 + \beta)} }$ labels under \cref{al_abs:def:noise_seeking_Tsybakov}.
\end{restatable}

The special case of the noise-seeking condition with $\zeta_0 = 0$ is recently studied in \citep{kpotufe2021nuances}, where the authors conclude that no active learners can outperform the passive counterparts in the \emph{nonparametric} regime.
\cref{al_abs:thm:standard_excess_error_noise} shows that, in the \emph{parametric} setting (with function approximation), \cref{al_abs:alg:epoch} provably overcomes these noise-seeking conditions.

\section{Extensions}
\label{al_abs:sec:extension}
We provide two adaptations of our main algorithm (\cref{al_abs:alg:epoch}) that can 
(1) achieve constant label complexity for a general set of regression functions (\cref{al_abs:sec:constant}); 
and (2) adapt to model misspecification (\cref{al_abs:sec:misspecified}).
These two adaptations can also be efficiently implemented via regression oracle and enjoy similar guarantees stated in \cref{al_abs:thm:epoch_efficient}. We defer computational analysis to \cref{al_abs:app:constant} and \cref{al_abs:app:mis}.

\subsection{Constant label Complexity}
\label{al_abs:sec:constant}

We start by considering a simple problem instance with $\cX = \crl{x}$, where active learning is reduced to mean estimation of $\eta(x)$.
Consider the Massart noise case where $\eta(x) \notin [\frac{1}{2} - \tau_0, \frac{1}{2} + \tau_0]$.
No matter how small the desired accuracy level $\epsilon>0$ is, the learner should not spend more than $O(\frac{\log(1/\delta)}{\tau_0^2})$ labels to correctly classify $x$ with probability at least $1-\delta$, which ensures $0$ excess error. 
In the general setting, but with Chow's excess error, a similar result follows: 
It takes at most $O(\frac{\log(1/\delta)}{\gamma^2})$ samples to verify if $\eta(x)$ is contained in $[\frac{1}{2}-\gamma, \frac{1}{2} + \gamma]$ or not. 
Taking the optimal action within $\crl{0, 1,\bot}$ (based on \cref{al_abs:eq:chow_error}) then leads to $0$ Chow's excess error.
This reasoning shows that, at least in simple cases, one should be able to achieve \emph{constant} label complexity no matter how small $\epsilon$ is. One natural question to ask is as follows.

\begin{minipage}[c]{\linewidth}
\vspace{15 pt}
\centering
  \emph{Can active learning achieve constant label complexity in more general cases?}
\vspace{15 pt}
\end{minipage}

We provide the first affirmative answer to the above question with a \emph{general} set of regression function $\cF$ (finite), and under \emph{general} action space $\cX$ and marginal distribution $\cD_{\cX}$.
The positive result is achieved by \cref{al_abs:alg:eluder} (deferred to \cref{al_abs:app:constant_alg}), which differs from \cref{al_abs:alg:epoch} in two aspects: 
(1) we drop the epoch scheduling, and 
(2) apply a tighter elimination step derived from an optimal stopping theorem.
Another change comes from the analysis of the algorithm: Instead of analyzing with respect to the disagreement coefficient, we work with the \emph{eluder dimension} $\mfe \ldef \sup_{f^{\star} \in \cF}\mfe_{f^\star}(\cF,\gamma/2)$.\footnote{We formally define eluder dimension in \cref{al_abs:app:star_eluder}. As examples, we have $\mfe = O(d \cdot \log \frac{1}{\gamma})$ for linear functions in $\R^d$, and $\mfe = O(C_{\textsf{link}} \cdot d \log \frac{1}{\gamma})$ for generalized linear functions (where $C_{\textsf{link}}$ is a quantity related to the link function).} 
To do that, we analyze active learning from the perspective of \emph{regret minimization with selective querying} \citep{dekel2012selective}, which allows us to incorporate techniques developed in the field of contextual bandits \citep{russo2013eluder, foster2020instance}.
We defer a detailed discussion to \cref{al_abs:app:constant_regret} and provide the following guarantees. 
\begin{restatable}{theorem}{thmConstant}
\label{al_abs:thm:constant}
With probability at least $1-2\delta$, \cref{al_abs:alg:eluder} returns a classifier with expected Chow's excess error at most $\epsilon$ and label complexity $O\paren{ \frac{\mfe \cdot \log ({\abs*{\cF}}/{\delta})}{\gamma^2}}$, which is independent of $\frac{1}{\eps}$.
\end{restatable}

Based on discussion in \cref{al_abs:sec:standard_excess_error}, we can immediately translate the above results into \emph{standard} excess error guarantees under the Massart noise (with $\gamma$ replaced by $\tau_0$).
We next discuss why existing algorithms/analyses do not guarantee constant label complexity, even in the linear case.
\begin{enumerate}
	\item \textbf{Epoch scheduling.} Many algorithms proceed in epochs and aim at \emph{halving} the excess error after each epoch \citep{balcan2007margin, zhang2014beyond, puchkin2021exponential}.
	One inevitably needs $\log \frac{1}{\eps}$ epochs to achieve $\epsilon$ excess error. 
	\item \textbf{Relating to disagreement coefficient.}
	The algorithm presented in \citet{krishnamurthy2019active} does not use epoch scheduling. However, their label complexity are analyzed with disagreement coefficient, which incurs a $\sum_{t=1}^{1/\eps} \frac{1}{t} = O(\log \frac{1}{\eps})$ term in the label complexity.
\end{enumerate}

\begin{remark}
    \cref{al_abs:alg:eluder} also provides guarantees when $x$ is selected by an adaptive adversary (instead of i.i.d. sampled $x \sim \cD_\cX$). In that case, we simultaneously upper bound the regret and the label complexity (see \cref{al_abs:thm:constant_adv} in \cref{al_abs:app:constant_alg}). Our results can be viewed as a generalization of the results developed in the linear case \citep{dekel2012selective}.
\end{remark}

\subsection{Dealing with Model Misspecification}
\label{al_abs:sec:misspecified}

Our main results are developed under realizability (\cref{al_abs:asmp:predictable}), which assumes that there exists a $f^\star \in \cF$ such that $f^\star = \eta$. 
In this section, we relax that assumption and allow model misspecification.
We assume the learner is given a set of regression function $\cF: \cX \to [0, 1]$ that may only \emph{approximates} the conditional probability $\eta$. 
More specifically, we make the following assumption.

\begin{assumption}[Model misspecification]
\label{al_abs:asmp:misspecified}
There exists a $\wb f \in \cF$ such that $\wb f$ approximate  $\eta$ up to $\kappa > 0$ accuracy, i.e., $\sup_{x \in \cX} \abs*{\bar f(x) - \eta(x)} \leq \kappa$.
\end{assumption}

We use a variation of \cref{al_abs:alg:epoch} to adapt to model misspecification (\cref{al_abs:alg:mis}, deferred to \cref{al_abs:app:mis_alg}). 
Compared to \cref{al_abs:alg:epoch}, the main change in \cref{al_abs:alg:mis} is to apply a more conservative step in determining the active set $\cF_m$ at each epoch:
We maintain a larger active set of regression function to ensure that $\wb f$ is not eliminated throughout all epochs.
Our algorithm proceeds \emph{without} knowing the misspecification level $\kappa$. 
However, the excess error bound presented next holds under the condition that $\kappa \leq \eps$ (i.e., it requires that the misspecification is no larger than the desired accuracy). 
Abbreviate $\wb \theta \ldef \sup_{\iota > 0}\theta_{\wb f}^{\val}(\cF, \gamma /2, \iota)$, we achieve the following guarantees.

\begin{restatable}{theorem}{thmMis}
\label{al_abs:thm:mis}
Suppose $\kappa\leq \eps$. With probability at least $1-2\delta$, \cref{al_abs:alg:mis} returns a classifier with Chow's excess error 
$O \prn{ \eps \cdot  \wb \theta \cdot {\log\prn{\frac{ \pseud(\cF)}{\eps \, \gamma \, \delta}}}}$
and label complexity
$O\prn{ \frac{\wb \theta \, \pseud(\cF)}{\gamma^2} \cdot \log^2 \prn{\frac{\pseud(\cF)}{\eps \, \gamma}}  \cdot  {\log\prn{\frac{ \pseud(\cF)}{\eps \, \gamma \, \delta}}}}$.
\end{restatable}

We only provide guarantee when $\kappa \leq \eps$, since the learned classifier suffers from an additive $\kappa$ term in the excess error (see \cref{al_abs:app:mis_partial} for more discussion).
On the other hand, the (inefficient) algorithm provided in \citet{puchkin2021exponential} works without any assumption on the approximation error.
An interesting future direction is to study the relation between computational efficiency and learning with \emph{general} approximation error.

\section{Proofs and Supporting Results}
\label{al_abs:al_abs:sec:proofs}

\subsection{Disagreement Coefficient, Star Number and Eluder Dimension}
\label{al_abs:app:star_eluder}

We provide formal definitions/guarantees of value function disagreement coefficient, eluder dimension and star number in this section. These results are developed in \citet{foster2020instance, russo2013eluder}. 
Since our guarantees are developed in terms of these complexity measures, any future developments on these complexity measures (e.g., with respect to richer function classes) directly lead to broader applications of our algorithms.

We first state known upper bound on value function disagreement coefficient with respect to nice sets of regression functions.

\begin{proposition}[\citet{foster2020instance}]
	\label{al_abs:prop:dis_coeff_bound}
	For any $f^{\star} \in \cF$ and $\gamma,\eps > 0$, let $\theta^{\val}_{f^{\star}}(\cF,\gamma,\eps)$ be the value function disagreement coefficient defined in \cref{al_abs:def:dis_coeff}. Let $\phi: \cX \rightarrow \R^d$ be a fixed feature mapping and $\cW \subseteq \R^d$ be a fixed set. The following upper bounds hold true.
	\begin{itemize}
		\item Suppose $\cF \ldef \crl{x \mapsto \ang{\phi(x),w} : w \in \cW }$ is a set of linear functions. We then have $\sup_{f \in \cF, \gamma > 0 ,\eps > 0} \theta^{\val}_f(\cF, \gamma, \eps) \leq d$.
		\item Suppose $\cF \ldef \crl{ x \mapsto \sigma( \ang{ \phi(x), w}): w \in \cW}$ is a set of generalized linear functions with any fixed link function $\sigma: \R \rightarrow \R$ such that $0 < c_l < \sigma^{\prime} \leq c_u$.
		We then have 
		$\sup_{f \in \cF, \gamma > 0 ,\eps > 0} \theta^{\val}_f(\cF, \gamma, \eps) \leq {\frac{c_u}{c_l}}^2 \cdot d$.
	\end{itemize}
\end{proposition}

We next provide the formal definition of value function eluder dimension and star number \citep{foster2020instance, russo2013eluder}.

\begin{definition}[Value function eluder dimension]
\label{al_abs:def:eluder}
For any $f^{\star} \in \cF$ and $\gamma > 0$, let $\check{\mathfrak{e}}_{f^{\star}}(\cF, \gamma)$ be the length of the longest sequence of data points $x^{1}, \dots, x^{m}$ such that for all $i$, there exists $f^{i} \in \cF$ such that 
\begin{align*}
    \abs{ f^{i}(x^{i}) - f^{\star}(x^{i}) } > \gamma, \quad \text{ and } \quad \sum_{j < i} \paren{ f^{i}(x^{j})  - f^{\star}(x^{j})}^2 \leq \gamma^2.
\end{align*}
The value function eluder dimension is defined as $\mathfrak{e}_{f^{\star}}(\cF, \gamma_0) \coloneqq \sup_{\gamma \geq \gamma_0} \check{\mathfrak{e}}_{f^{\star}}(\cF, \gamma)$. 
\end{definition}
\begin{definition}[Value function star number]
\label{al_abs:def:star}
For any $f^{\star} \in \cF$ and $\gamma> 0$, let $\check{\mathfrak{s}}_{f^{\star}}(\cF, \gamma)$ be the length of the longest sequence of data points $x^{1}, \dots, x^{m}$ such that for all $i$, there exists $f^{i} \in \cF$ such that 
\begin{align*}
    \abs{ f^{i}(x^{i}) - f^{\star}(x^{i}) } > \gamma, \quad \text{ and } \quad \sum_{j \neq i} \paren{ f^{i}(x^{j})  - f^{\star}(x^{j})}^2 \leq \gamma^2.
\end{align*}
The value function eluder dimension is defined as $\mathfrak{s}_{f^{\star}}(\cF, \gamma_0) \coloneqq \sup_{\gamma \geq \gamma_0} \check{\mathfrak{s}}_{f^{\star}}(\cF, \gamma)$. 
\end{definition}

Since the second constrain in the definition of star number is more stringent than the counterpart in the definition of eluder dimension, one immediately have that $\mfs_{f^{\star}}(\cF, \gamma) \leq \mfe_{f^{\star}}(\cF, \gamma)$. 
We provide known upper bounds for eluder dimension next.

\begin{proposition}[\citet{russo2013eluder}]
	\label{al_abs:prop:eluder_bound}
	Let $\phi: \cX \rightarrow \R^d$ be a fixed feature mapping and $\cW \subseteq \R^d$ be a fixed set. Suppose $\sup_{x \in \cX}\nrm{\phi(x)}_2 \leq 1$ and $\sup_{w \in \cW} \nrm{w}_2 \leq 1$. The following upper bounds hold true.
	\begin{itemize}
		\item Suppose $\cF \ldef \crl{x \mapsto \ang{\phi(x),w} : w \in \cW }$ is a set of linear functions. We then have $\sup_{f^{\star} \in \cF} \mfe_{f^{\star}}(\cF, \gamma) = O(d \log \frac{1}{\gamma})$.
		\item 
		Suppose $\cF \ldef \crl{ x \mapsto \sigma( \ang{ \phi(x), w}): w \in \cW}$ is a set of generalized linear functions with any fixed link function $\sigma: \R \rightarrow \R$ such that $0 < c_l < \sigma^{\prime} \leq c_u$.
        We then have 
	$\sup_{f^{\star} \in \cF} \mfe_{f^{\star}}(\cF, \gamma) = O \prn[\big]{ \prn[\big]{\frac{c_u}{c_l}}^2 d \log \prn[\big]{ \frac{c_u}{\gamma}} }$.
	\end{itemize}
\end{proposition}

The next result shows that the disagreement coefficient (with our \cref{al_abs:def:dis_coeff}) can be always upper bounded by (squared) star number and eluder dimension.

\begin{proposition}[\citet{foster2020instance}]
    \label{al_abs:prop:eluder_star_dis}
    Suppose $\cF$ is a uniform Glivenko-Cantelli class.
    For any $f^{\star}: \cX \rightarrow [0,1]$ and $\gamma, \eps >0$,
    we have $\theta^{\val}_{f^\star}(\cF, \gamma, \epsilon) \leq 4 \paren{\mfs_{f^\star}(\cF, \gamma)}^2$, and $\theta^{\val}_{f^\star}(\cF, \gamma, \epsilon) \leq 4 \, {\mfe_{f^\star}(\cF, \gamma)}$.
\end{proposition}

The requirement that $\cF$ is a uniform Glivenko-Cantelli class is rather weak: It is satisfied as long as  $\cF$ has finite Pseudo dimension \citep{anthony2002uniform}.

In our analysis, we sometimes work with sub probability measure (due to selective sampling). Our next result shows that defining the disagreement coefficient over all (sub) probability measures will not affect its value. 
More specifically, denote $\wt \theta^{\val}_{f^{\star}}(\cF, \gamma, \eps)$ be the disagreement coefficient defined in \cref{al_abs:def:dis_coeff}, but with  $\sup$ taking over all probability and sub probability measures. We then have the following equivalence.

\begin{proposition}
	\label{al_abs:prop:sub_measure}
	Fix any $\gamma_0, \eps_0 \geq 0$. We have  $\wt \theta^{\val}_{f^{\star}} (\cF, \gamma_0, \eps_0) = \theta^{\val}_{f^{\star}}  (\cF, \gamma_0, \eps_0)$.
\end{proposition}
\begin{proof}
We clearly have $\wt \theta^{\val}_{f^{\star}} (\cF, \gamma_0, \eps_0) \geq \theta^{\val}_{f^{\star}}  (\cF, \gamma_0, \eps_0)$ by additionally considering sub probability measures. We next show the opposite direction.

Fix any sub probability measure $\wt \cD_\cX$ that is non-zero (otherwise we have 
\linebreak
  $\P_{x \sim \wt \cD_\cX} \prn{\cdot}  = 0$).
Suppose $\E_{x \sim \wt \cD_\cX} \brk{1} = \kappa < 1$. We can now consider its normalized probability measure  $\wb \cD_\cX$ such that  $\wb \cD_\cX(\omega) = \frac{\wt \cD_\cX(\omega)}{\kappa}$ (for any $\omega$ in the sigma algebra).
Now fix any $\gamma > \gamma_0$ and $\eps > \eps_0$. We have 
\begin{align*}
	& \frac{\gamma^2}{\eps^2} \cdot 
	\P_{\wt \cD_\cX} \prn*{ \exists f \in \cF: \abs{f(x) - f^{\star}(x)} > \gamma,
	\nrm*{ f - f^{\star}}_{\wt \cD_\cX}^2 \leq \eps^2} \\
	& = \frac{\gamma^2}{\eps^2 / \kappa} \cdot 
	\P_{\wb \cD_\cX} \prn*{ \exists f \in \cF: \abs{f(x) - f^{\star}(x)} > \gamma,
	\nrm*{ f - f^{\star}}_{\wb \cD_\cX}^2 \leq \eps^2 /\kappa} \\
	& = \frac{\gamma^2}{\wb \eps^2 } \cdot 
	\P_{\wb \cD_\cX} \prn*{ \exists f \in \cF: \abs{f(x) - f^{\star}(x)} > \gamma,
	\nrm*{ f - f^{\star}}_{\wb \cD_\cX}^2 \leq \wb \eps^2 } \\
	& \leq \theta^{\val}_{f^{\star}}(\cF, \gamma_0, \eps_0),
\end{align*}
where we denote $\wb \eps \ldef \frac{\eps}{\sqrt{\kappa}} > \eps$, and the last follows from the fact that $\wb \cD_\cX$ is a probability measure. We then have 
$\wt \theta^{\val}_{f^{\star}} (\cF, \gamma_0, \eps_0) \leq \theta^{\val}_{f^{\star}}  (\cF, \gamma_0, \eps_0)$, and thus the desired result.
\end{proof}

\subsection{Concentration Results}
\label{al_abs:app:concentration}

\begin{lemma}[Freedman's inequality, \citep{freedman1975tail, agarwal2014taming}]
    \label{al_abs:lm:freedman}
    Let $(Z_t)_{t \leq T}$ be a real-valued martingale difference sequence adapted to a filtration $\mfF_t$, and let $\E_t \sq{\cdot} \ldef \E \sq{\cdot \mid \mfF_{t-1}}$. If $\abs{Z_t} \leq B$ almost surely, then for any $\eta \in (0,1/B)$ it holds with probability at least $1 - \delta$,
    \begin{align*}
        \sum_{t=1}^{T} Z_t \leq \eta \sum_{t=1}^{T} \E_{t} \sq{Z_t^2} + \frac{\log \delta^{-1}}{\eta}.
    \end{align*}
\end{lemma}

\begin{lemma}[\citep{foster2020instance}]
   \label{al_abs:lm:martingale_two_sides} 
   Let $(X_t)_{t \leq T}$ be a sequence of random variables adapted to a filtration $\mfF_t$. If $0 \leq {X_t} \leq B$ almost surely, then with probability at least $1-\delta$,
   \begin{align*}
       \sum_{t =1 }^T X_t \leq \frac{3}{2} \sum_{t=1}^T \E_{t}\sq{X_t} + 4B \log(2 \delta^{-1}),
   \end{align*}
   and 
   \begin{align*}
       \sum_{t =1 }^T \E_{t} \sq{X_t} \leq 2 \sum_{t=1}^T X_t + 8B \log(2 \delta^{-1}).
   \end{align*}
\end{lemma}
\begin{proof}
  These two inequalities are obtained by applying \cref{al_abs:lm:freedman} to $\prn{X_t - \E_t \sq{X_t}}_{t \leq T}$ and $\prn{\E_t \sq{X_t} - X_t}_{t \leq T}$, with $\eta = 1/2B$ and $\delta / 2$.	
  Note that $\E_t \sq{\prn{X_t - \E_t \sq{X_t}}^2} \leq \E_t \sq{X_t^2} \leq B \E_t\sq{X_t}$ if $0 \leq X_t \leq B$.
\end{proof}

We recall the definition of the Pseudo dimension of $\cF$.
\begin{definition}[Pseudo Dimension, \citet{pollard1984convergence, haussler1989decision, haussler1995sphere}]
\label{al_abs:def:pseudo_d}
Consider a set of real-valued function $\cF: \cX \rightarrow \R$. The pseudo-dimension $\pdim(\cF)$ of $\cF$ is defined as the VC dimension of the set of threshold functions 
$\crl{(x,\zeta) \mapsto \ind(f(x) > \zeta) : f \in \cF}$.
\end{definition}
We next provide concentration results with respect to a general set of regression function $\cF$ with finite Pseudo dimension.
We define/recall some notations.
Fix any epoch $m \in [M]$ and any time step $t$ within epoch $m$.
For any $f \in \cF$, we denote $M_t(f) \ldef Q_t \prn{ \prn{f(x_t) - y_t}^2 - \prn{f^\star(x_t) - y_t}^2}$, 
and $\wh R_m(f) \ldef \sum_{t=1}^{\tau_{m-1}} Q_t \prn{f(x_t) - y_t}^2$.
Recall that we have $Q_t = g_m(x_t)$.
We define filtration $\mfF_t \ldef \sigma \prn{ \prn{x_1, y_1}, \ldots , \prn{x_{t}, y_{t}}}$,\footnote{$y_t$ is not observed (and thus not included in the filtration) when $Q_t = 0$. Note that $Q_t$ is measurable with respect to $\sigma( (\mfF_{t-1}, x_t) ) $.} 
and denote $\E_t \sq{\cdot } \ldef \E \sq{\cdot \mid \mfF_{t-1}}$.
\begin{lemma}[\citet{krishnamurthy2019active}]
    \label{al_abs:lm:expected_sq_loss_pseudo}
    Suppose $\pseud(\cF) < \infty$.
   Fix any $\delta \in (0,1)$.  For any $\tau, \tau^\prime \in [T]$ such that $\tau < \tau^\prime$, with probability at least $1 - \delta $, we have 
   \begin{align*}
   	\sum_{t = \tau}^{\tau^\prime} M_t(f) \leq \sum_{t=\tau}^{\tau^\prime} \frac{3}{2} \E_t \brk{M_t(f)} + 
	C_\delta(\cF),
   \end{align*}
   and
   \begin{align*}
       \sum_{t = \tau}^{\tau^\prime} \E_t \sq{ M_t(f)}  \leq 2 \sum_{t = \tau}^{\tau^\prime} M_t(f) + C_\delta(\cF),
   \end{align*}
   where $C_\delta(\cF)  = C \cdot  \prn*{ \pdim(\cF) \cdot \log T +   \log \prn*{ \frac{\pdim(\cF)  \cdot  T}{\delta}} }  \leq C^{\prime} \cdot \prn*{\pseud(\cF) \cdot \log \prn*{\frac{T}{\delta}}}$, where $C, C^{\prime} >0$ are universal constants.
\end{lemma}

\subsection{Proofs and Supporting Results for \cref{al_abs:sec:epoch}}
\label{al_abs:app:epoch}
We give the proof of \cref{al_abs:thm:epoch} and \cref{al_abs:thm:epoch_efficient}.
Supporting lemmas used in the proofs are deferred to \cref{al_abs:app:epoch_lms}.

Fix any classifier $\wh h: \cX \rightarrow \crl{0, 1,\bot}$. For any $x\in\cX$, we introduce the notion
\begin{align}
& 	\exc_{\gamma}( \wh h;x) \ldef \nonumber\\
    &  \P_{y\mid x} \prn[\big]{y \neq \widehat h(x)} \cdot \ind \prn[\big]{ \widehat h(x) \neq \bot} + \prn[\big]{{1}/{2} - \gamma} \cdot \ind \prn[\big]{\widehat h(x) = \bot} - \P_{y\mid x} \prn[\big]{ y \neq h^\star(x) }\nonumber\\
    & = \ind \prn[\big]{ \widehat h(x) \neq \bot} \cdot \prn[\big]{\P_{y\mid x} \prn[\big]{y \neq \widehat h(x)} -  \P_{y\mid x} \prn[\big]{ y \neq h^\star(x) }} \nonumber \\
    & \quad + \ind \prn[\big]{ \widehat h(x) = \bot} \cdot \prn[\big]{ \prn[\big]{{1}/{2} - \gamma}  -  \P_{y\mid x} \prn[\big]{ y \neq h^\star(x) }} \label{al_abs:eq:excess_x}
\end{align}
to represent the excess error of $\wh h$ at point $x\in \cX$. Excess error of classifier $\wh h$ can be then written as $\exc_{\gamma}(\wh h) \ldef \err_\gamma(\wh h) - \err(h^{\star}) = \E_{x \sim \cD_\cX} \brk{ \exc_{\gamma}(\wh h;x)}$.

\thmEpoch*
\begin{proof}
	We analyze under the good event $\cE$ defined in \cref{al_abs:lm:expected_sq_loss_pseudo}, which holds with probability at least $1-{\delta}$. Note that all supporting lemmas stated in \cref{al_abs:app:epoch_lms} hold true under this event.

We analyze the Chow's excess error of $\wh h_m$, which is measurable with respect to $\mfF_{\tau_{m-1}}$. 
For any $x \in \cX$, if $g_m(x) = 0$, 
\cref{al_abs:lm:regret_no_query} implies that $\exc_{\gamma}(\wh h_m ;x) \leq 0$. 
If $g_m(x)= 1$, we know that $\wh h_m(x) \neq \bot$ and $\frac{1}{2} \in (\lcb(x;\cF_m),\ucb(x;\cF_m))$. 
Note that $\wh h_m(x) \neq h^{\star}(x)$ only if $\ind(f^{\star}(x) \geq 1/2) \neq \ind(\wh f_m(x) \geq 1/2)$.
Since $f^{\star}, \wh f_m \in \cF_m$ by \cref{al_abs:lm:set_f}. 
The error incurred in this case can be upper bounded by $2 \abs{ f^{\star}(x)- 1 /2} \leq 2 w(x;\cF_m)$, which results in $\exc_{\gamma}(\wh h_m; x) \leq 2 w(x;\cF_m)$. 
Combining these two cases together, we have 
\begin{align*}
	\exc_{\gamma}( \wh h_m) \leq 2 \E_{x \sim \cD_\cX} \brk{ \ind(g_m(x) = 1) \cdot w(x;\cF_m)}.	
\end{align*}
Take $m=M$ and apply \cref{al_abs:lm:per_round_regret_dis_coeff}, with notation $\rho_m \ldef 2 \beta_m + C_\delta$, leads to the following guarantee.
\begin{align*}
	\exc_{\gamma}( \wh h_M)
	& \leq  { \frac{ 8 \rho_M}{\tau_{M-1} \gamma} \cdot \theta^{\val}_{ f^{\star}}\prn*{\cF, \gamma/2, \sqrt{\rho_M/2 \tau_{M-1}}}}\\
	& =  O \prn*{ \frac{ \pseud(\cF) \cdot \log ( T / \delta)}{T \, \gamma} \cdot \theta^{\val}_{ f^{\star}}\prn*{\cF, \gamma/2, \sqrt{C_\delta/T}}},
\end{align*}
where we use the fact that $\frac{T}{2} \leq \tau_{M-1} \leq T$ and definitions of $\beta_m$ and $C_\delta$.
Simply considering $\theta \ldef \sup_{f^{\star} \in \cF, \iota > 0}\theta^{\val}_{f^{\star}} (\cF, \gamma / 2, \iota)$ as an upper bound of $\theta^{\val}_{f^{\star}} (\cF, \gamma / 2, \sqrt{C_\delta / T})$ 
and taking 
\begin{align*}
	T = O \prn*{ \frac{\theta \, \pseud(\cF) }{\eps \, \gamma} \cdot \log \prn*{ \frac{\theta \, \pseud(\cF) }{\eps \, \gamma \, \delta}} }	
\end{align*}
ensures that $\exc_{\gamma}(\wh h_M) \leq \eps$.

We now analyze the label complexity (note that the sampling process of \cref{al_abs:alg:epoch} stops at time $t = \tau_{M-1}$).
Note that $\E \brk{\ind(Q_t = 1) \mid \mfF_{t-1}} = \E_{x\sim\cD_\cX} \brk{ \ind(g_m(x) = 1) }$ for any epoch $m \geq 2$ and time step $t$ within epoch $m$. 
Combining \cref{al_abs:lm:martingale_two_sides} with \cref{al_abs:lm:conf_width_dis_coeff} leads to
    \begin{align*}
      & \sum_{t=1}^{\tau_{M-1}} \ind(Q_t = 1) \\
        & \leq \frac{3}{2} \sum_{t=1}^{\tau_{M-1}} \E \sq{\ind(Q_t = 1) \mid \mfF_{t-1}} + 4 \log \delta^{-1}\\
        & \leq 3 + \frac{3}{2}\sum_{m=2}^{M-1}\frac{(\tau_m - \tau_{m-1}) \cdot 4 \rho_m}{{\tau_{m-1}} \gamma^2} \cdot \theta^{\val}_{f^{\star}}\prn*{\cF, \gamma/2, \sqrt{\rho_m/2\tau_{m-1}}}  + 4 \log \delta^{-1} \\
        & \leq 3 + 6 \sum_{m=2}^{M-1}\frac{\rho_m}{ \gamma^2} \cdot \theta^{\val}_{f^{\star}}\prn*{\cF, \gamma/2, \sqrt{\rho_m/2\tau_{m-1}}}  + 4 \log \delta^{-1} \\
	& \leq 3 + 4 \log \delta^{-1} + \frac{18 \log T \cdot M \cdot C_\delta }{\gamma^2}
	\cdot \theta^{\val}_{f^{\star}}\prn*{\cF, \gamma/2, \sqrt{C_\delta/T }} \\ 
	& = O \prn*{ \frac{\theta \, \pseud(\cF) }{\gamma^2} \cdot \prn*{\log \prn*{ \frac{\theta \, \pseud(\cF) }{\eps \, \gamma}}}^{2} \cdot 
	\log \prn*{ \frac{\theta \,\pseud(\cF) }{\eps \, \gamma \, \delta}}} ,
    \end{align*}
    with probability at least $1-2\delta$ (due to an additional application of \cref{al_abs:lm:martingale_two_sides}); where we plug the above choice of $T$ and upper bound other terms as before.
\end{proof}

\paragraph{A slightly different guarantee for \cref{al_abs:alg:epoch}}
The stated \cref{al_abs:alg:epoch} takes $\theta \ldef \sup_{f^{\star} \in \cF, \iota > 0}\theta^{\val}_{f^{\star}} (\cF, \gamma / 2, \iota)$ as an input (the value of $\theta$ can be upper bounded for many function class $\cF$, as discussed in \cref{al_abs:app:star_eluder}).
However, we don't necessarily need to take $\theta$ as an input to the algorithm. Indeed, 
we can simply run a modified version of \cref{al_abs:alg:epoch} with $T = \frac{\pseud(\cF)}{\eps \, \gamma}$.
Following similar analyses in proof of \cref{al_abs:thm:epoch}, set $\iota \ldef \sqrt{C_\delta /T} \propto \sqrt{\gamma \eps}$, the modified version achieves excess error 
\begin{align*}
	\exc_{\gamma}(\wh h_M) = O \prn*{\eps \cdot \theta^{\val}_{f^{\star}}(\cF, \gamma /2, \iota) \cdot \log \prn*{\frac{\pseud(\cF)}{\eps \, \delta \, \gamma}}}
\end{align*}
with label complexity 
\begin{align*}
	 O \prn*{ \frac{\theta^{\val}_{f^{\star}}(\cF, \gamma /2, \iota) \cdot  \pseud(\cF) }{\gamma^2} \cdot \prn*{\log \prn*{ \frac{ \pseud(\cF) }{\eps \, \gamma}}}^{2} \cdot 
	\log \prn*{ \frac{\pseud(\cF) }{\eps \, \gamma \, \delta}}} .
\end{align*}

We now discuss the efficient implementation of \cref{al_abs:alg:epoch} and its computational complexity. 
We first state some known results in computing the confidence intervals with respect to a set of regression functions $\cF$.

\begin{proposition}[\citet{krishnamurthy2017active, foster2018practical, foster2020instance}] \label{al_abs:prop:CI_oracle}
Consider the setting studied in \cref{al_abs:alg:epoch}. 
Fix any epoch $m \in [M]$ and denote $\cB_m \ldef \crl{ (x_t,Q_t, y_t)}_{t=1}^{\tau_{m-1}}$.
Fix any $\alpha > 0$.
For any data point $x \in \cX$, there exists algorithms $\AlgLcb$ and $\AlgUcb$ that certify
\begin{align*}
    & \lcb(x;\cF_m) - \alpha \leq \AlgLcb(x;\cB_m,\beta_m,\alpha) \leq \lcb(x;\cF_m) \quad \text{and}\\
    &\ucb(x;\cF_m) \leq \AlgUcb(x;\cB_m,\beta_m,\alpha) \leq \ucb(x;\cF_m) + \alpha.
\end{align*}
The algorithms take 
 $O(\frac{1}{\alpha^2} \log \frac{1}{\alpha})$ calls of the regression oracle for general $\cF$
 and take $O(\log \frac{1}{\alpha})$ calls of the regression oracle if $\cF$ is convex and closed under pointwise convergence.
\end{proposition}
\begin{proof}
See Algorithm 2 in \citet{krishnamurthy2017active} for the general case; and Algorithm 3 in \citet{foster2018practical} for the case when $\cF$ is convex and closed under pointwise convergence.
\end{proof}

We next discuss the computational efficiency of \cref{al_abs:alg:epoch}. Recall that we redefine $\theta \ldef \sup_{f^{\star} \in \cF, \iota > 0} \theta^{\val}_{f^{\star}}(\cF,\gamma / 4, \iota)$ in the \cref{al_abs:thm:epoch_efficient} to account to approximation error.
\thmEpochEfficient*
\begin{proof}
Fix any epoch $m \in [M]$.
Denote $\wb \alpha \ldef \frac{\gamma}{4M}$ and $\alpha_m \ldef \frac{(M-m) \gamma}{4M}$.
With any observed $x \in \cX$, we construct the approximated confidence intervals $\wh \lcb(x;\cF_m)$ and 
$\wh \ucb(x; \cF_m)$ as follows.
\begin{align*}
&	\wh \lcb(x;\cF_m) \ldef \AlgLcb(x;\cB_m,\beta_m,\wb \alpha) - \alpha_m \quad \text{and}\\	
    & \wh \ucb(x;\cF_m) \ldef\AlgUcb(x;\cB_m,\beta_m,\wb \alpha)+ \alpha_m. 
\end{align*}
For efficient implementation of \cref{al_abs:alg:epoch}, we replace $\lcb(x;\cF_m)$ and $\ucb(x;\cF_m)$ with $\wh \lcb(x;\cF_m)$ and $\wh \ucb(x;\cF_m)$ in the construction of $\wh h_m$ and $g_m$.

Based on \cref{al_abs:prop:CI_oracle}, we know that 
\begin{align*}
    & \lcb(x;\cF_m) - \alpha_m - \wb \alpha \leq 	\wh \lcb(x;\cF_m) \leq \lcb(x;\cF_m) - \alpha_m \quad \text{and}\\
    &\ucb(x;\cF_m) + \alpha_m \leq \wh \ucb(x;\cF_m) \leq \ucb(x;\cF_m) + \alpha_m + \wb \alpha .
\end{align*}
Since $\alpha_m + \wb \alpha  \leq \frac{\gamma}{4}$ for any $m \in [M]$, the guarantee in \cref{al_abs:lm:query_implies_width} can be modified as $g_m(x)= 1 \implies w(x;\cF_m)\geq \frac{\gamma}{2}$. 

Fix any $m \geq 2$. Since $\cF_{m} \subseteq \cF_{m-1}$ by \cref{al_abs:lm:set_f}, we have 
\begin{align*}
	& \wh \lcb(x;\cF_m) \geq \lcb(x;\cF_m) -  \alpha_m - \wb \alpha \geq \lcb(x;\cF_{m-1}) - \alpha_{m-1} \geq \wh\lcb (x;\cF_{m-1}) \quad \text{and} \\
	& \wh \ucb(x;\cF_m) \leq \ucb(x;\cF_m) +  \alpha_m + \wb \alpha \leq \ucb(x;\cF_{m-1}) + \alpha_{m-1} \leq \wh\ucb (x;\cF_{m-1}).
\end{align*}
These ensure $\ind(g_m(x) = 1) \leq \ind(g_{m-1}(x)=1)$. Thus, the guarantees stated in \cref{al_abs:lm:conf_width_dis_coeff} and \cref{al_abs:lm:per_round_regret_dis_coeff} still hold (with $\frac{\gamma}{2}$ replaced by $\frac{\gamma}{4}$ due to modification of \cref{al_abs:lm:query_implies_width}).
The guarantee stated in \cref{al_abs:lm:regret_no_query} also hold since  $\wh \lcb(x;\cF_m) \leq \lcb(x;\cF_m)$ and $\wh \ucb(x;\cF_m) \geq \ucb(x;\cF_m)$ by construction. As a result, the guarantees stated in \cref{al_abs:thm:epoch} hold true with changes only in constant terms.

We now discuss the computational complexity of the efficient implementation. 
At the beginning of each epoch $m$. We use one oracle call to compute $\widehat f_m = \argmin_{f \in \cF} \sum_{t =1}^{ \tau_{m-1}} Q_t \paren{f(x_t) - y_t}^2 $. 
The main computational cost comes from computing $\wh \lcb$ and $\wh \ucb$ at each time step.
We take $\alpha = \wb \alpha \ldef \frac{\gamma}{4M}$ into \cref{al_abs:prop:CI_oracle}, which leads to 
$O \prn{ \frac{(\log T)^2}{\gamma^2}\cdot \log \prn{ \frac{\log T}{\gamma}}}$ calls of the regression oracle for general $\cF$ and 
$O \prn{ \log \prn{ \frac{\log T}{\gamma}}}$ calls of the regression oracle for any convex $\cF$ that is closed under pointwise convergence. This also serves as the per-example inference time for $\wh h_{M}$. The total computational cost of \cref{al_abs:alg:epoch} is then derived by multiplying the per-round cost by $T$ and plugging $T = \wt O( \frac{\theta \, \pseud(\cF)}{\eps \, \gamma})$ into the bound (for any parameter, we only keep $\poly$ factors in the total computational cost and keep $\poly$ or $\polylog$ dependence in the per-example computational cost).
\end{proof}

\subsubsection{Supporting Lemmas}
\label{al_abs:app:epoch_lms}
We use $\cE$ to denote the good event considered in \cref{al_abs:lm:expected_sq_loss_pseudo}, and analyze under this event in this section. We abbreviate $C_\delta \ldef C_\delta(\cF)$ in the following analysis.

\begin{lemma}
\label{al_abs:lm:set_f}
The followings hold true:
\begin{enumerate}
	\item $f^\star \in \cF_m$ for any $m \in [M]$.
	\item $\sum_{t=1}^{\tau_{m-1}} \E_t \brk{M_t(f)} \leq 2 \beta_m +  C_\delta$ for any $f \in \cF_m$. 
	\item $\cF_{m+1} \subseteq \cF_m$ for any $m \in [M-1]$.
\end{enumerate}
\end{lemma}
\begin{proof}
\begin{enumerate}
	\item 
	Fix any epoch $m \in [M]$ and time step $t$ within epoch $m$.
	Since $ \E [y_t] = f^\star(x_t)$, we have $\E_t \brk{ M_t(f)}  = \E \brk{ Q_t \prn{f(x) - f^\star(x) }^2 }=\E \brk{ g_m(x)\prn{f(x) - f^\star(x) }^2 } \geq 0$ for any $f \in \cF$.
By \cref{al_abs:lm:expected_sq_loss_pseudo}, we then have $\wh R_m (f^\star) \leq \wh R_m(f) + C_\delta /2 \leq \wh R_m(f) + \beta_m$ for any $f \in \cF$.
The elimination rule in \cref{al_abs:alg:eluder} then implies that $f^\star \in \cF_m$ for any $m \in [M]$.
\item  Fix any $f \in \cF_m$. With \cref{al_abs:lm:expected_sq_loss_pseudo}, we have 
	\begin{align*}
		\sum_{t=1}^{\tau_{m-1}} \E_t [M_t(f)] & \leq 2 \sum_{t=1}^{\tau_{m-1}} M_t(f) + C_\delta \\
			& = 2 \wh R_{m }(f) - 2\wh R_{m}(f^{\star}) + C_\delta \\
			& \leq 2 \wh R_{m}(f) - 2\wh R_{m} ( \wh f_m) + C_\delta\\
			& \leq 2 \beta_m + C_\delta , 
	\end{align*}
	where the third line comes from the fact that $\wh f_m$ is the minimizer of $\wh R_{m} (\cdot) $; and the last line comes from the fact that $f \in \cF_m$.
	\item Fix any $f \in \cF_{m+1}$. We have 
	\begin{align*}
		\wh R_{m} (f) - \wh R_{m} (\wh f_m) & \leq   
		\wh R_{m} (f) - \wh R_{m} (f^{\star}) + \frac{C_\delta}{2}\\
		& = \wh R_{m+1}(f) - \wh R_{m+1}(f^{\star}) 
		- \sum_{t=\tau_{m-1}+1}^{\tau_{m}} M_t(f) + \frac{C_\delta}{2}\\
		& \leq \wh R_{m+1} (f) - \wh R_{m+1} (\wh f_{m+1}) 
		- \sum_{t=\tau_{m-1}+1}^{\tau_{m}} \E_t [M_t(f)] /2 + {C_\delta}\\
		& \leq \beta_{m+1} + C_\delta \\
		& = \beta_m,
	\end{align*}	
	where the first line comes from \cref{al_abs:lm:expected_sq_loss_pseudo}; the third line comes from the fact that $\wh f_{m+1}$ is the minimizer with respect to $\wh R_{m+1}$ and \cref{al_abs:lm:expected_sq_loss_pseudo}; the last line comes from the definition of $\beta_m$.
\end{enumerate}
\end{proof}
\begin{lemma}
\label{al_abs:lm:query_implies_width}
For any $m \in [M]$, we have $g_m(x)= 1 \implies w(x;\cF_m) > \gamma$.
\end{lemma}
\begin{proof}
We only need to show that $\ucb(x;\cF_m) - \lcb(x;\cF_m) \leq \gamma \implies g_m(x) = 0$. Suppose otherwise $g_m(x) = 1$, which implies that both 
\begin{align}
\label{al_abs:eq:query_condition}
\frac{1}{2} \in \prn*{\lcb(x;\cF_m), \ucb(x;\cF_m)}  \quad \text{ and } \quad {\brk*{\lcb(x;\cF_m), \ucb(x;\cF_m)} \nsubseteq \brk*{  \frac{1}{2}-\gamma, \frac{1}{2} +\gamma } } .
\end{align}
If $\frac{1}{2} \in (\lcb(x;\cF_m), \ucb(x;\cF_m))$ and $\ucb(x;\cF_m) - \lcb(x;\cF_m) \leq \gamma$, we must have $\lcb(x;\cF_m) \geq  \frac{1}{2}- \gamma$ and $\ucb(x;\cF_m) \leq \frac{1}{2} + \gamma$, which contradicts with \cref{al_abs:eq:query_condition}.
\end{proof}

We introduce more notations.
Fix any $m \in [M]$.
We use $n_m \ldef \tau_m - \tau_{m-1}$ to denote the length of epoch $m$,
and use abbreviation $\rho_m \ldef 2 \beta_m + C_\delta$.
    Denote $\prn{\cX, \Sigma, \cD_\cX}$ as the (marginal) probability space,
    and denote $\wb \cX_m \ldef \crl{x \in \cX: g_m(x) = 1} \in \Sigma$ be the region where query \emph{is} requested within epoch $m$.
    Since we have $\cF_{m+1} \subseteq \cF_m$ by \cref{al_abs:lm:set_f}, we clearly have $\wb \cX_{m+1} \subseteq \wb \cX_m$.
    We now define a sub probability measure $\wb \mu_m \ldef ({\cD_\cX})_{\mid \wb \cX_m}$ such that $\wb \mu_m(\omega) = \cD_{\cX}\prn{ \omega \cap \wb \cX_m}$ for any $\omega \in \Sigma$. 
    Fix any time step $t$ within epoch $m$ and any $\wb m \leq m$.
    Consider any measurable function $F$ (that is $\cD_\cX$ integrable), we have 
    \begin{align}
    	\E_{x \sim \cD_\cX} \brk*{ \ind(g_m(x) = 1) \cdot F(x)}
	& = \int_{x \in \wb \cX_{m}} F(x) \, d \cD_\cX(x) \nonumber \\ 
	& \leq \int_{x \in \wb \cX_{\wb m}} F(x) \, d \cD_\cX(x)\nonumber \\ 
	& = \int_{x \in \cX} F(x) \, d \wb \mu_{\wb m} (x) \nonumber \\ 
	& \rdef \E_{x \sim \wb \mu_{\wb m}} \brk*{ F(x)}, \label{al_abs:eq:change_of_measure} 
    \end{align}
    where, by a slightly abuse of notations, we use $\E_{x \sim \mu} \sq{\cdot}$ to denote the integration with any sub probability measure $\mu$. 
    In particular, \cref{al_abs:eq:change_of_measure} holds with equality when $\wb m = m$.
\begin{lemma}
    \label{al_abs:lm:conf_width_dis_coeff}
 Fix any epoch $m \geq 2$. 
 We have
	\begin{align*}
    \E_{x \sim \cD_\cX} \sq{\ind (g_m(x) = 1)} 
	 \leq  \frac{4 \rho_m}{{\tau_{m-1}} \gamma^2} \cdot \theta^{\val}_{f^{\star}}\prn*{\cF, \gamma/2, \sqrt{\rho_m/2\tau_{m-1}}}.
	\end{align*}
\end{lemma}
\begin{proof}
We know that $\ind(g_m(x) = 1) = \ind (g_m(x)= 1) \cdot \ind(w(x;\cF_m) > \gamma )$ from \cref{al_abs:lm:query_implies_width}. Thus, for any $\wb m \leq m$, we have 
    \begin{align}
	    \E_{x \sim \cD_\cX} \sq{\ind(g_m(x)= 1)} 
	    &  = \E_{x \sim \cD_\cX} \sq{\ind(g_m(x)= 1) \cdot \ind(w(x;\cF_m)> \gamma )}\nonumber \\ 
	    & \leq \E_{x \sim \wb \mu_{\wb m} } \sq{\ind(w(x;\cF_m)> \gamma  )}\nonumber \\
    & \leq \E_{x \sim \wb \mu_{\wb m}} \prn[\Big]{ \ind \prn[\big]{\exists f \in \cF_m, \abs*{f(x) - f^{\star}(x)} > \gamma/2}} , \label{al_abs:eq:conf_width_dis_coeff_1}
    \end{align}
where the second line uses \cref{al_abs:eq:change_of_measure} and the last line comes from the facts that
    $f^{\star} \in \cF_m$ and $w(x;\cF_m) > \gamma  \implies \exists f \in \cF_m, \abs{f(x) - f^{\star}(x)} > {\gamma}/ {2}$. 

     For any time step $t$, let  $m(t)$ denote the epoch where  $t$ belongs to.
From \cref{al_abs:lm:set_f}, we know that, $\forall f \in \cF_m$,  
    \begin{align}
\rho_m &
\geq \sum_{t=1}^{\tau_{m -1}} \E_{t} \sq[\Big]{ Q_t \prn[\big]{f(x_t) - f^{\star}(x_t)}^2} \nonumber \\
       &  = \sum_{t=1}^{\tau_{m -1}} \E_{x \sim \cD_\cX} \sq[\Big]{\ind(g_{m(t)}(x)=1) \cdot \prn[\big]{f(x) - f^{\star}(x)}^2} \nonumber \\
	  & = \sum_{\wb m=1}^{m-1} n_{\wb m} \cdot \E_{x \sim \wb \mu_{\wb m}} \brk*{  \prn*{f(x) - f^{\star}(x)}^2}\nonumber \\
	& = \tau_{m-1} \E_{x \sim \wb \nu_m} \brk*{  \prn*{f(x) - f^{\star}(x)}^2}, \label{al_abs:eq:conf_width_dis_coeff_2}
    \end{align}
    where we use $Q_t = g_{m(t)}(x_t) = \ind(g_{m(t)}(x) = 1)$ and \cref{al_abs:eq:change_of_measure} on the second line, and define a new sub probability measure 
    $$\wb \nu_m \ldef \frac{1}{\tau_{m-1}} \sum_{\wb m =1}^{m-1} n_{\wb m} \cdot \wb \mu_{\wb m}$$ on the third line. 

    Plugging \cref{al_abs:eq:conf_width_dis_coeff_2} into \cref{al_abs:eq:conf_width_dis_coeff_1} leads to the bound 
    \begin{align*}
	& \E_{x \sim \cD_\cX} \sq{\ind(g_m(x)= 1)} \\
	& \leq \E_{x \sim \wb \nu_{m}} \sq[\bigg]{\ind \prn[\Big]{\exists f \in \cF, \abs[\big]{f(x) - f^{\star}(x)} > \gamma/2, \E_{x \sim \wb \nu_m} \sq[\Big]{\prn[\big]{f(x) - f^{\star}(x)}^2} \leq \frac{\rho_m}{\tau_{m-1}}}},
    \end{align*}
    where we use the definition of $\wb \nu_m$ again (note that \cref{al_abs:eq:conf_width_dis_coeff_1} works with any $\wb m \leq m$).  
    Combining the above result with the discussion around \cref{al_abs:prop:sub_measure} and \cref{al_abs:def:dis_coeff}, we then have 
    \begin{align*}
	 \E_{x \sim \cD_\cX} \sq{\ind(g_m(x)= 1)}  
	 \leq  \frac{4 \rho_m}{{\tau_{m-1}} \, \gamma^2} \cdot \theta^{\val}_{f^{\star}}\prn*{\cF, \gamma/2, \sqrt{\rho_m/2\tau_{m-1}}}.
    \end{align*}
\end{proof}

\begin{lemma}
    \label{al_abs:lm:per_round_regret_dis_coeff}
Fix any epoch $m\geq 2$. We have 
    \begin{align*}
    	\E_{x \sim \cD_\cX} \sq{\ind(g_m(x)= 1)\cdot w(x;\cF_m)} 
 \leq { \frac{4 \rho_m}{\tau_{m-1}\,  \gamma} \cdot \theta^{\val}_{f^{\star}}\prn*{\cF, \gamma/2, \sqrt{\rho_m/2\tau_{m-1}}}}.
    \end{align*}
\end{lemma}
\begin{proof}
    Similar to the proof of \cref{al_abs:lm:conf_width_dis_coeff}, we have 
    \begin{align*}
      & \E_{x \sim \cD_\cX} \sq{\ind(g_m(x)= 1)\cdot w(x;\cF_m)} \\
	& = \E_{x \sim \cD_\cX} \sq{\ind(g_m(x)=1) \cdot \ind(w(x;\cF_m)> \gamma )\cdot w(x;\cF_m)} \\
	& \leq \E_{x \sim \wb \mu_{\wb m}} \sq{\ind(w(x;\cF_m)> \gamma )\cdot w(x;\cF_m)}
    \end{align*}
    for any $\wb m \leq m$. 
With $\wb \nu_m = \frac{1}{\tau_{m-1}} \sum_{\wb m =1}^{m-1} n_{\wb m} \cdot \wb \mu_{\wb m}$, we then have 
\begin{align*}
	 & \E_{x \sim \cD_\cX} \sq{\ind(g_m(x)= 1)\cdot w(x;\cF_m)} \\
	& \leq \E_{x \sim \wb \nu_{m}} \sq{\ind(w(x;\cF_m)> \gamma )\cdot w(x;\cF_m)}\\
        & \leq \E_{x \sim \wb \nu_{m}} \sq*{\ind(\exists f \in \cF_m, \abs[\big]{f(x) - f^{\star}(x)} > \gamma/2)\cdot \prn*{\sup_{f , f^\prime \in \cF_m} \abs*{f(x) - f^\prime(x)}}} \\
        & \leq 2\E_{x \sim \wb \nu_{m}} \sq*{\ind(\exists f \in \cF_m, \abs[\big]{f(x) - f^{\star}(x)} > \gamma/2)\cdot \prn*{\sup_{f \in \cF_m} \abs{f(x) - f^{\star}(x)}}} \\
        & \leq 2 \int_{\gamma/2}^1 \E_{x \sim \wb \nu_{m}} \sq*{\ind \prn*{\sup_{f \in \cF_m} \abs[\big]{f(x) - f^{\star}(x)} \geq \omega}} \, d \, \omega \\
& \leq  2 \int_{\gamma/2}^1  \frac{1}{\omega^2} \, d \, \omega \cdot \prn*{ \frac{\rho_m}{\tau_{m-1}} \cdot \theta^{\val}_{f^{\star}}\prn*{\cF, \gamma/2, \sqrt{\rho_m/2\tau_{m-1}}}}\\
& \leq { \frac{4 \rho_m}{\tau_{m-1} \, \gamma} \cdot \theta^{\val}_{f^{\star}}\prn*{\cF, \gamma/2, \sqrt{\rho_m/2\tau_{m-1}}}},
\end{align*}
where we use similar steps as in the proof of \cref{al_abs:lm:conf_width_dis_coeff}.
\end{proof}
\begin{lemma}
\label{al_abs:lm:regret_no_query}
Fix any $m \in [M]$. We have $\exc_{\gamma}(\wh h_m ;x) \leq 0$ if $g_m(x) = 0$.
\end{lemma}
\begin{proof}
	Recall that
\begin{align*}
	\exc_{\gamma}( \wh h;x) & =  \nonumber
      \ind \prn[\big]{ \widehat h(x) \neq \bot} \cdot \prn[\big]{\P_{y \mid x} \prn[\big]{y \neq \widehat h(x)} -  \P_{y\mid x} \prn[\big]{ y \neq h^\star(x) }} \nonumber \\
    & \quad + \ind \prn[\big]{ \widehat h(x) = \bot} \cdot \prn[\big]{ \prn[\big]{{1}/{2} - \gamma}  -  \P_{y \mid x} \prn[\big]{ y \neq h^\star(x) }} .
\end{align*}
We now analyze the event $\curly*{g_m(x)= 0}$ in two cases. 

\textbf{Case 1: ${\widehat h_m(x) = \bot} $.} 

Since $\eta(x) = f^{\star}(x) \in [\lcb(x;\cF_m), \ucb(x;\cF_m)]$, we know that $\eta(x) \in \sq{ \frac{1}{2} - \gamma, \frac{1}{2} + \gamma }$ and thus $\P_{y \mid x} \prn[\big]{ y\neq h^\star(x) } \geq \frac{1}{2} - \gamma$. 
As a result, we have $\exc_{\gamma}(\wh h_m;x) \leq 0$.

\textbf{Case 2: ${\widehat h_m(x) \neq \bot}$ but ${\frac{1}{2} \notin (\lcb(x;\cF_m), \ucb(x;\cF_m))} $.} 

In this case, we know that $\widehat h_m (x) = h^\star(x)$ whenever $\eta(x) \in [\lcb(x;\cF_m), \ucb(x;\cF_m)]$. 
As a result, we have $\exc_{\gamma}(\wh h_m;x) \leq 0$ as well.
\end{proof}

\subsection{Proofs and Supporting Results for \cref{al_abs:sec:standard_excess_error}}
\label{al_abs:app:standard}
\propProperAbs*
\begin{proof}
The proper abstention property of $\wh h$ returned by \cref{al_abs:alg:epoch} is achieved via conservation: $\wh h$ will avoid abstention unless it is absolutely sure that abstention is the optimal choice.
The proper abstention property implies that $\P_{x \sim \cD_\cX} (\wh h(x) = \bot) \leq \P_{x \sim \cD_\cX} (x \in \cX_\gamma)$. The desired result follows by combining this inequality with \cref{al_abs:eq:randomization}.
\end{proof}

\thmStandardExcessError*
\begin{proof}
	The results follow by taking the corresponding $\gamma$ in \cref{al_abs:alg:epoch} and then apply \cref{al_abs:prop:proper_abstention}.
	In the case with Massart noise, we have $\P_{x \sim \cD_\cX} (x \in \cX_{\gamma}) =0$ when $\gamma = \tau_0$; and the corresponding label complexity scales as $\wt O(\tau_0^{-2})$. 
	In the case with Tsybakov noise, we have $\gamma \cdot \P_{x \sim \cD_\cX} (x \in \cX_{\gamma}) = \frac{\eps}{2}$ when $\gamma = (\frac{\eps}{2c})^{1 /(1+\beta)}$. Applying \cref{al_abs:alg:epoch} to achieve $\frac{\eps}{2}$ Chow's excess error thus leads to $\frac{\eps}{2} + \frac{\eps}{2} = \eps$ standard excess error. The corresponding label complexity scales as $\wt O(\eps^{-2 /(1+\beta)})$. 
\end{proof}

\thmStandardExcessErrorNoise*
\begin{proof}
For any abstention parameter $\gamma > 0$, we denote $\cX_{\zeta_0, \gamma} \ldef \crl{x \in \cX: \eta(x) \in \brk{\frac{1}{2} - \gamma , \frac{1}{2} + \gamma  }, \abs{\eta(x) - 1 /2 } > \zeta_0 }$ as the intersection of the region controlled by noise-seeking conditions and the (possible) abstention region. 
Let $\wh h$ be the classifier returned by \cref{al_abs:alg:epoch} and $\check h$ be its randomized version (over the abstention region).
We denote $\cS \ldef \crl{x \in \cX: \wh h(x) = \bot}$ be the abstention region of $\wh h$. 
Since $\wh h$ abstains properly, we have $\cS \subseteq \crl{x \in \cX: \abs{\eta(x) - 1 /2} \leq \gamma} \rdef \cX_{\gamma} $. 
We write  $\cS_0 \ldef \cS \cap \cX_{\zeta_0, \gamma}$, $\cS_1 \ldef \cS \setminus \cS_0$ and $\cS_2 \ldef \cX \setminus \cS$.
For any $h: \cX \rightarrow \cY$, we use the notation $\exc(h;x) \ldef \prn{\P_{y\mid x} \prn[\big]{y \neq h(x)} -  \P_{y\mid x} \prn[\big]{ y \neq h^\star(x) }}$, and have $\exc(h) = \E_{x \sim \cD_\cX} \brk{\exc(h;x)}$.
We then have 
 \begin{align*}
	\exc(\check h)
	& = \E_{x \sim \cD_\cX} \brk*{\exc(\check h; x) \cdot \ind \prn{x \in \cS_0}} 
	+ \E_{x \sim \cD_\cX} \brk*{\exc(\check h; x) \cdot \ind \prn{x \in \cS_1}}\\
   & \quad + \E_{x \sim \cD_\cX} \brk*{\exc(\check h; x) \cdot \ind \prn{x \in \cS_2}}\\
	& \leq \gamma \cdot \E_{x \sim \cD_\cX} \brk{\ind(x \in \cS_0)}
	+ \zeta_0 \cdot \E_{x \sim \cD_\cX} \brk{\ind(x \in \cS_1)}\\
   & \quad + \E_{x \sim \cD_\cX} \brk{\exc_{\gamma}(\wh h;x) \cdot \ind(x \in \cS_2)}\\
	& \leq \gamma \cdot \E_{x \sim \cD_\cX} \brk{\ind(x \in \cX_{\zeta_0,\gamma})} + \zeta_0 + \eps /2 ,
\end{align*}
where the bound on the third term comes from the same analysis that appears in the proof of \cref{al_abs:thm:epoch} (with $\eps /2$ accuracy).
One can then tune $\gamma$ in ways discussed in the proof of \cref{al_abs:thm:standard_excess_error} to bound the first term by $\eps /2$, i.e., 
$\gamma \cdot \E_{x \sim \cD_\cX} \brk{\ind(x \in \cX_{\zeta_0,\gamma})} \leq \eps /2$, with similar label complexity.
\end{proof}

\propBudget*

Before proving \cref{al_abs:prop:budget}, we first construct a simple problem with linear regression function and give the formal definition of ``uncertainty-based'' active learner.

\begin{example}
   \label{al_abs:ex:linear} 
   We consider the case where $\cX = [0,1]$ and $\cD_\cX = \unif(\cX)$. 
   We consider feature embedding $\phi:\cX \rightarrow \R^2$, i.e., $\phi(x) = \brk{ \phi_1(x), \phi_2(x)}^{\trn}$. 
   We take $\phi_1(x) \ldef 1$ for any $x \in \cX$, and define $\phi_2(x)$ as 
   \begin{align*}
       \phi_2(x) \ldef  \begin{cases}
	   0, & x \in \cX_{\hard}, \\
	   1, & x \in \cX_{\easy},
       \end{cases} 
   \end{align*}
   where $\cX_{\easy} \subseteq \cX$ is any subset such that $\cD_\cX(\cX_{\easy}) = p$, for some constant $p \in (0,1)$, and $\cX_{\hard} = \cX \setminus \cX_{\easy}$.
   We consider a set of linear regression function $\cF \ldef \crl{ f_\theta: f_\theta(x) = \ang{ \phi(x), \theta}, \nrm{\theta}_2 \leq 1}$.
   We set  $f^{\star} = f_{\theta^{\star}}$, where $\theta^{\star}= \brk{ \theta^{\star}_1, \theta_2^{\star}}^{\trn}$ is selected such that $\theta_1^{\star} = \frac{1}{2}$ and $\theta_2^{\star} = \unif(\crl{\pm \frac{1}{2}})$.
\end{example}

\begin{definition}
	\label{al_abs:def:proper_learner}
	We say an algorithm is a ``uncertainty-based'' active learner if, for any $x \in \cX$, the learner
\begin{itemize}
	\item 	constructs an open confidence interval $(\lcb(x), \ucb(x))$ with $\eta(x) \in (\lcb(x), \ucb(x))$;\footnote{By restricting to learners that construct an open confidence interval containing $\eta(x)$, we do not consider the corner cases when $\lcb(x) = \frac{1}{2}$ or $\ucb(x) = \frac{1}{2}$ and the confidence interval close.}
	\item  queries the label of $x \in \cX$ if $\frac{1}{2} \in (\lcb(x), \ucb(x))$.
\end{itemize}
\end{definition}
\begin{proof}
    With any given labeling budget $B$, we consider the problem instance described in \cref{al_abs:ex:linear} with 
   $p = B^{-1}/2$.
   We can easily see that this problem instance satisfy \cref{al_abs:def:noise_seeking_Massart} and \cref{al_abs:def:noise_seeking_Tsybakov}.
    
    We first consider any ``uncertainty-based'' active learner. Let $Z$ denote the number of data points lie in $\cX_{\easy}$ among the first $B$ random draw of examples. 
    We see that $Z \sim \cB(B, B^{-1}/2)$ follows a binomial distribution with $B$ trials and $B^{-1}/2$ success rate.
    By Markov inequality, we have 
    \begin{align*}
	\P \prn*{Z \geq \frac{3}{2} \E[Z]} =  
	\P \prn*{Z \geq \frac{3}{4} } \leq \frac{2}{3}.
    \end{align*}
    That being said, with probability at least $1/3$, there will be $Z=0$ data point that randomly drawn from the easy region $\cX_{\easy}$. We denote that event as $\cE$. 
    Since $\eta(x) = f^{\star}(x) = \frac{1}{2}$ for any $x \in \cX_{\hard}$, any ``uncertainty-based'' active learner will query the label of any data point $x \in \cX_{\hard}$.
    As a result, under event $\cE$, the active learner will use up all the labeling budget in the first $B$ rounds and observe zero label for any data point $x \in \cX_{\easy}$.
    Since the easy region $\cX_{\easy}$ has measure $B^{-1} / 2$ and $\theta^{\star}_2 = \unif( \crl{ \pm \frac{1}{2}})$, any classification rule over the easy region would results in expected excess error lower bounded by
     $B^{-1}/ 4$.
     To summarize, with probability at least $\frac{1}{3}$, any ``uncertainty-based'' active learner without abstention suffers expected excess error $\Omega(B^{-1})$.

    We now consider the classifier returned by \cref{al_abs:alg:epoch}.
    For the linear function considered in \cref{al_abs:ex:linear}, we have $\pdim(\cF) \leq 2$ \citep{haussler1989decision} and $\theta^{\val}_{f^\star}(\cF, \gamma / 2, \eps) \leq 2$ for any $\eps \geq 0$ (see \cref{al_abs:app:star_eluder}).
    Thus, by setting $T = O \prn{ \frac{1}{\eps \, \gamma} \cdot \log \prn{ \frac{1}{\eps \, \gamma \, \delta}}}$, with probability at least $1 - \delta$, \cref{al_abs:alg:epoch} return a classifier $\wh h$ with Chow's excess error at most $\eps$ and label complexity $O \prn{ \frac{1}{\gamma^2} \cdot \log^2 \prn{ \frac{1}{\eps \, \gamma}} \cdot \log \prn{ \frac{1}{\eps \, \gamma \, \delta}}} = \poly \prn{ \frac{1}{\gamma}, \log \prn{ \frac{1}{\eps\, \gamma \, \delta}}} $. 
    Since $\wh h$ enjoys proper abstention, it never abstains for $x \in \cX_{\easy}$. 
    Note that we have $\eta(x) = \frac{1}{2}$ for any $x \in \cX_{\hard}$.
    By randomizing the decision of $\wh h$ over the abstention region, we obtain a classifier with standard excess error at most $\eps$. 
    \looseness=-1

\end{proof}
\subsection{Proofs and Supporting Results for \cref{al_abs:sec:constant}}
\label{al_abs:app:constant}
We introduce a new perspective for designing and analyzing active learning algorithms in \cref{al_abs:app:constant_regret}.
We present our algorithm and its theoretical guarantees in \cref{al_abs:app:constant_alg}, and defer  
supporting lemmas to \cref{al_abs:app:constant_lms}.

\subsubsection{The Perspective: Regret Minimization with Selective Sampling}
\label{al_abs:app:constant_regret}
We view active learning as a decision making problem: at each round, the learner selects an action, suffers a loss (that may not be observable), and decides to query the label or not. 
At a high level, the learner aims at \emph{simultaneously} minimizing the regret and the number of queries. The leaner returns a (randomized) classifier/decision rule at the end of the learning process. 

The perspective is inspired by the seminal results derived in \citet{dekel2012selective}, where the authors study active learning with linear functions and focus on developing standard excess error guarantees.
With this regret minimization perspective, we can also take advantage of fruitful results developed in the field of contextual bandits \citep{russo2013eluder, foster2020instance}.

\paragraph{Decision making for regret minimization} To formulate the regret minimization problem, we consider the action set $\cA = \crl{0, 1, \bot}$, where the action $1$ (resp. $0$) represents labeling any data point $x\in \cX$ as $1$ (resp. $0$); and the action $\bot$ represents abstention. 
At each round $t \in [T]$, the learner observes a data point $x_t \in \cX$ (which can be chosen by an adaptive adversary), takes an action $a_t \in \cA$, and then suffers a loss, which is defined as  
\begin{align*}
    \ell_t(a_t) = \ind(y_t \neq a_t, a_t \neq \bot) + \prn*{\frac{1}{2} - \gamma} \cdot \ind (a_t = \bot).
\end{align*}
We use $a^{\star}_t \ldef \ind(f^{\star}(x_t) \geq 1/2) = \ind( \eta(x_t)\geq 1/2)$ to denote the action taken by the Bayes optimal classifier $h^{\star}\in \cH$. 
Denote filtration $\mfF_{t} \ldef \sigma(\paren*{x_i, y_i}_{i=1}^{t})$. We define the (conditional) expected regret at time step $t \in [T]$ as 
\begin{align*}
    {\reg}_t \ldef \E \sq{\ell_t(a_t) - \ell_t(a^\star_t)  \mid  \mfF_{t-1}}.
\end{align*}
The (conditional) expected cumulative regret across $T$ rounds is defined as
\begin{align*}
	{\reg}(T) \ldef \sum_{t=1}^T {\reg}_t,
\end{align*}
which is the target that the learner aims at minimizing.

\paragraph{Selective querying for label efficiency} Besides choosing an action $a_t \in \cA$ at each time step, our algorithm also determines \emph{whether or not} to query the label $y_t$ with respect to $x_t$. 
Note that such selective querying protocol makes our problem different from contextual bandits \citep{russo2013eluder, foster2020instance}:
The loss $\ell_t(a_t)$ of an chosen $a_t$ may not be even observed.

We use $Q_t$ to indicate the query status at round $t$, i.e., 
\begin{align*}
    Q_t = \ind \paren*{\text{label $y_t$ of $x_t$ is queried}}.
\end{align*}
The learner also aims at minimizing the total number of queries across $T$ rounds, i.e., $\sum_{t=1}^{T} Q_t$.

\paragraph{Connection to active learning} 
We consider the following learner for the above mentioned decision making problem with $(x, y) \sim \cD_{\cX \cY}$. 
At each round, the learner constructs a classifier $\wh h_t: \cX \rightarrow \crl{0,1,\bot}$ and
a query function $g_t:\cX \rightarrow \crl{0,1}$; the learner then takes action $a_t = \wh h_t(x_t)$ and decides the query status as $Q_t = g_t(x_t)$.

Conditioned on $\mfF_{t-1}$, taking expectation over $\ell_t(a_t)$ leads to the following equivalence:
\begin{align*}
   \E \brk*{\ell_t(a_t) \mid \mfF_{t-1}} & = \E \brk*{ \ind(y_t \neq a_t, a_t \neq \bot) + \prn*{\frac{1}{2} - \gamma} \cdot \ind (a_t = \bot) \mid \mfF_{t-1}}\\
   & =  \E \brk*{ \ind \prn[\big]{y_t \neq \widehat h(x_t), \widehat h(x_t) \neq \bot} + \prn*{\frac{1}{2} - \gamma} \cdot \ind \prn[\big]{\widehat h(x_t) = \bot}  \mid \mfF_{t-1}} \\
   & =  \P_{(x,y) \sim \cD_{\cX \cY}} \prn[\big]{y \neq \wh h(x), \wh h(x) \neq \bot} + \prn*{\frac{1}{2} - \gamma} \cdot \P \paren{\widehat h(x) = \bot}\\
   & = \err_{\gamma}\prn{\wh h_t}.
\end{align*}
This shows that the (conditional) expected instantaneous loss precisely captures the Chow's error of classifier $\widehat h_t$. Similarly, we have 
\begin{align*}
   \E \sq*{\ell_t(a_t^\star) \mid \mfF_{t-1} } = \P_{(x,y) \sim \cD_{\cX \cY}} \paren*{ \ind \paren{ y \neq \ind(\eta(x) \geq 1/2) }} = \err(h^\star).
\end{align*}
Combining the above two results, we notice that the (conditional) expected instantaneous \emph{regret} exactly captures the Chow's excess error of classifier $\widehat h_t$, i.e., 
\begin{align*}
     \reg_t  = \err_{\gamma}(\widehat h_t) - \err(h^\star).
\end{align*}
Let $\wh h \sim \unif( \crl{\wh h_t}_{t=1}^{T})$ be a classifier randomly selected from all the constructed classifiers. Taking expectation with respect to this random selection procedure, we then have 
\begin{align}
\E_{\wh h \sim \unif( \crl{\wh h_t}_{t=1}^{T})} \brk{ \err_{\gamma}(\widehat h) - \err(h^\star) } = \sum_{t=1}^T \paren{ \err_{\gamma}(\widehat h_t) - \err(h^\star) } / T = \reg(T)/T . 
\label{al_abs:eq:expect_chow}
\end{align}
If we manage to guarantee that the cumulative regret is sublinear in $T$ and the total number of queries is logarithmic in $T$, we would achieve the goal of active learning with exponential savings in label complexity.

For analysis purpose, we also consider another classifier $\wh h_t^\star$, which is defined as 
\begin{align*}
  \wh h_t^\star(x) \ldef \begin{cases}
    \bot, & \text{if } \wh h_t(x) = \bot;\\
    h^\star(x), & \text{o.w.}
  \end{cases}
\end{align*}
That is, $\wh h_t^\star$ abstains whenever $\wh h_t$ abstains, and follows the Bayes optimal classifier otherwise. We use $\wh a_t^\star = \wh h_t(x_t)$ to denote the action of $\wh h_t^\star$ at round $t$ and have $\E \brk{\ell_t(a_t) \mid \mfF_{t-1}} = \err_\gamma(\wh h_t^\star)$.

\subsubsection{Algorithm and Main Results}
\label{al_abs:app:constant_alg}

We present our algorithm that achieves constant label complexity in \cref{al_abs:alg:eluder}. 
Compared to \cref{al_abs:alg:epoch}, \cref{al_abs:alg:eluder} drops the epoch scheduling, uses a sharper elimination rule for the active set (note that $\beta$ doesn't depend on $T$, thanks to the optimal stopping theorem in \cref{al_abs:lm:opt_stopping}), and is analyzed with respect to eluder dimension (\cref{al_abs:def:eluder}) instead of disagreement coefficient.
As a result, we shave all three sources of $\log \frac{1}{\eps}$, and achieve constant label complexity for general $\cF$ (as long as it's finite and has finite eluder dimension). 
We abbreviate $\mfe \ldef \sup_{f^{\star} \in \cF}\mfe_{f^{\star}}(\cF, \gamma /2)$.

\begin{algorithm}[H]
	\caption{Efficient Active Learning with Abstention (Constant Label Complexity)}
	\label{al_abs:alg:eluder} 
	\renewcommand{\algorithmicrequire}{\textbf{Input:}}
	\renewcommand{\algorithmicensure}{\textbf{Output:}}
	\newcommand{\algorithmicbreak}{\textbf{break}}
    \newcommand{\BREAK}{\STATE \algorithmicbreak}
	\begin{algorithmic}[1]
		\REQUIRE Time horizon $T \in \N$, abstention parameter $\gamma \in (0, 1/2)$ and confidence level $\delta \in (0, 1)$.
		\STATE Initialize $\wh \cH \ldef \emptyset$. Set $T \ldef O \prn{\frac{\mfe}{\eps \, \gamma} \cdot \log \prn{\frac{\abs{\cF}}{\delta}}}$ and $\beta\ldef{2}	\log\prn[\big]{\frac{2 \abs*{\cF}}{\delta}}$. 
		\FOR{$t = 1, 2, \dots, T$}
		\STATE Get $\widehat f_t \ldef \argmin_{f \in \cF} \sum_{i < t} Q_i \paren{f(x_i) - y_i}^2 $.\\
		\hfill \algcommentlight{We use $Q_t \in \crl{0,1}$ to indicate whether the label of  $x_t$ is queried.}
		\STATE (Implicitly) Construct active set of regression function $\cF_t \subseteq \cF$ as 
		\begin{align*}
		    \cF_t \ldef \crl*{ f \in \cF:  \sum_{i = 1}^{t-1} Q_i \prn*{f(x_i) - y_i}^2 \leq \sum_{i = 1}^{t-1} Q_i \paren{\widehat f_t(x_i) - y_i}^2 + \beta}. 
		\end{align*}
		\STATE Construct classifier $\wh h_t:\cX \rightarrow \crl{0, 1,\bot}$ as 
		\begin{align*}
			\wh h_t (x) \ldef 
			\begin{cases}
				\bot, & \text{ if } \brk { \lcb(x;\cF_t), \ucb(x;\cF_t)} \subseteq 
				\brk*{ \frac{1}{2} - \gamma, \frac{1}{2} + \gamma}; \\
        \ind(\wh f_t(x) \geq \frac{1}{2} ) , & \text{o.w.}
			\end{cases}
		\end{align*}
		Update $\wh \cH = \wh \cH \cup \crl{\wh h_t}$.	
		Construct query function $g_m:\cX \rightarrow \crl{0,1}$ as
		\begin{align*}
		g_t(x)\ldef \ind \prn*{ \frac{1}{2} \in \prn{\lcb(x;\cF_t), \ucb(x;\cF_t)} } \cdot 	
		\ind \prn{\wh h_t(x) \neq \bot} .
		\end{align*}
		\STATE Observe $x_t \sim \cD_{\cX}$. Take action $a_t \ldef \wh h_t (x_t)$. Set $Q_t \ldef g_t(x_t)$. 
		\IF{$Q_t = 1$}
		\STATE Query the label $y_t$ of $x_t$.
		\ENDIF
		\ENDFOR
		\STATE \textbf{Return} $\wh h \ldef \unif(\wh \cH)$.
	\end{algorithmic}
\end{algorithm}

Before proving \cref{al_abs:thm:constant}. We define some notations that are specialized to \cref{al_abs:app:constant}.
We define filtrations $\mfF_{t-1} \ldef  \sigma ( x_1,y_1,\ldots, x_{t-1} ,y_{t-1})$ and 
$\wb \mfF_{t-1} \ldef  \sigma ( x_1,y_1,\ldots, x_{t-1}, y_{t-1}, x_t )$.
Note that we additionally include the data point $x_t$ in the filtration $\wb \mfF_{t-1}$ at time step $t-1$.
We denote $\E_t [\cdot] \ldef \E [\cdot \mid \wb \mfF_{t-1}]$. 
For any $t \in [T]$, we denote $M_t(f) \ldef Q_t \prn{ \prn{f(x_t) - y_t}^2 - \prn{f^\star(x_t) - y_t}^2}$. 
We have $\sum_{i=1}^{\tau} \E_t[M_t(f)] = \sum_{t=1}^{\tau} Q_t \prn{f(x_t) - f^\star(x_t)}^2$.
For any given data point $x_t \in \cX$, we use abbreviations
\begin{align*}
    \ucb_t \ldef \ucb(x_t; \cF_t) = \sup_{f \in \cF_t} f(x_t) \quad  \text{ and }  \quad \lcb_t \ldef \lcb(x_t; \cF_t) = \inf_{f \in \cF_t} f(x_t)
\end{align*}
to denote the upper and lower confidence bounds of $\eta(x_t) = f^\star(x_t)$. We also denote 
\begin{align*}
    w_t \ldef \ucb_t - \lcb_t = \sup_{f, f^\prime \in \cF_t} \abs*{f(x_t) - f^\prime(x_t)}
\end{align*}
as the width of confidence interval.

\thmConstant*

\begin{proof}
	We first analyze the label complexity of \cref{al_abs:alg:eluder}.
	Note that \cref{al_abs:alg:eluder} constructs $\wh h_t$ and $g_t$ in forms similar to the ones constructed in \cref{al_abs:alg:epoch}, and \cref{al_abs:lm:query_implies_width} holds for \cref{al_abs:alg:eluder} as well.
	Based on \cref{al_abs:lm:query_implies_width}, we have $Q_t = g_t(x_t) = 1 \implies w_t > \gamma$. 
	Thus, taking $\zeta = \gamma$ in \cref{al_abs:lm:conf_width_eluder} leads to 
	$$
	\sum_{t=1}^{T} \ind (Q_t =1) < \frac{17 \log (2 \abs{\cF}/ \delta)}{2 \gamma^2} \cdot \mfe_{f^{\star}}(\cF, \gamma / 2),
	$$ 
	\emph{with probability one}. 
	The label complexity of \cref{al_abs:alg:eluder} is then upper bounded by a constant as long as $\mfe_{f^{\star}}(\cF, \gamma / 2)$ is upper bounded by a constant (which has no dependence on $T$ or $\frac{1}{\eps}$).

We next analyze the excess error of $\wh h$. We consider the good event $\cE$ defined in \cref{al_abs:lm:set_f_eluder}, which holds true with probability at least $1-\delta$.
Under event $\cE$, \cref{al_abs:lm:regret_eluder_cond_wh_a} shows that 
\begin{align*}
	\sum_{t=1}^{T} \E \brk{\ell_t(a_t) - \ell_t(\wh a_t^\star) \mid \wb \mfF_{t-1} } \leq  
	\frac{17 \sqrt{2} \beta}{\gamma} \cdot \mfe_{f^{\star}}(\cF, \gamma / 2). 
\end{align*}
Since 
\begin{align*}
	\E \brk[\Big] {\E \brk{\ell_t(a_t) - \ell_t(\wh a_t^\star) \mid \wb \mfF_{t-1} } \mid \mfF_{t-1}}
	= \E \brk{\ell_t(a_t) - \ell_t(\wh a_t^\star) \mid \mfF_{t-1} },
\end{align*}
and 
  $ 0 \leq {\E \brk{\ell_t(a_t) - \ell_t(\wh a_t^\star) \mid \wb \mfF_{t-1} }}  \leq 1$ by \cref{al_abs:lm:wh_a_ineq}, 
applying \cref{al_abs:lm:martingale_two_sides} with respect to $\E \brk{\ell_t(a_t) - \ell_t(\wh a_t^\star) \mid \wb \mfF_{t-1} }$ leads to 
\begin{align*}
  \sum_{t=1}^{T} \E \brk{\ell_t(a_t) - \ell_t(\wh a_t^\star) \mid \mfF_{t-1} } \leq  
	\frac{34 \sqrt{2} \beta}{\gamma} \cdot \mfe_{f^{\star}}(\cF, \gamma / 2) + 8 \log(2 \delta^{-1}).
\end{align*}
  From \cref{al_abs:lm:wh_a_ineq}, we know that 
  \begin{align*}
    \E \brk{\ell_t(\wh a^\star_t) - \ell_t(a_t^\star) \mid \mfF_{t-1} } = 
    \E \brk[\Big]{\E \brk{\ell_t(\wh a^\star_t) - \ell_t(a_t^\star) \mid \wb \mfF_{t-1} } \mid \mfF_{t-1}} \leq 0. 
  \end{align*}
  We then have 
\begin{align*}
  \reg(T) &=	\sum_{t=1}^{T} \E \brk{\ell_t(a_t) - \ell_t(a_t^\star) \mid \mfF_{t-1} }\\
  &=	\sum_{t=1}^{T} \E \brk{\ell_t(a_t) - \ell_t(\wh a_t^\star) \mid \mfF_{t-1} } + \sum_{t=1}^{T} \E \brk{\ell_t(\wh a^\star_t) - \ell_t(a_t^\star) \mid \mfF_{t-1} }\\
  &\leq  
	\frac{34 \sqrt{2} \beta}{\gamma} \cdot \mfe_{f^{\star}}(\cF, \gamma / 2) + 8 \log(2 \delta^{-1}),
\end{align*}
with probability at least $1-2\delta$ (due to the additional application of \cref{al_abs:lm:martingale_two_sides}).
Since $\wh h \sim \unif(\wh \cH)$, based on \cref{al_abs:eq:expect_chow}, we thus know that 
\begin{align*}
\E_{\wh h \sim \unif(\wh \cH)} \brk{ \err_\gamma(\wh h) - \err(h^{\star})} & = \sum_{t=1}^{T}\prn[\big]{\err_\gamma(\wh h_t) - \err( h^{\star})}/ T \\
	&\leq \prn*{\frac{34 \sqrt{2} \beta}{ \gamma} \cdot \mfe_{f^{\star}}(\cF, \gamma / 2) + 8 \log \prn*{2 \delta^{-1}}}/T 
\end{align*}
With $T \ldef  O \prn{\frac{\mfe}{\eps \, \gamma} \cdot \log \prn{ \frac{\abs{\cF}}{\delta}}}$, we can control the expected Chow's excess error to be at most $\eps$.
\end{proof}

\begin{theorem}
	\label{al_abs:thm:constant_adv}
	Consider the setting where the data points $\crl{x_t}_{t=1}^{T}$ are chosen by an adaptive adversary with $y_t \sim \cD_{\cY \mid x_t}$. With probability at least $1-\delta$, \cref{al_abs:alg:eluder} simultaneously guarantees 
\begin{align*}
	\sum_{t=1}^{T} \E \brk{\ell_t(a_t) - \ell_t(a_t^\star) \mid \wb \mfF_{t-1} } \leq  
	\frac{34 \sqrt{2} \beta}{\gamma} \cdot \mfe_{f^{\star}}(\cF, \gamma / 2),
\end{align*}
and 
	$$
	\sum_{t=1}^{T} \ind (Q_t =1) < \frac{17 \log (2 \abs{\cF}/ \delta)}{2 \gamma^2} \cdot \mfe_{f^{\star}}(\cF, \gamma / 2).
	$$ 
\end{theorem}
\begin{proof}
  The label complexity follows the same analysis as in the proof of \cref{al_abs:thm:constant}.

  To analyze the regret, we consider the good event $\cE$ defined in \cref{al_abs:lm:set_f_eluder}, which holds true with probability at least $1-\delta$.
Under event $\cE$, \cref{al_abs:lm:regret_eluder_cond} shows that 
\begin{align*}
	\sum_{t=1}^{T} \E \brk{\ell_t(a_t) - \ell_t(a_t^\star) \mid \wb \mfF_{t-1} } \leq  
	\frac{17 \sqrt{2} \beta}{\gamma} \cdot \mfe_{f^{\star}}(\cF, \gamma / 2). 
\end{align*}
\end{proof}

We redefine $\mfe \ldef \sup_{f^{\star} \in \cF}\mfe_{f^{\star}}(\cF, \gamma /4)$ in the following \cref{al_abs:thm:eluder_efficient} to account for the induced approximation error in efficient implementation.
\begin{restatable}{theorem}{thmEluderEfficient}
	\label{al_abs:thm:eluder_efficient}
	\cref{al_abs:alg:eluder} can be efficiently implemented via the regression oracle and enjoys the same theoretical guarantees stated in \cref{al_abs:thm:constant} or \cref{al_abs:thm:constant_adv}.
	The number of oracle calls needed is $ O(\frac{\mfe}{\eps \, \gamma^{3}} \cdot \log \prn{\frac{\abs{\cF}}{\delta}}\cdot \log \prn{\frac{1}{\gamma}})$ for a general set of regression functions $\cF$, and $ O(\frac{\mfe}{\eps \, \gamma} \cdot \log \prn{\frac{\abs{\cF}}{\delta}} \cdot \log \prn{\frac{1}{\gamma}})$ when $\cF$ is convex and closed under pointwise convergence.
	The per-example inference time of the learned $\wh h_{M}$ is $O ( \frac{1}{\gamma^2} \log \frac{1}{\gamma})$ for general $\cF$, and $O ( \log \frac{1}{\gamma}) $ when $\cF$ is convex and closed under pointwise convergence.
\end{restatable}
\begin{proof}
Denote $\cB_t \ldef \crl{ (x_i,Q_i, y_i)}_{i=1}^{\tau_{t-1}}$ 
At any time step $t \in [T]$ of \cref{al_abs:alg:eluder}, we construct classifier $\wh h_t$ and query function $g_t$ with approximated confidence bounds, i.e.,
\begin{align*}
&	\wh \lcb(x;\cF_t) \ldef \AlgLcb(x;\cB_t,\beta_t,\alpha)  \quad \text{and}	
    & \wh \ucb(x;\cF_t) \ldef\AlgUcb(x;\cB_t,\beta_t,\alpha), 
\end{align*}
where $\AlgLcb$ and $\AlgUcb$ are subroutines discussed in \cref{al_abs:prop:CI_oracle} and $\alpha \ldef \frac{\gamma}{4}$.

Since the theoretical analysis of \cref{al_abs:thm:constant} and \cref{al_abs:thm:constant_adv} do not require an non-increasing (with respect to time step $t$) sampling region, i.e., $\crl{x \in \cX: g_t(x) = 1}$, we only need to approximate the confidence intervals at $\frac{\gamma}{4}$ level.
This slightly save the computational complexity 
compared to \cref{al_abs:thm:epoch_efficient}, which approximates the confidence interval at $\frac{\gamma}{4 \ceil{\log T}}$ level. The rest of the analysis of computational complexity follows similar steps in the proof of \cref{al_abs:thm:epoch_efficient}.
\end{proof}

\subsubsection{Supporting Lemmas}
\label{al_abs:app:constant_lms}

Consider a sequence of random variables $\prn{Z_t}_{t \in \N}$ adapted to filtration ${\wb \mfF_t}$. 
We assume that $\E \sq*{\exp(\lambda Z_t)} < \infty$ for all $\lambda$. Denote $\mu_t \ldef \E \sq*{ Z_t  \mid \wb \mfF_{t-1} }$ 
and
$$ \psi_t(\lambda) \ldef \log  \E \sq*{ \exp( \lambda \cdot \paren*{ Z_t - \mu_t } )  \mid \wb \mfF_{t-1} } 
.$$ 

\begin{lemma}[\citet{russo2013eluder}]
	\label{al_abs:lm:opt_stopping}
With notations defined above. For any $\lambda \geq 0$ and $\delta > 0$, we have 
\begin{align}
    \P \prn*{ \forall \tau \in \N, \sum_{t = 1}^{\tau} \lambda Z_t \leq \sum_{t=1}^{\tau} \prn*{ \lambda \mu_t + \psi_t(\lambda) } + \log \prn*{\frac{1}{\delta}} } \geq 1 -\delta.
\end{align}
\end{lemma}

\begin{lemma}
    \label{al_abs:lm:expected_sq_loss_opt}
   Fix any $\delta \in (0,1)$.  For any $\tau\in [T]$, with probability at least $1 - \delta $, we have 
   \begin{align*}
   	\sum_{t = 1}^{\tau} M_t(f) \leq \sum_{t=1}^{\tau} \frac{3}{2} \E_t \brk{M_t(f)} + 
	C_\delta,
   \end{align*}
   and
   \begin{align*}
       \sum_{t = 1}^{\tau} \E_t \sq{ M_t(f)}  \leq 2 \sum_{t = 1}^{\tau} M_t(f) + C_\delta,
   \end{align*}
   where $C_\delta \ldef  4 \log \prn*{ \frac{2 \abs{\cF} }{\delta}}$.
\end{lemma}
\begin{proof}
	Fix any $f \in \cF$. We take $Z_t = M_t(f)\ldef Q_t \prn{ \prn{f(x_t) - y_t}^2 - \prn{f^\star(x_t) - y_t}^2}$ in \cref{al_abs:lm:opt_stopping}. 
	We can rewrite 
	$$Z_t = Q_t \prn*{ \prn{f(x_t) - f^{\star}(x_t) }^2 + 2 \prn{f(x_t) - f^{\star}(x_t)} \eps_t},$$ 
	where we use the notation $\eps_t \ldef f^{\star}(x_t) - y_t$. Since $\E_t [\eps_t] = 0$ and $\E_t \brk{\exp(\lambda \eps_t) \mid \wb \mfF_{t-1} } \leq \exp(\frac{\lambda^2}{2})$ by Hoeffding Lemma, we have 
	\begin{align*}
		\mu_t  = \E_t \brk{Z_t}  = Q_t \prn*{ f(x_t) - f^{\star}(x_t)}^2 ,
	\end{align*}
	and 
	\begin{align*}
		\psi_t(\lambda) & = \log  \E \sq*{ \exp( \lambda \cdot \paren*{ Z_t - \mu_t } )  \mid \wb \mfF_{t-1} }\\
		& = \log \E_t \brk*{ \exp \prn*{2 \lambda Q_t \prn*{f(x_t) - f^{\star}(x_t) \cdot \eps_t }}} \\
		& \leq { \frac{\prn*{2 \lambda Q_t ( f(x_t) - f^{\star}(x_t) }^2 }{2}} \\
		& = {2 \lambda^2 \mu_t},
	\end{align*}
	where the last line comes from the fact that $Q_t \in \crl{0,1}$. 
	Plugging these results into \cref{al_abs:lm:opt_stopping} with $\lambda = 1 / 4$ leads to 
	\begin{align*}
		\sum_{t=1}^{\tau} M_t (f) \leq \sum_{t=1}^{\tau} \frac{3}{2} \E_t \brk*{ M_t(f)} + 4 \log \delta^{-1}. 
	\end{align*}
    	Following the same procedures above with $Z_t = - M_t(f)$ and $\lambda = 1/ 4$ leads to 
	\begin{align*}
		\sum_{t=1}^{\tau} \E_t \brk*{ M_t(f)} \leq 2 \sum_{t=1}^{\tau} M_t (f) + 4 \log \delta^{-1}. 
	\end{align*}
	The final guarantees come from taking a union abound over $f \in \cF$ and splitting the probability for both directions.
\end{proof}
We use $\cE$ to denote the good event considered in \cref{al_abs:lm:expected_sq_loss_opt}, we use it through out the rest of this section.

\begin{lemma}
\label{al_abs:lm:set_f_eluder}
With probability at least $1-\delta$, the followings hold true:
\begin{enumerate}
	\item $f^\star \in \cF_t$ for any $t \in [T]$.
	\item $\sum_{t=1}^{\tau-1} \E_t \brk{M_t(f)} \leq 2   C_\delta$ for any $f \in \cF_{\tau}$. 
\end{enumerate}
\end{lemma}
\begin{proof}
	The first statement immediately follows from \cref{al_abs:lm:expected_sq_loss_opt} (the second inequality) and the fact that $\beta \ldef C_{\delta}/{2}$ in \cref{al_abs:alg:eluder}.

For any $f \in \cF_{\tau}$, we have 
\begin{align}
\sum_{t=1}^{\tau-1} \E_t \brk{M_t(f)} & \leq 2\sum_{t=1}^{\tau-1} Q_t \prn*{ (f(x_t) - y_t)^2 - ( f^{\star}(x_t) - y_t)^2 } + C_\delta\nonumber\\
    & \leq 2\sum_{t=1}^{\tau-1} Q_t \prn*{ (f(x_t) - y_t)^2 - ( \wh f_\tau(x_t) - y_t)^2 }+ C_\delta \nonumber \\
    & \leq 2 C_\delta, \label{al_abs:eq:squared_diff_f_star}
\end{align}
where the first line comes from \cref{al_abs:lm:expected_sq_loss_opt}, the second line comes from the fact that $\wh f_\tau$ is the minimize among $\cF_\tau$, and the third line comes from the fact that $f \in \cF_\tau$ and $2 \beta = C_\delta$. 
\end{proof}

\begin{lemma}
\label{al_abs:lm:conf_width_eluder}
 For any $\zeta > 0$, with probability $1$, we have 
\begin{align*}
	\sum_{t = 1}^{T} \ind \prn*{Q_t = 1} \cdot \ind \prn*{ w_t > \zeta } < \prn*{ \frac{16\beta}{\zeta^2} + 1 } \cdot \mfe_{f^{\star}}(\cF, \zeta/2).
\end{align*}
\end{lemma}
\begin{remark}
	Similar upper bound has been established in the contextual bandit settings for $\sum_{t=1}^{T} \ind(w_t > \zeta)$ \citep{russo2013eluder, foster2020instance}. 
	 We develop our results with an additional $\ind(Q_t=1)$ term to account for selective querying in active learning.
\end{remark}
\begin{proof}
We give some definitions first. We say that $x$ is $\zeta$-independent of a sequence $x_1, \dots, x_{\tau}$ if there exists a $f \in \cF$ such that $\abs*{f(x) - f^\star(x)} > \zeta$ and $\sum_{i \leq {\tau}} \paren{f(x_i) - f^\star(x_i)}^2 \leq \zeta^2$. We say that $x$ is $\zeta$-dependent of $x_1, \dots, x_{\tau}$ if we have $\abs*{f(x) -f^\star(x)} \leq \zeta$ for all $f \in \cF$ such that $\sum_{i \leq {\tau}} \paren{f(x_i) - f^\star(x_i)}^2 \leq \zeta^2$. 

For any $t \in [T]$, and we denote $\cS_{t} = \curly*{x_i: Q_i = g_i(x_i)=1 , i \in [t]}$ as the \emph{queried} data points up to time step $t$. We assume that $\abs*{\cS_t} = \tau$ and denote $\cS_{t} = (x_{g(1)}, \dots, x_{g(\tau)})$, where $g(i)$ represents the time step where the $i$-th \emph{queried} data point is queried.

\textbf{Claim 1.} For any $j \in [\tau]$, $x_{g(j)}$ is $\frac{\zeta}{2}$-dependent on at most $\frac{16 \beta}{\zeta^2}$ disjoint subsequences of $x_{g(1)}, \dots, x_{g(j-1)}$.

For any $x_{g(j)} \in \cS_t$, recall that
$$w_{g(j)} = \ucb_{g(j)} - \lcb_{g(j)} = \max_{f, f^\prime \in \cF_{g(j)}} \abs*{f(x_t) - f^\prime(x_t)}.$$ 
If $w_{g(j)} > \zeta$, there must exists a $f \in \cF_{g(j)}$ such that $\abs*{f(x_{g(j)}) - f^\star(x_{g(j)})} > \frac{\zeta}{2}$. Focus on this specific $f \in \cF_{g(j)} \subseteq \cF$. 
If $x_{g(j)}$ is $\frac{\zeta}{2}$-dependent on a subsequence $x_{g(i_1)}, \dots, x_{g(i_{m})}$ (of $x_{g(1)}, \dots, x_{g(j-1)}$), we must have 
\begin{align*}
    \sum_{k \leq m} \paren{ f(x_{g(i_k)}) - f^\star(x_{g(i_k)}) }^2 > \frac{\zeta^2}{4}.
\end{align*}
Suppose $x_{g(j)}$ is $\frac{\zeta}{2}$-dependent on $K$ \emph{disjoint} subsequences of $x_{g(1)}, \dots, x_{g(j-1)}$, according to \cref{al_abs:lm:set_f_eluder}, we must have
\begin{align*}
    K \cdot \frac{\zeta^2}{4} < \sum_{i < j} \paren{f(x_{g(i)}) - f^\star(x_{g(i)})}^2 = \sum_{k < g(j)} Q_k \paren{f(x_k) - f^\star(x_k)}^2  \leq 4 \beta,
\end{align*}
which implies that $K < \frac{16 \beta}{\zeta^2}$.

\textbf{Claim 2.} Denote $d \ldef \check \mfe_{f^\star}(\cF, \zeta/2) \geq 1$ and $K \ldef \floor*{\frac{\tau-1}{d}}$. There must exists a $j \in [\tau]$ such that $x_{g(j)}$ is $\frac{\zeta}{2}$-dependent on at least $K$ disjoint subsequences of $x_{g(1)}, \dots, x_{g(j-1)}$.

We initialize $K$ subsequences $\cC_i = \crl{x_{g(i)}}$. If $x_{g(K+1)}$ is $\frac{\zeta}{2}$-dependent on each $\cC_i$, we are done. If not, select a subsequence $\cC_i$ such that $x_{g(K+1)}$ is $\frac{\zeta}{2}$-independent of and add $x_{g(K+1)}$ into this subsequence. Repeat this procedure with $j > K+1$ until $x_{g(j)}$ is $\frac{\zeta}{2}$-dependent of all $\cC_i$ or $j = \tau$. In the first case we prove the claim. In the later case, we have $\sum_{i \leq K} \abs*{\cC_i} = \tau - 1 \geq Kd$. Since $\abs*{\cC_i} \leq d$ by the construction of $\cC_i$ and the definition of $\check \mfe_{f^\star}(\cF, \zeta/2)$, we must have $\abs*{\cC_i} = d$ for all $i \in [K]$. As a result, $x_{g(\tau)}$ must be $\frac{\zeta}{2}$-dependent of all $\cC_i$.

It's easy to check that $\floor*{\frac{\tau - 1}{d}} \geq \frac{\tau}{d} - 1$. Combining Claim 1 and 2, we have 
\begin{align*}
    \frac{\tau}{d} -1 \leq \floor*{\frac{\tau - 1}{d}} \leq K < \frac{16 \beta}{\zeta^2}.
\end{align*}
Rearranging leads to the desired result.
\end{proof}
The following \cref{al_abs:lm:regret_no_query_constant} is a restatement of \cref{al_abs:lm:regret_no_query} in the regret minimization setting.
\begin{lemma}
\label{al_abs:lm:regret_no_query_constant}
If $Q_t = 0$, we have 
    $\E \sq*{\ell_t(a_t) - \ell_t(a_t^\star) \mid \wb \mfF_{t-1}} \leq 0$.
\end{lemma}
\begin{proof}
	Recall we have $a_t = \wh h_t (x_t)$. We then have  
\begin{align*}
    & \E  \sq*{\ell_t(a_t) - \ell_t(a_t^\star) \mid \wb \mfF_{t-1} } \\
    & = \P_{y_t \mid x_t} \prn[\big]{y_t \neq \widehat h_t(x_t)} \cdot \ind \prn[\big]{ \widehat h_t(x_t) \neq \bot} + \prn[\big]{{1}/{2} - \gamma} \cdot \1 \prn[\big]{\widehat h_t(x_t) = \bot} \\
    &\quad - \P_{y_t\mid x_t} \prn[\big]{ y_t \neq h^\star(x_t) }\\
    & = \ind \prn[\big]{ \widehat h_t(x_t) \neq \bot} \cdot \prn[\big]{\P_{y_t\mid x_t} \prn[\big]{y_t \neq \widehat h_t(x_t)} -  \P_{y_t\mid x_t} \prn[\big]{ y_t \neq h^\star(x_t) }} \\
    & \quad + \ind \prn[\big]{ \widehat h_t(x_t) = \bot} \cdot \prn[\big]{ \prn[\big]{{1}/{2} - \gamma}  -  \P_{y_t \mid x_t} \prn[\big]{ y_t \neq h^\star(x_t) }}.
\end{align*}
We now analyze the event $\curly*{Q_t = 0}$ in two cases. 

\textbf{Case 1: ${\widehat h_t(x_t) = \bot} $.} 

Since $\eta(x_t) = f^{\star}(x_t) \in [\lcb_t, \ucb_t]$, we further know that $\eta(x_t) \in \sq{ \frac{1}{2} - \gamma, \frac{1}{2} + \gamma }$ and thus $\P_{y_t\mid x_t} \prn[\big]{ y_t \neq h^\star(x_t) } \geq \frac{1}{2} - \gamma$. As a result, we have $\E \sq*{\ell_t(a_t) - \ell_t(a_t^\star) \mid \wb \mfF_{t-1} } \leq 0$.

\textbf{Case 2: ${\widehat h_t(x_t) \neq \bot}$ but ${\frac{1}{2} \notin (\lcb_t, \ucb_t)} $.} 

In this case, we know that $\widehat h_t (x_t) = h^\star(x_t)$ whenever $\eta(x_t) \in [\lcb_t, \ucb_t]$. As a result, we have \\
  $\E \sq*{\ell_t(a_t) - \ell_t(a_t^\star) \mid \wb \mfF_{t-1} } = 0$.
\end{proof}
\begin{lemma}
\label{al_abs:lm:regret_eluder_cond}
 Assume $\mu(x_t) \in [\lcb_t, \ucb_t]$ and $f^{\star}$ is not eliminated across all $t \in [T]$. We have 
\begin{align*}
	\sum_{t=1}^{T} \E \brk{\ell_t(a_t) - \ell_t(a_t^\star) \mid \wb \mfF_{t-1} } \leq 
	\frac{17 \sqrt{2} \beta}{\gamma} \cdot \mfe_{f^{\star}}(\cF, \gamma / 2).
\end{align*}
\end{lemma}
\begin{proof}
	\cref{al_abs:lm:regret_no_query_constant} shows that non-positive conditional regret is incurred at whenever $Q_t = 0$, we then have 
	\begin{align*}
	\sum_{t=1}^{T} \E \brk{\ell_t(a_t) - \ell_t(a_t^\star) \mid \wb \mfF_{t-1} } & 
	\leq \sum_{t=1}^{T} \ind(Q_t =1) \E \brk*{ \ell_t(a_t) - \ell_t(a_t^{\star}) \mid \wb \mfF_{t-1}}\\
	& \leq \sum_{t=1}^{T} \ind(Q_t = 1) \cdot \ind (w_t > \gamma) \cdot \abs*{ 2 f^{\star}(x_t) - 1}\\
	& \leq \sum_{t=1}^{T}\ind(Q_t = 1) \cdot \ind (w_t > \gamma) \cdot 2 w_t,
	\end{align*}
  where the second line comes from the fact that, under the event $\crl{Q_t = 1}$, we have $w_t > \gamma$ (using a similar analysis as in \cref{al_abs:lm:query_implies_width}) and $\E \brk*{ \ell_t(a_t) - \ell_t(a_t^{\star}) \mid \wb \mfF_{t-1}} \leq \abs{2 f^\star(x_t) - 1}$ (since $a_t \neq \bot$), the last line comes from the fact that $ \abs{f^{\star}(x_t) - \frac{1}{2}} \leq  w_t$ whenever $f^{\star}$ is not eliminated and $Q_t = 1$.
	We can directly apply $w_t \leq 1$ and \cref{al_abs:lm:conf_width_eluder} to bound the above terms by 
	$\wt O(\frac{ \mfe_{f^{\star}}(\cF, \gamma / 2)}{\gamma^2})$, which has slightly worse dependence on $\gamma$. Following \citet{foster2020instance}, we take a slightly tighter analysis below. 

	Let $\cS_T \ldef \crl{ x_i: Q_i =1, i \in [T] } $ denote the set of queried data points. Suppose $\abs{\cS_T} = \tau$. 
	Let $i_1, \ldots, i_\tau$ be a reordering of indices within $\cS_T$ such that $w_{i_1}(x_{i_1}) \geq w_{i_2}(x_{i_2}) \geq \ldots \geq w_{i_\tau} (x_{i_\tau})$. 
	Consider any index $t \in [\tau]$ such that $w_{i_t} (x_{i_t}) \geq \gamma$. For any $\zeta \geq \gamma$, \cref{al_abs:lm:conf_width_eluder} implies that 
	\begin{align}
	t \leq \sum_{t=1}^{T} \ind(Q_t = 1) \cdot \ind ( w_t(x_t) > \zeta) \leq \frac{17 \beta}{\zeta^2} \cdot \mfe_{f^{\star}}\prn*{\cF, {\zeta}/{2}} \leq \frac{17 \beta}{\zeta^2} \cdot \mfe_{f^{\star}}\prn*{\cF, {\gamma}/{2}} . \label{al_abs:eq:regret_eluder_cond}
	\end{align}
  Taking $\zeta = w_{i_t}(x_{i_t})$ in \cref{al_abs:eq:regret_eluder_cond} leads to the following inequality on $w_{i_t} (x_{i_t})$: 
$$
w_{i_t}(x_{i_t}) \leq \sqrt{\frac{17 \beta \cdot \mfe_{f^{\star}}(\cF, \gamma/ 2)}{t}}
.$$ 
  Taking $\zeta = \gamma$ in \cref{al_abs:eq:regret_eluder_cond} leads to the following inequality on $\tau$:
$$
\tau \leq \frac{17 \beta}{\gamma^2} \cdot \mfe_{f^{\star}}(\cF, \gamma / 2)
.$$ 
We then have 
\begin{align*}
	\sum_{t=1}^{T}\ind(Q_t = 1) \cdot \ind (w_t > \gamma) \cdot 2 w_t & = 
	\sum_{t=1}^{\tau}  \ind(w_{i_t} > \gamma) \cdot 2 w_{i_t}(x_{i_t}) \\
	& \leq 2 \,\sum_{t=1}^{\tau} \sqrt{\frac{17 \beta \cdot \mfe_{f^{\star}}(\cF, \gamma/ 2)}{t}}\\
	& \leq \sqrt{34 \beta \cdot \mfe_{f^{\star}}(\cF, \gamma / 2) \cdot \tau}	\\
	& \leq \frac{17 \sqrt{2} \beta}{\gamma} \cdot \mfe_{f^{\star}}(\cF, \gamma / 2).
\end{align*}
\end{proof}

\begin{lemma}
  \label{al_abs:lm:wh_a_ineq}
  We have 
  $$0 \leq \E \brk{\ell_t(a_t) - \ell_t(\wh a_t^\star) \mid \wb \mfF_{t-1}} \leq 1,$$ and
  $$\E \brk{\ell_t(\wh a_t^\star) - \ell_t(a_t^\star) \mid \wb \mfF_{t-1}} \leq 0.$$
\end{lemma}
\begin{proof}
  By construction, we have $\wh a_t^\star = \bot$ if $a_t = \bot$, and $\wh a_t^\star = a_t^\star$ otherwise.
  Similar to the analysis in \cref{al_abs:lm:regret_no_query_constant}, we have
\begin{align*}
  & \E  \sq*{\ell_t(a_t) - \ell_t(\wh a_t^\star) \mid \wb \mfF_{t-1} } \\
     & = \ind \prn[\big]{ \widehat h_t(x_t) \neq \bot} \cdot \prn[\big]{\P_{y_t\mid x_t} \prn[\big]{y_t \neq \widehat h_t(x_t)} -  \P_{y_t\mid x_t} \prn[\big]{ y_t \neq h^\star(x_t) }}, 
\end{align*}
and 
\begin{align*}
     \E  \sq*{\ell_t(\wh a^\star_t) - \ell_t(a_t^\star) \mid \wb \mfF_{t-1} } 
     = \ind \prn[\big]{ \widehat h_t(x_t) = \bot} \cdot \prn[\big]{ \prn[\big]{{1}/{2} - \gamma}  -  \P_{y_t \mid x_t} \prn[\big]{ y_t \neq h^\star(x_t)}}.
\end{align*}
  The statement $0 \leq \E \brk{\ell_t(a_t) - \ell_t(\wh a_t^\star) \mid \wb \mfF_{t-1}} \leq 1$ follows from the fact that $0 \leq {\P_{y_t\mid x_t} \prn[\big]{y_t \neq \widehat h_t(x_t)} -  \P_{y_t\mid x_t} \prn[\big]{ y_t \neq h^\star(x_t) }} \leq 1$ when $\wh h_t(x_t) \neq \bot$.

  Similar to the analysis in \cref{al_abs:lm:regret_no_query_constant}, we have $\P_{y_t\mid x_t} \prn[\big]{ y_t \neq h^\star(x_t) } \geq \frac{1}{2} - \gamma$ when $\wh h_t(x_t) = \bot$. This leads to $\E \brk{\ell_t(\wh a_t^\star) - \ell_t(a_t^\star) \mid \wb \mfF_{t-1}} \leq 0$.
\end{proof}

\begin{lemma}
\label{al_abs:lm:regret_eluder_cond_wh_a}
 Assume $\mu(x_t) \in [\lcb_t, \ucb_t]$ and $f^{\star}$ is not eliminated across all $t \in [T]$. We have 
\begin{align*}
  \sum_{t=1}^{T} \E \brk{\ell_t(a_t) - \ell_t(\wh a_t^\star) \mid \wb \mfF_{t-1} } \leq 
	\frac{17 \sqrt{2} \beta}{\gamma} \cdot \mfe_{f^{\star}}(\cF, \gamma / 2).
\end{align*}
\end{lemma}

\begin{proof}
  We first consider the event $\crl{Q_t = 0}$. We have 
\begin{align*}
  & \E  \sq*{\ell_t(a_t) - \ell_t(\wh a_t^\star) \mid \wb \mfF_{t-1} } \\
     & = \ind \prn[\big]{ \widehat h_t(x_t) \neq \bot} \cdot \prn[\big]{\P_{y_t\mid x_t} \prn[\big]{y_t \neq \widehat h_t(x_t)} -  \P_{y_t\mid x_t} \prn[\big]{ y_t \neq h^\star(x_t) }}. 
\end{align*}
  When $\wh h_t(x_t) \neq \bot$ and $Q_t = 0$, we must have $\frac{1}{2} \notin \prn{\lcb_t, \ucb_t}$. We then have $\wh h_t(x_t) = h^\star(x_t)$, which leads to $\E  \sq*{\ell_t(a_t) - \ell_t(\wh a_t^\star) \mid \wb \mfF_{t-1} }=0$.

  With the above results on the event $\crl{Q_t = 0}$, the rest of the analysis are the same as the analysis as in \cref{al_abs:lm:regret_eluder_cond} since $\wh a_t^\star = a_t^\star$ under event $\crl{Q_t = 1}$.
\end{proof}
	
\subsection{Proofs and Supporting Results for \cref{al_abs:sec:misspecified}}
\label{al_abs:app:mis}
\subsubsection{Algorithm and Main Results}
\label{al_abs:app:mis_alg}
\begin{algorithm}[]
	\caption{Efficient Active Learning with Abstention under Misspecification}
	\label{al_abs:alg:mis} 
	\renewcommand{\algorithmicrequire}{\textbf{Input:}}
	\renewcommand{\algorithmicensure}{\textbf{Output:}}
	\newcommand{\algorithmicbreak}{\textbf{break}}
    \newcommand{\BREAK}{\STATE \algorithmicbreak}
	\begin{algorithmic}[1]
		\REQUIRE Accuracy level $\epsilon > 0$, abstention parameter $\gamma \in (\eps, 1/2)$ and confidence level $\delta \in (0, 1)$.
		\STATE Define $T \ldef \frac{\pseud(\cF)}{\eps \, \gamma}$, $M \ldef \ceil{\log_2 T}$ and $C_\delta \ldef O \prn{\pseud(\cF) \cdot \log(T /\delta)}$.
		\STATE Define $\tau_m \ldef 2^m$ for $m\geq1$, $\tau_0 = 0$ and $\beta_m \ldef \prn*{M-m+1}\cdot \prn*{2\eps^2 \tau_{M-1} +  2C_\delta  }$.  
		\FOR{epoch $m = 1, 2, \dots, M$}
		\STATE Get $\widehat f_m \ldef \argmin_{f \in \cF} \sum_{t=1}^{\tau_{m-1}} Q_t \paren{f(x_t) - y_t}^2 $.\\
		\hfill \algcommentlight{We use $Q_t \in \crl{0,1}$ to indicate whether the label of  $x_t$ is queried.}
		\STATE (Implicitly) Construct active set of regression function $\cF_m \subseteq \cF$ as 
		\begin{align*}
		    \cF_m \ldef \crl*{ f \in \cF:  \sum_{t = 1}^{\tau_{m-1}} Q_t \prn*{f(x_t) - y_t}^2 \leq \sum_{t = 1}^{\tau_{m-1}} Q_t \paren{\widehat f_m(x_t) - y_t}^2 + \beta_m }. 
		\end{align*}
		\STATE Construct classifier $\wh h_m: \cX \rightarrow \crl{0,1,\bot}$ as
		\begin{align*}
			\wh h_m (x) \ldef 
			\begin{cases}
				\bot, & \text{ if } \brk { \lcb(x;\cF_m), \ucb(x;\cF_m) } \subseteq 
				\brk*{ \frac{1}{2} - \gamma, \frac{1}{2} + \gamma}; \\
        \ind(\wh f_m(x) \geq \frac{1}{2} ) , & \text{ o.w. }
			\end{cases}
		\end{align*}
		and query function $g_m: \cX \rightarrow \crl{0,1}$ as
		\begin{align*}
		g_m(x) \ldef\ind \prn*{ \frac{1}{2} \in \prn{\lcb(x;\cF_m) , \ucb(x;\cF_m) } } \cdot
		\ind \prn{\wh h_m(x) \neq \bot} .
		\end{align*}
		\IF{epoch $m=M$ }
		\STATE \textbf{Return} classifier $\wh h_{M} $.
		\ENDIF
		\FOR{time $t = \tau_{m-1} + 1 ,\ldots , \tau_{m} $} 
		\STATE Observe $x_t \sim \cD_{\cX}$. Set $Q_t \ldef g_m(x_t)$. 
		\IF{$Q_t = 1$}
		\STATE Query the label $y_t$ of $x_t$.
		\ENDIF
		\ENDFOR
		\ENDFOR
	\end{algorithmic}
\end{algorithm}

\cref{al_abs:alg:mis} achieves the guarantees stated in \cref{al_abs:thm:mis}. 
\cref{al_abs:thm:mis} is proved based on supporting lemmas derived in \cref{al_abs:app:mis_lms}. 
Note that, under the condition $\kappa \leq \eps$, we still compete against the Bayes classifier $h^{\star} = h_{f^{\star}}$ in the analysis of Chow's excess error \cref{al_abs:eq:chow_error}.

\thmMis*

\begin{proof}
	We analyze under the good event $\cE$ defined in \cref{al_abs:lm:expected_sq_loss_pseudo}, which holds with probability at least $1-\delta$. Note that all supporting lemmas stated in \cref{al_abs:app:mis_lms} hold true under this event.

We analyze the Chow's excess error of $\wh h_m$, which is measurable with respect to $\mfF_{\tau_{m-1}}$. 
For any $x \in \cX$, if $g_m(x) = 0$, 
\cref{al_abs:lm:regret_no_query_mis} implies that $\exc_{\gamma}(\wh h_m ;x) \leq 2 \kappa $. 
If $g_m(x)= 1$, we know that $\wh h_m(x) \neq \bot$ and $\frac{1}{2} \in (\lcb(x;\cF_m),\ucb(x;\cF_m))$. 
Since $\wb f \in \cF_m$ by \cref{al_abs:lm:set_f_mis} and $\sup_{x \in \cX} \abs{ \wb f(x) - f^{\star}(x)} \leq \kappa$ by assumption. 
The error incurred in this case is upper bounded by 
\begin{align*}
	\exc_{\gamma}(\wh h_m; x) 
	& \leq 2 \abs{ f^{\star}(x)- 1 /2}\\
	& \leq 2\kappa + 2 \abs{ \wb f(x)- 1 /2}\\
	& \leq 2\kappa + 2 w(x;\cF_m).
\end{align*}
Combining these two cases together, we have 
\begin{align*}
	\exc_{\gamma}( \wh h_m) \leq 2 \kappa +  2 \E_{x \sim \cD_\cX} \brk{ \ind(g_m(x) = 1) \cdot w(x;\cF_m)}.	
\end{align*}
Take $m=M$ and apply \cref{al_abs:lm:per_round_regret_dis_coeff_mis} leads to the following guarantee.
\begin{align*}
	\exc_{\gamma}( \wh h_M)
	& \leq  2 \kappa + { \frac{ 72 \beta_M}{\tau_{M-1} \gamma} \cdot \theta^{\val}_{\wb  f}\prn*{\cF, \gamma/2, \sqrt{\beta_M/ \tau_{M-1}}}}\\
	& \leq 2 \kappa +  O \prn*{ \frac{\eps^2}{\gamma} + \frac{ \pseud(\cF) \cdot \log ( T / \delta)}{T \, \gamma}} \cdot \theta^{\val}_{ \wb f}\prn*{\cF, \gamma/2, \sqrt{C_\delta/T}} \\
	& = O \prn*{ \eps \cdot \wb \theta \cdot \log \prn*{ \frac{\pseud(\cF) }{\eps \, \gamma \, \delta}}},
\end{align*}
where we 
take $\wb \theta \ldef \sup_{\iota > 0}\theta^{\val}_{\wb f} (\cF, \gamma / 2, \iota)$ as an upper bound of $\theta^{\val}_{\wb f} (\cF, \gamma / 2, \sqrt{C_\delta / T})$,
and use the fact that $T = {\frac{\pseud(\cF) }{\eps \, \gamma}}$ and the assumptions that $\kappa \leq \eps < \gamma$.

We now analyze the label complexity (note that the sampling process of \cref{al_abs:alg:mis} stops at time $t = \tau_{M-1}$).
Note that $\E \brk{\ind(Q_t = 1) \mid \mfF_{t-1}} = \E_{x\sim\cD_\cX} \brk{ \ind(g_m(x) = 1) }$ for any epoch $m \geq 2$ and time step $t$ within epoch $m$. 
Combining \cref{al_abs:lm:martingale_two_sides} with \cref{al_abs:lm:conf_width_dis_coeff_mis} leads to
    \begin{align*}
      & \sum_{t=1}^{\tau_{M-1}} \ind(Q_t = 1)\\
      & \leq \frac{3}{2} \sum_{t=1}^{\tau_{M-1}} \E \sq{\ind(Q_t = 1) \mid \mfF_{t-1}} + 4 \log \delta^{-1}\\
        & \leq 3 + \frac{3}{2}\sum_{m=2}^{M-1}\frac{(\tau_m - \tau_{m-1}) \cdot 36 \beta_m}{{\tau_{m-1}} \, \gamma^2} \cdot \theta^{\val}_{\wb f}\prn*{\cF, \gamma/2, \sqrt{\beta_m/\tau_{m-1}}}  + 4 \log \delta^{-1} \\
        & \leq 3 + 48 \sum_{m=2}^{M-1}\frac{\beta_m}{ \gamma^2} \cdot \theta^{\val}_{\wb f}\prn*{\cF, \gamma/2, \sqrt{\beta_m/\tau_{m-1}}}  + 4 \log \delta^{-1} \\
	& \leq 3 + 4 \log \delta^{-1} + O \prn*{ \frac{M^2\cdot \eps^2 \cdot T}{\gamma^2} + \frac{M^2 \cdot C_\delta }{\gamma^2}}
	\cdot \theta^{\val}_{\wb f}\prn*{\cF, \gamma/2, \sqrt{C_\delta/T }} \\ 
	& = O \prn*{ \frac{\wb \theta \, \pseud(\cF)}{\gamma^2} \cdot \prn*{\log \prn*{ \frac{\pseud(\cF)}{\eps\, \gamma }}}^{2} \cdot 
	\log \prn*{ \frac{ \pseud(\cF) }{\eps \,\gamma \,  \delta}}} 
    \end{align*}
    with probability at least $1-2\delta$ (due to an additional application of \cref{al_abs:lm:martingale_two_sides}); where we use the fact that $T = {\frac{\pseud(\cF) }{\eps \, \gamma}}$ and the assumptions that $\kappa \leq \eps < \gamma$ as before.
\end{proof}

\begin{restatable}{theorem}{thmMisEfficient}
	\label{al_abs:thm:mis_efficient}
	\cref{al_abs:alg:mis} can be efficiently implemented via the regression oracle and enjoys the same theoretical guarantees stated in \cref{al_abs:thm:mis}.
	The number of oracle calls needed is $\wt O(\frac{\pseud(\cF) }{\eps \, \gamma^{3}})$ for a general set of regression functions $\cF$, and $\wt O(\frac{\pseud(\cF)}{\eps \, \gamma})$ when $\cF$ is convex and closed under pointwise convergence.
	The per-example inference time of the learned $\wh h_{M}$ is $\wt O ( \frac{1}{\gamma^2} \log^2 \prn{\frac{\pseud(\cF)}{\eps }})$ for general $\cF$, and $\wt O ( \log \frac{1}{\gamma}) $ when $\cF$ is convex and closed under pointwise convergence.
\end{restatable}

\begin{proof}
Note that classifier $\wh h_m$ and query function $q_m$ in \cref{al_abs:alg:mis} are constructed in the way as the ones in \cref{al_abs:alg:epoch},
Thus, \cref{al_abs:alg:mis} can be efficiently implemented in the same way as discussed in \cref{al_abs:thm:epoch_efficient}, and enjoys the same per-round computational complexities.
The total computational complexity is then achieved by multiplying the per-round computational complexity by $T = {\frac{\pseud(\cF) }{\eps \, \gamma}}$.
\end{proof}

\subsubsection{Discussion on $\kappa \leq \eps$}
\label{al_abs:app:mis_partial}

We provide guarantees (in \cref{al_abs:thm:mis}) when $\kappa \leq \eps$ since the learned classifier suffers from an additive  $\kappa$ term in the excess error, as shown in the proof of \cref{al_abs:thm:mis}.
We next give preliminary discussions on this issue by relating active learning with to a (specific) regret minimization problem and connecting to existing lower bound in the literature.
More specifically, we consider the perspective and notations discussed in \cref{al_abs:app:constant_regret}.
Fix any epoch $m \geq 2$ and time step $t$ within epoch $m$. 
We have 
\begin{align*}
  \reg_t &= \E \sq{\ell_t(a_t) - \ell_t(a^\star_t)  \mid  \mfF_{t-1}}\\
    &= \err_{\gamma}(\wh h_m) - \err(h^\star)\\
    & = \exc_{\gamma}(\wh h_m) \\
    &= \wt O \prn*{ \kappa + \frac{\wb \theta}{2^m \, \gamma} },
\end{align*}
where the bound comes from similar analysis as in the proof of \cref{al_abs:thm:mis}. Summing the instantaneous regret over $T$ rounds, we have 
\begin{align*}
    \reg(T) & = \sum_{t=1}^T \reg_t\\
    & \leq 2 + \sum_{m=2}^M (\tau_m - \tau_{m-1}) \cdot \exc_{\gamma}(\wh h_m)\\
    & \leq \wt O \prn*{\kappa \cdot T + \frac{\wb \theta}{\gamma}}.
\end{align*}
The above bound indicates an additive regret term scales as $\kappa \cdot T$. On the other hand, 
it is known that an additive $\kappa \cdot T$ regret is in general unavoidable in linear bandits under model misspecification \citep{lattimore2020learning}.
This connection partially explains/justifies why we only provide guarantee for \cref{al_abs:thm:mis} under $\kappa \leq \eps$.

There are, however, many differences between the two learning problems. We list some distinctions below.
\begin{enumerate}
    \item The regret minimization problem considered in \cref{al_abs:app:constant_regret} only takes three actions $\cA = \crl{0, 1,\bot}$, yet the lower bound in linear bandits is established with a large action set \citep{lattimore2020learning};
    \item A standard contextual bandit problem will observe loss (with respect to the pulled action) at each step $t \in [T]$, however, the active learning problem will only observe (full) feedback at time steps when a query is issued, i.e., $\crl{t \in [T]: Q_t =1}$.
\end{enumerate}
We leave a comprehensive study of the problem for feature work.

\subsubsection{Supporting Lemmas}
\label{al_abs:app:mis_lms}

We use the same notations defined in \cref{al_abs:app:epoch}, except $\wh h_m$, $g_m$ and $\beta_m$ are defined differently.
We adapt the proofs \cref{al_abs:thm:epoch} (in \cref{al_abs:app:epoch}) to deal with model misspecification.

Note that although we do not have $f^{\star}\in \cF	$ anymore, one can still define random variables of the form $M_t(f)$, and guarantees in \cref{al_abs:lm:expected_sq_loss_pseudo} still hold.
We use $\cE$ to denote the good event considered in \cref{al_abs:lm:expected_sq_loss_pseudo}, we analyze under this event through out the rest of this section.
We also only analyze under the assumption of \cref{al_abs:thm:mis}, i.e., $\kappa^2 \leq \eps$.
\begin{lemma}
	\label{al_abs:lm:f_mis}
	Fix any epoch $m \in [M]$. We have 
	$$
	\wh R_m(\wb f) \leq \wh R_m(f^{\star}) + \frac{3 }{2} \cdot \kappa^2 \tau_{m-1} + C_\delta
	,$$ 
	where $C_\delta \ldef 8 \log \prn*{ \frac{\abs{\cF}\cdot T^2}{\delta}}$.
\end{lemma}
\begin{proof}
	From \cref{al_abs:lm:expected_sq_loss_pseudo} we know that 
	\begin{align*}
		\wh R_m(\wb f) - \wh R_m(f^{\star}) & \leq \sum_{t=1}^{\tau_{m-1}} \frac{3}{2} \cdot
		\E_{t} \brk*{ Q_t\prn*{\wb f(x_t) - f^{\star}(x_t)}^2} + C_\delta \\
		& \leq \frac{3 }{2} \cdot \kappa^2 \tau_{m-1} + C_\delta,
	\end{align*}
	where we use the fact that $\E_t \brk{y_t \mid x_t} = f^{\star}(x_t)$ 
	(and thus $\E_t \brk{ M_t(\wb f)} = \E_{t}\brk{Q_t \prn{\wb f(x_t) - f^{\star}(x_t)}^2}$) on the first line; 
	and use the fact $\sup_{x} \abs{\wb f(x) - f^{\star}(x)} \leq \kappa$ on the second line.
\end{proof}
\begin{lemma}
\label{al_abs:lm:set_f_mis}
The followings hold true:
\begin{enumerate}
	\item $\wb f\in \cF_m$ for any $m \in [M]$.
	\item $\sum_{t=1}^{\tau_{m-1}} \E_t \brk{M_t(f)} \leq 4 \beta_m $ for any $f \in \cF_m$. 
	\item $ \sum_{t=1}^{\tau_{m-1}} \E \brk{ Q_t(x_t) \prn{ f(x_t) - \wb f(x_t) }^2} \leq 9 \beta_m$ for any $f \in \cF_m$.
	\item $\cF_{m+1} \subseteq \cF_m$ for any $m \in [M-1]$.
\end{enumerate}
\end{lemma}
\begin{proof}
\begin{enumerate}
	\item Fix any epoch $m \in [M]$. By \cref{al_abs:lm:expected_sq_loss_pseudo}, we have 
$\wh R_m (f^\star) \leq \wh R_m(f) + C_\delta /2 $ for any $f \in \cF$.
Combining this with \cref{al_abs:lm:f_mis} leads to 
\begin{align*}
	\wh R_m(\wb f) 
	& \leq \wh R_m(f) + \frac{3}{2}\cdot \prn*{ \kappa^2 \tau_{m-1} + C_\delta}\\
	& \leq \wh R_m(f) + \beta_m,
\end{align*}
for any $f \in \cF$, where the second line comes from the definition of $\beta_m$ (recall that we have $\kappa\leq \eps$ by assumption).
We thus have $\wb f \in \cF_m$ for any $m \in [M]$.
\item  Fix any $f \in \cF_m$. With \cref{al_abs:lm:expected_sq_loss_pseudo}, we have 
	\begin{align*}
		\sum_{t=1}^{\tau_{m-1}} \E_t [M_t(f)] & \leq 2 \sum_{t=1}^{\tau_{m-1}} M_t(f) + C_\delta \\
			& = 2 \wh R_{m }(f) - 2\wh R_{m}(f^{\star}) + C_\delta \\
			& \leq 2 \wh R_{m }(f) - 2\wh R_{m}(\wb f)+ 3 \kappa^2 \tau_{m-1} + 3 C_\delta \\
			& \leq 2 \wh R_{m }(f) - 2\wh R_{m}(\wh f_m)+ 3 \kappa^2 \tau_{m-1} + 3 C_\delta \\
			& \leq 2 \beta_m + 3 \kappa^2 \tau_{m-1} + 3 C_\delta \\
			& \leq 4 \beta_m , 
	\end{align*}
	where the third line comes from \cref{al_abs:lm:f_mis}; the fourth line comes from the fact that $\wh f_m$ is the minimizer of $\wh R_{m} (\cdot) $; and the fifth line comes from the fact that $f \in \cF_m$.
\item  Fix any $f \in \cF_m$. With \cref{al_abs:lm:expected_sq_loss_pseudo}, we have 
	\begin{align*}
    &\sum_{t=1}^{\tau_{m-1}} \E_t \brk{ Q_t(x_t) \prn{ f(x_t) - \wb f(x_t) }^2}\\
		& = \sum_{t=1}^{\tau_{m-1}} \E_t \brk{ Q_t(x_t) \prn{ (f(x_t) - f^{\star}(x_t)) + 
		( f^{\star}(x_t) - \wb f(x_t)) }^2} \\
		& \leq 2 \sum_{t=1}^{\tau_{m-1}} \E_t \brk{ Q_t(x_t) \prn{ f(x_t) -  f^{\star}(x_t) }^2} + 2 \tau_{m-1} \kappa^2\\
		&  = 2 \sum_{t=1}^{\tau_{m-1}} \E_t [M_t(f)] + 2 \tau_{m-1} \kappa^2  \\
		& \leq 8 \beta_m + 2 \tau_{m-1} \kappa^2\\
		& \leq 9 \beta_m,
	\end{align*}
	where we use $\prn{a+b}^2 \leq a^2 + b^2$ on the second line; and use statement $2$ on the fourth line.
	\item Fix any $f \in \cF_{m+1}$. We have 
	\begin{align*}
    &\wh R_{m} (f) - \wh R_{m} (\wh f_m)\\
    & \leq   
		\wh R_{m} (f) - \wh R_{m} (f^{\star}) + \frac{C_\delta}{2}\\
		& = \wh R_{m+1}(f) - \wh R_{m+1}(f^{\star}) 
		- \sum_{t=\tau_{m-1}+1}^{\tau_{m}} M_t(f) + \frac{C_\delta}{2}\\
		& \leq \wh R_{m+1}(f) - \wh R_{m+1} (\wb f) + \frac{3}{2} \kappa^2 \tau_m + C_\delta
		- \sum_{t=\tau_{m-1}+1}^{\tau_{m}} \E_t [M_t(f)] /2 + {C_\delta}\\
		& \leq \wh R_{m+1}(f) - \wh R_{m+1} (\wh f_{m+1}) + \frac{3}{2} \kappa^2 \tau_m + 2C_\delta\\
		& \leq \beta_{m+1} + \frac{3}{2} \kappa^2 \tau_m + 2C_\delta\\
		& \leq \beta_m,
	\end{align*}	
	where the first line comes from \cref{al_abs:lm:expected_sq_loss_pseudo}; the third line comes from \cref{al_abs:lm:f_mis} and \cref{al_abs:lm:expected_sq_loss_pseudo}; the fourth line comes from the fact that $\wh f_{m+1}$ is the minimizer with respect to $\wh R_{m+1}$ and \cref{al_abs:lm:expected_sq_loss_pseudo}; the last line comes from the definition of $\beta_m$.
\end{enumerate}
\end{proof}
Since the classifier $\wh h_m$ and query function $g_m$ are defined in the same way as in \cref{al_abs:alg:epoch}, \cref{al_abs:lm:query_implies_width} holds true for \cref{al_abs:alg:mis} as well.
As a result of that, \cref{al_abs:lm:conf_width_dis_coeff} and \cref{al_abs:lm:per_round_regret_dis_coeff} hold true with minor modifications. We present the modified versions below, whose proofs follow similar steps as in \cref{al_abs:lm:conf_width_dis_coeff} and \cref{al_abs:lm:per_round_regret_dis_coeff} but replace $f^{\star}$ with $\wh f$ (and thus using concentration results derived in \cref{al_abs:lm:set_f_mis}).
\begin{lemma}
    \label{al_abs:lm:conf_width_dis_coeff_mis}
 Fix any epoch $m \geq 2$. We have 
	\begin{align*}
    \E_{x \sim \cD_\cX} \sq{\ind (g_m(x)= 1)} \leq \frac{36 \beta_m}{{\tau_{m-1}} \, \gamma^2} \cdot \theta^{\val}_{\wb f}\prn*{\cF, \gamma/2, \sqrt{\beta_m/\tau_{m-1}}}.
	\end{align*}
\end{lemma}
\begin{lemma}
    \label{al_abs:lm:per_round_regret_dis_coeff_mis}
 Fix any epoch $m \geq 2$. We have 
    \begin{align*}
    	\E_{x \sim \cD_\cX} \sq{\ind(g_m(x) = 1)\cdot w(x;\cF_m)} \leq { \frac{36 \beta_m}{\tau_{m-1} \gamma} \cdot \theta^{\val}_{\wb f}\prn*{\cF, \gamma/2, \sqrt{\beta_m/\tau_{m-1}}}}.
    \end{align*}
\end{lemma}
\begin{lemma}
\label{al_abs:lm:regret_no_query_mis}
Fix any $m \in [M]$. We have $\exc_{\gamma}(\wh h_m ;x) \leq 2 \kappa$ if $g_m(x) = 0$.
\end{lemma}
\begin{proof}
	Recall that
\begin{align*}
	\exc_{\gamma}( \wh h;x) & =  \nonumber
      \ind \prn[\big]{ \widehat h(x) \neq \bot} \cdot \prn[\big]{\P_{y \mid x} \prn[\big]{y \neq \widehat h(x)} -  \P_{y \mid x} \prn[\big]{ y \neq h^\star(x) }} \nonumber \\
    & \quad + \ind \prn[\big]{ \widehat h(x) = \bot} \cdot \prn[\big]{ \prn[\big]{{1}/{2} - \gamma}  -  \P_{y\mid x} \prn[\big]{ y \neq h^\star(x) }} .
\end{align*}
We now analyze the event $\curly*{g_m(x)= 0}$ in two cases. 

\textbf{Case 1: ${\widehat h_m(x) = \bot} $.} 

Since $\wb f(x) \in [\lcb(x;\cF_m), \ucb(x;\cF_m)]$ by \cref{al_abs:lm:set_f_mis}, we know that $\eta(x) = f^{\star}(x) \in \sq{ \frac{1}{2} - \gamma - \kappa, \frac{1}{2} + \gamma + \kappa}$ and thus $\P_{y} \prn[\big]{ y\neq h^\star(x) } \geq \frac{1}{2} - \gamma-\kappa$. 
As a result, we have $\exc_{\gamma}(\wh h_m;x) \leq \kappa $.

\textbf{Case 2: ${\widehat h_m(x) \neq \bot}$ but ${\frac{1}{2} \notin (\lcb(x;\cF_m), \ucb(x;\cF_m))} $.} 

We clearly have $\exc_{\gamma}(\wh h_m;x) \leq 0$ if $\widehat h_m (x) = h^\star(x)$. 
Now consider the case when $\widehat h_m (x) \neq h^\star(x)$. 
Since $\wb f(x) \in [\lcb(x;\cF_m), \ucb(x;\cF_m)]$ and $\abs{ \wb f(x) - f^{\star}(x)} \leq \kappa$, we must have $\abs*{ f^{\star}(x) - 1 /2  } \leq \kappa$ in that case, which leads to $\exc_{\gamma}(\wh h_m;x) \leq 2 \abs{ f^{\star}(x)-1 / 2} \leq 2 \kappa$.
\end{proof}

\chapter{Active Learning with Neural Networks}
\label{chapter:active:deep}

	Deep neural networks have great representation power, but typically require large numbers of training examples. This motivates deep active learning methods that can significantly reduce the amount of labeled training data. Empirical successes of deep active learning have been recently reported in the literature, however, rigorous label complexity guarantees of deep active learning have remained elusive. This constitutes a significant gap between theory and practice. This chapter tackles this gap by providing the first near-optimal label complexity guarantees for deep active learning. The key insight is to study deep active learning from the nonparametric classification perspective. Under standard low noise conditions, we show that active learning with neural networks can provably achieve the minimax label complexity, up to disagreement coefficient and other logarithmic terms. When equipped with an abstention option, we further develop an efficient deep active learning algorithm that achieves $\mathsf{polylog}(\frac{1}{\epsilon})$ label complexity, without any low noise assumptions.  We also provide extensions of our results beyond the commonly studied Sobolev/H\"older spaces and develop label complexity guarantees for learning in Radon $\mathsf{BV}^2$ spaces, which have recently been proposed as natural function spaces associated with neural networks.

\section{Introduction}
\label{al_deep:sec:intro}

We study active learning with neural network hypothesis classes, sometimes known as \emph{deep active learning}.
Active learning agent proceeds by selecting the most informative data points to label: The goal of active learning is to achieve the same accuracy achievable by passive learning, but with much fewer label queries \citep{settles2009active, hanneke2014theory}.
When the hypothesis class is a set of neural networks, the learner further benefits from the representation power of deep neural networks, which has driven the successes of passive learning in the past decade \citep{krizhevsky2012imagenet, lecun2015deep}.
With these added benefits,
deep active learning has become a popular research area, with empirical successes observed in many recent papers \citep{sener2018active, ash2019deep, citovsky2021batch, ash2021gone, kothawade2021similar, emam2021active, ren2021survey}.
However, due to the difficulty of analyzing a set of neural networks, rigorous label complexity guarantees for deep active learning have remained largely elusive. 

To the best of our knowledge, there are only two papers \citep{karzand2020maximin, wang2021neural} that have made the attempts at theoretically quantifying active learning gains with neural networks.
While insightful views are provided, these two works have their own limitations. The guarantees provided in \citet{karzand2020maximin} only work in the $1d$ case where data points are uniformly sampled from $[0,1]$ and labeled by a well-seperated piece-wise constant function in a noise-free way (i.e., without any labeling noise).
\citet{wang2021neural} study deep active learning by linearizing the neural network at its random initialization and then analyzing it as a linear function; moreover, as the authors agree, their error bounds and label complexity guarantees can in fact be \emph{vacuous} in certain cases.
Thus, it's fair to say that up to now researchers have not identified cases where deep active learning are provably near minimax optimal (or even with provably non-vacuous guarantees), which constitutes a significant gap between theory and practice.

In this chapter, we bridge this gap by providing the first near-optimal label complexity guarantees for deep active learning.
We obtain insights from the nonparametric setting where the conditional probability (of taking a label of $1$) is assumed to be a smooth function \citep{tsybakov2004optimal, audibert2007fast}.
Previous nonparametric active learning algorithms proceed by partitioning the action space into exponentially many sub-regions (e.g., partitioning the unit cube $[0,1]^{d}$ into $\eps^{-d}$ sub-cubes each with volume $\eps^{d}$), and then conducting local mean (or some higher-order statistics) estimation  within each sub-region \citep{castro2008minimax, minsker2012plug, locatelli2017adaptivity, locatelli2018adaptive, shekhar2021active, kpotufe2021nuances}.
We show that, with an appropriately chosen set of neural networks that \emph{globally} approximates the smooth regression function, one can in fact recover the minimax label complexity for active learning, up to disagreement coefficient \citep{hanneke2007bound, hanneke2014theory} and other logarithmic factors.
Our results are established by (i) identifying the ``right tools'' to study neural networks (ranging from approximation results \citep{yarotsky2017error,yarotsky2018optimal} to complexity measure of neural networks \citep{bartlett2019nearly}), and (ii) developing novel extensions of agnostic active learning algorithms \citep{balcan2006agnostic, hanneke2007bound, hanneke2014theory} to work with a set of neural networks.

While matching the minimax label complexity in nonparametric active learning is existing, such minimax results scale as $\Theta(\poly(\frac{1}{\eps}))$ \citep{castro2008minimax, locatelli2017adaptivity} and do not resemble what is practically observed in deep active learning: A fairly accurate neural network classifier can be obtained by training with only a few labeled data points.
Inspired by recent results in \emph{parametric} active learning with abstention \citep{puchkin2021exponential, zhu2022efficient}, we develop an oracle-efficient algorithm showing that deep active learning provably achieves $\polylog(\frac{1}{\eps})$ label complexity when equipped with an abstention option \citep{chow1970optimum}.
Our algorithm not only achieves an exponential saving in label complexity (\emph{without any low noise assumptions}), but is also highly practical: 
In real-world scenarios such as medical imaging, it makes more sense for the classifier to abstain from making prediction on hard examples (e.g., those that are close to the boundary), and ask medical experts to make the judgments.

\subsection{Problem Setting}
\label{al_deep:sec:setting}

Let $\cX$ denote the instance space and $\cY$ denote the label space. 
We focus on the binary classification problem where $\cY \ldef \curly*{0, 1}$. The joint distribution over $\cX \times \cY$ is denoted as $\cD_{\cX \cY}$. 
We use $\cD_{\cX}$ to denote the marginal distribution over the instance space $\cX$, and use $\cD_{\cY \vert x}$ to denote the conditional distribution of $\cY$ with respect to any $x \in \cX$.   
We consider the standard active learning setup where $x \sim \cD_\cX$ but its label  $y \sim \cD_{\cY \vert x}$ is only observed after issuing a label query.
We define $\eta(x) \ldef \P_{y \sim \cD_{\cY \vert x}} (y = 1)$ as the conditional probability of taking a label of $1$.
The Bayes optimal classifier $h^{\star}$ can thus be expressed as $h^{\star}(x) \ldef \ind(\eta(x) \geq 1/2)$.
For any classifier $h: \cX \rightarrow \cY$, 
its (standard) error is calculated as $\err(h) \ldef \P_{(x,y) \sim \cD_{\cX \cY}} (h(x) \neq y)$;  
and its (standard) excess error is defined as 
$\exc(h) \ldef \err(h) - \err(h^{\star})$.
Our goal is to learn an accurate classifier with a small number of label querying.

\paragraph{The nonparametric setting}
We consider the nonparametric setting where the conditional probability $\eta$ is characterized by a smooth function.
Fix any $\alpha \in \N_+$, the \emph{Sobolev norm} of a function  $f: \cX \rightarrow \R$ is defined as 
\linebreak
$\nrm{f}_{\cW^{\alpha, \infty}} \ldef \max_{\wb \alpha, \abs{\wb \alpha} \leq \alpha} \esup_{x \in \cX} \abs{\wderi^{\alpha}f (x)}$,
where $\alpha = \prn{ \alpha_1, \ldots, \alpha_d}$, $\abs{\alpha} = \sum_{i=1}^{d} \alpha_i$ and $\wderi^{\alpha}f$ denotes the standard $\alpha$-th weak derivative of $f$.
The unit ball in the Sobolev space is defined as 
$
	\cW^{\alpha, \infty}_1 (\cX) \ldef \crl{ f : \nrm{f}_{\cW^{\alpha, \infty}} \leq 1}.
$
Following the convention of nonparametric active learning \citep{castro2008minimax, minsker2012plug, locatelli2017adaptivity, locatelli2018adaptive, shekhar2021active, kpotufe2021nuances}, we assume $\cX = [0,1]^{d}$ and $\eta \in \cW^{\alpha, \infty}_1(\cX)$ (except in \cref{al_deep:sec:extension}).

\paragraph{Neural networks}
We consider \emph{feedforward neural networks} with Rectified Linear Unit (ReLU) activation function, which is defined as $\relu(x) \ldef \max \crl{x, 0}$. Each neural network $f_\dnn: \cX \rightarrow \R$ consists of several input units (which corresponds to the covariates of $x \in \cX$), one output unit (which corresponds to the prediction in $\R$), and multiple hidden computational units.
Each hidden computational unit takes inputs $\crl{\wb x_i}_{i=1}^{N}$ (which are outputs from previous layers) and perform the computation $\relu(\sum_{i=1}^{N} w_i \wb x_i + b)$ with \emph{adjustable} parameters $\crl{w_i}_{i=1}^{N}$ and $b$; 
the output unit performs the same operation, but without the ReLU nonlinearity.  
We use $W$ to denote the total number of parameters of a neural network, and  $L$ to denote the depth of the neural network.

\subsection{Contributions and Organization}
\label{al_deep:sec:contributions}

Neural networks are known to be universal approximators \citep{cybenko1989approximation, hornik1991approximation}. 
In this chapter, we argue that, in both passive and active regimes, 
the universal approximatability makes neural networks ``universal classifiers'' for classification problems: With an appropriately chosen set of neural networks, one can recover known minimax rates (up to disagreement coefficients in the active setting) in the rich nonparametric regimes.\footnote{
As a byproduct, our results also provide a new perspective on nonparametric active learning through the lens of neural network approximations. Nonparametric active learning was previously tackled through space partitioning and local estimations over exponentially many sub-regions \citep{castro2008minimax, minsker2012plug, locatelli2017adaptivity, locatelli2018adaptive, shekhar2021active, kpotufe2021nuances}.}
We provide informal statements of our main results in the sequel, 
with detailed statements and associated definitions/algorithms deferred to later sections.

In \cref{al_deep:sec:noise}, we analyze the label complexity of deep active learning under the standard Tsybakov noise condition with smoothness parameter $\beta \geq 0$ \citep{tsybakov2004optimal}. Let $\cH_\dnn$ be an appropriately chosen set of neural network classifiers and denote $\theta_{\cH_\dnn}(\eps)$ as the disagreement coefficient \citep{hanneke2007bound, hanneke2014theory} at level $\eps$. 
We develop the following label complexity guarantees for deep active learning.

\begin{theorem}
[Informal]
\label{al_deep:thm:noise_informal}
There exists an algorithm that returns a neural network classifier $\wh h \in \cH_\dnn$ with excess error $\wt O(\eps)$ after querying $\wt O \prn{ \theta_{\cH_\dnn} \prn{ \eps^{\frac{\beta}{1 +  \beta}}} \cdot \eps^{- \frac{d + 2 \alpha}{\alpha + \alpha\beta}} }$ labels.
\end{theorem}

The label complexity presented in \cref{al_deep:thm:noise_informal} matches the active learning lower bound $\Omega(\eps^{- \frac{d+2\alpha}{\alpha + \alpha \beta}})$ \citep{locatelli2017adaptivity} up to the dependence on the disagreement coefficient (and other logarithmic factors).
Since $\theta_{\cH_\dnn}(\eps) \leq \eps^{-1}$ by definition, the label complexity presented in \cref{al_deep:thm:noise_informal} is never worse than the passive learning rates $\wt \Theta(\eps^{-\frac{d+2\alpha + \alpha \beta}{\alpha + \alpha \beta}})$ \citep{audibert2007fast}. 
We also discover conditions under which the disagreement coefficient with respect to a set of neural network classifiers can be properly bounded, i.e., $\theta_{\cH_\dnn} (\eps) = o(\eps^{-1})$ (implying strict improvement over passive learning) and $\theta_{\cH_\dnn} (\eps) = o(1)$ (implying matching active learning lower bound).

In \cref{al_deep:sec:abstention}, we develop label complexity guarantees for deep active learning when an additional abstention option is allowed \citep{chow1970optimum, puchkin2021exponential, zhu2022efficient}.
Suppose a cost (e.g. $0.49$) that is marginally smaller than random guessing (which has expected cost $0.5$) is incurred whenever the classifier abstains from making a predication, we develop the following label complexity guarantees for deep active learning. 
\looseness=-1

\begin{theorem}
	[Informal]
	\label{al_deep:thm:abs_informal}
	There exists an efficient algorithm that constructs a neural network classifier $\wh h_\dnn$ with Chow's excess error $\wt O(\eps)$ after querying $\polylog(\frac{1}{\eps})$ labels.
\end{theorem}
The above $\polylog(\frac{1}{\eps})$ label complexity bound is achieved \emph{without any low noise assumptions}.
Such exponential label savings theoretically justify the great empirical performances of deep active learning observed in practice (e.g., in \citet{sener2018active}): 
It suffices to label a few data points to achieve a high accuracy level.
Moreover, apart from an initialization step, our algorithm (\cref{al_deep:alg:abs}) developed for \cref{al_deep:thm:abs_informal}  can be \emph{efficiently} implemented in $\wt O(\eps^{-1})$ time, given a convex loss regression oracle over an appropriately chosen set of neural networks; in practice, the regression oracle can be approximated by running stochastic gradient descent.
\looseness=-1

\paragraph{Technical contributions}
Besides identifying the ``right tools'' (ranging from approximation results \citep{yarotsky2017error,yarotsky2018optimal} to complexity analyses \citep{bartlett2019nearly}) to analyze deep active learning, our theoretical guarantees are empowered by novel extensions of active learning algorithms \emph{under neural network approximations}. In particular, we deal with approximation error in active learning under Tsybakov noise, and identify conditions that greatly relax the approximation requirement in the learning with abstention setup; we also analyze the disagreement coefficient, both classifier-based and value function-based, with a set of neural networks.These analyses together lead to our main results for deep active learning (e.g., \cref{al_deep:thm:noise_informal} and \cref{al_deep:thm:abs_informal}).
More generally, we establish a bridge between approximation theory and active learning; we provide these general guarantees in \cref{al_deep:app:RCAL_gen} (under Tsybakov noise) and \cref{al_deep:app:abs_gen} (with the abstention option), which can be of independent interests.
Benefited from these generic algorithms and guarantees, in \cref{al_deep:sec:extension}, we extend our results into learning smooth functions in the Radon $\BV^2$ space \citep{ongie2020function, parhi2021banach, parhi2022kinds, parhi2022near, unser2022ridges}, which is recently proposed as a natural space to analyze neural networks.
\looseness=-1

\subsection{Additional Related Work}
\label{al_deep:sec:related}

Active learning concerns about learning accurate classifiers without extensive human labeling.
One of the earliest work of active learning dates back to the \textCAL algorithm proposed by \citet{cohn1994improving}, which set the cornerstone for \emph{disagreement-based} active learning.
Since then, a long line of work have been developed, either directly working with a set classifier \citep{balcan2006agnostic, hanneke2007bound, dasgupta2007general, beygelzimer2009importance, beygelzimer2010agnostic, huang2015efficient, cortes2019active} or work with a set of regression functions \citep{krishnamurthy2017active,krishnamurthy2019active}.
These work mainly focus on the parametric regime (e.g., learning with a set of linear classifiers), and their label complexities rely on the boundedness of the so-called disagreement coefficient \citep{hanneke2007bound, hanneke2014theory, friedman2009active}.
Active learning in the nonparametric regime has been analyzed in \citet{castro2008minimax, minsker2012plug, locatelli2017adaptivity, locatelli2018adaptive, kpotufe2021nuances}. These algorithms rely on partitioning of the input space $\cX \subseteq [0,1]^{d}$ into exponentially (in dimension) many small cubes, and then conduct local mean (or some higher-order statistics) estimation within each small cube.
\looseness=-1

It is well known that, in the worst case, active learning exhibits no label complexity gains over the passive counterpart \citep{kaariainen2006active}.
To bypass these worst-case scenarios, active learning has been popularly analyzed under the so-called Tsybakov low noise conditions \citep{tsybakov2004optimal}.
Under Tsybakov noise conditions, active learning has been shown to be strictly superior than passive learning in terms of label complexity \citep{castro2008minimax, locatelli2017adaptivity}.
Besides analyzing active learning under favorable low noise assumptions, more recently, researchers consider active learning with an abstention option and analyze its label complexity under Chow's error \citep{chow1970optimum}.
In particular, \citet{puchkin2021exponential, zhu2022efficient} develop active learning algorithms with $\polylog(\frac{1}{\eps})$ label complexity when analyzed under Chow's excess error.
\citet{shekhar2021active} study nonparametric active learning under a different notion of the Chow's excess error, and propose algorithms with $\poly(\frac{1}{\eps})$ label complexity; their algorithms follow similar procedures of those partition-based nonparametric active learning algorithms (e.g., \citet{minsker2012plug, locatelli2017adaptivity}).\looseness=-1

Inspired by the success of deep learning in the passive regime, active learning with neural networks has been extensively explored in recent years 
\citep{sener2018active, ash2019deep, citovsky2021batch, ash2021gone, kothawade2021similar, emam2021active, ren2021survey}.
Great empirical performances are observed in these papers, however, rigorous label complexity guarantees have largely remains elusive (except in \citet{karzand2020maximin, wang2021neural}, with limitations discussed before).
We bridge the gap between practice and theory by providing the first near-optimal label complexity guarantees for deep active learning.
Our results are built upon approximation results of deep neural networks \citep{yarotsky2017error, yarotsky2018optimal, parhi2022near} and VC/pseudo dimension analyses of neural networks with given structures \citep{bartlett2019nearly}.\looseness=-1

\section{Label Complexity of Deep Active Learning}
\label{al_deep:sec:noise}

We analyze the label complexity of deep active learning in this section.
We first introduce the Tsybakov noise condition in \cref{al_deep:sec:tsybakov}, and then identify the ``right tools'' to analyze classification problems with neural network classifiers in \cref{al_deep:sec:noise_passive} (where we also provide passive learning guarantees).
We establish our main active learning guarantees in \cref{al_deep:sec:noise_active}.

\subsection{Tsybakov Noise Condition}
\label{al_deep:sec:tsybakov}
It is well known that active learning exhibits no label complexity gains over the passive counterpart without additional low noise assumptions \citep{kaariainen2006active}.
We next introduce the Tsybokov low noise condition \citep{tsybakov2004optimal}, which has been extensively analyzed in active learning literature.

\begin{definition}[Tsybakov noise]
	\label{al_deep:def:Tsybakov}
	A distribution $\cD_{\cX\cY}$ satisfies the Tsybakov noise condition with parameter $\beta \geq 0$ and a universal constant $c \geq 1$ if, $\forall \tau > 0$,
	\begin{align*}
	\P_{x \sim \cD_\cX} \prn{ \abs{ \eta(x) - 1 / 2} \leq \tau} \leq  c \, \tau^\beta.
	\end{align*}
\end{definition}

The case with $\beta = 0$ corresponds to the general case \emph{without} any low noise conditions, 
where no active learning algorithm can outperform the passive counterpart \citep{audibert2007fast, locatelli2017adaptivity}.
We use $\cP(\alpha, \beta)$ to denote the set of distributions satisfying:
(i) the smoothness conditions introduced in \cref{al_deep:sec:setting} with parameter $\alpha > 0$; 
and (ii) the Tsybakov low noise condition (i.e., \cref{al_deep:def:Tsybakov}) with parameter $\beta \geq 0$. 
We assume 
$\cD_{\cX \cY} \in \cP(\alpha, \beta)$ in the rest of \cref{al_deep:sec:noise}.
As in \citet{castro2008minimax, hanneke2014theory}, we assume the knowledge of noise/smoothness parameters.

\subsection{Approximation and Expressiveness of Neural Networks}
\label{al_deep:sec:noise_passive}

Neural networks are known to be universal approximators \citep{cybenko1989approximation, hornik1991approximation}: For any continuous function $g : \cX \rightarrow \R$ and any error tolerance $\kappa > 0$, there exists a large enough neural network $f_{\dnn}$ such that $\nrm{f_\dnn - g}_{\infty} \ldef \sup_{x \in \cX} \abs{f_{\dnn}(x)-g(x)} \leq \kappa$.
Recently, \emph{non-asympototic} approximation rates by ReLU neural networks have been developed for smooth functions in the Sobolev space, which we restate in the following.\footnote{As in \citet{yarotsky2017error}, we hide constants that are potentially $\alpha$-dependent and $d$-dependent into the Big-Oh notation.} 
\begin{theorem}[\citet{yarotsky2017error}]
	\label{al_deep:thm:approx_sobolev}
	Fix any $\kappa>0$. For any $f^{\star} = \eta \in \cW^{\alpha, \infty}_1([0,1]^{d})$, there exists a neural network $f_\dnn$ with $W = O \prn{\kappa^{- \frac{d}{\alpha}} \log \frac{1}{\kappa} }$ total number of parameters arranged in $L = O ( \log \frac{1}{\kappa})$ layers such that $\nrm{f_\dnn - f^{\star} }_\infty \leq \kappa$. 
\end{theorem}

The architecture of the neural network $f_\dnn$ appearing in the above theorem only depends on the smooth function space $\cW^{\alpha, \infty}_1 \prn{[0,1]^{d}}$, but otherwise is independent of the true regression function $f^{\star}$; also see \citet{yarotsky2017error} for details.
Let $\cF_{\dnn}$ denote the set of neural network \emph{regression functions} with the same architecture. We construct a set of neural network \emph{classifiers} by thresholding the regression function at $\frac{1}{2}$, i.e.,  
$\cH_{\dnn} \ldef \crl{ h_f \ldef \ind(f(x) \geq 1/2): f \in \cF_{\dnn}} $.
The next result concerns about the expressiveness of the neural network classifiers, in terms of a well-known complexity measure: the VC dimension \citep{vapnik1971uniform}.

\begin{theorem}[\citet{bartlett2019nearly}]
	\label{al_deep:thm:vcd_nn}
	Let $\cH_{\dnn}$ be a set of neural network classifiers of the same architecture and with $W$ parameters arranged in  $L$ layers.
	We then have 
	\begin{align*}
	\Omega(WL \log \prn*{{W}/{L}}) \leq \vcd(\cH_\dnn)  \leq O(WL \log \prn*{ W}). 
	\end{align*}
\end{theorem}

With these tools, we can construct a set of neural network classifiers $\cH_{\dnn}$ such that (i) the best in-class classifier $\check h \in \cH_{\dnn}$ has small excess error, and (ii) $\cH_{\dnn}$  has a well-controlled VC dimension that is proportional to smooth/noise parameters. 
More specifically, we have the following proposition.
\begin{restatable}{proposition}{propVCApprox}
	\label{al_deep:prop:vc_approx}
Suppose $\cD_{\cX \cY} \in \cP(\alpha, \beta)$.
One can construct a set of neural network classifier $\cH_{\dnn}$ such that the following two properties hold simultaneously: 
\begin{align*}
	\inf_{h \in \cH_{\dnn} }\err(h) - \err(h^{\star}) = O \prn{\eps} \quad \text{ and }
\quad 	\vcd( \cH_{\dnn}) = \wt O \prn{ \eps^{- \frac{d}{\alpha(1+\beta)}}}.
\end{align*}
\end{restatable}

With the approximation results obtained above, to learn a classifier with $O(\eps)$ excess error, one only needs to focus on a set of neural networks $\cH_{\dnn}$ with a well-controlled VC dimension.
As a warm-up, we first analyze the label complexity of such procedure in the passive regime (with fast rates).

\begin{restatable}{theorem}{thmPassiveNoise}
	\label{al_deep:thm:passive_noise}
	Suppose $\cD_{\cX \cY} \in \cP(\alpha, \beta)$.
	Fix any $\eps, \delta > 0$.
	Let $\cH_{\dnn}$ be the set of neural network classifiers constructed in \cref{al_deep:prop:vc_approx}.
	With $n = \wt O ( \eps^{- \frac{d+2\alpha + \alpha \beta}{\alpha(1+\beta)} })$ i.i.d. sampled points, with probability at least $1-\delta$,
	the empirical risk minimizer $\wh h \in \cH_{\dnn}$ achieves excess error $ O(\eps)$.
\end{restatable}

The label complexity results obtained in \cref{al_deep:thm:passive_noise} matches, up to logarithmic factors, the passive learning lower bound $\Omega ( \eps^{- \frac{d+2\alpha + \alpha \beta}{\alpha(1+\beta)} })$ established in \citet{audibert2007fast}, indicating that our proposed learning procedure \emph{with a set of neural networks} is near minimax optimal.\footnote{Similar passive learning guarantees have been developed with different tools and analyses, e.g., see results in \citet{kim2021fast}.}

\subsection{Deep Active Learning and Guarantees}
\label{al_deep:sec:noise_active}

The passive learning procedure presented in the previous section treats every data point equally, i.e., it requests the label of every data point. 
Active learning reduces the label complexity by only querying labels of data points that are ``more important''.
We present deep active learning results in this section. 
Our algorithm (\cref{al_deep:alg:NCAL}) is inspired by \textRCAL \citep{balcan2006agnostic, hanneke2007bound, hanneke2014theory} and the seminal \textCAL algorithm \citep{cohn1994improving}; we call our algorithm \textNCAL to emphasize that it works with a set of neural networks.
\looseness=-1

For any accuracy level $\eps > 0$, \textNCAL first initialize a set of neural network classifiers $\cH_0 \ldef \cH_{\dnn}$ such that (i) the best in-class classifier $\check h \ldef \argmin_{h \in \cH_{\dnn}} \err(h)$ has excess error at most  $O(\eps)$, and (ii) the VC dimension of  $\cH_\dnn$ is upper bounded by  $\wt O(\eps^{ - \frac{d}{\alpha(1+\beta)}})$ (see \cref{al_deep:sec:noise_passive} for more details).
\textNCAL then runs in epochs of geometrically increasing lengths.
At the beginning of epoch $m$, based on previously \emph{labeled} data points, \textNCAL updates a set of active classifier $\cH_m$ such that, with high probability, the best classifier $\check h$ remains \emph{uneliminated}.
Within each epoch $m$, \textNCAL only queries the label $y$ of a data point $x$ if it lies in the \emph{region of disagreement} with respect to the current active set of classifier  $\cH_m$, i.e., 
	$\DIS(\cH_m) \ldef \crl{x \in \cX: \exists h_1, h_2 \in \cH_m \text{ s.t. } h_1(x) \neq h_2(x)}$.
\textNCAL returns any classifier $\wh h \in \cH_m$ that remains uneliminated after  $M-1$ epoch.

\begin{algorithm}[H]
	\caption{\textNCAL}
	\label{al_deep:alg:NCAL} 
	\renewcommand{\algorithmicrequire}{\textbf{Input:}}
	\renewcommand{\algorithmicensure}{\textbf{Output:}}
	\newcommand{\algorithmicbreak}{\textbf{break}}
    \newcommand{\BREAK}{\STATE \algorithmicbreak}
	\begin{algorithmic}[1]
		\REQUIRE Accuracy level $\epsilon \in (0, 1)$, confidence level $\delta \in (0, 1)$.
		\STATE Let $\cH_\dnn$ be a set of neural networks classifiers constructed in \cref{al_deep:prop:vc_approx}.
		\STATE Define $T \ldef \eps^{- \frac{2+\beta}{1+\beta}} \cdot \vcd(\cH_{\dnn}) $, $M \ldef \ceil{\log_2 T}$, $\tau_m \ldef 2^m$ for $m\geq1$ and $\tau_0 \ldef 0$. 
		\STATE Define $\rho_m \ldef O \prn*{ \prn*{\frac{\vcd(\cH_{\dnn}) \cdot \log (\tau_{m-1}) \cdot  \log (M /\delta) }{\tau_{m-1}}}^{\frac{1+\beta}{2+\beta}} }$ for $m \geq 2$ and  $\rho_1 \ldef 1$.
		\STATE Define $\wh R_m(h) \ldef \sum_{t = 1}^{\tau_{m-1}} Q_t \ind\prn*{h(x_t) \neq y_t}$ with the convention that $\sum_{t=1}^{0} \ldots = 0$.
		\STATE Initialize $\cH_0 \ldef \cH_{\dnn}$.
		\FOR{epoch $m = 1, 2, \dots, M$}
		\STATE Update active set 
		    $\cH_m \ldef \crl*{ h \in \cH_{m-1}:  \wh R_m(h) \leq \inf_{h \in \cH_{m-1}} \wh R_m(h) + \tau_{m-1} \cdot \rho_m}$
		\IF{epoch $m=M$}
		\STATE \textbf{Return} any classifier $\wh h \in \cH_M$.
		\ENDIF
		\FOR{time $t = \tau_{m-1} + 1 ,\ldots , \tau_{m} $} 
		\STATE Observe $x_t \sim \cD_{\cX}$. Set $Q_t \ldef \ind(x_t \in \DIS(\cH_m))$.
		\IF{$Q_t = 1$}
		\STATE Query the label $y_t$ of $x_t$.
		\ENDIF
		\ENDFOR
		\ENDFOR

	\end{algorithmic}
\end{algorithm}

Since \textNCAL only queries labels of data points lying in the region of disagreement, its label complexity should intuitively be related to how fast the region of disagreement shrinks. More formally, the rate of collapse of the (probability measure of) region of disagreement is captured by the \emph{(classifier-based) disagreement coefficient} \citep{hanneke2007bound, hanneke2014theory}, which 
we introduce next.

\begin{definition}[Classifier-based disagreement coefficient]
	\label{al_deep:def:dis_coeff_classifier}
	For any $\eps_0$ and classifier  $h \in \cH$, the classifier-based disagreement coefficient of  $h$ is defined as 
	 \begin{align*}
		\theta_{\cH, h} (\eps_0)
        \ldef \sup_{\eps > \eps_0} \frac{\P_{ x \sim \cD_{\cX}} 
	\prn{ \DIS \prn{ \cB_{\cH}( h, \eps)}}}{\eps} \vee 1,
	\end{align*}
  where $\cB_{\cH}(h,\eps) \ldef \crl{ g \in \cH: \P_{x \sim \cD_\cX} \prn{  g(x) \neq h(x)} \leq \eps}$.
	We also define $\theta_{\cH}(\eps_0) \ldef \sup_{h \in \cH} \theta_{\cH, h}(\eps_0)$.
\end{definition}

The guarantees of \textNCAL follows from a more general analysis of \textRCAL under function approximation.
In particular, to achieve fast rates under Tsybakov noise, previous analysis of \textRCAL requires that the Bayes optimal classifier lies within the hypothesis class \citep{hanneke2014theory}.
This requirement is typically not satisfied in our setting with neural network approximations.
Our analysis broadens the understanding of \textRCAL under function approximation; 
we defer the general analysis to \cref{al_deep:app:RCAL_gen} and present the guarantees below.
\looseness=-1

\begin{restatable}{theorem}{thmActiveNoise}
	\label{al_deep:thm:active_noise}
	Suppose $\cD_{\cX \cY} \in \cP(\alpha, \beta)$.
	Fix any $\eps,\delta > 0$.
	With probability at least $1-\delta$, \cref{al_deep:alg:NCAL} 
	returns a classifier $\wh h \in \cH_{\dnn}$ with excess error $\wt O(\eps)$ after querying 
	$\wt O \prn{ \theta_{\cH_\dnn} \prn{ \eps^{\frac{\beta}{1 +  \beta}}} \cdot \eps^{- \frac{d + 2 \alpha}{\alpha + \alpha\beta}} }$ labels.
\end{restatable}

We next discuss in detail the label complexity of deep active learning proved in \cref{al_deep:thm:active_noise}.

\begin{itemize}
\item 
Ignoring the dependence on disagreement coefficient, the label complexity appearing in \cref{al_deep:thm:active_noise} matches, up to logarithmic factors, the lower bound $\Omega(\eps^{- \frac{d+ 2 \alpha}{\alpha + \alpha \beta}})$ for active learning \citep{locatelli2017adaptivity}.
	At the same time, the label complexity appearing in \cref{al_deep:thm:active_noise} is \emph{never worse} than the passive counterpart (i.e., $\wt \Theta( \eps^{- \frac{d+2\alpha + \alpha \beta}{\alpha(1+\beta)} })$ since $\theta_{\cH_\dnn}(\eps^{\frac{\beta}{1+\beta}}) \leq \eps^{- \frac{\beta}{1 + \beta}}$.
\item 
We also identify cases when $\theta_{\cH_\dnn}(\eps^{\frac{\beta}{1+\beta}}) = o ( \eps^{- \frac{\beta}{1+\beta}})$, indicating \emph{strict} improvement over passive learning (e.g., when $\cD_\cX$ is supported on countably many data points), 
and when $\theta_{\cH_\dnn}(\eps^{\frac{\beta}{1+\beta}}) = O(1)$, indicating matching the minimax active lower bound (e.g., when $\cD_{\cX \cY}$ satisfies conditions such as \emph{decomposibility} defined in \cref{al_deep:def:decomposable}. See \cref{al_deep:app:dis_coeff} for detailed discussion).\footnote{We remark that disagreement coefficient is usually bounded/analyzed under additional assumptions on $\cD_{\cX \cY}$, even for simple cases with a set of linear classifiers \citep{friedman2009active, hanneke2014theory}. 
The label complexity guarantees of partition-based nonparametric active algorithms (e.g., \citet{castro2008minimax}) do not depend on the disagreement coefficient, but they are analyzed under stronger assumptions, e.g., they require the strictly stronger membership querying oracle. See \citet{wang2011smoothness} for a discussion.
We left a comprehensive analysis of the disagreement coefficient with a set of neural network classifiers for future work.}
\end{itemize}

Our algorithm and theorems lead to the following results, 
which could benefit both deep active learning and nonparametric learning communities.
\begin{itemize}
	\item \textbf{Near minimax optimal label complexity for deep active learning.}
	While empirical successes of deep active learning have been observed, rigorous label complexity analysis remains elusive except for two attempts made in \citet{karzand2020maximin, wang2021neural}.
	The guarantees provided in \citet{karzand2020maximin} only work in very special cases (i.e., data uniformly sampled from $[0,1]$ and labeled by well-separated piece-constant functions in a noise-free way). \citet{wang2021neural} study deep active learning in the NTK regime by linearizing the neural network at its random initialization and analyzing it as a linear function; moreover, as the authors agree, their error bounds and label complexity guarantees are \emph{vacuous} in certain cases. 
	On the other hand, our guarantees are minimax optimal, up to disagreement coefficient and other logarithmic factors, which bridge the gap between theory and practice in deep active learning.
  \looseness=-1
	\item \textbf{New perspective on nonparametric learning.}
	Nonparametric learning of smooth functions have been mainly approached by partitioning-based methods \citep{tsybakov2004optimal, audibert2007fast, castro2008minimax, minsker2012plug, locatelli2017adaptivity, locatelli2018adaptive, kpotufe2021nuances} : Partition the unit cube  $[0,1]^{d}$ into exponentially (in dimension) many sub-cubes and conduct local mean estimation within each sub-cube (which additionally requires a strictly stronger membership querying oracle).
Our results show that, in both passive and active settings, one can learn \emph{globally} with a set of neural networks and achieve near minimax optimal label complexities.
\looseness=-1
\end{itemize}

\section{Deep Active Learning with Abstention: Exponential Speedups}
\label{al_deep:sec:abstention}

While the theoretical guarantees provided in \cref{al_deep:sec:noise} are near minimax optimal, the label complexity scales as $\poly(\frac{1}{\eps})$, which doesn't match the great empirical performance observed in deep active learning.
In this section, we fill in this gap by leveraging the idea of abstention and provide a deep active learning algorithm that achieves exponential label savings.
We introduce the concepts of abstention and Chow's excess error in \cref{al_deep:sec:abs_chow}, and provide our label complexity guarantees in \cref{al_deep:sec:exponential}.\looseness=-1

\subsection{Active Learning without Low Noise Conditions}
\label{al_deep:sec:abs_chow}

The previous section analyzes active learning under Tsybakov noise, which has been extensively studied in the literature since \citet{castro2008minimax}.
More recently, promising results are observed in active learning under Chow's excess error, but otherwise \emph{without any low noise assumption} \citep{puchkin2021exponential, zhu2022efficient}.
We introduce this setting in the following.

\paragraph{Abstention and Chow's error \citep{chow1970optimum}} 
We consider classifier of the form $\widehat h: \cX \rightarrow \cY \cup \curly*{\bot}$ where $\bot$ denotes the action of abstention. For any fixed $0 < \gamma < \frac{1}{2}$, the Chow's error is defined as 
\begin{align*}
    \err_{\gamma}(\widehat h)  \ldef \P_{(x,y) \sim \cD_{\cX \cY}} \prn{\widehat h (x) \neq y,  \widehat h(x) \neq \bot} + \prn*{{1}/{2} - \gamma} \cdot \P_{(x,y) \sim \cD_{\cX \cY}} \prn{\widehat h(x) = \bot}.
\end{align*}
The parameter $\gamma$ can be chosen as a small constant, e.g., $\gamma = 0.01$, to avoid excessive abstention: The price of abstention is only marginally smaller than random guess (which incurs cost $0.5$).
The \emph{Chow's excess error} is then defined as $\exc_\gamma(\wh h) \ldef \err_{\gamma}(\wh h) - \err(h^\star)$ \citep{puchkin2021exponential}.

At a high level, analyzing with Chow's excess error allows slackness in predications of hard examples (e.g., data points whose $\eta(x)$ is close to $\frac{1}{2}$) by leveraging the power of abstention.
\citet{puchkin2021exponential,zhu2022efficient} show that $\polylog(\frac{1}{\eps})$ is always achievable in the \emph{parametric} settings.
We generalize their results to the \emph{nonparametric} setting and analyze active learning with a set of neural networks.

\subsection{Exponential Speedups with Abstention}
\label{al_deep:sec:exponential}

In this section, we work with a set of neural network \emph{regression functions}  $\cF_\dnn: \cX \rightarrow [0,1]$ (that approximates  $\eta$) and then \emph{construct} classifiers  $h: \cX \rightarrow \cY \cup \crl{\bot}$ \emph{with an additional abstention action}. 
To work with a set of regression functions $\cF_{\dnn}$, we analyze its ``complexity'' from the lenses of \emph{pseudo dimension} $\pseud(\cF_{\dnn})$ \citep{pollard1984convergence, haussler1989decision, haussler1995sphere} and \emph{value function disagreement coefficient} $\theta^{\val}_{\cF_\dnn} (\iota)$ (for some $\iota >0$) \citep{foster2020instance}.
We defer detailed definitions of these complexity measures to \cref{al_deep:app:value_func_complexity}.

\begin{algorithm}[H]
	\caption{\textNCALP}
	\label{al_deep:alg:NCALP} 
	\renewcommand{\algorithmicrequire}{\textbf{Input:}}
	\renewcommand{\algorithmicensure}{\textbf{Output:}}
	\newcommand{\algorithmicbreak}{\textbf{break}}
    \newcommand{\BREAK}{\STATE \algorithmicbreak}
	\begin{algorithmic}[1]
		\REQUIRE Accuracy level $\epsilon \in (0, 1)$, confidence level $\delta \in (0, 1)$, abstention parameter $\gamma \in (0, 1/2)$.
		\STATE Let $\cF_\dnn$ be a set of neural network regression functions obtained by (i) applying \cref{al_deep:thm:approx_sobolev} with an appropriate approximation level $\kappa$ (which satisfies $\frac{1}{\kappa} = \poly(\frac{1}{\gamma}) \, \polylog(\frac{1}{\eps \, \gamma})$), and (ii) applying a preprocessing step on the set of neural networks obtained from step (i). See \cref{al_deep:app:abs} for details.
		\STATE Define $T \ldef \frac{\theta^{\val}_{\cF_\dnn}(\gamma /4) \cdot \pseud(\cF_\dnn) }{\eps \, \gamma }$, $M \ldef \ceil{\log_2 T}$, and $C_\delta \ldef O\prn{\pseud(\cF_\dnn) \cdot \log(T /\delta)}$.
		\STATE Define $\tau_m \ldef 2^m$ for $m\geq1$, $\tau_0 \ldef 0$, and $\beta_m \ldef 3(M-m+1) C_\delta$. 
		\STATE Define  $\wh R_m(f) \ldef \sum_{t=1}^{\tau_{m-1}} Q_t \prn{\wh f(x_t) - y_t}^2 $ with the convention that $\sum_{t=1}^{0} \ldots = 0$.
		\FOR{epoch $m = 1, 2, \dots, M$}
		\STATE Get $\widehat f_m \ldef \argmin_{f \in \cF_\dnn} \sum_{t=1}^{\tau_{m-1}} Q_t \paren{f(x_t) - y_t}^2 $.
		\STATE 
		(Implicitely) Construct active set 
		   $ \cF_m \ldef \crl*{ f \in \cF_\dnn:  \wh R_m(f) \leq  \wh R_m(\wh f_m) + \beta_m }$.
		\STATE Construct classifier $\wh h_m: \cX \rightarrow \crl{0, 1, \bot}$ as 
		\begin{align*}
      &\wh h_m (x) \ldef\\ 
      &\begin{cases}
				\bot, & \text{ if } \brk { \lcb(x;\cF_m) - \frac{\gamma}{4}, \ucb(x;\cF_m) + \frac{\gamma}{4}} \subseteq 
				\brk*{ \frac{1}{2} - \gamma, \frac{1}{2} + \gamma}; \\
        \ind(\wh f_m(x) \geq \frac{1}{2} ) , & \text{o.w.}
			\end{cases}
		\end{align*}
		and query function 
			$g_m(x)\ldef \ind \prn*{ \frac{1}{2} \in \prn*{\lcb(x;\cF_m) - \frac{\gamma}{4}, \ucb(x;\cF_m) + \frac{\gamma}{4}} } \cdot
		\ind \prn{\wh h_m(x) \neq \bot}$.
		\IF{epoch $m=M$}
		\STATE \textbf{Return} classifier $\wh h_M$.
		\ENDIF
		\FOR{time $t = \tau_{m-1} + 1 ,\ldots , \tau_{m} $} 
		\STATE Observe $x_t \sim \cD_{\cX}$. Set $Q_t \ldef g_m(x_t)$.
		\IF{$Q_t = 1$}
		\STATE Query the label $y_t$ of $x_t$.
		\ENDIF
		\ENDFOR
		\ENDFOR

	\end{algorithmic}
\end{algorithm}

We now present \textNCALP (\cref{al_deep:alg:NCALP}), a deep active learning algorithm that leverages the power of abstention.
\textNCALP first initialize a set of set of neural network regression functions $\cF_\dnn$ by applying a preprocessing step on top of the set of regression functions obtained from \cref{al_deep:thm:approx_sobolev} with a carefully chosen approximation level $\kappa$.
The preprocessing step mainly contains two actions: 
(1) clipping  $f_\dnn: \cX \rightarrow \R$ into  $\check f_\dnn: \cX \rightarrow [0,1]$ (since we obviously have $\eta(x) \in [0,1]$); and
(2) filtering out  $f_\dnn \in \cF_\dnn$ that are clearly not a good approximation of  $\eta$.
After initialization, \textNCALP runs in epochs of geometrically increasing lengths.
At the beginning of epoch $m \in [M]$, \textNCALP (implicitly) constructs an active set of regression functions $\cF_m$ that are ``close'' to the true conditional probability $\eta$.
For any $x \sim \cD_\cX$, \textNCALP constructs a lower bound $\lcb(x;\cF_m) \ldef \inf_{f \in \cF_m} f(x)$ and an upper bound $\ucb(x;\cF_m) \ldef \sup_{f \in \cF_m} f(x)$ as a confidence range of $\eta(x)$ (based on $\cF_m$).  
An empirical classifier with an abstention option $\wh h_m: \cX \rightarrow \crl{0, 1,\bot}$ and a query function $g_m:\cX \rightarrow \crl{0, 1}$ are then constructed based on the confidence range (and the abstention parameter $\gamma$).
For any time step $t$ within epoch $m$, \textNCALP queries the label of the observed data point $x_t$ if and only if $Q_t \ldef g_m(x_t) = 1$.
\textNCALP returns $\wh h_M$ as the learned classifier.

\textNCALP is adapted from the algorithm developed in \citet{zhu2022efficient}, but with novel extensions. In particular, the algorithm presented in \citet{zhu2022efficient} requires the existence of a $\wb f \in \cF$ such that $\nrm{\wb f - \eta}_\infty \leq \eps$ (to achieve  $\eps$ Chow's excess error),
Such an approximation requirement directly leads to $\poly(\frac{1}{\eps})$ label complexity
\emph{in the nonparametric setting}, which is unacceptable.
The initialization step of \textNCALP (line 1) is carefully chosen to ensure that $\pseud(\cF_\dnn), \theta^{\val}_{\cF_\dnn}(\frac{\gamma}{4}) = \poly(\frac{1}{\gamma}) \cdot \polylog(\frac{1}{\eps})$; together with a sharper analysis of concentration results, these conditions help us derive the following deep active learning guarantees (also see \cref{al_deep:app:abs_gen} for a more general guarantee).

\begin{restatable}{theorem}{thmAbs}
	\label{al_deep:thm:abs}
	Fix any $\eps, \delta, \gamma >0$.
	With probability at least $1-\delta$, \cref{al_deep:alg:NCALP} (with an appropriate initialization at line 1) returns a classifier  $\wh h$ with Chow's excess error $\wt O(\eps)$ after querying $\poly(\frac{1}{\gamma}) \cdot \polylog(\frac{1}{\eps \, \delta})$ labels.
\end{restatable}

We discuss two important aspects of \cref{al_deep:alg:NCALP}/\cref{al_deep:thm:abs} in the following, i.e., exponential savings and computational efficiency. We defer more detailed discussions to \cref{al_deep:app:proper} and \cref{al_deep:app:computational}.

\begin{itemize}
	\item 
	\textbf{Exponential speedups.}
	\cref{al_deep:thm:abs} shows that, equipped with an abstention option, deep active learning enjoys $\polylog(\frac{1}{\eps})$ label complexity. This provides theoretical justifications for great empirical results of deep active learning observed in practice.
	Moreover, \cref{al_deep:alg:NCALP} outputs a classifier that abstains \emph{properly}, i.e., it abstains only if abstention is the optimal choice; such a property further implies $\polylog(\frac{1}{\eps})$ label complexity under \emph{standard} excess error and Massart noise \citep{massart2006risk}.\looseness=-1
	 \item  
	\textbf{Computational efficiency.}
	Suppose one can efficiently implement a (weighted) square loss regression oracle over the \emph{initialized} set of neural networks $\cF_\dnn$:  
		Given any set $\cS$ of weighted examples $(w, x, y) \in \R_+ \times \cX \times \cY$, the regression oracle outputs
$\widehat f_\dnn \ldef \argmin_{f \in \cF_\dnn} \sum_{(w, x, y) \in \cS} w \paren*{f(x) - y}^2$
	.\footnote{In practice, this oracle can be approximated using gradient descent or its variants.} \cref{al_deep:alg:NCALP} can then be \emph{efficiently} implemented with $\poly(\frac{1}{ \gamma}) \cdot \frac{1}{\eps}$ oracle calls.
\end{itemize}

While the label complexity obtained in \cref{al_deep:thm:abs} has desired dependence on $\polylog(\frac{1}{\eps})$, its dependence on $\gamma$ can be of order  $\gamma^{-\poly(d)}$.
Our next result shows that, however, such dependence is unavoidable even in the case of learning a single ReLU function. 
\begin{restatable}{theorem}{thmSingleReLULB}
	\label{al_deep:thm:single_relu_lb}
Fix any $\gamma \in (0,1 /8)$.
For any accuracy level $\eps$ sufficiently small, there exists a problem instance such that 
(1)  $\eta \in \cW^{1,\infty}_1 (\cX)$ and is of the form $\eta(x) \ldef \relu(\ang{w,x}+a)+b$;
  and (2) for any active learning algorithm, it takes at least $\gamma^{-\Omega(d)}$ labels to identify an $\eps$-optimal classifier, for either standard excess error or Chow's excess error (with parameter  $\gamma$).\looseness=-1
\end{restatable}

\section{Extensions}
\label{al_deep:sec:extension}

Previous results are developed in the commonly studied Sobolev/H\"older spaces.
Our techniques, however, are generic and can be adapted to other function spaces, given neural network approximation results.
In this section, we provide extensions of our results to the Radon $\BV^{2}$ space, which was recently proposed as the natural function space associated with ReLU neural networks \citep{ongie2020function, parhi2021banach, parhi2022kinds, parhi2022near, unser2022ridges}.\footnote{Other extensions are also possible given neural network approximation results, e.g., recent results established in \citet{lu2021deep}.}

\paragraph{The Radon $\BV^2$ space}
The Radon $\BV^2$ unit ball over domain $\cX$ is defined as 
	$\RBV^2_1(\cX) \ldef \crl{f: \nrm{f}_{\RBV^2(\cX)} \leq 1}$,
where $\nrm{f}_{\RBV^2(\cX)}$ denotes the Radon $\BV^2$ norm of $f$ over domain $\cX$.\footnote{We provide more mathematical backgrounds and associated definitions in \cref{al_deep:app:extension}.}
Following \citet{parhi2022near}, we assume $\cX = \crl{x \in \R^{d}: \nrm{x}_2 \leq 1}$ and $\eta \in \RBV^2_1(\cX)$.\looseness=-1

The Radon $\BV^2$ space naturally contains neural networks of the form $f_{\dnn}(x) = \sum_{k=1}^{K} v_i \cdot \relu(w_i^{\trn}x + b_i) $.
On the contrary, such $f_\dnn$ doesn't lie in any Sobolev space of order $\alpha \geq 2$ (since  $f_\dnn$ doesn't have second order \emph{weak} derivative).
Thus, if  $\eta$ takes the form of the aforementioned neural network (e.g., $\eta = f_\dnn$), approximating  $\eta$ up to $\kappa $ from a Sobolev perspective requires  $\wt O(\kappa^{-{d}})$ total parameters, which suffers from the curse of dimensionality.
On the other side, however, such bad dependence on dimensionality goes away when approximating from a Radon $\BV^2$ perspective, as shown in the following theorem.

\begin{theorem}[\citet{parhi2022near}]
\label{al_deep:thm:approx_RBV2}	
	Fix any $\kappa>0$. For any $f^{\star} \in \RBV^2_1(\cX)$, there exists a one-hidden layer neural network $f_\dnn$ of width $K = O \prn{\kappa^{- \frac{2d}{d+3}}  }$ such that $\nrm{f^{\star} - f_\dnn}_\infty \leq \kappa$. 
\end{theorem}

Equipped with this approximation result, we provide the active learning guarantees for learning a smooth function within the Radon $\BV^2$ unit ball as follows.

\begin{restatable}{theorem}{thmActiveNoiseRBV}
	\label{al_deep:thm:active_noise_RBV}
	Suppose $\eta \in \RBV^2_1(\cX)$ and the Tsybakov noise condition is satisfied with parameter $\beta \geq 0$.
	Fix any $\eps, \delta > 0$.
	There exists an algorithm such that, with probability at least $1-\delta$, it learns 
	a classifier $\wh h \in \cH_\dnn$ with excess error $\wt O(\eps)$ after querying  
	$\wt O \prn{ \theta_{\cH_\dnn} \prn{ \eps^{\frac{\beta}{1 +  \beta}}} \cdot \eps^{- \frac{4d + 6}{(1+\beta)(d+3)}} }$ labels.
\end{restatable}

Compared to the label complexity obtained in \cref{al_deep:thm:active_noise}, the label complexity obtained in the above theorem doesn't suffer from the curse of dimensionality: For $d$ large enough, the above label complexity scales as  $\eps^{-O(1)}$ yet label complexity in \cref{al_deep:thm:active_noise} scales as $\eps^{-O(d)}$.
Active learning guarantees under Chow's excess error in the Radon $\BV^2$ space are similar to results presented in \cref{al_deep:thm:abs}, and are thus deferred to \cref{al_deep:app:extension}.

\section{Discussion}
\label{al_deep:sec:discussion}
We provide the first near-optimal deep active learning guarantees, under both standard excess error and Chow's excess error. 
Our results are powered by generic algorithms and analyses developed for active learning that bridge approximation guarantees into label complexity guarantees. 
We outline some natural directions for future research below.
\begin{itemize}
	\item \textbf{Disagreement coefficients for neural networks.}
		While we have provided some results regarding the disagreement coefficients for neural networks, we believe a comprehensive investigation on this topic is needed.
	For instance, can we discover more general settings where the classifier-based disagreement coefficient can be upper bounded by $O(1)$?
	It is also interesting to explore sharper analyses on the value function disagreement coefficient.
\item \textbf{Adaptivity in deep active learning.}
	Our current results are established with the knowledge of some problem-dependent parameters, e.g., the smoothness parameters regarding the function spaces and the noise levels.
	It will be interesting to see if one can develop algorithms that can automatically adapt to unknown parameters, e.g., by leveraging techniques developed in \citet{locatelli2017adaptivity, locatelli2018adaptive}.
\end{itemize}

\section{Generic Version of \cref{al_deep:alg:NCAL} and Its Guarantees}
\label{al_deep:app:RCAL_gen}
We present \cref{al_deep:alg:RCAL} below, a generic version of \cref{al_deep:alg:NCAL} that doesn't require the approximating classifiers to be neural networks.
The guarantees of \cref{al_deep:alg:RCAL} are provided in \cref{al_deep:thm:RCAL_gen}, which is proved in \cref{al_deep:app:RCAL_gen_proof} based on 
supporting lemmas provided in \cref{al_deep:app:RCAL_gen_lemma}.

\begin{algorithm}[H]
	\caption{\textRCAL with Approximation}
	\label{al_deep:alg:RCAL} 
	\renewcommand{\algorithmicrequire}{\textbf{Input:}}
	\renewcommand{\algorithmicensure}{\textbf{Output:}}
	\newcommand{\algorithmicbreak}{\textbf{break}}
    \newcommand{\BREAK}{\STATE \algorithmicbreak}
	\begin{algorithmic}[1]
		\REQUIRE Accuracy level $\epsilon \in (0, 1)$, confidence level $\delta \in (0, 1)$.
		\STATE Let $\cH$ be a set of approximating classifiers such that $\inf_{h \in \cH }\err(h) - \err(h^{\star}) = O \prn{\eps}$.
		\STATE Define $T \ldef \eps^{- \frac{2+\beta}{1+\beta}} \cdot \vcd(\cH) $, $M \ldef \ceil{\log_2 T}$, $\tau_m \ldef 2^m$ for $m\geq1$ and $\tau_0 \ldef 0$. 
		\STATE Define $\rho_m \ldef O \prn*{ \prn*{\frac{\vcd(\cH) \cdot \log (\tau_{m-1}) \cdot  \log (M /\delta) }{\tau_{m-1}}}^{\frac{1+\beta}{2+\beta}} }$ for $m \geq 2$ and  $\rho_1 \ldef 1$.
		\STATE Define $\wh R_m(h) \ldef \sum_{t = 1}^{\tau_{m-1}} Q_t \ind\prn*{h(x_t) \neq y_t}$ with the convention that $\sum_{t=1}^{0} \ldots = 0$.
		\STATE Initialize $\cH_0 \ldef \cH$.
		\FOR{epoch $m = 1, 2, \dots, M$}
		\STATE Update active set 
    \begin{align*}
        \cH_m \ldef \crl*{ h \in \cH_{m-1}:  \wh R_m(h) \leq \inf_{h \in \cH_{m-1}} \wh R_m(h) + \tau_{m-1} \cdot \rho_m}.
    \end{align*}
		\IF{epoch $m=M$}
		\STATE \textbf{Return} any classifier $\wh h \in \cH_M$.
		\ENDIF
		\FOR{time $t = \tau_{m-1} + 1 ,\ldots , \tau_{m} $} 
		\STATE Observe $x_t \sim \cD_{\cX}$. Set $Q_t \ldef \ind(x_t \in \DIS(\cH_m))$.
		\IF{$Q_t = 1$}
		\STATE Query the label $y_t$ of $x_t$.
		\ENDIF
		\ENDFOR
		\ENDFOR

	\end{algorithmic}
\end{algorithm}
We provide guarantees for \cref{al_deep:alg:RCAL}, and then specialize them to the settings with neural network approximation, i.e., in \cref{al_deep:thm:active_noise} and \cref{al_deep:thm:active_noise_RBV}.
Our proofs build on the analysis of \textRCAL \citep{hanneke2014theory}, with additional arguments to handle function approximation. We note that the original analysis assumes $h^\star \in \cH$, i.e., the Bayes optimal classifier is contained in the hypothesis class.

 \begin{restatable}{theorem}{thmRCALGen}
	\label{al_deep:thm:RCAL_gen}
	Fix $\eps, \delta > 0$.
	Suppose $\inf_{h \in \cH} \err(h) - \err(h^{\star}) = O(\eps)$.
	With probability at least $1-\delta$, 
	\cref{al_deep:alg:RCAL} returns a classifier $\wh h \in \cH$ with excess error $\wt O(\eps)$ 
	after querying 
	 \begin{align*}
		 \wt O \prn*{ \theta_\cH(\eps^{\frac{\beta}{1+\beta}}) \cdot \eps^{-\frac{2}{1+\beta}}\cdot \vcd(\cH)  }
	\end{align*}
	labels.
\end{restatable}

\subsection{Supporting Lemmas}
\label{al_deep:app:RCAL_gen_lemma}

\begin{lemma}[\citet{tsybakov2004optimal, hanneke2014theory}]
	\label{al_deep:lm:bernstein}
  Let $h^\star$ denote the Bayes optimal classifier.
	Suppose $\cD_{\cX \cY}$ satisfies the Tsybakov noise condition with parameter $\beta \geq 0$, then there exists an universal constant $c^{\prime} > 0$ such that we have 
\begin{align*}
\P_{x \sim \cD_\cX} \prn{h(x) \neq h^{\star}(x)} \leq c^{\prime} \prn{ \err(h) - \err(h^{\star})}^{\frac{\beta}{1+\beta}}	
\end{align*}
for any measurable $h: \cX \rightarrow \cY$.
\end{lemma}

We next present a lemma in the passive learning setting, which will later be incorporated into the active learning setting. 
We first define some notations. 
Suppose $D_n = \crl{(x_i, y_i)}_{i=1}^{n}$ are $n$ i.i.d. data points drawn from $\cD_{\cX \cY}$.
For any measurable $h: \cX \rightarrow \cY$, we denote  $\wb R_n(h) \ldef \sum_{i=1}^{n} \ind(h(x_i) \neq y_i)$ as the empirical error of $h$ over dataset $D_n$. We clearly have $\E\brk{\wb R_n(h)} = n \cdot \err(h)$ by i.i.d. assumption.

\begin{lemma}
	\label{al_deep:lm:tsy_concentration}
Fix $\eps, \wb \delta > 0$.
Suppose $\cD_{\cX \cY}$ satisfies Tsybakov noise condition with parameter $\beta \geq 0$ and $\err(\check h) - \err(h^{\star}) = O(\eps)$, where $\check h = \argmin_{h \in \cH} \err(h)$ and $h^{\star}$ is the Bayes classifier.
Let $D_n = \crl{(x_i, y_i)}_{i=1}^{n}$ be a set of $n$ i.i.d. data points drawn from $\cD_{\cX \cY}$. 
If $\beta > 0$, suppose $n$ satisfies 
\begin{align*}
	n \leq \eps^{- \frac{2+\beta}{1+\beta}} \cdot \vcd(\cH)^{\frac{2+2\beta}{\beta}} \cdot \log (\wb \delta^{-1}) \cdot \prn{\log n}^{\frac{2+2\beta}{\beta}}.
\end{align*}
With probability at least $1-\wb \delta$, we have the following inequalities hold:
 \begin{align}
	 n \cdot \prn{ \err(h) - \err(h^{\star})} & \leq 2 \cdot \prn{\wb R_n(h) - \wb R_n(\check h)} + n \cdot \rho(n,\wb\delta) , \quad \forall h \in \cH, \label{al_deep:eq:tsy_concentration_1} \\
	 \wb R_n(\check h)  - \min_{h \in\cH} \wb R_n(h) & \leq  n \cdot \rho(n, \wb \delta), \label{al_deep:eq:tsy_concentration_2}
\end{align}
  where $\rho(n,\wb \delta) \ldef C \cdot  \prn*{ \prn*{\frac{\vcd(\cH) \cdot \log n \cdot  \log \wb \delta^{-1} }{n}}^{\frac{1+\beta}{2+\beta}} + \eps }$ with a universal constant $C >0$.\footnote{The logarithmic factors in this bound might be further optimized; however, we do not focus on optimizing logarithmic factors in this work.}
\end{lemma}
\begin{proof}
	Denote $\wb \cH \ldef \cH \cup \crl{h^{\star}}$. We know that $\vcd(\wb \cH) \leq \vcd(\cH) + 1 = O(\vcd(\cH))$.
  Since $\cD_{\cX \cY}$ satisfies Tsybakov noise condition and $h^\star \in \wb \cH$, the condition in \cref{al_deep:lm:bernstein} is satisfied by all $h \in \wb \cH$.
Invoking Lemma 3.1 in \citet{hanneke2014theory}, with probability at least $1-\frac{\wb \delta}{2}$, $\forall h \in \wb \cH$, we have 
 \begin{align}
n \cdot \prn{ \err(h) - \err(h^{\star})} & \leq \max \crl*{2 \cdot \prn{\wb R_n(h) - \wb R_n(h^{\star})} , n \cdot \wb \rho(n, \wb \delta)} , \label{al_deep:eq:tsy_concentration_3}\\
\wb R_n(h)  - \min_{h \in \wb \cH} \wb R_n(h) & \leq  \max \crl*{ 2n \cdot \prn{\err(h) - \err(h^{\star})}  , n \cdot \wb \rho(n, \wb \delta)} \label{al_deep:eq:tsy_concentration_4},
\end{align}
where $\wb \rho(n, \wb \delta) = O \prn*{ \prn*{\frac{\vcd(\wb \cH) \cdot \log n + \log \wb \delta^{-1} }{n}}^{\frac{1+\beta}{2+\beta}}  }= O \prn*{ \prn*{\frac{\vcd(\cH) \cdot \log n \cdot \log \wb \delta^{-1} }{n}}^{\frac{1+\beta}{2+\beta}}  }$.

\cref{al_deep:eq:tsy_concentration_2} follows by taking $h = \check h$ in \cref{al_deep:eq:tsy_concentration_4} and noticing that 
 \begin{align*}
\wb R_h(\check h) - \min_{h \in \cH} \wb R_n(h) 
& \leq \wb R_n (\check h) - \min_{h \in \wb \cH} \wb R_n(h)\\
& \leq \max \crl*{ 2n \cdot O(\eps) , n \cdot \wb \rho(n,\wb \delta)},
\end{align*}
where we use the assumption that $\err(\check h) - \err(h^{\star}) = O(\eps)$.

To derive \cref{al_deep:eq:tsy_concentration_1}, we first notice that applying \cref{al_deep:eq:tsy_concentration_3} for any $h \in \cH$, we have 
\begin{align*}
	n \cdot \prn{\err(h) - \err(h^{\star})}
	& \leq 2 \cdot \prn{ \wb R_n(h) -\wb R_n(\check h) + \wb R_n(\check h) - \wb R_n(h^{\star}) } + n \cdot \wb \rho(n, \wb \delta).
\end{align*}
We next only need to upper bound $\wb R_n(\check h) - \wb R_n(h^{\star})$, and show that it is order-wise smaller than $n \cdot \rho(n ,\wb \delta)$.
We consider random variable $g_i \ldef \ind(\check h(x_i) \neq y_i) - \ind(h^{\star}(x_i) \neq y_i)$.
We have 
\begin{align*}
	\V(g_i) 
	& \leq \E \brk{g_i^2}\\
	& = \E \brk{\ind(\check h(x_i) \neq h^{\star}(x_i)} \\
	& = O \prn*{ \eps^{\frac{\beta}{1+\beta}}  },
\end{align*}
where the last line follows from \cref{al_deep:lm:bernstein} and the assumption that $\err(\check h) - \err(h^{\star}) = O(\eps)$.
Denote $g = \frac{1}{n} \sum_{i=1}^{n} g_i = \frac{1}{n} \prn{\wb R_n(\check h) - \wb R_n(h^{\star})}$, and notice that $\E \brk{g} = \err(\check h) - \err(h^{\star})$. 
  Applying Bernstein inequality (e.g., Lemma B.9 in \citet{shalev2014understanding}) on $g - \E\sq{g}$, with probability at least  $1- \frac{\wb \delta}{2}$, we have 
\begin{align*}
	g - \E \brk{g} \leq O \prn*{\prn*{\frac{\eps^{\frac{\beta}{1+\beta}} \log \wb \delta^{-1}} {n}}^{\frac{1}{2}} + \frac{\log \wb \delta^{-1}}{n} },
\end{align*}
which further leads to 
\begin{align*}
	\wb R_n(\check h) - \wb R_n (h^{\star}) \leq n \cdot O\prn*{ \eps + \prn*{\frac{\eps^{\frac{\beta}{1+\beta}} \log \wb \delta^{-1}} {n}}^{\frac{1}{2}} + \frac{\log \wb \delta^{-1}}{n}}.
\end{align*}

The RHS is order-wise smaller than $\rho_n$ when  $\beta = 0$. We consider the case when  $\beta > 0$ next.
Since $\log (\wb \delta^{-1}) /n $ is clearly a lower-order term compared to $\rho_n$, we only need to show that 
$\prn*{\frac{\eps^{\frac{\beta}{1+\beta}} \log \wb \delta^{-1}} {n}}^{\frac{1}{2}} $ is order-wise smaller than  $\rho_n$.
We can easily check that 
\begin{align*}
	\prn*{\frac{\eps^{\frac{\beta}{1+\beta}} \log \wb \delta^{-1}} {n}}^{\frac{1}{2}} \leq \prn*{\frac{\vcd(\cH) \cdot \log n \cdot  \log \wb \delta^{-1} }{n}}^{\frac{1+\beta}{2+\beta}}
\end{align*}
whenever $n$ satisfies the following condition
\begin{align*}
	n \leq \eps^{- \frac{2+\beta}{1+\beta}} \cdot \vcd(\cH)^{\frac{2+2\beta}{\beta}} \cdot \log (\wb \delta^{-1}) \cdot \prn{\log n}^{\frac{2+2\beta}{\beta}}.
\end{align*}
\end{proof}

We denote $\check h \ldef \argmin_{h \in \cH} \err(h)$, which satisfies $\err(\check h) - \err(h^{\star}) = O(\eps)$ (as assumed in \cref{al_deep:thm:RCAL_gen}).
For any $h \in \cH$,
we also use the shorthand $\wb R_m(h) = \wb R_{\tau_{m-1}} (h) \ldef \sum_{t=1}^{\tau_{m-1}} \ind(h(x_t) \neq y_t)$. Note that $\wb R_m$ is only used in analysis since some  $y_t$ are not observable.

\begin{lemma}
	\label{al_deep:lm:RCAL_set}
	With probability at least $1-\frac{\delta}{2}$, the following holds true for all epochs $m \in [M]$:
	\begin{enumerate}
		\item $\check h \in \cH_m$.
		\item $\err(h) - \err(h^{\star}) \leq 3 \rho_m, \forall h \in \cH_m$. 
	\end{enumerate}

\end{lemma}
\begin{proof}
	For each $m = 2, 3, \ldots, M$, we invoke \cref{al_deep:lm:tsy_concentration} with $n = \tau_{m-1}$ and  $\wb \delta = \delta /2M$, which guarantees that 
 \begin{align}
	 \tau_{m-1} \cdot \prn{ \err(h) - \err(h^{\star})} & \leq 2 \cdot \prn{\wb R_m(h) - \wb R_m(\check h)} + \tau_{m-1} \cdot \rho_m, \quad \forall h \in \cH, \label{al_deep:eq:RCAL_set_1} \\
	 \wb R_m(\check h)  - \min_{h \in\cH} \wb R_m(h) & \leq  \tau_{m-1}\cdot \rho_m. \label{al_deep:eq:RCAL_set_2}
\end{align}
  Note that the choice of $T$ used in \cref{al_deep:alg:RCAL} ensures that (1) the requirement needed for $n $ in \cref{al_deep:lm:tsy_concentration} when  $\beta > 0$ is satisfied, and (2) the second term $\eps$ in $\rho(\tau_{m-1}, \delta /2M)$ (see \cref{al_deep:lm:tsy_concentration} for definition of $\rho(\tau_{m-1}, \delta /2M)$) is a lower-order term compared to the first term.
We use $\cE$ to denote the good event where \cref{al_deep:eq:RCAL_set_1} and  \cref{al_deep:eq:RCAL_set_2} hold true across $m = 2, 3, \ldots, M$. This good event happens with probability at least $1-\frac{\delta}{2}$. We analyze under $\cE$ in the following.

We prove \cref{al_deep:lm:RCAL_set} through induction. The statements clearly hold true for  $m = 1$. Suppose  the statements hold true up to epoch $m$, we next prove the correctness for epoch  $m+1$. 

We know that $\check h \in \cH_m$ by assumption. Based on the querying criteria of  \cref{al_deep:alg:RCAL}, we know that 
\begin{align}
	\label{al_deep:eq:RCAL_equiv}
	\wh R_{m+1} (\check h) - \wh R_{m + 1} (h) 
	&  = \wb R_{m+1} (\check h) - \wb R_{m+1} (h), \quad \forall h \in \cH_m
\end{align}
  From \cref{al_deep:eq:RCAL_set_2} (at epoch $m+1$), we also have 
\begin{align*}
\wb R_{m+1} (\check h) - \min_{h \in \cH_m} R_{m+1} (h)
& \leq 
\wb R_{m+1} (\check h) - \min_{h \in \cH} R_{m+1} (h)\\
& \leq \tau_m \cdot \rho_{m+1}.
\end{align*}
Combining the above two inequalities leads to
\begin{align*}
	\wh R_{m+1} (\check h) - \wh R_{m + 1} (h)  \leq \tau_m \cdot \rho_{m+1},
\end{align*}
implying that $\check h \in \cH_{m+1}$ (due to the construction of $\cH_{m+1}$ in \cref{al_deep:alg:RCAL}).

Based on \cref{al_deep:eq:RCAL_equiv}, the construction of $\cH_{m+1}$, and the fact that  $\check h \in \cH_{m}$,
we know that, for any $h \in \cH_{m+1} \subseteq \cH_m$, 
\begin{align*}
	\wb R_{m+1} (h) - \wb R_{m+1} (\check h)
	& = \wh R_{m+1}(h) - \wh R_{m+1} (\check h)\\
	& \leq \wh R_{m+1} (h) - \min_{h \in \cH_m} \wh R_{m+1} (h)\\
	& \leq \tau_m \cdot \rho_{m+1}.
\end{align*}
Plugging the above inequality into \cref{al_deep:eq:RCAL_set_1} (at epoch $m+1$) leads to  $\err(h) - \err(h^{\star}) \leq 3 \rho_{m+1}$ for any $h \in \cH_{m+1}$.  
We thus prove the desired statements at epoch  $m+1$.
\end{proof}

\subsection{Proof of \cref{al_deep:thm:RCAL_gen}}
\label{al_deep:app:RCAL_gen_proof}

\thmRCALGen*
\begin{proof}
	Based on \cref{al_deep:lm:RCAL_set}, we know that, with probability at least $1- \frac{\delta}{2}$, we have 
	\begin{align*}
		\err(\wh h) - \err(h^{\star})
		& \leq 3 \rho_M \\
		& = O \prn*{ \prn*{\frac{\vcd(\cH) \cdot \log (\tau_{M-1}) \cdot  \log (M /\delta) }{\tau_{M-1}}}^{\frac{1+\beta}{2+\beta}}  }\\
		& = \wt O (\eps),
	\end{align*}
	where we use the definition of $T$ and  $\tau_M$.

	We next analyze the label complexity of \cref{al_deep:alg:RCAL}. 
	Since \cref{al_deep:alg:RCAL} stops and the beginning at epoch $M$, we only need to calculated the label complexity in the first  $M-1$ epochs.
	We have 
	\begin{align*}
		\sum_{t=1}^{\tau_{M-1}} Q_t
    & = \sum_{m=1}^{M-1} \sum_{t = \tau_{m-1}+ 1}^{\tau_m}\ind \prn{x_t \in \DIS(\cH_m)}\\
		& \leq \sum_{m=1}^{M-1} \sum_{t = \tau_{m-1}+ 1}^{\tau_m} \ind \prn*{x_t \in \DIS(\cB_{\cH}(h^{\star}, c^{\prime} \prn{3 \rho_m}^{\frac{\beta}{1+\beta}}))},
	\end{align*}
	where on the second line we use the facts
	(1) $\err(h) - \err(h^{\star}) \leq 3 \rho_m, \forall h \in \cH_m$ from \cref{al_deep:lm:RCAL_set},
  and (2) $\P_{x \sim \cD_\cX}\prn{ h(x) \neq h^{\star}(x)} \leq c^{\prime} \prn{\err(h) - \err(h^{\star})}^{\frac{\beta}{1+\beta}}$ from \cref{al_deep:lm:bernstein} (with the same constant  $c^{\prime}$).
	Suppose $\err(\check h ) - \err(h^{\star}) = c^{\prime \prime} \eps$ with a universal constant $c^{\prime \prime}$ by assumption. Applying \cref{al_deep:lm:bernstein} on $\check h$ leads to the fact that 
	 $h^{\star} \in \cB_{\cH} ( \check h, c^{\prime}\prn{c^{\prime \prime} \eps}^{\frac{\beta}{1+\beta}})$.
   Since $\P_{x \sim \cD_\cX}( h(x) \neq \check h(x)) \leq \P_{x \sim \cD_\cX} ( h(x) \neq h^{\star}(x)) + \P_{x \sim \cD_\cX} (h^{\star}(x) \neq \check h(x))$, 
	 we further have
	 \begin{align*}
	 	\sum_{t=1}^{\tau_{M-1}}Q_t
		& \leq \sum_{m=1}^{M-1}\sum_{t = \tau_{m-1}+ 1}^{\tau_m} \ind \prn*{x_t \in \DIS(\cB_{\cH}(\check h, \wb c\cdot {\rho_m}^{\frac{\beta}{1+\beta}}))},
	 \end{align*}
	 with a universal constant $\wb c > 0$.
	Noticing that the RHS is a sum of independent Bernoulli random variables,
applying Chernoff bound leads to the following guarantees on an event $\cE^{\prime}$ that happens with probability at least $1-\frac{\delta}{2}$:
\begin{align*}
	 	\sum_{t=1}^{\tau_{M-1}}Q_t
		& \leq 2e \sum_{m=1}^{M-1}\sum_{t = \tau_{m-1}+ 1}^{\tau_m} \P \prn*{x \in \DIS(\cB_{\cH}(\check h, \wb c \cdot {\rho_m}^{\frac{\beta}{1+\beta}}))} + 2 \log(2 /\delta) \\
  & = 2e \sum_{m=1}^{M-1} \prn{\tau_m - \tau_{m-1}} \cdot \P \prn*{x \in \DIS(\cB_{\cH}(\check h, \wb c \cdot {\rho_m}^{\frac{\beta}{1+\beta}}))} + 2 \log(2 /\delta) \\
		& \leq 2e \sum_{m=2}^{M-1} { \tau_{m-1} } \cdot \theta_{\cH, \check h}\prn*{\wb c \cdot \rho_m^{\frac{\beta}{1+\beta}}} \cdot \wb c \cdot \rho_m^{\frac{\beta}{1+\beta}} + 2 \log (2 /\delta) + 4e\\
		& \leq 2e M \cdot \theta_{\cH, \check h}\prn*{\wb c \cdot \rho_M^{\frac{\beta}{1+\beta}}} \cdot  \prn*{\wb c\cdot \tau_{M-1}\cdot \rho_M^{\frac{\beta}{1+\beta}} } + 2 \log( 2 /\delta) + 4e,
\end{align*}
  where the third line follows from the definition of disagreement coefficient, and the last line follows from the facts that $\crl{\rho_m}$ is a non-increasing sequence yet $\crl{\tau_{m-1} \cdot \rho_m}$ is an increasing sequence.
Basic algebra and basic properties of the disagreement coefficient (i.e., Theorem 7.1 and Corollary 7.2 in \citet{hanneke2014theory}) shows that 
\begin{align*}
	\sum_{t=1}^{\tau_{M-1}} Q_t \leq \wt O \prn*{ \theta_\cH(\eps^{\frac{\beta}{1+\beta}}) \cdot \eps^{-\frac{2}{1+\beta}}\cdot \vcd(\cH)  } ,
\end{align*}
under event $\cE \cap \cE^{\prime}$, which happens with probability at least $1-\delta$.
\end{proof}

\section{Generic Version of \cref{al_deep:alg:NCALP} and Its Guarantees}
\label{al_deep:app:abs_gen}

This section is organized as follows.
We first introduce some complexity measures in \cref{al_deep:app:value_func_complexity}. 
We then provide the generic algorithm (\cref{al_deep:alg:abs}) and state its theoretical guarantees (\cref{al_deep:thm:abs_gen}) in \cref{al_deep:app:abs_gen_class}.

\subsection{Complexity Measures}
\label{al_deep:app:value_func_complexity}

We first introduce \emph{pseudo dimension} \citep{pollard1984convergence, haussler1989decision, haussler1995sphere}, a complexity measure used to analyze real-valued functions.

\begin{definition}[Pseudo dimension]
\label{al_deep:def:pseudo_d}
Consider a set of real-valued function $\cF: \cX \rightarrow \R$. The pseudo dimension $\pseud(\cF)$ of $\cF$ is defined as the VC dimension of the set of threshold functions 
$\crl{(x,\zeta) \mapsto \ind(f(x) > \zeta) : f \in \cF}$.
\end{definition}

As discussed in \citet{bartlett2019nearly}, similar results as in \cref{al_deep:thm:vcd_nn} holds true for $\pseud(\cF)$ as well.

\begin{theorem}[\citet{bartlett2019nearly}]
	\label{al_deep:thm:pdim_nn}
	Let $\cF_{\dnn}$ be a set of neural network regression functions of the same architecture and with $W$ parameters arranged in  $L$ layers.
	We then have 
	\begin{align*}
	\Omega(WL \log \prn*{{W}/{L}}) \leq \pseud(\cF_\dnn)  \leq O(WL \log \prn*{ W}). 
	\end{align*}
\end{theorem}

We now introduce \emph{value function disagreement coefficient}, which is proposed by \citet{foster2020instance} in contextual bandits and then adapted to active learning by \citet{zhu2022efficient} with additional supreme over the marginal distribution $\cD_\cX$ to deal with distributional shifts caused by selective sampling.

\begin{definition}[Value function disagreement coefficient]
    \label{al_deep:def:dis_coeff_value}
    For any $f^{\star} \in \cF$ and $\gamma_0 ,\epsilon_0 > 0$, the value function disagreement coefficient $	 \theta^{\val}_{f^{\star}}(\cF, \gamma_0, \eps_0)$ is defined as 
    \begin{align*}
 \sup_{\cD_\cX}\sup_{\gamma> \gamma_0, \eps> \eps_0} 
	 \crl*{ \frac{\gamma^2}{\epsilon^2} \cdot 
	\P_{\cD_\cX} \prn*{ \exists f \in \cF: \abs{f(x) - f^{\star}(x)} > \gamma,
	\nrm*{ f - f^{\star}}_{\cD_\cX} \leq \eps} } \vee 1,
    \end{align*}
    where $\nrm{f}^2_{\cD_\cX} \ldef \E_{x \sim \cD_\cX} \brk{f^2(x)}$. We also define $\theta^{\val}_\cF (\gamma_0) \ldef \sup_{f^{\star} \in \cF, \eps_0 > 0} \theta^{\val}_{f^{\star}}(\cF, \gamma_0, \eps_0)$.
\end{definition}

\subsection{The Generic Algorithm and Its Guarantees}
\label{al_deep:app:abs_gen_class}
We present \cref{al_deep:alg:abs}, a generic version of \cref{al_deep:alg:NCALP} that doesn't require the approximating classifiers to be neural networks.

\begin{algorithm}[H]
	\caption{\textNCALP (Generic Version)}
	\label{al_deep:alg:abs} 
	\renewcommand{\algorithmicrequire}{\textbf{Input:}}
	\renewcommand{\algorithmicensure}{\textbf{Output:}}
	\newcommand{\algorithmicbreak}{\textbf{break}}
    \newcommand{\BREAK}{\STATE \algorithmicbreak}
	\begin{algorithmic}[1]
		\REQUIRE Accuracy level $\epsilon \in (0, 1)$, confidence level $\delta \in (0, 1)$, abstention parameter $\gamma \in (0, 1/2)$.
		\STATE Let $\cF: \cX \rightarrow [0, 1]$ be a set of regression functions such that there exists a regression function $\wb f \in \cF$ with $\nrm{\wb f - \eta}_\infty \leq \kappa \leq \gamma / 4$.
		\STATE Define $T \ldef \frac{\theta^{\val}_\cF(\gamma / 4) \cdot \pseud(\cF) }{\eps \, \gamma }$, $M \ldef \ceil{\log_2 T}$, and $C_\delta \ldef O\prn{\pseud(\cF) \cdot \log(T /\delta)}$.
		\STATE Define $\tau_m \ldef 2^m$ for $m\geq1$, $\tau_0 \ldef 0$, and $\beta_m \ldef 3(M-m+1) C_\delta$. 
		\STATE Define  $\wh R_m(f) \ldef \sum_{t=1}^{\tau_{m-1}} Q_t \prn{\wh f(x_t) - y_t}^2 $ with the convention that $\sum_{t=1}^{0} \ldots = 0$.
		\FOR{epoch $m = 1, 2, \dots, M$}
		\STATE Get $\widehat f_m \ldef \argmin_{f \in \cF} \sum_{t=1}^{\tau_{m-1}} Q_t \paren{f(x_t) - y_t}^2 $.
		\STATE 
		(Implicitely) Construct active set 
		   $ \cF_m \ldef \crl*{ f \in \cF:  \wh R_m(f) \leq  \wh R_m(\wh f_m) + \beta_m }$.
		\STATE Construct classifier $\wh h_m: \cX \rightarrow \crl{0, 1,\bot}$ as 
		\begin{align*}
      &\wh h_m (x) \ldef \\
      &\begin{cases}
				\bot, & \text{ if } \brk { \lcb(x;\cF_m) - \frac{\gamma}{4}, \ucb(x;\cF_m) + \frac{\gamma}{4}} \subseteq 
				\brk*{ \frac{1}{2} - \gamma, \frac{1}{2} + \gamma}; \\
        \ind(\wh f_m(x) \geq \frac{1}{2} ) , & \text{o.w.}
			\end{cases}
		\end{align*}
		and query function 
			$g_m(x)\ldef \ind \prn*{ \frac{1}{2} \in \prn*{\lcb(x;\cF_m) - \frac{\gamma}{4}, \ucb(x;\cF_m) + \frac{\gamma}{4}} } \cdot
		\ind \prn{\wh h_m(x) \neq \bot}$.
		\IF{epoch $m=M$}
		\STATE \textbf{Return} classifier $\wh h_M$.
		\ENDIF
		\FOR{time $t = \tau_{m-1} + 1 ,\ldots , \tau_{m} $} 
		\STATE Observe $x_t \sim \cD_{\cX}$. Set $Q_t \ldef g_m(x_t)$.
		\IF{$Q_t = 1$}
		\STATE Query the label $y_t$ of $x_t$.
		\ENDIF
		\ENDFOR
		\ENDFOR

	\end{algorithmic}
\end{algorithm}

We next state the theoretical guarantees for \cref{al_deep:alg:abs}.

\begin{restatable}{theorem}{thmAbsGen}
	\label{al_deep:thm:abs_gen}
	Suppose $\theta^{\val}_\cF(\gamma / 4) \leq \wb \theta$ and the approximation level $\kappa \in (0, \gamma /4]$ satisfies 
 \begin{align}
	 \label{al_deep:eq:kappa_requirement}
	\prn*{\frac{432 \wb \theta \cdot M^2}{\gamma^2}} \cdot \kappa^2  
	\leq \frac{1}{10}.
\end{align}
	With probability at least  $1-\delta$, \cref{al_deep:alg:abs} returns a classifier $\wh h: \cX \rightarrow \crl{0, 1, \bot}$ with Chow's excess error 
\begin{align*}
	\exc_\gamma(\wh h) =
	 O \prn*{  \eps \cdot \log \prn*{\frac{\wb \theta \cdot \pseud(\cF)}{\eps \, \gamma \, \delta}}}, 
\end{align*}
after querying at most 
\begin{align*}
O\prn*{  \frac{ M^2 \cdot \pseud(\cF) \cdot \log(T /\delta) \cdot \wb \theta }{\gamma^2}} 
\end{align*}
labels.
\end{restatable}

\cref{al_deep:thm:abs_gen} is proved in \cref{al_deep:app:abs_gen_proof}, based on supporting lemmas and theorems established in \cref{al_deep:app:concentration_abs} and \cref{al_deep:app:supporting_abs}.
The general result (\cref{al_deep:thm:abs_gen}) will be used to prove results in specific settings (e.g., \cref{al_deep:thm:abs} and \cref{al_deep:thm:abs_RBV}).

\subsubsection{Concentration Results}
\label{al_deep:app:concentration_abs}

\begin{lemma}[Freedman's inequality \citep{freedman1975tail, agarwal2014taming}]
    \label{al_deep:lm:freedman}
    Let $(X_t)_{t \leq T}$ be a real-valued martingale difference sequence adapted to a filtration $\mfF_t$, and let $\E_t \sq{\cdot} \ldef \E \sq{\cdot \mid \mfF_{t-1}}$. If $\abs{X_t} \leq B$ almost surely, then for any $\eta \in (0,1/B)$ it holds with probability at least $1 - \delta$,
    \begin{align*}
        \sum_{t=1}^{T} X_t \leq \eta \sum_{t=1}^{T} \E_{t} \sq{X_t^2} + \frac{\log \delta^{-1}}{\eta}.
    \end{align*}
\end{lemma}

\begin{lemma}[\citep{foster2020instance}]
   \label{al_deep:lm:martingale_two_sides} 
   Let $(X_t)_{t \leq T}$ be a sequence of random variables adapted to a filtration $\mfF_t$. If $0 \leq {X_t} \leq B$ almost surely, then with probability at least $1-\delta$,
   \begin{align*}
       \sum_{t =1 }^T X_t \leq \frac{3}{2} \sum_{t=1}^T \E_{t}\sq{X_t} + 4B \log(2 \delta^{-1}),
   \end{align*}
   and 
   \begin{align*}
       \sum_{t =1 }^T \E_{t} \sq{X_t} \leq 2 \sum_{t=1}^T X_t + 8B \log(2 \delta^{-1}).
   \end{align*}
\end{lemma}
\begin{proof}
  These two inequalities are obtained by applying \cref{al_deep:lm:freedman} to $\prn{X_t - \E_t \sq{X_t}}_{t \leq T}$ and $\prn{\E_t \sq{X_t} - X_t}_{t \leq T}$, with $\eta = 1/2B$ and $\delta / 2$.	
  Note that $\E_t \sq{\prn{X_t - \E_t \sq{X_t}}^2} \leq \E_t \sq{X_t^2} \leq B \E_t\sq{X_t}$ if $0 \leq X_t \leq B$.
\end{proof}

We now define/recall some notations.
Denote $n_m \ldef \tau_{m} - \tau_{m-1}$.
Fix any epoch $m \in [M]$ and any time step $t$ within epoch $m$.
We have $f^{\star} = \eta$.
For any $f \in \cF$, we denote $M_t(f) \ldef Q_t \prn{ \prn{f(x_t) - y_t}^2 - \prn{f^\star(x_t) - y_t}^2}$, 
and $\wh R_m(f) \ldef \sum_{t=1}^{\tau_{m-1}} Q_t \prn{f(x_t) - y_t}^2$.
Recall that we have $Q_t = g_m(x_t)$.
We define filtration $\mfF_t \ldef \sigma \prn{ \prn{x_1, y_1}, \ldots , \prn{x_{t}, y_{t}}}$,\footnote{$y_t$ is not observed (and thus not included in the filtration) when $Q_t = 0$. Note that $Q_t$ is measurable with respect to $\sigma( (\mfF_{t-1}, x_t) ) $.} 
and denote $\E_t \sq{\cdot } \ldef \E \sq{\cdot \mid \mfF_{t-1}}$.
We next present concentration results with respect to a general set of regression function $\cF$ with finite pseudo dimension.

\begin{lemma}[\citet{krishnamurthy2019active}]
    \label{al_deep:lm:expected_sq_loss_pseudo}
    Consider an infinite set of regression function $\cF$.
   Fix any $\delta \in (0,1)$.  For any $\tau, \tau^\prime \in [T]$ such that $\tau < \tau^\prime$, with probability at least $1 - \frac{\delta}{2} $, we have 
   \begin{align*}
   	\sum_{t = \tau}^{\tau^\prime} M_t(f) \leq \sum_{t=\tau}^{\tau^\prime} \frac{3}{2} \E_t \brk{M_t(f)} + 
	C_\delta,
   \end{align*}
   and
   \begin{align*}
       \sum_{t = \tau}^{\tau^\prime} \E_t \sq{ M_t(f)}  \leq 2 \sum_{t = \tau}^{\tau^\prime} M_t(f) + C_\delta,
   \end{align*}
   where $C_\delta  = C \cdot \prn*{ \pseud(\cF) \cdot \log T +   \log \prn*{ \frac{\pseud(\cF)  \cdot  T}{\delta}} } $ with a universal constant $C >0$.

\end{lemma}

\subsubsection{Supporting Lemmas for \cref{al_deep:thm:abs_gen}}
\label{al_deep:app:supporting_abs}
Fix any classifier $\wh h: \cX \rightarrow \crl{0, 1,\bot}$. For any $x\in\cX$, we use the notion
\begin{align}
& 	\exc_\gamma ( \wh h;x) \ldef \nonumber\\
    &  \P_{y\mid x} \prn[\big]{y \neq \widehat h(x)} \cdot \ind \prn[\big]{ \widehat h(x) \neq \bot} + \prn[\big]{{1}/{2} - \gamma} \cdot \1 \prn[\big]{\widehat h(x) = \bot} - \P_{y\mid x} \prn[\big]{ y \neq h^\star(x) }\nonumber\\
    & = \ind \prn[\big]{ \widehat h(x) \neq \bot} \cdot \prn[\big]{\P_{y\mid x} \prn[\big]{y \neq \widehat h(x)} -  \P_{y\mid x} \prn[\big]{ y \neq h^\star(x) }} \nonumber \\
    & \quad + \ind \prn[\big]{ \widehat h(x) = \bot} \cdot \prn[\big]{ \prn[\big]{{1}/{2} - \gamma}  -  \P_{y\mid x} \prn[\big]{ y \neq h^\star(x) }} \label{al_deep:eq:excess_x}
\end{align}
to represent the excess error of $\wh h$ at point $x\in \cX$. Excess error of classifier $\wh h$ can be then written as $\exc_\gamma (\wh h) \ldef \err_\gamma(\wh h) - \err(h^{\star}) = \E_{x \sim \cD_\cX} \brk{ \exc_\gamma (\wh h;x)}$.

We let $\cE$ denote the good event considered in \cref{al_deep:lm:expected_sq_loss_pseudo}, we analyze under this event through out the rest of this section.
Most lemmas presented in this section are inspired by results provided \citet{zhu2022efficient}. Our main innovation is an inductive analysis of lemmas that eventually relaxes the requirements for approximation error for \cref{al_deep:thm:abs_gen}.

\paragraph{General lemmas} We introduce some general lemmas for \cref{al_deep:thm:abs_gen}. 

\begin{lemma}
\label{al_deep:lm:query_implies_width}
For any $m \in [M]$, we have $g_m(x)= 1 \implies w(x;\cF_m) > \frac{\gamma}{2}$.
\end{lemma}
\begin{proof}
We only need to show that $\ucb(x;\cF_m) - \lcb(x;\cF_m) \leq \frac{\gamma}{2} \implies g_m(x) = 0$. Suppose otherwise $g_m(x) = 1$, which implies that both 
\begin{align}
\label{al_deep:eq:query_condition}
&\frac{1}{2} \in \prn*{\lcb(x;\cF_m)-\frac{\gamma}{4}, \ucb(x;\cF_m) + \frac{\gamma}{4}}  \quad \text{ and } \nonumber \\
&{\brk*{\lcb(x;\cF_m) - \frac{\gamma}{4}, \ucb(x;\cF_m) + \frac{\gamma}{4}} \nsubseteq \brk*{  \frac{1}{2}-\gamma, \frac{1}{2} +\gamma } } .
\end{align}
If $\frac{1}{2} \in \prn*{\lcb(x;\cF_m)- \frac{\gamma}{4}, \ucb(x;\cF_m)+ \frac{\gamma}{4}}$ and $\ucb(x;\cF_m) - \lcb(x;\cF_m) \leq \frac{\gamma}{2}$, we must have $\lcb(x;\cF_m) \geq  \frac{1}{2}- \frac{3}{4}\gamma$ and $\ucb(x;\cF_m) \leq \frac{1}{2} + \frac{3}{4}\gamma$, which contradicts with \cref{al_deep:eq:query_condition}.
\end{proof}

\begin{lemma}
\label{al_deep:lm:regret_no_query_mis}
Fix any $m \in [M]$.
Suppose $\wb f \in \cF_m$,
we have $\exc_{\gamma}(\wh h_m ;x) \leq 0 $ if $g_m(x) = 0$.
\end{lemma}
\begin{proof}
	Recall that
\begin{align*}
	\exc_{\gamma}( \wh h;x) & =  \nonumber
      \ind \prn[\big]{ \widehat h(x) \neq \bot} \cdot \prn[\big]{\P_{y \mid x} \prn[\big]{y \neq \widehat h(x)} -  \P_{y \mid x} \prn[\big]{ y \neq  h^{\star}(x) }} \nonumber \\
    & \quad + \ind \prn[\big]{ \widehat h(x) = \bot} \cdot \prn[\big]{ \prn[\big]{{1}/{2} - \gamma}  -  \P_{y\mid x} \prn[\big]{ y \neq  h^{\star}(x) }} .
\end{align*}
We now analyze the event $\curly*{g_m(x)= 0}$ in two cases. 

\textbf{Case 1: ${\widehat h_m(x) = \bot} $.} 

Since $\wb f(x) \in [\lcb(x;\cF_m), \ucb(x;\cF_m)]$ and $\kappa \leq \frac{\gamma}{4}$ by assumption, 
we know that $\eta(x) = f^{\star}(x) \in \sq{ \frac{1}{2} - \gamma, \frac{1}{2} + \gamma}$ and thus $\P_{y} \prn[\big]{ y\neq h^\star(x) } \geq \frac{1}{2} - \gamma$. 
As a result, we have $\exc_\gamma(\wh h_m;x) \leq 0$.

\textbf{Case 2: ${\widehat h_m(x) \neq \bot}$ but ${\frac{1}{2} \notin \prn{\lcb(x;\cF_m) - \frac{\gamma}{4}, \ucb(x;\cF_m) + \frac{\gamma}{4}}} $.} 

Since $\wb f(x) \in [\lcb(x;\cF_m), \ucb(x;\cF_m)]$ and $\kappa \leq \frac{\gamma}{4}$ by assumption, 
we clearly have $\widehat h_m (x) = h^\star(x)$ when ${\frac{1}{2} \notin \prn{\lcb(x;\cF_m) - \frac{\gamma}{4}, \ucb(x;\cF_m) + \frac{\gamma}{4}}} $. We thus have 
$\exc_\gamma(\wh h_m;x) \leq 0$.
\end{proof}

\paragraph{Inductive lemmas}
We prove a set of statements for \cref{al_deep:thm:abs_gen} in an inductive way. 
Fix any epoch $m \in [M]$,
we consider 
\begin{align}
	\label{al_deep:eq:inductive1}
\begin{dcases}
	 \wh R_m(\wb f) - \wh R_m(f^{\star}) \leq \E_{t} \brk*{ Q_t\prn*{\wb f(x_t) - f^{\star}(x_t)}^2} + C_\delta  \leq \frac{3}{2} C_\delta \\
 \wb f\in \cF_m\\
 \sum_{t=1}^{\tau_{m-1}} \E_t \brk{M_t(f)} \leq 4 \beta_m , \forall f \in \cF_m \\
\sum_{t=1}^{\tau_{m-1}} \E \brk{ Q_t(x_t) \prn{ f(x_t) - \wb f(x_t) }^2} \leq 9 \beta_m,  \forall f \in \cF_m\\
\cF_{m} \subseteq \cF_{m-1}
\end{dcases},
\end{align}
\begin{align}
	\label{al_deep:eq:inductive2}
    \E_{x \sim \cD_\cX} \sq{\ind (g_m(x)= 1)} \leq \frac{144 \beta_m}{{\tau_{m-1}} \, \gamma^2} \cdot \theta^{\val}_{\wb f}\prn*{\cF, \gamma/4, \sqrt{\beta_m/\tau_{m-1}}} \leq \frac{144 \beta_m}{{\tau_{m-1}} \, \gamma^2} \cdot \wb \theta,
\end{align}
and 
\begin{align}
	\label{al_deep:eq:inductive3}
    	\E_{x \sim \cD_\cX} \sq{\ind(g_m(x) = 1)\cdot w(x;\cF_m)} \leq  \frac{72 \beta_m}{\tau_{m-1} \gamma} \cdot \theta^{\val}_{\wb f}\prn*{\cF, \gamma/4, \sqrt{\beta_m/\tau_{m-1}}}\leq \frac{72 \beta_m}{\tau_{m-1} \gamma} \cdot \wb \theta.
\end{align}
\begin{lemma}
\label{al_deep:lm:set_f_mis}
Fix any $\wb m = [M]$.
When $\wb m=1, 2$ or when \cref{al_deep:eq:inductive2} holds true for epochs $m = 2,3,\dots,\wb m-1$, then 
\cref{al_deep:eq:inductive1} holds true for epoch $m = \wb m$.
\end{lemma}
\begin{proof}
The statements in \cref{al_deep:eq:inductive1} clearly hold true for $m = \wb m =1$ since, by definition,  $\cF_0 = \cF$ and  $\sum_{t=1}^{0} \ldots = 0 $.
We thus only need to consider the case when $\wb m\geq 2$.
We next prove each of the five statements in \cref{al_deep:eq:inductive1} for epoch $m = \wb m$.
\begin{enumerate}
	\item 
	In the case when $\wb m = 2$, from \cref{al_deep:lm:expected_sq_loss_pseudo}, we know that 
	\begin{align*}
		\wh R_{\wb m}(\wb f) - \wh R_{\wb m}(f^{\star}) & \leq \sum_{t=1}^{\tau_{\wb m-1}} \frac{3}{2} \cdot
		\E_{t} \brk*{ Q_t\prn*{\wb f(x_t) - f^{\star}(x_t)}^2} + C_\delta \\
	& \leq 3 + C_\delta \leq \frac{3}{2} C_\delta,
	\end{align*}
	where the second line follows from the fact that $\tau_1 = 2$ (without loss of generality, we assume $C_\delta \geq 6$ here).

	We now focus on the case when  $\wb m \geq 3$.
	We have 
	\begin{align*}
		\wh R_{\wb m}(\wb f) - \wh R_{\wb m}(f^{\star}) & \leq \sum_{t=1}^{\tau_{\wb m-1}} \frac{3}{2} \cdot
		\E_{t} \brk*{ Q_t\prn*{\wb f(x_t) - f^{\star}(x_t)}^2} + C_\delta \\
	& \leq \frac{3}{2} \sum_{\check m = 1}^{\wb m-1} n_{\check m} \E_{x \sim \cD_\cX} \brk{ \ind(g_{\check m}(x) = 1)} \cdot \kappa^2 + C_\delta\\
	& \leq \frac{3}{2} \prn*{2+ \sum_{\check m = 2}^{\wb m-1} n_{\check m} \frac{144 \beta_{\check m} \cdot \wb \theta}{\tau_{\check m-1} \gamma^2}} \cdot  \kappa^2  + C_\delta \\
	& \leq \prn*{3+ \frac{144 \wb \theta}{\gamma^2} \cdot \prn*{\sum_{\check m = 2}^{\wb m-1}\beta_{\check m}} } \cdot  \kappa^2  + C_\delta \\
	& \leq \prn*{3+ \frac{432 \wb \theta \cdot M^2}{\gamma^2} \cdot  C_\delta } \cdot  \kappa^2  + C_\delta \\
	& \leq  \frac{3}{2} C_\delta,
	\end{align*}
	where the first line follows from \cref{al_deep:lm:expected_sq_loss_pseudo}; the second line follows from the fact that $\nrm{\wb f - f^{\star}}_{\infty} \leq \kappa$; the third line follows from \cref{al_deep:eq:inductive2}; the forth line follows from $n_{\check m} = \tau_{\check m - 1}$; the fifth line follows from the definition of $\beta_{\check m}$; and
	the last line follows from the choice of  $\kappa$ in \cref{al_deep:eq:kappa_requirement}
\item 
Since $\E_t \brk{ M_t(f)} = \E_{t}\brk{Q_t \prn{f(x_t) - f^{\star}(x_t)}^2}$,
by \cref{al_deep:lm:expected_sq_loss_pseudo}, we have 
$\wh R_{\wb m} (f^\star) \leq \wh R_{\wb m}(f) + C_\delta /2 $ for any $f \in \cF$.
Combining this with statement 1 leads to 
\begin{align*}
	\wh R_{\wb m}(\wb f) 
	& \leq \wh R_{\wb m}(f)  + 2 C_\delta\\
	& \leq \wh R_{\wb m}(f) + \beta_{\wb m}
\end{align*}
for any $f \in \cF$, where the second line follows from the definition of $\beta_{\wb m}$.
We thus have $\wb f \in \cF_{\wb m}$ based on the elimination rule.
\item  Fix any $f \in \cF_{\wb m}$. We have 
	\begin{align*}
		\sum_{t=1}^{\tau_{\wb m-1}} \E_t [M_t(f)] & \leq 2 \sum_{t=1}^{\tau_{\wb m-1}} M_t(f) + C_\delta \\
			& = 2 \wh R_{\wb m }(f) - 2\wh R_{\wb m}(f^{\star}) + C_\delta \\
			& \leq 2 \wh R_{\wb m}(f) - 2\wh R_{\wb m}(\wb f) + 4 C_\delta \\
			& \leq 2 \wh R_{\wb m }(f) - 2\wh R_{\wb m}(\wh f_{\wb m})+ 4 C_\delta \\
			& \leq 2 \beta_{\wb m} + 4 C_\delta \\
			& \leq 4 \beta_{\wb m} , 
	\end{align*}
	where 
the first line follows from \cref{al_deep:lm:expected_sq_loss_pseudo};
	the third line follows from statement 1; the fourth line follows from the fact that $\wh f_{\wb m}$ is the minimizer of $\wh R_{\wb m} (\cdot) $; and the fifth line follows from the fact that $f \in \cF_{\wb m}$.
\item  Fix any $f \in \cF_{\wb m}$. We have 
	\begin{align*}
    &\sum_{t=1}^{\tau_{\wb m-1}} \E_t \brk{ Q_t(x_t) \prn{ f(x_t) - \wb f(x_t) }^2}\\
		& = \sum_{t=1}^{\tau_{\wb m-1}} \E_t \brk{ Q_t(x_t) \prn{ (f(x_t) - f^{\star}(x_t)) + 
		( f^{\star}(x_t) - \wb f(x_t)) }^2} \\
		& \leq 2 \sum_{t=1}^{\tau_{\wb m-1}} \E_t \brk{ Q_t(x_t) \prn{ f(x_t) -  f^{\star}(x_t) }^2} + 2C_\delta\\
		&  = 2 \sum_{t=1}^{\tau_{\wb m-1}} \E_t [M_t(f)] + 2C_\delta  \\
		& \leq 8 \beta_{\wb m} + 2C_\delta\\
		& \leq 9 \beta_{\wb m},
	\end{align*}
	where the second line follows from $\prn{a+b}^2 \leq 2(a^2 + b^2)$ and (the proof of) statement 1 on the second line; and the fourth line follows from statement 3.
	\item Fix any $f \in \cF_{\wb m}$. We have 
	\begin{align*}
    & \wh R_{\wb m-1} (f) - \wh R_{\wb m-1} (\wh f_{\wb m-1})\\
    & \leq   
		\wh R_{\wb m-1} (f) - \wh R_{\wb m-1} (f^{\star}) + \frac{C_\delta}{2}\\
		& = \wh R_{\wb m}(f) - \wh R_{\wb m}(f^{\star}) 
		- \sum_{t=\tau_{\wb m-2}+1}^{\tau_{\wb m-1}} M_t(f) + \frac{C_\delta}{2}\\
		& \leq \wh R_{\wb m}(f) - \wh R_{\wb m} (\wb f) + \frac{3}{2} C_\delta
		- \sum_{t=\tau_{\wb m-2}+1}^{\tau_{\wb m-1}} \E_t [M_t(f)] /2 + {C_\delta}\\
		& \leq \wh R_{\wb m}(f) - \wh R_{\wb m} (\wh f_{\wb m}) + \frac{5}{2}C_\delta\\
		& \leq \beta_{\wb m} + 3 C_\delta\\
		& \leq \beta_{\wb m-1},
	\end{align*}	
	where the first line follows from \cref{al_deep:lm:expected_sq_loss_pseudo}; 
	the third line follows from statement 1 and \cref{al_deep:lm:expected_sq_loss_pseudo}; the fourth line follows from the fact that $\wh f_{\wb m}$ is the minimizer with respect to $\wh R_{\wb m}$ and \cref{al_deep:lm:expected_sq_loss_pseudo}; the last line follows from the construction of $\beta_{\wb m}$.
\end{enumerate}
\end{proof}

We introduce more notations.
    Denote $\prn{\cX, \Sigma, \cD_\cX}$ as the (marginal) probability space,
    and denote $\wb \cX_m \ldef \crl{x \in \cX: g_m(x) = 1} \in \Sigma$ be the region where query \emph{is} requested within epoch $m$.
    Under the prerequisites of \cref{al_deep:lm:conf_width_dis_coeff_mis} and \cref{al_deep:lm:per_round_regret_dis_coeff_mis} (i.e., \cref{al_deep:eq:inductive1} holds true for epochs $m = 1,2,\ldots,\wb m$), we have $\cF_{m} \subseteq \cF_{m - 1}$ for  $m = 1, 2, \ldots, \wb m$, which leads to
    $\wb \cX_{m} \subseteq \wb \cX_{m-1}$ for $m = 1, 2, \ldots, \wb m$.
    We now define a sub probability measure $\wb \mu_{m} \ldef ({\cD_\cX})_{\mid \wb \cX_{m}}$ such that $\wb \mu_{m}(\omega) = \cD_{\cX}\prn{ \omega \cap \wb \cX_{m}}$ for any $\omega \in \Sigma$. 
    Fix any epoch $m \leq \wb m$ and
    consider any measurable function $F$ (that is $\cD_\cX$ integrable), we have 
    \begin{align}
    	\E_{x \sim \cD_\cX} \brk*{ \ind(g_{\wb m}(x) = 1) \cdot F(x)}
	& = \int_{x \in \wb \cX_{\wb m}} F(x) \, d \cD_\cX(x) \nonumber \\ 
	& \leq \int_{x \in \wb \cX_{m}} F(x) \, d \cD_\cX(x)\nonumber \\ 
	& = \int_{x \in \cX} F(x) \, d \wb \mu_{m} (x) \nonumber \\ 
	& \rdef \E_{x \sim \wb \mu_{m}} \brk*{ F(x)}, \label{al_deep:eq:change_of_measure} 
    \end{align}
    where, by a slightly abuse of notations, we use $\E_{x \sim \mu} \sq{\cdot}$ to denote the integration with any sub probability measure $\mu$. 
    In particular, \cref{al_deep:eq:change_of_measure} holds with equality when $m = \wb m$.

\begin{lemma}
    \label{al_deep:lm:conf_width_dis_coeff_mis}
 Fix any epoch $\wb m \geq 2$. Suppose \cref{al_deep:eq:inductive1} holds true for epochs $m = 1,2,\ldots,\wb m$, we then have \cref{al_deep:eq:inductive2} holds true for epoch $m = \wb m$.
\end{lemma}

\begin{proof}
We prove \cref{al_deep:eq:inductive2} for epoch $m = \wb m$.
We know that $\ind(g_{\wb m}(x) = 1) = \ind (g_{\wb m}(x)= 1) \cdot \ind(w(x;\cF_{\wb m}) > \gamma/2 )$ from \cref{al_deep:lm:query_implies_width}. Thus, for any $\check m \leq \wb m$, we have 
    \begin{align}
	    \E_{x \sim \cD_\cX} \sq{\ind(g_{\wb m}(x)= 1)} 
	    &  = \E_{x \sim \cD_\cX} \sq{\ind(g_{\wb m}(x)= 1) \cdot \ind(w(x;\cF_{\wb m})> \gamma/2 )}\nonumber \\ 
	    & \leq \E_{x \sim \wb \mu_{\check m} } \sq{\ind(w(x;\cF_{\wb m})> \gamma/2  )}\nonumber \\
    & \leq \E_{x \sim \wb \mu_{\check m}} \prn[\Big]{ \ind \prn[\big]{\sup_{f \in \cF_{\wb m}} \abs*{f(x) - \wb f(x)} > \gamma/4}} , \label{al_deep:eq:conf_width_dis_coeff_1}
    \end{align}
where the second line uses \cref{al_deep:eq:change_of_measure} and the last line follows from the facts that
    $\wb f \in \cF_{\wb m}$ (by \cref{al_deep:eq:inductive1}) and $w(x;\cF_{\wb m}) > \gamma/2  \implies \exists f \in \cF_{\wb m}, \abs{f(x) - \wb f (x)} > {\gamma}/ {4}$. 

    For any time step $t$, let  $m(t)$ denote the epoch where  $t$ belongs to.
From \cref{al_deep:eq:inductive1}, we know that, $\forall f \in \cF_{\wb m}$,  
    \begin{align}
9 \beta_{\wb m} &
\geq \sum_{t=1}^{\tau_{\wb m -1}} \E_{t} \sq[\Big]{ Q_t \prn[\big]{f(x_t) - \wb f(x_t)}^2} \nonumber \\
       &  = \sum_{t=1}^{\tau_{\wb m -1}} \E_{x \sim \cD_\cX} \sq[\Big]{\ind(g_{m(t)}(x)=1) \cdot \prn[\big]{f(x) - \wb f(x)}^2} \nonumber \\
	  & = \sum_{\check m=1}^{\wb m-1} n_{\check m} \cdot \E_{x \sim \wb \mu_{\check m}} \brk*{  \prn*{f(x) - \wb f(x)}^2}\nonumber \\
	& = \tau_{\wb m-1} \E_{x \sim \wb \nu_{\wb m}} \brk*{  \prn*{f(x) - \wb f(x)}^2}, \label{al_deep:eq:conf_width_dis_coeff_2}
    \end{align}
    where we use $Q_t = g_{m(t)}(x_t) = \ind(g_{m(t)}(x) = 1)$ and \cref{al_deep:eq:change_of_measure} on the second line, and define a new sub probability measure 
    $$\wb \nu_{\wb m} \ldef \frac{1}{\tau_{\wb m-1}} \sum_{\check m =1}^{\wb m-1} n_{\check m} \cdot \wb \mu_{\check m}$$ on the third line.

    Plugging \cref{al_deep:eq:conf_width_dis_coeff_2} into \cref{al_deep:eq:conf_width_dis_coeff_1} leads to the bound 
    \begin{align*}
	& \E_{x \sim \cD_\cX} \sq{\ind(g_{\wb m}(x)= 1)} \\
	& \leq \E_{x \sim \wb \nu_{\wb m}} \sq[\bigg]{\ind \prn[\Big]{\exists f \in \cF, \abs[\big]{f(x) - \wb f(x)} > \gamma/4, \E_{x \sim \wb \nu_{\wb m}} \sq[\Big]{\prn[\big]{f(x) - \wb f(x)}^2} \leq \frac{9 \beta_{\wb m}}{\tau_{\wb m-1}}}},
    \end{align*}
    where we use the definition of $\wb \nu_{\wb m}$ again (note that \cref{al_deep:eq:conf_width_dis_coeff_1} works with any $\check m \leq \wb m$).  
    Based on the \cref{al_deep:def:dis_coeff_value},\footnote{Note that analyzing with a sub probability measure $\wb \nu$ does not cause any problem. See \citet{zhu2022efficient} for a detailed discussion.} we then have
    \begin{align*}
	& \E_{x \sim \cD_\cX} \sq{\ind(g_{\wb m}(x)= 1)}  \\
	& \leq  \frac{144 \beta_{\wb m}}{{\tau_{\wb m-1}} \, \gamma^2} \cdot \theta^{\val}_{\wb f}\prn*{\cF, \gamma/4, \sqrt{9 \beta_{\wb m}/2\tau_{\wb m-1}}}\\
	& \leq  \frac{144 \beta_{\wb m}}{{\tau_{\wb m-1}} \, \gamma^2} \cdot \theta^{\val}_{\wb f}\prn*{\cF, \gamma/4, \sqrt{ \beta_{\wb m}/\tau_{\wb m-1}}}\\
	& \leq \frac{144 \beta_{\wb m}}{{\tau_{\wb m-1}} \, \gamma^2} \cdot \wb \theta.
    \end{align*}
\end{proof}

\begin{lemma}
    \label{al_deep:lm:per_round_regret_dis_coeff_mis}
 Fix any epoch $\wb m \geq 2$. Suppose \cref{al_deep:eq:inductive1} holds true for epochs $m = 1,2,\ldots,\wb m$, we then have \cref{al_deep:eq:inductive3} holds true for epoch $m = \wb m$.
\end{lemma}

\begin{proof}
	We prove \cref{al_deep:eq:inductive3} for epoch $m = \wb m$.
    Similar to the proof of \cref{al_deep:lm:conf_width_dis_coeff_mis}, we have 
    \begin{align*}
      & \E_{x \sim \cD_\cX} \sq{\ind(g_{\wb m}(x)= 1)\cdot w(x;\cF_{\wb m})} \\
	& = \E_{x \sim \cD_\cX} \sq{\ind(g_{\wb m}(x)=1) \cdot \ind(w(x;\cF_{\wb m})> \gamma/2 )\cdot w(x;\cF_{\wb m})} \\
	& \leq \E_{x \sim \wb \mu_{\check m}} \sq{\ind(w(x;\cF_{\wb m})> \gamma/2 )\cdot w(x;\cF_{\wb m})}
    \end{align*}
    for any $\check m \leq \wb m$. 
With $\wb \nu_{\wb m} \ldef \frac{1}{\tau_{\wb m-1}} \sum_{\check m =1}^{\wb m-1} n_{\check m} \cdot \wb \mu_{\check m}$, we then have 
\begin{align*}
	 & \E_{x \sim \cD_\cX} \sq{\ind(g_{\wb m}(x)= 1)\cdot w(x;\cF_{\wb m})} \\
	& \leq \E_{x \sim \wb \nu_{\wb m}} \sq{\ind(w(x;\cF_{\wb m})> \gamma/2 )\cdot w(x;\cF_{\wb m})}\\
        & \leq \E_{x \sim \wb \nu_{\wb m}} \sq*{\ind \prn*{\sup_{f \in \cF_{\wb m}} \abs[\big]{f(x) - \wb f(x)} > \gamma/4}\cdot \prn*{\sup_{f , f^\prime \in \cF_{\wb m}} \abs*{f(x) - f^\prime(x)}}} \\
        & \leq 2\E_{x \sim \wb \nu_{\wb m}} \sq*{\ind \prn*{\sup_{f \in \cF_{\wb m}} \abs[\big]{f(x) - \wb f(x)} > \gamma/4}\cdot \prn*{\sup_{f \in \cF_{\wb m}} \abs{f(x) - \wb f(x)}}} \\
        & \leq 2 \int_{\gamma/4}^1 \E_{x \sim \wb \nu_{\wb m}} \sq*{\ind \prn*{\sup_{f \in \cF_{\wb m}} \abs[\big]{f(x) - \wb f(x)} \geq \omega}} \, d \, \omega \\
& \leq  2 \int_{\gamma/4}^1  \frac{1}{\omega^2} \, d \, \omega \cdot \prn*{ \frac{9 \beta_{\wb m}}{\tau_{\wb m-1}} \cdot \theta^{\val}_{\wb f}\prn*{\cF, \gamma/4, \sqrt{9 \beta_{\wb m}/2\tau_{\wb m-1}}}}\\
& \leq { \frac{72 \beta_{\wb m}}{\tau_{\wb m-1} \, \gamma} \cdot \theta^{\val}_{\wb f}\prn*{\cF, \gamma/4, \sqrt{\beta_{\wb m}/\tau_{\wb m-1}}}}\\
& \leq  \frac{72 \beta_{\wb m}}{\tau_{\wb m-1} \, \gamma} \cdot \wb \theta,
\end{align*}
where we follow similar steps as in the proof of \cref{al_deep:lm:conf_width_dis_coeff_mis} and use some basic arithmetic facts.
\end{proof}

\begin{lemma}
	\label{al_deep:lm:induction}
	\cref{al_deep:eq:inductive1}, \cref{al_deep:eq:inductive2} and \cref{al_deep:eq:inductive3} hold true for all $m \in [M]$.
\end{lemma}
\begin{proof}
	We first notice that, by \cref{al_deep:lm:set_f_mis}, \cref{al_deep:eq:inductive1} holds true for epochs $\wb m=1,2$ unconditionally.
	We also know that, by \cref{al_deep:lm:conf_width_dis_coeff_mis} and \cref{al_deep:lm:per_round_regret_dis_coeff_mis}, once \cref{al_deep:eq:inductive1} holds true for epochs $m = 1,2, \ldots, \wb m$, \cref{al_deep:eq:inductive2} and \cref{al_deep:eq:inductive3} hold true for epochs $m = \wb m$ as well; 
	at the same time, by \cref{al_deep:lm:set_f_mis}, once \cref{al_deep:eq:inductive2} holds true for epochs $m = 2, 3, \ldots, \wb m$, \cref{al_deep:eq:inductive1} will hold true for epoch $m = \wb m+1$.

	We thus can start the induction procedure from  $\wb m = 2$, and make sure that 
	\cref{al_deep:eq:inductive1}, \cref{al_deep:eq:inductive2} and \cref{al_deep:eq:inductive3} hold true for all $m \in [M]$.
\end{proof}

\subsection{Proof of \cref{al_deep:thm:abs_gen}}
\label{al_deep:app:abs_gen_proof}

\thmAbsGen*

\begin{proof}
	We analyze under the good event $\cE$ defined in \cref{al_deep:lm:expected_sq_loss_pseudo}, which holds with probability at least $1-\frac{\delta}{2}$. Note that all supporting lemmas stated in \cref{al_deep:app:supporting_abs} hold true under this event.

Fix any $m \in [M]$.
We analyze the Chow's excess error of $\wh h_m$, which is measurable with respect to $\mfF_{\tau_{m-1}}$. 
For any $x \in \cX$, if $g_m(x) = 0$, 
\cref{al_deep:lm:regret_no_query_mis} implies that $\exc_\gamma (\wh h_m ;x) \leq 0 $. 
If $g_m(x)= 1$, we know that $\wh h_m(x) \neq \bot$ and $\frac{1}{2} \in (\lcb(x;\cF_m) - \frac{\gamma}{4},\ucb(x;\cF_m) + \frac{\gamma}{4})$. 
Since $\wb f \in \cF_m$ by \cref{al_deep:lm:induction} (with \cref{al_deep:eq:inductive1}) and $\sup_{x \in \cX} \abs{ \wb f(x) - f^{\star}(x)} \leq \kappa \leq \gamma /4$ by construction. 
The error incurred in this case is upper bounded by 
\begin{align*}
	\exc(\wh h_m; x) 
	& \leq 2 \abs{ f^{\star}(x)- 1 /2}\\
	& \leq 2\kappa + 2 \abs{ \wb f(x)- 1 /2}\\
	& \leq 2\kappa + 2 w(x;\cF_m) + \frac{\gamma}{2} \\
	& \leq 4 w(x;\cF_m),
\end{align*}
where we use \cref{al_deep:lm:query_implies_width} in the last line.

Combining these two cases together, we have 
\begin{align*}
	\exc( \wh h_m) \leq 4 \, \E_{x \sim \cD_\cX} \brk{ \ind(g_m(x) = 1) \cdot w(x;\cF_m)}.	
\end{align*}
Take $m=M$ and apply \cref{al_deep:lm:induction} (with \cref{al_deep:eq:inductive3}) leads to the following guarantee.
\begin{align*}
	\exc( \wh h_M)
	& \leq   { \frac{ 576 \beta_M}{\tau_{M-1} \gamma} \cdot \theta^{\val}_{\wb  f}\prn*{\cF, \gamma/4, \sqrt{\beta_M/ \tau_{M-1}}}}\\
	& \leq  O \prn*{  \frac{ \pseud(\cF) \log ( T / \delta)}{T \, \gamma} \cdot \wb \theta} \\
	& = O \prn*{  \eps \cdot \log \prn*{\frac{\wb \theta \cdot \pseud(\cF)}{\eps \, \gamma \, \delta}}}, 
\end{align*}
where we 
use the fact that $T = \frac{\wb \theta \cdot \pseud(\cF)}{\eps \, \gamma}$.

We now analyze the label complexity (note that the sampling process of \cref{al_deep:alg:abs} stops at time $t = \tau_{M-1}$).
Note that $\E \brk{\ind(Q_t = 1) \mid \mfF_{t-1}} = \E_{x\sim\cD_\cX} \brk{ \ind(g_m(x) = 1) }$ for any epoch $m \geq 2$ and time step $t$ within epoch $m$. 
  Combining \cref{al_deep:lm:martingale_two_sides} and \cref{al_deep:lm:induction} (with \cref{al_deep:eq:inductive2}) leads to
    \begin{align*}
        \sum_{t=1}^{\tau_{M-1}} \ind(Q_t = 1) & \leq \frac{3}{2} \sum_{t=1}^{\tau_{M-1}} \E \sq{\ind(Q_t = 1) \mid \mfF_{t-1}} + 4 \log (2 /\delta)\\
        & \leq 3 + \frac{3}{2}\sum_{m=2}^{M-1}\frac{(\tau_m - \tau_{m-1}) \cdot 144 \beta_m}{{\tau_{m-1}} \, \gamma^2} \cdot \wb \theta + 4 \log (2 /\delta) \\
	& \leq 3 + 4 \log (2 /\delta) + O\prn*{  \frac{ M^2 \cdot \pseud(\cF) \cdot \log(T /\delta) \cdot \wb \theta }{\gamma^2}}
	\\ 
	& =O\prn*{  \frac{ M^2 \cdot \pseud(\cF) \cdot \log(T /\delta) \cdot \wb \theta }{\gamma^2}} 
    \end{align*}
    with probability at least $1-\delta$ (due to another application of \cref{al_deep:lm:martingale_two_sides} with confidence level $\delta /2$), 
  where we use the fact that $\beta_m \ldef 3 (M - m + 1) C_\delta$ and $C_\delta \ldef O\prn{\pseud(\cF) \cdot \log(T /\delta)}$.
\end{proof}

\section{Other Proofs and Supporting Results}

\subsection{Proofs and Supporting Results for \cref{al_deep:sec:noise_passive}}

\propVCApprox*
\begin{proof}
We take $\kappa = \eps^{\frac{1}{1+\beta}}$ in \cref{al_deep:thm:approx_sobolev} to construct a set of neural network classifiers $\cH_\dnn$ with  $W = O \prn{ \eps^{- \frac{d}{\alpha(1+\beta)}} \log \frac{1}{\eps}}$ total parameters arranged in $L = O \prn{ \log \frac{1}{\eps}}$ layers.	
According to \cref{al_deep:thm:vcd_nn}, we know 
\begin{align*}
\vcd( \cH_{\dnn}) = O \prn{\eps^{- \frac{d}{\alpha(1+\beta)}} \cdot \log^2 \prn{\eps^{-1}}} = \wt O \prn{ \eps^{- \frac{d}{\alpha(1+\beta)}}}.
\end{align*}
We now show that there exists a classifier $\wb h \in \cH_{\dnn}$ with small excess error.
Let $\wb h = h_{\wb f}$ be the classifier such that  $\nrm{ \wb f - \eta}_{\infty} \leq \kappa$. We can see that 
 \begin{align*}
	\exc(\wb h)
	& = \E \brk*{ \ind( \wb h(x) \neq y) - \ind( h^{\star}(x) \neq y)}\\
	& = \E \brk*{ \abs{ 2 \eta(x) - 1} \cdot \ind( \wb h(x) \neq h^{\star}(x))}\\
	& \leq 2 \kappa \cdot \P_{x \sim \cD_{\cX}} \prn*{ x \in \cX: \abs{\eta(x) - {1}/{2}} \leq \kappa}\\
	& = O \prn{ \kappa^{1 + \beta}}\\
	& = O(\eps),
\end{align*}
where the third line follows from the fact that $\wb h$ and  $h^{\star}$ disagrees only within region $\crl{x \in \cX: \abs{\eta(x) - 1 /2} \leq \kappa}$ and the incurred error is at most  $2 \kappa$ on each disagreed data point.
The fourth line follows from the Tsybakov noise condition and the last line follows from the selection of $\kappa$.
\end{proof}

Before proving \cref{al_deep:thm:passive_noise}, we first recall the excess error guarantee for empirical risk minimization under Tsybakov noise condition.

\begin{theorem}[\citet{boucheron2005theory}]
	\label{al_deep:thm:erm_tsy}
	Suppose $\cD_{\cX \cY}$ satisfies Tsybakov noise condition with parameter  $\beta \geq 0$.
	Consider a datatset $D_n = \crl{(x_i, y_i)}_{i=1}^{n}$ of $n$ points i.i.d. sampled from  $\cD_{\cX \cY}$.
	Let $\wh h \in \cH$ be the empirical risk minimizer on $D_n$.
	For any constant  $\rho > 0$, we have
	 \begin{align*}
		 & \err(\wh h)  - \min_{h \in \cH} \err(h) \\
		 &\leq \rho \cdot \prn{ \min_{h \in \cH}\err(h) - \err (h^{\star})}
		 + O \prn*{ \frac{\prn*{1 + \rho}^2}{\rho} \cdot \prn*{ \frac{\vcd(\cH) \cdot \log n}{n}}^{\frac{1+\beta}{2+\beta}} + \frac{\log\delta^{-1}}{n}},
	\end{align*}
	with probability at least $1-\delta$.
\end{theorem}

\thmPassiveNoise*

\begin{proof}
\cref{al_deep:prop:vc_approx} certifies $\min_{h \in \cH_\dnn} \err(h) - \err(h^{\star}) = O(\eps)$
and $\vcd(\cH_\dnn) = O \prn*{\eps^{-\frac{d}{\alpha(1+\beta)}} \cdot \log^2(\eps^{-1})}$.
Take $\rho = 1$ in \cref{al_deep:thm:erm_tsy}, leads to 
 \begin{align*}
& \err(\wh h) - \err(h^{\star}) \leq   O \prn*{ \eps + \prn*{ \eps^{-\frac{d}{\alpha(1+\beta)}} \cdot \log^2(\eps^{-1}) \cdot \frac{\log n}{n}}^{\frac{1+\beta}{2+\beta}} + \frac{\log\delta^{-1}}{n}},
\end{align*}
Taking $n = O \prn{\eps^{-\frac{d+2\alpha + \alpha \beta}{\alpha(1+\beta)}}\cdot \log (\eps^{-1}) + \eps^{-1} \cdot \log(\delta^{-1})} = \wt O \prn{\eps^{-\frac{d+2\alpha + \alpha \beta}{\alpha(1+\beta)}}}$ thus ensures that $\err(\wh h) - \err(h^{\star}) = O(\eps)$.
\end{proof}

\subsection{Proofs and Supporting Results for \cref{al_deep:sec:noise_active}}
We prove \cref{al_deep:thm:active_noise} in \cref{al_deep:app:active_noise_proof} and discuss the disagreement coefficient in \cref{al_deep:app:dis_coeff}.

\subsubsection{Proof of \cref{al_deep:thm:active_noise}}
\label{al_deep:app:active_noise_proof}

\thmActiveNoise*
\begin{proof}
	Construct $\cH_\dnn$ based on \cref{al_deep:prop:vc_approx} such that $\min_{h \in \cH_\dnn} \err(h) - \err(h^{\star}) = O(\eps)$ and $\vcd(\cH_\dnn) = \wt O \prn{ \eps^{-\frac{d}{\alpha(1+\beta)}} }$. Taking such $\cH_\dnn$ into \cref{al_deep:thm:RCAL_gen} leads to the desired result.
\end{proof}

\subsubsection{Discussion on Disagreement Coefficient in \cref{al_deep:thm:active_noise}}
\label{al_deep:app:dis_coeff}
We discuss cases when the (classifier-based) disagreement coefficient with respect to a set of neural networks is well-bounded.
As mentioned before, even for simple classifiers such as linear functions, the disagreement coefficient has been analyzed under additional assumptions \citep{friedman2009active, hanneke2014theory}. 
In this section, we analyze the disagreement coefficient for a set of neural networks under additional assumptions on $\cD_{\cX \cY}$ and $\cH_{\dnn}$ (assumptions on $\cH_{\dnn}$ can be implemented via proper preprocessing steps).
We leave a more comprehensive investigation of the disagreement coefficient for future work.

The first case is when $\cD_\cX$ is supported on countably many data points. 
The following result show strict improvement over passive learning.
\begin{definition}[Disagreement core]
	\label{al_deep:def:dis_core}
	For any hypothesis class $\cH$ and classifier $h$, the disagreement core of  $h$ with respect to $\cH$  under $\cD_{\cX \cY}$ is defined as 
	\begin{align}
		\partial_{\cH} h \ldef \lim_{r \rightarrow 0} \DIS \prn{ \cB_{\cH} (h, r) }.
	\end{align}
\end{definition}

\begin{proposition}[Lemma 7.12 and Theorem 7.14 in \citet{hanneke2014theory}]
For any hypothesis class $\cH$ and classifier $h$, we have 
$\theta_h(\eps) = o(1 /\eps)$ if and only if  $\cD_{\cX} \prn{ \partial_{\cH} h} = 0$.
In particular, this implies that 
$\theta_{\cH} (\eps) = o ( 1/ \eps)$ whenever $\cD_\cX$ is supported on countably many data points.	
\end{proposition}

We now discuss conditions under which we can upper bound the disagreement coefficient by $O(1)$, which ensures results in \cref{al_deep:thm:active_noise} matching the minimax lower bound for active learning, up to logarithmic factors.
We introduce the following \emph{decomposable} condition.
\begin{definition}
	\label{al_deep:def:decomposable}
	A marginal distribution $\cD_\cX$ is $\eps$-decomposable if its (known) support $\supp(\cD_\cX)$ can be decomposed into connected subsets, i.e.,  $\supp(\cD_\cX) = \cup_{i \in \cI} \cX_i$, such that 
\begin{align*}
	\cD_\cX \prn{\cup_{i \in \cI^{\prime}} \cX_i} = O( \eps),
\end{align*}
where $\cI^{\prime} \ldef \crl{i \in \cI: \cD_\cX(\cX_i) \leq \eps}$.
\end{definition}

\begin{remark}
	Note that \cref{al_deep:def:decomposable} permits a decomposition such that $\abs{\wb \cI} = \Omega(\frac{1}{\eps})$ where $\wb \cI = \cI \setminus \cI^{\prime}$. 
	\cref{al_deep:def:decomposable} requires no knowledge of the index set $\cI$ or any $\cX_i$; it also places
	no restrictions on the conditional probability on each $\cX_i$.
\end{remark}

We first give results for a general hypothesis class $\cH$ as follows, and then discuss how to bound the disagreement coefficient for a set of neural networks. 

\begin{proposition}
	\label{al_deep:prop:dis_coeff_Oh1}
Suppose $\cD_\cX$ is decomposable (into $\cup_{i \in \cI} \cX_i$) and
the hypothesis class $\cH$ consists of classifiers whose predication on each $\cX_i$ is the same, i.e., $\abs{\crl{h(x): x \in \cX_i}} = 1$ for any $h \in \cH$ and $i \in \cI$.
We then have $\theta_\cH(\eps) = O(1)$ for  $\eps$ sufficiently small. 
\end{proposition}

\begin{proof}
Fix any $h \in \cH$. 
we know that for any $h^{\prime} \in \cB_\cH (h, \eps)$, we must have $\DIS(\crl{h, h^{\prime}}) \subseteq \cup_{i \in \cI^{\prime}} \cX_i$ since 
  $\cD_\cX \prn{x \in \cX: h(x) \neq h^{\prime}(x)} \leq \eps$, and $\abs{\crl{h(x): x \in \cX_i}} = 1$ for any  $h \in \cH$ and any $\cX_i$. 
This further implies that $\P \prn{ \DIS(\cB_{\cH} (h, \eps)} = O(\eps)$, and thus  $\theta_{\cH}(\eps) = O(1)$.
\end{proof}

We next discuss conditions under which we can satisfy the prerequisites of \cref{al_deep:prop:dis_coeff_Oh1}. 
Suppose $\cD_{\cX \cY} \in \cP(\alpha, \beta)$. 
We assume that $\cD_\cX$ is  $(\eps^{\frac{\beta}{1+\beta}})$-decomposable, and, for the desired accuracy level $\eps$, we have 
\begin{align}
	\label{al_deep:eq:supp_far_away_one_half}
	\abs{\eta(x) - {1}/{2}} \geq 2 \eps^{\frac{1}{1+\beta}}, \quad \forall x \in \supp(\cD_\cX).
\end{align}

With the above conditions satisfied, we can 
filter out neural networks that are clearly not ``close'' to $\eta$. Specifically, with $\kappa = \eps^{\frac{1}{1+\beta}}$ and $\cF_\dnn$ be the set of neural networks constructed from \cref{al_deep:prop:vc_approx}, we consider 
\begin{align}
	\wt \cF_\dnn \ldef \crl{ f \in \cF_\dnn: \abs{f(x) - 1 /2} \geq \eps^{\frac{1}{1+\beta}}, \forall x \in \supp(\cD_\cX)},
\end{align}
which is guaranteed to contain $\wb f \in \cF_\dnn$ such that $\nrm{\wb f - \eta}_\infty \leq \eps^{\frac{1}{1+\beta}}$.
Now focus on the subset 
\begin{align}
	\wt \cH_{\dnn} \ldef \crl{ h_f: f \in \wt\cF_\dnn}.
\end{align}
We clearly have $h_{\wb f} \in \wt \cH_{\dnn}$ (which ensures an $O(\eps)$-optimal classifier) and 
\linebreak
$\vcd(\wt \cH_\dnn) \leq \vcd(\cH_\dnn)$ (since $\wt \cH_\dnn \subseteq \cH_\dnn$).
We upper bound the disagreement coefficient $\theta_{\wt \cH_\dnn}(\eps^{\frac{\beta}{1+\beta}})$ next.

\begin{proposition}
	\label{al_deep:prop:dis_coeff_Oh1_nn}
	Suppose $\cD_{\cX \cY} \in \cP(\alpha, \beta)$ such that $\cD_\cX$ is  $(\eps ^{\frac{\beta}{1+\beta}})$-decomposable and \cref{al_deep:eq:supp_far_away_one_half} is satisfied (with the desired accuracy level $\eps$).
	We then have $\theta_{\wt \cH_\dnn}(\eps^{\frac{\beta}{1+\beta}}) = O(1)$.
\end{proposition}

\begin{proof}
The proof is similar to the proof of \cref{al_deep:prop:dis_coeff_Oh1}.	
Fix any $h = h_f \in \wt \cH_\dnn$. 
We first argue that, for any $i \in \cI$, under \cref{al_deep:eq:supp_far_away_one_half}, $\abs{\crl{h_f(x): x \in \cX_i}} = 1$, i.e.,  for $x \in \cX_i$, $h_f(x)$ equals either $1$ or  $0$, but not both:
This can be seen from the fact that any $f \in \wt \cF_\dnn$ is continuous and satisfies $\abs{f(x) - 1 /2} \geq \eps^{\frac{1}{1+\beta}}$ for any  $x \in \cX_i$.

Fix any $h \in \wt \cH_\dnn$. 
We know that for any $h^{\prime} \in \cH_{\wt \cH_\dnn} (h, \eps^{\frac{\beta}{1+\beta}})$, we must have $\DIS(\crl{h, h^{\prime}}) \subseteq {\cup_{i \in \cI^{\prime}} \cX_i} $ due to similar reasons argued in the proof of \cref{al_deep:prop:dis_coeff_Oh1}. 
This further implies that $\P \prn{ \DIS(\cB_{\wt \cH_\dnn} (h, \eps^{\frac{\beta}{1+\beta}})} = O(\eps^{\frac{\beta}{1+\beta}})$, and thus  $\theta_{\wt \cH_\dnn}(\eps^{\frac{\beta}{1+\beta}}) = O(1)$.
\end{proof}

We next argue that \cref{al_deep:eq:supp_far_away_one_half} is only needed in an approximate sense. 
We define the approximate decomposable condition in the following.

\begin{definition}
	\label{al_deep:def:decomposable_approx}
	A marginal distribution $\cD_\cX$ is $(\eps,\delta)$-decomposable if there exists a known subset $\wb \cX \subseteq \supp(\cD_\cX)$ such that 
\begin{align}
	\cD_\cX(\wb \cX) \geq 1- \delta,
\end{align}
and it can be decomposed into connected subsets, i.e.,  $\wb \cX = \cup_{i \in \cI} \cX_i$, such that 
\begin{align*}
	\cD_\cX \prn{\cup_{i \in \cI^{\prime}} \cX_i} = O( \eps),
\end{align*}
where $\cI^{\prime} \ldef \crl{i \in \cI: \cD_\cX(\cX_i) \leq \eps}$.
\end{definition}

Suppose $\cD_{\cX \cY} \in \cP(\alpha, \beta)$. 
We assume that $\cD_\cX$ is  $(\eps^{\frac{\beta}{1+\beta}}, \eps^{\frac{\beta}{1+\beta}})$-decomposable (wrt $\wb \cX \subseteq \cD_\cX$), and, for the desired accuracy level $\eps$, we have 
\begin{align}
	\label{al_deep:eq:supp_far_away_one_half_approx}
	\abs{\eta(x) - {1}/{2}} \geq 2 \eps^{\frac{1}{1+\beta}}, \quad \forall x \in \wb \cX.
\end{align}

With the above conditions satisfied, we can 
filter out neural networks that are clearly not ``close'' to $\eta$. Specifically, with $\kappa = \eps^{\frac{1}{1+\beta}}$ and $\cF_\dnn$ be the set of neural networks constructed from \cref{al_deep:prop:vc_approx}, we consider 
\begin{align}
	\wb \cF_\dnn \ldef \crl{ f \in \cF_\dnn: \abs{f(x) - 1 /2} \geq \eps^{\frac{1}{1+\beta}}, \forall x \in \wb \cX},
\end{align}
which is guaranteed to contain $\wb f \in \cF_\dnn$ such that $\nrm{\wb f - \eta}_\infty \leq \eps^{\frac{1}{1+\beta}}$.
Now focus on the subset 
\begin{align}
	\wb \cH_{\dnn} \ldef \crl{ h_f: f \in \wb \cF_\dnn}.
\end{align}
We clearly have $h_{\wb f} \in \wb \cH_{\dnn}$ (which ensures an $O(\eps)$-optimal classifier) and
\linebreak
$\vcd(\wb \cH_\dnn) \leq \vcd(\cH_\dnn)$ (since $\wb \cH_\dnn \subseteq \cH_\dnn$).
We upper bound the disagreement coefficient $\theta_{\wb \cH_\dnn}(\eps^{\frac{\beta}{1+\beta}})$ next.

\begin{proposition}
	\label{al_deep:prop:dis_coeff_Oh1_nn_approx}
	Suppose $\cD_{\cX \cY} \in \cP(\alpha, \beta)$ such that $\cD_\cX$ is  $(\eps ^{\frac{1}{1+\beta}}, \eps)$-decomposable (wrt known $\wb \cX \subseteq \supp(\cD_\cX)$) and \cref{al_deep:eq:supp_far_away_one_half_approx} is satisfied (with the desired accuracy level $\eps$).
	We then have $\theta_{\wb \cH_\dnn}(\eps^{\frac{\beta}{1+\beta}}) = O(1)$.
\end{proposition}

\begin{proof}
The proof is the same as the proof of \cref{al_deep:prop:dis_coeff_Oh1_nn_approx} except 
for any $h^{\prime} \in \cH_{\wb \cH_\dnn} (h, \eps^{\frac{\beta}{1+\beta}})$, we must have $\DIS(\crl{h, h^{\prime}}) \subseteq \prn{\cup_{i \in \cI^{\prime}} \cX_i} \cup \prn{\supp(\cD_\cX) \setminus \wb \cX}$. 
Based on the assumption that $\cD_\cX$ is  $(\eps ^{\frac{1}{1+\beta}}, \eps)$-decomposable, this also leads to
\linebreak
  $\theta_{\wb \cH_\dnn}(\eps^{\frac{\beta}{1+\beta}}) = O(1)$.
\end{proof}

\subsection{Proofs and Supporting Results for \cref{al_deep:thm:abs}}
\label{al_deep:app:abs}

We provide prerequisites in \cref{al_deep:app:abs_prereq,al_deep:app:abs_prereq2} and the preprocessing procedures in \cref{al_deep:app:filtering}. We give the proof of \cref{al_deep:thm:abs} in \cref{al_deep:app:abs_proof}.

\subsubsection{Upper Bounds on Pseudo Dimension}
\label{al_deep:app:abs_prereq}

We present a result regarding the approximation and an upper bound on the pseudo dimension (i.e., \cref{al_deep:def:pseudo_d}). 
\begin{proposition}
	\label{al_deep:prop:pd_approx}
	Suppose $\cD_{\cX \cY} \in \cP(\alpha, \beta)$. 
	One can construct a set of neural network regression functions  $\cF_\dnn$ such that the following two properties hold simultaneously:
	 \begin{align*}
		\exists f \in \cF_\dnn \text{ s.t. } \nrm{ f - f^{\star}}_\infty \leq \kappa, \quad \text{ and } \quad \pseud(\cF_\dnn) \leq c \cdot {\kappa^{-\frac{d}{\alpha}} \log^2 (\kappa^{-1})},
	\end{align*}
	where $c > 0$ is a universal constant.
\end{proposition}

\begin{proof}
	The result follows by combining \cref{al_deep:thm:approx_sobolev} and \cref{al_deep:thm:pdim_nn}.
\end{proof}

\subsubsection{Upper Bounds on Value Function Disagreement Coefficient}
\label{al_deep:app:abs_prereq2}

We derive upper bounds on the value function disagreement coefficient (i.e., \cref{al_deep:def:dis_coeff_value}). We first introduce the (value function) eluder dimension, a complexity measure that is closely related to the value function disagreement coefficient \citet{russo2013eluder, foster2020instance}.

\begin{definition}[Value function eluder dimension]
\label{al_deep:def:eluder}
For any $f^{\star} \in \cF$ and $\gamma_0 > 0$, let $\check{\mathfrak{e}}_{f^{\star}}(\cF, \gamma)$ be the length of the longest sequence of data points $x^{1}, \dots, x^{m}$ such that for all $i$, there exists $f^{i} \in \cF$ such that 
\begin{align*}
    \abs{ f^{i}(x^{i}) - f^{\star}(x^{i}) } > \gamma, \quad \text{ and } \quad \sum_{j < i} \paren{ f^{i}(x^{j})  - f^{\star}(x^{j})}^2 \leq \gamma^2.
\end{align*}
The value function eluder dimension is defined as $\mathfrak{e}_{f^{\star}}(\cF, \gamma_0) \coloneqq \sup_{\gamma > \gamma_0} \check{\mathfrak{e}}_{f^{\star}}(\cF, \gamma)$. 
\end{definition}

The next result shows that the value function disagreement coefficient can be upper bounded by eluder dimension.
\begin{proposition}[\citet{foster2020instance}]
    \label{al_deep:prop:eluder_star_dis}
    Suppose $\cF$ is a uniform Glivenko-Cantelli class.
    For any $f^{\star}: \cX \rightarrow [0,1]$ and $\gamma, \eps > 0$,
    we have $\theta^{\val}_{f^\star}(\cF, \gamma, \epsilon) \leq 4 \, {\mfe_{f^\star}(\cF, \gamma)}$.
\end{proposition}

We remark here that the requirement that $\cF$ is a uniform Glivenko-Cantelli class is rather weak: It is satisfied as long as  $\cF$ has finite pseudo dimension \citep{anthony2002uniform}.

In the following, we only need to derive upper bounds on the value function eluder dimension, which upper bounds on the value function disagreement coefficient.\footnote{We focus on Euclidean geometry on $\cX$ (i.e., using $\nrm{\cdot}_2$ norm) in deriving the upper bound. Slightly tighter bounds might be possible with other norms.}
We first define two definitions: (i) the standard definition of covering number (e.g., see \citet{wainwright2019high}), and (ii) a newly-proposed definition of approximate Lipschitzness.

\begin{definition}
	\label{al_deep:def:covering}
An $\iota$-covering of a set  $\cX$ with respect to a metric  $\rho$ is a set  $\crl{x_1, \ldots,x_N} \subseteq \cX$ such that for each $x \in \cX$, there exists some $i \in [N]$ such that $\rho(x, x_i) \leq \iota$. The $\iota$-covering number  $\cN(\iota; \cX, \rho)$ is the cardinality of the smallest  $\iota$-cover.
\end{definition}

\begin{definition}
	\label{al_deep:def:lip_approx}
	We call a function $f: \cX \rightarrow \R$ $(L,\kappa)$-approximate Lipschitz if 
	\begin{align*}
		\abs{ f(x) - f(x^{\prime}) } \leq L \cdot \nrm{x - x^{\prime}}_2 + \kappa
	\end{align*}
	for any $x, x^{\prime} \in \cX$.
\end{definition}

We next provide upper bounds on value function eluder dimension and value function disagreement coefficient.

\begin{theorem}
\label{al_deep:thm:eluder_lip_approx}
Suppose $\cF$ is a set of $(L,\kappa /4)$-approximate Lipschitz functions. 
For any $\kappa^{\prime} \geq \kappa$,
we have $\sup_{f \in \cF} \mfe_f(\cF, \kappa^{\prime}) \leq 17 \cdot \cN(\frac{\kappa^{\prime}}{8L}; \cX, \nrm{\cdot}_2)$.
\end{theorem}
\begin{proof}
Fix any $f \in \cF$ and $\wb \kappa \geq \kappa^{\prime}$. We first give upper bounds on $\check \mfe_f(\cF, \wb \kappa)$.

We construct $\cG \ldef \cF - f$, which is a set of $(2L,\kappa /2)$-Lipschitz functions. 
Fix any eluder sequence $x^1, \dots, x^m$ at scale $\wb \kappa$ and any $\check x \in \cX$. We claim that $\abs{ \crl{x_j}_{j\leq m} \cap \cS  } \leq 17$ where $\cS \ldef \crl{ x \in \cX: \nrm{x-\check x}_2 \leq \frac{\wb \kappa}{8L}}$.
Suppose $\crl{x_j}_{j\leq m} \cap \cS = x_{j_1}, \dots, x_{j_k}$ ($j_i$ is ordered based on the ordering of $\crl{x_j}_{j\leq m}$). 
Since $x^{j_k}$ is added into the eluder sequence, there must exists a $g^{j_k} \in \cG$ such that  
\begin{align}
\label{al_deep:eq:eluder_lip}
    \abs{g^{j_k}(x^{j_k})} > \wb \kappa, \quad \text{ and } \quad 
    \sum_{j<j_k} \prn{g^{j_k}(x^j)}^2 \leq \wb \kappa^2.
\end{align}
Since $g^{j_k}$ is $(2L,\kappa /2)$-Lipschitz, $\wb \kappa \geq \kappa^{\prime}\geq \kappa$ and $x^{j_k} \in \cS$, we must have $g^{j_k}(x) \geq \frac{\wb \kappa}{4}$ for any $x \in \cS$. As a result, we must have $\abs{ \crl{x_j}_{j<j_k} \cap \cS^i  } \leq 16$
as otherwise the second constraint in \cref{al_deep:eq:eluder_lip} will be violated.
We cover the space $\cX$ with $\cN(\frac{\wb \kappa}{8L}; \cX, \nrm{\cdot}_2)$ balls of radius $\frac{\wb \kappa}{8L}$. Since the eluder sequence contains at most $17$ data points within each ball, 
we know that $\check \mfe_f(\cF, \wb \kappa) \leq 17 \cdot\cN(\frac{\wb \kappa}{8L}; \cX, \nrm{\cdot}_2) $.

The desired result follows by noticing that $17 \cdot\cN(\frac{\wb \kappa}{8L}; \cX, \nrm{\cdot}_2)$ is non-increasing in $\wb \kappa$.
\end{proof}

\begin{corollary}
	\label{al_deep:cor:eluder_lip_approx}
	Suppose $\cX \subseteq \B^{d}_r \ldef \crl{x \in \R^{d}: \nrm{x}_2 \leq r}$
and $\cF$ is a set of $(L,\kappa /4)$-approximate Lipschitz functions. 
For any $\kappa^{\prime} \geq \kappa$,
there exists a universal constant $c > 0$, such that
$\theta^{\val}_\cF (\kappa^{\prime}) \ldef \sup_{f \in \cF, \iota > 0} \theta_f^{\val}(\cF, \kappa^{\prime}, \iota) \leq c \cdot \prn{\frac{Lr}{\kappa^{\prime}}}^{d}$.
\end{corollary}
\begin{proof}
It is well-known that $\cN(\iota; \B^{d}_r, \nrm{\cdot}_2) \leq \prn*{1 + 2r / \iota}^{d}$ \citep{wainwright2019high}. The desired result thus follows from combining \cref{al_deep:thm:eluder_lip_approx} with \cref{al_deep:prop:eluder_star_dis}.	
\end{proof}

\subsubsection{The Preprocessing Step: Clipping and Filtering}
\label{al_deep:app:filtering}

Let $\eta: \cX \rightarrow [0,1]$ denote the true conditional probability and 
$\cF_\dnn$ denote a set of neural network regression functions (e.g., constructed based on \cref{al_deep:thm:approx_sobolev}).
We assume that (i) $\eta$ is  $L$-Lipschitz, and (ii) there exists a $f \in \cF$ such that $\nrm{f - \eta}_\infty \leq \kappa$ for some approximation factor $\kappa > 0$.
We present the preprocessing step below in \cref{al_deep:alg:preprocess}.

\begin{algorithm}[H]
	\caption{The Preprocessing Step: Clipping and Filtering}
	\label{al_deep:alg:preprocess} 
	\renewcommand{\algorithmicrequire}{\textbf{Input:}}
	\renewcommand{\algorithmicensure}{\textbf{Output:}}
	\newcommand{\algorithmicbreak}{\textbf{break}}
    \newcommand{\BREAK}{\STATE \algorithmicbreak}
	\begin{algorithmic}[1]
		\REQUIRE A set of regression functions $\cF$, Lipschitz parameter $L > 0$, approximation factor  $\kappa > 0$.
		\STATE \textbf{Clipping.}
		Set $\check \cF \ldef \crl{\check f: f \in \cF} $, where, for any $f \in \cF$, we denote
\begin{align*}
	\check f(x) \ldef 
	\begin{cases}
		1, & \text{ if } f(x) \geq 1;\\
		0, & \text{ if } f(x) \leq 0;\\
		f(x), & \text{ o.w. }
	\end{cases}
\end{align*}
		\STATE \textbf{Filtering.} Set $\wt \cF \ldef \crl{\check f \in \check \cF: \check f \text{ is $(L,2\kappa)$-approximate Lipschitz}}$
		\STATE \textbf{Return} $\wt \cF$.
	\end{algorithmic}
\end{algorithm}

\begin{proposition}
	\label{al_deep:prop:preprocess}
Suppose $\eta$ is  $L$-Lipschitz and  $\cF_\dnn$ is a set of neural networks (of the same architecture) with $W$ parameters arranged in  $L$ layers such that there exists a  $f \in \cF_\dnn$ with  $\nrm{f - \eta}_\infty \leq \kappa$. Let  $\wt \cF_\dnn$ be the set of functions obtained by applying \cref{al_deep:alg:preprocess} on  $\cF_\dnn$, we then have 
(i)  $\pseud(\wt \cF_\dnn) = O \prn{WL \log (W)}$, and (ii) there exists a $\wt f \in \wt \cF_\dnn$ such that $\nrm{\wt f - \eta}_\infty \leq \kappa$.
\end{proposition}
\begin{proof}
Suppose $f$ is a neural network function, we first notice that the ``clipping'' step can be implemented by adding one additional layer with $O(1)$ additional parameters for each neural network function. More specifically, fix any $f: \cX \rightarrow \R$, we can set $\check f(x) \ldef \relu(f(x)) - \relu(f(x) - 1)$. 
Set $\check \cF_\dnn \ldef \crl{\check f: f \in \cF_\dnn} $, we then have $\pseud(\check \cF_\dnn) = O (WL \log (W)) $ based on \cref{al_deep:thm:pdim_nn}. Let $\wt \cF_\dnn$ be the filtered version of  $\check \cF_\dnn$.
Since $\wt \cF_\dnn \subseteq \check \cF_\dnn$, we have $\pseud(\wt \cF_\dnn) = O \prn{WL \log (W)}$.

Since $\eta: \cX \rightarrow [0,1]$, we have 
	$\nrm{\check f - \eta }_\infty \leq \nrm{f - \eta }_\infty$, which implies that there must exists a $\check f \in \check \cF_\dnn$ such  $\nrm{\check f - \eta}_\infty \leq \kappa$. To prove the second statement, it suffices to show that the $\check f \in \check \cF$ that achieves  $\kappa$ approximation error is not removed in the ``filtering'' step, i.e., $\check f$ is  $(L, 2\kappa)$-approximate Lipschitz.
For any $x, x^\prime \in \cX$, we have 
 \begin{align*}
	 \abs{ \check f(x) - \check f(x^{\prime})}
	 & = \abs{ \check f(x) - \eta(x) + \eta(x) - \eta(x^{\prime}) + \eta(x^{\prime}) - \check f(x^{\prime})}\\
	 & \leq L \nrm{x - x^{\prime}}_2 + 2 \kappa,
\end{align*}
where we use the $L$-Lipschitzness of $\eta$ and the fact that $\nrm{ \check f - \eta}_{\infty} \leq \kappa$.
\end{proof}
\begin{proposition}
	\label{al_deep:prop:clip_filter}
	Suppose $\eta$ is  $L$-Lipschitz and  $\cX \subseteq \B^{d}_r$.
	Fix any $\kappa \in (0, \gamma / 32]$. There exists a set of neural network regression functions $\cF_\dnn$ such that the followings hold simultaneously.
	\begin{enumerate}
		\item 	$\pseud(\cF_\dnn) \leq c \cdot {\kappa^{-\frac{d}{\alpha}} \log^2(\kappa^{-1})}$ with a universal constant $c > 0$.
		\item There exists a $\wb f \in \cF_\dnn$ such that $\nrm{\wb f - \eta}_\infty \leq \kappa$.
		\item $ \theta^{\val}_{\cF_\dnn}(\gamma / 4) \ldef \sup_{f \in \cF_\dnn, \iota > 0} \theta^{\val}_f (\cF_\dnn, \gamma /4, \iota) \leq c^{\prime} \cdot \prn{\frac{Lr}{\gamma}}^{d}$ with a universal constant $c^{\prime} > 0$.
	\end{enumerate}
\end{proposition}
\begin{proof}
Let $\cF_\dnn$ be obtained by (i) invoking \cref{al_deep:thm:approx_sobolev} with approximation level $\kappa $, and (ii) invoking \cref{al_deep:alg:preprocess} on the set of functions obtained in step (i).
The first two statements follow from \cref{al_deep:prop:preprocess}, and the third statement follows from \cref{al_deep:cor:eluder_lip_approx} (note that to achieve guarantees for disagreement coefficient at level $\gamma / 4$, we need to have  $\kappa \leq \gamma / 32$ when invoking \cref{al_deep:thm:approx_sobolev}). 
\end{proof}

\subsubsection{Proof of \cref{al_deep:thm:abs}}
\label{al_deep:app:abs_proof}

\thmAbs*

\begin{proof}
Let line 1 of \cref{al_deep:alg:NCALP} be the set of neural networks $\cF_\dnn$ generated from \cref{al_deep:prop:clip_filter} with approximation level $\kappa \in (0, \gamma / 32]$ (and constants $c, c^{\prime}$ specified therein).
To apply results derived in \cref{al_deep:thm:abs_gen}, we need to satisfying \cref{al_deep:eq:kappa_requirement}, i.e., specifying an approximation level $\kappa \in (0, \gamma / 32]$ such that the following holds true
 \begin{align*}
	\frac{1}{\kappa^2} \geq \frac{4320 \cdot c^{\prime} \cdot \prn{\frac{Lr}{\gamma}}^{d} \cdot \prn*{\ceil*{\log_2 \prn*{\frac{c^{\prime} \cdot \prn{\frac{Lr}{\gamma}}^{d} \cdot c \cdot \prn{\kappa^{-\frac{d}{\alpha}} \log^2 (\kappa^{-1})}}{\eps \, \gamma}}}}^{2} }{\gamma^2}
\end{align*}

For the setting we considered, i.e., $\cX = [0,1]^{d}$ and $\eta \in \cW_{1}^{\alpha, \infty}(\cX)$, we have $r = \sqrt{d} = O(1)$ and $L \leq \sqrt{d} = O(1)$ (e.g., see Theorem 4.1 in \citet{heinonen2005lectures}).\footnote{Recall that we ignore constants that can be potentially $\alpha$-dependent and $d$-dependent.}
We thus only need to select a $\kappa \in (0, \gamma /32]$ such that 
\begin{align*}
	\frac{1}{\kappa} \geq \wb c \cdot \prn*{\frac{1}{\gamma}}^{\frac{d}{2}+1} \cdot \prn*{\log \frac{1}{\eps \, \gamma} + \log \frac{1}{\kappa}},
\end{align*}
with a universal constant $\wb c > 0$ (that is possibly  $d$-dependent and $\alpha$-dependent).
Since $x \geq 2a \log a \implies x \geq a \log x$ for any  $a > 0$, we can select a  $\kappa > 0$ such that 
 \begin{align*}
	\frac{1}{\kappa} = {\check c \cdot \prn*{\frac{1}{\gamma}}^{\frac{d}{2} + 1} \cdot \log \frac{1}{\eps \, \gamma}}
\end{align*}
with a universal constant $\check c > 0$.
With such choice of $\kappa$, from \cref{al_deep:prop:clip_filter}, we have  
\begin{align*}
\pseud(\cF_\dnn) = O\prn*{\prn*{\frac{1}{\gamma}}^{\frac{d^2+d}{2\alpha}} \cdot \polylog \prn*{\frac{1}{\eps \, \gamma}} }.
\end{align*}
Plugging this bound on $\pseud(\cF_\dnn)$ and the upper bound on $\theta^{\val}_{\cF_\dnn}(\gamma / 4)$ from \cref{al_deep:prop:clip_filter} into the guarantee of \cref{al_deep:thm:abs_gen} leads to $\exc_\gamma(\wh h) = O \prn{\eps \cdot \log \prn{\frac{1}{\eps \, \gamma \, \delta}}}$ after querying 
\begin{align*}
	O \prn*{\prn*{\frac{1}{\gamma}}^{d + 2 + \frac{d^2+d}{2\alpha}} \cdot \polylog \prn*{\frac{1}{\eps \, \gamma \, \delta}} }
\end{align*}
labels.
\end{proof}

\subsection{Other Proofs and Supporting Results for \cref{al_deep:sec:abstention}}
We discuss the proper abstention property of classifier learned in \cref{al_deep:alg:NCALP} and its exponential speedups under standard excess error and Massart noise in \cref{al_deep:app:proper}. We discuss the computational efficiency of \cref{al_deep:alg:NCALP} in \cref{al_deep:app:computational}. We provide the proof of \cref{al_deep:thm:single_relu_lb} in \cref{al_deep:app:single_relu_lb}.

\subsubsection{Proper Abstention and Exponential Speedups under Massart Noise}
\label{al_deep:app:proper}

We first recall the definition of \emph{proper abstention} introduced in \citet{zhu2022efficient}.
\begin{definition}[Proper abstention]
\label{al_deep:def:proper_abstention}
A classifier $\widehat h : \cX \rightarrow \cY \cup \curly*{\bot}$ enjoys proper abstention if and only if it abstains in regions where abstention is indeed the optimal choice, i.e., 
$\crl[\big]{x \in \cX: \widehat h(x) = \bot} \subseteq \crl*{x \in \cX: \eta(x) \in \brk*{\frac{1}{2} - \gamma , \frac{1}{2} + \gamma  } } \rdef \cX_\gamma$.
\end{definition}

We next show that the classifier $\wh h$ returned by \cref{al_deep:alg:abs} enjoys the proper abstention property. 
We also convert the abstaining classifier $\widehat h: \cX \rightarrow \cY \cup \curly*{\bot}$ into a standard classifier $\check h: \cX \rightarrow \cY$ and quantify its standard excess error.
The conversion is through randomizing the prediction of $\wh h$ over its abstention region, i.e., if $\wh h(x) = \bot$, then its randomized version $\check h(x)$ predicts $0$ and $1$ with equal probability \citep{puchkin2021exponential}.

\begin{proposition}
\label{al_deep:prop:proper_abstention}
The classifier $\wh h$ returned by \cref{al_deep:alg:abs} enjoys proper abstention. With randomization over the abstention region, we have the following upper bound on its standard excess error
\begin{align}
	\label{al_deep:eq:prop_abstention}
    \err(\check h) - \err(h^\star)  
     = \err_{\gamma}(\widehat h) - \err(h^\star) + \gamma \cdot \P_{x \sim \cD_{\cX}} (x \in \cX_{\gamma}).
\end{align}
\end{proposition}

\begin{proof}
The proper abstention property of $\wh h$ returned by \cref{al_deep:alg:abs} is achieved via conservation: $\wh h$ will avoid abstention unless it is absolutely sure that abstention is the optimal choice (also see the proof of \cref{al_deep:lm:regret_no_query_mis}.

Let $\check h: \cX \rightarrow \cY$ be the randomized version of $\wb h: \cX \rightarrow \crl{0, 1, \bot}$ (over the abstention region  $\crl{x \in \cX: \wh h(x) = \bot} \subseteq \cX_\gamma$).
We can see that, compared to the Chow's abstention error $1 /2 - \gamma$, the additional error incurred over the abstention region is exactly $\gamma \cdot \P_{x \sim \cD_{\cX}} (x \in \cX_{\gamma})$. 
We thus have
\begin{align*}
    \err(\widehat h) - \err(h^\star)  
     \leq \err_{\gamma}(\widehat h) - \err(h^\star) + \gamma \cdot \P_{x \sim \cD_{\cX}} (x \in \cX_{\gamma}).
\end{align*}
\end{proof}

To characterize the standard excess error of classifier with proper abstention, we only need to upper bound the term $ \P_{x \sim \cD_{\cX}} (x \in \cX_{\gamma})$, which does \emph{not} depends on the (random) classifier $\wh h$. Instead, it only depends on the marginal distribution. 

We next introduce the Massart \citep{massart2006risk}, which can be viewed as the extreme version of the Tsybakov noise by sending $\beta \rightarrow \infty$. 
\begin{definition}[Massart noise]
	\label{al_deep:def:massart}
  A distribution $\cD_{\cX \cY}$ satisfies the Massart noise condition with parameter $\tau_0> 0$ if
  $\P_{x \sim \cD_\cX} \paren*{\abs*{\eta(x) - 1 / 2} \leq \tau_0} = 0$.
\end{definition}
\begin{restatable}{proposition}{propProperAbstention}
\label{al_deep:prop:proper_abstention_2}
Suppose Massart noise holds.
By setting the abstention parameter $\gamma = \tau_0$ in \cref{al_deep:alg:abs} (and randomization over the abstention region), with probability at least $1-\delta$, we obtain a classifier with standard excess error $\wt O(\eps)$ after querying $\poly(\frac{1}{\tau_0}) \cdot \polylog(\frac{1}{\eps \, \delta})$ labels.
\end{restatable}
\begin{proof}
	This is a direct consequence of \cref{al_deep:thm:abs} and \cref{al_deep:prop:proper_abstention}.
\end{proof}

\subsubsection{Computational Efficiency}
\label{al_deep:app:computational}
    We discuss the efficient implementation of \cref{al_deep:alg:abs} and its computational complexity in the section. 
The computational efficiency of \cref{al_deep:alg:abs} mainly follows from the analysis in \citet{zhu2022efficient}. We provide the discussion here for completeness.

\paragraph{Regression orcale}
We introduce the regression oracle over the set of initialized neural networks $\cF_\dnn$ (line 1 at \cref{al_deep:alg:NCALP}).
Given any set $\cS$ of weighted examples $(w, x, y) \in \R_+ \times \cX \times \cY$ as input, the regression oracle outputs 
\begin{align*}
    \widehat f_\dnn \ldef \argmin_{f \in \cF_\dnn} \sum_{(w, x, y) \in \cS} w \paren*{f(x) - y}^2.
\end{align*}
While the exact computational complexity of such oracle with a set of neural networks remains elusive, in practice, running stochastic gradient descent often leads to great approximations.
We quantify the computational complexity in terms of the number of calls to the regression oracle. Any future analysis on such oracle can be incorporated into our guarantees.

We first state some known results in computing the confidence intervals with respect to a general set of regression functions $\cF$.

\begin{proposition}[\citet{krishnamurthy2017active, foster2018practical, foster2020instance}] \label{al_deep:prop:CI_oracle}
Consider the setting studied in \cref{al_deep:alg:abs}. 
Fix any epoch $m \in [M]$ and denote $\cB_m \ldef \crl{ (x_t,Q_t, y_t)}_{t=1}^{\tau_{m-1}}$.
Fix any $\iota > 0$.
For any data point $x \in \cX$, there exists algorithms $\AlgLcb$ and $\AlgUcb$ that certify
\begin{align*}
    & \lcb(x;\cF_m) - \iota \leq \AlgLcb(x;\cB_m,\beta_m,\iota) \leq \lcb(x;\cF_m) \quad \text{and}\\
    &\ucb(x;\cF_m) \leq \AlgUcb(x;\cB_m,\beta_m,\iota) \leq \ucb(x;\cF_m) + \iota.
\end{align*}
The algorithms take 
 $O(\frac{1}{\iota^2} \log \frac{1}{\iota})$ calls of the regression oracle for general $\cF$
 and take $O(\log \frac{1}{\iota})$ calls of the regression oracle if $\cF$ is convex and closed under pointwise convergence.
\end{proposition}
\begin{proof}
See Algorithm 2 in \citet{krishnamurthy2017active} for the general case; and Algorithm 3 in \citet{foster2018practical} for the case when $\cF$ is convex and closed under pointwise convergence.
\end{proof}

We now state the computational guarantee of \cref{al_deep:alg:abs}, given the regression oracle introduced above.
\begin{restatable}{theorem}{thmAbsEfficient}
	\label{al_deep:thm:abs_efficient}
	\cref{al_deep:alg:abs} can be efficiently implemented via the regression oracle and enjoys the same theoretical guarantees stated in \cref{al_deep:thm:abs}.
	The number of oracle calls needed is $\poly(\frac{1}{\gamma}) \cdot \frac{1}{\eps}$; the per-example inference time of the learned $\wh h_{M}$ is $\wt O ( \frac{1}{\gamma^2} \cdot \polylog \prn{\frac{1}{\eps \, \gamma}})$ for general $\cF$, and $\wt O ( \polylog \prn{\frac{1}{\eps \, \gamma}}) $ when $\cF$ is convex.
\end{restatable}

\begin{proof}
Fix any epoch $m \in [M]$.
Denote $\wb \iota \ldef \frac{\gamma}{8M}$ and $\iota_m \ldef \frac{(M-m) \gamma}{8M}$.
With any observed $x \in \cX$, we construct the approximated confidence intervals $\wh \lcb(x;\cF_m)$ and 
$\wh \ucb(x; \cF_m)$ as follows.
\begin{align*}
&	\wh \lcb(x;\cF_m) \ldef \AlgLcb(x;\cB_m,\beta_m,\wb \iota) - \iota_m \quad \text{and}\\	
    & \wh \ucb(x;\cF_m) \ldef\AlgUcb(x;\cB_m,\beta_m,\wb \iota)+ \iota_m. 
\end{align*}
For efficient implementation of \cref{al_deep:alg:abs}, we replace $\lcb(x;\cF_m)$ and $\ucb(x;\cF_m)$ with $\wh \lcb(x;\cF_m)$ and $\wh \ucb(x;\cF_m)$ in the construction of $\wh h_m$ and $g_m$.

Based on \cref{al_deep:prop:CI_oracle}, we know that 
\begin{align*}
    & \lcb(x;\cF_m) - \iota_m - \wb \iota \leq 	\wh \lcb(x;\cF_m) \leq \lcb(x;\cF_m) - \iota_m \quad \text{and}\\
    &\ucb(x;\cF_m) + \iota_m \leq \wh \ucb(x;\cF_m) \leq \ucb(x;\cF_m) + \iota_m + \wb \iota .
\end{align*}
Since $\iota_m + \wb \iota  \leq \frac{\gamma}{8}$ for any $m \in [M]$, the guarantee stated in \cref{al_deep:lm:query_implies_width} can be modified as $g_m(x)= 1 \implies w(x;\cF_m)\geq \frac{\gamma}{4}$. 
The guarantee stated in \cref{al_deep:lm:regret_no_query_mis} also holds true since we have $\wh \lcb(x;\cF_m) \leq \lcb(x;\cF_m)$ and $\wh \ucb(x;\cF_m) \geq \ucb(x;\cF_m)$ by construction. 
Suppose $\cF_{m} \subseteq \cF_{m-1}$ (as in \cref{al_deep:lm:set_f_mis}), we have 
\begin{align*}
	& \wh \lcb(x;\cF_m) \geq \lcb(x;\cF_m) -  \iota_m - \wb \iota \geq \lcb(x;\cF_{m-1}) - \iota_{m-1} \geq \wh\lcb (x;\cF_{m-1}) \quad \text{and} \\
	& \wh \ucb(x;\cF_m) \leq \ucb(x;\cF_m) +  \iota_m + \wb \iota \leq \ucb(x;\cF_{m-1}) + \iota_{m-1} \leq \wh\ucb (x;\cF_{m-1}),
\end{align*}
which ensures that $\ind(g_m(x) = 1) \leq \ind(g_{m-1}(x)=1)$. 
Thus, the inductive lemmas appearing in \cref{al_deep:app:supporting_abs} can be proved similarly with changes only in constant terms (also change the constant terms in the definition of $\wb \theta$ and in \cref{al_deep:eq:kappa_requirement}, since $\frac{\gamma}{2}$ is replaced by $\frac{\gamma}{4}$ in \cref{al_deep:lm:query_implies_width}).
As a result, the guarantees stated in \cref{al_deep:thm:abs_gen} (and \cref{al_deep:thm:abs}) hold true with changes only in constant terms.

We now discuss the computational complexity of the efficient implementation. 
At the beginning of each epoch $m$. We use one oracle call to compute $\widehat f_m \ldef \argmin_{f \in \cF} \sum_{t =1}^{ \tau_{m-1}} Q_t \paren{f(x_t) - y_t}^2 $. 
The main computational cost comes from computing $\wh \lcb$ and $\wh \ucb$ at each time step.
We take $\iota = \wb \iota \ldef \frac{\gamma}{8M}$ into \cref{al_deep:prop:CI_oracle}, which leads to 
$O \prn{ \frac{(\log T)^2}{\gamma^2}\cdot \log \prn{ \frac{\log T}{\gamma}}}$ calls of the regression oracle for general $\cF$ and 
$O \prn{ \log \prn{ \frac{\log T}{\gamma}}}$ calls of the regression oracle for any convex $\cF$ that is closed under pointwise convergence. This also serves as the per-example inference time for $\wh h_{M}$. 
The total computational cost of \cref{al_deep:alg:abs} is then derived by multiplying the per-round cost by $T$ and plugging $T = \frac{\theta \, \pseud(\cF)}{\eps \, \gamma} = \wt O \prn{\poly(\frac{1}{\gamma}) \cdot \frac{1}{\eps}}$ into the bound.
\end{proof}

\subsubsection{Proof of \cref{al_deep:thm:single_relu_lb}}
\label{al_deep:app:single_relu_lb}

For ease of construction, we suppose the instance space is $\cX = \B^{d}_1 \ldef \crl{x \in \R^{d}: \nrm{x}_2 \leq 1}$. 
Part of our construction is inspired by \citet{li2021eluder}.

\thmSingleReLULB*

\begin{proof}
	Fix any $\gamma \in (0,1 /8)$.
We first claim that we can find a discrete subset $\wb \cX \subseteq \cX$ with cardinality $\abs{\wb \cX} \geq (1/8\gamma)^{d/2}$ such that $\nrm{x_i}_2 =1$ and $\ang{x_1,x_2} \leq 1 - 4\gamma$ for any $x_i \in \wb \cX$. 
To prove this, we first notice that $\nrm{x_1 - x_2}_2 \geq \tau \iff \ang{x_1, x_2} \leq 1 - \tau^2/2$. Since the $\tau$-packing number on the unit sphere is at least $(1/\tau)^d$, setting $\tau = \sqrt{8 \gamma}$ leads to the desired claim.

We set $\cD_\cX \ldef \unif(\wb \cX)$ and $\cF_\dnn \ldef \crl{ \relu(\ang*{w, \cdot} - (1-4 \gamma))+ (1 /2 - 2\gamma): w \in \wb \cX}$. We have $\cF_\dnn \subseteq \cW^{1, \infty}_1(\cX)$ since $\nrm{w}_2 \leq$ for any $w \in \wb \cX$. We randomly select a $w^\star \in \cX$ and set $f^\star(\cdot) = \eta(\cdot) = \relu(\ang{w^\star,\cdot} - (1-4\gamma)) + (1 /2 - 2 \gamma)$. 
We assume that the labeling feedback is the conditional expectation, i.e., $\eta(x)$ is provided if $x$ is queried.
We see that $f^\star(x) = 1 /2 - 2 \gamma$ for any $x \in \cX$ but $x \neq w^\star$, and $f^\star(w^\star) = 1 /2 + 2 \gamma$. 
We can see that mistakenly select the wrong $\wh f \neq f^\star$ leads to $\frac{\gamma}{4} \cdot \frac{2}{\abs{\wb \cX}} = \frac{\gamma}{2 \abs{\wb \cX}}$ excess error. Note that the excess error holds true in both standard excess error and Chow's excess error (with parameter $\gamma$) since $\cD_\cX \prn{x \in \cX: \eta(x) \in [1 /2 - \gamma, 1 /2 + \gamma]} = 0$ by construction.

We suppose the desired access error $\eps$ is sufficiently small (e.g.,  $\eps \leq \frac{\gamma}{8 \abs{\wb \cX}}$).
We now show that, with label complexity at most $K \ldef \floor{{\abs{\wb \cX}}/{2}} = \Omega(\gamma^{- d /2})$, any active learning algorithm will, in expectation, pick a classifier that has $\Omega(\eps)$ excess error.
Since the worst case error of any randomized algorithm is lower bounded by the expected error of the best deterministic algorithm against a input distribution \citep{yao1977probabilistic}, we only need to analyze a deterministic learner.
We set the input distribution as the uniform distribution over instances with parameter $w^\star \in \wb \cX$.
For any deterministic algorithm, we use $s \ldef (x_{i_1}, \dots, x_{i_K})$ to denote the data points queried under the constraint that at most $K$ labels can be queried. We denote $\wh f \in \cF$ as the learned classifier conditioned on $s$.
Since $w^\star \sim \unif(\wb \cX)$, we know that, with probability at least $\frac{1}{2}$, $w^\star \notin s$. Conditioned on that event, we know that, with probability at least $\frac{1}{2}$, the learner will output $\wh f \neq f^\star$ since more than half of the data points remains unqueried. The deterministic algorithm thus outputs the wrong $\wh f \neq f^\star$ with probability at least $\frac{1}{2} \cdot \frac{1}{2} = \frac{1}{4}$, which has $\frac{\gamma}{2 \abs{\wb \cX}}$ excess error as previously discussed.
When $\eps \leq \frac{\gamma}{8 \abs{\wb \cX}}$, this leads to $\Omega(\eps)$ excess error in expectation.
\end{proof}

\subsection{Proofs and Supporting Results for \cref{al_deep:sec:extension}}
\label{al_deep:app:extension}
We provide mathematical backgrounds for the Radon $\BV^{2}$ space in \cref{al_deep:app:radon}, derive approximation results and passive learning results in \cref{al_deep:app:radon_passive}, and derive active learning results in \cref{al_deep:app:radon_active}.

\subsubsection{The Radon $\BV^2$ Space}
\label{al_deep:app:radon}
We provide explicit definition of the $\nrm{f}_{\RBV^2(\cX)}$ and associated mathematical backgrounds in this section. 
Also see \citet{ongie2020function, parhi2021banach, parhi2022kinds, parhi2022near, unser2022ridges} for more discussions.

We first introduce the \emph{Radon transform} of a function $f: \R^{d} \rightarrow \R$ as 
\begin{align*}
\mathscr{R} \crl{f} (\gamma ,t) \ldef \int_{\crl{x: \gamma^{\trn}x = t}} f(x)\, \mathsf{d} s(x), \quad (\gamma,t) \in \S^{d-1} \times \R,
\end{align*}
where $s$ denotes the surface measure on the hyperplane $\crl{x: \gamma^{\trn}x = t}$.
The Radon domain is parameterized by a \emph{direction} $\gamma \in \S^{d-1}$ and an \emph{offset} $t \in \R$.
We also introduce the \emph{ramp filter} as 
\begin{align*}
	\Lambda^{d-1} \ldef \prn{- \partial^{2}_t}^{\frac{d-1}{2}},
\end{align*}
where $\partial_t$ denotes the partial derivative with respect to the offset variable,  $t$, of the Radon domain, and the fractional powers are defined in terms of Riesz potentials.

With the above preparations, we can define the  $\RTV^2$-seminorm as 
\begin{align*}
\RTV^2(f) \ldef c_d \nrm{\partial^2_t \Lambda^{d-1} \mathscr{R}f}_{\cM(\S^{d-1} \times \R)},	
\end{align*}
where $c_d = 1 / (2 (2 \pi)^{d-1})$ is a dimension-dependent constant, and $\nrm{\cdot}_{\cM(\S^{d-1} \times \R)}$ denotes the \emph{total variation norm} (in terms of measures) over the bounded domain  $\S^{d-1} \times \R$.
The $\RBV^2$ norm of $f$ over  $\R^{d}$ is defined as 
\begin{align*}
\nrm{f}_{\RBV^2(\R^{d})} \ldef \RTV^2(f) + \abs{f(0)} + \sum_{k=1}^{d} \abs{f(e_k) - f(0)},	
\end{align*}
where $\crl{e_k}_{k=1}^{d}$ denotes the canonical basis of $\R ^{d}$.
The $\RBV^2(\R^{d})$ space is then defined as 
\begin{align*}
	\RBV^2(\R^{d}) \ldef \crl{f \in L^{\infty,1}(\R^{d}): \RBV^2(f) < \infty},
\end{align*}
where $L^{\infty,1}(\R^{d})$ is the Banach space of functions mapping $\R^{d} \rightarrow \R$ of at most linear growth.
To define the $\RBV^2$ norm of $f$ over a bounded domain  $\cX \subseteq \R ^{d}$, we use the standard approach of considering restrictions of functions in $\RBV^2(\R^{d})$, i.e.,
\begin{align*}
	\nrm{f}_{\RBV^2(\cX)} \ldef \inf_{g \in \RBV^2(\R^{d})} \nrm{g}_{\RBV^2(\R^{d})} \quad \text{ s.t. }
	\quad g \vert_{\cX} = f.
\end{align*}

In the rest of \cref{al_deep:app:extension}, we use $\cP(\beta)$ to denote the set of distributions that satisfy (1) Tsybakov noise condition with parameter  $\beta \geq 0$; and (2)  $\eta \in \RBV^2_1(\cX)$.
\subsubsection{Approximation and Passive Learning Results}
\label{al_deep:app:radon_passive}

\begin{proposition}
	\label{al_deep:prop:vc_approx_RBV}
Suppose $\cD_{\cX \cY} \in \cP(\beta)$. 	
One can construct a set of neural network classifier $\cH_{\dnn}$ such that the following two properties hold simultaneously: 
\begin{align*}
	\min_{h \in \cH_{\dnn} }\err(h) - \err(h^{\star}) = O \prn{\eps} \quad \text{ and }
\quad 	\vcd( \cH_{\dnn}) = \wt O \prn{ \eps^{- \frac{2d}{(1+\beta)(d+3)}}}.
\end{align*}
\end{proposition}
\begin{proof}
We take $\kappa = \eps^{\frac{1}{1+\beta}}$ in \cref{al_deep:thm:approx_RBV2} to construct a set of neural network classifiers $\cH_\dnn$ with  $W = O \prn{ \eps^{- \frac{2d}{(1+\beta)(d+3)}} }$ total parameters arranged in $L = O \prn{ 1}$ layers.	
According to \cref{al_deep:thm:vcd_nn}, we know 
\begin{align*}
\vcd( \cH_{\dnn}) = O \prn{\eps^{- \frac{2d}{(1+\beta)(d+3)}} \cdot \log \prn{\eps^{-1}}} = \wt O \prn{ \eps^{- \frac{2d}{(1+\beta)(d+3)}}}.
\end{align*}
We now show that there exists a classifier $\wb h \in \cH_{\dnn}$ with small excess error.
Let $\wb h = h_{\wb f}$ be the classifier such that  $\nrm{ \wb f - \eta}_{\infty} \leq \kappa$. We can see that 
 \begin{align*}
	\exc(\wb h)
	& = \E \brk*{ \ind( \wb h(x) \neq y) - \ind( h^{\star}(x) \neq y)}\\
	& = \E \brk*{ \abs{ 2 \eta(x) - 1} \cdot \ind( \wb h(x) \neq h^{\star}(x))}\\
	& \leq 2 \kappa \cdot \P_{x \sim \cD_{\cX}} \prn*{ x \in \cX: \abs{\eta(x) - {1}/{2}} \leq \kappa}\\
	& = O \prn{ \kappa^{1 + \beta}}\\
	& = O(\eps),
\end{align*}
where the third line follows from the fact that $\wb h$ and  $h^{\star}$ disagrees only within region $\crl{x \in \cX: \abs{\eta(x) - 1 /2} \leq \kappa}$ and the incurred error is at most  $2 \kappa$ on each disagreed data point.
The fourth line follows from the Tsybakov noise condition and the last line follows from the selection of $\kappa$.
\end{proof}

\begin{theorem}
	\label{al_deep:thm:passive_noise_RBV}
	Suppose $\cD_{\cX \cY} \in \cP(\beta)$.
	Fix any $\eps, \delta > 0$.
	Let $\cH_{\dnn}$ be the set of neural network classifiers constructed in \cref{al_deep:prop:vc_approx_RBV}.
	With $n =\wt O \prn{\eps^{-\frac{4d+6+ \beta(d+3)}{(1+\beta)(d+3)}}} $ i.i.d. sampled data points, with probability at least $1-\delta$,
	the empirical risk minimizer $\wh h \in \cH_{\dnn}$ achieves excess error $ O(\eps)$.
\end{theorem}

\begin{proof}
\cref{al_deep:prop:vc_approx_RBV} certifies $\min_{h \in \cH_\dnn} \err(h) - \err(h^{\star}) = O(\eps)$
and
\linebreak
  $\vcd(\cH_\dnn) = O \prn*{\eps^{-\frac{2d}{(1+\beta)(d+3)}} \cdot \log(\eps^{-1})}$.
Take $\rho = 1$ in \cref{al_deep:thm:erm_tsy}, leads to 
 \begin{align*}
& \err(\wh h) - \err(h^{\star}) \leq   O \prn*{ \eps + \prn*{ \eps^{-\frac{2d}{(1+\beta)(d+3)}} \cdot \log(\eps^{-1}) \cdot \frac{\log n}{n}}^{\frac{1+\beta}{2+\beta}} + \frac{\log\delta^{-1}}{n}},
\end{align*}
Taking $n = O \prn{\eps^{-\frac{4d+6+ \beta(d+3)}{(1+\beta)(d+3)}}\cdot \log (\eps^{-1}) + \eps^{-1} \cdot \log(\delta^{-1})} = \wt O \prn{\eps^{-\frac{4d+6+ \beta(d+3)}{(1+\beta)(d+3)}}}$ thus ensures that $\err(\wh h) - \err(h^{\star}) = O(\eps)$.
\end{proof}

\subsubsection{Active Learning Results}
\label{al_deep:app:radon_active}

\thmActiveNoiseRBV*
\begin{proof}
	Construct $\cH_\dnn$ based on \cref{al_deep:prop:vc_approx_RBV} such that
\linebreak
  $\min_{h \in \cH_\dnn} \err(h) - \err(h^{\star}) = O(\eps)$ and $\vcd(\cH_\dnn) = \wt O \prn{ \eps^{-\frac{2d}{(1+\beta)(d+3)}} }$. Taking such $\cH_\dnn$ as the initialization of \cref{al_deep:alg:RCAL} (line 1) and applying \cref{al_deep:thm:RCAL_gen} leads to the desired result.
\end{proof}

To derive deep active learning guarantee with abstention in the Radon $\BV^2$ space, 
we first present two supporting results below.
\begin{proposition}
	\label{al_deep:prop:pd_approx_RBV}
	Suppose $\cD_{\cX \cY} \in \cP(\beta)$. 
	One can construct a set of neural network regression functions  $\cF_\dnn$ such that the following two properties hold simultaneously:
	 \begin{align*}
		\exists f \in \cF_\dnn \text{ s.t. } \nrm{ f - f^{\star}}_\infty \leq \kappa, \quad \text{ and } \quad \pseud(\cF_\dnn) \leq c \cdot {\kappa^{-\frac{2d}{d+3}} \log^2 (\kappa^{-1})},
	\end{align*}
	where $c > 0$ is a universal constant.
\end{proposition}
\begin{proof}
	The result follows by combining \cref{al_deep:thm:approx_RBV2} and \cref{al_deep:thm:pdim_nn}.
\end{proof}

\begin{proposition}
	\label{al_deep:prop:clip_filter_radon}
	Suppose $\eta$ is  $L$-Lipschitz and  $\cX \subseteq \B^{d}_r$.
	Fix any $\kappa \in (0, \gamma / 32]$. There exists a set of neural network regression functions $\cF_\dnn$ such that the followings hold simultaneously.
	\begin{enumerate}
		\item 	$\pseud(\cF_\dnn) \leq c \cdot {\kappa^{-\frac{2d}{d+3}} \log^2(\kappa^{-1})}$ with a universal constant $c > 0$.
		\item There exists a $\wb f \in \cF_\dnn$ such that $\nrm{\wb f - \eta}_\infty \leq \kappa$.
		\item $ \theta^{\val}_{\cF_\dnn}(\gamma / 4) \ldef \sup_{f \in \cF_\dnn, \iota > 0} \theta^{\val}_f (\cF_\dnn, \gamma /4, \iota) \leq c^{\prime} \cdot \prn{\frac{Lr}{\gamma}}^{d}$ with a universal constant $c^{\prime} > 0$.
	\end{enumerate}
\end{proposition}
\begin{proof}
	The implementation and proof are similar to those in \cref{al_deep:prop:clip_filter}, except we use \cref{al_deep:prop:pd_approx_RBV} instead of \cref{al_deep:prop:pd_approx}.
\end{proof}

We now state and prove deep active learning guarantees in the Radon $\BV^{2}$ space.
\begin{restatable}{theorem}{thmAbsRBV}
	\label{al_deep:thm:abs_RBV}
	Suppose $\eta \in \RBV^2_1(\cX)$.
	Fix any $\eps, \delta, \gamma >0$.
	There exists an algorithm such that,
	with probability at least $1-\delta$, it learns a classifier  $\wh h$ with Chow's excess error $\wt O(\eps)$ after querying $\poly(\frac{1}{\gamma}) \cdot \polylog(\frac{1}{\eps \, \delta})$ labels.
\end{restatable}
\begin{proof}
The result is obtained by applying \cref{al_deep:alg:abs} with line 1 be the set of neural networks $\cF_\dnn$ generated from \cref{al_deep:prop:clip_filter_radon} with approximation level $\kappa \in (0, \gamma / 32]$ (and constants $c, c^{\prime}$ specified therein).
The rest of the proof proceeds in a similar way as the proof \cref{al_deep:thm:abs}.
Since we have $r=1$ and  $L\leq 1$ \citep{parhi2022near},
we only need to choose a $\kappa > 0$ such that 
 \begin{align*}
	\frac{1}{\kappa} = {\check c \cdot \prn*{\frac{1}{\gamma}}^{\frac{d}{2} + 1} \cdot \log \frac{1}{\eps \, \gamma}}
\end{align*}
with a universal constant $\check c > 0$.
With such choice of $\kappa$, we have  
\begin{align*}
\pseud(\cF_\dnn) = O\prn*{\prn*{\frac{1}{\gamma}}^{\frac{d^2+2d}{d+3}} \polylog \prn*{\frac{1}{\eps \, \gamma}}}.
\end{align*}
Plugging this bound on $\pseud(\cF_\dnn)$ and the upper bound on $\theta^{\val}_{\cF_\dnn}(\gamma / 4)$ from \cref{al_deep:prop:clip_filter_radon} into the guarantee of \cref{al_deep:thm:abs_gen} leads to $\exc_\gamma(\wh h) = O \prn{\eps \cdot \log \prn{\frac{1}{\eps \, \gamma \, \delta}}}$ after querying 
\begin{align*}
	O \prn*{\prn*{\frac{1}{\gamma}}^{d + 2 + \frac{d^2 + 2d}{d+3}} \cdot \polylog \prn*{\frac{1}{\eps \, \gamma \, \delta}} }
\end{align*}
labels.
\end{proof}

\part{Sequential Decision Making with Large Action Spaces}
\label{part:large}

\chapter{Contextual Bandits with Large Action Spaces: Made Practical}
\label{chapter:large:linear}
A central problem in sequential decision making is to develop algorithms that are practical and computationally efficient, yet support the use of flexible, general-purpose models. Focusing on the contextual bandit problem, recent progress provides provably efficient algorithms with strong empirical performance when the number of possible alternatives (``actions'') is small, but guarantees for decision making in large, continuous action spaces have remained elusive, leading to a significant gap between theory and practice. We present the first efficient, general-purpose algorithm for contextual bandits with continuous, linearly structured action spaces. Our algorithm makes use of computational oracles for (i) supervised learning, and (ii) optimization over the action space, and achieves sample complexity, runtime, and memory independent of the size of the action space. In addition, it is simple and practical. We perform a large-scale empirical evaluation, and show that our approach typically enjoys superior performance and efficiency compared to standard baselines. 

\section{Introduction}
\label{dm_linear:sec:intro}

We consider the design of practical, theoretically motivated algorithms for sequential decision making with contextual information, better known as the \emph{contextual bandit problem}. Here, a learning agent repeatedly receives a \emph{context} (e.g., a user's profile), selects an \emph{action} (e.g., a news article to display), and receives a \emph{reward} (e.g., whether the article was clicked). Contextual bandits are a useful model for decision making in unknown environments in which both exploration and generalization are required, but pose significant algorithm design challenges beyond classical supervised learning. Recent years have seen development on two fronts:
On the theoretical side, extensive research into finite-action contextual bandits has resulted in practical, provably efficient algorithms capable of supporting flexible, general-purpose models \citep{langford2007epoch,agarwal2014taming,foster2020beyond,simchi2021bypassing,foster2021efficient}.
Empirically, contextual bandits have been widely deployed in practice for online personalization and recommendation tasks \citep{li2010contextual,agarwal2016making,tewari2017ads,cai2021bandit}, leveraging the availability of high-quality action slates (e.g., subsets of candidate articles selected by an editor).

The developments above critically rely on the existence of a small number of possible decisions or alternatives. However, many applications demand the ability to make contextual decisions in large, potentially continuous spaces, where actions might correspond to images in a database or high-dimensional embeddings of rich documents such as webpages. Contextual bandits in large (e.g., million-action) settings remains a major challenge---both statistically and computationally---and constitutes a substantial gap between theory and practice. In particular:

\begin{itemize}
\item Existing \emph{general-purpose} algorithms \citep{langford2007epoch,agarwal2014taming,foster2020beyond,simchi2021bypassing,foster2021efficient} allow for the use of flexible models (e.g., neural networks, forests, or kernels) to facilitate generalization across contexts, but have sample complexity and computational requirements linear in the number of actions. These approaches can degrade in performance under benign operations such as duplicating actions.

\item While certain recent approaches extend the general-purpose methods above to accommodate large action spaces, they either require sample complexity exponential in action dimension~\citep{krishnamurthy2020contextual}, or require additional distributional assumptions \citep{sen2021top}.
\item Various results efficiently handle large or continuous action spaces \citep{dani2008stochastic, jun2017scalable, yang2021linear} with specific types of function approximation, but do not accommodate general-purpose models.
\end{itemize}
As a result of these algorithmic limitations, empirical aspects of contextual decision making in large action spaces have remained relatively unexplored compared to the small-action regime \citep{bietti2021contextual}, with little in the way of readily deployable out-of-the-box solutions.

\paragraph{Contributions}
We provide the first efficient algorithms for contextual bandits with continuous, linearly structured action spaces and general function approximation. Following \cite{chernozhukov2019semi,xu2020upper,foster2020adapting}, we adopt a modeling approach, and assume rewards for each context-action pair $(x,a)$ are structured as
\begin{equation}
  \label{dm_linear:eq:linear}
  \fstar(x,a)=\tri*{\phi(x,a),\gstar(x)}.
\end{equation}
Here $\phi(x,a)\in\bbR^{d}$ is a known context-action embedding (or feature map) and $\gstar\in\cG$ is a context embedding to be learned online, which belongs to an arbitrary, user-specified function class $\cG$. Our algorithm, \mainalg, is computationally efficient (in particular, the runtime and memory are \emph{independent} of the number of actions) whenever the user has access to (i) an \emph{online regression oracle} for supervised learning over the reward function class, and (ii) an \emph{action optimization oracle} capable of solving problems of the form
\begin{align*}
\argmax_{a\in\cA}\tri*{\phi(x,a),\theta}
\end{align*}
for any $\theta\in\bbR^{d}$. The former oracle follows prior approaches to finite-action contextual bandits \citep{foster2020beyond,simchi2021bypassing,foster2021efficient}, while the latter generalizes efficient approaches to (non-contextual) linear bandits \citep{mcmahan2004online,dani2008stochastic,bubeck2012towards,hazan2016volumetric}. We provide a regret bound for \mainalg which scales as $\sqrt{\poly(d)\cdot T}$, and---like the computational complexity---is independent of the number of actions. Beyond these results, we provide a particularly practical variant of \mainalg (\greedyalg), which enjoys even faster runtime at the cost of slightly worse ($\poly(d)\cdot{}T^{2/3}$-type) regret.

\paragraph{Our techniques}
On the technical side, we show how to
\emph{efficiently} combine the inverse gap weighting technique \citep{abe1999associative,foster2020beyond} previously used in the finite-action setting with optimal design-based approaches for exploration with linearly structured actions. This offers a computational improvement upon the results of \cite{xu2020upper,foster2020adapting}, which provide algorithms with $\sqrt{\poly(d)\cdot{}T}$-regret for the setting we consider, but require enumeration over the action space. Conceptually, our results expand upon the class of problems for which minimax approaches to exploration \citep{foster2021statistical} can be made efficient.

\paragraph{Empirical performance}
As with previous approaches based on regression oracles, \mainalg is simple, practical, and well-suited to flexible, general-purpose function approximation. In extensive experiments ranging from thousands to millions of actions, we find that our methods typically enjoy superior performance compared to existing baselines. In addition, our experiments validate the statistical model in \cref{dm_linear:eq:linear} 
which we find to be well-suited to learning with large-scale language models \citep{devlin2019bert}.

\subsection{Organization}
This chapter is organized as follows. In \cref{dm_linear:sec:setting}, we formally introduce our statistical model and the computational oracles upon which our algorithms are built; we also discuss additional related work in \cref{dm_linear:sec:related}. Subsequent sections are dedicated to our main results.
\begin{itemize}
\item As a warm-up, \cref{dm_linear:sec:warmup} presents a simplified algorithm, \greedyalg, which illustrates the principle of exploration over an approximate optimal design. This algorithm is practical and oracle-efficient, but has suboptimal $\poly(d)\cdot{}T^{2/3}$-type regret.
\item 
  Building on these ideas, \cref{dm_linear:sec:minimax} presents our main algorithm, \mainalg, which combines the idea of approximate optimal design used by \greedyalg with the inverse gap weighting method \citep{abe1999associative,foster2020beyond}, resulting in an oracle-efficient algorithm with $\sqrt{\poly(d)\cdot{}T}$-regret.
\end{itemize}
\cref{dm_linear:sec:experiments} presents empirical results for both
algorithms. We close with discussion of 
future directions in \cref{dm_linear:sec:discussion}.
All proofs are deferred to \cref{dm_linear:sec:proofs}.

\section{Problem Setting}

\label{dm_linear:sec:setting}

The contextual bandit problem proceeds over $T$ rounds. At each round $t\in\brk{T}$, the learner receives a context $x_t \in \cX$ (the \emph{context space}), selects an action $a_t\in\cA$ (the \emph{action space}), and then observes a reward $r_t(a_t)$, where $r_t:\cA\to\brk{-1,1}$ is the underlying reward function.
We assume that for each round $t$, conditioned on $x_t$, the reward $r_t$ is sampled from a (unknown) distribution $\bbP_{r_t}(\cdot\mid{}x_t)$. We allow both the contexts $x_{1},\ldots,x_T$ and the distributions $\bbP_{r_1},\ldots,\bbP_{r_T}$ to be selected in an arbitrary, potentially adaptive fashion based on the history.

\paragraph{Function approximation}

Following a standard approach to developing efficient contextual bandit methods, we take a modeling approach, and work with a user-specified class of regression functions $\cF \subseteq (\cX \times \cA \rightarrow [-1,1])$ that aims to model the underlying mean reward function. We make the following realizability assumption \citep{agarwal2012contextual, foster2018practical, foster2020beyond, simchi2021bypassing}.
\begin{assumption}[Realizability]
  \label{dm_linear:asm:realizability}
  There exists a regression function $f^\star \in \cF$ such that $ \E \sq{r_t(a) \mid x_t=x} = f^\star(x, a)$ for all $a\in\cA$ and $t\in\brk{T}$.
\end{assumption}
Without further assumptions, there exist function classes $\cF$ for which the regret of any algorithm must grow proportionally to $\abs{\cA}$ (e.g., \cite{agarwal2012contextual}). In order to facilitate generalization across actions and achieve sample complexity and computational complexity independent of $\abs{\cA}$, we assume that each function $f\in\cF$ is linear in a known (context-dependent) feature embedding of the action. Following \citet{xu2020upper,foster2020adapting}, we assume that $\cF$ takes the form
\begin{align*}
  \cF = \crl*{f_g \prn{x,a} = \ang{\phi \prn{x,a}, g \prn{x}}: g \in \cG  },
\end{align*}
where $\phi(x,a)\in\bbR^{d}$ is a known, context-dependent action embedding and $\cG$ is a user-specified class of context embedding functions.

This formulation assumes linearity in the action space (after featurization), but allows for nonlinear, \emph{learned} dependence on the context $x$ through the function class $\cG$, which can be taken to consist of neural networks, forests, or any other flexible function class a user chooses. For example, in news article recommendation, $\phi(x,a)=\phi(a)$ might correspond to an embedding of an article $a$ obtained using a large pre-trained language-model, while $g(x)$ might correspond to a task-dependent embedding of a user $x$, which our methods can learn online. Well-studied special cases include the linear contextual bandit setting \citep{chu2011contextual,abbasi2011improved}, which corresponds to the special case where each $g\in\cG$ has the form $g \prn{x} = \theta$ for some fixed $\theta \in \R^d$, as well as the standard finite-action contextual bandit setting, where $d=\abs{\cA}$ and $\phi(x,a)=e_a$.

We let $\gstar\in\cG$ denote the embedding for which $\fstar=f_{\gstar}$. We assume that $\sup_{x \in \cX,a \in \cA}\nrm{\phi(x,a)} \leq 1$ and $\sup_{g \in \cG, x \in \cX} \nrm{g(x)} \leq 1$. In addition, we assume that $\spn(\crl{\phi(x,a)}) = \R^d$ for all $x \in \cX$.

\paragraph{Regret}
For each regression function $f\in\cF$, let $\pi_f(x_t) \ldef \argmax_{a \in \cA} f(x_t,a)$ denote the induced policy, and define $\pi^\star \ldef \pi_{f^\star}$ as the optimal policy. We measure the performance of the learner in terms of regret: %
\begin{align*}
    \regcb(T) \ldef  \sum_{t=1}^T  r_t(\pi^\star(x_t)) - r_t(a_t).
\end{align*}

\subsection{Computational Oracles}

To derive efficient algorithms with sublinear runtime, we make use of two computational oracles: First, following \cite{foster2020beyond,simchi2021bypassing, foster2020adapting, foster2021instance}, we use an \emph{online regression oracle} for supervised learning over the reward function class $\cF$. Second, we use an \emph{action optimization oracle}, which facilitates linear optimization over the action space $\cA$ \citep{mcmahan2004online,dani2008stochastic,bubeck2012towards,hazan2016volumetric}.

\paragraph{Function approximation: Regression oracles}

A fruitful approach to designing efficient contextual bandit algorithms is through reduction to supervised regression with the class $\cF$, which facilitates the use of off-the-shelf supervised learning algorithms and models \citep{foster2020beyond, simchi2021bypassing, foster2020adapting, foster2021instance}. Following \citet{foster2020beyond}, we assume access to an \emph{online regression oracle} \sqalgtext, which is an algorithm for online learning (or, sequential prediction) with the square loss.

We consider the following protocol. At each round $t \in [T]$, the oracle produces an estimator $\fhat_t=f_{\ghat_t}$,
then receives a context-action-reward tuple $(x_t, a_t,
r_t(a_t))$. The goal of the oracle is to accurately predict the reward
as a function of the context and action, and we evaluate its
prediction error via the square loss $(\fhat_t(x_t,a_t)-r_t)^2$. We measure the oracle's cumulative performance through square-loss regret to $\cF$.%
\begin{assumption}[Bounded square-loss regret]
\label{dm_linear:asm:regression_oracle}
The regression oracle \sqalgtext guarantees that for any (potentially adaptively chosen) sequence $\curly*{(x_t, a_t, r_t(a_t))}_{t=1}^T$, 
\begin{align*}
  \sum_{t=1}^T \paren*{\widehat f_t(x_t, a_t) - r_t(a_t)}^2
                   -\inf_{f \in \cF}\sum_{t=1}^T \paren*{f(x_t, a_t) - r_t(a_t)}^2 
  \leq \regsq(T),\notag
\end{align*}
for some (non-data-dependent) function $\regsq(T)$.
\end{assumption}
We let $\Tsq$ denote an upper bound on the time required to (i) query the oracle's estimator $\ghat_t$ with $x_t$ and receive the {vector} $\ghat_t(x_t)\in\bbR^{d}$, and (ii) update the oracle with the example $(x_t, a_t, r_t(a_t))$.
{We let $\Msq$ denote the maximum memory used by the oracle throughout its execution.}

Online regression is a well-studied problem, with computationally efficient algorithms for many models. Basic examples include finite classes $\cF$, where one can attain $\regsq(T) = O(\log \abs*{\cF})$ \citep{vovk1998game}, and linear models ($g(x)= \theta$), where the online Newton step algorithm \citep{hazan2007logarithmic} satisfies \cref{dm_linear:asm:regression_oracle} with $\regsq(T) = O(d \log T)$. More generally, even for classes such as deep neural networks for which provable guarantees may not be available, regression is well-suited to gradient-based methods. We refer to \citet{foster2020beyond, foster2020adapting} for more comprehensive discussion.

\paragraph{Large action spaces: Action optimization oracles}
\label{dm_linear:sec:linear_opt_oracle}
The regression oracle setup in the prequel is identical to that considered in the finite-action setting \citep{foster2020beyond}. In order to develop efficient algorithms for large or infinite action spaces, we assume access to an oracle for linear optimization over actions.
\begin{definition}[Action optimization oracle]
  \label{dm_linear:def:action_oracle}
  An {action optimization oracle} \optalgtext takes as input a context $x \in \cX$, and vector $\theta \in \R^d$ and returns
\begin{align}
    a^\star \ldef \argmax_{a \in \cA} \ang*{\phi(x,a), \theta}. \label{dm_linear:eq:linear_opt_oracle}
\end{align}
\end{definition}
For a single query to the oracle, 
We let $\Topt$ denote a bound on the runtime for a single query to the oracle.
We {let  $\Mopt$ denote the maximum memory used by the oracle throughout its execution.}

The action optimization oracle in \cref{dm_linear:eq:linear_opt_oracle} is widely used throughout the literature on linear bandits \citep{dani2008stochastic, chen2017nearly, cao2019disagreement,katz2020empirical}, and can be implemented in polynomial time for standard combinatorial action spaces. It is a basic computational primitive in the theory of convex optimization, and when $\cA$ is convex, it is equivalent (up to polynomial-time reductions) to other standard primitives such as separation oracles and membership oracles \citep{schrijver1998theory, grotschel2012geometric}. It also equivalent to the well-known Maximum Inner Product Search (MIPS) problem \citep{shrivastava2014asymmetric}, for which sublinear-time hashing based methods are available.

\begin{example}
  \label{dm_linear:ex:combinatorial}
  Let $G=(V,E)$ be a graph, and let $\phi(x,a) \in \crl{0,1}^{\abs{E}}$ represent a matching and $\theta \in \R^{\abs{E}}$ be a vector of edge weights. The problem of finding the maximum-weight matching for a given set of edge weights can be written as a linear optimization problem of the form in \cref{dm_linear:eq:linear_opt_oracle}, and Edmonds' algorithm \citep{edmonds1965paths} can be used to find the maximum-weight matching in $O(\abs{V}^2 \cdot \abs{E})$ time.  
\end{example}
Other combinatorial problems that admit polynomial-time action optimization oracles include the maximum-weight spanning tree problem, the assignment problem, and others \citep{awerbuch2008online, cesa2012combinatorial}.

\paragraph{Action representation}
We define $b_\cA$ as the number of bits used to represent actions in $\cA$, which is always upper bounded by $O(\log \abs{\cA})$ for finite action sets, and by $\wt{O}(d)$ {for actions that can be represented as vectors in $\R^d$}. Tighter bounds are possible with additional structual assumptions. Since representing actions is a minimal assumption, we hide the dependence on $b_\cA$ in big-$\bigoh$ notation for our runtime and memory analysis.

\subsection{Additional Related Work}
\label{dm_linear:sec:related}

In this section we highlight some relevant lines of research not already discussed.

\paragraph{Efficient general-purpose contextual bandit algorithms}
There is a long line of research on computationally efficient methods for contextual bandits with general function approximation, typically based on reduction to either cost-sensitive classification oracles \citep{langford2007epoch, dudik2011efficient, agarwal2014taming} or regression oracles \citep{foster2018practical, foster2020beyond,simchi2021bypassing}. Most of these works deal with a finite action spaces and have regret scaling with the number of actions, which is necessary without further structural assumptions \citep{agarwal2012contextual}. An exception is the works of \citet{foster2020adapting} and \citet{xu2020upper}, both of which consider the same setting as this chapter. Both of the algorithms in these works require solving subproblems based on maximizing quadratic forms (which is NP-hard in general \citep{sahni1974computationally}), and cannot directly take advantage of the linear optimization oracle we consider.
Also related is the work of \citet{zhang2021feel}, which proposes a posterior sampling-style algorithm for the setting we consider. This algorithm is not fully comparable computationally, as it requires sampling from specific posterior distribution; it is unclear whether this can be achieved in a provably efficient fashion.

\paragraph{Linear contextual bandits} The linear contextual bandit problem is a special case of our setting in which $\gstar(x)=\theta\in\bbR^{d}$ is constant (that is, the reward function only depends on the context through the feature map $\phi$). The most well-studied families of algorithms for this setting are UCB-style algorithms and posterior sampling. With a well-chosen prior and posterior distribution, posterior sampling can be implemented efficiently \citep{agrawal2013thompson}, but it is unclear how to efficiently adapt this approach to accomodate general function approximation. Existing UCB-type algorithms require solving sub-problems based on maximizing quadratic forms, which is NP-hard in general \citep{sahni1974computationally}. One line of research aims to make UCB efficient by using hashing-based methods (MIPS) to approximate the maximum inner product \citep{yang2021linear, jun2017scalable}. These methods have runtime sublinear (but still polynomial) in the number of actions.

\paragraph{Non-contextual linear bandits} For the problem of \emph{non-contextual} linear bandits (with either stochastic or adversarial rewards), there is a long line of research on efficient algorithms that can take advantage of linear optimization oracles \citep{awerbuch2008online, mcmahan2004online, dani2006robbing, dani2008stochastic,bubeck2012towards, hazan2016volumetric, ito2019oracle}; see also work on the closely related problem of combinatorial pure exploration \citep{chen2017nearly, cao2019disagreement,katz2020empirical,wagenmaker2021experimental}.
In general, it is not clear how to lift these techniques to contextual bandits with linearly-structured actions and general function approximation.
{We also mention that optimal design has been applied in the context of linear bandits, but these algorithms are restricted to the non-contextual setting \citep{lattimore2020bandit, lattimore2020learning}, or to pure exploration \citep{soare2014best, fiez2019sequential}.
The only exception we are aware of is \citet{ruan2021linear}, who
extend these developments to linear contextual bandits (i.e., where
$\gstar(x) = \theta$), but critically use that contexts are stochastic. 
}

\paragraph{Other approaches} Another line of research provides efficient contextual bandit methods under specific modeling assumptions on the context space or action space that differ from the ones we consider here. \citet{zhou2020neural, xu2020neural, zhang2021neural, kassraie2022neural} provide generalizations of the UCB algorithm and posterior sampling based on the Neural Tangent Kernel (NTK). These algorithms can be used to learn context embeddings (i.e., $g(x)$) with general function approximation, but only lead to theoretical guarantees under strong RKHS-based assumptions. For large action spaces, these algorithms typically require enumeration over actions. \citet{majzoubi2020efficient} consider a setting with nonparametric action spaces and design an efficient tree-based learner; their guarantees, however, scale exponentially in the dimensionality of action space. 
\citet{sen2021top} provide heuristically-motivated but empirically-effective tree-based algorithms for contextual bandits with large action spaces, with theoretical guarantees when the actions satisfy certain tree-structured properties. Lastly, another empirically-successful approach is the policy gradient method (e.g., \citet{williams1992simple, bhatnagar2009natural, pan2019policy}). On the theoretical side, policy gradient methods do not address the issue of systematic exploration, and---to our knowledge---do not lead to provable guarantees for the setting considered in this chapter.

\section{Warm-Up: Efficient Algorithms via Uniform Exploration}
\label{dm_linear:sec:warmup}

In this section, we present our first result: an efficient algorithm based on uniform exploration over a representative basis (\spannerGreedy; \cref{dm_linear:alg:greedy}). This algorithm achieves computational efficiency by taking advantage of an online regression oracle, but its regret bound has sub-optimal dependence on $T$.
Beyond being practically useful in its own right, this result serves as a warm-up for \cref{dm_linear:sec:minimax}.

Our algorithm is based on exploration with a \emph{G-optimal design} for the embedding $\phi$, which is a distribution over actions that minimizes a certain notion of worse-case variance \citep{kiefer1960equivalence,atwood1969optimal}.
\begin{definition}[G-optimal design]
    \label{dm_linear:def:g_opt_design}
   Let a set $\cZ \subseteq \R^d$ be given. A distribution $q \in \Delta(\cZ)$ is said to be a G-optimal design with approximation factor $C_{\opt} \geq 1$ if 
   \begin{align*}
       \sup_{z \in \cZ} \nrm{z}_{V(q)^{-1}}^2 \leq C_{\opt} \cdot d,
   \end{align*} 
   where $V(q) \ldef \E_{z \sim q} \brk[\big]{z z^\top}$.
\end{definition}
The following classical result guarantees existence of a G-optimal design.
\begin{lemma}[\citet{kiefer1960equivalence}]
    \label{dm_linear:prop:kiefer_wolfowitz}
    For any compact set $\cZ \subseteq \R^d$, there exists an optimal design with $C_{\opt} = 1$. 
\end{lemma}

\begin{algorithm}[H]
	\caption{\spannerGreedy}
	\label{dm_linear:alg:greedy} 
	\renewcommand{\algorithmicrequire}{\textbf{Input:}}
	\renewcommand{\algorithmicensure}{\textbf{Output:}}
	\newcommand{\algorithmicbreak}{\textbf{break}}
    \newcommand{\BREAK}{\STATE \algorithmicbreak}
	\begin{algorithmic}[1]
		\REQUIRE Exploration parameter $\epsilon \in (0, 1]$, online regression oracle \sqalgtext, action optimization oracle \optalgtext.
		\FOR{$t = 1, 2, \dots, T$}
		\STATE Observe context $x_t$.
		\STATE Receive $\widehat f_t = f_{\widehat g_t}$ from regression oracle \sqalgtext.
\STATE Get $\widehat a_t \gets \argmax_{a \in \cA} \ang*{\phi(x_t, a), \widehat g_t(x_t)}$.
\STATE Call subroutine to compute $C_{\opt}$-approximate optimal design $\des_t \in \Delta(\cA)$ for set $\curly*{\phi(x_t, a)}_{a\in\cA}$.\\
\hfill \algcommentlight{See \cref{dm_linear:alg:barycentric} for efficient solver.}
  \STATE Define $p_t \ldef \eps \cdot \des_t + (1-\epsilon) \cdot \indic_{\wh a_t}$.
  \STATE Sample $a_t \sim p_t$ and observe reward $r_t(a_t)$.
  \STATE Update oracle \sqalgtext with $(x_t, a_t, r_t(a_t))$.
\ENDFOR
	\end{algorithmic}
\end{algorithm}

\cref{dm_linear:alg:greedy} uses optimal design as a basis for exploration: At each round, the learner obtains an estimator $\wh f_t$ from the regression oracle \sqalgtext, then appeals to a subroutine to compute an (approximate) G-optimal design $\des_t \in \Delta(\cA)$ for the action embedding $\crl*{\phi(x_t,a)}_{a\in\cA}$.
Fix an exploration parameter $\eps > 0$, the algorithm then samples an action $a \sim \des_t$ from the optimal design with probability $\veps$ (``exploration''), or plays the greedy action $\wh a_t \ldef \argmax_{a \in \cA} \wh f_t(x_t,a)$ with probability $1-\eps$ (``exploitation''). \cref{dm_linear:alg:greedy} is efficient whenever an approximate optimal design can be computed efficiently, which can be achieved using \cref{dm_linear:alg:barycentric}. We defer a detailed discussion of efficiency for a moment, and first state the main regret bound for the algorithm.

\begin{restatable}{theorem}{thmGreedy}
	\label{dm_linear:thm:greedy}
With a $C_{\opt}$-approximate optimal design subroutine and an appropriate choice for $\veps\in(0,1]$, \cref{dm_linear:alg:greedy}, with probability at least $1-\delta$, enjoys regret
\begin{align*}
     \regcb(T)
      = \bigoh \prn*{ (C_{\opt} \cdot d)^{1/3} T^{2/3} (\regsq(T) + \log(\delta^{-1}))^{1/3}}.
\end{align*}	
In particular, when invoked with \cref{dm_linear:alg:barycentric} {(with $C=2$)} as a subroutine, the algorithm enjoys regret 
\begin{align*}
  \regcb(T)
   = \bigoh\prn*{ d^{2/3}T^{2/3} (\regsq(T) + \log(\delta^{-1}))^{1/3}}.
\end{align*}	
and has per-round runtime $O(\Tsq  + \Topt \cdot d^2 \log d  + d^4 \log d)$ and maximum memory $O( \Msq + \Mopt + d^2)$.
\end{restatable}

\paragraph{Computational efficiency}
The computational efficiency of \cref{dm_linear:alg:greedy} hinges on the ability to efficiently compute an approximate optimal design (or, by convex duality, the John ellipsoid \citep{john1948extremum}) for the set $\crl*{\phi(x_t,a)}_{a\in\cA}$.
All off-the-shelf optimal design solvers that we are aware of require solving quadratic maximization subproblems, which in general cannot be reduced to a linear optimization oracle (\cref{dm_linear:def:action_oracle}). While there are some special cases where efficient solvers exist (e.g., when $\cA$ is a polytope (\citet{cohen2019near} and references therein)), computing an exact optimal design is NP-hard in general \citep{grotschel2012geometric,summa2014largest}.
To overcome this issue, we use the notion of a \emph{barycentric spanner}, which acts as an approximate optimal design and can be computed efficiently using an action optimization oracle.

\begin{definition}[\citet{awerbuch2008online}]
\label{dm_linear:def:barycentric}
Let a compact set $\cZ \subseteq \R^d$ of full dimension be given. For $C\geq{}1$, a subset of points $\cS = \curly*{z_1, \dots, z_d} \subseteq \cZ$ is said to be a $C$-approximate barycentric spanner for $\cZ$ if every point $z \in \cZ$ can be expressed as a weighted combination of points in $\cS$ with coefficients in $[-C, C]$.
\end{definition}

The following result shows that any barycentric spanner yields an approximate optimal design.

\begin{restatable}{lemma}{propBary}
	\label{dm_linear:prop:barycentric_approx_opt_design}
    If $\cS = \crl{z_1, \dots, z_d}$ is a $C$-approximate barycentric spanner for $\cZ \subseteq \R^d$, then $q \ldef \unif(\cS)$ is a $(C^2 \cdot d)$-approximate optimal design.
\end{restatable}
Using an algorithm introduced by \citet{awerbuch2008online}, one can efficiently compute the $C$-approximate barycentric spanner for the set $\crl*{\phi(x,a)}_{a\in\cA}$ using $O(d^2 \log_C d)$ calls to the action optimization oracle; their method is restated as \cref{dm_linear:alg:barycentric} in \cref{dm_linear:app:warmup}.

\paragraph{Key features of \cref{dm_linear:alg:greedy}}
While the regret bound for \cref{dm_linear:alg:greedy} scales with $T^{2/3}$, which is not optimal, this result constitutes the first computationally efficient algorithm for contextual bandits with linearly structured actions and general function approximation. Additional features include:
\begin{itemize}
      \item \emph{Simplicity and practicality.} Appealing to uniform exploration makes \cref{dm_linear:alg:greedy} easy to implement and highly practical. In particular, in the case where the action embedding does not depend on the context (i.e., $\phi(x,a) = \phi(a)$) an approximate design can be precomputed and reused, reducing the per-round runtime to $\wt O(\Tsq + \Topt)$ and the maximum memory to $O(\Msq + d)$.
      \item \emph{Lifting optimal design to contextual bandits.} Previous bandit algorithms based on optimal design are limited to the non-contextual setting, and to pure exploration. Our result highlights for the first time that optimal design can be efficiently combined with general function approximation.

\end{itemize}

\paragraph{Proof sketch for \cref{dm_linear:thm:greedy}}
To analyze \cref{dm_linear:alg:greedy}, we follow a recipe introduced by \citet{foster2020beyond,foster2021statistical} based on the \emph{\dectext} (DEC),\footnote{The original definition of the Decision-Estimation Coefficient in \citet{foster2021statistical} uses Hellinger distance rather than squared error. The squared error version we consider here leads to tighter guarantees for bandit problems where the mean rewards serve as a sufficient statistic.} defined as $\dec_\gamma(\cF) \ldef \sup_{\wh{f} \in \conv(\cF), x \in \cX} \dec_{\gamma}(\cF;\wh{f},x)$, where
\begin{align}
\label{dm_linear:eq:dec}
	 \dec_{\gamma}(\cF; \wh{f}, x) \ldef \inf_{p \in \Delta(\cA)} \sup_{a^\star \in \cA} \sup_{f^\star \in \cF} 
	 \E_{a \sim p} \bigg[ f^\star(x, a^\star) - f^\star(x, a) - \gamma \cdot \prn{\wh{f}(x,a) - f^\star(x,a)}^2 \bigg] .
\end{align}
\citet{foster2021statistical} consider a meta-algorithm which, at each round $t$, (i) computes $\fhat_t$ by appealing to a regression oracle, (ii) computes a distribution $p_t\in\Delta(\cA)$ that solves the minimax problem in \cref{dm_linear:eq:dec} with $x_t$ and $\fhat_t$ plugged in, and (iii) chooses the action $a_t$ by sampling from this distribution. One can show (\cref{dm_linear:lm:regret_decomp} in \cref{dm_linear:app:warmup}) that for any $\gamma > 0$, this strategy enjoys the following regret bound:
\begin{align}
	\regcb(T) \approxleq T \cdot \dec_\gamma(\cF) + \gamma \cdot \regsq(T), \label{dm_linear:eq:decomposition}
\end{align}
More generally, if one computes a distribution that does not solve \cref{dm_linear:eq:dec} exactly, but instead certifies an upper bound on the DEC of the form $\dec_\gamma(\cF) \leq \wb \dec_\gamma(\cF)$, the same result holds with $\dec_\gamma(\cF)$ replaced by $\wb \dec_\gamma(\cF)$. \cref{dm_linear:alg:greedy} is a special case of this meta-algorithm, so to bound the regret it suffices to show that the exploration strategy in the algorithm certifies a bound on the DEC.
\begin{restatable}{lemma}{propDecGreedy}
  \label{dm_linear:prop:dec_greedy}
  For any $\gamma\geq{}1$, by choosing $\eps = \sqrt{C_{\opt} \cdot d/ 4\gamma} \wedge 1$, the exploration strategy in \cref{dm_linear:alg:greedy} certifies that $\dec_\gamma(\cF) = \bigoh(\sqrt{C_{\opt} \cdot d/\gamma})$.
\end{restatable}
Using \cref{dm_linear:prop:dec_greedy}, one can upper bound the first term in \cref{dm_linear:eq:decomposition} by \linebreak 
$O(T\sqrt{C_{\opt}d/\gamma})$.
The regret bound in \cref{dm_linear:thm:greedy} follows by choosing $\gamma$ to balance the two terms.

\section{Efficient, Near-Optimal Algorithms}
\label{dm_linear:sec:minimax}

In this section we present \spannerIGW (\cref{dm_linear:alg:igw}), an efficient algorithm with $\widetilde O(\sqrt{T})$ regret (\cref{dm_linear:alg:igw}). We provide the algorithm and statistical guarantees in \cref{dm_linear:sec:minimax_statistical}, then discuss computational efficiency in \cref{dm_linear:sec:minimax_computational}.

\subsection{Algorithm and Statistical Guarantees}
\label{dm_linear:sec:minimax_statistical}

Building on the approach in \cref{dm_linear:sec:warmup}, \spannerIGW uses the idea of exploration with an optimal design. However, in order to achieve $\sqrt{T}$ regret, we combine optimal design with the \emph{inverse gap weighting} (IGW) technique. previously used in the finite-action contextual bandit setting \citep{abe1999associative, foster2020beyond}.

Recall that for finite-action contextual bandits, the inverse gap weighting technique works as follows. Given a context $x_t$ and estimator $\fhat_t$ from the regression oracle \sqalgtext, we assign a distribution to actions in $\cA$ via the rule
\begin{align*}
p_t(a) \ldef \frac{1}{\lambda + \gamma\cdot\prn*{\fhat_t(x_t,\ahat_t) - \fhat_t(x_t,a)}},
\end{align*}
where $\ahat_t\ldef\argmax_{a\in\cA}\fhat_t(x_t,a)$ and $\lambda>0$ is chosen such that $\sum_{a}p_t(a)=1$. This strategy certifies that $\dec_{\gamma}(\cF; \fhat_t,x_t)\leq\frac{\abs{\cA}}{\gamma}$, which leads to regret $\bigoh\prn[\big]{\sqrt{\abs{\cA}T\cdot\regsq(T)}}$. While this is essentially optimal for the finite-action setting, the linear dependence on $\abs{\cA}$ makes it unsuitable for the large-action setting we consider.

To lift the IGW strategy to the large-action setting, \cref{dm_linear:alg:igw} combines it with optimal design with respect to a \emph{reweighted embedding}. Let $\fhat\in\cF$ be given. For each action $a \in \cA$, we define a reweighted embedding via
\begin{align}
    \wb{\phi}(x, a) \ldef  \frac{\phi(x,a)}{\sqrt{1 + \eta \paren*{ \wh f(x,\wh a) - \wh f(x, a)}}}, \label{dm_linear:eq:reweighting}
\end{align}
where $\ahat\ldef{}\argmax_{a\in\cA}\fhat(x,a)$ and $\eta >0$ is a reweighting parameter to be tuned later. This reweighting is \emph{action-dependent} since $\wh f(x,a)$ term appears on the denominator. Within \cref{dm_linear:alg:igw}, we compute a new reweighted embedding at each round $t \in [T]$ using $\wh{f}_t = f_{\wh{g}_t}$, the output of the regression oracle \sqalgtext.

\cref{dm_linear:alg:igw} proceeds by computing an optimal design $\des_t \in \Delta(\cA)$ with respect to the reweighted embedding defined in \cref{dm_linear:eq:reweighting}. The algorithm then creates a distribution $q_t \ldef \frac{1}{2} \des_t + \frac{1}{2} \indic_{\wh a_t}$ by mixing the optimal design with a delta mass at the greedy action $\wh a_t$. Finally, in \cref{dm_linear:eq:igw_reweight}, the algorithm computes an augmented version of the inverse gap weighting distribution by reweighting according to $q_t$. This approach certifies the following bound on the \dectext.
\begin{restatable}{lemma}{propDecIGW}
    \label{dm_linear:prop:dec_igw}
For any $\gamma>0$, by setting $\eta = \gamma/ (C_{\opt}\cdot d)$, the exploration strategy used in \cref{dm_linear:alg:igw} certifies that $\dec_\gamma(\cF) = O \prn{{C_{\opt} \cdot d}/{\gamma}}$.
\end{restatable}
This lemma shows that the reweighted IGW strategy enjoys the best of both worlds: By leveraging optimal design, we ensure good coverage for all actions, leading to $\bigoh(d)$ (rather than $\bigoh(\abs{\cA})$) scaling, and by leveraging inverse gap weighting, we avoid excessive exploration, leading $\bigoh(1/\gamma)$ rather than $\bigoh(1/\sqrt{\gamma})$ scaling. Combining this result with \cref{dm_linear:lm:regret_decomp} leads to our main regret bound for \spannerIGW.

\begin{algorithm}[H]
	\caption{\spannerIGW}
	\label{dm_linear:alg:igw} 
	\renewcommand{\algorithmicrequire}{\textbf{Input:}}
	\renewcommand{\algorithmicensure}{\textbf{Output:}}
	\newcommand{\algorithmicbreak}{\textbf{break}}
    \newcommand{\BREAK}{\STATE \algorithmicbreak}
	\begin{algorithmic}[1]
		\REQUIRE Exploration parameter $\gamma > 0$, online regression
                oracle \sqalgtext, action optimization oracle \optalgtext.
                \STATE Define $\eta\ldef{}\frac{\gamma}{C_{\opt}\cdot{}d}$.
\FOR{$t = 1, 2, \dots, T$}
		\STATE Observe context $x_t$.
		\STATE Receive $\widehat f_t = f_{\widehat g_t}$ from regression oracle \sqalgtext.
		\STATE Get $\widehat a_t \gets \argmax_{a \in \cA} \ang*{\phi(x_t, a), \widehat g_t(x_t)}$.
	\STATE Call subroutine to compute $C_{\opt}$-approximate optimal design $\des_t \in \Delta(\cA)$ for reweighted embedding $\curly*{\wb \phi(x_t, a)}_{a\in\cA}$ (\cref{dm_linear:eq:reweighting} with $\wh f = \wh f_t$). 
\hfill \algcommentlight{See \cref{dm_linear:alg:barycentric_reweight} for efficient solver.}
                  \STATE Define $q_t \ldef \frac{1}{2} \des_t + \frac{1}{2} \indic_{\wh a_t}$.
\STATE For each $a \in \supp(q_t)$, define \begin{align}
		    p_t(a) \ldef \frac{q_t(a)}{\lambda + \eta \prn*{ \wh f_t(x_t,\wh a_t) - \wh f_t(x_t, a)}}, \label{dm_linear:eq:igw_reweight}
		\end{align}
                where $\lambda \in [\frac{1}{2}, 1]$ is chosen so that $\sum_{a \in \supp(q_t)} p_t(a) = 1$. 
		 \STATE Sample $a_t \sim p_t$ and observe reward $r_t(a_t)$.
		 \STATE Update \sqalgtext with $(x_t, a_t, r_t(a_t))$.
\ENDFOR
	\end{algorithmic}
\end{algorithm}

\begin{restatable}{theorem}{thmIGW}
  \label{dm_linear:thm:igw}
  Let $\delta\in(0,1)$ be given. With a $C_{\opt}$-approximate optimal design subroutine and an appropriate choice for $\gamma>0$, \cref{dm_linear:alg:igw} ensures that with probability at least $1-\delta$, 
\begin{align*}
\regcb(T) & = \bigoh \prn*{ \,{\sqrt{C_{\opt} \cdot d \,T \, \prn[\big]{\regsq(T)+ \log( \delta^{-1})}}} }.
\end{align*}	
In particular, when invoked with \cref{dm_linear:alg:barycentric_reweight} {(with $C= 2$)} as a subroutine, the algorithm has
\begin{align*}
    \regcb(T) = \bigoh \prn*{d \, \sqrt{T \, \prn[\big]{\regsq(T)+ \log(\delta^{-1})}}},
\end{align*}
and has per-round runtime $O(\Tsq + \prn{\Topt \cdot d^{3} +d^4} \cdot \log^{2} \prn[\big]{ \frac{T}{r}})$ and the maximum memory $O(\Msq + \Mopt + d^2 + d \log \prn[\big]{\frac{T}{r}} )$.
\end{restatable}
\cref{dm_linear:alg:igw} is the first computationally efficient algorithm with $\sqrt{T}$-regret for contextual bandits with
general function approximation and linearly structured action
spaces. In what follows, we show how to leverage the action optimization oracle (\cref{dm_linear:def:action_oracle}) to achieve this efficiency.

\subsection{Computational Efficiency}
\label{dm_linear:sec:minimax_computational}
The computational efficiency of \cref{dm_linear:alg:igw} hinges on the ability to efficiently compute an optimal design. As with \cref{dm_linear:alg:greedy}, we address this issue by appealing to the notion of a barycentric spanner, which serves as an approximate optimal design. However, compared to \cref{dm_linear:alg:greedy}, a substantial additional challenge is that \cref{dm_linear:alg:igw} requires an approximate optimal design for the \emph{reweighted} embeddings. Since the reweighting is action-dependent, the action optimization oracle \optalgtext cannot be directly applied to optimize over the reweighted embeddings, which prevents us from appealing to an out-of-the-box solver (\cref{dm_linear:alg:barycentric}) in the same fashion as the prequel.

\begin{algorithm}[H]
\caption{\textsf{ReweightedSpanner}}
	\label{dm_linear:alg:barycentric_reweight} 
	\renewcommand{\algorithmicrequire}{\textbf{Input:}}
	\renewcommand{\algorithmicensure}{\textbf{Output:}}
	\newcommand{\algorithmicbreak}{\textbf{break}}
	\newcommand{\BREAK}{\STATE \algorithmicbreak}
\begin{algorithmic}[1]
		\REQUIRE Context $x \in \cX$, oracle prediction $\wh g(x) \in \R^d$, 
		{action $\wh a \ldef \argmax_{a \in \cA} \ang{\phi(x,a), \wh g(x)}$,}
		reweighting parameter $\eta > 0$, approximation factor $C > \sqrt{2}$, initial set $\cS = \prn{a_1, \dots, a_d}$ with $\abs{\det(\phi(x, \cS))} \geq r^d$ for $r \in (0, 1)$.
		\WHILE{not break}
		\FOR{$i = 1, \dots, d$}
		\STATE Compute $\theta \in \R^d$ representing linear function $\wb \phi(x,a) \mapsto \det(\wb \phi(x,\cS_i(a)))$, where $\cS_i(a) \ldef (a_1, \ldots,a_{i-1}, a, a_{i+1}, \ldots, a_d)$.
\hfill\algcommentlight{$\wb \phi$ is computed from $f_{\ghat}$, $\ahat$, and $\eta$ via \cref{dm_linear:eq:reweighting}.}
		\STATE Get $a \gets \argmaxIGW(\theta; x, \wh g(x), \eta, r)$. 
                \hfill\algcommentlight{\cref{dm_linear:alg:igw_argmax}.}
		\IF{\mbox{$\abs*{\det( \wb \phi(x, \cS_{i}(a)))} \geq \frac{\sqrt{2}C}{2} \abs*{\det(\wb \phi(x, \cS))}$}}
		\STATE Update $a_i \gets a$.\STATE \textbf{continue} to line 2.
		\ENDIF
		\ENDFOR
		\BREAK
		\ENDWHILE
		\RETURN $C$-approximate barycentric spanner $\cS$.
	\end{algorithmic}
\end{algorithm}

To address the challenges above, we introduce \rwalgtext (\cref{dm_linear:alg:barycentric_reweight}), a barycentric spanner computation algorithm which is tailored to the reweighted embedding $\phibar$. 
{To describe the algorithm, let us introduce some additional notation. For a set $\cS\subseteq\cA$ of $d$ actions, we let $\det(\wb \phi(x, \cS))$ denote the determinant of the $d$-by-$d$ matrix whose columns are $\crl*{\phibar(x,a)}_{a\in\cA}$. \rwalgtext adapts the barycentric spanner computation approach of \citet{awerbuch2008online}, which aims to identify a subset $\cS\subseteq\cA$ with $\abs{\cS}=d$ that approximately maximizes $\abs{\det(\wb \phi(x, \cS))}$.}
The key feature of \rwalgtext is a subroutine, \argmaxIGW (\cref{dm_linear:alg:igw_argmax}), which implements an (approximate) action optimization oracle for the reweighted embedding:
\begin{equation}
  \argmax_{a\in\cA}\tri*{\phibar(x,a),\theta}.\label{dm_linear:eq:reweighted_argmax}
\end{equation}
\argmaxIGW uses line search reduce the problem in \cref{dm_linear:eq:reweighted_argmax} to a sequence of linear optimization problems with respect to the \emph{unweighted} embeddings, each of which can be solved using \optalgtext. This yields the following guarantee for \cref{dm_linear:alg:barycentric_reweight}.

\begin{restatable}{theorem}{thmBarycentricReweight}
	\label{dm_linear:thm:barycentric_reweight}
	Suppose that \cref{dm_linear:alg:barycentric_reweight} is invoked with parameters $\eta > 0$, $r\in(0,1)$, and $C > \sqrt{2}$, and that the initialization set $\cS$ satisfies $\abs{\det(\phi(x,\cS))} \geq r^d$. Then the algorithm returns a $C$-approximate barycentric spanner with respect to the reweighted embedding set $\crl*{\wb \phi(x,a)}_{a\in\cA}$, and does so with 
	$O(\prn{\Topt \cdot d^{3} +d^4} \cdot \log^{2} (e \vee \frac{\eta}{r}))$ runtime and $O(\Mopt + d^2 + d \log (e \vee \frac{\eta}{r}) )$ memory.	
      \end{restatable}
We refer to \cref{dm_linear:app:argmax_reweight} for self-contained analysis of \argmaxIGW.

\begin{algorithm}[]
  \caption{\argmaxIGW}
  \label{dm_linear:alg:igw_argmax} 
  \renewcommand{\algorithmicrequire}{\textbf{Input:}}
  \renewcommand{\algorithmicensure}{\textbf{Output:}}
  \newcommand{\algorithmicbreak}{\textbf{break}}
  \newcommand{\BREAK}{\STATE \algorithmicbreak}
  \begin{algorithmic}[1]
	  \REQUIRE Linear parameter $\theta \in \R^{d}$, context $x \in \cX$, oracle prediction $\wh g(x) \in \R^d$, reweighting parameter $\eta >0$, {initialization} constant $r \in (0,1)$. 
	  \STATE Define $N \ldef \ceil{d \log_{\frac{4}{3}} \prn{\frac{2\eta + 1}{r}}}$.
\STATE Define \mbox{$\cE \ldef  \crl{\prn{\frac{3}{4}}^i}_{i=1}^N \cup \crl{- \prn{\frac{3}{4}}^i}_{i=1}^N$}.
	  \STATE Initialize $\wh \cA = \emptyset$.
	  \FOR{each $\eps \in \cE$}
	  \STATE Compute $\wb \theta \gets 2 \eps \theta + {\eps^2 \eta} \cdot  \wh g(x)$.
\STATE \mbox{Get $a \gets \argmax_{a \in \cA} \ang{\phi(x,a), \wb \theta}$; add $a$ to $\wh \cA$.}
	  \ENDFOR
	  \RETURN $\argmax_{a \in \wh{\cA}} {\ang{\wb \phi(x,a), \theta}}^2$ 
\hfill \algcommentlight{$\wt O(d)$ candidates.}
  \end{algorithmic}
\end{algorithm}

\paragraph{On the initialization requirement}
The runtime for \cref{dm_linear:alg:barycentric_reweight} scales with $\log(r^{-1})$, where $r\in(0,1)$ is such that $\det(\phi(x,\cS))\geq{}r^{d}$ for the initial set $\cS$. In \cref{dm_linear:app:det_assumption}, we provide computationally efficient algorithms for initialization under various assumptions on the action space.

\section{Empirical Results}
\label{dm_linear:sec:experiments}
In this section we investigate the empirical performance of \greedyalg
and \spannerIGW through three experiments. First, we compare the spanner-based
algorithms to state-of-the art finite-action algorithms on a
large-action dataset; this experiment features nonlinear, learned
context embeddings $g\in\cG$.
Next, we study the impact of redundant actions on the statistical
performance of said algorithms.  Finally, we experiment with a large-scale large-action contextual
bandit benchmark, where we find that the spanner-based methods exhibit
excellent performance.

\paragraph{Preliminaries}
{We conduct experiments on three datasets, whose details are summarized in \cref{dm_linear:tab:datasets}.
\oneshotwiki \citep{singh12:wiki-links,oneshotwiki} is a
named-entity recognition task where contexts are text phrases
preceding and following the mention text, and where actions are text
phrases corresponding to the concept names.  
\amazon-3m \citep{Bhatia16} is an extreme multi-label dataset whose
contexts are text phrases corresponding to the title and description
of an item, and whose actions are integers corresponding to item tags.
Actions are embedded into $\R^{d}$ with $d$ specified in \cref{dm_linear:tab:datasets}.
We construct binary rewards for each dataset, and report 90\% bootstrap confidence intervals (CIs) 
of the rewards in the experiments.
We defer other experimental details to \cref{dm_linear:app:exp_basic}.
}
Code to reproduce all results is available at \url{https://github.com/pmineiro/linrepcb}.
\begin{table}[ht]
\caption{Details of datasets used in experiments.}
\label{dm_linear:tab:datasets}
\centering
\begin{tabular}{c c c c}
\toprule
Dataset & $T$ & $|\cA|$ & $d$ \\
\midrule
\oneshotwiki-\textsf{311} & 622000 & 311 & 50 \\
\oneshotwiki-\textsf{14031} & 2806200 & 14031 & 50 \\
\amazon-\textsf{3m} & 1717899 & 2812281 & 800 \\
\bottomrule
\end{tabular}
\end{table}

\paragraph{Comparison with finite-action baselines}  We compare \greedyalg
and \mainalg with their finite-action counterparts \greedy and
\squarecb \citep{foster2020beyond} on the \oneshotwiki-\textsf{14031}
dataset. We consider \emph{bilinear models} in which regression functions take the form $f(x, a) = \ang{\phi(a), W x}$
where $W$ is a matrix of learned parameters; the \emph{deep
  models} of the form $f(x, a) = \ang{\phi(a), W \wb g(x)}$, where
$\wb g$ is a learned two-layer neural network {and $W$ contains learned parameters as before}.\footnote{Also see \cref{dm_linear:app:exp_basic} for details.}
\cref{dm_linear:tab:oneshotresults} presents our results. We find that \mainalg performs best, and that both spanner-based algorithms
either tie or exceed their finite-action counterparts. In addition, we
find that working with deep models uniformly improves performance for
all methods. We refer to \cref{dm_linear:tab:timings} in \cref{dm_linear:app:exp_timing} for timing
information.

\begin{table}[ht]
\caption{Comparison on \oneshotwiki-\textsf{14031}.  Values are
the average progressive rewards {(confidence intervals)}, scaled by 1000. We include the performance of the best constant predictor (as a baseline) and the supervised learner (as a skyline).}
\label{dm_linear:tab:oneshotresults}
\centering
\begin{tabular}{c c c }
\toprule
Algorithm & \multicolumn{2}{c}{Regression Function} \\
 & Bilinear & Deep \\
\midrule
\textsf{best constant} & \multicolumn{2}{c}{$0.07127$} \\
\greedy & $[ 5.00, 6.27 ]$ & $[ 7.15, 8.52 ]$ \\
\greedyalg & $[ 6.29, 7.08 ]$ & $[ 6.67, 8.30 ]$ \\
\squarecb & $[ 7.57, 8.59 ]$ & $[ 10.4, 11.3 ]$ \\
\mainalg & $[ 8.84, 9.68 ]$ & $[ 11.2, 12.2 ]$ \\
\supervised & $[ 31.2, 31.3 ]$ & $[ 36.7, 36.8 ]$ \\
\bottomrule
\end{tabular}
	\end{table}

\paragraph{Impact of redundancy}
Finite-action contextual bandit algorithms can explore excessively in
the presence of redundant actions. To evaluate performance in the face
of redundancy, we augment
\oneshotwiki-\textsf{311} by duplicating action the final
action. \cref{dm_linear:tab:duplicateaction} displays the performance of
\spannerIGW and its finite-action counterpart, \squarecb, with a
varying number of duplicates. We find that \mainalg is completely
invariant to duplicates (in fact, the algorithm produces numerically
identical output when the random seed is fixed), but \squarecb is
negatively impacted and over-explores the duplicated action.  \greedyalg
and \greedy behave analogously (not shown).

\begin{table}[ht]
\caption{Redundancy study on \oneshotwiki-\textsf{311}.  Values are
the average progressive rewards {(confidence intervals)}, scaled by 100.}
\label{dm_linear:tab:duplicateaction}
\centering
\begin{tabular}{c c c }
\toprule
Duplicates & \mainalg & \squarecb \\
\midrule
0 & $[ 12.6, 13.0 ]$ & $[ 12.2, 12.6 ]$ \\
16 & $[ 12.6, 13.0 ]$ & $[ 12.1, 12.4 ]$ \\
256 & $[ 12.6, 13.0 ]$ & $[ 10.2, 10.6 ]$ \\
1024 & $[ 12.6, 13.0 ]$ & $[ 8.3, 8.6 ]$ \\
\bottomrule
\end{tabular}
\end{table}

\paragraph{Large scale exhibition}
We conduct a large scale experiment using the \amazon-\textsf{3m} dataset.
Following \citet{sen2021top}, we study the top-$k$ setting where $k$ actions are selected 
at each round.
Out of the total number of actions sampled, we let $r$ denote the number of actions sampled for exploration.
We apply \greedyalg for this dataset 
{and consider regression functions similar to the deep models discussed before.}
The setting $(k=1)$ corresponds to running our algorithm
unmodified, and $(k=5, r=3)$ corresponds to selecting 5 actions per round and using 3 exploration slots.
\cref{dm_linear:fig:amazon3m} in
\cref{dm_linear:app:exp_figure} displays the results. For $(k=1)$ the final CI is
$[0.1041, 0.1046]$, and for $(k=5,r=3)$ the final CI is $[0.438, 0.440]$.

{In the setup} with $(k=5, r=3)$, our results are directly comparable to
\citet{sen2021top}, who evaluated a tree-based contextual bandit
method on the same dataset. The best result from \cite{sen2021top}
achieves roughly 0.19 reward with $(k=5, r=3)$,  which we exceed by a factor of 2.
This indicates that our use of embeddings provides favorable inductive bias for
this problem, and underscores the broad utility of our techniques (which leverage embeddings).  For $(k=5,
r=3)$, our inference time on a commodity CPU with batch size 1 is
160ms per example, which is slower than the time of 7.85ms per example
reported in \citet{sen2021top}.

\section{Discussion}
\label{dm_linear:sec:discussion}

We provide the first efficient algorithms for contextual bandits with
continuous, linearly structured action spaces and general-purpose
function approximation.
We highlight some natural directions for future research below.

\begin{itemize}
  \item \textbf{Efficient algorithms for nonlinear action 
      spaces.} Our algorithms take advantage of linearly structured
    action spaces by appealing to optimal design. Can we develop
    computationally efficient methods for contextual bandits with
    nonlinear dependence on the action space?
\item \textbf{Reinforcement learning.} The contextual bandit problem is a special case of the
    reinforcement learning problem with horizon one. Given our
    positive results in the contextual bandit setting, a natural next
    step is to extend our methods to reinforcement learning problems
    with large action/decision spaces. For example,
    \citet{foster2021statistical} build on our computational tools to
    provide efficient algorithms for reinforcement learning with bilinear classes.

\end{itemize}
  Beyond these directions, natural domains in which to extend our
  techniques include pure exploration and off-policy learning with linearly structured actions.

\section{Proofs and Supporting Results}
\label{dm_linear:sec:proofs}

\subsection{Proofs and Supporting Results for \cref{dm_linear:sec:warmup}}
\label{dm_linear:app:warmup}
This section is organized as follows. We provide supporting results in \cref{dm_linear:app:warmup_support}, then give the proof of \cref{dm_linear:thm:greedy} in \cref{dm_linear:app:warmup_greedy}.

\subsubsection{Supporting Results}
\label{dm_linear:app:warmup_support}

\paragraph{Barycentric Spanner and Optimal Design}
\paranewline

\noindent
\cref{dm_linear:alg:barycentric} restates an algorithm of
\citet{awerbuch2008online}, which efficiently computes a barycentric
spanner (\cref{dm_linear:def:barycentric}) given access to a linear optimization
oracle (\cref{dm_linear:def:action_oracle}). 
{Recall that, for a set $\cS\subset\cA$ of $d$ actions, the notation $\det(\wb \phi(x, \cS))$ (resp. $\det(\phi(x,\cS))$) denotes the determinant of the $d$-by-$d$ matrix whose columns are the $\wb \phi$ (resp. $\phi$) embeddings of actions.} 
\begin{algorithm}[H]
    \caption{Approximate Barycentric Spanner \citep{awerbuch2008online}}
    \label{dm_linear:alg:barycentric} 
    \renewcommand{\algorithmicrequire}{\textbf{Input:}}
    \renewcommand{\algorithmicensure}{\textbf{Output:}}
    \newcommand{\algorithmicbreak}{\textbf{break}}
    \newcommand{\BREAK}{\STATE \algorithmicbreak}
    \begin{algorithmic}[1]
        \REQUIRE Context $x \in \cX$ and approximation factor $C > 1$.
\FOR{$i = 1, \dots, d$}
        \STATE Compute $\theta \in \R^d$ representing linear function $\phi(x,a) \mapsto \det(\phi(x,a_1), \ldots, \phi(x, a_{i-1}), \phi(x,a), e_{i+1}, \ldots, e_d)$.

\STATE Get $a_i \gets \argmax_{a \in \cA} \abs{\ang{\phi(x,a), \theta}}$.\ENDFOR
	\STATE Construct $\cS = \prn{a_1, \ldots, a_d}$. \hfill \algcommentlight{Initial set of actions $\cS \subseteq \cA$ such that $\abs{\cS} = d$ and $\abs{\det(\phi(x,\cS))} > 0$.}
        \WHILE{not break}
		\FOR{$i = 1, \dots, d$}
		\STATE Compute $\theta \in \R^d$ representing linear function $\phi(x,a) \mapsto \det(\phi(x, \cS_{i}(a)))$, where $\cS_i(a) \ldef ( a_1, \ldots , a_{i-1}, a, a_{i+1}, \ldots, a_d )$.
\STATE Get $a \gets \argmax_{a \in \cA} \abs{\ang{\phi(x,a), \theta}}$.
		\IF{$\abs{\det( \phi(x,\cS_{i}(a)))} \geq C \abs{\det(\phi(x, \cS))}$}
		\STATE Update $a_i \gets a$. \STATE \textbf{continue} to line 5.
		\ENDIF
		\ENDFOR
		\BREAK
		\ENDWHILE
        \RETURN $C$-approximate barycentric spanner $\cS$.
    \end{algorithmic}
\end{algorithm}

\begin{lemma}[\citet{awerbuch2008online}]
  \label{dm_linear:prop:barycentric_basic}
For any $x \in \cX$, \cref{dm_linear:alg:barycentric} computes a $C$-approximate barycentric spanner for $\crl{\phi(x,a): a \in \cA}$ within $O(d \log_C d)$ iterations of the while-loop.
\end{lemma}

\begin{lemma}
\label{dm_linear:prop:barycentric_basic_computational}	
Fix any constant $C > 1$.
   \cref{dm_linear:alg:barycentric} can be implemented with 
   runtime $O(\Topt \cdot d^2 \log d + d^4 \log d)$ and memory $O(\Mopt + d^2)$.
\end{lemma}

\begin{proof}[\pfref{dm_linear:prop:barycentric_basic_computational}]
We provide the computational complexity analysis starting from the while-loop (line 5-12) in the following.
The computational complexity regarding the first for-loop (line 1-3) can be similarly analyzed.
\begin{itemize}
	\item \emph{Outer loops (lines 5-6).}
From \cref{dm_linear:prop:barycentric_basic}, we know that \cref{dm_linear:alg:barycentric} terminates within $O(d \log d)$  iterations of the while-loop (line 5). 
	It is also clear that the for-loop (line 6) is invoked at most $d$ times.
	 \item \emph{Computational complexity for lines 7-10.}
		 We discuss how to efficiently implement this part using rank-one updates.
We analyze the computational complexity for each line in the following.
\begin{itemize}
	\item \emph{Line 7.}
We discuss how to efficiently compute the linear function $\theta$ through rank-one updates.
	Fix any $Y \in \R^{d}$. 
   Let $\Phi_\cS$ denote the invertible (by construction) matrix whose $k$-th column is $\phi(x,a_k)$ (with $a_k \in \cS$). Using the rank-one update formula for the determinant \citep{meyer2000matrix}, we have 
   \begin{align}
& \det( \phi(x,a_1),\ldots, \phi(x,a_{i-1}), Y, \phi(x,a_{i+1}),\ldots, \phi(x, a_d) ) \nonumber \\
       & = \det \prn[\Big]{\Phi_\cS + \prn[\big]{ Y - \phi(x,a_i)} e_i^\top} \nonumber\\
       & = \det(\Phi_\cS) \cdot \prn[\Big]{1 + e_i^\top \Phi_\cS^{-1} \prn[\big]{Y - \phi(x,a_i)}} \nonumber \\
       & = \ang[\big]{Y, \det(\Phi_\cS) \cdot \prn*{\Phi_\cS^{-1}}^\top e_i}  + \det(\Phi_\cS) \cdot \prn[\big]{ 1 - e_i^\top \Phi_\cS^{-1} \phi(x,a_i) }. \label{dm_linear:eq:det_rank_one} 
   \end{align}
   We first notice that $\det(\Phi_\cS) \cdot \prn[\big]{ 1 - e_i^\top \Phi_\cS^{-1} \phi(x,a_i) } = 0$ since one can take $Y = 0 \in \R ^{d}$. 
   We can then write 
   \begin{align*}
 \det( \phi(x,a_1),\ldots, \phi(x,a_{i-1}), Y, \phi(x,a_{i+1}),\ldots, \phi(x, a_d) ) = \ang{Y, \theta} 
   \end{align*}
   where $\theta = \det(\Phi_\cS) \cdot \prn*{\Phi_\cS^{-1}}^\top e_i$.
   Thus, whenever $\det(\Phi_\cS)$ and  $\Phi_\cS^{-1}$ are known, compute $\theta$ takes  $O(d)$ time. 
   The maximum memory requirement is $O(d^2)$, following from the storage of $\Phi_\cS^{-1}$.
   \item \emph{Line 8.} When $\theta$ is computed, we can compute  $a$ by first compute  $a_+ \ldef \argmax_{a \in \cA} \ang{\phi(x,a), \theta}$ and $a_- \ldef \argmax_{a \in \cA} - \ang{\phi(x,a), \theta}$ and then compare the two. This process takes two oracle calls to \optalgtext, which takes $O(\Topt)$ time.
	   The maximum memory requirement is $O(\Mopt + d)$, following from the memory requirement of \optalgtext and the storage of $\theta$.
\item \emph{Line 9.}
	Once $\theta$ and $\det(\Phi_\cS)$ are computed, checking the updating criteria takes  $O(d)$ time. The maximum memory requirement is $O(d)$, following from the storage of $\phi(x, a)$ and $\theta$.
 \item \emph{Line 10.}
We discuss how to efficiently update $\det(\Phi_\cS)$ and  $\Phi_\cS^{-1}$ through rank-one updates.
	If an update $a_i = a$ is made, we can update the determinant using rank-one update (as in \cref{dm_linear:eq:det_rank_one}) with runtime $O(d)$ and memory  $O(d^2)$; and update the inverse matrix using the Sherman-Morrison rank-one update formula \citep{sherman1950adjustment}, i.e.,
\begin{align*}
       \prn[\Big]{\Phi_\cS + \prn[\big]{ \phi(x,a) - \phi(x,a_i)} e_i^\top}^{-1} = \Phi_\cS^{-1} -
       \frac{\Phi_\cS^{-1}\prn[\big]{ \phi(x,a) - \phi(x,a_i)} e_i^\top \Phi_\cS^{-1}}{1 + e_i \Phi_\cS^{-1} \prn[\big]{ \phi(x,a) - \phi(x,a_i)} },
\end{align*}
which can be implemented in $O(d^2)$ time and memory.
Note that the updated matrix must be invertible by construction.
\end{itemize}
Thus, using rank-one updates, the total runtime adds up to $O(\Topt + d^2)$ and the maximum memory requirement is $O(\Mopt + d^2)$.
We also remark that the initial matrix determinant and inverse can be computed cheaply since the first iteration of the first for-loop (i.e., line 2 with $i=1$) is updated from the identity matrix.
\end{itemize}
   To summarize,
   \cref{dm_linear:alg:barycentric} has runtime $O(\Topt \cdot d^2 \log d + d^4 \log d)$ 
   and uses at most $O(\Mopt + d^2) $ units of memory. 
\end{proof}

The next proposition shows that a barycentric spanner implies an approximate optimal design. The result is well-known (e.g., \citet{hazan2016volumetric}), but we provide a proof here for completeness.

\propBary*

\begin{proof}[\pfref{dm_linear:prop:barycentric_approx_opt_design}]
  Assume without loss of generality that $\cZ \subseteq \R^d$ spans $\R^d$. By \cref{dm_linear:def:barycentric}, we know that for any $z \in \cZ$, we can represent $z$ as a weighted sum of elements in $\cS$ with coefficients in the range $[-C, C]$. Let $\Phi_\cS \in \R^{d \times d}$ be the matrix whose columns are the vectors in $\cS$. For any $z \in \cZ$, we can find $\theta \in [-C,C]^d$ such that $z = \Phi_\cS \theta$.
   Since $\Phi_\cS$ is invertible (by construction), we can write $\theta = \Phi_\cS^{-1} z$, which implies the result via
   \begin{align*}
       C^2 \cdot d \geq \norm{\theta}_2^2 = \norm{z}^2_{(\Phi_\cS \Phi_\cS^\top)^{-1}} = \frac{1}{d}\cdot \norm{z}^2_{V(q)^{-1}}.
   \end{align*}
\end{proof}

\paragraph{Regret Decomposition}
\paranewline

\noindent
Fix any $\gamma > 0$.
We consider the following meta algorithm that utilizes the online regression oracle \sqalgtext defined in \cref{dm_linear:asm:regression_oracle}.

For $t = 1, 2, \ldots, T$:
\begin{itemize}
\item Get context $x_t \in \cX$ from the environment and regression function $\wh f_t \in \conv(\cF)$ from the online regression oracle \sqalgtext.
\item Identify the distribution $p_t \in \Delta(\cA)$ that solves the
  minimax problem \linebreak 
  $\dec_\gamma(\cF; \wh f_t, x_t)$ (defined in
  \cref{dm_linear:eq:dec}) and play action $a_t \sim p_t$.
\item Observe reward $r_t$ and update regression oracle with example
  $(x_t, a_t, r_t)$.
\end{itemize}

The following result bounds the contextual bandit regret for the meta algorithm described above. 
The result is a variant of the regret decomposition based on
the \dectext given in \citet{foster2021statistical}, which generalizes
\citet{foster2020beyond}. The slight
differences in constant terms are due to the difference in reward range.
\begin{lemma}[\citet{foster2020beyond,foster2021statistical}]
    \label{dm_linear:lm:regret_decomp}
   Suppose that \cref{dm_linear:asm:regression_oracle} holds. Then probability at least $1-\delta$, the contextual bandit regret is upper bounded as follows:
   \begin{align*}
       \regcb(T) \leq \dec_{\gamma}(\cF)\cdot{}T + 2 \gamma \cdot \regsq(T) + 64 \gamma \cdot \log (2\delta^{-1}) + \sqrt{8T \log(2 \delta^{-1})}.
   \end{align*} 
\end{lemma}

In general, identifying a distribution that \emph{exactly} solves the
minimax problem corresponding to the DEC may be impractical. However,
if one can identify a distribution that instead certifies an \emph{upper bound}
$\wb{\dec}_{\gamma}(\cF)$ on the \dectext (in the sense that
$\dec_\gamma(\cF) \leq \wb \dec_\gamma(\cF)$), the regret bound in
\cref{dm_linear:lm:regret_decomp} continues to hold with $\dec_\gamma(\cF)$
replaced by $\wb \dec_\gamma(\cF)$.

\paragraph{\pfref{dm_linear:prop:dec_greedy}}

\noindent
\propDecGreedy*
\begin{proof}[\pfref{dm_linear:prop:dec_greedy}]
  Fix a context $x \in \cX$. In our setting, where actions are
  linearly structured, we can equivalently write the \dectext
  $\dec_{\gamma}(\cF;\fhat,x)$ as
\begin{align}
& \dec_{\gamma}  (\cG; \wh{g}, x) \ldef \nonumber \\
&\inf_{p \in \Delta(\cA)} \sup_{a^\star \in \cA} \sup_{g^\star \in \cG} \E_{a \sim p} \Big[\ang[\big]{\phi(x,a^\star) - \phi(x, a), g^\star(x)}  - {\gamma} \cdot \prn[\big]{\ang[\big]{\phi(x,a), g^\star(x) - \widehat g(x)}}^2 \bigg]. \label{dm_linear:eq:dec_linear}
\end{align}
Recall that within our algorithms, $\wh g \in
\conv(\cG)$ is obtained from the estimator $\wh f = f_{\wh g}$ output
by \sqalgtext. We will bound the quantity in
  \cref{dm_linear:eq:dec_linear} uniformly
  for all $x\in\cX$ and $\wh{g}:\cX\to\bbR^d$ with $\nrm{\ghat}\leq{}1$.
  Recall that we assume $\sup_{g \in \cG, x \in \cX} \nrm{g(x)} \leq 1$.
  
Denote $\wh a \ldef \argmax_{a \in \cA} \ang[\big]{\phi(x,a), \wh g(x)}$ and $a^\star \ldef \argmax_{a \in \cA} \ang[\big]{\phi(x,a), g^\star(x)}$. 
For any $\eps \leq 1$,
let $p \ldef \eps \cdot \des + (1 - \eps)\cdot \indic_{\wh a}$, where
$\des \in \Delta(\cA)$ is any $C_{\opt}$-approximate optimal design
for the embedding $\crl*{\phi(x,a)}_{a\in\cA}$. We have the following decomposition.
\begin{align}
    \E_{a \sim p} \sq[\Big]{\ang[\big]{\phi(x,a^\star) - \phi(x, a), g^\star(x)}} & = 
    \E_{a \sim p} \sq[\Big]{\ang[\big]{\phi(x, \wh a) - \phi(x,a), \wh g(x)}} \nonumber \\
										  & \quad + \E_{a \sim p} \sq[\Big]{\ang[\big]{\phi(x,a), \wh g(x) - g^\star(x)}} \nonumber\\ 
    & \quad +  \prn[\Big]{\ang[\big]{\phi(x,a^\star), g^\star(x)}
    - \ang[\big]{\phi(x,\wh a), \wh g(x)}}.
    \label{dm_linear:eq:dec_greedy_decompostion}
\end{align}
For the first term in \cref{dm_linear:eq:dec_greedy_decompostion}, we have
\begin{align*}
\E_{a \sim p} \sq[\Big]{\ang[\big]{\phi(x, \wh a) - \phi(x,a), \wh g(x)}} & = 
    \eps \cdot \E_{a \sim \des} \sq[\Big]{\ang[\big]{\phi(x, \wh a) - \phi(x,a), \wh g(x)}}  \\
									  & \leq 2 \eps \cdot \sup_{x \in \cX,a \in \cA}  \nrm{\phi(x,a)} \cdot \sup_{x \in \cX}  \nrm{\wh g(x)} \\
									  &\leq  2 \eps.
\end{align*}
Next, since
\begin{align*}
    \ang[\big]{\phi(x,a), \wh g(x) - g^\star(x)} \leq \frac{\gamma}{2} \cdot \prn[\big]{\ang[\big]{\phi(x,a), \wh g(x) - g^\star(x)}}^2 + \frac{1}{2 \gamma}
\end{align*}
by AM-GM inequality, we can bound the second term in
\cref{dm_linear:eq:dec_greedy_decompostion} by
\begin{align*}
    \E_{a \sim p} \sq[\Big]{\ang[\big]{\phi(x,a), \wh g(x) - g^\star(x)}} 
    \leq \frac{\gamma}{2} \cdot \E_{a \sim p} \sq[\Big]{\prn[\big]{\ang[\big]{\phi(x,a), \wh g(x) - g^\star(x)}}^2 }
    + \frac{1}{2 \gamma}.
\end{align*}
We now turn our attention to the third term. Observe that since
$\wh{a}$ is \linebreak 
optimal for $\ghat$, $\ang[\big]{\phi(x,\wh a), \wh g(x)}
\geq \ang[\big]{\phi(x,a^\star), \wh g(x)}$. As a result, defining \linebreak
$V(\des) \ldef \E_{a \sim
  \des} \sq{\phi(x,a) \phi(x,a)^\top}$, we have 
\begin{align*}
     \ang[\big]{\phi(x,a^\star), g^\star(x)} - \ang[\big]{\phi(x,\wh a), \wh g(x)} & \leq  \ang[\big]{\phi(x,a^\star) , g^\star(x) - \wh g(x) }\\
     & \leq \nrm[\big]{\phi(x,a^\star)}_{V(\des)^{-1}} \cdot \nrm[\big]{g^\star(x) - \wh g(x)  }_{V(\des)} \\
     & = \frac{1}{2 \gamma \eps}\cdot \nrm[\big]{\phi(x,a^\star)}^2_{V(\des)^{-1}} \\
     & \quad + \frac{\gamma}{2} \cdot \eps \cdot \E_{a \sim \des} \sq[\Big]{\prn[\big]{\phi(x,a), g^\star(x) - \wh g(x)}^2} \\
     & \leq \frac{C_{\opt} \cdot d}{2 \gamma \eps} + \frac{\gamma}{2}\cdot \E_{a \sim p} \sq[\Big]{\prn[\big]{\phi(x,a), g^\star(x) - \wh g(x)}^2}.
\end{align*}
Here, the third line follows from the AM-GM inequality, and the last
line follows from the ($C_{\opt}$-approximate) optimal design property and the definition of $p$.

Combining these bounds, we have
\begin{align*}
    \dec_\gamma(\cF) = \inf_{p \in \Delta(\cA)} \sup_{a^\star \in \cA} \sup_{g^\star \in \cG}\dec_\gamma(\cG; \wh g, x) \leq 2 \eps + \frac{1}{2 \gamma} + \frac{C_{\opt} \cdot d}{2 \gamma \eps}.
\end{align*}
Since $\gamma \geq 1$, taking $\eps \ldef \sqrt{C_{\opt} \cdot d/ 4\gamma} \wedge 1$ gives
\begin{align*}
    \dec_\gamma (\cF) \leq 2 \sqrt{\frac{C_{\opt} \cdot d}{\gamma}} + \frac{1}{2 \gamma} \leq 3 \sqrt{\frac{C_{\opt} \cdot d}{\gamma}}
\end{align*}
whenever $\veps<1$. On the other hand, when $\eps=1$, this bound holds trivially.
\end{proof}

\subsubsection{\pfref{dm_linear:thm:greedy}}
\label{dm_linear:app:warmup_greedy}

\thmGreedy*

\begin{proof}[\pfref{dm_linear:thm:greedy}]
	Consider $\gamma \geq 1$.
    Combining \cref{dm_linear:prop:dec_greedy} with \cref{dm_linear:lm:regret_decomp}, we have 
   \begin{align*}
       \regcb(T) \leq 3T \cdot \sqrt{\frac{C_{\opt} \cdot d}{\gamma}}+ 2 \gamma \cdot \regsq(T) + 64 \gamma \cdot  \log (2\delta^{-1}) + \sqrt{8T \log(2 \delta^{-1})}.
   \end{align*} 
   The regret bound in \cref{dm_linear:thm:greedy} immediately follows by
   choosing \[\gamma = \prn*{\frac{3 T \sqrt{C_{\opt}\cdot d}}{2
     \regsq(T) + 64 \log(2\delta^{-1})}}^{2/3} \vee 1.\] In particular, when \cref{dm_linear:alg:barycentric} is
   invoked as a subroutine with parameter $C=2$,
   \cref{dm_linear:prop:barycentric_approx_opt_design} implies that we may take $C_{\opt}\leq{}4d$.

   \noindent\emph{Computational complexity.}
   We now bound the per-round computational complexity of
   \cref{dm_linear:alg:greedy} when \cref{dm_linear:alg:barycentric} is used as a
   subroutine to compute the approximate optimal design. Outside of
   the call to \cref{dm_linear:alg:barycentric}, \cref{dm_linear:alg:greedy} uses $O(1)$ calls
   to \sqalgtext to obtain $\wh g_t(x_t) \in \R^{d}$ and to update $\wh f_t$,
   and uses a single call to \optalgtext to compute $\wh a_t$. With the
   optimal design $\des_t$ returned by \cref{dm_linear:alg:barycentric}
   (represented as a barycentric spanner), sampling from $p_t$ 
   takes at most $O(d)$ time, since $\abs{\supp(p_t)} \leq d+1$.
outside of \cref{dm_linear:alg:barycentric} adds up to $O(\Tsq + \Topt +d)$. In terms of memory, calling \sqalgtext and \optalgtext takes $O(\Msq + \Mopt)$ units, and maintaining the distribution $p_t$ (the barycentric spanner) takes
   $O(d)$ units,
   so the maximum memory (outside of \cref{dm_linear:alg:barycentric}) is $O(\Msq + \Mopt + d)$.
   The stated results follow from combining the computational complexities analyzed in \cref{dm_linear:prop:barycentric_basic_computational}.
\end{proof}

\subsection{Proofs and Supporting Results for \cref{dm_linear:sec:minimax_statistical}}
\label{dm_linear:app:minimax_statistical}
In this section we provide supporting results concerning
\cref{dm_linear:alg:igw} (\cref{dm_linear:app:igw_support}), and then give the proof of \cref{dm_linear:thm:igw} (\cref{dm_linear:app:igw_thm}).

\subsubsection{Supporting Results}
\label{dm_linear:app:igw_support}

\begin{lemma}
    \label{dm_linear:prop:lambda_range}
    In \cref{dm_linear:alg:igw} (\cref{dm_linear:eq:igw_reweight}),
    there exists a unique choice of $\lambda > 0$ such that $ \sum_{a \in \cA}^{} p_t(a) = 1$, and its value lies in $[\frac{1}{2}, 1]$.
  \end{lemma}
\begin{proof}[\pfref{dm_linear:prop:lambda_range}]
Define $h(\lambda) \ldef \sum_{a \in \supp(q_t) }^{} \frac{q_t(a)}{\lambda + \eta \prn{\wh f_t(x_t, \wh a_t) - \wh f_t(x_t, a)}} $. We first notice that $h(\lambda)$ is continuous and strictly decreasing over  $(0, \infty)$. 
We further have 
\begin{align*}
	h ({1}/{2})
	& \geq \frac{q_t(\wh a_t)}{{1}/{2} + \eta \prn{\wh f_t(x_t, \wh a_t) - \wh f_t(x_t, \wh a_t)}} \geq \frac{1 /2}{1 /2} =1;
\end{align*}
and 
\begin{align*}
        h(1) \leq \sum_{a \in \supp(q_t) } q_t(a) = \frac{1}{2} + \frac{1}{2}\sum_{a \in \supp(\des_t)} \des_t(a) = 1.
\end{align*}
As a result, there exists a unique normalization constant $\lambda^{\star} \in [\frac{1}{2}, 1]$ such that $ h(\lambda^{\star}) = 1 $.
    \end{proof}
\propDecIGW* 
\begin{proof}[\pfref{dm_linear:prop:dec_igw}]
As in the proof of \cref{dm_linear:prop:dec_greedy}, we use the linear structure
of the action space to rewrite the \dectext
  $\dec_{\gamma}(\cF;\fhat,x)$ as
\begin{align*}
& \dec_{\gamma}  (\cG; \wh{g}, x)  \ldef \\
&\inf_{p \in \Delta(\cA)} \sup_{a^\star \in \cA} \sup_{g^\star \in \cG} \E_{a \sim p} \Big[\ang[\big]{\phi(x,a^\star) - \phi(x, a), g^\star(x)}  - {\gamma} \cdot \prn[\big]{\ang[\big]{\phi(x,a), g^\star(x) - \widehat g(x)}}^2 \bigg],
\end{align*}
Where $\ghat$ is such that $\fhat=f_{\ghat}$. We will bound the quantity above uniformly
  for all $x\in\cX$ and $\wh{g}:\cX\to\bbR^{d}$.

   Denote $\wh a \ldef \argmax_{a \in \cA} \ang[\big]{\phi(x,a), \wh g(x)}$, $a^\star \ldef \argmax_{a \in \cA} \ang[\big]{\phi(x,a), g^\star(x)}$
   and $\des \in \Delta(\cA)$ be a $C_{\opt}$-approximate optimal
   design with respect to the reweighted embedding $\wb
   \phi(x,\cdot)$). We use the setting $\eta = \frac{\gamma}{C_{\opt} \cdot d}$ throughout the proof.
   Recall that for the sampling distribution in \cref{dm_linear:alg:igw}, we set
   $q \ldef \frac{1}{2} \des + \frac{1}{2} \indic_{\wh a}$ and define
		\begin{align}
		    p(a) = \frac{q(a)}{\lambda + \frac{\gamma}{C_{\opt}\cdot d} \paren*{ \ang[\big]{\phi(x,\wh a) - \phi(x,a), \wh g(x)}}}, \label{dm_linear:eq:igw_reweight_linear}
		\end{align}
                where $\lambda \in [\frac{1}{2}, 1]$ is a
                normalization constant (cf. \cref{dm_linear:prop:lambda_range}).

We decompose the regret of the distribution $p$ in
                \cref{dm_linear:eq:igw_reweight_linear} as
\begin{align}
 &   \E_{a \sim p} \sq[\Big]{\ang[\big]{\phi(x,a^\star) - \phi(x, a), g^\star(x)}} \nonumber \\
 & = \E_{a \sim p} \sq[\Big]{\ang[\big]{\phi(x, \wh a) - \phi(x,a), \wh g(x)}} 
    + \E_{a \sim p} \sq[\Big]{\ang[\big]{\phi(x,a), \wh g(x) - g^\star(x)}} \nonumber\\ 
    & \quad +  \ang[\big]{\phi(x,a^\star),  g^\star(x) - \wh g(x)}
    + \ang[\big]{\phi(x,a^\star) - \phi(x,\wh a), \wh g(x)}.
    \label{dm_linear:eq:dec_igw_decompostion}
\end{align}
    Writing out the expectation, the first term in \cref{dm_linear:eq:dec_igw_decompostion} is upper bounded as follows.
   \begin{align*}
       & \E_{a \sim p} \sq[\Big]{\ang[\big]{\phi(x, \wh a) - \phi(x,a), \wh g(x)}} \\ 
& = \sum_{a \in \supp(\des) \cup \crl{\wh a}} p(a) \cdot  \ang[\big]{\phi(x, \wh a) - \phi(x,a), \wh g(x)} \\
       & < \sum_{a \in \supp(\des)} \frac{\des(a)/2}{\frac{\gamma}{C_{\opt}\cdot d} \paren*{ \ang[\big]{\phi(x,\wh a) - \phi(x,a), \wh g(x)}}}  \cdot \ang[\big]{\phi(x, \wh a) - \phi(x,a), \wh g(x)}\\ 
       & \leq \frac{C_{\opt}\cdot d}{ 2 \gamma}, 
   \end{align*}
   where we use that $\lambda > 0$ in the second inequality (with the
   convention that $\frac{0}{0} = 0$).
   
   The second term in \cref{dm_linear:eq:dec_igw_decompostion} can be upper
   bounded as in the proof of \cref{dm_linear:prop:dec_greedy}, by applying the
   AM-GM inequality:
   \begin{align*}
       \E_{a \sim p} \sq[\Big]{\ang[\big]{\phi(x,a), g^\star(x) - \wh g(x)}} 
       \leq \frac{\gamma}{2} \cdot \E_{a \sim p} \sq[\Big]{\prn[\big]{\ang[\big]{\phi(x,a), \wh g(x) - g^\star(x)}}^2 }
       + \frac{1}{2 \gamma}.
   \end{align*}
   
The third term in \cref{dm_linear:eq:dec_igw_decompostion} is the most
   involved. To begin, we define $V(p) \ldef \E_{a \sim p}
   \sq{\phi(x,a) \phi(x,a)^\top}$ and apply the following standard bound:
   \begin{align}
        \ang[\big]{\phi(x,a^\star), \wh g(x) - g^\star(x)} 
        & \leq \nrm[\big]{\phi(x,a^\star)}_{V(p)^{-1}} \cdot \nrm[\big]{g^\star(x) - \wh g(x)  }_{V(p)} \nonumber\\
        & \leq \frac{1}{2 \gamma} \cdot \nrm[\big]{\phi(x,a^\star)}^2_{V(p)^{-1}} + \frac{\gamma}{2} \cdot \nrm[\big]{g^\star(x) - \wh g(x)}^2_{V(p)} \nonumber \\
        & = \frac{1}{2 \gamma} \cdot \nrm[\big]{\phi(x,a^\star)}^2_{V(p)^{-1}} + \frac{\gamma}{2}\cdot \E_{a \sim p} \sq[\Big]{\prn[\big]{\phi(x,a), g^\star(x) - \wh g(x)}^2}, \label{dm_linear:eq:dec_igw_1}
   \end{align}
   where the second line follows from the AM-GM inequality. The second
   term in \cref{dm_linear:eq:dec_igw_1} matches the bound we desired, so it
   remains to bound the first term. Let $\dess$ be the following sub-probability measure:
   \begin{align*}
       \dess(a) \ldef
		\frac{\des(a)/2}{\lambda + \frac{\gamma}{C_{\opt}\cdot d} \paren*{ \ang[\big]{\phi(x,\wh a) - \phi(x,a), \wh g(x)}}},
   \end{align*}
   and let $V(\dess) \ldef \E_{a \sim \dess} \sq{\phi(x,a)
     \phi(x,a)^\top}$. We clearly have $V(p) \succeq V(\dess)$ from
   the definition of $p$ (cf. \cref{dm_linear:eq:igw_reweight_linear}). We
   observe that
   \begin{align*}
       V(\dess) & = \sum_{a \in \supp(\dess)} \dess(a) \phi(x,a) \phi(x,a)^\top \\
       & = \frac{1}{2} \cdot \sum_{a \in \supp(\des)} \des(a) \wb \phi(x,a) \wb \phi(x,a)^\top \cdot \frac{1+ \frac{\gamma}{C_{\opt}\cdot d} \paren*{ \ang[\big]{\phi(x,\wh a) - \phi(x,a), \wh g(x)}}}{\lambda + \frac{\gamma}{C_{\opt}\cdot d} \paren*{ \ang[\big]{\phi(x,\wh a) - \phi(x,a), \wh g(x)}}}\\
       & \succeq \frac{1}{2} \cdot \sum_{a \in \supp(\des)} \des(a) \wb \phi(x,a) \wb \phi(x,a)^\top \rdef \frac{1}{2} \wb V(\des), 
   \end{align*}
   where the last line uses that $\lambda \leq 1$. Since $\wb V(\des)$
   is positive-definite by construction, we have that $V(p)^{-1} \preceq V(\dess)^{-1} \preceq 2 \cdot \wb V(\des)^{-1}$. As a result, 
   \begin{align}
      \frac{1}{2 \gamma} \cdot \nrm[\big]{\phi(x,a^\star)}^2_{V(p)^{-1}} & \leq  
      \frac{1}{ \gamma} \cdot \nrm[\big]{\phi(x,a^\star)}^2_{\wb V(\des)^{-1}} \nonumber \\
      & = \frac{1 + \frac{\gamma}{ C_{\opt}\cdot d} \prn[\big]{\ang[\big]{\phi(x,\wh a) - \phi(x,a^\star), \wh g(x)}}}{\gamma} \cdot  \nrm[\big]{\wb \phi(x,a^\star)}^2_{\wb V(\des)^{-1}} \nonumber \\
      & \leq \frac{C_{\opt}\cdot d}{\gamma} + \ang[\big]{\phi(x,\wh a) - \phi(x,a^\star), \wh g(x)}, \label{dm_linear:eq:dec_igw_2}
   \end{align}
   where the last line uses that $\nrm[\big]{\wb
     \phi(x,a^\star)}^2_{\wb V(\des)^{-1}} \leq C_{\opt} \cdot d$,
   since $\des$ is a $C_{\opt}$-approximate optimal design for the set $\crl*{\wb \phi(x,a)}_{a\in\cA}$. Finally, we observe that the second term in \cref{dm_linear:eq:dec_igw_2} is cancelled out by the forth term in \cref{dm_linear:eq:dec_igw_decompostion}.
   
   Summarizing the bounds on the terms in
   \cref{dm_linear:eq:dec_igw_decompostion} leads to:
   \begin{align*}
       \dec_\gamma(\cF) = \inf_{p \in \Delta(\cA)} \sup_{a^\star \in \cA} \sup_{g^\star \in \cG}\dec_\gamma(\cG; \wh g, x) \leq \frac{C_{\opt}\cdot d}{2 \gamma} + \frac{1}{2 \gamma} + \frac{C_{\opt} \cdot d}{ \gamma } \leq \frac{2 \, C_{\opt} \cdot d}{ \gamma }.
   \end{align*}
\end{proof}

\subsubsection{\pfref{dm_linear:thm:igw}}
\label{dm_linear:app:igw_thm}
\thmIGW*
\begin{proof}
    Combining \cref{dm_linear:prop:dec_igw} with \cref{dm_linear:lm:regret_decomp}, we have 
    \begin{align*}
        \regcb(T) \leq 2T \cdot {\frac{C_{\opt} \cdot d}{\gamma}}+ 2 \gamma \cdot \regsq(T) + 64 \gamma \cdot \log (2\delta^{-1}) + \sqrt{8T \log(2 \delta^{-1})}.
    \end{align*} 
    The theorem follows by choosing 
    \begin{align*}
        \gamma = \prn*{ \frac{C_{\opt}\cdot d \, T}{\regsq(T) + 32 \log(2 \delta^{-1})}}^{1/2}.
    \end{align*}
    In particular, when \cref{dm_linear:alg:barycentric_reweight} is invoked
    as the subroutine with parameter $C=2$, we may take $C_{\opt} = 4
    d$.

   \noindent\emph{Computational complexity.}
    We now discuss the per-round computational complexity of
    \cref{dm_linear:alg:igw}. 
{We analyze a variant of the sampling rule specified in \cref{dm_linear:app:exp_modification} that does not require computation of the normalization constant.}
    Outside of the runtime and memory requirements
    required to compute the barycentric spanner using
    \cref{dm_linear:alg:barycentric_reweight}, which are stated in
    \cref{dm_linear:thm:barycentric_reweight}, \cref{dm_linear:alg:igw} uses $O(1)$ calls
    to the oracle \sqalgtext to obtain  $\wh g_t(x_t) \in \R^{d}$ and update $\wh f_t$,
    and uses a single call to \optalgtext to compute $\wh a_t$. 
With $\wh g_t(x_t)$ and $\wh a_t$, we can compute $\wh f_t(x_t, \wh a_t) - \wh f_t(x_t, a) = \ang{\phi(x_t,\wh a_t) - \phi(x_t,a), \wh g_t(x_t)}$ in $O(d)$ time for any $a \in \cA$; 
   thus, with the optimal design $\des_t$ returned by \cref{dm_linear:alg:barycentric_reweight}
   (represented as a barycentric spanner), 
we can construct the sampling distribution $p_t$ in  $O(d^2)$ time. 
Sampling from $p_t$ takes  $O(d)$ time since $\abs{\supp(p_t)} \leq d+1$. 
This adds up to runtime $O(\Tsq + \Topt + d^2)$.
   In terms of memory, calling \sqalgtext and \optalgtext takes $O(\Msq + \Mopt)$ units, and maintaining the distribution $p_t$ (the barycentric spanner) takes
   $O(d)$ units,
   so the maximum memory (outside of \cref{dm_linear:alg:barycentric_reweight}) is $O(\Msq + \Mopt + d)$.
   The stated results follow from combining the computational
   complexities analyzed in \cref{dm_linear:thm:barycentric_reweight}
, together with the choice of $\gamma$ described above.
\end{proof}

\subsection{Proofs and Supporting Results for \cref{dm_linear:sec:minimax_computational}}
\label{dm_linear:app:minimax_computational}
This section of the appendix is dedicated to the analysis of \cref{dm_linear:alg:barycentric_reweight}, and organized as follows.
\begin{itemize}
\item First, in \cref{dm_linear:app:argmax_reweight}, we analyze \cref{dm_linear:alg:igw_argmax}, a subroutine of \cref{dm_linear:alg:barycentric_reweight} which implements a linear optimization oracle for the reweighted action set used in the algorithm.
\item Next, in \cref{dm_linear:app:barycentric_reweight}, we prove \cref{dm_linear:thm:barycentric_reweight}, the main theorem concerning the performance of \cref{dm_linear:alg:barycentric_reweight}.
\item Finally, in \cref{dm_linear:app:det_assumption}, we discuss settings in which the initialization step required by \cref{dm_linear:alg:barycentric_reweight} can be performed efficiently.
\end{itemize}
Throughout this section of the appendix, we assume that the context $x\in\cX$ and estimator $\ghat:\cX\to\bbR^{d}$---which are arguments to \cref{dm_linear:alg:barycentric_reweight} and \cref{dm_linear:alg:igw_argmax}---are fixed.

\subsubsection{Analysis of \cref{dm_linear:alg:igw_argmax} (Linear Optimization Oracle for Reweighted Embeddings)}
\label{dm_linear:app:argmax_reweight}
A first step is to construct an (approximate) argmax oracle (after taking absolute value) with respect to the reweighted embedding $\wb \phi$.
Recall that the goal of \cref{dm_linear:alg:igw_argmax} is to implement a linear optimization oracle for the reweighted embeddings constructed by \cref{dm_linear:alg:barycentric_reweight}. That is, for any $\theta\in\bbR^{d}$,
we would like to compute an action that (approximately) solves
\begin{align*}
    \argmax_{a \in \cA} \abs[\big]{\ang[\big]{\wb \phi(x,a), \theta}} = 
    \argmax_{a \in \cA} {\ang[\big]{\wb \phi(x,a), \theta}}^2. 
\end{align*} 
Define
\begin{equation}
  \label{dm_linear:eq:iota}
  \iota(a) \ldef \ang{\wb \phi(x,a), \theta}^2,\quad\text{and}\quad
  a^\star \ldef \argmax_{a \in \cA} \iota(a).
\end{equation}
The main result of this section, \cref{dm_linear:prop:grid_search}, shows that \cref{dm_linear:alg:igw_argmax} identifies an action that achieves the maximum value in \cref{dm_linear:eq:iota} up to a multiplicative constant.

    \begin{theorem}
      \label{dm_linear:prop:grid_search}
      Fix any $\eta > {}0$, $r\in(0,1)$. Suppose $\zeta \leq \sqrt{\iota(a^{\star})} \leq 1$ for some $\zeta>0$. Then \cref{dm_linear:alg:igw_argmax} identifies an action $\check a$ such that $\sqrt{\iota(\check a)} \geq \frac{\sqrt{2}}{2}\cdot{}\sqrt{\iota(a^\star)}$, and does so with runtime $O(\prn{\Topt+d} \cdot \log (e \vee \frac{\eta}{\zeta}))$ and maximum memory $O(\Mopt + \log (e \vee \frac{\eta}{\zeta}) + d)$.
    \end{theorem}

    \begin{proof}[\pfref{dm_linear:prop:grid_search}]
      Recall from \cref{dm_linear:eq:reweighting} that we have
\begin{align*}
   {\ang[\big]{\wb \phi(x,a), \theta}}^2 = 
    \prn*{ \frac{{\ang[\big]{\phi(x,a), \theta}}}{\sqrt{1 + \eta \ang[\big]{\phi(x,\wh a) - \phi(x,a), \wh g(x)}}} }^2
    =  \frac{{\ang[\big]{\phi(x,a), \theta}}^2}{{1 + \eta \ang[\big]{\phi(x,\wh a) - \phi(x,a), \wh g(x)}}},
\end{align*}
where $\wh a \ldef \argmax_{a \in \cA} \tri*{\phi(x,a),\ghat(x)}$; note that the denominator is at least $1$. To proceed, we use that for any $X\in\bbR$ and $Y^2 > 0$, we have 
\begin{align*}
    \frac{X^2}{Y^2} = \sup_{\eps \in \R} \curly*{2 \eps X - \eps^2 Y^2}. 
\end{align*}
Taking $X = \ang[\big]{\phi(x,a), \theta}$ and $Y^2 = {1 + \eta \ang[\big]{\phi(x,\wh a) - \phi(x,a), \wh g(x)}}$ above, we can write
\begin{align}
     {\ang[\big]{\wb \phi(x,a), \theta}}^2 & = 
     \frac{{\ang[\big]{\phi(x,a), \theta}}^2}{{1 + \eta \ang[\big]{\phi(x,\wh a) - \phi(x,a), \wh g(x)}}} \nonumber \\
& = \sup_{\eps \in \bbR} \crl[\Big]{2 \eps \ang[\big]{\phi(x,a), \theta} - \eps^2 \cdot \prn[\big]{{1 + \eta \ang[\big]{\phi(x,\wh a) - \phi(x,a), \wh g(x)}}}} \label{dm_linear:eq:quotient_linear_o} \\
    & =  \sup_{\eps \in \bbR}\crl[\Big]{\ang[\big]{\phi(x,a), 2 \eps \theta + \eta \eps^2 \wh g(x)} - \eps^2 - \eta \eps^2 \ang[\big]{\phi(x,\wh a), \wh g(x)}} \label{dm_linear:eq:quotient_linear}.
\end{align}
The key property of this representation is that for any \emph{fixed} $\eps \in \bbR$, \cref{dm_linear:eq:quotient_linear} is a linear function of the \emph{unweighted} embedding $\phi$, and hence can be optimized using \optalgtext.
In particular, for any fixed $\eps \in \bbR$, consider the following linear optimization problem, which can be solved by calling \optalgtext:
\begin{align}
     \argmax_{a \in \cA} \crl[\Big]{2 \eps \ang[\big]{\phi(x,a), \theta} - \eps^2 \cdot \prn[\big]{{1 + \eta \ang[\big]{\phi(x,\wh a) - \phi(x,a), \wh g(x)}}}} \rdef \argmax_{a\in \cA} W(a;\eps)\label{dm_linear:eq:lin_opt_fix_eps}.
\end{align}
Define
\begin{align}
\label{dm_linear:eq:opt_eps}
    \eps^\star \ldef \frac{\ang{\phi(x,a^\star), \theta}}{{1 + \eta \ang{\phi(x,\wh a) - \phi(x,a^\star), \wh g(x)}}}.
\end{align}
If $\eps^{\star}$ was known (which is not the case, since $\astar$ is unknown), we could set $\eps = \eps^\star$ in \cref{dm_linear:eq:lin_opt_fix_eps} and compute an action $\wb a \ldef \argmax_{a \in \cA} W(a; \eps^{\star})$ using a single oracle call. We would then have $\iota(\wb a) \geq W(\wb a; \eps^{\star}) \geq W( a^{\star}; \eps^{\star}) = \iota(a^{\star})$, which follows because $\veps^{\star}$ is the maximizer in \cref{dm_linear:eq:quotient_linear_o} for $a=\astar$.

To get around the fact that $\veps^{\star}$ is unknown, \cref{dm_linear:alg:igw_argmax} performs a grid search over possible values of $\veps$. To show that the procedure succeeds, we begin by bounding the range of $\eps^\star$. With some rewriting, we have 
\begin{align*}
    \abs{\eps^\star} = \frac{\sqrt{\iota(a^\star)}}{\sqrt{{1 + \eta \ang{\phi(x,\wh a) - \phi(x,a^\star), \wh g(x)}}}}.
\end{align*}
Since $0 < \zeta \leq \sqrt{\iota(a^{\star})} \leq 1$, we have
        \begin{align*}
             \wb \zeta \ldef
            {\frac{\zeta}{\sqrt{1+2\eta}}} \leq  \abs{\eps^\star}  \leq 1.
        \end{align*}
\cref{dm_linear:alg:igw_argmax} performs a $(3/4)$-multiplicative grid search over the intervals $[\wb \zeta,1]$ and $[-1,- \wb\zeta]$, which uses $2\ceil{\log_{\frac{4}{3}} \prn{\wb \zeta^{-1}}} = O(\log (e \vee \frac{\eta}{\zeta}))$ grid points. It is immediate to that the grid contains $\wb \eps\in\bbR$ such that $\wb \eps \cdot \eps^{\star} > 0$ and
        $\frac{3}{4} \abs*{\eps^\star} \leq \abs*{  \wb \eps} \leq \abs*{ \eps^\star}$. Invoking \cref{dm_linear:lm:reweighted_opt_approx} (stated and proven in the sequel) with $\bar{a} \ldef \argmax_{a\in\cA}W(a; \wb \veps)$ implies that $\iota(\wb a) \geq \frac{1}{2} \iota(a^\star)$. To conclude, recall that \cref{dm_linear:alg:igw_argmax} outputs the maximizer
        \[
          \acheck \ldef \argmax_{a\in\wh{\cA}}\iota(a),
        \]
        where $\wh{\cA}$ is the set of argmax actions encountered by the grid search. Since $\bar{a}\in\wh{\cA}$, we have $\iota(\check a) \geq \iota (\wb a) \geq \frac{1}{2}\iota(a^\star)$ as desired.

	\noindent \emph{Computational complexity.}
        Finally, we bound the computational complexity of \cref{dm_linear:alg:igw_argmax}. \cref{dm_linear:alg:igw_argmax} maintains a grid of $O(\log (e \vee \frac{\eta}{\zeta}))$ points, and hence calls the oracle \optalgtext $O(\log (e \vee \frac{\eta}{\zeta}))$ in total; this takes $O(\Topt \cdot \log (e \vee \frac{\eta}{\zeta}) )$ time.
	Computing the final maximizer from the set $\wh{\cA}$, which contains $O(\log (e \vee \frac{\eta}{\zeta}))$ actions, takes $O(d \log (e \vee \frac{\eta}{\zeta}))$ time (compute each $\ang{\wb \phi(x,a), \wb \theta}^2$ takes $O(d)$ time). Hence, the total runtime of \cref{dm_linear:alg:igw_argmax} adds up to $O(\prn{\Topt+d} \cdot \log (e \vee \frac{\eta}{\zeta}))$.
	The maximum memory requirement is $O(\Mopt + \log (e \vee \frac{\eta}{\zeta}) + d)$, follows from calling \optalgtext, and storing $\cE, \wh \cA$ and other terms such as $\wh g(x), \theta, \wb \theta, \phi(x,a), \wb \phi(x,a)$.
    \end{proof}

\paragraph{Supporting Results}
\noindent
\begin{lemma}
    \label{dm_linear:lm:reweighted_opt_approx}
    Let $\eps^{\star}$ be defined as in \cref{dm_linear:eq:opt_eps}.
Suppose $\wb \eps \in \R$ has $  \wb \eps\cdot \eps^\star > 0$ and $\frac{3}{4} \abs*{\eps^\star} \leq \abs*{  \wb \eps} \leq \abs*{ \eps^\star} $. 
Then, if $  \wb a \ldef \argmax_{a\in\cA}W(a;\wb{\veps})$, we have $\iota(\wb  a) \geq \frac{1}{2} \iota(a^\star)$.
\end{lemma}

\begin{proof}[\pfref{dm_linear:lm:reweighted_opt_approx}]
  First observe that using the definition of $\iota(a)$, along with \cref{dm_linear:eq:quotient_linear_o} and \cref{dm_linear:eq:lin_opt_fix_eps}, we have $\iota(\wb a) \geq W(\wb a; \wb \eps) \geq W(a^\star; \wb \eps)$,
  where the second inequality uses that $\abar\ldef \argmax_{a\in\cA}W(a;\vepsbar)$. Since $\wb \eps \cdot \eps^\star > 0$, we have $\sgn(\wb \eps \cdot \ang{\phi(x,a^\star), \theta}) = \sgn( \eps^\star \cdot \ang{\phi(x,a^\star), \theta})$.
  If $\sgn(\wb \eps \cdot \ang{\phi(x,a^\star), \theta}) \geq 0$, then since $\frac{3}{4} \abs*{\eps^\star} \leq \abs*{  \wb \eps} \leq \abs*{ \eps^\star} $, we have  
    \begin{align*}
      W(\astar;\vepsbar)&=2 \wb \eps \cdot \ang[\big]{\phi(x,a^\star), \theta} - \wb \eps^2 \cdot \prn[\big]{{1 + \eta \ang[\big]{\phi(x,\wh a) - \phi(x,a^\star), \wh g(x)}}}\\
      &\geq 
     \frac{3}{2} \eps^\star \cdot \ang[\big]{\phi(x,a^\star), \theta} - (\eps^\star)^2 \cdot \prn[\big]{{1 + \eta \ang[\big]{\phi(x,\wh a) - \phi(x,a^\star), \wh g(x)}}} \\
      &= \frac{1}{2} \frac{\ang{\phi(x,a^\star), \theta}^2}{{1 + \eta \ang{\phi(x,\wh a) - \phi(x,a^\star), \wh g(x)}}} 
      = \frac{1}{2} \iota(a^\star),
    \end{align*}
    where we use that ${{1 + \eta \ang[\big]{\phi(x,\wh a) - \phi(x,a^\star), \wh g(x)}}} \geq 1$ for the first inequality and use the definition of $\eps^\star$ for the second equality.

    On the other hand, when $\sgn(\wb \eps \cdot \ang{\phi(x,a^\star), \theta}) < 0$, we similarly have  
    \begin{align*}
      W(\astar;\vepsbar)&=2 \wb \eps \cdot \ang[\big]{\phi(x,a^\star), \theta} - \wb \eps^2 \cdot \prn[\big]{{1 + \eta \ang[\big]{\phi(x,\wh a) - \phi(x,a^\star), \wh g(x)}}}\\
      & \geq 
     2\eps^\star  \cdot\ang[\big]{\phi(x,a^\star), \theta} - (\eps^\star)^2 \cdot \prn[\big]{{1 + \eta \ang[\big]{\phi(x,\wh a) - \phi(x,a^\star), \wh g(x)}}} 
       = \iota(a^\star).
    \end{align*}
   Summarizing both cases, we have $\iota(\bar a) \geq \frac{1}{2}\iota(a^\star)$.
    \end{proof}

\subsubsection{\pfref{dm_linear:thm:barycentric_reweight}}
\label{dm_linear:app:barycentric_reweight}

\thmBarycentricReweight*

\begin{proof}[\pfref{dm_linear:thm:barycentric_reweight}]
We begin by examining the range of $\sqrt{\iota(a^{\star})}$ used in \cref{dm_linear:prop:grid_search}.
Note that the linear function $\theta$ passed as an argument to \cref{dm_linear:alg:barycentric_reweight} takes the form $\wb \phi(x,a) \mapsto \det(\wb \phi(x,\cS_i(a)))$, i.e., ${\ang{\wb \phi(x,a), \theta}} =  {\det(\wb \phi(x,\cS_i(a)))}$, where $\cS_i(a) \ldef \prn{a_1, \ldots, a_{i-1}, a, a_{i+1}, \ldots , a_d}$. 
For the upper bound, we have 
\begin{align*}
\abs{\ang{\wb \phi(x,a^{\star}), \theta}} = \abs{\det(\wb \phi(x,\cS_i(a^{\star})))} \leq \prod_{a \in \cS_i(a^{\star})}^{}  \nrm{\wb \phi(x,a)}_2^d \leq \sup_{a \in \cA} \nrm{\phi(x,a)}_2^d \leq 1
\end{align*}
by Hadamard's inequality and the fact that the reweighting appearing in \cref{dm_linear:eq:reweighting} enjoys $\nrm{\wb \phi(x,a)}_2 \leq \nrm{\phi(x,a)}_2$. This shows that $\sqrt{\iota(a^\star)} \leq 1$.
For the lower bound, we first recall that in \cref{dm_linear:alg:barycentric_reweight}, the set $\cS$ is initialized to have $\abs{\det(\phi(x, \cS))} \geq r^{d}$, and thus $\abs{\det(\wb \phi(x, \cS))} \geq \wb r^{d}$, where $\wb r \ldef \frac{r}{\sqrt{1 + 2 \eta}}$ accounts for the reweighting in \cref{dm_linear:eq:reweighting}. Next, we observe that as a consequence of the update rule in \cref{dm_linear:alg:barycentric_reweight}, we are guaranteed that $\abs{\det(\wb \phi(x,\cS))} \geq \wb r^{d}$ across all rounds.
Thus, whenever \cref{dm_linear:alg:igw_argmax} is invoked with the linear function $\theta$ described above, there must exist an action $a\in\cA$ such that $\abs{\ang{\wb \phi(x,a), \theta}} \geq \wb r^{d}$, which implies that $\sqrt{\iota(a^\star)} \geq \wb r^{d}$ and we can take $\zeta \ldef \wb r^{d}$ in \cref{dm_linear:prop:grid_search}.

  We next bound the number of iterations of the while-loop before the algorithm terminates. Let $\wb C \ldef \frac{\sqrt{2}}{2} \cdot C > 1$.  At each iteration (beginning from line 3) of \cref{dm_linear:alg:barycentric_reweight}, one of two outcomes occurs:
    \begin{enumerate}
        \item We find an index $ i \in [d]$ and an action $a \in \cA$ such that $\abs{\det(\wb \phi(x,\cS_i(a)))} > \wb C \abs{\det(\wb \phi(x, \cS))}$, and update $a_i = a$.
\item We conclude that $\sup_{a \in \cA} \max_{i \in [d]} \abs{\det \prn{\wb \phi(x,\cS_i(a))}} \leq C \abs{\det \prn{\wb \phi(x,\cS)}}$ and terminate the algorithm.
    \end{enumerate}
    We observe that 
    (i) the initial set $\cS$ has $\abs{\det(\wb \phi(x, \cS))} \geq \wb r^d$ with $\wb r \ldef \frac{r}{\sqrt{1+2\eta}}$ (as discussed before), 
    (ii) $\sup_{\cS \subseteq \cA, \abs{\cS} = d} \abs{\det(\wb \phi(x,\cS))} \leq 1$ by Hadamard's inequality,
    and (iii)
    each update of $\cS$ increases the (absolute) determinant by a factor of $\wb C$. 
    Thus, fix any $C > \sqrt{2}$, we are guaranteed that \cref{dm_linear:alg:barycentric_reweight} terminates within $O(d \log \prn{ e \vee \frac{\eta}{r}})$ iterations of the while-loop.

    We now discuss the correctness of \cref{dm_linear:alg:barycentric_reweight}, i.e., when terminated, the set $\cS$ is a $C$-approximate barycentric spanner with respect to the reweighted embedding $\wb \phi$.
First, note that by \cref{dm_linear:prop:grid_search}, \cref{dm_linear:alg:igw_argmax} is guaranteed to identify an action $\check a\in\cA$ such that $\abs{\det(\wb \phi(x,\cS_i(\check a)))} > \wb C \abs{\det(\wb \phi(x,\cS))}$ as long as there exists an action $a^\star\in\cA$ such that $\abs{\det(\wb \phi(x, \cS_i(a^\star)))} > C \abs{\det(\wb \phi(x,\cS))}$. As a result, by Observation 2.3 in \citet{awerbuch2008online}, if no update is made and \cref{dm_linear:alg:barycentric_reweight} terminates, we have identified a $C$-approximate barycentric spanner with respect to embedding $\wb \phi$.

\noindent\emph{Computational complexity.}
We provide the computational complexity analysis for \cref{dm_linear:alg:barycentric_reweight} in the following. 
We use $\wb \Phi_\cS$ to denote the matrix whose  $k$-th column is  $\wb \phi(x,a_k)$ with  $a_k \in \cS$.
\begin{itemize}
	\item \emph{Initialization.}
We first notice that, given $\wh g(x) \in \R^d$ and  \linebreak
$\wh a \ldef \argmax_{a \in \cA} \ang{\phi(x,a), \wh g(x)}$, it takes $O(d)$ time to compute  $\wb \phi(x,a)$ for any  $a \in \cA$.
Thus, computing  $\det(\wb \Phi_\cS)$ and  $\wb \Phi_\cS^{-1}$ takes $O(d^2 + d^{\omega}) = O(d^{\omega})$ time, where we use $O(d^{\omega})$ (with $2 \leq \omega \leq 3$) to denote the time of computing matrix determinant/inversion.
The maximum memory requirement is  $O(d^2)$, following from the storage of $\crl{\wb \phi(x,a)}_{a \in \cS}$ and  $\wb \Phi_\cS^{-1}$.
	\item \emph{Outer loops (lines 1-2).}
We have already shown that \cref{dm_linear:alg:barycentric} terminates within $O(d \log \prn{e \vee \frac{\eta}{r}})$  iterations of the while-loop (line 2). 
	It is also clear that the for-loop (line 2) is invoked at most $d$ times.
	 \item \emph{Computational complexity for lines 3-7.}
		 We discuss how to efficiently implement this part using rank-one updates.
We analyze the computational complexity for each line in the following.
The analysis largely follows from the proof of \cref{dm_linear:prop:barycentric_basic_computational}.
\begin{itemize}
	\item \emph{Line 3.}
Using rank-one update of the matrix determinant (as discussed in the proof of \cref{dm_linear:prop:barycentric_basic_computational}), we have  
   \begin{align*}
 \det( \wb \phi(x,a_1),\ldots, \wb \phi(x,a_{i-1}), Y, \wb \phi(x,a_{i+1}),\ldots, \wb \phi(x, a_d) ) = \ang{Y, \theta}, 
   \end{align*}
   where $\theta = \det(\wb \Phi_\cS) \cdot \prn{\wb \Phi_\cS^{-1}}^\top e_i$.
   Thus, whenever $\det(\wb \Phi_\cS)$ and  $\wb \Phi_\cS^{-1}$ are known, compute $\theta$ takes  $O(d)$ time. 
   The maximum memory requirement is $O(d^2)$, following from the storage of $\wb \Phi_\cS^{-1}$.
   \item \emph{Line 4.} When $\theta$ is computed, we can compute  $a$ by 
invoking \argmaxIGW (\cref{dm_linear:alg:igw_argmax}). As discussed in \cref{dm_linear:prop:grid_search}, this step takes runtime $O(\prn{\Topt \cdot d +d^2} \cdot \log (e \vee \frac{\eta}{r}))$ and maximum memory $O(\Mopt + d \log (e \vee \frac{\eta}{r}) + d)$ (by taking $\zeta = \wb r^{d}$ as discussed before).
\item \emph{Line 5.}
	Once $\theta$ and $\det(\wb \Phi_\cS)$ are computed, checking the updating criteria takes  $O(d)$ time. The maximum memory requirement is $O(d)$, following from the storage of $\wb \phi(x, a)$ and $\theta$.
 \item \emph{Line 6.}
	 As discussed in the proof of \cref{dm_linear:prop:barycentric_basic_computational}, if an update $a_i = a$ is made, we can update $\det(\wb \Phi_\cS)$ and $\wb \Phi_\cS^{-1}$ using rank-one updates with $O(d^{2})$ time and memory.
\end{itemize}
Thus, using rank-one updates, the total runtime for line 3-7 adds up to $O(\prn{\Topt \cdot d +d^2} \cdot \log (e \vee \frac{\eta}{r}))$ and maximum memory requirement is $O(\Mopt + d^2+ d \log (e \vee \frac{\eta}{r}) )$.
\end{itemize}
   To summarize,
   \cref{dm_linear:alg:barycentric} has runtime $O(\prn{\Topt \cdot d^{3} +d^4} \cdot \log^{2} (e \vee \frac{\eta}{r}))$
   and uses at most $O(\Mopt + d^2 + d \log (e \vee \frac{\eta}{r}) )$ units of memory. 
    \end{proof}

    \subsubsection{Efficient Initializations for \cref{dm_linear:alg:barycentric_reweight}}
    \label{dm_linear:app:det_assumption}
    
    In this section we discuss specific settings in which the initialization required by \cref{dm_linear:alg:barycentric_reweight} can be computed efficiently. For the first result, we let $\ball(0,r)\ldef\crl*{x\in\bbR^{d}\mid{}\nrm*{x}_2\leq{}r}$ denote the ball of radius $r$ in $\bbR^{d}$.
\begin{example}
  \label{dm_linear:ex:init1}
  Suppose that there exists $r \in (0,1)$ such that $\ball(0, r) \subseteq \crl{\phi(x,a): a \in \cA}$. Then by choosing $\cS \ldef \crl{r e_1, \dots, r e_d} \subseteq \cA$, we have $\abs{\det \prn{\phi(\cS)}} = r^d$.
\end{example}

    The next example is stronger, and shows that we can efficiently compute a set with large determinant whenever such a set exists.

    \begin{example}
      \label{dm_linear:prop:determinant_initialization}
        Suppose there exists a set $\cS^\star \subseteq \cA$ such that $\abs{\det(\phi(\cS^\star))} \geq \wb r^d$ for some $\wb r > 0$. Then there exists an efficient algorithm that identifies a set $\cS \subseteq \cA$ with $\abs{\det(\phi(\cS))} \geq r^d$ for $r \ldef \frac{\wb r}{8d}$, and does so with 
   runtime $O(\Topt \cdot d^2 \log d + d^4 \log d)$ and memory $O(\Mopt + d^2)$.
      \end{example}
\begin{proof}[Proof for \cref{dm_linear:prop:determinant_initialization}]
The guarantee is achieved by running \cref{dm_linear:alg:barycentric} with $C=2$.
One can show that this strategy achieves the desired approximation guarantee by slightly generalizing the proof of a similar result in \citet{mahabadi2019composable}. In more detail, \citet{mahabadi2019composable} study the problem of identifying a subset $\cS\subseteq\cA$ such that $\abs*{\cS} = k$ and $\det(\Phi_{\cS}^\top \Phi_{\cS})$ is (approximately) maximized, where $\Phi_\cS \in \R^{d \times \abs{\cS}}$ denotes the matrix whose columns are $\phi(x,a)$ for  $a \in \cS$.
We consider the case when $k=d$, and
make the following observations.
 \begin{itemize}
	\item We have 
            $\det(\Phi_{\cS}^\top \Phi_{\cS}) = \paren*{\det(\Phi_{\cS})}^2 = \paren{\det(\phi(x, \cS))}^2$. Thus, maximizing $\det(\Phi_{\cS}^\top \Phi_{\cS})$ is equivalent to maximizing $\abs{\det(\phi(x,\cS))}$.
	 \item 
           The Local Search Algorithm provided in \citet{mahabadi2019composable} (Algorithm 4.1 therein) has the same update and termination condition as \cref{dm_linear:alg:barycentric}. As a result, one can show that the conclusion of their Lemma 4.1 also applies to \cref{dm_linear:alg:barycentric}.
\end{itemize}
    \end{proof}

\subsection{Other Details for Experiments}
\label{dm_linear:app:experiment}
\subsubsection{Basic Details}
\label{dm_linear:app:exp_basic}

\paragraph{Datasets}

\oneshotwiki \citep{singh12:wiki-links,oneshotwiki} is a
named-entity recognition task where contexts are text phrases
preceding and following the mention text, and where actions are text
phrases corresponding to the concept names.  We use the python
package \sentencetransformers \citep{reimers-2019-sentence-bert} to
separately embed the text preceding and following the reference into $\R^{768}$,
and then concatenate, resulting in a context embedding in $\R^{1536}$.
We embed the action (mentioned entity) text into $\R^{768}$ and then use
SVD on the collection of embedded actions to reduce the dimensionality
to $\R^{50}$.  The reward function is an indicator function for
whether the action corresponds to the actual entity
mentioned.  \oneshotwiki-311 (resp. \oneshotwiki-14031) is a subset of
this dataset obtained by taking all actions with at least 2000
(resp. 200) examples. 

\amazon-3m \citep{Bhatia16} is an extreme multi-label dataset whose
contexts are text phrases corresponding to the title and description
of an item, and whose actions are integers corresponding to item tags.
We separately embed the title and description phrases using \sentencetransformers,
which leads to a context embedding in $\R^{1536}$.
Following the protocol
used in \citet{sen2021top}, the first 50000 examples are fully supervised,
and subsequent examples have bandit feedback.  We use Hellinger
PCA~\citep{lebret2014word} on the supervised data label cooccurrences to
construct the action embeddings {in $\R^{800}$}. Rewards are binary, and indicate
whether a given item has the chosen tag. Actions that do not occur in the supervised
portion of the dataset cannot be output by the model, but are retained
for evaluation: For example, if during the bandit feedback phase, an example
consists solely of tags that did not occur during the supervised phase,
the algorithm will experience a reward of 0 for every feasible action
on the example. For a typical seed, this results in roughly 890,000 feasible actions for the model. In the $(k=5, r=3)$ setup, we take
the top-$k$ actions as the greedy slate, and then independently
decide whether to explore for each exploration slot {(the bottom $r$ slots)}. For exploration, we
sample from the spanner set without replacement.

\paragraph{Regression functions and oracles}
For bilinear models, regression functions take the form $f(x, a) = \ang{\phi(a), Wx}$, 
where $W$ is a matrix of learned parameters.  For deep
models, regression functions pass the original context through 2 residual
leaky \textsf{ReLU} layers before applying the bilinear layer, $f(x, a) =
\ang{\phi(a), W \wb g(x)}$, where $\wb g$ {is a learned two-layer neural network, and $W$ is a matrix of learned parameters}.
{For experiments with respect to \oneshotwiki datasets, we add a learned bias term for regression functions (same for every action); for experiments with respect to the \amazon-\textsf{3m} dataset, we additionally add an action-dependent bias term that is obtained from the supervised examples.}
The online regression oracle is implemented using \textsf{PyTorch}'s Adam optimizer with log loss (recall that rewards are 0/1).

\paragraph{Hyperparameters}
For each algorithm, we optimize its hyperparameters using random search \citep{bergstra2012random}. Speccifically, hyperparameters are tuned by taking the best of 59 randomly selected configurations for a fixed seed (this seed is not used for evaluation).
A seed determines both dataset shuffling, initialization of
regressor parameters, and random choices made by any action
sampling scheme. 

\paragraph{Evaluation}
We evaluate each algorithm on 32 seeds.
All reported confidence intervals are 90\% bootstrap CIs for the mean.

\subsubsection{Practical Modification to Sampling Procedure in \spannerIGW}
\label{dm_linear:app:exp_modification}
For experiments with \spannerIGW, we slightly modify the action
sampling distribution so as to avoid computing the normalization
constant $\lambda$. First, we modify the weighted embedding scheme
given in \cref{dm_linear:eq:reweighting} using the following expression:
\begin{align*}
    \wb{\phi}(x_t, a) \ldef \frac{\phi(x_t,a)}{\sqrt{1 + d + \frac{\gamma}{4d} \prn[\big]{ \wh f_t(x_t,\wh a_t) - \wh f_t(x_t, a)}}}. 
\end{align*}
We obtain a $4d$-approximate optimal design for the reweighted
embeddings by first computing a $2$-approximate barycentric spanner
$\cS$, then taking $q_t^{\mathrm{opt}} \ldef \unif(\cS)$. To proceed, let $\wh a_t \ldef \argmax_{a \in \cA} \wh f(x_t,a)$ and $\wb d \ldef
\abs{\cS \cup \crl{\wh a_t}}$. 
We construct the sampling distribution $p_t \in \Delta(\cA)$ as follows:
\begin{itemize}
    \item Set $p_t(a) \ldef \frac{1}{\wb d + \frac{\gamma}{4d}
        \prn[\big]{\wh f_t(x_t,\wh a_t) - \wh f_t(x_t, a)}}$ for each
      $a \in \supp(\cS)$.
\item Assign remaining probability mass to $\wh a_t$.
\end{itemize}
With a small modification to the proof of \cref{dm_linear:prop:dec_igw}, one
can show that this construction certifies that $\dec_\gamma(\cF) = 
\bigoh\prn[\big]{\frac{d^2}{\gamma}}$. Thus, the regret bound
in \cref{dm_linear:thm:igw} holds up to a constant factor.
{Similarly, with a small modification to the proof of \cref{dm_linear:thm:barycentric_reweight}, we can also show that
---with respect to this new embedding---\cref{dm_linear:alg:barycentric_reweight} has $O(\prn{\Topt \cdot d^{3} +d^4} \cdot \log^{2} \prn[\big]{ \frac{d + \gamma /d}{r}})$ runtime and $O(\Mopt + d^2 + d \log \prn[\big]{\frac{d + \gamma /d}{r}} )$ memory.}

\subsubsection{Timing Information}
\label{dm_linear:app:exp_timing}

\begin{table}[H]
    \centering
    \caption{Per-example inference timings for \oneshotwiki-14031.
      CPU timings use batch size $1$ on an Azure
      \textsf{STANDARD\_D4\_V2} machine. GPU timings use batch size
      1024 on an Azure \textsf{STANDARD\_NC6S\_V2} (Nvidia P100-based)
      machine.}
    \label{dm_linear:tab:timings}
    \begin{tabular}{c c c}
    \toprule
         Algorithm & CPU & GPU \\
         \midrule
         \greedy & 2 ms & 10 $\mu$s \\
         \greedyalg & 2 ms & 10 $\mu$s \\
         \squarecb & 2 ms & 10 $\mu$s\\
         \mainalg & 25 ms & 180 $\mu$s \\
    \bottomrule
    \end{tabular}
  \end{table}

\cref{dm_linear:tab:timings} contains timing information the
$\oneshotwiki$-14031 dataset with a bilinear model.  The CPU timings
are most relevant for practical scenarios such as information
retrieval and recommendation systems, while the GPU timings are
relevant for scenarios where simulation is possible.  Timings for
\greedyalg do not include the one-time cost to compute the spanner
set.  Timings for all algorithms use precomputed context and action
embeddings.  For all but algorithms but \mainalg, timings reflect the
major bottleneck of computing the argmax action, since all subsequent
steps take $O(1)$ time with respect to $\abs{\cA}$.  In particular,
\squarecb is implemented using rejection sampling, which does not
require explicit construction of the action distribution.  For
\mainalg, the additional overhead is due to the time required to construct an approximate optimal design for each example.

  \subsubsection{Additional Figures}
  \label{dm_linear:app:exp_figure}

  \begin{figure}[H]
\centering\includegraphics[width=.8\textwidth]{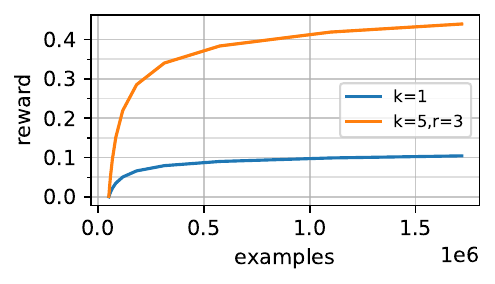}
\caption{Performance of \greedyalg on \amazon-3m.}
\label{dm_linear:fig:amazon3m}
\end{figure}

In \cref{dm_linear:fig:amazon3m}, we show the empirical performance of \greedyalg on \amazon-3m.
Confidence intervals are
  rendered, but are but too small to visualize. For $(k=1)$, the final CI is
$[0.1041, 0.1046]$, and for $(k=5,r=3)$, the final CI is $[0.438, 0.440]$.

\chapter{Contextual Bandits with Smooth Regret}
\label{chapter:large:smooth}
	Designing efficient general-purpose contextual bandit algorithms that work with large---or even continuous---action spaces would facilitate application to important scenarios such as information retrieval, recommendation systems, and continuous control. While obtaining standard regret guarantees can be hopeless, alternative regret notions have been proposed to tackle the large action setting. We propose a smooth regret notion for contextual bandits, which dominates previously proposed alternatives. We design a statistically and computationally efficient algorithm---for the proposed smooth regret---that works with general function approximation under standard supervised oracles. We also present an adaptive algorithm that automatically adapts to any smoothness level. Our algorithms can be used to recover the previous minimax/Pareto optimal guarantees under the standard regret, e.g., in bandit problems with multiple best arms and Lipschitz/H{\"o}lder bandits. We conduct large-scale empirical evaluations demonstrating the efficacy of our proposed algorithms.

\section{Introduction}

Contextual bandits concern the problem of sequential decision making with contextual information.
Provably efficient contextual bandit algorithms have been proposed over the past decade \citep{langford2007epoch,agarwal2014taming,foster2020beyond,simchi2021bypassing,foster2021efficient}.
However, these developments only work in setting with a small number of actions,
and their theoretical guarantees become vacuous when working with a large action space \citep{agarwal2012contextual}. 
The hardness result can be intuitively understood through a ``needle in the haystack'' construction: 
When good actions are extremely rare, identifying any good action demands trying almost all alternatives.
This prevents naive direct application of contextual bandit algorithms to large action problems, e.g., in information retrieval, recommendation systems, and continuous control.

To bypass the hardness result, one approach is to assume structure on the model class.
For example, in the standard linear contextual bandit \citep{auer2002using, chu2011contextual, abbasi2011improved}, learning the $d$ components of the reward vector---rather than examining every single action---effectively guides the learner to the optimal action.
Additional structural assumptions have been studied in the literature, e.g., linearly structured actions and general function approximation \citep{foster2020adapting, xu2020upper}, Lipschitz/H\"older regression functions \citep{kleinberg2004nearly, hadiji2019polynomial}, and convex functions \citep{lattimore2020improved}.
While these assumptions are fruitful theoretically, they might be violated in practice.

An alternative approach is to compete against a less demanding benchmark.
Rather than competing against a policy that always plays the best action, one can compete against a policy that plays the best smoothed distribution over the actions: a smoothed distribution---by definition---cannot concentrate on the best actions when they are in fact rare.  Thus, for the previously mentioned ``needle in the haystack'' construction, the benchmark is weak as well.  This de-emphasizes such constructions and focuses algorithm design on scenarios where intuition suggests good solutions can be found without prohibitive statistical cost.

\paragraph{Contributions}
We study large action space problems under an alternate notion of regret.
Our first contribution is to propose a novel benchmark---the \smth regret---that formalizes the ``no needle in the haystack'' principle.  
We also show that our \smth regret dominates previously proposed regret notions along this line of work \citep{chaudhuri2018quantile, krishnamurthy2020contextual, majzoubi2020efficient}, i.e., any regret guarantees with respect to the \smth regret automatically holds for these previously proposed regrets.

We design efficient algorithms that work with the \smth regret and general function classes. 
Our first proposed algorithm, \smthigw, works with any fixed smoothness level $h>0$, and is efficient---both statistically and computationally---whenever the learner has access to standard oracles: (i) an online regression oracle for supervised learning, and (ii) a simple sampling oracle over the action space.
Statistically, \smthigw achieves $\sqrt{T /h}$-type regret for whatever action spaces; here ${1}/{h}$ should be viewed as the effective number of actions. Such guarantees can be verified to be minimax optimal when related back to the standard regret.
Computationally, the guarantee is achieved with $O(1)$ operations with respect to oracles, which can be usually efficiently implemented in practice.
Our second algorithm is a master algorithm which combines multiple \smthigw instances to compete against any unknown smoothness level. We show this master algorithm is Pareto optimal.

With our \smth regret and proposed algorithms, we exhibit guarantees under the standard regret in various scenarios, e.g., in problems with multiple best actions \citep{zhu2020regret} and in problems when the expected payoff function satisfies structural assumptions such as Lipchitz/H\"{o}lder continuity \citep{kleinberg2004nearly, hadiji2019polynomial}. Our algorithms are minimax/Pareto optimal when specialized to these settings.

\subsection{Organization}
We introduce our smooth regret in \cref{dm_smooth:sec:setting}, together with statistical and computational oracles upon which our algorithms are built. 
In \cref{dm_smooth:sec:alg}, we present our algorithm \smthigw, which illustrates the core ideas of learning with smooth regret at any fixed smoothness level. 
Built upon \smthigw, in \cref{dm_smooth:sec:adaptive}, we present a \corral-type of algorithm that can automatically adapt to any unknown smoothness level.
In \cref{dm_smooth:sec:extension}, we connect our proposed smooth regret to the standard regret over various scenarios.
We present empirical results in \cref{dm_smooth:sec:experiments}, and close with a discussion in \cref{dm_smooth:sec:discussion}.
We defer most proofs to \cref{dm_smooth:sec:proofs}.

\section{Problem Setting}
\label{dm_smooth:sec:setting}

We consider the following standard contextual bandit problems. At any time step $t \in [T]$, nature selects a context $x_t \in \cX$ and a distribution over loss functions $\ell_t: \cA \rightarrow [0,1]$ mapping from the (compact) action set $\cA$ to a loss value in $[0, 1]$.\footnote{For the convenience of leveraging existing results, in this chapter, we consider loss functions instead of reward functions. Nevertheless, for any action $a \in \cA$, its reward can be calculated as  $r_t(a) = 1 - \ell_t(a) \in [0,1]$.}  Conditioned on the context $x_t$, the loss function is stochastically generated, i.e., $\ell_t \sim \P_{\ell_t}(\cdot \mid x_t)$.
The learner selects an action $a_t \in \cA$ based on the revealed context $x_t$, and obtains (only) the loss $\ell_t(a_t)$ of the selected action. 
The learner has access to a set of measurable regression functions $\cF \subseteq (\cX \times \cA \rightarrow [0,1])$ to predict the loss of any context-action pair.
We make the following standard realizability assumption studied in the contextual bandit literature \citep{agarwal2012contextual, foster2018practical, foster2020beyond, simchi2021bypassing}.
\begin{assumption}[Realizability]
\label{dm_smooth:asm:realizability}
There exists a regression function $f^\star \in \cF$ such that $ \E \sq{\ell_t(a) \mid x_t} = f^\star(x_t, a)$ for any $a \in \cA$ and across all $t \in [T]$.
\end{assumption}

\paragraph{The smooth regret}
Let $(\cA, \sgm)$ be a measurable space of the action set and $\mu$ be a base probability measure over the actions. 
Let $\cQ_h$ denote the set of probability measures such that, for any measure $Q \in \cQ_h$, the following holds true: (i) $Q$ is absolutely continuous with respect to the base measure  $\mu$, i.e., $Q \ll \mu$; and (ii) The Radon-Nikodym derivative of $Q$ with respect to $\mu$ is no larger than $\frac{1}{h}$, i.e., $\frac{dQ}{d\mu} \leq 1/h$.
We call $\cQ_h$ the set of {smoothing kernels at smoothness level $h$, or simply put the set of $h$-smoothed kernels.}
For any context $x \in \cX$, we denote by $\smthh(x)$ the smallest loss incurred by any  $h$-smoothed kernel, i.e.,  
\begin{align*}
    \smthh(x) \ldef \inf_{Q \in \cQ_h}\E_{a \sim Q}\sq{f^\star(x,a)}.
\end{align*}
Rather than competing with $\argmin_{a\in\cA}f^\star(x,a)$---an impossible job in many cases---we take $\smthh(x)$ as the benchmark and define the \emph{smooth regret} as follows:
\begin{align}
    \regcbh(T) & \coloneqq  
    \E \sq*{ \sum_{t=1}^T  f^\star(x_t, a_t) - \smthh(x_t) } \label{dm_smooth:eq:smooth_regret}.
\end{align}
One important feature about the above definition is that the benchmark
$\smthh(x_t)$ automatically adapts to the context $x_t$: 
This gives the benchmark more power and makes it harder to compete against. 
In fact, our \smth regret dominates many existing regret measures with \emph{easier} benchmarks. We provide some examples in the following. 
\begin{itemize}
    \item \citet{chaudhuri2018quantile} propose the quantile regret, which aims at competing with the lower $h$-quantile of the loss function, i.e., $v_h(x) \ldef \inf \crl{\zeta: \mu(a \in \cA: f^\star(x,a) \leq \zeta) \geq h}$.
	    Consider $\cS_h \ldef \crl{a \in \cA: f^{\star}(x,a) \leq \nu_h(x)}$ such that $\mu(\cS_h) \geq h$.
	   Let $\wb Q_h \ldef \mu \vert_{\cS_h} / \mu(\cS_h)$ denote the (normalized) probability measure after restricting $\mu$ onto  $\cS_h$.
Since $\wb Q_h \in \cQ_h$, we clearly have $\smthh(x) \leq \E_{a \sim \wb Q_h} \brk{f^{\star}(x,a)} \leq \nu_h(x)$.
Besides, the (original) quantile was only studied in the non-contextual case.
    \item 
\citet{krishnamurthy2020contextual} study a notion of regret that is smoothed in a different way:
Their regret aims at competing with a known and \emph{fixed} smoothing kernel (on top of a {fixed} policy set) with Radon-Nikodym derivative at most ${1}/{h}$. Our benchmark is clearly harder to compete against since we consider any smoothing kernel with Radon-Nikodym derivative at most ${1}/{h}$. 
\end{itemize}
Besides being more competitive with respect to above benchmarks, \smth regret can also be naturally linked to the \emph{standard} regret under various settings previously studied in the bandit literature, e.g., 
in the discrete case with multiple best arms \citep{zhu2020regret} and in the continuous case with Lipschitz/H\"{o}lder continuous payoff functions \citep{kleinberg2004nearly,hadiji2019polynomial}.
We provide detailed discussion in \cref{dm_smooth:sec:extension}.

\subsection{Computational Oracles}
The first step towards designing computationally efficient algorithms is to identify reasonable oracle models to access the sets of regression functions or actions. Otherwise, enumeration over regression functions or actions (both can be exponentially large) immediately invalidate the computational efficiency. 
We consider two common oracle models: a regression oracle and a sampling oracle.

\paragraph{The regression oracles}
A fruitful approach to designing efficient contextual bandit algorithms is through reduction to supervised regression with the class $\cF$ \citep{foster2020beyond, simchi2021bypassing, foster2020adapting, foster2021instance}. 
Following \citet{foster2020beyond}, we assume that we have access to an \emph{online} regression oracle \sqalgtext, which is an algorithm for sequential predication under square loss.
More specifically, the oracle operates in the following protocol: At each round $t \in [T]$, the oracle makes a prediction $\wh f_t$, then receives context-action-loss tuple $(x_t, a_t, \ell_t(a_t))$.
The goal of the oracle is to accurately predict the loss as a function of the context and action, and we evaluate its performance via the square loss $\prn{\wh f_t(x_t,a_t) - \ell_t(a_t)}^2$.
We measure the oracle's cumulative performance through the square-loss regret to $\cF$, which is formalized below.

\begin{assumption}
\label{dm_smooth:assumption:regression_oracle}
The regression oracle \sqalgtext guarantees that, with probability at least $1-\delta$, for any (potentially adaptively chosen) sequence $\curly*{(x_t, a_t, \ell_t(a_t))}_{t=1}^T$, 
\begin{align*}
	\E \Bigg[\sum_{t=1}^T &  \prn*{\wh f_t(x_t, a_t) - \ell_t(a_t)}^2 - 
      \inf_{f \in \cF} \sum_{t=1}^T \prn*{f(x_t, a_t) - \ell_t(a_t)}^2 \Bigg]
    \leq \regsq(T, \delta),
\end{align*}
for some (non-data-dependent) function $\regsq(T, \delta)$.
\end{assumption}

Sometimes it's useful to consider a \emph{weighted} regression oracle, where the square errors are weighted differently. It is shown in \citet{foster2020adapting} (Theorem 5 therein) that any regression oracle satisfies \cref{dm_smooth:assumption:regression_oracle} can be used to generate a weighted regression oracle that satisfies the following assumption.

\newpage
\begin{assumption}
\label{dm_smooth:assumption:regression_oracle_weighted}
The regression oracle \sqalgtext guarantees that, with probability at least $1-\delta$, for any (potentially adaptively chosen) sequence $\curly*{(w_t, x_t, a_t, \ell_t(a_t))}_{t=1}^T$, 
\begin{align*}
	\E \Bigg[ \sum_{t=1}^T & w_t  \prn*{\wh f_t(x_t, a_t) - \ell_t(a_t)}^2 -
     \inf_{f \in \cF} \sum_{t=1}^T w_t \prn*{f(x_t, a_t) - \ell_t(a_t)}^2 \Bigg] \\
			       & \leq \E \sq*{ \max_{t \in [T]} w_t }  \regsq(T, {\delta}),
\end{align*}
for some (non-data-dependent) function $\regsq(T, \delta)$.
\end{assumption}

{For either regression oracle,} we let $\Tsq$ denote an upper bound on the time to (i) query the oracle's estimator $\wh f_t$ with context-action pair $(x_t,a)$ and receive its predicated value $\wh f_t(x_t,a) \in [0,1]$; (ii) query the oracle's estimator $\wh f_t$ with context $x_t$ and receive its argmin action $\wh a_t = \argmin_{a \in \cA} \wh f_t(x_t,a)$; and (iii) update the oracle with example $(x_t, a_t, r_t(a_t))$.
We let $\Msq$ denote the maximum memory used by the oracle throughout its execution.

Online regression is a well-studied problem, with known algorithms for many model classes \citep{foster2020beyond, foster2020adapting}: including linear models \citep{hazan2007logarithmic}, generalized linear models \citep{kakade2011efficient}, non-parametric models \citep{gaillard2015chaining}, and beyond. 
Using Vovk's aggregation algorithm \citep{vovk1998game}, one can show that $\regsq(T, \delta) = O(\log \prn{\abs{\cF} /\delta})$ for any finite set of regression functions $\cF$, which is the canonical setting studied in contextual bandits \citep{langford2007epoch, agarwal2012contextual}.
In the following of this chapter, we use abbreviation $\regsq(T) \ldef \regsq(T, T^{-1})$, and will keep the $\regsq(T)$ term in our regret bounds to accommodate for general set of regression functions.

\paragraph{The sampling oracles}
In order to design algorithms that work with large/continuous action spaces, we assume access to a
sampling oracle \samplealgtext to get access to the action space. 
In particular, the oracle \samplealgtext returns an action $a \sim \mu$ randomly drawn according to the base probability measure $\mu$ over the action space $\cA$.  
We let $\Tsample$ denote a bound on the runtime of single query to the oracle; and let $\Msample$ denote the maximum memory used by the oracle. 

\paragraph{Representing the actions} We use $b_\cA$ to denote the number of bits required to represent any action $a\in \cA$, which scales with $O(\log\abs{\cA})$ with a finite set of actions and $\wt O(d)$ for actions represented as vectors in $\R^d$.
Tighter bounds are possible with additional structual assumptions. Since representing actions is a minimal assumption, we hide the dependence on $b_\cA$ in big-$\bigoh$ notation for our runtime and memory analysis.

\section{Efficient Algorithm with Smooth Regret}
\label{dm_smooth:sec:alg}

We design an oracle-efficient (\smthigw, \cref{dm_smooth:alg:smooth}) algorithm that achieves a $\sqrt{T}$-type regret under the \smth regret defined in \cref{dm_smooth:eq:smooth_regret}. 
We focus on the case when the smoothness level $h>0$ is known in this section,
and leave the design of adaptive algorithms in \cref{dm_smooth:sec:adaptive}.

\cref{dm_smooth:alg:smooth} contains the pseudo code of our proposed \smthigw algorithm, which deploys a smoothed sampling distribution to balance exploration and exploitation.
At each round $t \in [T]$, the learner observes the context $x_t$ from the environment and obtains the estimator $\widehat f_t$ from the regression oracle \sqalgtext.
It then constructs a sampling distribution $P_t$ by mixing a smoothed distribution constructed using the \emph{inverse gap weighting} (IGW) technique \citep{abe1999associative, foster2020beyond} and a delta mass at the greedy action $\wh a_t \ldef \argmin_{a \in \cA} \wh f_t(x_t, a)$.
The algorithm samples an action $a_t \sim P_t$ and then update the regression oracle \sqalgtext. 
The key innovation of the algorithm lies in the construction of the smoothed IGW distribution, which we explain in detail next.

\begin{algorithm}[]
	\caption{\smthigw}
	\label{dm_smooth:alg:smooth} 
	\renewcommand{\algorithmicrequire}{\textbf{Input:}}
	\renewcommand{\algorithmicensure}{\textbf{Output:}}
	\newcommand{\algorithmicbreak}{\textbf{break}}
    \newcommand{\BREAK}{\STATE \algorithmicbreak}
	\begin{algorithmic}[1]
		\REQUIRE Exploration parameter $\gamma > 0$, online regression oracle \sqalgtext.
		\FOR{$t = 1, 2, \dots, T$}
		\STATE Observe context $x_t$.
		\STATE Receive $\widehat f_t$ from regression oracle \sqalgtext.
		\STATE Get $\widehat a_t \gets \argmin_{a \in \cA} \widehat f_t(x_t, a)$.
		\STATE Define 
		\begin{align}
		P_t \ldef M_t +  (1-M_t(\cA)) \cdot \indic_{\wh a_t}, 
    \label{dm_smooth:eq:sampling_dist}
		\end{align}
		where {$M_t$ is the measure defined in \cref{dm_smooth:eq:abe_long_measure}}
		\STATE Sample $a_t \sim P_t$ and observe loss $\ell_t(a_t)$. \algcommentlight{This can be done efficiently via \cref{dm_smooth:alg:reject_sampling}.}
        \STATE  Update \sqalgtext with $(x_t, a_t, \ell_t(a_t))$ 
		\ENDFOR
	\end{algorithmic}
\end{algorithm}

\paragraph{Smoothed variant of IGW} 
The IGW technique was previously used in the finite-action contextual bandit setting \citep{abe1999associative, foster2020beyond}, 
which assigns a probability mass to every action $a \in \cA$ inversely proportional to the estimated loss gap $(\wh f(x,a) - \wh f(x, \wh a))$. 
To extend this strategy to continuous action spaces we leverage Radon-Nikodym derivatives.  
Fix any constant $\gamma > 0$, we define a IGW-type function as 
\begin{align}
    m_t(a) \ldef \frac{1}{1+ h \gamma(\widehat f_t(x_t, a) - \widehat f_t(x_t, \widehat a_t))}. \label{dm_smooth:eq:density_m}
\end{align} 
For any $\omega \in \sgm$, we then define a new measure 
\begin{align}
    M_t(\omega) \ldef \int_{a \in \omega} m_t(a) \, d \mu (a) \label{dm_smooth:eq:abe_long_measure}
\end{align}
of the measurable action space $(\cA, \sgm)$, where $m(a)  = \frac{dM}{d\mu}(a)$ serves as the Radon-Nikodym derivative between the new measure $M$ and the base measure $\mu$.  
Since $m_t(a) \leq 1$ by construction, we have $M_t(\cA) \leq 1$, i.e., $M_t$ is a sub-probability measure.
\smthigw plays a probability measure $P_t \in \Delta(\cA)$ by mixing the sub-probability measure $M_t$ 
with a delta mass at the greedy action $\wh a_t$, as in \cref{dm_smooth:eq:sampling_dist}.

\begin{algorithm}[]
	\caption{Rejection Sampling for IGW}
	\label{dm_smooth:alg:reject_sampling} 
	\renewcommand{\algorithmicrequire}{\textbf{Input:}}
	\renewcommand{\algorithmicensure}{\textbf{Output:}}
	\newcommand{\algorithmicbreak}{\textbf{break}}
    \newcommand{\BREAK}{\STATE \algorithmicbreak}
	\begin{algorithmic}[1]
		\REQUIRE Sampling oracle \samplealgtext, greedy action $\wh a_t$, Radon-Nikodym derivative $m_t(a)$.
		\STATE Draw $a \sim \mu$ from sampling oracle \samplealgtext.
		\STATE Sample $Z$ from a Bernoulli random distribution with mean $m_t(a)$. 
		\IF{$Z=1$}
		\STATE Take action $a$.
	    \ELSE
		\STATE Take action $\wh a_t$.
		\ENDIF
	\end{algorithmic}
\end{algorithm}

\paragraph{Efficient sampling} We now discuss how to sample from the distribution of \cref{dm_smooth:eq:sampling_dist} using a single call to the sampling oracle, via rejection sampling. We first randomly sample an action $a \sim \mu$ from the sampling oracle \samplealgtext and with respect to the base measure $\mu$. We then compute $m_t(a)$ in \cref{dm_smooth:eq:density_m} with two evaluation calls to $\wh f_t$, one at $\wh f_t(x_t, {a})$ and the other at $\wh f_t(x_t, \wh a_t)$. Finally, we sample a random variable $Z$ from a Bernoulli distribution with expectation $m_t(a)$ and play either action $\wh{a}_t$ or action $a$ depending upon the realization of $Z$.
One can show that the sampling distribution described above coincides with the distribution defined in \cref{dm_smooth:eq:sampling_dist} (\cref{dm_smooth:prop:reject_sampling}).\footnote{The same idea can be immediately applied to the case of sampling from the IGW distribution with finite number of actions \citep{foster2020beyond}.}
We present the pseudo code for rejection sampling in \cref{dm_smooth:alg:reject_sampling}.

\begin{proposition}
	\label{dm_smooth:prop:reject_sampling}
    The sampling distribution generated from \cref{dm_smooth:alg:reject_sampling}
    coincides with the sampling distribution defined in \cref{dm_smooth:eq:sampling_dist}. 
\end{proposition}
\begin{proof}[\pfref{dm_smooth:prop:reject_sampling}]
    Let $\wb P_t$ denote the sampling distribution achieved by \cref{dm_smooth:alg:reject_sampling}.
    For any $\omega \in \sgm $, if $\wh a_t \notin \omega$, we have 
    \begin{align*}
        \wb P_t(\omega) = \int_{a \in \omega} m_t(a)\, d\mu(a) = M_t(\omega)
    \end{align*} 
    Now suppose that $\wh a_t \in \omega$: Then the rejection probability, which equals \linebreak 
$\E_{a \sim \mu} \sq*{ 1 - m_t(a)} = 1 - M_t(\cA)$, will be added to the above expression. \end{proof}

We now state the regret bound for \smthigw in the following.
\begin{restatable}{theorem}{thmRegret}
\label{dm_smooth:thm:regret}
Fix any smoothness level $h \in (0,1]$.
With an appropriate choice for  $\gamma > 0$, \cref{dm_smooth:alg:smooth} ensures that 
\begin{align*}
    \regcbh(T) \leq {\sqrt{4 T \, \regsq(T)/h} },
\end{align*}
with per-round runtime $O(\Tsq + \Tsample)$ and maximum memory $O(\Msq + \Msample)$.
\end{restatable}

\paragraph{Key features of \cref{dm_smooth:alg:smooth}}
\cref{dm_smooth:alg:smooth} achieves $\wt O \prn{\sqrt{T /h}}$ regret, which has no dependence on the number of actions.\footnote{We focus on the canonical case studied in contextual bandits with a finite $\cF$, and view $\regsq(T) = O(\log \abs{\cF})$.}  This suggests the \cref{dm_smooth:alg:smooth} can be used in large action spaces scenarios and only suffer regret scales with $1 /h$: the effective number of actions considered for \smth regret.
We next highlight the statistical and computational efficiencies of \cref{dm_smooth:alg:smooth}.
\begin{itemize}
    \item \emph{Statistical optimality.} 
	    It's not hard to prove a $\wt \Omega(\sqrt{T /h})$ lower bound for the \smth regret by relating it to the standard regret under a contextual bandit problem with finite actions: (i) the \smth regret and the standard regret coincides when $h = {1}/{\abs{\cA}}$; and (ii) the standard regret admits lower bound $\wt \Omega(\sqrt{\abs{\cA}T})$ \citep{agarwal2012contextual}.
	    In \cref{dm_smooth:sec:extension}, we further relate our \smth regret guarantee to standard regret guarantee under other scenarios and recover the minimax bounds.

    \item \emph{Computational efficiency.} \cref{dm_smooth:alg:smooth} is oracle-efficient and enjoys per-round runtime and maximum memory that scales linearly with oracle costs. To our knowledge, this leads to the first computationally efficient general-purpose algorithm that achieves a $\sqrt{T}$-type guarantee under \smth regret. The previously known efficient algorithm applies an \greedy-type of strategy and thus only achieves a $T^{2/3}$-type regret (\citet{majzoubi2020efficient}, and with respect to a weaker version of the \smth regret).
	
\end{itemize}

\paragraph{Proof sketch for \cref{dm_smooth:thm:regret}}
To analyze \cref{dm_smooth:alg:smooth}, we follow a recipe introduced by \citet{foster2020beyond, foster2020adapting, foster2021statistical} based on the \emph{\dectext} (DEC, adjusted to our setting), defined as $\dec_\gamma(\cF) \ldef \sup_{\wh{f} , x } \dec_{\gamma}(\cF;\wh{f},x)$, where
\begin{align}
	 \dec_{\gamma}(\cF; \wh{f}, x) \ldef \inf_{P \in \Delta(\cA)} \sup_{f^\star \in \cF} \E_{a \sim P}  
	 \bigg[ f^\star(x, a^\star) - \smthh(x) - \frac{\gamma}{4} \cdot \prn[\big]{\wh{f}(x,a) - f^\star(x,a)}^2 \bigg] .
\label{dm_smooth:eq:dec}
\end{align}
\citet{foster2020beyond, foster2020adapting, foster2021statistical} consider a meta-algorithm which, at each round $t$, (i) computes $\wh f_t$ by appealing to a regression oracle, (ii) computes a distribution $P_t\in\Delta(\cA)$ that solves the minimax problem in \cref{dm_smooth:eq:dec} with $x_t$ and $\wh f_t$ plugged in, and (iii) chooses the action $a_t$ by sampling from this distribution. One can show that for any $\gamma > 0$, this strategy enjoys the following regret bound:
\begin{align}
	\regcbh(T) \approxleq T \cdot \dec_\gamma(\cF) + \gamma \cdot \regsq(T), \label{dm_smooth:eq:decomposition}
\end{align}
More generally, if one computes a distribution that does not solve \cref{dm_smooth:eq:dec} exactly, but instead certifies an upper bound on the DEC of the form $\dec_\gamma(\cF) \leq \wb \dec_\gamma(\cF)$, the same result holds with $\dec_\gamma(\cF)$ replaced by $\wb \dec_\gamma(\cF)$. \cref{dm_smooth:alg:smooth} is a special case of this meta-algorithm, so to bound the regret it suffices to show that the exploration strategy in the algorithm certifies a bound on the DEC.

By applying principles of convex conjugate, we show that the IGW-type distribution of \cref{dm_smooth:eq:sampling_dist} certifies $\dec_\gamma(\cF) \leq \frac{2}{h\gamma}$ for any set of regression functions $\cF$ (\cref{dm_smooth:prop:dec_bound}, deferred to \cref{dm_smooth:app:smooth_supporting}).
With this bound on DEC, We can then bound the first term in \cref{dm_smooth:eq:decomposition} by $O(\frac{T}{h \gamma})$ and optimally tune $\gamma$ in \cref{dm_smooth:eq:decomposition} to obtain the desired regret guarantee.

Deriving the bound on the DEC is one of our key technical contributions, where we simultaneous eliminate the dependence on both the function class and (cardinality of) the action set.
Previous bounds on the DEC assume either a restricted function class $\cF$ or a finite action set.

\section{Adapting to Unknown Smoothness Parameters}
\label{dm_smooth:sec:adaptive}

Our results in \cref{dm_smooth:sec:alg} shows that, with a known $h$, one can achieve  \smth regret proportional to $\sqrt{T/h }$ against the optimal smoothing kernel in $\cQ_h$. The total loss achieved by the learner is the \smth regret plus the total loss suffered by playing the optimal smoothing kernel. One can notice that these two terms go into different directions: When $h$ gets smaller, the loss suffered by the optimal smoothing kernel gets smaller, yet the regret term gets larger. It is apriori unclear how to balance these terms, and therefore desirable to design algorithms that can automatically adapt to an unknown $h \in (0, 1]$. Note it is sufficient to adapt to unknown $h \in [1/T, 1]$, as the regret bound is vacuous for $h < 1/T$.  We provide such an algorithm in this section.

\paragraph{The \corral master algorithm}
Our algorithm follows the standard master-base algorithm structure: We run multiple base algorithms with different configurations in parallel, and then use a master algorithm to conduct model selection on top of base algorithms. The goal of the master algorithm is to balance the regret among base algorithms and eventually achieve a performance that is ``close'' to the best base algorithm (whose identity is unknown). We use the classical \corral algorithm \citep{agarwal2017corralling} as the master algorithm and initiate a collection of $B = \ceil*{\log T}$ (modified) \cref{dm_smooth:alg:smooth} as base algorithms. More specifically, for $b=1,2,\dots,B$, each base algorithm is initialized with smoothness level $h_b = 2^{-b}$. For any $h^\star \in [1/T,1]$, one can notice that there exists a base algorithm $i^\star$ that suits well to this (unknown) $h^\star$ in the sense that $h_{b^\star} \leq h^\star \leq 2 h_{b^\star}$. The goal of the master algorithm is thus to adapt to the base algorithm indexed by $b^\star$.

We provide a brief description of the \corral master algorithm, and direct the reader to \citet{agarwal2017corralling} for more details. The master algorithm maintains a distribution $q_t \in \Delta([B])$ over base algorithms. At each round, the master algorithm sample a base algorithm $I_t \sim q_t$ and passes the context $x_t$, the sampling probability $q_{t,I_t}$ and parameter $\rho_{t,I_t} \coloneqq 1/ \min_{i \leq t} q_{t,I_t}$ into the base algorithm $I_t$. The base algorithm $I_t$ then performs its learning process: it samples an arm $a_t$, observes its loss $\ell_t(a_{t,I_t})$, and then updates its internal state. The master algorithm is updated with respect to the importance-weighted loss $\frac{\ell_t(a_{t,I_t})}{q_{t,I_t}}$ and parameter $\rho_{t,I_t}$. In order to obtain theoretical guarantees, the base algorithms are required to be stable, which is defined as follows.

\begin{definition}
\label{dm_smooth:def:stable}
Suppose the base algorithm indexed by $b$ satisfies---when implemented alone---regret guarantee $\regcbhb(T) \leq R_{b}(T)$ for some non-decreasing $R_{b}(T):\N_+ \rightarrow \R_+$. 
Let $\regimph$ denote the \emph{importance-weighted} regret for base algorithm  $b$, i.e.,
\begin{align*}
\regimphb(T) \ldef 
\E \sq*{ \sum_{t = 1}^T \frac{\ind(I_t = b)}{q_{t,b}} \paren{ f^\star(x_t,a_t) - \smthhb(x_t) } }.
\end{align*}
The base algorithm $b$ is called $(\alpha, R_b(T))$ stable if 
    $\regimphb (T) \leq \E \sq*{\rho^\alpha_{T,b}} R_b(T)$. 
\end{definition}

\paragraph{A stable base algorithm}
Our treatment is inspired by \citet{foster2020adapting}. 
Let $\prn{\tau_1, \tau_2, \ldots} \subseteq [T]$ denote the time steps when the base algorithm $b$ is invoked,
i.e., when $I_t = b$. When invoked, the base algorithm receives $(x_t, q_{t,b}, \rho_{t,b})$ from the master algorithm. The base algorithm then sample from a distribution similar to \cref{dm_smooth:eq:sampling_dist} but with a customized learning rate $\gamma_{t,b} \ldef \sqrt{8T/ (h_b \cdot \rho_{t,b} \cdot \regsq(T))}$. 
After observing the loss $\ell_t(a_{t, b})$, the base algorithm then updates the weighted regression oracle satisfying \cref{dm_smooth:assumption:regression_oracle_weighted}. Our modified algorithm is summarized in \cref{dm_smooth:alg:stable}.

\begin{algorithm}[H]
	\caption{Stable Base Algorithm (Index $b$)}
	\label{dm_smooth:alg:stable} 
	\renewcommand{\algorithmicrequire}{\textbf{Input:}}
	\renewcommand{\algorithmicensure}{\textbf{Output:}}
	\newcommand{\algorithmicbreak}{\textbf{break}}
    \newcommand{\BREAK}{\STATE \algorithmicbreak}
	\begin{algorithmic}[1]
		\REQUIRE Weighted online regression oracle \sqalgtext.
		\STATE Initialize weighted regression oracle \sqalgtext.
		\FOR{$t \in \prn{\tau_1, \tau_2, \ldots} $}
		\STATE Receive context $x_t$, probability $q_{t,b}$ and parameter $\rho_{t,b}$ from the master algorithm.
		\STATE Receive $\widehat f_{t,b}$ from the \emph{weighted} online regression oracle \sqalgtext.
		\STATE Get $\widehat a_{t,b} \gets \argmin_{a \in \cA} \widehat f_{t,b}(x_t, a)$.
		\STATE Define $\gamma_{t,b} \ldef \sqrt{8 T/ (h_b \cdot \rho_{t,b} 
		\cdot \regsq(T))}$ and $w_{t,b} \ldef \ind(I_t=b) \cdot  \gamma_{t,b}/q_{t,b}$.
		\STATE Define $P_{t,b} \ldef M_{t,b} +  (1-M_{t,b}(\cA)) \cdot  \indic_{\wh a_{t,b}}$ according to \cref{dm_smooth:eq:sampling_dist} but with $\gamma_{t,b}$ defined above.
        \STATE Sample $a_{t,b} \sim P_{t,b}$ and observe loss $\ell_t(a_{t,b})$. \algcommentlight{This can be done efficiently via \cref{dm_smooth:alg:reject_sampling}.}
        \STATE  Update the weighted regression oracle \sqalgtext with $(w_{t,b}, x_t, a_t, \ell_t(a_{t,b}))$ 
		\ENDFOR
	\end{algorithmic}
\end{algorithm}

\begin{restatable}{proposition}{propStable}
\label{dm_smooth:prop:stable}
For any $b \in [B]$, \cref{dm_smooth:alg:stable} is $\prn*{\frac{1}{2},\sqrt{4T \, \regsq(T)/h_b}}$-stable, with per-round runtime $O(\Tsq + \Tsample)$ and maximum memory $O(\Msq + \Msample)$.
\end{restatable}

We now provide our model selection guarantees that adapt to unknown smoothness parameter $h \in (0,1]$. The result directly follows from combining the guarantee of \corral \citep{agarwal2017corralling} and our stable base algorithms.
\begin{restatable}{theorem}{thmAdaptive}
\label{dm_smooth:thm:adaptive}
Fix learning rate $\eta \in (0,1]$, the \corral algorithm with \cref{dm_smooth:alg:stable} as base algorithms guarantees that
\begin{align*}
     \regcbh(T) = \wt O \prn*{ \frac{1}{\eta} + \frac{\eta \, T  \, \regsq (T) }{h} }, \forall h \in (0,1].
\end{align*}
The \corral master algorithm has per-round runtime $\wt O(\Tsq + \Tsample)$ and maximum memory $\wt O(\Msq + \Msample)$.
\end{restatable}
\begin{remark}
\label{dm_smooth:rm:adaptive_2}
We keep the current form of \cref{dm_smooth:thm:adaptive} to better generalize to other settings, as explained in \cref{dm_smooth:sec:extension}.
With a slightly different analysis, we can recover the $\wt O \prn{ T^{\frac{1}{1+\beta}} h^{-\beta} \prn*{\log \abs*{\cF }}^{\frac{\beta}{1+ \beta}}}$ guarantee for any $\beta \in [0,1]$, which is known to be Pareto optimal \citep{krishnamurthy2020contextual}. We provide the proofs for this result in \cref{dm_smooth:app:adaptive_2}. 
\end{remark}

\section{Extensions to Standard Regret}
\label{dm_smooth:sec:extension}
We extend our results to various settings under the standard regret guarantee, including the discrete case with multiple best arms, and the continuous case under Lipschitz/H\"{o}lder continuity. Our results not only recover previously known minimax/Pareto optimal guarantees, but also generalize existing results in various ways. 

Although our guarantees are stated in terms of the \smth regret, they are naturally linked to the standard regret among various settings studied in this section. We thus primarily focus on the standard regret in this section. Let $a^\star_t \ldef \argmin_{a \in \cA} f^\star(x_t, a)$ denote the best action under context $x_t$.
The \emph{standard} (expected) regret is defined as
\begin{align*}
    \regcb(T) & \ldef \E \sq*{\sum_{t=1}^T  f^{\star}(x_t, a_t) - f^{\star}(x_t, a^\star_t)}. \label{dm_smooth:eq:regret}
\end{align*}
We focus on the canonical case with a finite set of regression functions $\cF$ and consider $\regsq(\cF) = O(\log \prn{\abs*{\cF}T})$ \citep{vovk1998game}.

\subsection{Discrete Case: Bandits with Multiple Best Arms}
\citet{zhu2020regret} study a non-contextual bandit problem with a large (discrete) action set $\cA$ which might contain multiple best arms. More specifically, suppose there exists a subset of optimal arms $\cA^\star \subseteq \cA$ with cardinalities $\abs*{\cA^\star}= K^\star$ and $\abs*{\cA} = K$, the goal is to adapt to the effective number of arms $\frac{K}{K^\star}$ and minimize the standard regret. Note that one could have $\frac{K}{K^\star} \ll K$ when $K^\star$ is large.

\textbf{Existing Results.} Suppose $\frac{K}{K^\star} = \Theta (T^\alpha)$ for some $\alpha \in [0,1]$. \citet{zhu2020regret} shows that: (i) when $\alpha$ is known, the minimax regret is $\wt \Theta(T^{(1+\alpha)/2})$; and (ii) when $\alpha$ is unknown, the Pareto optimal regret can be described by \linebreak
$\widetilde O( \max \curly*{T^{\beta}, T^{1+\alpha - \beta} })$ for any $\beta \in [0,1)$.

\textbf{Our Generalizations.} We extend the problem to the contextual setting: We use $\cA^\star_{x_t} \subseteq \cA$ to denote the \emph{subset} of optimal arms with respect to context $x_t$, and analogously assume that $\inf_{x \in \cX} \abs*{\cA^\star_{x}} = K^\star$ and $\frac{K}{K^\star}=T^\alpha$. 

Since $\frac{K^\star}{K}$ represents the proportion of actions that are optimal, by setting $h = \frac{K^\star}{K} = T^{-\alpha}$ (and under uniform measure), we can then relate the standard regret to the smooth regret, i.e., $\regcb(T) = \regcbh(T)$. In the case when $\alpha$ is known, \cref{dm_smooth:thm:regret} implies that $\regcb(T) = O \prn[\big]{T^{(1+\alpha)/2} \log^{1/2} \prn{\abs*{\cF}T}}$. In the case with unknown $\alpha$, by setting $\eta = T^{-\beta}$ in \cref{dm_smooth:thm:adaptive}, we have 
\begin{align*}
    \regcb(T) = O \paren[\big]{ \max \paren{ T^\beta, T^{1+\alpha-\beta} \log \prn{\abs*{\cF}T} }}.
\end{align*}
These results generalize the known minimax/Pareto optimal results in \citet{zhu2020regret} to the contextual bandit case, up to logarithmic factors.

\subsection{Continuous Case: Lipschitz/H\"older Bandits}

\citet{kleinberg2004nearly, hadiji2019polynomial} study non-contextual bandit problems with (non-contextual) mean payoff functions $f^\star(a)$ satisfying H\"older continuity. More specifically, let $\cA = [0,1]$ (with uniform measure) and $L, \alpha > 0$ be some H\"older smoothness parameters, the assumption is that
\begin{align*}
    \abs{ f^\star(a) - f^\star(a^\prime) } \leq L \, \abs{a - a^\prime}^\alpha,
\end{align*}
for any $a, a^\prime \in \cA$. The goal is to adapt to provide standard regret guarantee that adapts to the smoothness parameters $L$ and $\alpha$.

\textbf{Existing Results.} In the case when $L,\alpha$ are known, \citet{kleinberg2004nearly} shows that the minimax regret scales as $\Theta \prn{L^{1/(2\alpha+1)} T^{(\alpha+1)/(2\alpha+1)}}$; in the case with unknown $L, \alpha$, \citet{hadiji2019polynomial} shows that the Pareto optimal regret can be described by $\wt O \prn[\big]{\max \crl{T^\beta, L^{1/(1+\alpha)} T^{1- \frac{\alpha}{1+\alpha}\beta } }}$ for any $\beta \in [\frac{1}{2},1]$.

\textbf{Our Generalizations.} We extend the setting to the contextual bandit case and make the following analogous H\"older continuity assumption,\footnote{The special case with Lipschitz continuity ($\alpha = 1$) has been previously studied in the contextual setting, e.g., see \citet{krishnamurthy2020contextual}.} i.e.,
\begin{align*}
    \abs{ f^\star(x,a) - f^\star(x, a^\prime) } \leq L \, \abs{a - a^\prime}^\alpha, \quad \forall x \in \cX.
\end{align*}
We first divide the action set $\cA=[0,1]$ into $B=\ceil*{1/h}$ consecutive intervals $\crl{I_b}_{b=1}^{B}$ such that $I_b = [(b-1){h}, {b}{h}]$. Let $b_t$ denote the index of the interval where the best action $a^\star_t \ldef \argmin_{a \in \cA}f^{\star}(x_t, a)$ lies into, i.e., $a^\star_t \in I_{b_t}$. Our \smth regret (at level $h$) provides guarantees with respect to the smoothing kernel $\unif(I_{b_t})$. Since we have $\E_{a \sim \unif(I_{b_t})} \sq{f^\star(x_t, a)} \leq f^\star(x_t,a^\star_t)  + L h^\alpha$ under H\"older continuity, the following guarantee holds under the standard regret
\begin{equation}
    \regcb(T) \leq \regcbh(T) + L h^\alpha T. \label{dm_smooth:eq:holder} 
\end{equation}
When $L,\alpha$ are known, setting 
$h = \Theta \prn[\big]{L^{-2/(2\alpha+1)} T^{-1/(2\alpha + 1)} \log^{1/(2\alpha+1)} \prn{\abs*{\cF}T}}$ in \cref{dm_smooth:thm:regret} (together with \cref{dm_smooth:eq:holder}) leads to regret guarantee \linebreak
$O\prn[\big]{ L^{1/(2\alpha+1)} T^{(\alpha+1)/(2\alpha+1)} \log^{(\alpha/(2\alpha+1)} \prn{\abs*{\cF}T}}$, which is nearly minimax optimal \citep{kleinberg2004nearly}. In the case when $L,\alpha$ are unknown,
setting $\eta = T^{-\beta}$ in \cref{dm_smooth:thm:adaptive} (together with \cref{dm_smooth:eq:holder}) leads to
\begin{align*}
    \regcb(T) 
    = O \prn*{\max \crl*{T^\beta, L^{1/(1+2\alpha)} T^{1 - \frac{\alpha}{1+\alpha}\beta} \log^{\alpha/(1+\alpha)} \prn{\abs{\cF}T}}}, 
\end{align*} 
which matches the Pareto frontier obtained in \citet{hadiji2019polynomial} up to logarithmic factors.

\section{Experiments}
\label{dm_smooth:sec:experiments}

In this section we compare our technique empirically with prior art from the bandit and contextual bandit literature.  Code to reproduce these experiments is available at \url{https://github.com/pmineiro/smoothcb}.

\subsection{Comparison with Bandit Prior Art}

\begin{figure}[H]
    \centering
    \includegraphics[width=0.8\textwidth]{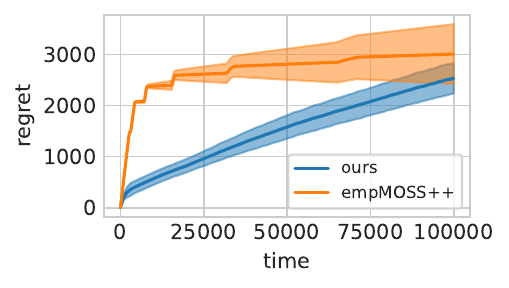}
    \caption{Comparison of regret on a bandit dataset with a discrete action space.}
    \label{dm_smooth:fig:newyorker}
\end{figure}

We replicate the real-world dataset experiment from \citet{zhu2020regret}.  The dataset consists of 10025 captions from the \emph{New Yorker Magazine} Cartoon Caption Contest and associated average ratings, normalized to [0, 1]. The caption text is discarded 
resulting in a non-contextual bandit problem with 10025 arms. When an arm is chosen, the algorithm experiences a Bernoulli loss realization whose mean is one minus the average rating for that arm.  The goal is to experience minimum regret over the planning horizon $T = 10^5$.  There are 54 arms in the dataset that have the minimal mean loss of 0.

For our algorithm, we used the uniform distribution over $[1, 2, \ldots, |\cA|]$ as a reference measure, for which $O(1)$ sampling is available.  We instantiated a tabular regression function, i.e., for each arm we maintained the empirical loss frequency observed for that arm.  
We use \corral with learning rate $\eta = 1$ and instantiated 8 subalgorithms with $\gamma h$ geometrically evenly spaced between $10^3$ and $10^6$.  These were our initial hyperparameter choices, but they worked well enough that no tuning was required.

In \cref{dm_smooth:fig:newyorker}, we compare our technique with \texttt{empMOSS++}, the best performing technique from \citet{zhu2020regret}. We plot the regret for both algorithms (smaller is better). Following the display convention of \citet{zhu2020regret}, shaded areas in the plot represent $0.5$ standard deviation (i.e., it captures around 38\% confidence region). Our technique is statistically equivalent.

\subsection{Comparison with Contextual Bandit Prior Art}

We replicate the online setting from \citet{majzoubi2020efficient}, where 5 large-scale OpenML regression datasets are converted into continuous action problems on $[0, 1]$ by shifting and scaling the target values into this range.  The context $x$ is a mix of numerical and categorical variables depending upon the particular OpenML dataset.  For any example, when the algorithm plays action $a$ and the true target is $y$, the algorithm experiences loss $|y - a|$ as bandit feedback.  

We use Lebesgue measure on $[0, 1]$ as our reference measure, for which $O(1)$ sampling is available.  To maintain $O(1)$ computation, we consider regression functions with (learned) parameters $\theta$ via $
f(x, a; \theta) \ldef g\left(\wh{a}\left(x; \theta\right) - a; \theta\right)$ where, for any $\theta$, $z=0$ is a global minimizer of $g(z; \theta)$.  Subject to this constraint, we are free to choose $g(\cdot; \theta)$ and $\wh{a}(\cdot; \theta)$ and yet are ensured that we can directly compute the minimizer of our loss predictor via $\wh a(x;\theta)$. For our experiments we use a logistic loss predictor and a linear argmin predictor with logistic link: Let $\theta \ldef \prn{v; w; \xi}$, we choose $$
\begin{aligned}
g(z; \theta) \ldef \sigma\left(|w| |z| + \xi \right), \quad \text{and}
\quad \wh{a}(x; \theta) \ldef \sigma\left(v^\top x\right),
\end{aligned}
$$ where $\sigma(\cdot)$ is the sigmoid function.

\begin{table}
     \caption{Average progressive loss on contextual bandits datasets with continuous action spaces, scaled by 1000.}
     \label{dm_smooth:tab:cats}
     \vspace{0.1in}
     \centering
     \begin{tabular}{c c c c}
     \toprule
     &\texttt{\small CATS}&\texttt{\small Ours (Linear)}&\texttt{\small Ours (RFF)} \\
          \midrule
     \texttt{Cpu}&$[55, 57]$ & $[40.6, 40.7]$ & $\bold{[38.6, 38.7]}$ \\
     \texttt{Fri}&$[183, 187]$ & $[161, 163]$& $\bold{[156, 157]}$ \\
     \texttt{Price}&$[108, 110]$ & $[70.2, 70.5]$ & $\bold{[66.1, 66.3]}$ \\
     \texttt{Wis}&$[172, 174]$ & $[138, 139]$ & $\bold{[136.2, 136.6]}$ \\
     \texttt{Zur}&$[24, 26]$ & $[24.3, 24.4]$ & $[25.4, 25.5]$\\
     \bottomrule
     \end{tabular}
     
\end{table}

In \cref{dm_smooth:tab:cats}, we compare our technique with \cats from \citet{majzoubi2020efficient}.  Following their protocol, we tune hyperparameters for each dataset to be optimal in-hindsight, and then report {95\%} bootstrap confidence intervals based upon the progressive loss of a single run.  Our algorithm outperforms \cats. 

To further exhibit the generality of our technique, we also include results for a nonlinear argmin predictor in \cref{dm_smooth:tab:cats} (last column), which uses a Laplace kernel regressor implemented via random Fourier features~\citep{rahimi2007random} 
to predict the argmin.  
This approach achieves even better empirical performance.
 
\section{Discussion}
\label{dm_smooth:sec:discussion}

This work presents simple and practical algorithms for contextual bandits with large---or even continuous---action spaces, continuing a line of research which assumes actions that achieve low loss are not rare. 
While our approach can be used to recover minimax/Pareto optimal guarantees under certain structural assumptions (e.g., with H\"older/Lipschitz continuity), it doesn't cover all cases.
For instance, on a large but finite action set with a linear reward function, the optimal smoothing kernel can be made to perform arbitrarily worse than the optimal action (e.g., by having one optimal action lying in an orthogonal space of all other actions); in this construction, algorithms provided in this chapter would perform poorly relative to specialized linear contextual bandit algorithms.

In future work we will focus on offline evaluation.  Our technique already generates data that is suitable for subsequent offline evaluation of policies absolutely continuous with the reference measure, but only when the submeasure sample is accepted (line 4 of \cref{dm_smooth:alg:reject_sampling}), i.e., only $M(\cA)$ fraction of the data is suitable for reuse. 
We plan to refine our sampling distribution so that the fraction of re-usable data can be increased, but presumably at the cost of additional computation.

We manage to achieve a $\sqrt{T}$-regret guarantee with respect to \smth regret, which dominates previously studied regret notions that competing against easier benchmarks.
A natural question to ask is, what is the strongest benchmark such that it is possible to still achieve a $\sqrt{T}$-type guarantee for problems with arbitrarily large action spaces?  Speculating, there might exist a regret notion which dominates \smth regret yet still admits a $\sqrt{T}$ guarantee.\looseness=-1

\section{Proofs and Supporting Results}
\label{dm_smooth:sec:proofs}

\subsection{Proofs and Supporting Results for \cref{dm_smooth:sec:alg}}
\label{dm_smooth:app:smooth}

This section is organized as follows. We provide supporting results in \cref{dm_smooth:app:smooth_supporting}, then give the proof of \cref{dm_smooth:thm:regret} in \cref{dm_smooth:app:smooth_reg}.

\subsubsection{Supporting Results}
\label{dm_smooth:app:smooth_supporting}

\paragraph{Preliminaries}
\paranewline

\noindent
We first introduce the concept of convex conjugate. 
For any function $\phi:\R \rightarrow \R \cup \curly*{- \infty, + \infty}$, its convex conjugate $\phi^\star: \R \rightarrow \R \cup \curly*{- \infty, + \infty}$ is defined as 
\begin{align*}
    \phi^\star(w) \ldef \sup_{v \in \R} \paren{ vw - \phi(v) }. 
\end{align*}
Since $(\phi^\star)^\star = \phi$, we have (Young-Fenchel inequality)
\begin{align}
    \phi(v) \geq vw - \phi^\star(w), \label{dm_smooth:eq:young_fenchel}
\end{align}
for any $w \in \dom(\phi^\star)$.
\begin{lemma}
\label{dm_smooth:prop:convex_conjugate}
$\phi(v) = \frac{1}{\gamma}(v-1)^2$ and $\phi^\star(w)=w + \frac{\gamma}{4}w^2$ are convex conjugates.
\end{lemma}
\begin{proof}[\pfref{dm_smooth:prop:convex_conjugate}]
    By definition of the convex conjugate, we have 
    \begin{align*}
        \phi^\star(w) & = \sup_{v \in \R} \prn*{-\frac{1}{\gamma} \cdot \prn*{v^2 - \prn{2 + \gamma w }v + 1}} \\
        & = w + \frac{\gamma}{4} w^2,
    \end{align*}
    where the second line follows from plugging in the maximizer $v = \frac{\gamma w }{2} + 1$. Note that the domain of $\phi^\star(w)$ is in fact $\R^d$ here. So, \cref{dm_smooth:eq:young_fenchel} holds for any $w \in \R^d$.
\end{proof}

We also introduce the concept of $\chi^2$ divergence. For probability measures $P$ and $Q$ on the same measurable space $(\cA, \sgm)$ such that $Q \ll P$, the $\chi^2$ divergence of $Q$ from $P$ is defined as 
\begin{align*}
    \chi^2 \infdiv{Q}{P} \ldef \E_{a \sim P} \sq*{\prn*{\frac{dQ}{dP}(a) - 1}^2},
\end{align*}
where $\frac{dQ}{dP}(a)$ denotes the Radon-Nikodym derivative of $Q$ with respect to $P$,
which is a function mapping from $a$ to $\R$.

\paragraph{Bounding the \dectext}
\paranewline

\noindent
We aim at bounding the \dectext in this section. 
We use expression $\inf_{Q \in \cQ_h} \E_{a^\star \sim Q} \sq{f^\star(x,a^\star)}$ for $\smthh(x)$.
With this expression, we rewrite the \dectext in the following: 
With respect to any context $x\in \cX$ and estimator $\wh f $ obtained from \sqalgtext, we denote
\begin{align*}
    &\dec_\gamma(\cF; \wh f, x) \ldef \\
    &\inf_{P \in \Delta(\cA)} \sup_{Q \in \cQ_h} \sup_{f \in \cF} \E_{a\sim P, a^\star \sim Q} \sq*{f(x, a) - f(x, a^\star) - \frac{\gamma}{4} \cdot \prn*{\widehat f(x,a ) - f(x, a)}^2}, 
\end{align*}
and define $\dec_\gamma(\cF) \ldef \sup_{\wh f, x } \dec_\gamma(\cF; \wh f, x)$ as the \dectext. 
We remark here that $\sup_{Q \in \cQ_h} \E_{a^\star \sim Q}   \sq{ - f(x,a^\star)} =  - \inf_{Q \in \cQ_h}\E_{a\sim Q} \sq{f^\star(x,a^\star)}$ so we are still compete with the best smoothing kernel within $\cQ_h$.

We first state a result that helps eliminate the unknown $f$ function in \dectext
(and thus the $\sup_{f\in \cF}$ term), and bound \dectext by the known $\wh f$ estimator (from the regression oracle \sqalgtext) and the $\chi^2$-divergence from $Q$ to $P$ (whenever $P$ and $Q$ are probability measures).

\begin{lemma}
    \label{dm_smooth:lm:dec_chi2}
    Fix constant $\gamma > 0$ and context $x\in\cX$ . For any measures $P$ and $Q$ such that $Q \ll P$, we have 
    \begin{align*}
        &  \sup_{f \in \cF} \E_{a\sim P, a^\star \sim Q} \sq*{f(x, a) - f(x, a^\star) - \frac{\gamma}{4} \cdot \prn*{\widehat f(x,a ) - f(x, a)}^2} \\
        & \leq \E_{a \sim P} \sq[\big]{\widehat f(x, a)} - \E_{a \sim Q} \sq[\big]{\widehat f(x, a)} +\frac{1}{\gamma} \cdot  \E_{a \sim P}\sq*{ \prn*{\frac{dQ}{dP}(a) - 1}^2 }. 
    \end{align*}
\end{lemma}

\begin{proof}[\pfref{dm_smooth:lm:dec_chi2}]
We omit the dependence on the context $x\in \cX$, and use abbreviations $f(a) \ldef f(x,a)$ and $\wh f(a) \ldef \wh f(x,a)$. Let $g \ldef f - \widehat f$, we re-write the expression as
\begin{align*}
    & \sup_{f \in \cF} \E_{a\sim P, a^\star \sim Q} \sq*{f( a) - f(a^\star) - \frac{\gamma}{4} \cdot \prn*{\widehat f(a ) - f(a)}^2}\\ 
    & = \sup_{g \in \cF - \widehat f} \E_{a\sim P} \sq[\big]{\widehat f(a)} - \E_{a^\star \sim Q} \sq[\big]{\widehat f(a^\star)} - \E_{a^\star \sim Q}\sq[\big]{ g(a^\star)} + \E_{a \sim P} \sq*{ g(a) - \frac{\gamma}{4} \cdot \prn*{g(a)}^2}\\
    & =  \E_{a\sim P} \sq[\big]{\widehat f(a)} - \E_{a \sim Q} \sq[\big]{\widehat f(a)} \nonumber \\
    & \quad + \sup_{g \in \cF - \widehat f} \prn*{ \E_{a \sim Q}\sq[\big]{ - g(a)} -\E_{a \sim P} \sq*{ \prn{-g(a)} + \frac{\gamma}{4} \cdot \prn*{ - g(a)}^2}} \\
    & =  \E_{a\sim P} \sq[\big]{\widehat f(a)} - \E_{a \sim Q} \sq[\big]{\widehat f(a)} \nonumber \\
    & \quad + \sup_{g \in \cF - \widehat f}  \E_{a \sim P}\sq*{ \frac{dQ}{dP}(a) \cdot \prn{- g(a)} -  \prn*{\prn{-g(a)} + \frac{\gamma}{4} \cdot \prn*{ - g(a)}^2}}\\
    & =  \E_{a\sim P} \sq[\big]{\widehat f(a)} - \E_{a \sim Q} \sq[\big]{\widehat f(a)} + \sup_{g \in \cF - \widehat f}  \E_{a \sim P}\sq*{ \frac{dQ}{dP}(a) \cdot \prn{- g(a)} -  \phi^\star \prn{-g(a)} },
\end{align*}
where we use the fact that $Q \ll P$ and $\phi^\star (w) = w + \frac{\gamma}{4} w^2$. Focus on the last term that depends on $g$ takes the form of the RHS of \cref{dm_smooth:eq:young_fenchel}: Consider $v = \frac{dQ}{dP}(a)$ and $w = -g(a)$ and apply \cref{dm_smooth:eq:young_fenchel} (with \cref{dm_smooth:prop:convex_conjugate}) eliminates the dependence on $g$ (since it works for any $w = - g(a)$) and leads to the following bound 
\begin{align*}
    & \sup_{f \in \cF} \E_{a\sim P, a^\star \sim Q} \sq*{f( a) - f(a^\star) - \frac{\gamma}{4} \cdot \prn*{\widehat f(a ) - f(a)}^2}\\  
    & \leq \E_{a\sim P} \sq[\big]{\widehat f(a)} - \E_{a \sim Q} \sq[\big]{\widehat f(a)} + \frac{1}{\gamma} \cdot  \E_{a \sim P}\sq*{ \prn*{\frac{dQ}{dP}(a) - 1}^2 }. 
\end{align*}
\end{proof}

We now bound the \dectext with sampling distribution defined in \cref{dm_smooth:eq:sampling_dist}. We drop the dependence on $t$ and define the sampling distribution in the generic form: Fix any constant $\gamma > 0$, context $x \in \cX$ and estimator $\wh f $, we define sampling distribution 
\begin{align}
	P \ldef M + (1 - M(\cA)) \cdot \indic_{\wh a} \label{dm_smooth:eq:sampling_dist_gen},
\end{align}
where $\wh a \ldef \argmin_{a \in \cA} \wh f(x,a)$ and the measure $M$ is defined through $M(\omega) \ldef \int_{a \in \omega} m(a) \, d \mu(a) $ with
\begin{align}
    m(a) \ldef \frac{1}{1+ h \gamma(\widehat f(x, a) - \widehat f(x, \widehat a))}. 
    \label{dm_smooth:eq:abe_long_measure_gen}
\end{align} 

\begin{restatable}{lemma}{propDecBound}
    \label{dm_smooth:prop:dec_bound}
    Fix any constant $\gamma > 0$ and any set of regression function $\cF$. Let $P$ be the sampling distribution defined in \cref{dm_smooth:eq:sampling_dist_gen}, we then have $\dec_\gamma (\cF) \leq \frac{2}{h \, \gamma}$.
\end{restatable}

\begin{proof}[\pfref{dm_smooth:prop:dec_bound}]
    As in the proof of \cref{dm_smooth:lm:dec_chi2}, we omit the dependence on the context $x \in \cX$ and use abbreviations $f(a) \ldef f(x,a)$ and $\wh f(a) \ldef \wh f(x,a)$.

    We first notice that for any $Q \in \cQ_h$ we have $Q \ll M$ for $M$ defined in \cref{dm_smooth:eq:abe_long_measure_gen}: we have (i) $Q \ll \mu$ by definition, and (ii) $\mu \ll M$ (since $m(a) \geq \frac{1}{1+h\gamma} > 0$).\footnote{We thus have $Q \ll P$ as well since $P$ contains the component $M$ by definition. We will, however, mostly be working with $M$ due to its nice connection with the base measure $\mu$, as defined in \cref{dm_smooth:eq:abe_long_measure_gen}.} 
    On the other side, however, we do not necessarily have $P \ll \mu$ for $P$ defined in \cref{dm_smooth:eq:sampling_dist_gen}: It's possible to have $P \prn{\crl{a^\star}} > 0$ yet $\mu \prn{\crl{a^\star}} = 0$, e.g., $\mu$ is some continuous measure.
    To isolate the corner case, we first give the following decomposition for any $Q \in \cQ_h$ and $f \in \cF$.
    With $P \ldef M + (1 - M(\cA)) \cdot \indic_{\wh a}$, we have 
    \begin{align}
        & \E_{a\sim P, a^\star \sim Q} \sq*{f( a) - f(a^\star) - \frac{\gamma}{4} \cdot \prn*{\widehat f(a ) - f(a)}^2}\nonumber\\
        & = \prn{1-M(\cA)} \cdot \prn*{ f(\wh a) - \frac{\gamma}{4} \cdot \prn[\big]{\wh f(\wh a) - f(\wh a)}^2} \nonumber \\
	& \quad + \E_{a\sim M, a^\star \sim Q} \sq*{f( a) - f(a^\star) - \frac{\gamma}{4} \cdot \prn*{\widehat f(a ) - f(a)}^2} \nonumber\\
        & = \prn{1-M(\cA)} \cdot \prn*{ \wh f(\wh a) + \prn[\big]{f(\wh a) - \wh f(\wh a)} - \frac{\gamma}{4} \cdot \prn[\big]{\wh f(\wh a) - f(\wh a)}^2} \nonumber \\
	& \quad + \E_{a\sim M, a^\star \sim Q} \sq*{f( a) - f(a^\star) - \frac{\gamma}{4} \cdot \prn*{\widehat f(a ) - f(a)}^2} \nonumber\\
        & \leq \prn{1-M(\cA)} \cdot \prn[\Big]{ \wh f(\wh a) + \frac{1}{\gamma}} 
        + \E_{a\sim M, a^\star \sim Q} \sq*{f( a) - f(a^\star) - \frac{\gamma}{4} \cdot \prn*{\widehat f(a ) - f(a)}^2} \nonumber \\
        & \leq \frac{1 - M(\cA)}{\gamma} + \prn{1-M(\cA)} \cdot { \wh f(\wh a)  } 
        + \E_{a\sim M} \sq[\big]{\widehat f(a)} - \E_{a \sim Q} \sq[\big]{\widehat f(a)} \nonumber \\
	& \quad + \frac{1}{\gamma} \cdot  \E_{a \sim M}\sq*{ \prn*{\frac{dQ}{dM}(a) - 1}^2 }, \label{dm_smooth:eq:prop_dec_bound_1}
    \end{align}
    where the fourth line follows from applying AM-GM inequality and the fifth line follows from applying \cref{dm_smooth:lm:dec_chi2} with $Q \ll M$.\footnote{With a slight abuse of notation, we use $\E_{a \sim M}[\cdot]$ denote the integration with respect to the sub-probability measure $M$.} 
We now focus on the last four terms in \cref{dm_smooth:eq:prop_dec_bound_1}. Denote $m(a) \ldef \frac{dM}{d \mu} (a)$ and $q(a) \ldef \frac{dQ}{d \mu}(a)$, with change of measures, we have 
    \begin{align}
        & (1 - M(\cA) \cdot \prn[\big]{\wh f(\wh a)} + \E_{a\sim M} \sq[\big]{\widehat f(a)} - \E_{a \sim Q} \sq[\big]{\widehat f(a)} + \frac{1}{\gamma} \cdot  \E_{a \sim M}\sq*{ \prn*{\frac{dQ}{dM}(a) - 1}^2 } \nonumber\\
        & = \E_{a \sim \mu} \sq*{m(a) \cdot \prn[\Big]{\wh f(a) - \wh f(\wh a)} } - \E_{a \sim \mu} \sq*{q(a) \cdot \prn[\Big]{\wh f(a) - \wh f(\wh a)} } \nonumber \\
	& \quad + \frac{1}{\gamma} \cdot \E_{a \sim \mu} \sq*{m(a) \cdot \prn*{\frac{q(a)}{m(a)} - 1}^2 } \nonumber \\
        & = \E_{a \sim \mu} \sq*{m(a) \cdot \prn[\Big]{\wh f(a) - \wh f(\wh a)} } - \E_{a \sim \mu} \sq*{q(a) \cdot \prn[\Big]{\wh f(a) - \wh f(\wh a)} } \nonumber \\
	& \quad + \frac{1}{\gamma} \cdot \E_{a \sim \mu} \sq*{q(a) \cdot \frac{q(a)}{m(a)} - 2 q(a) + m(a)  } \nonumber \\
	& = \E_{a \sim \mu} \sq*{m(a) \cdot \prn[\Big]{\wh f(a) - \wh f(\wh a)} } + 
	\frac{1}{\gamma} \cdot \E_{a \sim Q} \brk*{\frac{q(a)}{m(a)} - \gamma \cdot \prn*{\wh f(a) - \wh f(\wh a)} } \nonumber \\
	& \quad + \frac{M(\cA) - 2}{ \gamma}
 \label{dm_smooth:eq:prop_dec_bound_2}
    \end{align}
   Plugging \cref{dm_smooth:eq:prop_dec_bound_2} into \cref{dm_smooth:eq:prop_dec_bound_1} leads to 
   \begin{align}
	   &	   \E_{a\sim P, a^\star \sim Q}  \sq*{f( a) - f(a^\star) - \frac{\gamma}{4} \cdot \prn*{\widehat f(a ) - f(a)}^2} \nonumber \\
	& \leq
\E_{a \sim \mu} \sq*{m(a) \cdot \prn[\Big]{\wh f(a) - \wh f(\wh a)} } +
	\frac{1}{\gamma} \cdot \E_{a \sim Q} \brk*{\frac{q(a)}{m(a)} - \gamma \cdot \prn*{\wh f(a) - \wh f(\wh a)} } \nonumber \\
	& \leq  \frac{2}{h\gamma},
\label{dm_smooth:eq:prop_dec_bound_3}
   \end{align}
   where \cref{dm_smooth:eq:prop_dec_bound_3} follows from the fact that $m(a) \ldef \frac{dM}{d \mu}(a) = \frac{1}{1 + h \gamma \prn{\wh f(a) - \wh f(\wh a)}}$ and $q(a) \ldef \frac{dQ}{d \mu}(a) \leq \frac{1}{h}$ for any $Q \in \cQ_h$.
   This certifies that $\dec_\gamma(\cF) \leq \frac{2}{h \gamma}$.
\end{proof}

\subsubsection{\pfref{dm_smooth:thm:regret}}
\label{dm_smooth:app:smooth_reg}

\thmRegret*

\begin{proof}[\pfref{dm_smooth:thm:regret}]
    We use abbreviation $f_t(a) \ldef f(x_t,a)$ for any $f \in \cF$.
    Let $a^\star_t$ denote the action sampled according to the best smoothing kernel within $\cQ_h$ (which could change from round to round). 
    We let $\cE$ denote the good event where the regret guarantee stated in \cref{dm_smooth:assumption:regression_oracle} (i.e., $\regsq(T) \ldef \regsq(T, T^{-1})$) holds with probability at least  $1- T^{-1}$. Conditioned on this good event, following the analysis provided in \citet{foster2020adapting}, we decompose the contextual bandit regret as follows.
    \begin{align*}
	   & \E \brk*{\sum_{t=1}^T f_t^\star(a_t) - f_t^\star(a^\star_t)}\\
        & = \E \sq*{\sum_{t=1}^T f_t^\star(a_t) - f_t^\star(a^\star_t) - \frac{\gamma}{4} \cdot \prn*{\wh f_t( a_t) - f_t^\star(a_t)}^2}\\
	& \quad + \frac{\gamma}{4} \cdot  \E \sq*{\sum_{t=1}^T \prn*{\wh f_t(a_t) - f_t^\star(a_t)}^2} \nonumber \\
        & \leq T \cdot \frac{2}{h \gamma} + \frac{\gamma}{4} \cdot  \E \sq*{\sum_{t=1}^T \prn*{\wh f_t(a_t) - f_t^\star(a_t)}^2},
    \end{align*}
    where the bound on the first term follows from \cref{dm_smooth:prop:dec_bound}. We analyze the second term below.
    \begin{align*}
        & \frac{\gamma}{4} \cdot \E \Biggl[\sum_{t=1}^T \biggl( \prn*{\wh f_t(a_t) - \ell_t(a_t)}^2 -\prn[\Big]{f^{\star}(a_t) - \ell_t(a_t)}^2 \\
	& \quad + 2 \prn[\Big]{\ell_t(a_t) - f^\star_t(a_t)} \cdot \prn[\Big]{\wh f_t(a_t) - f^\star_t(a_t)} \biggr)  \Biggr] \\
        & =  \frac{\gamma}{4} \cdot \E \sq*{\sum_{t=1}^T \prn*{ \prn*{\wh f_t(a_t) - \ell_t(a_t)}^2 -\prn[\Big]{f^\star_t(a_t) - \ell_t(a_t)}^2  }} \\
        & \leq \frac{\gamma}{4} \cdot \regsq(T), 
    \end{align*}
    where on the second line follows from the fact that $\E \sq{\ell_t(a) \mid x_t} = f^\star(x_t,a)$ and $\ell_t$ is conditionally independent of $a_t$, and the third line follows from the bound on regression oracle stated in \cref{dm_smooth:assumption:regression_oracle}. 
    As a result, we have 
    \begin{align*}
        \regcbh(T) \leq \frac{2T}{h \gamma} + \frac{\gamma}{4} \cdot \regsq(T) + O(1),
    \end{align*}
    where the additional term $O(1)$ accounts for the expected regret suffered under event  $\neg \cE$.
    Taking $\gamma = \sqrt{8 T/\prn{h \cdot \regsq(T)}}$ leads to the desired result.

    \noindent\emph{Computational complexity.}
    We now discuss the computational complexity of \cref{dm_smooth:alg:smooth}. At each round \cref{dm_smooth:alg:smooth} takes $O(1)$ calls to \sqalgtext to obtain estimator $\wh f_t$ and the best action $\wh a_t$. Instead of directly form the action distribution defined in \cref{dm_smooth:eq:sampling_dist}, \cref{dm_smooth:alg:smooth} uses \cref{dm_smooth:alg:reject_sampling} to sample action $a_t \sim P_t$, which takes one call of the sampling oracle \samplealgtext to draw a random action and $O(1)$ calls of the regression oracle \sqalgtext to compute the mean of the Bernoulli random variable. Altogether, \cref{dm_smooth:alg:smooth} has per-round runtime $O(\Tsq + \Tsample)$ and maximum memory $O(\Msq + \Msample)$.
\end{proof}

\subsection{Proofs and Supporting Results for \cref{dm_smooth:sec:adaptive}}
This section is organized as follows. We first prove \cref{dm_smooth:prop:stable} in \cref{dm_smooth:app:stable_base}, then prove \cref{dm_smooth:thm:adaptive} in \cref{dm_smooth:app:adaptive}.

\subsubsection{\pfref{dm_smooth:prop:stable}}
\label{dm_smooth:app:stable_base}
The proof of \cref{dm_smooth:prop:stable} follows similar analysis as in \citet{foster2020adapting}, with minor changes to adapt to our settings.

\propStable*

\begin{proof}[\pfref{dm_smooth:prop:stable}]
   Fix the index $b \in [B]$ of the subroutine. We use shorthands $h = h_b$, $q_t = q_{t,b}$, $\rho_t = \rho_{t,b}$, $\gamma_{t} = \gamma_{t,b}$, and so forth. 
   We also write $Z_{t} = Z_{t,b} \ldef {\ind(I_t = b)}$. Similar to the proof of \cref{dm_smooth:thm:regret}, we use abbreviation $f_t(a) \ldef f(x_t,a)$ for any $f \in \cF$.
   Let $a^\star_t$ denote the action sampled according to the best smoothing kernel within $\cQ_h$ (which could change from round to round). 

    We let $\cE$ denote the good event where the regret guarantee stated in \cref{dm_smooth:assumption:regression_oracle_weighted} (with $\regsq(T) \ldef \regsq(T, T^{-1})$) holds with probability at least  $1- T^{-1}$. Conditioned on this good event, similar to the proof of \cref{dm_smooth:thm:regret} (and following \citet{foster2020adapting}), we decompose the contextual bandit regret as follows.
    \begin{align*}
    & \E \sq*{ \sum_{t = 1}^T \frac{Z_t}{q_{t}} \paren{ f^\star_t(a_t) - f_t^\star(a^\star_t) } } \\
    & = \E \sq*{ \sum_{t = 1}^T \frac{Z_t}{q_{t}} \prn*{ f^\star_t(a_t) - f_t^\star(a^\star_t)  - \frac{\gamma_t}{4}\cdot \prn*{\wh f_t(a_t) - f^\star_t(a_t)}^2} } \\
    & \quad + \E \sq*{ \sum_{t=1}^T \frac{Z_t}{q_t} \cdot \frac{\gamma_t}{4}\cdot \prn*{\wh f_t(a_t) - f^\star_t(a_t)}^2}\nonumber \\
    & \leq \E \sq*{\sum_{t=1}^T \frac{Z_t}{q_t}\cdot \frac{2}{h \gamma_t}}
    + \E \sq*{ \sum_{t=1}^T \frac{Z_t}{q_t} \cdot \frac{\gamma_t}{4}\cdot \prn*{\wh f_t(a_t) - f^\star_t(a_t)}^2}\nonumber \\
    & \leq \E\sq*{\max_{t \in [T]} \gamma_t^{-1}} \cdot \frac{2T}{h} 
    + \E \sq*{ \sum_{t=1}^T \frac{Z_t}{q_t} \cdot \frac{\gamma_t}{4}\cdot \prn*{\wh f_t(a_t) - f^\star_t(a_t)}^2}\nonumber ,
    \end{align*}
    where the bound on the first term follows from \cref{dm_smooth:prop:dec_bound} (the third line, conditioned on $Z_t$). We bound the second term next.
    \begin{align*}
    & \E \sq*{ \sum_{t=1}^T \frac{Z_t}{q_t} \cdot \frac{\gamma_t}{4}\cdot \prn*{\wh f_t(a_t) - f^\star_t(a_t)}^2}\nonumber \\
    & = \frac{1}{4} \cdot \E \Biggl[\sum_{t=1}^T \frac{Z_t}{q_t} \gamma_t \biggl( \prn*{\wh f_t(a_t) - \ell_t(a_t)}^2 -\prn[\Big]{f^\star_t(a_t) - \ell_t(a_t)}^2  \\
    & \quad + 2 \prn[\Big]{\ell_t(a_t) - f^\star_t(a_t)} \cdot \prn[\Big]{\wh f_t(a_t) - f^\star_t(a_t)} \biggr) \Biggr]\\
    & =  \frac{1}{4} \cdot \E \sq*{\sum_{t=1}^T \frac{Z_t}{q_t} \gamma_t \prn*{ \prn*{f_t(a_t) - \ell_t(a_t)}^2 -\prn[\Big]{f^\star_t(a_t) - \ell_t(a_t)}^2  }} \\
    & \leq \frac{1}{4} \cdot \E \sq*{\max_{t\in[T]}\frac{\gamma_t}{q_t}} \cdot \regsq(T), 
    \end{align*}
    where the last line follows from \cref{dm_smooth:assumption:regression_oracle_weighted}. As a result, we have 
    \begin{align*}
	    \regimph(T) \leq 
    \E\sq*{\max_{t \in [T]} \gamma_t^{-1}} \cdot \frac{2T}{h} 
    + \frac{1}{4} \cdot \E \sq*{\max_{t\in[T]}\frac{\gamma_t}{q_t}} \cdot \regsq(T) + O(1),
    \end{align*}
    where the additional $O(1)$ term is to account for the expected regret under event  $\neg \cE$.
    Notice that $\gamma_{t} \ldef \sqrt{8 T/ (h \cdot \rho_{t}
    \cdot \regsq(T))}$, which is non-increasing in $t$; and $\frac{\gamma_t}{q_t} \leq \gamma_t \rho_t$, which is non-decreasing in $t$. Thus, we have  
    \begin{align*}
	    \regimph(T) & \leq 
    \E\sq*{ \gamma_T^{-1}} \cdot \frac{2T}{h} 
    + \frac{1}{4} \cdot \E \sq*{{\gamma_T}{\rho_T}} \cdot \regsq(T) + O(1)\\ 
    & = \E \sq*{\sqrt{\rho_T}} \cdot \sqrt{T \regsq(T)/ 2 h} + \E \sq*{\sqrt{\rho_T}} \sqrt{T \regsq(T)/2h} + O(1)\\
    & \leq \E \sq*{\sqrt{\rho_T}} \cdot  \sqrt{4T \regsq(T)/ h}.
    \end{align*}

    \emph{Computational complexity.} The computational compleity of \cref{dm_smooth:alg:stable} can be analyzed in a similar way as the computational complexity of \cref{dm_smooth:alg:smooth}, except with a \emph{weighted} regression oracle \sqalgtext this time.
\end{proof}

\subsubsection{\pfref{dm_smooth:thm:adaptive}}
\label{dm_smooth:app:adaptive}

We first restate the guarantee of \corral, specialized to our setting.

\begin{theorem}[\citet{agarwal2017corralling}]
    \label{dm_smooth:thm:corral}
    Fix an index $b \in [B]$. Suppose base algorithm $b$ is $(\alpha_b, R_b(T))$-stable with respect to decision space indexed by $b$. If $\alpha_b < 1$, the \corral master algorithm, with learning rate $\eta>0$, guarantees that 
    \begin{align*}
        \E \sq*{\sum_{t=1}^T f^\star(x_t,a_t) - \inf_{Q_t \in \cQ_{h_b}} \E_{a^\star_t \sim Q_t} \sq*{f^\star(x_t, a^\star_t)} } = \wt O \prn*{\frac{B}{\eta} + T \eta + \prn*{R_b(T)}^{\frac{1}{1 - \alpha_b}} \eta^{\frac{\alpha_b}{1-\alpha_b}}}.
    \end{align*}
\end{theorem}

\thmAdaptive*
\begin{proof}[\pfref{dm_smooth:thm:adaptive}]
    We prove the guarantee for any $h^\star \in [1/T,1]$ as the otherwise the bound simply becomes vacuous.
    Recall that we initialize $B = \ceil{\log T}$ \cref{dm_smooth:alg:stable} as base algorithms, each with a fixed smoothness parameter $h_b = 2^{-b}$, for $b \in [B]$. 
    Using such geometric grid guarantees that there exists an $b^\star \in [B]$ such that $h_{b^\star} \leq h^\star \leq 2 h_{b^\star}$. 
    To obtain guarantee with respect to $h^\star$, it suffices to compete with subroutine $b^\star$ since $\cQ_{h^\star} \subseteq \cQ_{h_{b^\star}}$ by definition. \cref{dm_smooth:prop:stable} shows that the base algorithm indexed by $b^\star$ is $(\frac{1}{2}, \sqrt{4 T \regsq(T)/ h_{b^\star}})$-stable. Plugging this result into \cref{dm_smooth:thm:corral} leads to the following guarantee:
    \begin{align*}
	& \E \sq*{\sum_{t=1}^T f^\star(x_t,a_t) - \inf_{Q_t \in \cQ_{h^\star}} \E_{a^\star_t \sim Q_t} \sq*{f^\star(x_t, a^\star_t)} } \\
	&\leq \E \sq*{\sum_{t=1}^T f^\star(x_t,a_t) - \inf_{Q_t \in \cQ_{h_{b^\star}}} \E_{a^\star_t \sim Q_t} \sq*{f^\star(x_t, a^\star_t)} } \\ 
        & = \wt O \prn*{\frac{B}{\eta} + T \eta + \frac{\eta \, T \, \regsq(T)}{h_{b^\star}}} \\
        & = \wt O \prn*{\frac{1}{\eta} + T \eta + \frac{\eta \, T \, \regsq(T)}{h^\star}} .
    \end{align*}

    \emph{Computational complexity.}
    The computational complexities (both runtime and memory) of the \corral master algorithm can be upper bounded by $\wt O(B \cdot \cC)$ where we use $\cC$ denote the complexities of the base algorithms. We have $B = O(\log T)$ in our setting. Thus, directly plugging in the computational complexities of \cref{dm_smooth:alg:stable} leads to the results. 
\end{proof}

\paragraph{Recovering Adaptive Bounds in \citet{krishnamurthy2020contextual}}
\paranewline

\noindent
\label{dm_smooth:app:adaptive_2}
We discuss how our algorithms can also recover the adaptive regret bounds stated in \citet{krishnamurthy2020contextual} (Theorems 4 and 15), i.e.,
\begin{align*}
    \regcbh(T) = \wt O \prn*{ T^{\frac{1}{1+\beta}} (h^\star)^{-\beta} \prn*{\log \abs*{\cF }}^{\frac{\beta}{1+ \beta}}},
\end{align*}
for any $h^\star \in (0,1]$ and $\beta \in [0, 1]$. This line of analysis directly follows the proof used in \citet{krishnamurthy2020contextual}. 

We focus on the case with $\regsq(T) = O \prn{\log \prn{\abs{\cF}T}}$.
For base algorithm (\cref{dm_smooth:alg:stable}), following the analysis used in \citet{krishnamurthy2020contextual}, we have 
    \begin{align*}
	    \regimph(T) &
	    \leq   \min \crl*{T, \E \sq*{\sqrt{\rho_T}} \cdot \sqrt{4 T \regsq(T)/ h}} \\
     & \leq \min \crl*{T,  \sqrt{\E \sq{\rho_T}}\cdot \sqrt{4 T \regsq(T)/ h}} \\
     & = O \prn*{  T^{\frac{1}{1 + \beta}} \cdot \prn*{\E \sq{\rho_T} \regsq(T) / h}^{\frac{\beta}{1 + \beta}} },
    \end{align*}
    where on the first line we combine the regret obtained from \cref{dm_smooth:prop:stable} with a trivial upper bound $T$; on the second line we use the fact that $\sqrt{\cdot}$ is concave; and on the third line we use that fact that $\min \crl{A,B} \leq A^\gamma B^{1 - \gamma}$ for $A,B > 0$ and $\gamma \in [0,1]$ (taking $A= T$, $B = \sqrt{\E \sq{\rho_T} \cdot 4 T \regsq(T) / h}$ and $\gamma = \frac{1 - \beta}{1 + \beta}$). This line of analysis thus shows that \cref{dm_smooth:alg:stable} is $\prn*{\frac{\beta}{1+\beta}, \wt O \prn*{  T^{\frac{1}{1 + \beta}} \cdot \prn*{ \regsq(T) / h}^{\frac{\beta}{1 + \beta}} }}$-stable for any $\beta \in [0,1]$.\footnote{As remarked in \citet{krishnamurthy2020contextual}, the \corral algorithm works with both $\E \sq{\rho_T^\alpha}$ and $\prn*{ \E \sq{\rho_T}}^\alpha$.}
    
    Now following the similar analysis as in the proof of \cref{dm_smooth:thm:adaptive}, and consider $\regsq(T) = O(\log \prn{\abs{\cF}T})$ for the case with a finite set of regression functions, we have 
        \begin{align*}
        \E \sq*{\sum_{t=1}^T f^\star(x_t,a_t) - \inf_{Q_t \in \cQ_{h^\star}} \E_{a^\star_t \sim Q_t} \sq*{f^\star(x_t, a^\star_t)} } 
        = \wt O \prn*{\frac{1}{\eta} + T \eta + T \cdot \prn*{\frac{\log \prn{\abs*{\cF}T} \, \eta}{h^\star}}^\beta},
    \end{align*}
    for any $h^\star \in (0,1]$. Taking $\eta = T^{- \frac{1}{1+\beta}}\cdot \prn*{\log \prn{ \abs*{\cF}T}}^{- \frac{\beta}{1+\beta}}$ recovers the results presented in \citet{krishnamurthy2020contextual}.

\part{\, Model Selection in Sequential Decision Making}
\label{part:model_selection}

\chapter{Bandit Learning with Multiple Best Arms}
\label{chapter:model:multiple}

We study a regret minimization problem with the existence of multiple best/near-optimal arms in the multi-armed bandit setting. We consider the case when the number of arms/actions is comparable or much larger than the time horizon, and make \emph{no} assumptions about the structure of the bandit instance. Our goal is to design algorithms that can automatically adapt to the \emph{unknown} hardness of the problem, i.e., the number of best arms. Our setting captures many modern applications of bandit algorithms where the action space is enormous and the information about the underlying instance/structure is unavailable. We first propose an adaptive algorithm that is agnostic to the hardness level and theoretically derive its regret bound. We then prove a lower bound for our problem setting, which indicates: (1) no algorithm can be minimax optimal simultaneously over all hardness levels; and (2) our algorithm achieves a rate function that is Pareto optimal. With additional knowledge of the expected reward of the best arm, we propose another adaptive algorithm that is minimax optimal, up to polylog factors, over \emph{all} hardness levels. Experimental results confirm our theoretical guarantees and show advantages of our algorithms over the previous state-of-the-art.

\section{Introduction}
\label{ms_multi:section:intro}

Multi-armed bandit problems describe exploration-exploitation trade-offs in sequential decision making. Most existing bandit algorithms tend to provide regret guarantees when the number of available arms/actions is smaller than the time horizon. In modern applications of bandit algorithm, however, the action space is usually comparable or even much larger than the allowed time horizon so that many existing bandit algorithms cannot even complete their initial exploration phases. Consider a problem of personalized recommendations, for example. For most users, the total number of movies, or even the amount of sub-categories, far exceeds the number of times they visit a recommendation site.  Similarly, the enormous amount of user-generated content on YouTube and Twitter makes it increasingly challenging to make optimal recommendations. The tension between a very large action space and a limited time horizon poses a realistic problem in which deploying algorithms that converge to an optimal solution over an asymptotically long time horizon \emph{do not} give satisfying results. There is a need to design algorithms that can exploit the highest possible reward within a \emph{limited} time horizon. Past work has partially addressed this challenge.  The quantile regret proposed in \cite{chaudhuri2018quantile} to calculate regret with respect to an satisfactory action rather than the best one. The discounted regret analyzed in \cite{ryzhov2012knowledge, russo2018satisficing} is used to emphasize short time horizon performance. Other existing works consider the extreme case when the number of actions is indeed infinite, and tackle such problems with one of two main assumptions: (1) the discovery of  a near-optimal/best arm follows some probability measure with \emph{known}  parameters \cite{berry1997bandit, wang2009algorithms,aziz2018pure, ghalme2020ballooning}; (2) the existence of a \emph{smooth} function represents the mean-payoff over a continuous subset \cite{agrawal1995continuum, kleinberg2005nearly, kleinberg2008multi, bubeck2011x, locatelli2018adaptivity, hadiji2019polynomial}. However, in many situations, neither assumption may be realistic. We make minimal assumptions in this chapter. We study the regret minimization problem over a time horizon $T$, which might be unknown, with respect to a bandit instance with $n$ total arms, out of which $m$ are best/near-optimal arms. We emphasize that the allowed time horizon and the given bandit instance should be viewed as features of \emph{one} problem and together they indicate an intrinsic hardness level. We consider the case when the number of arms $n$ is comparable or larger than the time horizon $T$ so that no standard algorithm provides satisfying result. Our goal is to design algorithms that could adapt to the \emph{unknown} $m$ and achieve optimal regret.

\subsection{Contributions and Organization}
We make the following contributions. In \cref{ms_multi:section:problem_setting}, we formally define the regret minimization problem that represents the tension between a very large action space and a limited time horizon; and capture the hardness level in terms of the number of best arms. We provide an adaptive algorithm that is agnostic to the \emph{unknown} number of best arms in \cref{ms_multi:section:adaptive}, and theoretically derive its regret bound. In \cref{ms_multi:section:lower_bound_pareto_optimal}, we prove a lower bound for our problem setting that indicates that there is no algorithm that can be optimal simultaneously over all hardness levels. Our lower bound also shows that our algorithm provided in \cref{ms_multi:section:adaptive} is Pareto optimal. With additional knowledge of the expected reward of the best arm, in \cref{ms_multi:section:extra_info}, we provide an algorithm that achieves the non-adaptive minimax optimal regret, up to polylog factors, without the knowledge of the number of best arms. Experiments conducted in \cref{ms_multi:section:experiment} confirm our theoretical guarantees and show advantages of our algorithms over previous state-of-the-art. We conclude this chapter in \cref{ms_multi:section:conclusion}. Most of the proofs are deferred to the Appendix due to lack of space.

\subsection{Additional Related Work}

\textbf{Time sensitivity and large action space.} As bandit models are getting much more complex, usually with large or infinite action spaces, researchers have begun to pay attention to tradeoffs between regret and time horizons when deploying such models. \cite{deshpande2012linear} study a linear bandit problem with ultra-high dimension, and provide algorithms that, under various assumptions, can achieve good reward within short time horizon. \cite{russo2018satisficing} also take time horizon into account and model time preference by analyzing a discounted regret. \cite{chaudhuri2018quantile} consider a quantile regret minimization problem where they define their regret with respect to expected reward ranked at $(1-\rho)$-th quantile. One could easily transfer their problem to our setting; however, their regret guarantee is sub-optimal. \cite{katz2019true, aziz2018pure} also consider the problem with $m$ best/near-optimal arms with no other assumptions, but they focus on the pure exploration setting; \cite{aziz2018pure} additionally requires the knowledge of $m$. Another line of research considers the extreme case when the number arms is infinite, but with some \emph{known} regularities. \cite{berry1997bandit} proposes an algorithm with a minimax optimality guarantee under the situation where the reward of each arm follows \emph{strictly} Bernoulli distribution; \cite{teytaud:inria-00173263} provides an anytime algorithm that works under the same assumption. \cite{wang2009algorithms} relaxes the assumption on Bernoulli reward distribution, however, some other parameters are assumed to be known in their setting.

\textbf{Continuum-armed bandit.} Many papers also study bandit problems with continuous action spaces, where they embed each arm $x$ into a bounded subset ${\mathcal X} \subseteq \R^d$ and assume there exists a smooth function $f$ governing the mean-payoff for each arm. This setting is firstly introduced by \cite{agrawal1995continuum}. When the smoothness parameters are known to the learner or under various assumptions, there exists algorithms \cite{kleinberg2005nearly, kleinberg2008multi, bubeck2011x} with near-optimal regret guarantees. When the smoothness parameters are unknown, however, 
\cite{locatelli2018adaptivity} proves a lower bound indicating no strategy can be optimal simultaneously over all smoothness classes; under extra information, they provide adaptive algorithms with near-optimal regret guarantees. Although achieving optimal regret for all settings is impossible, \cite{hadiji2019polynomial} design adaptive algorithms and prove that they are Pareto optimal. Our algorithms are mainly inspired by the ones in \cite{hadiji2019polynomial, locatelli2018adaptivity}. A closely related line of work \cite{valko2013stochastic, grill2015black, bartlett2018simple, shang2019general} aims at minimizing simple regret in the continuum-armed bandit setting.

\textbf{Adaptivity to unknown parameters.} \cite{bubeck2011lipschitz} argues the awareness of regularity is flawed and one should design algorithms that can \emph{adapt} to the unknown environment. In situations where the goal is pure exploration or simple regret minimization, \cite{katz2019true, valko2013stochastic, grill2015black, bartlett2018simple, shang2019general} achieve near-optimal guarantees with unknown regularity because their objectives trade-off exploitation in favor of exploration. In the case of cumulative regret minimization, however, \cite{locatelli2018adaptivity} shows no strategy can be optimal simultaneously over all smoothness classes. In special situations or under extra information, \cite{bubeck2011lipschitz, bull2015adaptive, locatelli2018adaptivity} provide algorithms that adapt in different ways. \cite{hadiji2019polynomial} borrows the concept of Pareto optimality from economics and provide algorithms with rate functions that are Pareto optimal. Adaptivity is studied in statistics as well: in some cases, only additional logarithmic factors are required  \cite{lepskii1991problem, birge1997model}; in others, however, there exists an additional polynomial cost of adaptation \cite{cai2005adaptive}.

\section{Problem Setting}
\label{ms_multi:section:problem_setting}

We consider the multi-armed bandit instance $\underline{\nu} = (\nu_1, \dots, \nu_n)$ with $n$ probability distributions with means $\mu_i = \E_{X \sim \nu_i} [X] \in [0, 1]$. Let $\mu_\star = \max_{i \in [n]} \{\mu_i\}$ be the highest mean and $S_{\star} = \{i \in [n]: \mu_i = \mu_\star\}$ denote the subset of best arms. The cardinality $| S_\star|  = m$ is \emph{unknown} to the learner. {We could also generalize our setting to $S^{\prime}_\star = \{ i \in [n]: \mu_i \geq \mu_\star - \epsilon(T) \}$ with unknown $|S^\prime_\star|$ (i.e., situations where there is an unknown number of near-optimal arms). Setting $\epsilon$ to be dependent on $T$ is to avoid an additive term linear in $T$, e.g., $\epsilon \leq 1/\sqrt{T} \Rightarrow \epsilon T \leq \sqrt{T}$. All theoretical results and algorithms presented in this chapter are applicable to this generalized setting with minor modifications. For ease of exposition, we focus on the case with multiple best arms throughout this chapter.} At each time step $t \in [T]$, the algorithm/learner selects an action $A_t \in [n]$ and receives an independent reward $X_t \sim \nu_{A_t}$. We assume that $X_t - \mu_{A_t}$ is $(1/2)$-sub-Gaussian conditioned on $A_t$.\footnote{We say a random variable $X$ is $\sigma$-sub-Gaussian if $\E[\exp(\lambda X)] \leq \exp(\sigma^2 \lambda^2/2)$ for all $\lambda \in \R$.} We measure the success of an algorithm through the expected cumulative (pseudo) regret:
\begin{align}
	R_T  = T \cdot \mu_\star -  \E \left[ \sum_{t=1}^T \mu_{A_t}\right].\nonumber
\end{align}

We use ${\mathcal R}(T, n, m)$ to denote the set of regret minimization problems with allowed time horizon $T$ and any bandit instance $\underline{\nu}$ with $n$ total arms and $m$ best arms.\footnote{Our setting could be generalized to the case with infinite arms: one can consider embedding arms into an arm space ${\mathcal X}$ and let $p$ be the probability that an arm sampled uniformly at random is (near-) optimal. $1/p$ will then serve a similar role as $n/m$ does in the original definition.} We emphasize that $T$ is part of the problem instance. We are particularly interested in the case when $n$ is comparable or even larger than $T$, which captures many modern applications where the available action space far exceeds the allowed time horizon. Although learning algorithms may not be able to pull each arm once, one should notice that the true/intrinsic hardness level of the problem could be viewed as $n/m$: selecting a subset uniformly at random with cardinality ${\Theta}(n/m)$ guarantees, with constant probability, the access to at least one best arm; but of course it is impossible to do this without knowing $m$. We quantify the \emph{intrinsic} hardness level over a set of regret minimization problems ${\mathcal R}(T, n, m)$ as
\begin{align}
    \psi({\mathcal R}(T, n, m)) = \inf \{ \alpha \geq 0: n/m \leq 2T^{\alpha}  \}, \nonumber
\end{align}
where the constant $2$ in front of $T^\alpha$ is added to avoid otherwise the trivial case with all best arms when the infimum is $0$. $\psi({\mathcal R}(T, n, m))$ is used here as it captures the \emph{minimax} optimal regret over the set of regret minimization problem ${\mathcal R}(T, n, m)$, as explained later in our review of the \moss algorithm and the lower bound. As smaller $\psi({\mathcal R}(T, n, m))$ indicates easier problems, we then define the family of regret minimization problems with hardness level at most $\alpha$ as 
\begin{align}
	{\mathcal H}_T(\alpha) = \{ \cup {\mathcal R}(T, n, m) : \psi({\mathcal R}(T, n, m))\leq \alpha \}, \nonumber
\end{align}
with $\alpha \in [0, 1]$. Although $T$ is necessary to define a regret minimization problem, we actually encode the hardness level into a single parameter $\alpha$, which captures the \emph{tension} between the complexity of bandit instance at hand and the allowed time horizon $T$: problems with different time horizons but the same $\alpha$ are equally difficult in terms of the achievable minimax regret (the exponent of $T$). We thus mainly study problems with $T$ large enough so that we could mainly focus on the polynomial terms of $T$. We are interested in designing algorithms with \emph{minimax} guarantees over ${\mathcal H}_T(\alpha)$, but \emph{without} the knowledge of $\alpha$.

 \textbf{\moss and upper bound.} In the classical setting, \moss, proposed by \cite{audibert2009minimax} and further generalized to the sub-Gaussian case \cite{lattimore2020bandit} and improved in terms of constant factors \cite{garivier2018kl}, achieves the minimax optimal regret. In this chapter, we will use \moss as a subroutine with regret upper bound $O (\sqrt{nT} )$ when $T \geq n$. For any problem in ${\mathcal H}_T(\alpha)$ with \emph{known} $\alpha$, one could run \moss on a subset selected uniformly at random with cardinality $\widetilde{O}(T^{\alpha})$ and achieve regret $\widetilde{O}(T^{(1+\alpha)/2})$. 
 
 \textbf{Lower bound.} The lower bound $\Omega(\sqrt{nT})$ in the classical setting does not work for our setting as its proof heavily relies on the existence of single best arm \cite{lattimore2020bandit}. However, for problems in ${\mathcal H}_T(\alpha)$, we do have a matching lower bound $\Omega(T^{(1+\alpha)/2})$ as one could always apply the standard lower bound on an bandit instance with $n = \lfloor T^\alpha \rfloor$ and $m=1$. For general value of $m$, a lower bound of the order $\Omega(\sqrt{T(n-m)/m})=\Omega(T^{(1+\alpha)/2})$ for the $m$-best arms case could be obtained following similar analysis in Chapter 15 of \cite{lattimore2020bandit}.

Although $\log T$ may appear in our bounds, throughout this chapter, we focus on problems with $T \geq 2$ as otherwise the bound is trivial.

\section{An Adaptive Algorithm}
\label{ms_multi:section:adaptive}

\cref{ms_multi:alg:regret_multi} takes time horizon $T$ and a user-specified $\beta \in [1/2, 1)$ as input, and it is mainly inspired by \cite{hadiji2019polynomial}. \cref{ms_multi:alg:regret_multi} operates in iterations with geometrically-increasing length (roughly) $\Delta T_i = 2^{p+i}$ with $p = \lceil \log_2 T^\beta \rceil$. At each iteration $i$, it restarts \moss on a set $S_i$ consisting of $K_i = 2^{p+2-i}$ real arms selected uniformly at random \emph{plus} a set of ``virtual'' \emph{mixture-arms} (one from each of the $1\leq j<i$ previous iterations, none if $i=1$). The mixture-arms are constructed as follows. After each iteration $i$, let $\widehat{p}_i$ denote the vector of empirical sampling frequencies of the arms in that iteration (i.e., the $k$-th element of $\widehat{p}_i$ is the number of times arm $k$, including all previously constructed mixture-arms, was sampled in iteration $i$ divided by the total number of samples $\Delta T_i$).  The mixture-arm for iteration $i$ is the $\widehat{p}_i$-mixture of the arms, denoted by $\widetilde{\nu}_i$.  When \moss samples from $\widetilde{\nu}_i$ it first draws $i_t \sim \widehat{p}_i$, then draws a sample from the corresponding arm $\nu_{i_t}$ (or $\widetilde{\nu}_{i_t}$).  The mixture-arms provide a convenient summary of the information gained in the previous iterations, which is key to our theoretical analysis. Although our algorithm is working on fewer regular arms in later iterations, information summarized in mixture-arms is good enough to provide guarantees. We name our algorithm \mossPlus as it restarts \moss at each iteration with past information summarized in mixture-arms. We provide an anytime version of \cref{ms_multi:alg:regret_multi} in \cref{ms_multi:appendix:anytime_multi} via the standard doubling trick.

\begin{algorithm}[]
	\caption{\mossPlus}
	\label{ms_multi:alg:regret_multi} 
	\renewcommand{\algorithmicrequire}{\textbf{Input:}}
	\renewcommand{\algorithmicensure}{\textbf{Output:}}
	\begin{algorithmic}[1]
		\REQUIRE Time horizon $T$ and user-specified parameter $\beta \in [1/2, 1)$.
		\STATE \textbf{Set:} $p = \lceil \log_2 T^\beta \rceil$, $K_i = 2^{p+2-i}$ and $\Delta T_i = \min \{2^{p + i}, T\}$.
		\FOR {$i = 1, \dots, p$}
		\STATE Run \moss on a subset of arms $S_i$ for $\Delta T_i$ rounds. $S_i$ contains $K_i$ real arms selected uniformly at random \emph{and} the set of virtual mixture-arms from previous iterations, i.e., $\{\widetilde{\nu}_j\}_{j < i}$.
		\STATE Construct a virtual mixture-arm $\widetilde{\nu}_i$ based on empirical sampling frequencies of \moss above.
		\ENDFOR 
	\end{algorithmic}
\end{algorithm}

\subsection{Analysis and Discussion}
\label{ms_multi:section:adaptive_analysis}

We use $\mu_{S} = \max_{\nu \in S} \{\E_{X \sim \nu}[X]\}$ to denote the highest expected reward over a set of distributions/arms $S$. For any algorithm that only works on $S$, we can decompose the regret into approximation error and learning error:
\begin{align}
\label{ms_multi:eq:regret_decomposition}
R_T &= \underbrace{ \E \left[ T \cdot (\mu_\star - \mu_S) \right]}_{\text{expected approximation error due to the selection of $S$}} \\
    & \quad + \underbrace{\E \left[ T\cdot \mu_S -  \sum_{t=1}^T \mu_{A_t}\right]}_{\text{expected learning error due to the  sampling rule $\{A_t\}_{t=1}^T$}}. \nonumber
\end{align}

This type of regret decomposition was previously used in \cite{kleinberg2005nearly, auer2007improved, hadiji2019polynomial} to deal with the continuum-armed bandit problem. We consider here a probabilistic version, with randomness in the selection of $S$, for the classical setting.

The main idea behind providing guarantees for \mossPlus is to decompose its regret at each iteration, using \cref{ms_multi:eq:regret_decomposition}, and then bound the expected approximation error and learning error separately. The expected learning error at each iteration could always be controlled as $\widetilde{O}(T^\beta)$ thanks to regret guarantees for \moss and specifically chosen parameters $p$, $K_i$, $\Delta T_i$. Let $i_\star$ be the largest integer such that $K_i \geq 2T^\alpha \log \sqrt{T}$ still holds. The expected approximation error in iteration $i \leq i_\star$ could be upper bounded by $\sqrt{T}$ following an analysis on hypergeometric distribution. As a result, the expected regret in iteration $i \leq i_\star$ is $\widetilde{O}(T^\beta)$. Since the mixture-arm $\widetilde{\nu}_{i\star}$ is included in all following iterations, we could further bound the expected approximation error in iteration $i > i_\star$ by $\widetilde{O}(T^{1+\alpha-\beta})$ after a careful analysis on $\Delta T_i/ \Delta T_{i_\star}$. This intuition is formally stated and proved in \cref{ms_multi:thm:regret_multi}.

\begin{restatable}{theorem}{regretMulti}
	\label{ms_multi:thm:regret_multi}
	Run \mossPlus with time horizon $T$ and an user-specified parameter $\beta \in [1/2, 1)$ leads to the following regret upper bound:
	\begin{align}
	\sup_{\omega \in {\mathcal H}_T(\alpha)} R_T \leq C \, (\log_2 T)^{5/2} \cdot T^{\min\{\max \{\beta, 1+\alpha - \beta \},1\}}, \nonumber
	\end{align} 
	where $C$ is a universal constant.
\end{restatable}

\begin{remark}
    We primarily focus on the polynomial terms in $T$ when deriving the bound, but put no effort in optimizing the polylog term. The $5/2$ exponent of $\log_2 T$ might be tightened as well.
\end{remark}

The theoretical guarantee is closely related to the user-specified parameter $\beta$: when $\beta > \alpha$, we suffer a multiplicative cost of adaptation  $\widetilde{O}(T^{|(2\beta - \alpha  - 1)/2|})$, with $\beta = (1+\alpha)/2$ hitting the sweet spot, comparing to non-adaptive minimax regret; when $\beta \leq \alpha$, there is essentially no guarantees. One may hope to improve this result. However, our analysis in \cref{ms_multi:section:lower_bound_pareto_optimal} indicates: (1) achieving minimax optimal regret for all settings simultaneously is \emph{impossible}; and (2) the rate function achieved by \mossPlus is already \emph{Pareto optimal}.

\section{Lower Bound and Pareto Optimality}
\label{ms_multi:section:lower_bound_pareto_optimal}

\subsection{Lower Bound}
\label{ms_multi:section:lower_bound}
In this section, we show that designing algorithms with the non-adaptive minimax optimal guarantee over all values of $\alpha$ is impossible. We first state the result in the following general theorem.

\begin{restatable}{theorem}{lowerBound}
	\label{ms_multi:thm:lower_bound}
	For any $0 \leq \alpha^\prime < \alpha \leq 1$, assume $T^{\alpha} \leq B$ and $\lfloor T^\alpha  \rfloor -1 \geq \max\{ T^\alpha/4, 2\}$. If an algorithm is such that $\sup_{\omega \in {\mathcal H}_T(\alpha^\prime)} R_T \leq B$, then the regret of this algorithm is lower bounded on ${\mathcal H}_T(\alpha)$:
	\begin{align}
	\sup_{\omega \in {\mathcal H}_T(\alpha)} R_T \geq 2^{-10} T^{1+\alpha} B^{-1}. \label{ms_multi:eq:lower_bound}
	\end{align}
\end{restatable}

To give an interpretation of \cref{ms_multi:thm:lower_bound}, we consider any algorithm/policy $\pi$ together with regret minimization problems ${\mathcal H}_T(\alpha^\prime)$ and ${\mathcal H}_T(\alpha)$ satisfying corresponding requirements. On one hand, if algorithm $\pi$ achieves a regret that is order-wise larger than $\widetilde{O}(T^{(1+\alpha^\prime)/2})$ over ${\mathcal H}_T(\alpha^\prime)$, it is already not minimax optimal for ${\mathcal H}_T(\alpha^\prime)$. Now suppose $\pi$ achieves a near-optimal regret, i.e., $\widetilde{O}(T^{(1+\alpha^\prime)/2})$, over ${\mathcal H}_T(\alpha^\prime)$; then, according to \cref{ms_multi:eq:lower_bound}, $\pi$ must incur a regret of order at least $\widetilde{\Omega}(T^{1/2 + \alpha - \alpha^\prime/2})$ on one problem in ${\mathcal H}_T(\alpha^\prime)$. This, on the other hand, makes algorithm $\pi$ strictly sub-optimal over ${\mathcal H}_T(\alpha)$.

\subsection{Pareto Optimality}
\label{ms_multi:section:pareto_optimality}

We capture the performance of any algorithm by its dependence on polynomial terms of $T$ in the asymptotic sense. Note that the hardness level of a problem is encoded in $\alpha$.

\begin{definition}
	\label{ms_multi:def:rate}
	Let $\theta:[0,1] \rightarrow [0, 1]$ denote a non-decreasing function. An algorithm achieves the rate function $\theta$ if 
	\begin{align}
		\forall \epsilon > 0, \forall \alpha \in [0, 1], \quad \limsup_{T \rightarrow \infty} \frac{\sup_{\omega \in {\mathcal H}_T(\alpha)} R_T}{T^{\theta(\alpha)+\epsilon}} < + \infty. \nonumber 
	\end{align}
\end{definition}

Recall that a function $\theta^\prime$ is strictly smaller than another function $\theta$ in pointwise order if ${\theta^\prime}(\alpha) \leq \theta(\alpha)$ for all $\alpha$ and ${\theta^\prime}(\alpha_0) < \theta(\alpha_0)$ for at least one value of $\alpha_0$. As there may not always exist a pointwise ordering over rate functions, following \cite{hadiji2019polynomial}, we consider the notion of Pareto optimality over rate functions achieved by some algorithms.

\begin{definition}
	\label{ms_multi:def:pareto}
	A rate function $\theta$ is Pareto optimal if it is achieved by an algorithm, and there is no other algorithm achieving a strictly smaller rate function ${\theta^\prime}$ in pointwise order. An algorithm is Pareto optimal if it achieves a Pareto optimal rate function.
\end{definition}

Combining the results in \cref{ms_multi:thm:regret_multi} and \cref{ms_multi:thm:lower_bound} with above definitions, we could further obtain the following result in \cref{ms_multi:thm:pareto}.

\begin{figure}
    \centering
    \includegraphics[width=.8\textwidth]{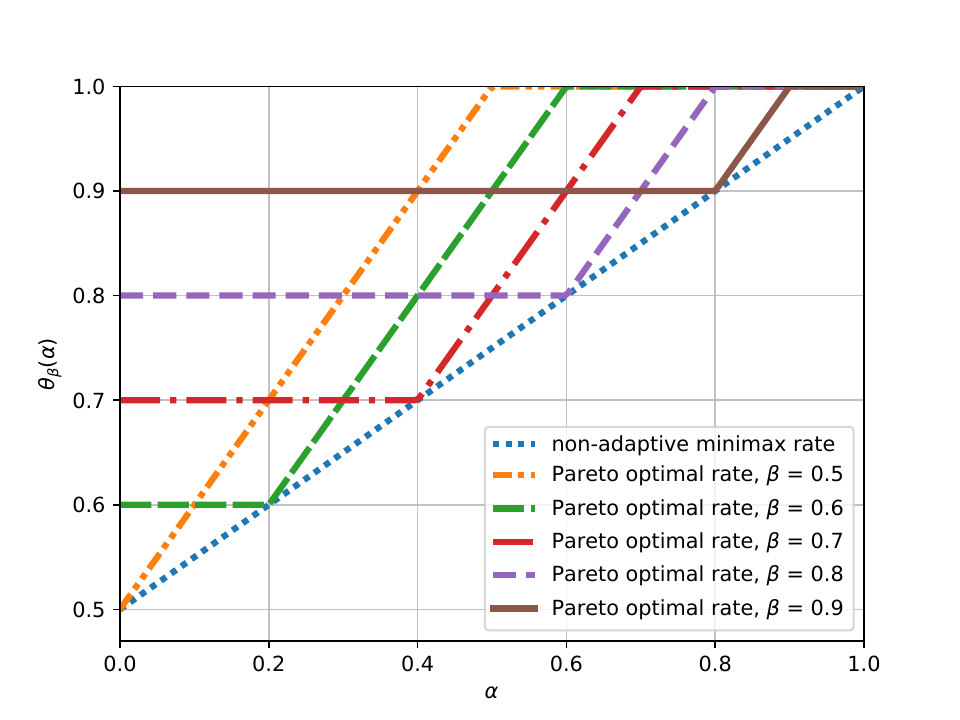}
    \caption{Pareto optimal rates for bandit learning with multiple best arms.}
    \label{ms_multi:fig:pareto}
\end{figure}

\begin{restatable}{theorem}{pareto}
	\label{ms_multi:thm:pareto}
	The rate function achieved by \mossPlus with any $ \beta \in [1/2, 1)$, i.e.,
	\begin{align}
	\theta_{\beta}: \alpha \mapsto \min \{ \max \{ \beta, 1 + \alpha - \beta\}, 1\}, \label{ms_multi:eq:admissible_rate}
	\end{align}
	is Pareto optimal.
\end{restatable}

\cref{ms_multi:fig:pareto} provides an illustration of the rate functions achieved by \mossPlus with different $\beta$ as input, as well as the non-adaptive minimax optimal rate.

\begin{remark}
One should notice that the naive algorithm running \moss on a subset selected uniformly at random with cardinality $\widetilde{O}(T^{\beta^\prime})$ is not Pareto optimal, since running \mossPlus with $\beta = (1+\beta^{\prime})/2$ leads to a strictly smaller rate function. The algorithm provided in \cite{chaudhuri2018quantile}, if transferred to our setting and allowing time horizon dependent quantile, is not Pareto optimal as well since it corresponds to the rate function $\theta(\alpha) = \max\{2.89 \, \alpha, 0.674\}$.

\end{remark}

\section{Learning with Extra Information}
\label{ms_multi:section:extra_info}
Although previous \cref{ms_multi:section:lower_bound_pareto_optimal} gives negative results on designing algorithms that could optimally adapt to all settings, one could actually design such an algorithm \emph{with} extra information. In this section, we provide an algorithm that takes the expected reward of the best arm $\mu_{\star}$ (or an estimated one with error up to $1/\sqrt{T}$) as extra information, and achieves near minimax optimal regret over all settings simultaneously. Our algorithm is mainly inspired by \cite{locatelli2018adaptivity}.

\subsection{Algorithm}
\label{ms_multi:section:extra_algo}

We name our \cref{ms_multi:alg:extra_info} \algParallel as it maintains $\lceil \log T \rceil$ instances of subroutine, i.e., \cref{ms_multi:alg:moss_subroutine}, in parallel. Each subroutine $\sr_i$ is initialized with time horizon $T$ and hardness level $\alpha_i = i/\lceil \log T \rceil$. We use $T_{i, t}$ to denote the number of samples allocated to $\sr_i$ up to time $t$, and represent its empirical regret at time $t$ as $\widehat{R}_{i,t} = T_{i, t}\cdot  \mu_{\star} - \sum_{t=1}^{T_{i,t}} X_{i, t}$ with $X_{i,t} \sim \nu_{A_{i,t}}$ being the $t$-th empirical reward obtained \emph{by} $\sr_i$ and $A_{i,t}$ being the index of the $t$-th arm pulled \emph{by} $\sr_i$.

\begin{algorithm}[]
	\caption{\moss Subroutine ($\sr$)}
	\label{ms_multi:alg:moss_subroutine} 
	\renewcommand{\algorithmicrequire}{\textbf{Input:}}
	\renewcommand{\algorithmicensure}{\textbf{Output:}}
	\begin{algorithmic}[1]
		\REQUIRE Time horizon $T$ and hardness level $\alpha$.
		\STATE Select a subset of arms $S_{\alpha}$ uniformly at random with $|S_{\alpha}| = \lceil 2T^{\alpha} \log \sqrt{T} \rceil$ and run \moss on $S_{\alpha}$.
	\end{algorithmic}
\end{algorithm}

\begin{algorithm}[]
	\caption{\algParallel}
	\label{ms_multi:alg:extra_info} 
	\renewcommand{\algorithmicrequire}{\textbf{Input:}}
	\renewcommand{\algorithmicensure}{\textbf{Output:}}
	\begin{algorithmic}[1]
		\REQUIRE Time horizon $T$ and the optimal reward $\mu_\star$.
		\STATE \textbf{set:} $p = \lceil \log T \rceil$, $\Delta = \lceil \sqrt{T} \rceil$ and $t = 0$.
		\FOR {$i = 1, \dots, p$}
		\STATE Set $\alpha_i = i/p$, initialize $\sr_i$ with $\alpha_i$, $T$; set $T_{i, t}=0$, and $\widehat{R}_{i, t} = 0$.
		\ENDFOR
		\FOR {$i = 1, \dots, \Delta-1$}
		\STATE  Select $k = \argmin_{i \in [p]} \widehat{R}_{i, t}$ and run $\sr_k$ for $\Delta$ rounds.
		\STATE Update $ T_{k, t}   = T_{k, t} + \Delta , \,
		\widehat{R}_{k, t}  = T_{k, t} \cdot \mu_\star - \sum_{t=1}^{T_{k, t}} X_{k, t} , \,
		t  = t + \Delta$.
		\ENDFOR 
	\end{algorithmic}
\end{algorithm}

\algParallel operates in iterations of length $\lceil \sqrt{T} \rceil$. At the beginning of each iteration, i.e., at time $t = i \cdot \lceil \sqrt{T} \rceil$ for $i \in \{0\} \cup [\lceil \sqrt{T} \rceil - 1]$, \algParallel first selects the subroutine with the lowest (breaking ties arbitrarily) empirical regret so far, i.e., $k = \argmin_{i \in [\lceil \log T \rceil]} \widehat{R}_{i, t}$; it then \emph{resumes} the learning process of $\sr_k$, from where it halted, for another $\lceil \sqrt{T} \rceil$ more pulls. All the information is updated at the end of that iteration. An anytime version of \cref{ms_multi:alg:extra_info} is provided in \cref{ms_multi:appendix_anytime_extra}.

\subsection{Analysis}
\label{ms_multi:section:extra_analysis}

As \algParallel discretizes the hardness parameter over a grid with interval $1/\lceil \log T \rceil$, we first show that running the best subroutine alone leads to regret $\widetilde{O} (T^{(1+\alpha)/2})$.

\begin{restatable}{lemma}{subroutineBest}
	\label{ms_multi:lm:subroutine_best}
	Suppose $\alpha$ is the true hardness parameter and $\alpha_i - 1/\lceil \log T \rceil < \alpha \leq \alpha_i$, run \cref{ms_multi:alg:moss_subroutine} with time horizon $T$ and $\alpha_i$ leads to the following regret bound:
	\begin{align}
	\sup_{\omega \in {\mathcal H}_T(\alpha)} R_T \leq C\, \log T \cdot T^{(1+\alpha )/2},\nonumber
	\end{align}
	where $C$ is a universal constant.
\end{restatable}

Since \algParallel always allocates new samples to the subroutine with the lowest empirical regret so far, we know that the regret of every subroutine should be roughly of the same order at time $T$. In particular, all subroutines should achieve regret $\widetilde{O}(T^{(1+\alpha)/2})$, as the best subroutine does. \algParallel then achieves the non-adaptive minimax optimal regret, up to polylog factors, \emph{without} knowing the true hardness level $\alpha$.

\begin{restatable}{theorem}{extraInfo}
	\label{ms_multi:thm:extra_info}
	For any $\alpha \in [0, 1]$ unknown to the learner, run \algParallel with time horizon $T$ and optimal expected reward $\mu_\star$ leads to the following regret upper bound:
	\begin{align}
	\sup_{\omega \in {\mathcal H}_T(\alpha)} R_T \leq C \, \left( \log T \right)^{2} T^{(1+ \alpha )/ 2}, \nonumber
	\end{align}
	where $C$ is a universal constant.
\end{restatable}

\section{Experiments}
\label{ms_multi:section:experiment}
We conduct three experiments to compare our algorithms with baselines. In \cref{ms_multi:sec:experiment_hardness}, we compare the performance of each algorithm on problems with varying hardness levels. We examine how the regret curve of each algorithm increases on synthetic and real-world datasets in \cref{ms_multi:sec:experiment_real_time} and \cref{ms_multi:sec:experiment_real_data}, respectively.

We first introduce the nomenclature of the algorithms. We use \moss to denote the standard \moss algorithm; and \mossOracle to denote \cref{ms_multi:alg:moss_subroutine} with \emph{known} $\alpha$. \quantile represents the algorithm (QRM2) proposed by \cite{chaudhuri2018quantile} to minimize the regret with respect to the $(1-\rho)$-th quantile of means among arms, without the knowledge of $\rho$. One could easily transfer \quantile to our settings with top-$\rho$ fraction of arms treated as best arms. As suggested in \cite{chaudhuri2018quantile}, we reuse the statistics obtained in previous iterations of \quantile to improve its sample efficiency. We use \mossPlus to represent the vanilla version of \cref{ms_multi:alg:regret_multi}; and use \mossPlusEmp to represent an empirical version such that: (1) \mossPlusEmp reuse statistics obtained in previous round, as did in \quantile; and (2) instead of selecting $K_i$ real arms uniformly at random at the $i$-th iteration, \mossPlusEmp selects $K_i$ arms with the highest empirical mean for $i > 1$. We choose $\beta = 0.5$ for \mossPlus and \mossPlusEmp in all experiments.\footnote{Increasing $\beta$ generally leads to worse performance on problems with small $\alpha$ but better performance on problems with large $\alpha$.} All results are averaged over 100 experiments. Shaded area represents 0.5 standard deviation for each algorithm.

\subsection{Adaptivity to Hardness Level}
\label{ms_multi:sec:experiment_hardness}

\begin{figure}[h]
     \centering
     \subfloat[]{\includegraphics[width=.5\textwidth]{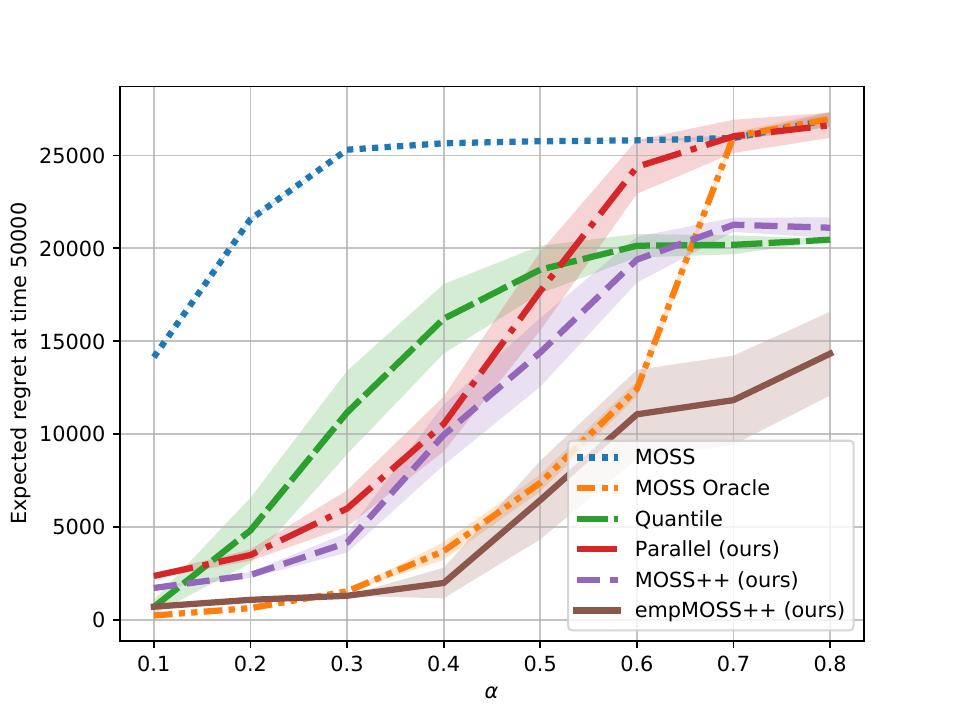}}
     \subfloat[]{\includegraphics[width=.5\textwidth]{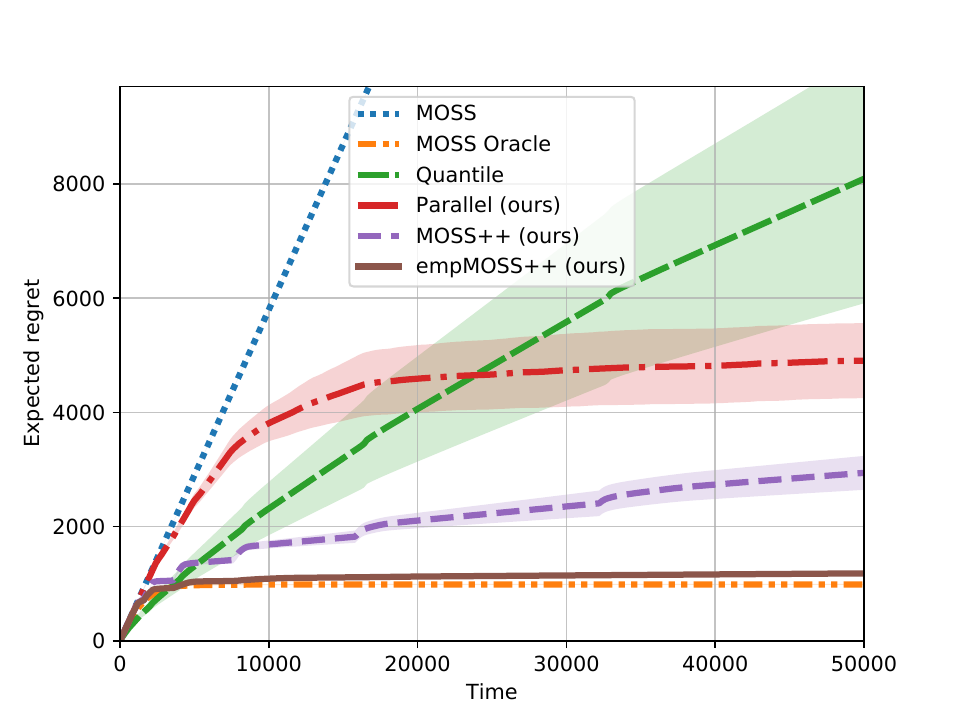}}
     \caption{Experiments on synthetic dataset. (a) Comparison of regret with varying hardness level $\alpha$ (b) Comparison of progressive regret curve with $\alpha = 0.25$.}
	\label{ms_multi:fig:synthetic}
\end{figure}

We compare our algorithms with baselines on regret minimization problems with different hardness levels. For this experiment, we generate best arms with expected reward 0.9 and sub-optimal arms with expected reward evenly distributed among $\{0.1, 0.2, 0.3, 0.4, 0.5\}$. All arms follow Bernoulli distribution. We set the time horizon to $T = 50000$ and consider the total number of arms $n = 20000$. We vary $\alpha$ from 0.1 to 0.8 (with interval 0.1) to control the number of best arms $m = \lceil n/ 2T^\alpha \rceil$ and thus the hardness level. In \cref{ms_multi:fig:synthetic}(a), the regret of any algorithm gets larger as $\alpha$ increases, which is expected. \moss does not provide satisfying performance due to the large action space and the relatively small time horizon. Although implemented in an anytime fashion, \quantile could be roughly viewed as an algorithm that runs \moss on a subset selected uniformly at random with cardinality $T^{0.347}$. \quantile displays good performance when $\alpha = 0.1$, but suffers regret much worse than \mossPlus and \mossPlusEmp when $\alpha$ gets larger. Note that the regret curve of \quantile gets flattened at $20000$ is expected: it simply learns the best sub-optimal arm and suffers a regret $50000 \times (0.9 -0.5)$. Although \algParallel enjoys near minimax optimal regret, the regret it suffers from is the summation of 11 subroutines, which hurts its empirical performance. \mossPlusEmp achieves performance comparable to \mossOracle when $\alpha$ is small, and achieve the best empirical performance when $\alpha \geq 0.3$. When $\alpha \geq 0.7$, \mossOracle needs to explore most/all of the arms to statistically guarantee the finding of at least one best arm, which hurts its empirical performance.

\subsection{Comparison of Progressive Regret Curve}
\label{ms_multi:sec:experiment_real_time}

We compare how the regret curve of each algorithm increases in \cref{ms_multi:fig:synthetic}(b). We consider the same regret minimization configurations as described in \cref{ms_multi:sec:experiment_hardness} with $\alpha = 0.25$. \mossPlusEmp, \mossPlus and \algParallel all outperform \quantile with \mossPlusEmp achieving the performance closest to \mossOracle. \mossOracle, \algParallel and \mossPlusEmp have flattened their regret curve indicating they could confidently recommend the best arm. The regret curves of \mossPlus and \quantile do not flat as the random-sampling component in each of their iterations encourage them to explore new arms. Comparing to \mossPlus, \quantile keeps increasing its regret at a much faster rate and with a much larger variance, which empirically confirms the sub-optimality of their regret guarantees.

\subsection{Real-World Dataset}
\label{ms_multi:sec:experiment_real_data}

We also compare all algorithms in a realistic setting of recommending funny captions to website visitors. We use a real-world dataset from the \emph{New Yorker Magazine} Cartoon Caption Contest\footnote{\url{https://www.newyorker.com/cartoons/contest}.}. The dataset of 1-3 star caption ratings/rewards for Contest 652 consists of $n=10025$ captions\footnote{Available online at \url{https://nextml.github.io/caption-contest-data}.}.  We use the ratings to compute Bernoulli reward distributions for each caption as follows. The mean of each caption/arm $i$ is calculated as the percentage $p_i$ of its ratings that were funny or somewhat funny (i.e., 2 or 3 stars). We normalize each $p_i$ with the best one and then threshold each: if $p_i\geq 0.8$, then put $p_i=1$; otherwise leave $p_i$ unaltered.  This produces a set of $m=54$ best arms  with rewards 1 and all other $9971$ arms with rewards among $[0, 0.8]$. We set $T=10^5$ and this results in a hardness level around $\alpha \approx 0.43$. 

\begin{figure}
    \centering
    \includegraphics[width=.8\textwidth]{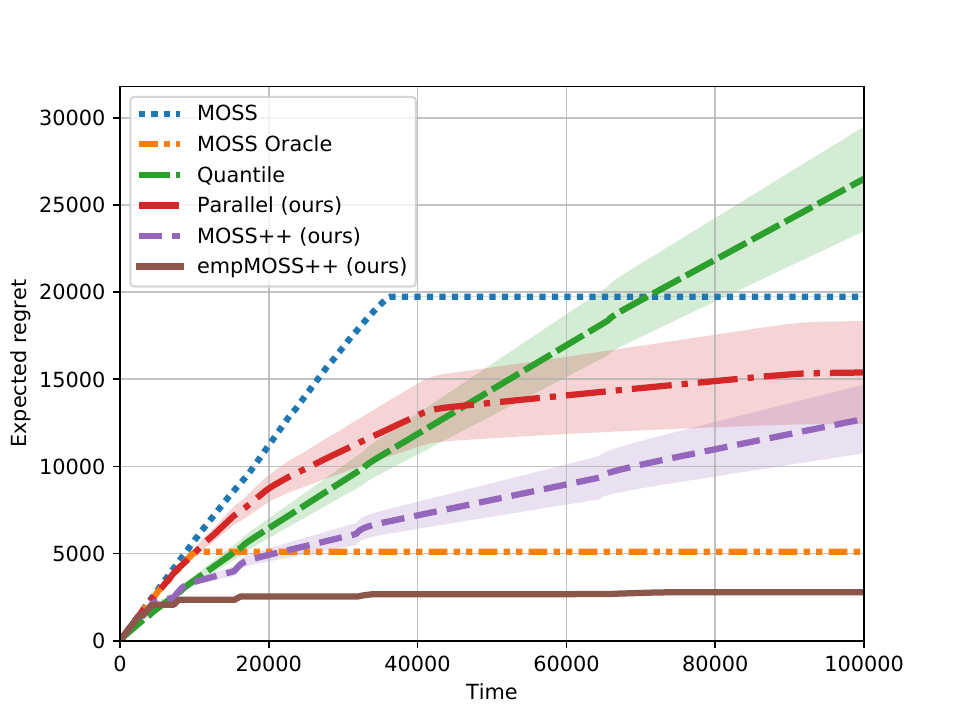}
    \caption{Comparison of progressive regret curve on a real-world dataset from the \emph{New Yorker Magazine} Cartoon Caption Contest.}
    \label{ms_multi:fig:real_data}
\end{figure}

Using these Bernoulli reward models, we compare the performance of each algorithm, as shown in \cref{ms_multi:fig:real_data}. \moss, \mossOracle, \algParallel and \mossPlusEmp have flattened their regret curve indicating they could confidently recommend the funny captions (i.e., best arms).  Although \moss could eventually identify a best arm in this problem, its cumulative regret is more than 7x of the regret achieved by \mossPlusEmp due to its initial exploration phase. The performance of \quantile is even worse, and its cumulative regret is more than 9x of the regret achieved by \mossPlusEmp. One surprising phenomenon is that \mossPlusEmp outperforms \mossOracle in this realistic setting. Our hypothesis is that \mossOracle is a little bit conservative and selects an initial set with cardinality too large. This experiment demonstrates the effectiveness of \mossPlusEmp and \mossPlus in modern applications of bandit algorithm with large action space and limited time horizon.

\section{Discussion}
\label{ms_multi:section:conclusion}

We study a regret minimization problem with large action space but limited time horizon, which captures many modern applications of bandit algorithms. Depending on the number of best/near-optimal arms, we encode the hardness level, in terms of minimax regret achievable, of the given regret minimization problem into a single parameter $\alpha$, and we design algorithms that could adapt to this \emph{unknown} hardness level. Our first algorithm \mossPlus takes a user-specified parameter $\beta$ as input and provides guarantees as long as $\alpha < \beta$; our lower bound further indicates the rate function achieved by \mossPlus is Pareto optimal. Although no algorithm can achieve near minimax optimal regret over all $\alpha$ simultaneously, as demonstrated by our lower bound, we overcome this limitation with an (often) easily-obtained extra information and propose \algParallel that is near-optimal for all settings. Inspired by \mossPlus, We also propose \mossPlusEmp with excellent empirical performance. Experiments on both synthetic and real-world datasets demonstrate the efficiency of our algorithms over the previous state-of-the-art.

\section{Proofs and Supporting Results}
\label{ms_multi:sec:proofs}

\subsection{Proofs and Supporting Results for \cref{ms_multi:section:adaptive}}

We introduce the notation $R_{T \vert {\mathcal F}} = T \cdot \mu_\star - \E [ \sum_{t=1}^T X_t \vert {\mathcal F}]$ for any $\sigma$-algebra ${\mathcal F}$. One should also notice that $\E[R_{T\vert {\mathcal F}}] = R_T$.

\subsubsection{Proof of \cref{ms_multi:thm:regret_multi}}

\begin{lemma}
	\label{ms_multi:lm:prob_replacement}
	For an instance with $n$ total arms and $m$ best arms, and for a subset $S$ selected uniformly at random with cardinality $k$, the probability that none of the best arms are selected in $S$ is upper bounded by $\exp(- mk / n)$. 

\end{lemma}{}
\begin{proof}
	Consider selecting $k$ items out of $n$ items without replacement; and suppose there are $m$ target items. Let ${\mathcal E}$ denote the event where none of the target items are selected, we then have 
	\begin{align}
	\Prob \left({\mathcal E} \right) & = \frac{\binom{n-m}{k}}{\binom{n}{k}} = \frac{ \frac{(n-m)!}{(n-m-k)! k!} }{\frac{n!}{(n-k)! k!}}\nonumber \\
	& = \frac{(n-m)!}{(n-m-k)!} \cdot \frac{(n-k)!}{n!} \nonumber \\
	& = \prod_{i=0}^{k-1} \frac{n-m-i}{n-i} \nonumber \\
	& \leq \left( \frac{n-m}{n} \right)^k \label{ms_multi:eq:prob_decreasing} \\
	& \leq \exp\left({- \frac{m}{n} \cdot k}\right), \label{ms_multi:eq:prob_exp}
	\end{align}
	where \cref{ms_multi:eq:prob_decreasing} comes from the fact that $\frac{n-m-i}{n-i}$ is decreasing in $i$; and \cref{ms_multi:eq:prob_exp} comes from the fact that $1-x \leq \exp(-x)$ for all $x \in \R$. 
	
	Selecting arms with replacement gives the same guarantee (which directly goes to \cref{ms_multi:eq:prob_decreasing}), and can be used in corner cases when $k >n$.
\end{proof}{}

\regretMulti*

\begin{proof} 

	Let $T_i = \sum_{j = 1}^{i} \Delta T_j$. We first notice that \cref{ms_multi:alg:regret_multi} is a valid algorithm in the sense that it selects an arm $A_t$ for any $t \in [T]$, i.e., it does not terminate before time $T$: the argument is clearly true if there exists $i \in [p]$ such that $\Delta T_i = T$; otherwise, we can show that 
	\begin{align*}
	    T_p  = \sum_{i=1}^p \Delta T_i = 2(2^{2p} - 1) \geq 2^{2p} \geq T,
	\end{align*}
	for all $\beta \in [1/2, 1)$. 

	We will only consider the case when $\alpha < \beta$ in the following since otherwise \cref{ms_multi:thm:regret_multi} trivially holds due to $T^{1 + \alpha - \beta} \geq T$.

	Let ${\mathcal F}_{i-1}$ represents information collected up to the beginning of iteration $i$, including the random selection of $S_i$. We use $\mu_{S_i \vert \cF_i} = \max_{\nu \in S_i} \{\E_{X \sim \nu}[X \vert \cF_{i-1}]\}$ to denote the maximum expected reward among arms in $S_i$ conditioned on $\cF_{i-1}$. We use $R_{\Delta T_i \vert \cF_{i-1}} = \Delta T_i \cdot \mu_\star - \E [\sum_{t = T_{i-1}+1}^{T_i} X_t \vert \cF_{i-1}]$ to denote the conditional expected cumulative regret at iteration $i$; and further have $R_{\Delta T_i}=\E[R_{\Delta T_i \vert {\mathcal F}_{i-1}}]$. 

	For any virtual mixture-arm $\widetilde{\nu}_j$ created before iteration $i$ (i.e., $j < i$), we use $\widetilde{\mu}_{j \vert \cF_j} = \E_{X \sim \widetilde{\nu}_{j}}[X \vert {\mathcal F}_{j}]$ to denote its expected reward conditioned on $\cF_j$. Conditioning on $\cF_j$, let $X$ be a sample from a virtual mixture-arm $\widetilde{\nu}_i$, which is realized by first sampling an index $j_t$ (of a real arm) from the empirical measure, and then draw $X$ from the real arm $\nu_{j_t}$. We then know that $X - \widetilde{\mu}_{j \vert \cF_j}$ is (conditional) $(\sqrt{2}/2)$-sub-Gaussian: $X - \widetilde{\mu}_{j \vert \cF_j} = \left(X - \mu_{j_t} \right) + \left(\mu_{j_t} - \widetilde{\mu}_{j \vert \cF_j} \right)$ and thus for any $\lambda \in \R$,
	\begin{align}
		\E \Big[ \exp \left(\lambda \left(X - \widetilde{\mu}_{j \vert \cF_j} \right) \right) \Big\vert \cF_{j} \Big] & = \E \Big[ \E \left[ \exp \left(\lambda \left(X - \widetilde{\mu}_{j \vert \cF_j} \right) \right) \vert j_t \right] \Big\vert \cF_{j}\Big] \nonumber \\
		& = \E \Big[ \exp \left( \lambda(\mu_{j_t} - \widetilde{\mu}_{j \vert \cF_j}) \right) \E \left[ \exp \left(\lambda \left(X - \mu_{j_t}\right) \right) \vert j_t \right] \Big\vert \cF_{j}\Big] \nonumber \\
		& \leq \exp \left(\frac{\lambda^2/4}{2} \right) \E \Big[ \exp \left( \lambda(\mu_{j_t} - \widetilde{\mu}_{j \vert \cF_j}) \right) \Big\vert \cF_j \Big] \nonumber \\
		& \leq \exp \left( \frac{\lambda^2/4}{2} + \frac{\lambda^2/4}{2} \right) \label{ms_multi:eq:bounded_rv}\\
		& = \exp \left( \frac{\lambda^2 /2}{2}\right) \nonumber 
	\end{align}
	where \cref{ms_multi:eq:bounded_rv} comes from the fact that $\mu_{j_t} \in [0, 1]$ and $\E [\mu_{j_t} \vert \cF_j] = \widetilde{\mu}_{j \vert \cF_j}$. In the following, we'll directly plug in the regret bound of \moss for the $1$-sub-Gaussian case.

	Applying \cref{ms_multi:eq:regret_decomposition} on $R_{\Delta T_i \vert {\mathcal F}_{i-1}}$ leads to 
	\begin{align}
	\label{ms_multi:eq:regret_decomposition_internal}
	R_{\Delta T_i\vert {\mathcal F}_{i-1}} = \Delta T_i \cdot  \left(\mu_\star - \mu_{S_i \vert \cF_{i-1}} \right) + \left( \Delta T_i \cdot \mu_{S_i \vert \cF_{i-1}} - \E \left[ \sum_{t=T_{i-1}+1}^{T_i} \mu_{A_t} \, \bigg\vert \, {\mathcal F}_{i-1} \right] \right),
	\end{align}
	where, by a slightly abuse of notations, we use $\mu_{A_t}$ to refer to the mean of arm $A_t \in S_i$, which could also be the mean of a virtual arm constructed in one of the previous iterations.

	We first consider the learning error for any iteration $i \in [p]$. $\mu_{S_i \vert \cF_{i-1}}$ is measurable with respect to ${\mathcal F}_{i-1}$ and thus can be thought as fixed at time $T_{i-1}+1$ (conditioned on ${\mathcal F}_{i-1}$). Since \moss restarts at each iteration, conditioning on the information available at the beginning of the $i$-th iteration, i.e., ${\mathcal F}_{i-1}$, and apply the regret bound for \moss, we have:
	\begin{align}
\Delta T_i \cdot \mu_{S_i \vert \cF_{i-1}} - \E \left[ \sum_{t=T_{i-1}+1}^{T_i} \mu_{A_t} \, \bigg\vert \, {\mathcal F}_{i-1} \right]& \leq 39 \sqrt{|S_i| \Delta T_i} + |S_i| \label{ms_multi:eq:regret_learning_0} \\
	& = 39 \sqrt{( K_i + i - 1 ) \Delta T_i}  + (K_i + i -1) \nonumber \\
	& \leq 39 \sqrt{K_i \Delta T_i + (p-1) \Delta T_i} + (K_i + p-1) \label{ms_multi:eq:regret_learning_1}\\
	& \leq 39\sqrt{ 2^{2p+2} + (p-1)T} + 2^{p+1} + (p-1) \label{ms_multi:eq:regret_learning_2} \\
	& \leq 39\sqrt{ 16  T^{2\beta} + \log_2 (T^\beta)\, T} + 4  T^\beta + \log_2 T^\beta \label{ms_multi:eq:regret_learning_3} \\
	& \leq  166 \, (\log_2 T)^{1/2} \cdot T^\beta, \label{ms_multi:eq:regret_learning}
	\end{align}
	where \cref{ms_multi:eq:regret_learning_0} comes from the guarantee of \moss \cite{lattimore2020bandit}; \cref{ms_multi:eq:regret_learning_1} comes from $i \leq p$; \cref{ms_multi:eq:regret_learning_2} comes from the definition of $K_i$ and $\Delta T_i$; \cref{ms_multi:eq:regret_learning_3} comes from the fact that $p = \lceil \log_2 T^\beta \rceil \leq \log_2 T^\beta  + 1$; \cref{ms_multi:eq:regret_learning} comes from some trivial boundings on the constant.\footnote{One can remove the $(\log_2 T)^{1/2}$ term in many cases, e.g., when $\beta > 1/2$ and $T$ is large enough (with respect to $\beta$). However, we mainly focus on the polynomial terms here.}
	
	Taking expectation over randomness in $\cF_{i-1}$ in \cref{ms_multi:eq:regret_decomposition_internal}, we obtain
	\begin{align}
	\label{ms_multi:eq:regret_decomposition_internal_expectation}
	R_{\Delta T_i}  \leq  \Delta T_i \cdot \E \left[ (\mu_\star - \mu_{S_i \vert \cF_{i-1}}) \right] + 166 \, (\log_2 T)^{1/2} \cdot T^\beta.
	\end{align}

	Now, we only need to consider the first term, i.e., the expected approximation error over the $i$-th iteration. Let ${\mathcal E}_i$ denote the event that none of the best arms, among regular arms, is selected in $S_i$, according to \cref{ms_multi:lm:prob_replacement}, we further have
	\begin{align}
	\Delta T_i \cdot \E \left[ (\mu_\star - \mu_{S_i \vert \cF_{i-1}}) \right] & \leq \Delta T_i \cdot \left( 0 \cdot \Prob(\neg {\mathcal E}_i) + 1 \cdot \Prob({\mathcal E}_i) \right) \label{ms_multi:eq:regret_approximation_expectation_1} \\
	& \leq \Delta T_i \cdot \exp(- K_i/(2T^\alpha)), \label{ms_multi:eq:regret_approximation_expectation}
	\end{align}
	where we use the fact the $\mu_i \in [0,1]$ in \cref{ms_multi:eq:regret_approximation_expectation_1}; and directly plug $n/m \leq  2T^\alpha$ into \cref{ms_multi:eq:prob_exp} to get \cref{ms_multi:eq:regret_approximation_expectation}.
	
	Let $i_\star \in [p]$ be the largest integer, if exists, such that $K_{i_\star} \geq 2T^\alpha \log \sqrt{T}$, we then have that, for any $i \leq i_\star$,
	\begin{align}
	\Delta T_i \cdot \E \left[ (\mu_\star - \mu_{S_i \vert \cF_{i-1}}) \right] \leq \Delta T_i/ \sqrt{T} \leq T / \sqrt{T} \leq \sqrt{T}. \label{ms_multi:eq:regret_approximation_i0}
	\end{align}
	Note that this choice of $i_\star$ indicates $T^\alpha \log T \leq K_{i_\star} < 2 T^\alpha \log T$.
	
	If we have $K_1 < 2T^{\alpha} \log \sqrt{T}$, we then set $i_\star = 1$. Notice that $K_1 = 2^{p+1} =  2^{\lceil \log_2 T^\beta \rceil + 1} \geq 2 T^{\beta} > 2 T^\alpha$, we then have 
	\begin{align}
	\Delta T_1 \cdot \E \left[ (\mu_\star - \mu_{S_1\vert \cF_0}) \right] \leq \Delta T_1 \exp(-1) \leq 2^{p+1} \exp(-1) < 2 T^\beta. \label{ms_multi:eq:regret_approximation_i0_1}
	\end{align}

	Combining \cref{ms_multi:eq:regret_decomposition_internal_expectation} with \cref{ms_multi:eq:regret_approximation_i0} or \cref{ms_multi:eq:regret_approximation_i0_1}, we have for any $i \leq i_\star$, and in particular for $i = i_\star$,
	\begin{align}
	 R_{\Delta T_{i}} & \leq \max\{\sqrt{T}, 2T^\beta\}+ 166 \, (\log_2 T)^{1/2} \cdot T^\beta \nonumber \\
	& \leq 168 \, (\log_2 T)^{1/2} \cdot T^\beta. \label{ms_multi:eq:regret_i0}
	\end{align}
	
	In the case when $i_\star = p$ or when $\Delta T_{i_\star} = \min \{2^{p+i}, T \}=T$, we know that \mossPlus will in fact stop at a time step no larger than $T_{i_\star}$ (since the allowed time horizon is $T$), and incur no regret in iterations $i > i_\star$. In the following, we only consider the case when $i_\star < p$ and $\Delta T_{i_\star} = 2^{p+i_\star}$. As a result, we have $K_{i_\star} \Delta T_{i_\star}  = 2^{2p+2}$ and thus
	\begin{align}
	\Delta T_{i_\star} = \frac{2^{2p + 2}}{K_{i_\star}} > \frac{2^{2p +1}}{T^\alpha \log T}, \label{ms_multi:eq:Ti0}
	\end{align}
	where \cref{ms_multi:eq:Ti0} comes from the fact that $ K_{i_\star} < \max \{ 2 T^\alpha \log T, 2T^\alpha \log \sqrt{T}\} = 2T^\alpha \log T$ by definition of $i_\star$. 
	
	We now analysis the expected approximation error for iteration $i > i_\star$. Since the sampling information during the $i_\star$-th iteration is summarized in the virtual mixture-arm $\widetilde{\nu}_{i_\star}$, and being added to all $S_i$ for all $i > i_\star$. Recall that $\widetilde{\mu}_{i_\star\vert {\mathcal F}_{i_\star}} = \E_{X \sim \widetilde{\nu}_{i_\star}}[X \vert {\mathcal F}_{i_\star}]$ denotes the expected reward of sampling according to the virtual mixture-arm $\widetilde{\nu}_{i_\star}$, conditioned on information collected in ${\mathcal F}_{i_\star}$. For any $i > i_\star$, we then have 
	\begin{align}
	\Delta T_i \cdot \E \left[ (\mu_\star - \mu_{S_i \vert \cF_{i-1}}) \right] & \leq \Delta T_i \cdot \E[(\mu_\star - \widetilde{\mu}_{i_\star\vert {\mathcal F}_{i_\star}})] \nonumber  \\
	& = \frac{\Delta T_i}{\Delta T_{i_\star}}\cdot  \E[ \Delta T_{i_\star} \cdot (\mu_\star - \widetilde{\mu}_{i_\star\vert {\mathcal F}_{i_\star}})]  \nonumber \\
	& = \frac{\Delta T_i}{\Delta T_{i_\star}}\cdot \E\left[\left( \Delta T_{i_\star} \cdot \mu_\star - \sum_{t = T_{i_\star -1}+1}^{T_{i_\star}} {\mu}_{A_t} \right] \right) \nonumber \\
	& = \frac{\Delta T_i}{\Delta T_{i_\star}}\cdot  R_{\Delta T_{i_\star}}\nonumber \\
	& < \frac{\Delta T_i}{\frac{2^{2p+1}}{T^\alpha  \log T}} \cdot 168 \, (\log_2 T)^{1/2} \cdot T^\beta \nonumber \\
	& \leq \frac{T^{1 + \alpha + \beta}}{{2^{2p}}} \cdot 84 \, (\log_2 T)^{3/2}  \label{ms_multi:eq:regret_approximation_important} \\
	& \leq 84 \, (\log_2 T)^{3/2} \cdot T^{1  + \alpha - \beta}, \label{ms_multi:eq:regret_approximation}
	\end{align}
	where \cref{ms_multi:eq:regret_approximation_important} comes from the fact that $\Delta T_i \leq T$ and some rewriting; \cref{ms_multi:eq:regret_approximation} comes from the fact that $p = \lceil \log_2 T^{\beta} \rceil \geq \log_2 T^{\beta}$.

Combining \cref{ms_multi:eq:regret_approximation} and \cref{ms_multi:eq:regret_decomposition_internal_expectation} gives the following regret bound for iterations $i > i_\star$:
\begin{align}
	R_{\Delta T_i}  \leq 250 \, (\log_2 T)^{3/2} \cdot  T^{\max \{\beta, 1+\alpha - \beta \}}, \nonumber 
\end{align}
where the constant $250$ simply comes from $84 + 166$.

Since the cumulative regret is non-decreasing in $t$, we have
\begin{align}
	R_{T} & \leq \sum_{i=1}^p R_{\Delta T_i} \nonumber \\
	& \leq 250 \, p \, (\log_2 T)^{3/2} \cdot T^{\max \{\beta, 1+\alpha - \beta \}} \nonumber   \\
	& \leq 250 \, (\log_2 T + 1) \cdot (\log_2 T)^{3/2} \cdot T^{\max \{\beta, 1+\alpha - \beta \}} \label{ms_multi:eq:total_regret_p}   \\
	& \leq 251 \, (\log_2 T)^{5/2} \cdot T^{\max \{\beta, 1+\alpha - \beta \}}, \nonumber 
\end{align}
where \cref{ms_multi:eq:total_regret_p} comes from the fact that $p = \lceil \log_2(T^\beta) \rceil \leq \log_2(T^\beta)  + 1 \leq \log_2 T  + 1$. Our results follows after noticing that $R_T \leq T$ is a trivial upper bound.
\end{proof}

\subsubsection{Anytime Version}
\label{ms_multi:appendix:anytime_multi}

\begin{algorithm}[H]
	\caption{Anytime version of \mossPlus}
	\label{ms_multi:alg:regret_multi_anytime} 
	\renewcommand{\algorithmicrequire}{\textbf{Input:}}
	\renewcommand{\algorithmicensure}{\textbf{Output:}}
	\begin{algorithmic}[1]
		\REQUIRE User specified parameter $\beta \in [1/2, 1)$.
		\FOR {$i = 0, 1, \dots$}
		\STATE Run \cref{ms_multi:alg:regret_multi} with parameter $\beta$ for $2^i$ rounds (note that we will set $p = \lceil \log_2 2^{i\beta} \rceil =\lceil i\beta \rceil$).
		\ENDFOR 
	\end{algorithmic}
\end{algorithm}

\begin{restatable}{corollary}{regretMultiAnytime}
	\label{ms_multi:corollary:regret_multi_anytime}
	For any unknown time horizon $T$, run \cref{ms_multi:alg:regret_multi_anytime} with an user-specified parameter $\beta \in [1/2, 1)$ leads to the following regret upper bound:
	\begin{align}
	\sup_{\omega \in {\mathcal H}_T(\alpha)} R_T \leq C \, (\log_2 T)^{5/2} \cdot T^{\min \{ \max\{ \beta, 1 + \alpha -\beta \}, 1\}}, \nonumber
	\end{align}
	where $C$ is a universal constant.
\end{restatable}

\begin{proof}
	Let $t_\star$ be the smallest integer such that 
	\begin{align}
	\sum_{i=0}^{t_\star} 2^i = 2^{t_\star+1} - 1\geq T. \nonumber
	\end{align}
	We then only need to run \cref{ms_multi:alg:regret_multi} for at most $t_\star$ times. By the definition of $t_\star$, we also know that $2^{t_\star} \leq T$, which leads to $t_\star \leq \log_2 T$.
	
	Let $\gamma = \min\{\max\{ \beta, 1+\alpha - \beta \},1\}$. From \cref{ms_multi:thm:regret_multi} we know that the regret at $i\in [t_\star]$-th round, denoted as $R_{2^i}$, could be upper bounded by
	\begin{align*}
		R_{2^i} & \leq 251 \, (\log_2 2^i)^{5/2} \cdot (2^{i})^{\gamma} \\
			&= 251 \, i^{5/2} \cdot (2^{\gamma})^i \\
			& \leq 251 \, t_\star^{5/2} \cdot (2^{\gamma})^i  \\
			& \leq 251 \, (\log_2 T)^{5/2} \cdot (2^{\gamma})^i. 
	\end{align*}
	For $i=0$, we have $R_{2^0} \leq 1 \leq 251 \, (\log_2 T)^{5/2} \cdot (2^{\gamma})^0$ as well as long as $T \geq 2$.
	
	Now for the unknown time horizon $T$, we could upper bound the regret by
	\begin{align}
	R_T & \leq \sum_{i=0}^{t_\star} R_{2^i} \nonumber \\
	& \leq 251 \, (\log_2 T)^{5/2} \cdot  \left( \sum_{i=0}^{t_\star} (2^{\gamma})^i\right) \nonumber \\
	& \leq  251 \, (\log_2 T)^{5/2} \cdot \int_{x=0}^{t_\star + 1} (2^{\gamma})^x dx \label{ms_multi:eq:anytime_integral}\\
	& = 251 \, (\log_2 T)^{5/2} \cdot \frac{1}{\log 2^\gamma} \cdot \left( (2^\gamma)^{t_\star + 1} - 1 \right) \nonumber \\
	& \leq \frac{2^\gamma}{\gamma \log 2} \, 251 \, (\log_2 T)^{5/2} \cdot T^\gamma \nonumber \\
	& \leq 1449 \, (\log_2 T)^{5/2} \cdot T^\gamma, \label{ms_multi:eq:anytime_gamma}
	\end{align}
	where \cref{ms_multi:eq:anytime_integral} comes from upper bounding summation by integral; and \cref{ms_multi:eq:anytime_gamma} comes from a trivial bound on the constant when $1/2 \leq \gamma \leq 1$.
\end{proof}

\subsection{Proofs and Supporting Results for \cref{ms_multi:section:lower_bound_pareto_optimal}}

\subsubsection{Proof of \cref{ms_multi:thm:lower_bound}}

\lowerBound*

The proof of \cref{ms_multi:thm:lower_bound} is mainly inspired by the proofs of lower bounds in \cite{locatelli2018adaptivity, hadiji2019polynomial}. Before the start of the proof, we first state a generalized version of Pinsker's inequality developed in \cite{hadiji2019polynomial} (Lemma 3 therein).
\begin{lemma} 
	\label{ms_multi:lm:pinsker}
	Let $\Prob$ and $\Q$ be two probability measures. For any random variable $Z \in [0, 1]$, we have
	\begin{align}
	|\E_{\Prob}[Z] - \E_{\Q}[Z]| \leq \sqrt{{\kl(\Prob, \Q)}/{2}}. \nonumber
	\end{align}
\end{lemma}

We consider $K+1$ bandit instances $\{  \underline{\nu}_i \}_{i=0}^K$ such that each bandit instance is a collection of $n$ distributions $\underline{\nu}_i = (\nu_{i1}, \nu_{i2}, \dots, \nu_{in})$ where each $\nu_{ij}$ represents a Gaussian distribution ${\mathcal N}(\mu_{ij}, 1/4)$ with $\mu_{ij} = \E [\nu_{ij}]$. For any given $0 \leq \alpha^\prime < \alpha \leq 1$ and time horizon $T$ large enough, we choose $n, m_0, m, K \in \N_{+}$ such that the following three conditions are satisfied:
\begin{enumerate}
	\item $n = m_0 + K m $;
	\item ${n}/{m_0} \leq 2T^{\alpha^\prime}$; 
	\item ${n}/{m} \leq 2T^{\alpha}$.
\end{enumerate}
\begin{proposition}
	\label{ms_multi:prop:construction}
	Integers satisfying the above three conditions exist. For instance, we could first fix $m \in \N_{+}$ and set $K = \lfloor T^\alpha \rfloor - 1 \geq 2$.\footnote{$K \geq 2$ holds for $T$ large enough.}  One could then set $m_0 = m\lceil T^{\alpha - \alpha^\prime} \rceil$ and $n = m_0 + Km$. 
\end{proposition}
\begin{proof}
	We notice that the first condition holds by construction. We now show that the second and the third conditions hold. 
	
	For the second condition, we have 
	\begin{align}
	\frac{n}{m_0} & = \frac{m_0+Km }{m_0} \nonumber \\
	& = 1 + \frac{m  (\lfloor T^\alpha \rfloor - 1)}{m\left\lceil T^{\alpha - \alpha^\prime} \right\rceil}  \nonumber \\
	& \leq 1 + \frac{T^{\alpha}}{T^{\alpha - \alpha^\prime}}  \nonumber\\
	& \leq 2 T^{\alpha^\prime}. \nonumber
	\end{align}
	
	For the third condition, we have 
	\begin{align}
	\frac{n}{m} & = \frac{m_0 + Km}{m} \nonumber \\
	& = \frac{m \lceil T^{\alpha - \alpha^\prime} \rceil + (\lfloor T^\alpha \rfloor -1) m }{m} \nonumber\\
	& = \lceil T^{\alpha - \alpha^\prime} \rceil + \lfloor T^\alpha \rfloor -1 \nonumber \\
	& = \left(\lceil T^{\alpha - \alpha^\prime} \rceil - 1 \right)+ \lfloor T^\alpha \rfloor  \nonumber \\
	& \leq T^{\alpha - \alpha^\prime} + T^{\alpha} \nonumber\\
	& \leq 2 T^{\alpha}. \nonumber
	\end{align}
\end{proof}

Now we group $n$ distribution into $K+1$ different groups based on their indices: $S_0 = [m_0]$ and $S_i = [m_0 + i \cdot m]\backslash [m_0 + (i-1)\cdot m]$. Let $\Delta \in (0,1]$ be a parameter to be tuned later, we then define $K+1$ bandit instances $\underline{\nu}_i$ for $i \in  \{ 0\} \cup  [K]$ by assigning different values to their means $\mu_{ij}$:
\begin{align}
\mu_{ij} = \begin{cases} {\Delta}/{2} & \mbox{if } j \in S_0,\\
\Delta & \mbox{if } j \in S_i \mbox{ and } i\neq 0,\\
0 & \mbox{otherwise}.
\end{cases}
\label{ms_multi:eq:lower_bound_means}
\end{align}
 We could clearly see there are $m_0$ best arms in instance $\underline{\nu}_0$ and $m$ best arms in instances $\underline{\nu}_i, \forall i \in [K]$. Based on our construction in \cref{ms_multi:prop:construction}, we could then conclude that, with time horizon $T$, the regret minimization problem with respect to $\underline{\nu}_0$ is in ${\mathcal H}_T(\alpha^\prime)$; and similarly the regret minimization problem with respect to $\underline{\nu}_i$ is in ${\mathcal H}_T(\alpha), \forall i \in [K]$.

For any $t \in [T]$, the tuple of random variables $H_t = (A_1, X_1, \dots, A_t, X_t)$ is the outcome of an algorithm interacting with an bandit instance up to time $t$. Let $\Omega_t = ([n] \times \R)^t \subseteq \R^{2t}$ and ${\mathcal F}_t = \mathfrak{B}(\Omega_t)$; one could then define a measurable space $(\Omega_t, {\mathcal F}_t)$ for $H_t$. The random variables $A_1, X_1, \dots, A_t, X_t$ that make up the outcome are defined by their coordinate projections:
\begin{align}
A_t(a_1, x_1, \dots, a_t, x_t) = a_t \quad \mbox{and} \quad X_t(a_1, x_1, \dots, a_t, x_t) = x_t. \nonumber
\end{align}
For any fixed algorithm/policy $\pi$ and bandit instance $\underline{\nu}_i$, $\forall i \in \{0\}\cup [K]$, we are now constructing a probability measure $\Prob_{i, t}$ over $(\Omega_t, {\mathcal F}_t)$. Note that a policy $\pi$ is a sequence $(\pi_t)_{t=1}^T$, where $\pi_t$ is a probability kernel from $(\Omega_{t-1}, {\mathcal F}_{t-1})$ to $([n], 2^{[n]})$. For each $i$, we define another probability kernel $p_{i, t}$ from $(\Omega_{t-1} \times [n], {\mathcal F}_{t-1} \otimes 2^{[n]})$ to $(\R, \mathfrak{B}(\R))$ that models the reward. Assuming the reward is distributed according to ${\mathcal N}(\mu_{i a_t}, 1/4 )$, we give its explicit expression for any $B \in \mathfrak{B}(\R)$ as:
\begin{align}
p_{i, t} \big( (a_1, x_1, \dots, a_t), B\big) =  \bigintssss_B \sqrt{\frac{2}{\pi}} \exp \big( - 2 (x-\mu_{ia_t} ) \big) dx. \nonumber 
\end{align}
The probability measure over $\Prob_{i, t}$ over $(\Omega_t, {\mathcal F}_t)$ could then be define recursively as $\Prob_{i, t} = p_{i, t} \big( \pi_{t} \Prob_{i, t-1} \big)$. We use $\E_i$ to denote the expectation taken with respect to $\Prob_{i, T}$. Apply the same analysis as on page 21 of \cite{hadiji2019polynomial}, we obtain the following proposition on $\kl$ decomposition.
\begin{proposition}
	\label{ms_multi:prop:KL_decomposition}
	\begin{align}
	\kl \left( \Prob_{0, T}, \Prob_{i, T} \right) = \E_{0} \left[ \sum_{t=1}^T \kl\left( {\mathcal N}(\mu_{0 A_t}, 1/4), {\mathcal N} \left( \mu_{i A_t}, 1/4 \right) \right) \right]. \nonumber
	\end{align}
\end{proposition}

With respect to notations and constructions described above, we now prove \cref{ms_multi:thm:lower_bound}.

\begin{proof}(\cref{ms_multi:thm:lower_bound})
	Let $N_{S_i}(T) = \sum_{t=1}^T \mathds{1}\left( A_t \in S_i\right)$ denote the number of times the algorithm $\pi$ selects an arm in $S_i$ up to time $T$. Let $R_{i, T}$ denote the expected (pseudo) regret achieved by the algorithm $\pi$ interacting with the bandit instance $\underline{\nu}_i$. Based on the construction of bandit instance in \cref{ms_multi:eq:lower_bound_means}, we have 
	\begin{align}
	R_{0, T} \geq \frac{\Delta}{2} \sum_{i=1}^K \E_{0} \left[ N_{S_i}(T) \right], \label{ms_multi:eq:regret_0}
	\end{align}
	and $\forall i \in [K]$,
	\begin{align}
	R_{i, T} \geq \frac{\Delta}{2} \left( T - \E_{i} [N_{S_i}(T)] \right) = \frac{T\Delta}{2} \left( 1- \frac{\E_{i} [N_{S_i}(T)]}{T} \right). \label{ms_multi:eq:regret_i}
	\end{align}
	According to \cref{ms_multi:prop:KL_decomposition} and the calculation of $\kl$-divergence between two Gaussian distributions, we further have
	\begin{align}
	\kl(\Prob_{0, T}, \Prob_{i, T}) & = \E_{0} \left[ \sum_{t=1}^T \kl\left( {\mathcal N}(\mu_{0 A_t}, 1/4), {\mathcal N} \left( \mu_{i A_t}, 1/4 \right) \right) \right] \nonumber\\
	& = \E_{0} \left[ \sum_{t=1}^T 2 \left( \mu_{0 A_t} - \mu_{i A_t} \right)^2 \right] \nonumber \\
	& = 2 \E_{0} \left[ N_{S_i}(T) \right] \Delta^2, \label{ms_multi:eq:kl_difference}
	\end{align}
	where \cref{ms_multi:eq:kl_difference} comes from the fact that $\mu_{0j}$ and $\mu_{ij}$ only differs for $j \in S_i$ and the difference is exactly $\Delta$.
	
	We now consider the average regret over $i \in [K]$:
	\begin{align}
	\frac{1}{K} \sum_{i=1}^K R_{i, T} &  \geq  \frac{T \Delta}{2} \left(1 - \frac{1}{K} \sum_{i=1}^K \frac{\E_i [N_{S_i}(T)]}{T} \right) \nonumber \\
	& \geq \frac{T \Delta}{2} \left(1- \frac{1}{K}\sum_{i=1}^K \left(\frac{\E_0 [N_{S_i}(T)]}{T} + \sqrt{\frac{\kl(\Prob_{0, T}, \Prob_{i, T})}{2}}  \right) \right) \label{ms_multi:eq:ave_regret_pinsker} \\
	& = \frac{T \Delta}{2} \left(1- \frac{1}{K} \frac{\sum_{i=1}^K \E_0 [N_{S_i}(T)]}{T} - \frac{1}{K}\sum_{i=1}^K \sqrt{{\E_{0} \left[ N_{S_i}(T) \right] \Delta^2}}   \right) \label{ms_multi:eq:ave_regret_decomposition} \\
	& \geq \frac{T \Delta}{2} \left(1 - \frac{1}{K} - \sqrt{\frac{\sum_{i=1}^K \E_{0} \left[ N_{S_i}(T) \right] \Delta^2}{K}} \right) \label{ms_multi:eq:ave_regret_concave}\\
	& \geq \frac{T \Delta}{2} \left(1 - \frac{1}{K} - \sqrt{\frac{2 \Delta R_{0, T}}{K}} \right) \label{ms_multi:eq:ave_regret_0} \\
	& \geq \frac{T \Delta}{2} \left(\frac{1}{2} - \sqrt{\frac{2 \Delta B}{K}} \right), \label{ms_multi:eq:ave_regret}
	\end{align}
	where \cref{ms_multi:eq:ave_regret_pinsker} comes from applying \cref{ms_multi:lm:pinsker} with $Z = {N_{S_i}(T)}/{T}$ and $\Prob = \Prob_{0, T}$ and $\Q = \Prob_{i, T}$; \cref{ms_multi:eq:ave_regret_decomposition} comes from applying \cref{ms_multi:eq:kl_difference}; \cref{ms_multi:eq:ave_regret_concave} comes from concavity of $\sqrt{\cdot}$ and the fact that ${\sum_{i=1}^K \E_0 [N_{S_i}(T)]} \leq {T}$; \cref{ms_multi:eq:ave_regret_0} comes from applying \cref{ms_multi:eq:regret_0}; and finally \cref{ms_multi:eq:ave_regret} comes from the fact that $K \geq 2$ by construction and the assumption that $R_{0, T} \leq B$.
	
	To obtain a large value for \cref{ms_multi:eq:ave_regret}, one could maximize $\Delta$ while still make $\sqrt{2 \Delta B/K} \leq 1/4$. Set $\Delta = 2^{-5}K B^{-1}$, following \cref{ms_multi:eq:ave_regret}, we obtain 
	\begin{align}
	\frac{1}{K} \sum_{i=1}^K R_{i, T} & \geq 2^{-8} T K B^{-1} \nonumber \\
	& = 2^{-8} T \left(\left\lfloor T^{\alpha}  \right\rfloor -1 \right) B^{-1} \label{ms_multi:eq:ave_regret_K}\\
	& \geq 2^{-10} T^{1 + \alpha} B^{-1}, \label{ms_multi:eq:ave_regret_final}
	\end{align}
	where \cref{ms_multi:eq:ave_regret_K} comes from the construction of $K$; and \cref{ms_multi:eq:ave_regret_final} comes from the assumption that $\left\lfloor T^{\alpha}  \right\rfloor -1 \geq T^{\alpha}/4$. 
	
	Now we only need to make sure $\Delta = 2^{-5}K B^{-1} \leq 1$. Since we have $K =\left\lfloor T^{\alpha}  \right\rfloor -1 \leq T^\alpha$ by construction and $T^{\alpha} \leq B$ by assumption, we obtain $\Delta = 2^{-5} K B^{-1} \leq 2^{-5} < 1$ as desired.
\end{proof}

\subsubsection{Proof of \cref{ms_multi:thm:pareto}}

\begin{lemma} 
	\label{ms_multi:lm:rate_comparison}
	Suppose an algorithm achieves rate function $\theta$, then for any $0 < \alpha \leq \theta(0)$, we have 
	\begin{align}
	\theta(\alpha) \geq 1 + \alpha - \theta(0). \label{ms_multi:eq:rate_func}
	\end{align}
\end{lemma}
\begin{proof}
	Fix $0 < \alpha \leq \theta(0)$. For any $\epsilon > 0$, there exists constant $c_1$ and $c_2$ such that for sufficiently large $T$,
	\begin{align}
	\sup_{\omega \in {\mathcal H}_T(0)} R_T \leq c_1 T^{\theta(0) + \epsilon} \quad \mbox{and} \quad \sup_{\omega \in {\mathcal H}_T(\alpha)} R_T \leq  c_2 T^{\theta(\alpha) + \epsilon}. \nonumber 
	\end{align}
	Let $B = \max\{c_1,1\} \cdot T^{\theta(0) + \epsilon}$, we could see that $T^{\alpha} \leq T^{\theta(0)} \leq B$ holds by assumption. For $T$ large enough, the condition $\lfloor T^\alpha  \rfloor -1 \geq \max \{ T^\alpha/4, 2\}$ of \cref{ms_multi:thm:lower_bound} holds. We then have
	\begin{align}
	c_2 T^{\theta(\alpha) + \epsilon} \geq 2^{-10} T^{1 + \alpha} \left( \max\{c_1,1\} \cdot T^{\theta(0) + \epsilon} \right)^{-1} = 2^{-10} T^{1+\alpha - \theta(0) - \epsilon} /\max\{c_1,1\}. \nonumber
	\end{align}
	For $T$ sufficiently large, we then must have
	\begin{align}
	\theta(\alpha) + \epsilon \geq 1 + \alpha - \theta(0) - \epsilon. \nonumber
	\end{align}
	Let $\epsilon \rightarrow 0$ leads to the desired result.
\end{proof}

\begin{lemma}
\label{ms_multi:lm:rate_lower_bound}
Suppose a rate function $\theta$ is achieved by an algorithm, then we must have 
\begin{align}
\label{ms_multi:eq:rate_lower_bound}
    \theta(\alpha) \geq \min \{ \max \{ \theta(0), 1 + \alpha - \theta(0) \}, 1 \},
\end{align}
with $\theta(0) \in [1/2, 1]$.
\end{lemma}
\begin{proof}
For any rate function $\theta$ achieved by an algorithm, we first notice that $\theta(\alpha) \geq \theta(\alpha^\prime)$ for any $0 \leq \alpha^\prime < \alpha \leq 1$ since ${\mathcal H}_T(\alpha^\prime) \subseteq {\mathcal H}_T(\alpha)$; this also implies $\theta(\alpha) \geq \theta(0)$. From \cref{ms_multi:lm:rate_comparison}, we further obtain $\theta(\alpha) \geq 1 + \alpha - \theta(0)$ if $\alpha \leq \theta(0)$. Thus, for any $\alpha \in (0, \theta(0)]$, we have 
\begin{align}
\theta(\alpha) \geq \max \{ \theta(0), 1 + \alpha - \theta(0) \}. \label{ms_multi:eq:rate_comparison}
\end{align}
Note that this indicates $\theta(\theta(0)) = 1$, as we trivially have $R_T \leq T$. For any $\alpha \in (\theta(0), 1]$, we have $\theta(\alpha) \geq \theta(\theta(0)) = 1$, which leads to $\theta(\alpha) = 1$ for $\alpha \in [\theta(0), 1]$. To summarize, we obtain the desired result in \cref{ms_multi:eq:rate_lower_bound}. We have $\theta(0) \in [1/2, 1]$ since the minimax optimal rate among problems in ${\mathcal H}_T(0)$ is $1/2$.
\end{proof}

\pareto*

\begin{proof}
From \cref{ms_multi:thm:regret_multi}, we know that the rate in \cref{ms_multi:eq:admissible_rate} is achieved by \cref{ms_multi:alg:regret_multi} with input $\beta$. We only need to prove that no other algorithms achieve strictly smaller rates in pointwise order.

Suppose, by contradiction, we have $\theta^\prime$ achieved by an algorithm such that $\theta^\prime(\alpha) \leq \theta_{\beta}(\alpha)$ for all $\alpha \in [0, 1]$ and $\theta^\prime(\alpha_0) < \theta(\alpha_0)$ for at least one $\alpha_0 \in [0, 1]$. We then must have $\theta^\prime(0) \leq \theta_{\beta}(0) = \beta$. We consider the following two exclusive cases.

\textbf{Case 1: $\theta^\prime(0) = \beta$.} According to \cref{ms_multi:lm:rate_lower_bound}, we must have $\theta^\prime \geq \theta_\beta$, which leads to a contradiction.

\textbf{Case 2: $\theta^\prime(0) = \beta^\prime < \beta$.} According \cref{ms_multi:lm:rate_lower_bound}, we must have $\theta^\prime \geq \theta_{\beta^\prime}$. However, $\theta_{\beta^\prime}$ is not strictly better than $\theta_{\beta}$, e.g., $\theta_{\beta^\prime}(2\beta - 1) =  2\beta - \beta^\prime > \beta = \theta_{\beta}(2\beta - 1)$, which also leads to a contradiction.
\end{proof}

\subsection{Proofs and Supporting Results for \cref{ms_multi:section:extra_info}}
\label{ms_multi:appendix:extra_info}

\subsubsection{Proof of \cref{ms_multi:lm:subroutine_best}}

\subroutineBest*

\begin{proof}

	Let ${\mathcal E}$ denote the event that none of the best arm is selected in $S_{\alpha_i}$. According to \cref{ms_multi:lm:prob_replacement}, the definition of $\alpha$ and the assumption that $\alpha\leq \alpha_i$, we know that $\Prob({\mathcal E}) \leq 1/\sqrt{T}$. We now upper bound the regret:
	\begin{align}
	R_T 
	& \leq \left( 39 \, \sqrt{|S_{\alpha_i}| T} + |S_{\alpha_i}|\right) \cdot \Prob(\neg {\mathcal E})  +  T \cdot \Prob( {\mathcal E})  \label{ms_multi:eq:subroutine_moss} \\
	& \leq  \left( 39 \, \sqrt{|S_{\alpha_i}| T} + |S_{\alpha_i}|\right) \cdot 1   + T \cdot \frac{1}{\sqrt{T}}  \nonumber \\
	& \leq 56 \, (\log T)^{1/2} \cdot T^{(1+\alpha_i)/2} + 2 \log T \cdot T^{\alpha_i} + \sqrt{T}  \nonumber \\
	& \leq 59 \, \log T \cdot T^{(1+\alpha_i)/2} \nonumber \\
	& < 59 \, \log T \cdot T^{(1+\alpha )/2} \cdot T^{1/(2 \, \lceil \log T \rceil)} \label{ms_multi:eq:subroutine_replce}\\
	& \leq 59  \sqrt{e}\, \log T \cdot T^{(1+\alpha )/2}, \label{ms_multi:eq:subroutine_logT}
	\end{align}
	where \cref{ms_multi:eq:subroutine_moss} comes from the regret bound of \moss; \cref{ms_multi:eq:subroutine_replce} comes from the assumption that $\alpha_i  < \alpha + 1/\lceil \log T \rceil$; and \cref{ms_multi:eq:subroutine_logT} comes from the fact that $T^{1/(2 \, \lceil \log T \rceil)} = e^{( \log T / (2\, \lceil \log T \rceil))} \leq \sqrt{e}$.\footnote{One can sharpen the $\log T$ term to $(\log T)^{1/2}$ in many cases, e.g., when $\alpha < 1$ and $T$ is large enough (with respect to $\alpha$). Again, we mainly focus on the polynomial terms here.}

\end{proof}

\subsubsection{Proof of \cref{ms_multi:thm:extra_info}}

We first provide a martingale (difference) concentration result from \cite{wainwright2019high} (a rewrite of Theorem 2.19).

\begin{lemma}
\label{ms_multi:lm:martingale}
Let $\{ D_t\}_{t=1}^{\infty}$ be a martingale difference sequence adapted to filtration $\{{\mathcal F}_{t}\}_{t=1}^{\infty}$. If $\E [\exp(\lambda D_t) \vert {\mathcal F}_{t-1}] \leq \exp(\lambda^2 \sigma^2 /2)$ almost surely for any $\lambda \in \R$, we then have 
\begin{align}
    \Prob \left( \bigg\vert \sum_{i=1}^t D_i \bigg\vert \geq \epsilon \right) \leq 2 \exp \left(- \frac{\epsilon^2}{2t \sigma^2} \right). \nonumber
\end{align}
\end{lemma}

\extraInfo*

\begin{proof}
	This proof largely follows the proof of Theorem 4 in \cite{locatelli2018adaptivity}. For any $T \in \N_+$ and $i \in [\lceil \log T \rceil]$, recall $\sr_i$ is the subroutine initialized with $T$ and $\alpha_i  = i/[\lceil \log T \rceil]$. We use $T_{i, t}$ to denote the number of samples allocated to $\sr_i$ up to time $t$, and represent its empirical regret at time $t$ as $\widehat{R}_{i,t} = T_{i, t}\cdot  \mu_{\star} - \sum_{t=1}^{T_{i,t}} X_{i, t}$ where $X_{i,t} \sim \nu_{A_{i,t}}$ is the $t$-th empirical reward obtained \emph{by} $\sr_i$ and $A_{i,t}$ is the index of the $t$-th arm pulled \emph{by} $\sr_i$. We consider the corresponding regret $R_{i, t}  = T_{i, t}\cdot  \mu_{\star} - \sum_{t=1}^{T_{i,t}} \E [\mu_{A_{i, t}}]$ (which is random in $T_{i, t}$). We choose $\delta = 1/\sqrt{T}$ as the confidence parameter and provide $\delta^\prime = \delta / \lceil \log T \rceil$ failure probability to each subroutine.
	
	Notice that $R_{i, t}  - \widehat{R}_{i,t} = \sum_{t=1}^{T_{i, t}}\left( X_{i,t} - \E[\mu_{A_{i, t}} ] \right)$ is a martingale with respect to filtration ${\mathcal F}_{t} = \sigma \big(\bigcup_{i \in [\lceil \log T \rceil]} \{T_{i, 1}, A_{i,1}, X_{i,1}, \dots, T_{i, t}, A_{i,T_{i,t}}, X_{i, T_{i, t}}\} \big)$; and $(R_{i, t} - \widehat{R}_{i,t} ) - (R_{i, t-1} - \widehat{R}_{i,t-1} )$ defines a martingale difference sequence. Since, no matter what value $T_{i, t}$ takes, $X_{i,T_{i, t}} - \E[\mu_{A_{i, T_{i, t}}} ] = (X_{i,T_{i, t}} - \mu_{A_{i, T_{i, t}}}) + (\mu_{A_{i, T_{i, t}}} - \E[\mu_{A_{i, T_{i, t}}} ])$ is $(\sqrt{2}/2)$-sub-Gaussian (following a similar analysis as in \cref{ms_multi:eq:bounded_rv}), applying \cref{ms_multi:lm:martingale} together with a union bound gives:
	\begin{align}
	\Prob \left( \forall i \in [\lceil \log T \rceil ], \forall t \in [T]: |\widehat{R}_{i,t} - R_{i, t} | \geq \sqrt{T_{i, t} \cdot \log \left( 2T \lceil \log T \rceil/\delta \right)} \right) \leq \delta.
	\end{align}
	We use ${\mathcal E} = \left\{ \forall i \in [\lceil \log T \rceil ], \forall t \in [T]: |\widehat{R}_{i,t} - R_{i, t} | < \sqrt{T_{i, t} \cdot \log \left( 2T \lceil \log T \rceil/\delta \right)} \right\}$ to denote the good event that holds true with probability at least $1-\delta$. Since the regret could be trivially upper bounded by $T \cdot \delta = \sqrt{T}$ when ${\mathcal E}$ doesn't hold, we only focus on the case when event ${\mathcal E}$ holds in the following.
	
	Fix any subroutine $k \in [\lceil \log T \rceil]$ and consider its empirical regret $\widehat{R}_{k,T}$ up to time $T$. For any $j \neq k$, let $T_j \leq T$ be the last time that the subroutine $\sr_j$ was invoked, we have 
	\begin{align}
	\widehat{R}_{j,T_j } & \leq \widehat{R}_{k, T_j } \nonumber \\
	& \leq R_{k, T_j } + \sqrt{T_{k, T_j} \cdot \log \left( 2T \lceil \log T \rceil/\delta \right)} \nonumber \\
	& \leq R_{k, T } + \sqrt{T \cdot \log \left( 2T \lceil \log T \rceil/\delta \right)}, \label{ms_multi:eq:extra_info_non_decreasing}
	\end{align}
	where \cref{ms_multi:eq:extra_info_non_decreasing} comes from the fact that the cumulative regret $R_{k,t }$ in non-decreasing in $t$. Since $\sr_j$ will only run additional $\lceil \sqrt{T} \rceil$ rounds after it was selected at time $T_j$, we further have 
	\begin{align}
	\widehat{R}_{j, T } & \leq 	\widehat{R}_{j, T_j }  + \left\lceil \sqrt{T} \right\rceil \nonumber \\
	& \leq R_{k, T } + \sqrt{5 T \cdot \log \left( 2T \lceil \log T \rceil/\delta \right)}, \label{ms_multi:eq:extra_info_ceiling}
	\end{align}
	where \cref{ms_multi:eq:extra_info_ceiling} comes from the combining \cref{ms_multi:eq:extra_info_non_decreasing} with a trivial bounding $\lceil \sqrt{T} \rceil \leq \sqrt{4T}$ for all $T \in \N_+$. Combining \cref{ms_multi:eq:extra_info_ceiling} with the fact that $R_{j, T} \leq \widehat{R}_{j, T} + \sqrt{T \cdot \log \left( 2T \lceil \log T \rceil/\delta \right)}$ leads to
	\begin{align}
	    R_{j, T} \leq R_{k, T} + 4\sqrt{T \cdot \log \left( 2T \lceil \log T \rceil/\delta \right)}. \label{ms_multi:eq:extra_info_bound}
	\end{align}
	
	Let $i_\star \in [\lceil \log T \rceil]$ denote the index such that $\alpha_{i_\star - 1} < \alpha \leq  \alpha_{i_\star}$. As the total regret is the sum of all subroutines, we have that, for some universal constant $C$,
	\begin{align}
	\sum_{i=1}^{\lceil \log T \rceil} {R}_{i, T }
	& \leq \lceil \log T \rceil \cdot \left( R_{i_\star, T } +   4\sqrt{ T \cdot \log \left( 2T \lceil \log T \rceil/\delta \right)} \right) \label{ms_multi:eq:extra_info_star} \\
	& \leq \lceil \log T \rceil \cdot \left( 59  \sqrt{e}\, \log T \cdot T^{(1+\alpha )/2} +  4\sqrt{ T \cdot \log \left( 2T^{3/2} \lceil \log T \rceil \right)} \right) \label{ms_multi:eq:extra_info_best} \\
	& \leq C \, \left( \log T \right)^{2} T^{(1+ \alpha )/ 2}, \nonumber
	\end{align}
	where \cref{ms_multi:eq:extra_info_star} comes from setting $k = i_\star$ in \cref{ms_multi:eq:extra_info_bound}; \cref{ms_multi:eq:extra_info_best} comes from applying \cref{ms_multi:lm:subroutine_best} with the non-decreasing nature of cumulative regret and taking $\delta = 1/\sqrt{T}$. Integrate once more leads to the desired result.
\end{proof}

\subsubsection{Anytime Version}
\label{ms_multi:appendix_anytime_extra}

The anytime version of \cref{ms_multi:alg:extra_info} could be constructed as following.

\begin{algorithm}[H]
	\caption{Anytime version of \algParallel}
	\label{ms_multi:alg:extra_info_anytime} 
	\renewcommand{\algorithmicrequire}{\textbf{Input:}}
	\renewcommand{\algorithmicensure}{\textbf{Output:}}
	\begin{algorithmic}[1]
		\FOR {$i = 0, 1, \dots$}
		\STATE Run \cref{ms_multi:alg:extra_info} with the optimal expected reward $\mu_\star$ for $2^i$ rounds.
		\ENDFOR 
	\end{algorithmic}
\end{algorithm}

\begin{corollary}
	For any time horizon $T$ and $\alpha \in [0, 1]$ unknown to the learner, run \cref{ms_multi:alg:extra_info_anytime} with optimal expected reward $\mu_\star$ leads to the following anytime regret upper:
	\begin{align}
	\sup_{\omega \in {\mathcal H}_T(\alpha)} R_T \leq C \, \left( \log T \right)^{2} T^{(1+ \alpha )/ 2}, \nonumber
	\end{align}
	where $C$ is a universal constant.
\end{corollary}
\begin{proof}
The proof is similar to the one for \cref{ms_multi:corollary:regret_multi_anytime}.
\end{proof}

\chapter{Model Selection in Linear Bandits}
\label{chapter:model:linear}
We study model selection in linear bandits, where the learner must adapt to the dimension (denoted by $d_\star$) of the smallest hypothesis class containing the true linear model while balancing exploration and exploitation. Previous papers provide various guarantees for this model selection problem, but have limitations; i.e., the analysis requires favorable conditions that allow for inexpensive statistical testing to locate the right hypothesis class or are based on the idea of ``corralling'' multiple base algorithms, which often performs relatively poorly in practice. These works also mainly focus on upper bounds. In this chapter, we establish the first lower bound for the model selection problem. Our lower bound implies that, even with a fixed action set, adaptation to the unknown dimension $d_\star$ comes at a cost: There is no algorithm that can achieve the regret bound $\widetilde{O}(\sqrt{d_\star T})$ simultaneously for all values of $d_\star$. We propose Pareto optimal algorithms that match the lower bound. Empirical evaluations show that our algorithm enjoys superior performance compared to existing ones.

\section{Introduction}
\label{ms_linear:sec:intro}

Model selection considers the problem of choosing an appropriate hypothesis class to conduct learning, and the hope is to optimally balance two types of error: the approximation error and the estimation error. In the supervised learning setting, the learner is provided with a (usually nested) sequence of hypothesis classes $\cH_d \subset \cH_{d+1}$. As an example, $\cH_d$ could be the hypothesis class consisting of polynomials of degree at most $d$. The goal is to design a learning algorithm that adaptively selects the best of these hypothesis classes, denoted by $\cH_\star$, to optimize the trade-off between approximation error and estimation error. Structural Risk Minimization (SRM) \citep{vapnik1974theory, vapnik1995nature, shawe1998structural} provides a principled way to conduct model selection in the standard supervised learning setting. SRM can automatically adapt to the complexity of the hypothesis class $\cH_\star$, with only additional logarithmic factors in sample complexity. Meanwhile, cross-validation \citep{stone1978cross, craven1978smoothing, shao1993linear} serves as a helpful tool to conduct model selection in practice.

Despite the importance and popularity of model selection in the supervised learning setting, only very recently have researchers started to study on model selection problems in interactive/sequential learning setting with bandit feedback. Two additional difficulties are highlighted in such bandit setting \citep{foster2019model}: (1) decisions/actions must be made online/sequentially without seeing the entire dataset; and (2) the learner's actions influence what data is observed, i.e., we only have partial/bandit feedback. In the simpler online learning setting with full information feedback, model selection results analogous to those in the supervised learning setting are obtained by several parameter-free online learning algorithms \citep{mcmahan2013minimax, orabona2014simultaneous, koolen2015second, luo2015achieving, orabona2016coin, foster2017parameter, cutkosky2017online, cutkosky2018black}.

The model selection problem for (contextual) linear bandits is first introduced by \cite{foster2019model}. They consider a sequence of nested linear classifiers in $\R^{d_i}$ as the set of hypothesis classes, with $d_1 < d_2 < \cdots < d_M = d$. The goal is to adapt to the smallest hypothesis class, with apriori \emph{unknown} dimension $d_\star$, that  preserves linearity in rewards. Equivalently, one can think of the model selection problem as learning a true reward parameter $\theta_\star \in \R^d$, but only the first $d_\star$ entries of $\theta_\star$ contain non-zero values. The goal is to design algorithms that could automatically adapt to the intrinsic dimension $d_\star$, rather than suffering the ambient dimension $d$. In favorable scenarios when one can cheaply test linearity, \cite{foster2019model} provide an algorithm with regret guarantee that scales as $\widetilde{O}(K^{1/4} T^{3/4}/\gamma^2 + \sqrt{K d_{\star} T }/\gamma^4)$, where $K$ is the number of arms and $\gamma$ is the smallest eigenvalue of the expected design matrix. The core idea therein is to conduct a sequential test, with sublinear sample complexity, to determine whether to step into a larger hypothesis class on the fly. Although this provides the first guarantee for model selection in the linear bandits, the regret bound is proportional to the number of arms $K$ and the reciprocal of the smallest eigenvalue, i.e., $\gamma^{-1}$. Both $K$ and $\gamma^{-1}$ can be quite large in practice, thus limiting the application of their algorithm. Recall that, when provided with the optimal hypothesis class, the classical algorithm \linucb \citep{chu2011contextual, auer2002using} for linear bandit achieves a regret bound $\widetilde{O}(\sqrt{d_\star T})$, with only polylogarithmic dependence on $K$ and no dependence on $\gamma^{-1}$.

The model selection problem in linear bandits was further studied in many subsequent papers. We roughly divide these methods into the following two sub-categories:
\begin{enumerate}
    \item \textbf{Testing in Favorable Scenarios.} The algorithm in \cite{ghosh2020problem} conducts a sequence of statistical tests to gradually estimate the true support (non-zero entries) of $\theta_\star$, and then applies standard linear bandit algorithms on identified support. The regret bound of their algorithm scales as $\widetilde{O}(d^2/\gamma^{4.65} + d^{1/2}_{\star}T^{1/2})$, where $\gamma = \min\{|\theta_{\star,i}|:\theta_{\star, i} \neq 0\}$ is the minimum magnitude of non-zero entries in $\theta_\star$. Their regret bound not only depends on the ambient dimension $d$ but also scales inversely proportional to a small quantity $\gamma$. Their guarantee becomes vacuous when $d$ and/or $\gamma^{-1}$ are large. \cite{chatterji2020osom} consider a different model selection problem where the rewards come from either a linear model or a model with $K$ independent arms. Their algorithm also relies on sequential statistical testing, which requires assumptions stronger than the ones used in \cite{foster2019model} (thus suffering from similar problems).

    \item \textbf{Corralling Multiple Base Algorithms.} Another approach maintains multiple base learners and use a master algorithm to determine sample allocation among base learners. This type of algorithm is initiated by the \corral algorithm \citep{agarwal2017corralling}. Focusing on our model selection setting, the base learners are usually constructed using standard linear bandit algorithms with respect to different hypothesis classes (dimensions). To give an example of the \corral-type of algorithm, the \smoothCorral algorithm developed in \cite{pacchiano2020model} enjoys regret guarantees $\widetilde{O}(d_{\star}\sqrt{T})$ or $\widetilde{O}(d^{1/2}_{\star}T^{2/3})$. Other algorithms of this type, including some concurrent works, can be found in \cite{abbasi2020regret, arora2020corralling, pacchiano2020regret, cutkosky2020upper, cutkosky2021dynamic}.

\end{enumerate}

Note that above algorithms either only work in favorable scenarios when some critical parameters, e.g., $\gamma^{-1}$ and $K$, are not too large or must balance over multiple base algorithms which often hurts the empirical performance. They also mainly focus on developing upper bounds for the model selection problem in linear bandits.
In this chapter, we explore the fundamental limits (lower bounds) of the model selection problem and design algorithms with matching guarantees (upper bounds). We establish a lower bound, using only a fixed action set, indicating that adaptation to the unknown intrinsic dimension $d_\star$ comes at a cost: There is no algorithm that can achieve the regret bound $\widetilde{O}(\sqrt{d_\star T})$ simultaneously for all values of $d_\star$. 
We also develop a Pareto optimal algorithm, with ideas fundamentally different from ``testing'' \citep{foster2019model, ghosh2020problem} and ``corralling'' \citep{pacchiano2020model, agarwal2017corralling}, to bear on the model selection problem in linear bandits. 
Our algorithm is built upon the construction of virtual mixture-arms, which is previously studied in continuum-armed bandits \citep{hadiji2019polynomial} and $K$-armed bandits \citep{zhu2020regret}. We adapt their methods to our setting, with new techniques developed to deal with the linear structure, e.g., the construction of virtual dimensions.

\subsection{Contribution and Organization}
We briefly summarize our contributions as follows.
\begin{itemize}
    \item We review the model selection problem in linear bandits, and additionally define a new parameter (in \cref{ms_linear:sec:setting}) that reflects the tension between time horizon and the intrinsic dimension. This parameter provides a convenient way to analyze high-dimensional linear bandits.

    \item We establish the first lower bound for the model selection problem in \cref{ms_linear:sec:lower_bound}. Our lower bound indicates that the model selection problem is strictly harder than the problem with given optimal hypothesis class: There is no algorithm that can achieve the non-adaptive $\widetilde{O}(\sqrt{d_\star T})$ regret bound simultaneously for all values of $d_\star$. We additionally characterize the exact Pareto frontier of the model selection problem.
    
    \item In \cref{ms_linear:sec:adaptivity}, we develop a Pareto optimal algorithm that is fundamentally different from existing ones relying on ``testing'' or ``corralling''. Our algorithm is built on the construction of virtual mixture-arms and virtual dimensions. Although our main algorithm is analyzed under a mild assumption, we also provide a workaround.
    
    \item We conduct experiments in \cref{ms_linear:sec:experiment} to evaluate our algorithms. Our main algorithm shows superior performance compared to existing ones. We also show that our main algorithm is fairly robust to the existence of the assumption used in our analysis.
\end{itemize}

\subsection{Additional Related Work}

\paragraph{Bandit with large/continuous action spaces} Adaptivity issues naturally arises in bandit problems with large or infinite action space. In continuum-armed bandit problems \citep{agrawal1995continuum}, actions are embedded into a bounded subset $\cX \subseteq \R^d$ with a smooth function $f$ governing the mean payoff for each arm. Achievable theoretical guarantees are usually influenced by some smoothness parameters, and an important question is to design algorithms that adapt to these \emph{unknown} parameters, as discussed in \cite{bubeck2011lipschitz}. \cite{locatelli2018adaptivity} show that, however, no strategy can be optimal simultaneously over all smoothness classes. \cite{hadiji2019polynomial} establishes the Pareto frontier for continuum-armed bandits with H\"older reward functions. 
Adaptivity is also studied in the discrete case with a large action space \citep{wang2008algorithms, lattimore2015pareto, chaudhuri2018quantile, russo2018satisficing, zhu2020regret}. \citet{lattimore2015pareto} studies the Pareto frontier in standard $K$-armed bandits. \citet{zhu2020regret} develop Pareto optimal algorithms for the case with multiple best arms.

\paragraph{High-dimensional linear bandits} As more and more complex data are being used and analyzed, modern applications of linear bandit algorithms usually involve dealing with ultra-high-dimensional data, sometimes with dimension even larger than time horizon \citep{deshpande2012linear}. 
To make progress in this high-dimensional regime, one natural idea is to study (or assume) sparsity in the reward vector and try to adapt to the unknown true support (non-zero entries). The sparse bandit problem is strictly harder than the model selection setting considered here due to the absence of the hierarchical structures. Consequently, a lower bound on the regret of the form $\Omega(\sqrt{d T})$, which scales with the ambient dimension $d$, is indeed unavoidable in the sparse linear bandit problem \citep{abbasi2012online, lattimore2020bandit}. Other papers deal with the sparsity setting with additional feature feedback \citep{oswal2020linear} or further distributional/structual assumptions \citep{carpentier2012bandit, hao2020high} to circumvent the lower bound. These high-dimensional linear bandit problems motivate our investigation of the relationship between time horizon and data dimension.

\section{Problem Setting}
\label{ms_linear:sec:setting}

We consider a linear bandit problem with a finite action set $\cA \subseteq \R^d$ where $\abs*{\cA} = K$ \citep{auer2002using, chu2011contextual}. (The feature representation of) Each arm/action $a \in \cA$ is viewed as a $d$ dimensional vector, and its expected reward $f(a)$ is linear with respect to a reward parameter $\theta_\star \in \R^d$, i.e., $f(a) = \ang*{a, \theta_\star}$. As standard in the literature \citep{ lattimore2020bandit}, we assume $\max_{a \in \cA} \| a \| \leq 1$ and $\| \theta_\star \| \leq 1$. The bandit instance is said to have intrinsic dimension $d_\star$ if $\theta_\star$ only has non-zero entries on its first $d_\star \leq d$ coordinates. The model selection problem aims at designing algorithm that can automatically adapt to the \emph{unknown} intrinsic dimension $d_\star$ in the interactive learning setting with bandit feedback.

At each time step $t \in [T]$, the algorithm selects an action $A_t \in {\mathcal A}$ based on previous observations and receives a reward $X_t = \langle A_t, \theta_\star \rangle + \eta_t$, where $\eta_t$ is an independent $1$-sub-Gaussian noise. We define the pseudo regret (which is random, due to randomness in $A_t$) over time horizon $T$ as $\widehat{R}_T = \sum_{t=1}^T \left\langle \theta_\star,  a_\star - {A_t} \right\rangle$, where $a_\star$ corresponds to the best action in action set, i.e., $a_\star = \argmax_{a \in {\mathcal A}} \langle a, \theta_\star \rangle$. We measure the performance of any algorithm by its expected regret $R_{T} = \E \sq*{ \widehat{R}_T } = \E \sq*{ \sum_{t=1}^T \left\langle \theta_\star,  a_\star - {A_t} \right\rangle }$.

We primarily focus on the high-dimensional linear bandit setting with ambient dimension $d$ close to or even larger than (the allowed) time horizon $T$. We use ${\mathcal R}(T, d_\star)$ to denote the set of regret minimization problems with time horizon $T$ and any bandit instance with intrinsic dimension $d_\star$. We emphasize that $T$ is part of the problem instance, which was largely neglected in previous work focusing on the low dimensional regime where $T \gg d_\star$. To model the tension between the allowed time horizon and the intrinsic dimension, we define the hardness level as 
\begin{align*}
    \psi \left({\mathcal R}(T, d_\star) \right) = \min \{ \alpha \geq 0: d_\star \leq T^{\alpha}  \} = \log d_\star/ \log T.
\end{align*}
$\psi({\mathcal R}(T, d_\star))$ is used here since it precisely captures the regret over the set of regret minimization problem ${\mathcal R}(T, d_\star)$, as discussed later in our review of the \linucb algorithm and the lower bound. Since smaller $\psi({\mathcal R}(T, d_\star))$ indicates easier problem, we define the family of regret minimization problems with \emph{hardness level} at most $\alpha$ as 
\begin{align}
	{\mathcal H}_T(\alpha) = \{ \cup {\mathcal R}(T, d_\star) : \psi({\mathcal R}(T, d_\star))\leq \alpha \} \nonumber,
\end{align}
where $\alpha \in [0, 1]$. Although $T$ is necessary to define a regret minimization problem, the hardness of the problem is encoded into a single parameter $\alpha$: Problems with different time horizons but the same $\alpha$ are equally difficult in terms of the regret achieved by \linucb (the exponent of $T$). 
We explore the connection $d_\star \leq T^{\alpha}$ in the rest of this chapter and focus on (polynomial) dependence on $T$ (i.e., the dependence on $d_\star$ is translated into the dependence on $T^\alpha$).
We are interested in designing algorithms with worst case guarantees over ${\mathcal H}_T(\alpha)$, but \emph{without} the knowledge of $\alpha$.

\paragraph{\linucb and upper bounds} In the standard setting where $d_\star$ is known, \linucb \cite{chu2011contextual, auer2002using} achieves $\widetilde{O}(\sqrt{d_\star T})$ regret.\footnote{Technically, the regret bound is only achieved by a more complicated algorithm \suplinucb. However, it's common to use \linucb as the practical algorithm. See \citet{chu2011contextual} for detailed discussion.} For any problem in ${\mathcal H}_T(\alpha)$ with \emph{known} $\alpha$, one could run \linucb on the first $\floor*{T^{\alpha}}$ coordinates and achieve $\widetilde{O}(T^{(1+\alpha)/2})$ regret. The goal of model selection is to achieve the $\widetilde{O}(T^{(1+\alpha)/2})$ regret but without the knowledge of $\alpha$.

 \paragraph{Lower bounds} In the case when $d_\star \leq \sqrt{T}$, \cite{chu2011contextual} prove a $\Omega(\sqrt{d_\star T})$ lower bound for linear bandits. When $d_\star \geq \sqrt{T}$ is the case, a lower bound $\Omega(K^{1/4} T^{3/4})$ is developed in \cite{abe2003reinforcement}.

\section{Lower Bound and Pareto Optimality}
\label{ms_linear:sec:lower_bound}

We study lower bounds for model selection in this section. 
We show that simultaneously adapting to all hardness levels is impossible. Such fundamental limitation leads to the established of Pareto frontier.

Our lower bound is constructed by relating the regrets between two (sets of) closely related problems: We show that any algorithm achieves good performance on one of them necessarily performs bad on the other one. 
Similar ideas are previously explored in continuum-armed bandit and $K$-armed bandits \citep{locatelli2018adaptivity, hadiji2019polynomial, zhu2020regret}.
We study the linear case with model selection and establish the following lower bound.\footnote{Our lower bound is quantitatively similar to the one studied in $K$-armed bandits with multiple best arms \citep{zhu2020regret}.}
We use $\omega \in {\mathcal H}_T(\alpha)$ to represent any bandit regret minimization problem with time horizon $T$ and hardness level at most $\alpha$ (i.e., $d_\star \leq T^\alpha$).

\begin{restatable}{theorem}{lowerBound}
	\label{ms_linear:thm:lower_bound}
	Consider any $0 \leq \alpha^\prime < \alpha \leq 1$ and $B > 0$ satisfying $T^{\alpha} \leq B$ and $\lfloor T^\alpha/2 \rfloor \geq \max\{ T^\alpha/4, T^{\alpha^\prime}, 2\}$. If an algorithm is such that $\sup_{\omega \in {\mathcal H}_T(\alpha^\prime)} R_T \leq B$, then the regret of the same algorithm must satisfy
	\begin{align}
	\sup_{\omega \in {\mathcal H}_T(\alpha)} R_T \geq 2^{-10} \, T^{1+\alpha} B^{-1} .\label{ms_linear:eq:lower_bound}
	\end{align}
\end{restatable}

Our lower bound delivers important messages to the model selection problem in linear bandits. Most of the previous efforts and open problems \citep{foster2019model, pacchiano2020model} are made to match the usual non-adaptive regret with known $d_\star$ (or $\alpha$). Our lower bound, however, provides a negative answer towards the open problem of achieving regret guarantees $\widetilde{O}(T^{(1+\alpha)/2})$ simultaneously for all hardness levels $\alpha$. 
We interpret this result next.

\paragraph{Interpretation of \cref{ms_linear:thm:lower_bound}} Fix any linear bandit algorithm. We consider two problem instances with different hardness levels $0 \leq \alpha^\prime < \alpha \leq 1$ (and satisfy the constrains in \cref{ms_linear:thm:lower_bound}). On one hand, if the algorithm is such that $\sup_{\omega \in {\mathcal H}_T(\alpha^\prime)} R_T = \widetilde{\omega}(T^{(1+\alpha^\prime)/2})$, we know that this algorithm is already sub-optimal over problems with hardness level at most $\alpha^\prime$. On the other hand, suppose that the algorithm achieves the desired regret $\widetilde{O}(T^{(1+\alpha^\prime)/2})$ over ${\mathcal H}_T(\alpha^\prime)$. \cref{ms_linear:eq:lower_bound} then tells us that $\sup_{\omega \in {\mathcal H}_T(\alpha)} R_T = \widetilde{\Omega}(T^{(1+2\alpha - \alpha^\prime)/2})$, which is (asymptotically) larger than the desired regret $\widetilde{O}(T^{(1+\alpha)/2})$ over problems with hardness level at most $\alpha$.

If we aim at providing regret bounds with only polylogarithmic dependence on $K$ in linear bandits (which is usually the case for linear bandits with finite action set \citep{auer2002using, chu2011contextual}). our lower bound also provides a negative answer to the open problem of achieving a weaker guarantee $\widetilde{O}(T^\gamma d_\star^{1-\gamma}) = \widetilde{O}(T^{\gamma + \alpha(1-\gamma)})$, with $\gamma \in [1/2,1)$ \citep{foster2019model}, simultaneously for all $d_\star$ (or $\alpha$).

In the model selection setting, the performance of any algorithm should be a function of the hardness level $\alpha$: The algorithm needs to adapt the \emph{unknown} $\alpha$. To further explore the fundamental limit for model selection in linear bandits, following \cite{hadiji2019polynomial, zhu2020regret}, we define rate function to capture the performance of any algorithm (in terms of its regret dependence on polynomial terms of $T$).

\begin{definition}
	\label{ms_linear:def:rate}
	Let $\theta:[0,1] \rightarrow [0, 1]$ denote a non-decreasing function. An algorithm achieves the rate function $\theta$ if 
	\begin{align}
		\forall \epsilon > 0, \forall \alpha \in [0, 1], \quad \limsup_{T \rightarrow \infty} \frac{\sup_{\omega \in {\mathcal H}_T(\alpha)} R_T}{T^{\theta(\alpha)+\epsilon}} < + \infty. \nonumber 
	\end{align}
\end{definition}

Since there may not always exist a pointwise ordering over rate functions, we consider the notion of Pareto optimality over rate functions.

\begin{definition}
	\label{ms_linear:def:pareto}
	A rate function $\theta$ is Pareto optimal if it is achieved by an algorithm, and there is no other algorithm achieving a strictly smaller rate function $\theta^\prime$ in the pointwise order. An algorithm is Pareto optimal if it achieves a Pareto optimal rate function.
\end{definition}

We establish the following lower bound for any rate function that can be achieved by an algorithm designed for model selection in linear bandits.

\begin{restatable}{theorem}{thmRateLowerBound}
    \label{ms_linear:thm:rate_lower_bound}
Suppose a rate function $\theta$ is achieved by an algorithm, then we must have 
\begin{align}
\label{ms_linear:eq:rate_lower_bound}
    \theta(\alpha) \geq \min \{ \max \{ \theta(0), 1 + \alpha - \theta(0) \}, 1 \},
\end{align}
with $\theta(0) \in [1/2, 1]$.
\end{restatable}

\begin{figure}[H]
    \centering
    \includegraphics[width=.8\textwidth]{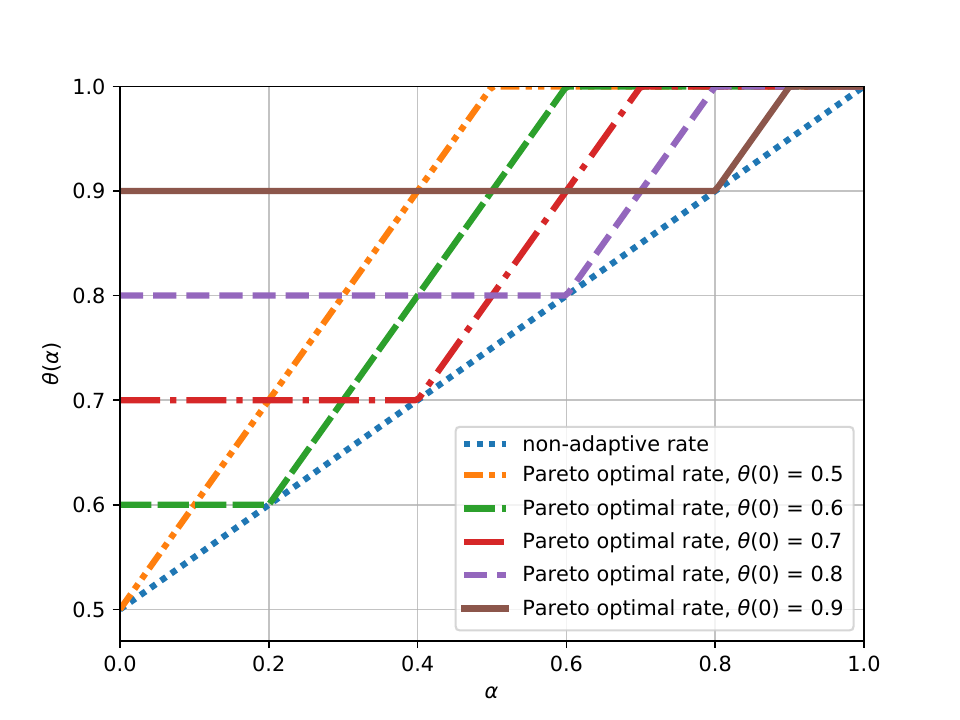}
    \caption{Pareto optimal rates for model selection in linear bandits.}
    \label{ms_linear:fig:pareto_rate}
\end{figure}

\cref{ms_linear:fig:pareto_rate} illustrates the Pareto frontiers for the model selection problem in linear bandits:
The blue dashed line represents the non-adaptive rate function achieved by \linucb with \emph{known} $\alpha$;
Other curves represent Pareto optimal rate functions (achieved by Pareto optimal algorithms introduced in \cref{ms_linear:sec:adaptivity}) for the model selection problem in linear bandits.
\cref{ms_linear:fig:pareto_rate} implies that no algorithm can achieve the non-adaptive rate simultaneously for all $\alpha$: any Pareto optimal curve has to be higher than the non-adaptive curve at least at some points.

\paragraph{Pareto optimality of \corral-type of algorithms} We remark that, accompanied with our lower bound, the \smoothCorral algorithm presented in \cite{pacchiano2020model} is also Pareto optimal. While only a $\widetilde O(d_\star \sqrt{T})$ regret bound is presented for the \smoothCorral algorithm, upon inspection of their analysis, we find that \smoothCorral can actually match the lower bound in \cref{ms_linear:eq:rate_lower_bound} by setting the learning rate as $\eta = T^{-\theta(0)}$, for any $\theta(0) \in [1/2,1)$. 
See \cref{ms_linear:app:corral} for a detailed discussion.

Although the \corral-type of algorithm (e.g., \smoothCorral) is Pareto optimal, they may not be effective in problems with specific structures \citep{papini2021leveraging}. We introduce a new Pareto optimal algorithm in the next section, which is shown to be more practical than \smoothCorral regarding model selection problems in linear bandits (see \cref{ms_linear:sec:experiment}).

\section{Pareto Optimality with New Ideas}
\label{ms_linear:sec:adaptivity}

We develop a Pareto optimal algorithm \linucbPlus (\cref{ms_linear:alg:linucbPlus}) that operates fundamentally different from algorithms rely on ``testing'' \citep{foster2019model, ghosh2020problem} or ``corralling'' \citep{pacchiano2020model, agarwal2017corralling}. 
Our algorithm is built upon the construction of virtual mixture-arms \citep{hadiji2019polynomial, zhu2020regret} and virtual dimensions.

We first introduce some additional notations. For any vector $a \in \R^d$ and $0 \leq d_i \leq d$, we use $a^{(d_i)} \in \R^{d_i}$ to represent the truncated version of $a$ that only keeps the first $d_i$ dimensions. We also use $[a_1; a_2]$ to represent the concatenated vector of $a_1$ and $a_2$. We denote $\cA^{(d_i)} \subseteq \R^{d_i}$ as the ``truncated" action (multi-) set, i.e., ${\mathcal A}^{(d_i)} = \curly*{ a^{(d_i)}   \in \R^{d_i}: a \in {\mathcal A} }$.
One can always \emph{manually} construct the truncated action set ${\mathcal A}^{(d_i)}$ and \emph{pretend} to work with arms with truncated feature representations (though their expected rewards may not be aligned with the truncated feature representations).

\begin{algorithm}[]
	\caption{\linucbPlus}
	\label{ms_linear:alg:linucbPlus} 
	\renewcommand{\algorithmicrequire}{\textbf{Input:}}
	\renewcommand{\algorithmicensure}{\textbf{Output:}}
	\begin{algorithmic}[1]
		\REQUIRE Time horizon $T$ and a user-specified parameter $\beta \in [1/2, 1)$.
		\STATE \textbf{Set:} $p = \lceil \log_2 T^\beta \rceil$, $d_i = \min\{2^{p+2-i}, d\}$ and $\Delta T_i = \min \{ 2^{p + i}, T\}$.
		\FOR {$i = 1, \dots, p$}
		\STATE Run \linucb on a set of arms $S_i$ for $\Delta T_i$ rounds, where $S_i$ contains all arms in ${\mathcal A}^{(d_i)}$ \emph{and} a set of virtual mixture-arms constructed from previous iterations, i.e., $\{\widetilde{\nu}_j\}_{j < i}$. \linucb is operated with respect to an modified linear bandit problem with added virtual dimensions.
		\STATE Construct a virtual mixture-arm $\widetilde{\nu}_i$ based on empirical sampling frequencies in iteration $i$.
		\ENDFOR 
	\end{algorithmic}
\end{algorithm}

We present \linucbPlus in \cref{ms_linear:alg:linucbPlus}. \linucbPlus operates in iterations with geometrically increasing length, and it invokes \linucb (\suplinucb) \citep{chu2011contextual, auer2002using} with (roughly) geometrically decreasing dimensions. 
The core steps of \linucbPlus are summarized at lines 3 and 4 in \cref{ms_linear:alg:linucbPlus}, which consists of construction of virtual mixture-arms and virtual dimensions (the modified linear bandit problem).
We next explain in detail these two core ideas.

\paragraph{The virtual mixture-arm} After each iteration $j$, let $\widehat{p}_j$ denote the vector of empirical sampling frequencies of the arms in that iteration, i.e., the $k$-th element of $\widehat{p}_j$ is the number of times arm $k$, including all previously constructed virtual mixture-arms, was sampled in iteration $j$ divided by the total number of time steps $\Delta T_j$. The virtual mixture-arm for iteration $j$ is the $\widehat{p}_j$-mixture of the arms played in iteration $j$, denoted by $\widetilde{\nu}_j$. When \linucb samples from $\widetilde{\nu}_j$, it first draws a real arm $j_t \sim \widehat{p}_j$ with feature representation $A_t$,\footnote{If the index of another virtual mixture-arm is returned, we sample from that virtual mixture-arm until a real arm is returned.} then pull the real arm $A_t$ to obtain a reward $X_t = \langle \theta_\star, A_t \rangle + \eta_t$. The expected reward of virtual mixture-arm $\widetilde{\nu}_j$ can be expressed as $\langle \theta_\star, a_\star \rangle -  R_{\Delta T_j}/\Delta T_j$, where we use $R_{\Delta T_j}$ to denote the expected regret suffered in iteration $j$. Virtual mixture-arms $\widetilde{\nu}_j$ provide a convenient summary of the information gained in the $j$-th iterations so that we don't need to explore arms in the (effectively) $d_j$ dimensional space again.

\paragraph{Linear bandits with added virtual dimensions} We consider the linear bandit problem in iteration $i$, where each arm in ${\mathcal A}^{(d_i)}$ is viewed as a vector in $\R^{d_i}$. Besides this simple truncation, we lift the feature representation of each arm into a slightly higher dimensional space to include the $i-1$ virtual mixture-arms constructed in previous iterations  (i.e., adding virtual dimensions). More specifically, we augment $i-1$ zeros to the feature representation of each truncated real arm $a \in \cA^{(d_i)}$; we also view each virtual mixture-arm $\widetilde{\nu}_j$ as a $d_i +i-1$ dimensional vector $\widetilde{\nu}^{\langle d_i \rangle }_j$ with its $(d_i + j)$-th entry being $1$ and all other entries being $0$. As a result, \linucb will operate on an modified linear bandit problem with action set ${\mathcal A}^{\langle d_i \rangle } \subseteq  \R^{d_i + i - 1}$, where $	{\mathcal A}^{\langle d_i \rangle } = \curly*{ [a^{(d_i)}; 0] \in \R^{d_i + i - 1}:  a \in {\mathcal A} } \cup \curly*{ \widetilde{\nu}_j^{\langle d_i \rangle} }$, and $\vert {\mathcal A}^{\langle d_i \rangle } \vert =  K + i-1$.
Working with added virtual dimensions allows us to incorporate information stored in virtual mixture-arms without too much additional cost since $i \leq p = O(\log T)$.

\begin{remark}
    \label{ms_linear:rm:modification}
    Previous application of the virtual mixture-arms only works in continuum-armed bandits or $K$-armed bandits \citep{zhu2020regret, hadiji2019polynomial}, where no further modifications are needed to incorporate information stored in virtual mixture-arms. Besides the construction of the virtual dimension, we also provide another way to incorporate the virtual mixture-arms in \cref{ms_linear:sec:remove_assumption}. These modifications are important for the linear bandit case.
\end{remark}

\subsection{Analysis}
\label{ms_linear:sec:adaptive_analysis}

We first analyze \linucbPlus with the following assumption. A modified version of \linucbPlus (\cref{ms_linear:alg:linucbPlus_modified}) is provided in \cref{ms_linear:sec:remove_assumption} and analyzed without the assumption. 

\begin{assumption}
	\label{ms_linear:assumption:action_set}
An action set ${\mathcal A} \subseteq \R^d$ is expressive if we have $a^{[d_i]} = [a^{(d_i)};0] \in {\mathcal A}$ for any $a \in {\mathcal A}$ and $d_i < d$.
\end{assumption}

\cref{ms_linear:assumption:action_set} is naturally satisfied when certain combinatorial structure and ranking information are associated with the action set.
This is best explained with an example. Suppose the arms are consumer products and each has a subset of $d$ possible features, i.e., the arms are \emph{binary vectors} in ${\mathbb R}^d$ indicating the features of the product (the combinatorial aspect). Think of the features as being ordered from base-level features to high-end features (the ranking information). In this case, \cref{ms_linear:assumption:action_set} means that if a product $a \in {\mathcal A}$, then ${\mathcal A}$ also contains all products with fewer high-end features, i.e., truncations of action $a$. We also make the following two comments regarding \cref{ms_linear:assumption:action_set}.

\begin{enumerate}
    \item The action set we used to construct the lower bound in \cref{ms_linear:thm:lower_bound} can be made expressive, as noted in \cref{ms_linear:rm:expressive_lower_bound} in \cref{ms_linear:app:lower_boud};
    \item Although the original version of \linucbPlus is analyzed with \cref{ms_linear:assumption:action_set}, it shows strong empirical performance even without such assumption (see \cref{ms_linear:sec:experiment}).
\end{enumerate}

Equipped with \cref{ms_linear:assumption:action_set}, we can replace the ``truncated'' action set $\cA^{(d_i)}$ with real arms that actually exist in the action set. As a result, the linearity in rewards is preserved in the modified linear bandit problem in $\R^{d_i+i-1}$ with added virtual dimensions.
The modified linear bandit problem is associated with reward vector ${\theta}_\star^{\langle d_i \rangle } =  \sq*{\theta_\star^{(d_i)}; \widetilde{\mu}_1; \dots;  \widetilde{\mu}_{i-1}} \in \R^{d_i + i -1}$, where we use $\widetilde{\mu}_{i} = \langle \theta_\star, a_\star \rangle -  R_{\Delta T_{i}}/\Delta T_{i}$ to denote the expected reward of mixture-arm $\widetilde \nu_i$. In the $i$-th iteration of \linucbPlus, we invoke \linucb to learn reward vector ${\theta}_\star^{\langle d_i \rangle}\in \R^{d_i + i -1}$, which takes worst case regret proportional to $d_i + i -1$ instead of the ambient dimension $d$.

Since there are at most $O(\log T)$ iterations of \linucbPlus, we only need to upper bound its regret at each iteration. Suppose $S_i$ is the set of actions that \linucbPlus is working on at iteration $i$. We use $a_{S_i} = \argmax_{a \in S_i} \langle \theta_\star, a \rangle$ to denote the arm with the highest expected reward; and decompose the regret into approximation error and learning error:
\begin{align}
\label{ms_linear:eq:regret_decomposition}
R_{\Delta T_i} & = \underbrace{ \E \left[ \Delta T_i \cdot \langle \theta_\star, a_\star - a_{S_i} \rangle  \right] }_{\text{expected approximation error due to the selection of $S_i$}} \\ \nonumber
& \quad + \underbrace{ \E \left[ \sum_{t=1}^{\Delta T_i} \langle \theta_\star, a_{S_i} - A_t \rangle\right]}_{\text{expected learning error due to the sampling rule $\{A_t\}_{t=1}^T$}}.
\end{align}

\paragraph{The learning error} At each iteration $i$, \linucbPlus invokes \linucb on a linear bandit problem in $\R^{d_i +i - 1}$ for $\Delta T_i$ time steps, where $d_i$ and $\Delta T_i$ are specifically chosen such that ${d_i \, \Delta T_i} \leq \widetilde{O}(T^{2\beta})$. The learning error is then upper bounded by $\widetilde{O} (\sqrt{d_i \, \Delta T_i}) = \widetilde{O}(T^\beta)$ based on the regret bound of \linucb (the norm of reward vector $\theta_\star^{\ang{d_i}}$ increases with iteration $i$ due to added virtual dimensions, we deal with that in \cref{ms_linear:appendix:modified_linucb}).

\paragraph{The approximation error} Let $i_\star \in [p]$ denote the largest integer such that $d_{i_\star} \geq d_\star$. For iterations $i \leq i_\star$, since $\theta_\star$ only has its first $d_\star \leq d_i$ coordinates being non-zero, we have $\max_{a \in \cA^{\ang{d_i}}} \{ \ang*{\theta_\star^{\ang{d_i}}, a} \} = \ang{\theta_\star, a_\star}$ and the expected approximation error equals zero. As a result, we upper bound the expected regret for iteration $i \leq i_\star$ by its expected learning error, i.e., $R_{\Delta T_i} \leq \widetilde{O}(T^\beta)$. Now consider any iteration $i > i_\star$. Since the virtual mixture-arm $\widetilde{\nu}_{i_\star}$ is constructed by then, and its expected reward is $\widetilde{\mu}_{i_\star} = \langle \theta_\star, a_\star \rangle -  R_{\Delta T_{i_\star}}/\Delta T_{i_\star}$, we can further bound the expected approximation error by $\Delta T_i R_{\Delta T_{i_\star}}/\Delta T_{i_\star} = \widetilde{O}(T^{1+\alpha - \beta})$ (detailed in \cref{ms_linear:app:thm_linucbPlus}).

We now present the formal guarantees of \linucbPlus.

\begin{restatable}{theorem}{thmLinucbPlus}
	\label{ms_linear:thm:linucbPlus}
	Run \linucbPlus with time horizon $T$ and any user-specified parameter $\beta \in [1/2, 1)$ leads to the following upper bound on the expected regret:
	\begin{align}
	& \sup_{\omega \in {\mathcal H}_T(\alpha)}  R_T  \nonumber \\
	& =  O \left(  \log^{7/2} \left( KT\log T\right)  \cdot T^{\min \{\max\{\beta, 1  + \alpha - \beta \},1 \}} \right). \nonumber
	\end{align} 
\end{restatable}

The next theorem shows that \linucbPlus is Pareto optimal with \emph{any} input $\beta \in [1/2, 1)$.

\begin{restatable}{theorem}{pareto}
	\label{ms_linear:thm:pareto}
	The rate function achieved by \linucbPlus with any input $ \beta \in [1/2, 1)$, i.e.,
	\begin{align}
	\theta_{\beta}: \alpha \mapsto \min \{ \max \{ \beta, 1 + \alpha - \beta\}, 1\}, \label{ms_linear:eq:pareto_rate}
	\end{align}
	is Pareto optimal.
\end{restatable}

\subsection{Removing \cref{ms_linear:assumption:action_set}}
\label{ms_linear:sec:remove_assumption}

\cref{ms_linear:assumption:action_set} is used to preserve linearity when working with truncated action sets. In general, one should not expect to deal with misspecified linear bandits without extra cost: \citet{lattimore2020learning} develop a regret lower bound $\Omega(\epsilon \sqrt{d} \, T)$ for misspecified linear bandits with misspecification level $\epsilon$. The lower bound scales linearly with $T$ if there is no extra control/assumptions on the misspecified level $\epsilon$.

Going back to our algorithm, however, we notice that there is a special structure in the source of misspecifications: the virtual-mixture arms are never misspecified. We explore this fact and provide a modified version of \cref{ms_linear:alg:linucbPlus} (i.e., \cref{ms_linear:alg:linucbPlus_modified}) that works \emph{without} \cref{ms_linear:assumption:action_set} and is Pareto optimal. The modified algorithm is less practical since it invokes \smoothCorral as a subroutine (see \cref{ms_linear:sec:experiment}).

\begin{algorithm}[]
	\caption{\linucbPlusCorral}
	\label{ms_linear:alg:linucbPlus_modified} 
	\renewcommand{\algorithmicrequire}{\textbf{Input:}}
	\renewcommand{\algorithmicensure}{\textbf{Output:}}
	\begin{algorithmic}[1]
		\REQUIRE Time horizon $T$ and a user-specified parameter $\beta \in [1/2, 1)$.
		\STATE \textbf{Set:} $p = \lceil \log_2 T^\beta \rceil$, $d_i = \min\{2^{p+2-i}, d\}$ and $\Delta T_i = \min \{ 2^{p + i}, T\}$.
		\FOR {$i = 1, \dots, p$}
		\STATE Construct two (smoothed) base algorithms: (1) a \linucb algorithm working with action set $\cA^{(d_i)}$; and (2) a \ucbalg algorithm working with the set of virtual mixture-arms (if any), i.e., $\{\widetilde{\nu}_j\}_{j < i}$. Invoke \smoothCorral as the master algorithm with learning rate $\eta = 1/\sqrt{d_i \Delta T_i}$.
		\STATE Construct a virtual mixture-arm $\widetilde{\nu}_i$ based on the empirical sampling frequencies in iteration $i$.
		\ENDFOR 
	\end{algorithmic}
\end{algorithm}

We defer detailed discussion on \cref{ms_linear:alg:linucbPlus_modified} and \smoothCorral to \cref{ms_linear:app:remove_assumption}. 
We state the guarantee of \cref{ms_linear:alg:linucbPlus_modified} next.

\begin{restatable}{theorem}{thmLinUCBPlusModified}
\label{ms_linear:thm:linucbPlus_modified}
With any input $\beta \in [1/2,1)$, the rate function achieved by \cref{ms_linear:alg:linucbPlus_modified} (without \cref{ms_linear:assumption:action_set}) is Pareto optimal.
\end{restatable}

\section{Empirical Results}
\label{ms_linear:sec:experiment}

We empirically evaluate our algorithms \linucbPlus and \linucbPlusCorral in this section. 
We find that \linucbPlus enjoys superior performance compared to existing algorithms.
Although \cref{ms_linear:assumption:action_set} is needed in the analysis of \linucbPlus, our experiments show that \linucbPlus is fairly robust to the existence of such assumption.

\begin{figure}[h]
     \centering
     \subfloat[]{\includegraphics[width=.5\textwidth]{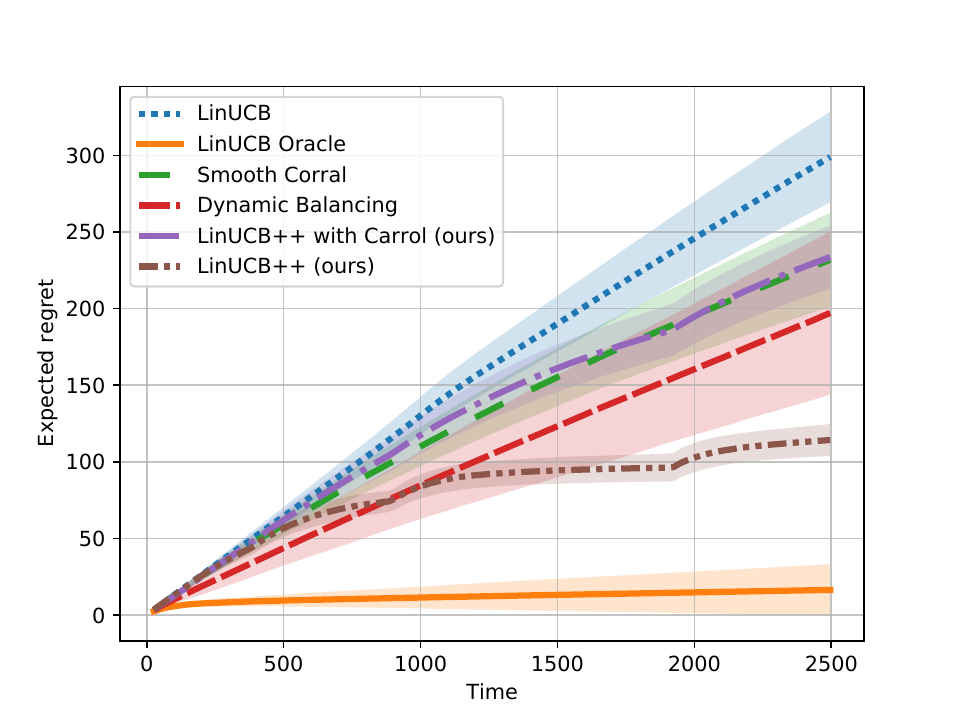}\label{ms_linear:fig:non_expressive_curve}}
     \subfloat[]{\includegraphics[width=.5\textwidth]{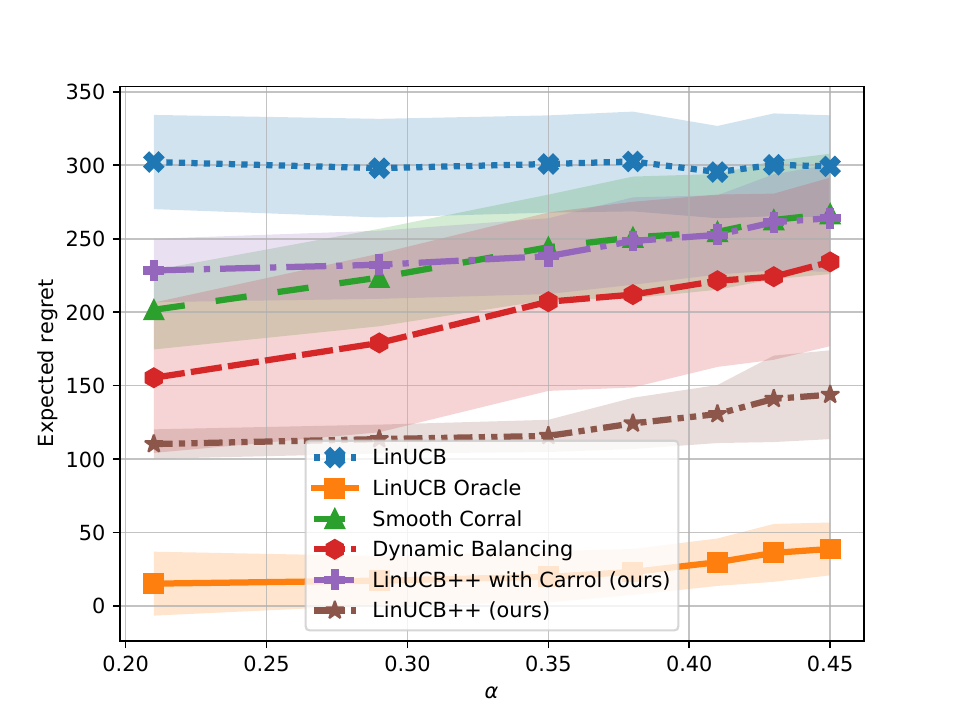}\label{ms_linear:fig:non_expressive_alpha}}
     \caption{Experiments \emph{without} \cref{ms_linear:assumption:action_set}. (a) Comparison of progressive regret curve with hardness level $\alpha \approx 0.32$. (b) Comparison of regret with varying $\alpha$.}
     \label{ms_linear:fig:non_expressive}
\end{figure}

We compare \linucbPlus and \linucbPlusCorral with four baselines: \linucb \citep{chu2011contextual}, \linucbOracle, \smoothCorral \citep{pacchiano2020model} and \dynamicBalancing \citep{cutkosky2021dynamic}.
\linucb is the standard linear bandit algorithm that works in the ambient dimension $\R^d$. \linucbOracle represents the oracle version of \linucb: it takes the knowledge of the instrinsic dimension $d_\star$ and works in $\R^{d_\star}$. \smoothCorral and \dynamicBalancing are implemented with $M = \ceil{\log_2 d}$ base \linucb learners with different dimensions $d_i \in \curly{2^0, 2^1, \dots, 2^{M-1}}$; their master algorithms conduct corraling/regret balancing on top of these base learners. 
We set $\beta = 0.5$ in \linucbPlus and \linucbPlusCorral.\footnote{In practice, we recommend taking $\beta = {(1+\widehat \alpha)}/{2}$ if an estimation $\widehat \alpha$ (of $\alpha$) is available; otherwise, we empirically find that taking $\beta = 0.5$ works well.}
The regularization parameter $\lambda$ for least squares in (all subroutines/base learners of) \linucb is set as $0.1$.

We first conduct experiments \emph{without} an expressive action set (i.e., without \cref{ms_linear:assumption:action_set}).
We consider a regret minimization problem with time horizon $T = 2500$ and a bandit instance consists of $K=1200$ arms selected uniformly at random in the $d=600$ dimensional unit ball. 
We set reward parameter $\theta_\star = [1/\sqrt{d_\star}, \dots, 1/\sqrt{d_\star}, 0, \dots, 0]^\top \in \R^d$ for any intrinsic dimension $d_\star$ (see \cref{ms_linear:app:experiment} for experiments with other choices of $\theta_\star$).
To prevent lengthy exploration over exploitation, we consider Gaussian noises with zero means and $0.1$ standard deviations.
We evaluate each algorithm on $100$ independent trials and average the results. \cref{ms_linear:fig:non_expressive_curve} shows how regret curves of different algorithms increase. The experiment is run with intrinsic dimension $d_\star = 12$, which corresponds to a hardness level $\alpha \approx 0.32$. \linucbPlus outperforms all other algorithms (except \linucbOracle), and enjoys the smallest variance. \linucbPlus (almost) flatten its regret curve at early stages, indicating that it has learned the true reward parameter. \cref{ms_linear:fig:non_expressive_alpha} illustrates the performance of algorithms with respect to different intrinsic dimensions. We run experiments with $d_\star \in \{5,10,15,20,25,30, 35\}$, and mark the corresponding $\alpha$ values in the plot. Across all $\alpha$ values, \linucbPlus shows superior performance compared to \linucb, \smoothCorral, \dynamicBalancing and \linucbPlusCorral. 
These results indicate that \linucbPlus can be practically applied without an expressive action set (thus without \cref{ms_linear:assumption:action_set}).

The empirically poor performance of \corral-type of algorithms might be due to the fact that they need to balance over multiple base algorithms. 
On the other hand, \linucbPlus invokes only one \linucb subroutine at each iteration. Although the subroutine is restarted at the beginning of each iteration, it runs on (roughly) geometrically decreasing dimensions. Such efficient learning procedure is backed by our construction of virtual mixture-arms and virtual dimensions.

\begin{figure}[h]
     \centering
     \subfloat[]{\includegraphics[width=.5\textwidth]{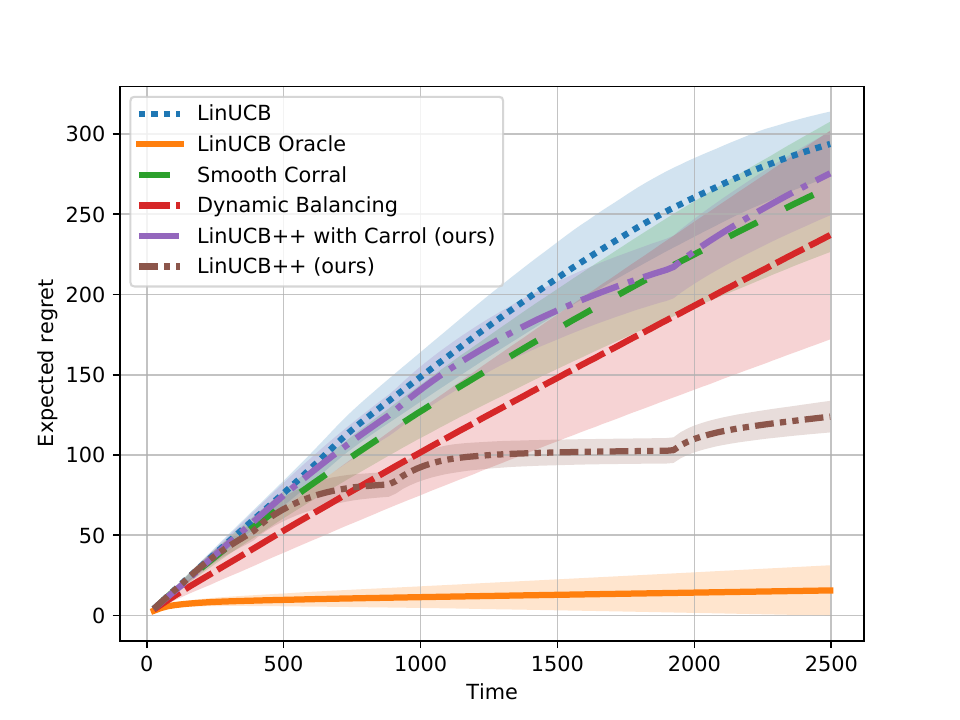}\label{ms_linear:fig:expressive_curve}}
     \subfloat[]{\includegraphics[width=.5\textwidth]{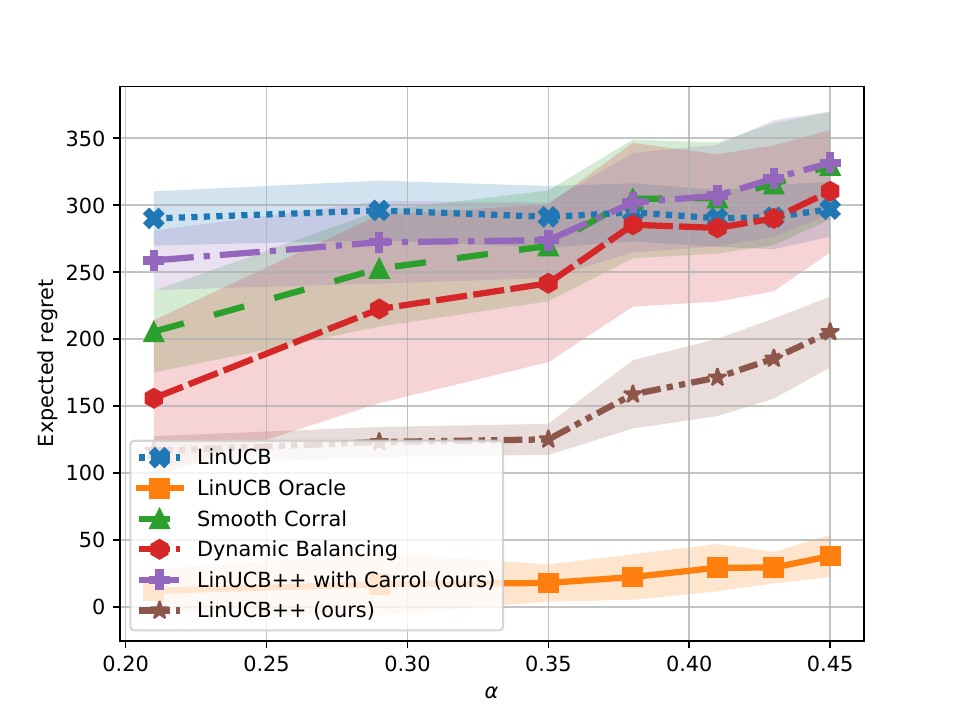}\label{ms_linear:fig:expressive_alpha}}
     \caption{Similar experiment setups to those shown in \cref{ms_linear:fig:non_expressive}, but with \cref{ms_linear:assumption:action_set}.}
     \label{ms_linear:fig:expressive}
\end{figure}

We also run experiments \emph{with} expressive action sets.
We first generate $K=800$ arms uniformly at random from a $d=400$ dimensional unit ball. The action set is then made \emph{expressive} by adding actions with truncated features.\footnote{We only truncate actions with respect to $d_i$\,s selected by \linucbPlus to avoid the computational burden of dealing with a large number of actions.}
We provide the expressive action set to all algorithms since the best reward could be achieved by a truncated arm.
Other experimental setups are similar to the ones described before.
The shape of curves appearing in both \cref{ms_linear:fig:expressive_curve} and \cref{ms_linear:fig:expressive_alpha} are resembles the ones in \cref{ms_linear:fig:non_expressive}, and \linucbPlus outperforms \linucb, \smoothCorral, \dynamicBalancing and \linucbPlusCorral. One slight difference is that \smoothCorral, \dynamicBalancing, \linucbPlusCorral and \linucbPlus have relatively worse performance when as $\alpha$ increases: 
The regret curves (in \cref{ms_linear:fig:expressive_alpha}) increase at faster speeds.
\smoothCorral, \dynamicBalancing and \linucbPlusCorral are outperformed by the standard \linucb when the hardness level $\alpha$ gets large.

\section{Discussion}
\label{ms_linear:sec:discussion}

We study the model selection problem in linear bandits where the goal is to adapt to the \emph{unknown} intrinsic dimension $d_\star$, rather than suffering from regret proportional to the ambient dimension $d$. We establish a lower bound indicating that adaptation to the unknown intrinsic dimension $d_\star$ comes at a cost: There is no algorithm that can achieve the regret bound $\widetilde{O}(\sqrt{d_\star T})$ simultaneously for all values of $d_\star$. Under a mild assumption, we design a Pareto optimal algorithm, with ideas fundamentally different from ``testing'' \citep{foster2019model, ghosh2020problem} and ``corralling'' \citep{pacchiano2020model, agarwal2017corralling}, to bear on the model selection problem in linear bandits. We also provide a workaround to remove the assumption. Experimental evaluations show superior performance of our main algorithm compared to existing ones.

Although linear bandits with a fixed action set are commonly studied in the literature \citep{lattimore2020learning, wagenmaker2021experimental}, an interesting direction is to generalize \linucbPlus to the contextual setting. The current version of \linucbPlus works in the setting with adversarial contexts under the following two additional assumptions: (1) we have a nested sequence of action sets $\cA_t \subseteq \cA_{t+1}$ with $|\cA_T| \leq K$; and (2) one of the best/near-optimal arm belongs to $\cA_1$. How to remove/weaken these assumptions is left to future work.
We also remark that, after our initial (arXiv) publication, \citet{marinov2021pareto} established the Pareto frontier for general contextual bandits, providing a negative answer to open problems raised in \citet{foster2020open}.

\section{Proofs and Supporting Results}
\label{ms_linear:sec:proofs}

\subsection{Proofs and Supporting Results for Section \ref{ms_linear:sec:lower_bound}}

Besides specific treatments for linear bandits (e.g., the lower bound construction for model selection), our proofs for this section largely follow the ones developed in \citet{hadiji2019polynomial, zhu2020regret}. We provide details here for completeness.

\subsubsection{Proof of \cref{ms_linear:thm:lower_bound}}
\label{ms_linear:app:lower_boud}
We consider $K+1$ linear bandit instances such that each is characterized by a reward vector $\theta_i \in \R^{d}$, $0 \leq i \leq K$, with different intrinsic dimensions $d_\star$ (or equivalently $\alpha$). For any action $a \in \R^d$, we obtain a reward $r = \langle \theta_i, a \rangle + \eta$ where $\eta$ is an independent $(1/2)$-sub-Gaussian noise. Time horizon $T$ is fixed and the ambient dimension $d$ is assumed to be large enough to avoid some trivial conflicts in the following construction (e.g., we need $d \geq T^\alpha$ to construct $\theta_i$) . For any $0 \leq \alpha^\prime < \alpha \leq 1$ so that $T^\alpha/2 \geq T^{\alpha^\prime}$, we now provide an explicit construction of $\{\theta_i\}_{i=0}^{K}$ as followings, with $\Delta \in \R$ to be specified later. 
\begin{enumerate}
    \item Let $\theta_0 \in \R^{d}$ be any vector such that it is only supported on one of its first $\lfloor T^{\alpha^\prime} \rfloor$ coordinates and $\| \theta_0 \|_2 = \Delta/2$. The regret minimization problem with respect to $\theta_0$ belongs to ${\mathcal H}_T(\alpha^\prime)$ by construction.
    
    \item For any $i \in [K]$, let $\theta_i = \theta_0 + \Delta \cdot e_{\rho(i)}$ where $e_j$ is the $j$-th canonical base and $\rho(i) = \lfloor T^{\alpha}/2 \rfloor + i$. We set $K = \lfloor T^{\alpha}/2 \rfloor = \Theta(T^\alpha)$ so that the regret minimization problem with respect to any $\theta_i$ belongs to ${\mathcal H}_T(\alpha)$.
\end{enumerate}
We consider a common \emph{fixed} action set ${\mathcal A} = \{a_i\}_{i=0}^K  = \{\theta_0 / \norm*{\theta_0}\} \cup \{ e_{\rho(i)} \}_{i=1}^{K}$ for all regret minimization problems (we set $a_0 = \theta_0 / \norm*{\theta_0}$ and $a_i = e_{\rho(i)}$ for convenience). We could notice that $a_0$ is the best arm with respect to $\theta_0$, which has expected reward $\Delta/2$; and $a_i$ is the best arm with respect to $\theta_i$, which has expected reward $\Delta$.

\begin{remark}
\label{ms_linear:rm:expressive_lower_bound}
    The action set $\cA$ can be made expressive by augmenting the action set with an all-zero action. The all-zero action will not affect our analysis since it always has zero expected reward.
\end{remark}

\begin{remark}
    One can also add other canonical bases into the action set $\cA$ so that $\curly*{\theta_i}_{i=1}^K$ becomes the unique reward vector for corresponding problems. These additional actions will not affect our analysis as well since they all have zero expected reward.
\end{remark}

For any $t \in [T]$, the tuple of random variables $H_t = (A_1, X_1, \dots, A_t, X_t)$ is the outcome of an algorithm interacting with an bandit instance up to time $t$. Let $\Omega_t = \prod_{i=1}^t (\cA \times \R)$ and ${\mathcal F}_t = \mathfrak{B}(\Omega_t)$; one could then define a measurable space $(\Omega_t, {\mathcal F}_t)$ for $H_t$. The random variables $A_1, X_1, \dots, A_t, X_t$ that make up the outcome are defined by their coordinate projections:
\begin{align}
A_t(a_1, x_1, \dots, a_t, x_t) = a_t \quad \mbox{and} \quad X_t(a_1, x_1, \dots, a_t, x_t) = x_t. \nonumber
\end{align}
For any fixed algorithm/policy $\pi$ and bandit instance $\theta_i$, we are now constructing a probability measure $\P_{i, t}$ over $(\Omega_t, {\mathcal F}_t)$. Note that a policy $\pi$ is a sequence $(\pi_t)_{t=1}^T$, where $\pi_t$ is a probability kernel from $(\Omega_{t-1}, {\mathcal F}_{t-1})$ to $(\cA, 2^{\cA})$ with the first probability kernel $\pi_1(\omega, \cdot)$ being defined arbitrarily over $(\cA, 2^{\cA})$, to model the selection of the first action. For each $i$, we define another probability kernel $p_{i, t}$ from $(\Omega_{t-1} \times \cA, {\mathcal F}_{t-1} \otimes 2^{\cA})$ to $(\R, \mathfrak{B}(\R))$ that models the reward. Since the reward is distributed according to ${\mathcal N}(\theta_i^{\top} a_t, 1/4 )$, we gives its explicit expression for any $B \in \mathfrak{B}(\R)$ as following
\begin{align}
p_{i, t} \big( (a_1, x_1, \dots, a_t), B\big) =  \bigintssss_B \sqrt{\frac{2}{\pi}} \exp \big( - 2 (x-\theta_i^{\top} a_t ) \big) dx .\nonumber 
\end{align}
The probability measure over $\P_{i, t}$ over $(\Omega_t, {\mathcal F}_t)$ could then be define recursively as $\P_{i, t} = p_{i, t} \big( \pi_{t} \P_{i, t-1} \big)$. We use $\E_i$ to denote the expectation taken with respect to $\P_{i, T}$. We have the following lemmas.

\begin{lemma}[\citet{lattimore2020bandit}]
	\label{ms_linear:lm:KL_decomposition}
	\begin{align}
	\kl \left( \P_{0, T}, \P_{i, T} \right) = \E_{0} \left[ \sum_{t=1}^T \kl\left( {\mathcal N}(\theta_0^{\top} A_t, 1/4), {\mathcal N} \left( \theta^{\top}_i A_t, 1/4 \right) \right) \right]. \label{ms_linear:eq:KL_decomposition}
	\end{align}
\end{lemma}
\begin{lemma}[\cite{hadiji2019polynomial}]
	\label{ms_linear:lm:pinsker}
	Let $\P$ and $\Q$ be two probability measures. For any random variable $Z \in [0, 1]$, we have \begin{align}
	|\E_{\P}[Z] - \E_{\Q}[Z]| \leq \sqrt{\frac{\kl(\P, \Q)}{2}}. \nonumber
	\end{align}
\end{lemma}

\lowerBound*

\begin{proof}
	Let $N_{i}(T) = \sum_{t=1}^T \mathds{1}\left( A_t = a_i\right)$ denote the number of times the algorithm $\pi$ selects arm $a_i$ up to time $T$. Let $R_{i, T}$ define the expected regret achieved by algorithm $\pi$ interacting with the bandit instance ${\theta}_i$. Based on the construction of bandit instances, we have 
	\begin{align}
	R_{0, T} \geq \frac{\Delta}{2} \sum_{i=1}^K \E_{0} \left[ N_{i}(T) \right], \label{ms_linear:eq:regret_0}
	\end{align}
	and for any $i \in [K]$
	\begin{align}
	R_{i, T} \geq \frac{\Delta}{2} \left( T - \E_{i} [N_{i}(T)] \right) = \frac{T\Delta}{2} \left( 1- \frac{\E_{i} [N_{i}(T)]}{T} \right). \label{ms_linear:eq:regret_i}
	\end{align}
	According to \cref{ms_linear:lm:KL_decomposition} and the calculation of $\kl$-divergence between two Gaussian distributions, we further have
	\begin{align}
	\kl(\P_{0, T}, \P_{i, T}) & = \E_{0} \left[ \sum_{t=1}^T \kl\left( {\mathcal N}(\theta_0^{\top}A_t, 1/4), {\mathcal N} \left( \theta_i^{\top}A_t, 1/4 \right) \right) \right] \nonumber\\
	& = \E_{0} \left[ \sum_{t=1}^T 2 \left\langle \theta_i - \theta_0, A_t \right \rangle^2 \right] \nonumber \\
	& = 2 \E_{0} \left[ N_{i}(T) \right] \Delta^2, \label{ms_linear:eq:kl_difference}
	\end{align}
	where \cref{ms_linear:eq:kl_difference} comes from the fact that $\theta_i = \theta_0 + \Delta \cdot e_{\rho(i)}$ and the only arm in ${\mathcal A}$ with non-zero value on the $\rho(i)$-th coordinate is $a_i = e_{\rho(i)}$, with $\ang*{\theta_i - \theta_0, a_i} = \Delta$.
	
	We now consider the average regret over $i \in [K]$:
	\begin{align}
	\frac{1}{K} \sum_{i=1}^K R_{i, T} &  \geq  \frac{T \Delta}{2} \left(1 - \frac{1}{K} \sum_{i=1}^K \frac{\E_i [N_{i}(T)]}{T} \right) \nonumber \\
	& \geq \frac{T \Delta}{2} \left(1- \frac{1}{K}\sum_{i=1}^K \left(\frac{\E_0 [N_{i}(T)]}{T} + \sqrt{\frac{\kl(\P_{i, T}, \P_{0, T})}{2}}  \right) \right) \label{ms_linear:eq:ave_regret_pinsker} \\
	& = \frac{T \Delta}{2} \left(1- \frac{1}{K} \frac{\sum_{i=1}^K \E_0 [N_{i}(T)]}{T} - \frac{1}{K}\sum_{i=1}^K \sqrt{{\E_{0} \left[ N_{i}(T) \right] \Delta^2}}   \right) \label{ms_linear:eq:ave_regret_decomposition} \\
	& \geq \frac{T \Delta}{2} \left(1 - \frac{1}{K} - \sqrt{\frac{\sum_{i=1}^K \E_{0} \left[ N_{i}(T) \right] \Delta^2}{K}} \right) \label{ms_linear:eq:ave_regret_concave}\\
	& \geq \frac{T \Delta}{2} \left(1 - \frac{1}{K} - \sqrt{\frac{2 \Delta R_{0, T}}{K}} \right) \label{ms_linear:eq:ave_regret_0} \\
	& \geq \frac{T \Delta}{2} \left(\frac{1}{2} - \sqrt{\frac{2 \Delta B}{K}} \right), \label{ms_linear:eq:ave_regret}
	\end{align}
	where \cref{ms_linear:eq:ave_regret_pinsker} comes from applying \cref{ms_linear:lm:pinsker} with $Z = {N_{i}(T)}/{T}$ and $\P = \P_{i, T}$ and $\Q = \P_{0, T}$; \cref{ms_linear:eq:ave_regret_decomposition} comes from \cref{ms_linear:lm:KL_decomposition}; \cref{ms_linear:eq:ave_regret_concave} comes from concavity of $\sqrt{\cdot}$; \cref{ms_linear:eq:ave_regret_0} comes from \cref{ms_linear:eq:regret_0}; and finally \cref{ms_linear:eq:ave_regret} comes from the fact that $K \geq 2$ by construction and the assumption that $R_{0, T} \leq B$.
	
	To obtain a large value for \cref{ms_linear:eq:ave_regret}, one could maximize $\Delta$ while still make sure $\sqrt{2 \Delta B/K} \leq 1/4$. Set $\Delta = 2^{-5}K B^{-1}$, following \cref{ms_linear:eq:ave_regret}, we obtain 
	\begin{align}
	\frac{1}{K} \sum_{i=1}^K R_{i, T} & \geq 2^{-8} T K B^{-1} \nonumber \\
	& = 2^{-8} T \left\lfloor T^{\alpha}/2  \right\rfloor B^{-1} \label{ms_linear:eq:ave_regret_K}\\
	& \geq 2^{-10} T^{1 + \alpha} B^{-1}, \label{ms_linear:eq:ave_regret_final}
	\end{align}
	where \cref{ms_linear:eq:ave_regret_K} comes from the construction of $K$; and \cref{ms_linear:eq:ave_regret_final} comes from the assumption that $\lfloor T^{\alpha}/2  \rfloor \geq T^{\alpha}/4$.
	
	It is clear that any action $a \in \cA$ satisfies $\norm*{a} \leq 1$ by construction, we now only need to make sure that $\norm*{\theta_i} \leq 1$ as well. Notice that $\norm*{\theta_i} \leq \sqrt{5}\Delta/2$ by construction, we only need to make sure $\Delta = 2^{-5}K B^{-1} \leq 2/\sqrt{5}$. Since on one hand $K =\lfloor T^{\alpha}/2 \rfloor \leq T^\alpha$, and on the other hand $T^{\alpha} \leq B$ by assumption, we have $\Delta = 2^{-5} K B^{-1} \leq 2^{-5} < 2/\sqrt{5}$, as desired.
\end{proof}

\subsubsection{Proof of \cref{ms_linear:thm:rate_lower_bound}}

\begin{lemma} 
	\label{ms_linear:lm:rate_comparison}
	Suppose an algorithm achieves rate function $\theta(\alpha)$ on ${\mathcal H}_T (\alpha)$, then for any $0 < \alpha \leq 1$ such that $\alpha \leq \theta(0)$, we have 
	\begin{align}
	\theta(\alpha) \geq 1 + \alpha - \theta(0). \label{ms_linear:eq:rate_func}
	\end{align}
\end{lemma}
\begin{proof}
	Fix $0 \leq \alpha \leq \theta(0)$. For any $\epsilon > 0$, there exists constant $c_1$ and $c_2$ such that
	\begin{align}
	\sup_{\omega \in {\mathcal H}_T(0)} R_T \leq c_1 T^{\theta(0) + \epsilon} \quad \mbox{and} \quad \sup_{\omega \in {\mathcal H}_T(\alpha)} R_T \leq  c_2 T^{\theta(\alpha) + \epsilon}, \nonumber 
	\end{align}
	for sufficiently large $T$.
	Let $B = \max\{c_1,1\} \cdot T^{\theta(0) + \epsilon}$, we could see that $T^{\alpha} \leq T^{\theta(0)} \leq B$ holds by assumption. For $T$ large enough, the condition $\lfloor T^\alpha/2 \rfloor \geq \max\{ T^\alpha/4, T^0, 2\}$ of \cref{ms_linear:thm:lower_bound} holds, and we then have
	\begin{align}
	c_2 T^{\theta(\alpha) + \epsilon} \geq 2^{-10} T^{1 + \alpha} \left( \max\{c_1,1\} \cdot T^{\theta(0) + \epsilon} \right)^{-1} = 2^{-10} T^{1+\alpha - \theta(0) - \epsilon} /\max\{c_1,1\} .\nonumber
	\end{align}
	For $T$ sufficiently large, we then must have
	\begin{align}
	\theta(\alpha) + \epsilon \geq 1 + \alpha - \theta(0) - \epsilon .\nonumber
	\end{align}
	Let $\epsilon \rightarrow 0$ leads to the desired result.
\end{proof}

\thmRateLowerBound*

\begin{proof}
For any adaptive rate function $\theta$ achieved by an algorithm, we first notice that $\theta(\alpha) \geq \theta(\alpha^\prime)$ for any $0 \leq \alpha^\prime \leq \alpha \leq 1$ as ${\mathcal H}_T(\alpha^\prime) \subseteq {\mathcal H}_T(\alpha)$, which also implies $\theta(\alpha) \geq \theta(0)$. From \cref{ms_linear:lm:rate_comparison}, we further obtain $\theta(\alpha) \geq 1 + \alpha - \theta(0)$ if $0 < \alpha \leq \theta(0)$. Thus, for any $\alpha \in (0, \theta(0)]$, we have 
\begin{align}
\theta(\alpha) \geq \max \{ \theta(0), 1 + \alpha - \theta(0) \}. \label{ms_linear:eq:rate_comparison}
\end{align}
Note that this indicates $\theta(\theta(0)) = 1$ since we trivially have $R_T \leq T$. For any $\alpha \in [\theta(0), 1]$, we have $\theta(\alpha) \geq \theta(\theta(0)) = 1$, which also leads to $\theta(\alpha) = 1$ for $\alpha \in [\theta(0), 1]$. To summarize, we obtain the desired result in \cref{ms_linear:eq:rate_lower_bound}. We have $\theta(0) \in [1/2, 1]$ as the minimax optimal rate among problems in ${\mathcal H}_T(0)$ is $1/2$ \citep{chu2011contextual}.
\end{proof}

\subsection{Proofs and Supporting Results for \cref{ms_linear:sec:adaptivity}}

\subsubsection{The virtual-mixture arm}

The expected reward of virtual mixture-arm $\widetilde{\nu}_j$ can be expressed as the total expected reward obtained in iteration $j$ divided by the corresponding time horizon $\Delta T_j$:
\begin{align}
    \widetilde{\mu}_j 
     = \E [\widetilde{\nu}_j] = \E \left[ \sum_{t  \text{ in iteration $j$}} X_t  \right] / \Delta T_j 
     =  \langle \theta_\star, a_\star \rangle -  R_{\Delta T_j}/\Delta T_j \in [-1, 1], \label{ms_linear:eq:expected_reward_virtual} 
\end{align}
where we use $R_{\Delta T_j}$ to denote the expected regret suffered in iteration $j$. Let $X_t$ be the reward obtained by pulling the virtual arm $\widetilde{\nu}_j$ (with $A_t$ being the feature representation of the drawn real arm), we then know that $X_t - \widetilde{\mu}_j$ is $\sqrt{2}$-sub-Gaussian since $X_t - \widetilde{\mu}_j = \left(X_t - \langle \theta_\star, A_t \rangle \right) + \left(\langle \theta_\star, A_t \rangle - \widetilde{\mu}_j  \right) = \eta_t + \left(\langle \theta_\star, A_t \rangle - \widetilde{\mu}_j  \right)$: $\eta_t$ is $1$-sub-Gaussian by assumption and $\left(\langle \theta_\star, A_t \rangle - \widetilde{\mu}_j  \right)$ is $1$-sub-Gaussian due to boundedness $\langle \theta_\star, A_t \rangle \in [-1, 1]$ and $\E [\langle \theta_\star, A_t \rangle] = \widetilde{\mu}_j$.

\subsubsection{Modifications of \linucb}
\label{ms_linear:appendix:modified_linucb}

Recall that, under \cref{ms_linear:assumption:action_set}, the linear reward structure is preserved in the modified linear bandit problem that \linucb will be working on in \cref{ms_linear:alg:linucbPlus}. Two main differences in the modified linear bandit problem from the original setting considered in \cite{chu2011contextual} are: (1) we will be working with $\sqrt{2}$-sub-Gaussian noise while they deal with strictly bounded noise; and (2) the norm of our reward parameter, i.e., $\| {\theta}_\star^{\langle d_i \rangle} \|$, could be as large as $1 + (p - 1) = p = \lceil \log_2(T^\beta) \rceil \leq \log_2(T) + 1 \leq 2 \log T $ when $T \geq 2$. 

To reduce clutters, we consider a $d$ dimensional linear bandit with time horizon $T$ and $K$ actions. We consider the reward structure $X_t = \langle \theta_\star, A_t \rangle + \eta_t$, where $\eta_t$ is an independent $\sqrt{2}$-sub-Gaussian noise, $\| \theta_\star\| \leq 2 \log T$ and $\| A_t \| \leq 1$.
The following \cref{ms_linear:thm:modified_linucb} takes care of these changes.

\begin{theorem}
    \label{ms_linear:thm:modified_linucb}
    For the modified setting introduced above, run \linucb with \linebreak 
    $\alpha = 2\sqrt{ \log (2TK/\delta)}$ leads to an upper bound 
    \begin{align}
        {O}\left( \log^2\left( K T \log ( T) / \delta \right) \cdot \sqrt{d  T } \right) \nonumber 
    \end{align}
    on the (pseudo) random regret with probability at least $1-\delta$.
\end{theorem}

\begin{corollary}
	\label{ms_linear:corollary:linucb}
    For the modified setting introduced above, run \linucb with \linebreak
    $\alpha = 2\sqrt{ \log (2T^{3/2}K)}$ leads to an upper bound 
    \begin{align}
        {O}\left( \log^2\left( K T \log ( T)  \right) \cdot \sqrt{d  T } \right) \nonumber 
    \end{align}
    on the expected regret.
\end{corollary}
\begin{proof}
	One can simply combine the result in \cref{ms_linear:thm:modified_linucb} with $\delta = 1/\sqrt{T}$.
\end{proof}

It turns out that in order to prove \cref{ms_linear:thm:modified_linucb}, we mainly need to modify Lemma 1 in \cite{chu2011contextual}, and the rest of the arguments go through smoothly. The changed exponent on the logarithmic term is due to $\| \theta_\star\| \leq 2 \log T$. We introduce the following notations. Let
\begin{align*}
    V_0 = I  \quad \text{and} \quad V_t = V_{t-1} + A_t A_t^\top
\end{align*}
denote the design matrix up to time $t$; and let
\begin{align*}
    \widehat{\theta}_t = V_t^{-1}\sum_{i=1}^t A_i X_i
\end{align*}
denote the estimate of $\theta_\star$ at time $t$. 

\begin{lemma}(modification of Lemma 1 in \cite{chu2011contextual})
    Suppose for any fixed sequence of selected actions $\{A_i\}_{i \leq t}$ the (random) rewards $\{X_i\}_{i\leq t}$ are independent. Then we have 
    \begin{align}
        \P \left( \forall A_{t+1} \in \cA_{t+1}: \vert \langle \widehat{\theta}_{t} - \theta_\star , A_{t+1} \rangle \vert \leq (\alpha + 2\log T) \sqrt{A_{t+1}^\top V_t^{-1} A_{t+1}} \right) \geq 1-\delta/T. \label{ms_linear:eq:lm_bound} 
    \end{align}
    
\end{lemma}

\begin{remark}
	The requirement of (conditional) independence is guaranted by the \suplinucb algorithm introduced in \cite{chu2011contextual}, and is not satisfied by the vanilla \linucb: the reveal/selection of a future arm $A_{t+1}$ makes previous rewards $\{X_i\}_{i \leq t}$ dependent. See Remark 4 in \cite{han2020sequential} for a detailed discussion.
\end{remark}

\begin{proof}
    For any fixed $A_t$, we first notice that 
    \begin{align}
        \abs*  {\ang {\widehat{\theta}_{t} - \theta_\star , A_{t+1} } } & = \abs { A^{\top}_{t+1} V_t^{-1}\sum_{i=1}^t A_i X_i - A^{\top}_{t+1} \theta_\star}\nonumber \\
        & = \abs*{ A^{\top}_{t+1} V_t^{-1}\sum_{i=1}^t A_i X_i - A^{\top}_{t+1} V_t^{-1} \left( I + \sum_{i=1}^t A_i A_i^\top \right)\theta_\star } \nonumber \\
        & \leq  \abs*{ \sum_{i=1}^t A^\top_{t+1} V^{-1}_t A_i \left( X_i - A_i^{\top} \theta_\star \right)} + \abs{A^{\top}_{t+1} V_t^{-1} \theta_\star} \nonumber \\
        & \leq \abs*{ \sum_{i=1}^t A^\top_{t+1} V^{-1}_t A_i \left( X_i - A_i^{\top} \theta_\star \right)} +  \norm{A^{\top}_{t+1} V_t^{-1} } \cdot \norm{\theta_\star}. \label{ms_linear:eq:lm_bound_two_terms}
    \end{align}
We next bound the two terms in \cref{ms_linear:eq:lm_bound_two_terms} seperately.

For the first term in \cref{ms_linear:eq:lm_bound_two_terms}, since $\left( X_i - A_i^\top \theta_\star \right)$ is $\sqrt{2}$-sub-Gaussian and $\{X_i\}_{i \leq t}$ are independent, we know that $\sum_{i=1}^t A^\top_{t+1} V^{-1}_t A_i \left( X_i - A_i^{\top} \theta_\star \right)$ is \linebreak
$\left( \sqrt{2 \sum_{i=1}^t \left( A^\top_{t+1} V^{-1}_t A_i \right)^2} \right)$-sub-Gaussian. Since 
\begin{align}
    \sqrt{\sum_{i=1}^t \left( A^\top_{t+1} V^{-1}_t A_i \right)^2}  & 
    = \sqrt{\sum_{i=1}^t A_{t+1}^\top V_t^{-1} A_i A_i^\top V_t^{-1} A_{t+1}} \nonumber \\
    & \leq \sqrt{A_{t+1}^\top V_t^{-1} \left( I + \sum_{i=1}^t A_i A_i^\top \right) V_t^{-1} A_{t+1}} \nonumber \\
    & = \sqrt{A_{t+1}^\top V_t^{-1} A_{t+1}}, \nonumber 
\end{align}
according to a standard Chernoff-Hoeffding bound, we have 
\begin{align}
    \P \left( \abs*{ \sum_{i=1}^t A^\top_{t+1} V^{-1}_t A_i \left( X_i - A_i^{\top} \theta_\star \right)} \geq \alpha \sqrt{A_{t+1}^\top V_t^{-1} A_{t+1}} \right) & \leq 2 \exp\left( -\frac{\alpha^2}{4} \right) \nonumber \\
    & = \frac{\delta}{TK}, \label{ms_linear:eq:lm_bound_alpha}
\end{align}
where \cref{ms_linear:eq:lm_bound_alpha} is due to $\alpha = 2\sqrt{\log (2TK/\delta)}$.

For the second term in \cref{ms_linear:eq:lm_bound_two_terms}, we have 
\begin{align}
   \norm{A^{\top}_{t+1} V_t^{-1} } \cdot \norm{\theta_\star} & \leq 2 \log T \, \sqrt{A^\top_{t+1} V^{-1}_t I V^{-1}_t A_{t+1}} \label{ms_linear:eq:lm_bound_norm} \\
   & \leq 2 \log T \, \sqrt{A^\top_{t+1} V^{-1}_t \left(I + \sum_{i=1}^t A_i A_i^\top \right) V^{-1}_t A_{t+1}} \nonumber \\
   & =  2 \log T \,\sqrt{A_{t+1}^\top V_t^{-1} A_{t+1}} . \nonumber 
\end{align}
where \cref{ms_linear:eq:lm_bound_norm} comes from the fact that $\norm{\theta_\star} \leq 2 \log T$.

The desired result in \cref{ms_linear:eq:lm_bound} follows from a union bound argument together with the two upper bounds derived above. 
\end{proof}

\begin{remark}
	Technically, regret guarantees are for a more complicated version of \linucb that ensures statistical independence \citep{chu2011contextual}. However, as recommended by \cite{chu2011contextual}, we will use the more practical \linucb as our subroutine. 
\end{remark}

\subsubsection{Notations and Preliminaries for Analysis of \linucbPlus}
\label{ms_linear:app:preliminary_linucbPlus}

We provide some notations and preliminaries for analysis of \linucbPlus that will be used in the following two subsections, i.e., the proofs of \cref{ms_linear:lm:linucb_learning_error} and \cref{ms_linear:thm:linucbPlus}.

We define $T_i = \sum_{j = 1}^{i} \Delta T_j$ so that the $i$-th iteration of \linucbPlus goes from $T_{i-1} + 1$ to $T_i$. We first notice that \cref{ms_linear:alg:linucbPlus} is a valid algorithm in the sense that it selects an arm $A_t$ for any $t \in [T]$, i.e., it does not terminate before time $T$: the argument is clearly true if there exists $i \in [p]$ such that $\Delta T_i = T$; otherwise, we can show that 
	\begin{align*}
	    T_p  = \sum_{i=1}^p \Delta T_i = 2(2^{2p} - 1) \geq 2^{2p} \geq T,
	\end{align*}
    for all $\beta \in [1/2, 1]$. 

We use $R_{\Delta T_i} = \Delta T_i \cdot \mu_\star - \E [\sum_{t = T_{i-1}+1}^{T_i} X_t ]$ to denote the expected cumulative regret at iteration $i$. Let ${\mathcal F}_{i}$ denote the information collected up to the end of iteration $i$, we further use $R_{\Delta T_i \vert {\mathcal F}_{i-1}}$ to represent the expected regret conditioned on ${\mathcal F}_{i-1}$ and have $\E[R_{\Delta T_i \vert {\mathcal F}_{i-1}}] = R_{\Delta T_i}$.

In the modified linear bandit problem at each iteration $i$, we will be applying \linucb with respect to a $d_i +i -1$ dimensional problem with an action set $\cA^{\ang{d_i}}$ such that $\abs*{\cA^{\ang{d_i}}} \leq K + i -1$. Let $a^{\langle d_i \rangle }_{\star} = \argmax_{a \in {\mathcal A}^{\langle d_i \rangle}} \{\langle \theta^{\langle d_i \rangle}_{\star}, a \rangle \}$ denote the best arm in the $i$-th iteration. Applying \cref{ms_linear:eq:regret_decomposition} on $R_{\Delta T_i \vert {\mathcal F}_{i-1}}$ leads to 
\begin{align}
	\label{ms_linear:eq:regret_decomposition_internal}
	R_{\Delta T_i\vert {\mathcal F}_{i-1}} = \Delta T_i \cdot  \left(\langle \theta_\star, a_\star \rangle - \langle \theta_{\star}^{\langle d_i \rangle}, a_{\star}^{\langle d_i \rangle} \rangle \right) +  \E \left[ \sum_{t=T_{i-1}+1}^{T_i} \langle \theta_{\star}^{\langle d_i \rangle}, a_{\star}^{\langle d_i \rangle} - A_t \rangle \, \bigg\vert \, {\mathcal F}_{i-1} \right] ,
\end{align}
where $A_t \in {\mathcal A}^{\langle d_i \rangle}$ and $\langle \theta_{\star}^{\langle d_i \rangle}, A_t \rangle$ represents the expected reward of pulling arm $A_t$.

\subsubsection{Proof of \cref{ms_linear:lm:linucb_learning_error}}
The proof of \cref{ms_linear:lm:linucb_learning_error} follows the notations and preliminaries introduced in \cref{ms_linear:app:preliminary_linucbPlus}.

\begin{restatable}{lemma}{lmLinucbLearningError}
    \label{ms_linear:lm:linucb_learning_error}
    At each iteration $i \in [p]$, the learning error suffered from subroutine \linucb is upper bounded by ${O} \paren*{\log^{5/2} \left( KT\log T\right) \cdot T^\beta}$.
\end{restatable}

\begin{proof}

    We focus on the second term in \cref{ms_linear:eq:regret_decomposition_internal}, i.e., the (conditional) learning error during iteration $i$. Conditioning on $\cF_{i-1}$, both $\theta^{\langle d_i \rangle}_\star$ and $a_\star^{\ang{d_i}}$ can be treated as fixed quantities. Applying the regret bound in \cref{ms_linear:corollary:linucb}, we have:
	\begin{align}
	& \E \left[ \sum_{t=T_{i-1}+1}^{T_i} \langle \theta_{\star}^{\langle d_i \rangle}, a_{\star}^{\langle d_i \rangle} - A_t \rangle \, \bigg\vert \, {\mathcal F}_{i-1} \right] \nonumber \\
	& = {O} \left( \log^2 \left( (K+i-1) \Delta T_i \log (\Delta T_i) \right) \cdot \sqrt{(d_i +i -1 ) \Delta T_i  } \right) \label{ms_linear:eq:regret_learning_0} \\
	& = {O} \left( \log^2 \left( (K+p) \Delta T_i \log (\Delta T_i) \right) \cdot  \sqrt{(d_i+p) \Delta T_i}  \right) \label{ms_linear:eq:regret_learning_1}\\
	& = {O} \left(\log^2 \left( (K+p) T \log T \right)  \cdot \sqrt{2^{2p+2} + p T} \right) \label{ms_linear:eq:regret_learning_2} \\
	& = {O} \left(\log^2 \left( K T \log T \right)  \cdot \sqrt{T^{2\beta} + \log T \cdot T} \right) \label{ms_linear:eq:regret_learning_3} \\
	& = {O} \left(\log^{5/2} \left( KT\log T\right) \cdot T^\beta  \right), \label{ms_linear:eq:regret_learning}
	\end{align}
	where \cref{ms_linear:eq:regret_learning_0} comes from the guarantee of \linucb in \cref{ms_linear:corollary:linucb}; \cref{ms_linear:eq:regret_learning_1} uses the fact that $i \leq p $; \cref{ms_linear:eq:regret_learning_2} comes from the definition of $d_i$ and $\Delta T_i$; \cref{ms_linear:eq:regret_learning_3} comes from the fact that $p = \ceil*{ \log_2 T^\beta}$; \cref{ms_linear:eq:regret_learning} comes from trivially bounding $\sqrt{T^{2\beta} + \log T \cdot T} = O((\log T)^{1/2} \cdot T^\beta)$.\footnote{One can improve the bound to $\sqrt{T^{2\beta} + \log T \cdot T} = O(T^\beta)$ in many cases, e.g., when $\beta > 1/2$. We mainly focus on the polynomial terms here.} The desired result follows from taking another expectation over randomness in $\cF_{i-1}$.
\end{proof}

\subsubsection{Proof of \cref{ms_linear:thm:linucbPlus}}
\label{ms_linear:app:thm_linucbPlus}
The proof of \cref{ms_linear:thm:linucbPlus} follows the notations and preliminaries introduced in \cref{ms_linear:app:preliminary_linucbPlus}.

\thmLinucbPlus*

\begin{proof}

    When $\alpha \geq \beta$, one could see that \cref{ms_linear:thm:linucbPlus} trivially holds since $T^{1 + \alpha - \beta} \geq T$. In the following, we only consider the case when $\alpha < \beta$.

	Taking expectation on \cref{ms_linear:eq:regret_decomposition_internal} and combining the result in \cref{ms_linear:lm:linucb_learning_error}, we obtain
	\begin{align}
	\label{ms_linear:eq:regret_decomposition_internal_expectation}
	R_{\Delta T_i}  =  \Delta T_i \cdot  \E \left[ \left( \langle \theta_\star, a_\star \rangle -\langle \theta_{\star}^{\langle d_i \rangle}, a_{\star}^{\langle d_i \rangle} \rangle  \right) \right]  + {O} \left( \log^{5/2} \left( KT\log T\right) \cdot T^\beta \right).
	\end{align}

	We now focus on the first term, i.e., the expected approximation error over the $i$-th iteration. Notice that, according to the definition of $a^{\langle d_i \rangle}_{\star}$ and $\theta^{\langle d_i \rangle}_\star$, we have $\langle \theta_{\star}^{\langle d_i \rangle}, a_{\star}^{\langle d_i \rangle} \rangle = \langle \theta_\star, a_\star \rangle$ if $d_i \geq d_\star$, i.e., the optimal arm is contained in the action set $\cA^{\ang{d_i}}$. Let $i_\star \in [p]$ be the largest integer such that $d_{i_\star} \geq d_{\star}$, we then have that, for any $i \leq i_\star$ and in particular for $i = i_\star$,
	\begin{align}
	 R_{\Delta T_{i}} =  {O} \left(T^\beta \log^{5/2} \left( KT\log T\right)\right). \label{ms_linear:eq:regret_i0}
	\end{align}
	
	In the case when $\Delta T_{i_\star} = \min \{2^{p+i_\star}, T \}=T$ or $i_\star = p$, we know that \linucbPlus will in fact stop at a time step no larger than $T_{i_\star}$ (since the allowed time horizon is $T$), and incur no regret in iterations $i > i_\star$. In the following, we only consider the case when $\Delta T_{i_\star} = 2^{p+i_\star}$ and $i_\star < p$. To incooperate another possible corner case when $d_{i_\star} = \min\{2^{p+2-i_\star}, d\} = d$, we consider $d_{i_\star + 1} = 2^{p + 1 - i_\star} < d_{i_\star}$. As a result, we have $d_{i_\star} \Delta T_{i_\star} > d_{i_\star + 1} \Delta T_{i_\star} = 2^{2p+1}$, which leads to
	\begin{align}
	\Delta T_{i_\star} > \frac{2^{2p + 1}}{d_{i_\star}} > \frac{2^{2p }}{d_\star} = \frac{2^{2p }}{T^{\alpha}}, \label{ms_linear:eq:Ti0}
	\end{align}
	where \cref{ms_linear:eq:Ti0} comes from the fact that $ d_{i_\star} < 2 d_\star$ according to the definition of $i_\star$.\footnote{We will have $\Delta T_{i_\star}  \geq 2^{2p+1}/T^\alpha > 2^{2p}/T^\alpha$ if $d_{i_\star} = \min\{2^{p+2-i_\star}, d\}  = 2^{p+2-i_\star}$.}
	
	We now analysis the expected approximation error for iteration $i > i_\star$. Since the sampling information during $i_\star$-th iteration is summarized in the virtual mixture-arm $\widetilde{\nu}_{i_\star}$, and its representation $\widetilde{\nu}^{\langle d_i \rangle}_{i_\star}$ is added to ${\mathcal A}^{\langle d_i \rangle}$. For any $i > i_\star$, we then have 
	\begin{align}
    \Delta T_i \cdot  \E \left[ \left( \langle \theta_\star, a_\star \rangle -\langle \theta_{\star}^{\langle d_i \rangle}, a_{\star}^{\langle d_i \rangle} \rangle \right) \right]  & \leq \Delta T_i \cdot \E \left[\left(\langle \theta_\star, a_\star \rangle - \langle \theta_{\star}^{\langle d_i \rangle}, \widetilde{\nu}^{\langle d_i \rangle}_{i_\star} \rangle \right) \right]  \nonumber  \\
    & = \Delta T_i \cdot \left(\langle \theta_\star, a_\star \rangle -  \widetilde{\mu}_{i_\star} \right) \label{ms_linear:eq:regret_approximation_equivalent_form}  \\
	& = \frac{\Delta T_i}{\Delta T_{i_\star}}\cdot  R_{\Delta T_{i_\star}} \label{ms_linear:eq:regret_approximation_expected_reward} \\
	& = \frac{\Delta T_i}{\frac{2^{2p}}{T^\alpha }} \cdot {O} \left(\log^{5/2} \left( KT\log T\right) \cdot T^\beta  \right)\label{ms_linear:eq:regret_approximation_i_star} \\
	& = \frac{ {O} \left(  \log^{5/2} \left( KT\log T\right)  \cdot T^{1 + \alpha + \beta} \right) }{{2^{2p}}} \label{ms_linear:eq:regret_approximation_important} \\
	& = {O} \left( \log^{5/2} \left( KT\log T\right) \cdot T^{1  + \alpha - \beta} \right), \label{ms_linear:eq:regret_approximation}
	\end{align}
	where \cref{ms_linear:eq:regret_approximation_equivalent_form} comes from the formulation of the modified linear bandit problem; \cref{ms_linear:eq:regret_approximation_expected_reward} comes from that fact that $\widetilde{\mu}_j = \E [\widetilde{\mu}_{j \vert \cF_j}]=\langle \theta_\star, a_\star \rangle -  R_{\Delta T_j}/\Delta T_j$ derived from \cref{ms_linear:eq:expected_reward_virtual};
	\cref{ms_linear:eq:regret_approximation_i_star} comes from the bound in \cref{ms_linear:eq:regret_i0} with $i=i_\star$; \cref{ms_linear:eq:regret_approximation_important} comes from the fact that $\Delta T_i \leq T$ and some rewriting; \cref{ms_linear:eq:regret_approximation} comes from the fact that $p = \lceil \log_2 T^{\beta} \rceil \geq \log_2 T^{\beta}$.
	
Combining \cref{ms_linear:eq:regret_approximation} and \cref{ms_linear:eq:regret_decomposition_internal_expectation} for cases when $i > i_\star$ (or the corner case algorithm stops before $T_{i_\star}$ and incurs no regret in iterations $i \geq i_\star$), and together with \cref{ms_linear:eq:regret_i0} for cases when $i \leq i_\star$, we have that $\forall i \in [p]$,
\begin{align}
    R_{\Delta T_i}  & =  {O} \left( \log^{5/2} \left( KT\log T\right) \cdot T^{\max\{\beta, 1  + \alpha - \beta \}} \right) . \nonumber 
\end{align}

Since the cumulative regret is non-decreasing in $t$, we have
\begin{align}
	R_{T} & \leq \sum_{i=1}^p R_{\Delta T_i} \nonumber \\
	& = \sum_{i=1}^p {O} \left( \log^{5/2} \left( KT\log T\right) \cdot T^{\max\{\beta, 1  + \alpha - \beta \}} \right) \nonumber \\
    & = {O} \left( \log^{7/2} \left( KT\log T\right) \cdot T^{\max\{\beta, 1  + \alpha - \beta \}} \right),\nonumber
\end{align}
where we use the fact that $p = \lceil \log_2(T^\beta) \rceil = O(\log T)$. Our results follows after noticing $R_T \leq T$ is a trivial upper bound.
\end{proof}

\subsubsection{Proof of \cref{ms_linear:thm:pareto}}

\pareto*

\begin{proof}
From \cref{ms_linear:thm:linucbPlus}, we know that the rate in \cref{ms_linear:eq:pareto_rate} is achieved by \cref{ms_linear:alg:linucbPlus} with input $\beta$. We only need to prove that no other algorithms achieve strictly smaller rates in pointwise order.

Suppose, by contradiction, we have $\theta^\prime$ achieved by an algorithm such that $\theta^\prime(\alpha) \leq \theta_{\beta}(\alpha)$ for all $\alpha \in [0, 1]$ and $\theta^\prime(\alpha_0) < \theta(\alpha_0)$ for at least one $\alpha_0 \in [0, 1]$. We then must have $\theta^\prime(0) \leq \theta_{\beta}(0) = \beta$. We consider the following two exclusive cases.

\textbf{Case 1: $\theta^\prime(0) = \beta$.} According to \cref{ms_linear:thm:rate_lower_bound}, we must have $\theta^\prime \geq \theta_\beta$, which leads to a contradiction.

\textbf{Case 2: $\theta^\prime(0) = \beta^\prime < \beta$.} According \cref{ms_linear:thm:rate_lower_bound}, we must have $\theta^\prime \geq \theta_{\beta^\prime}$. However, $\theta_{\beta^\prime}$ is not strictly better than $\theta_{\beta}$, e.g., $\theta_{\beta^\prime}(2\beta - 1) =  2\beta - \beta^\prime > \beta = \theta_{\beta}(2\beta - 1)$, which also leads to a contradiction.
\end{proof}

\subsection{Proofs and Supporting Results for \cref{ms_linear:sec:remove_assumption}}
\label{ms_linear:app:remove_assumption}

\subsubsection{Discussion on \cref{ms_linear:alg:linucbPlus_modified}}
We construct the following two (smoothed) base algorithms \citep{pacchiano2020model} at each iteration of \linucbPlus: (1) a \linucb algorithm that works with truncated feature representations in $\R^{d_i}$, with possible mis-specifications; and (2) a \ucbalg algorithm that works only with virtual mixture-arms, if there exists any. We use \smoothCorral from \cite{pacchiano2020model} as the master algorithm and always optimally tune it with respect to the \linucb base, i.e., set the learning rate as $\eta = 1/\sqrt{d_i \Delta T_i}$. For iterations such that $d_i \geq d_\star$, the \linucb is the optimal base and we incur $\widetilde{O}(\sqrt{d_i \Delta T_i}) = \widetilde{O}(T^\beta)$ regret; a good enough virtual mixture-arm $\widetilde{\nu}_{i_\star}$ is then constructed as before. For later iterations such that $d_i < d_\star$, \smoothCorral incurs regret $\widetilde{O}(\max \{T^{1+\alpha - \beta}, T^{\beta}\})$ thanks to guarantees of the \ucbalg base: the $\widetilde{O}(T^{1+\alpha - \beta})$ term is due to the approximation error and the $\widetilde{O}(T^\beta)$ term is due to the learning error. Although the learning error of \ucbalg is enlarged from $\widetilde{O}(T^{1/2})$ to $\widetilde{O}(T^\beta)$, as \smoothCorral is always tuned with respect to the \linucb base, this won't affect the resulted Pareto optimality.

\subsubsection{Proof of \cref{ms_linear:thm:linucbPlus_modified}}

\thmLinUCBPlusModified* 

\begin{proof}

At each iteration $i \in [p]$ of \linucbPlus, we applying \smoothCorral as the master algorithm with two smoothed base algorithms: (1) a \linucb algorithm that works with truncated feature representations in $\R^{d_i}$, with possible mis-specifications; and (2) a \ucbalg algorithm that works only with virtual mixture-arms, if there exists any. The learning rate of \smoothCorral is always optimally tuned with respect to the \linucb base, i.e., $\eta = 1/\sqrt{d_i \Delta T_i}$. Since there are at most $p = O(\log T)$ iterations, we only need to bound the expected regret at each iteration $R_{\Delta T_i}$. As before, we use $i_\star \in [p]$ to denote the largest integer such that $d_{i_\star} \geq d_\star$.

For $i \leq i_\star$, the \linucb base works on a well-specified linear bandit problem. Theorem 5.3 in \cite{pacchiano2020model} gives the following guarantees:
\begin{align*}
    R_{\Delta T_i} = \widetilde{O} \left(\sqrt{\Delta T_i} + \eta^{-1} + \Delta T_i \eta + \Delta T_i d_i \eta \right) = \widetilde{O} \left(\sqrt{d_i \Delta T_i} \right) = \widetilde{O} \left(T^\beta \right).
\end{align*}
Good enough virtual mixture-arm $\widetilde{\nu}_{i_\star}$ is then constructed with conditional expectation $\widetilde{\mu}_{i_\star \vert \cF_{i_\star}} = \E [\widetilde{\nu}_{i_\star} \vert \cF_{i_\star}] = \langle \theta_\star, a_\star \rangle -  \widehat R_{\Delta T_{i_\star}}/\Delta T_{i_\star}$.

We now analyze the regret incurred for iteration $i > i_\star$. Conditioning on past information $\cF_{i-1}$ and let $r(\pi_t)$ denote the (conditional) expected reward of applying policy $\pi_t$, we have 
\begin{align}
    R_{\Delta T_i \vert \cF_{i-1}} & = \Delta T_i \cdot   \left( \langle \theta_\star, a_\star \rangle -\widetilde{\mu}_{i_\star \vert \cF_{i_\star}} \right) + \E \left[ \sum_{t \text{ in iteration } i} \widetilde{\mu}_{i_\star\vert \cF_{i_\star} } - r(\pi_t) \, \bigg\vert \, {\mathcal F}_{i-1}  \right] \nonumber \\
    & = \Delta T_i \cdot   \left( \langle \theta_\star, a_\star \rangle -\widetilde{\mu}_{i_\star \vert \cF_{i_\star}} \right) + \widetilde{O} \left(\sqrt{\Delta T_i} + \eta^{-1} + \Delta T_i \eta + \Delta T_i \eta \right) \nonumber ,
\end{align}
where the second term comes from the guarantee of \smoothCorral with respect to the \ucbalg base. Taking expectation over randomness in $\cF_{i-1}$ leads to
\begin{align}
      R_{\Delta T_i }  & = \widetilde{O}\left(T^{1+\alpha - \beta} \right) + \widetilde{O}\left( T^\beta \right) \nonumber ,
\end{align}
where the first term follows from a similar analysis as in \cref{ms_linear:eq:regret_approximation}, and the second term follows by setting $\eta = 1/\sqrt{d_i \Delta T_i}$. A similar analysis as in \cref{ms_linear:thm:pareto} thus show \cref{ms_linear:alg:linucbPlus_modified} is Pareto optimal, even without \cref{ms_linear:assumption:action_set}.
\end{proof}

\subsubsection{Discussion on \smoothCorral}
\label{ms_linear:app:corral}

\cite{pacchiano2020model} tackles the model selection problem in linear bandit by applying \smoothCorral with $O(\log d)$ base \linucb learners working with different dimensions $d_i \in \{2^0, 2^1, \dots, 2^{\floor*{\log d}}\}$. Let $d_{i_\star}$ denote the smallest dimension that satisfies $d_{i_\star} \geq d_\star$. With respect to the base \linucb working on the first $d_{i_\star}$ dimensions, \smoothCorral enjoys regret guarantee
\begin{align}
    R_{T} = \widetilde{O} \left(\sqrt{T} + \eta^{-1} + T \eta + T d_\star \eta \right) \nonumber.
\end{align}
\smoothCorral then achieves the rate function in \cref{ms_linear:eq:pareto_rate} by setting the learning rate $\eta = T^{-\beta}$ (and also noticing that $d_\star \leq T^{\alpha}$).

\subsection{Other Details for Experiments}
\label{ms_linear:app:experiment}
\begin{figure}[H]
     \centering
     \subfloat[]{\includegraphics[width=.5\textwidth]{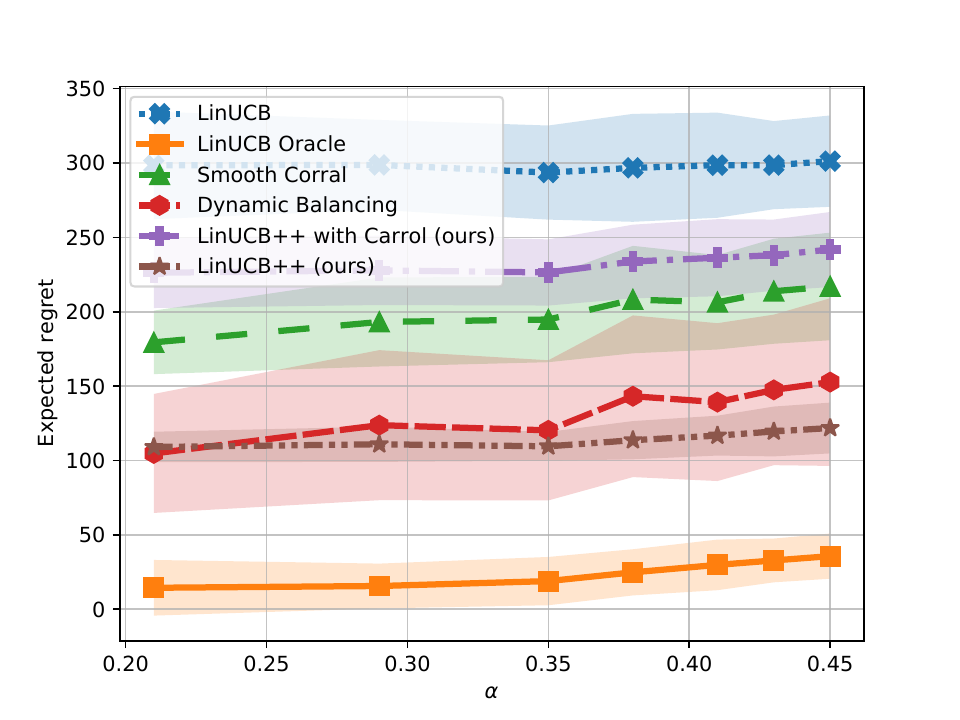}\label{ms_linear:fig:decay}}
     \subfloat[]{\includegraphics[width=.5\textwidth]{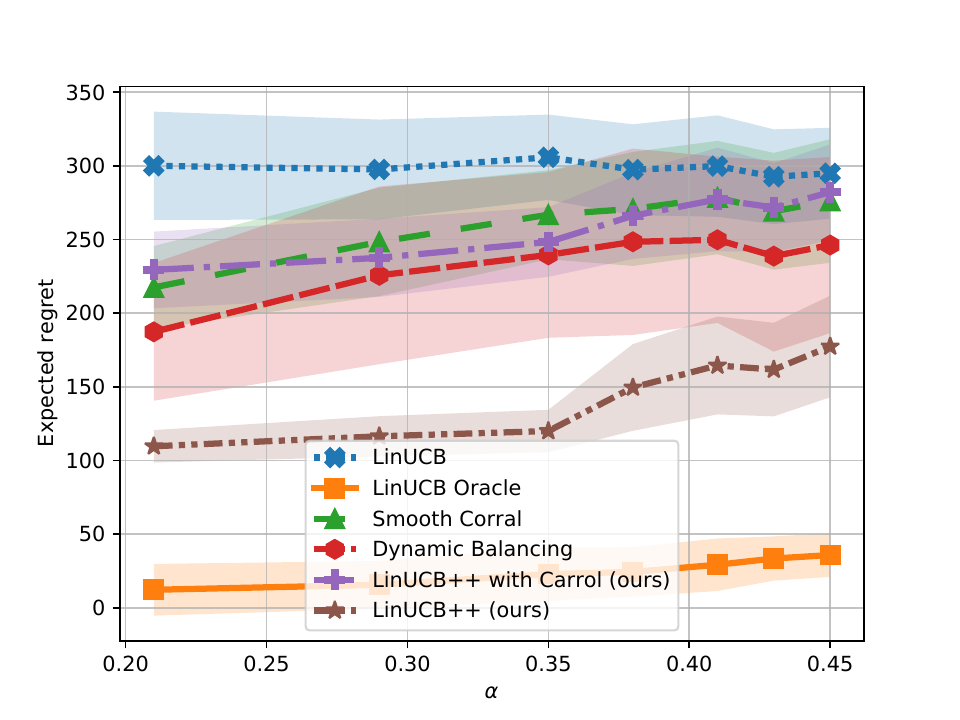}\label{ms_linear:fig:flip}}
     \caption{Similar experiment setups to those shown in \cref{ms_linear:fig:non_expressive_alpha}, but with different reward parameters $\theta_\star$.}
     \label{ms_linear:fig:additional}
\end{figure}

We conduct additional experiments with setups similar to the ones shown in \cref{ms_linear:fig:non_expressive_alpha}, but with different reward parameters $\theta_\star$. 
We set $\theta_\star$ as (the normalized version of) $[\frac{1}{\sqrt{1}}, \frac{1}{\sqrt{2}}, \dots, \frac{1}{\sqrt{d_\star}}, 0, \dots, 0]^\top \in \R^d$ in \cref{ms_linear:fig:decay}; and $\theta_\star$ as (the normalized version of) $[\frac{1}{\sqrt{d_\star}}, \frac{1}{\sqrt{d_\star - 1}}, \dots, \frac{1}{\sqrt{1}}, 0, \dots, 0]^\top \in \R^d$ in \cref{ms_linear:fig:flip}. 
With $\theta_\star$ selected in \cref{ms_linear:fig:decay}, \dynamicBalancing shows comparable performance to \linucbPlus in terms of averaged regret (but with larger variance). \linucbPlus outperforms \dynamicBalancing when $\theta_\star$ is ``flipped'' (i.e., the one used in \cref{ms_linear:fig:flip}) but with the same intrinsic dimension $d_\star$.

\chapter{Model Selection in Best Action Identification}
\label{chapter:model:bai}

We introduce the model selection problem in pure exploration linear bandits, where the learner needs to adapt to the instance-dependent complexity measure of the smallest hypothesis class containing the true model. We design algorithms in both fixed confidence and fixed budget settings with near instance optimal guarantees. The core of our algorithms is a new optimization problem based on experimental design that leverages the geometry of the action set to identify a near-optimal hypothesis class. Our fixed budget algorithm is developed based on a novel selection-validation procedure, which provides a new way to study the understudied fixed budget setting (even without the added challenge of model selection). We adapt our algorithms, in both fixed confidence and fixed budget settings, to problems with model misspecification.

\section{Introduction}
\label{ms_bai:sec:intro}

The pure exploration linear bandit problem considers a set of arms whose expected rewards are linear in their \emph{given} feature representation, and aims to identify the optimal arm through adaptive sampling. 
Two settings, i.e., fixed confidence and fixed budget settings, are studied in the literature.
In the fixed confidence setting, the learner continues sampling arms until a desired confidence level is reached, and the goal is to minimize the total number of samples \citep{soare2014best, xu2018fully, tao2018best, fiez2019sequential, degenne2020gamification, katz2020empirical}. 
In the fixed budget setting, the learner is forced to output a recommendation within a pre-fixed sampling budget, and the goal is to minimize the error probability \citep{hoffman2014correlation, katz2020empirical, alieva2021robust, yang2021towards}.
Applications of pure exploration linear bandits include content recommendation, digital advertisement and A/B/n testing (see aforementioned papers for more discussions on applications).

All existing works, however, focus on linear models with the \emph{given} feature representations and fail to adapt to cases when the problem can be explained with a much simpler model, i.e., a linear model based on a subset of the features. In this chapter, we introduce the model selection problem in pure exploration linear bandits. We consider a sequence of nested linear hypothesis classes $\cH_1 \subseteq \cH_2 \subseteq \dots \subseteq \cH_D$ and assume that $\cH_{d_\star}$ is the smallest hypothesis class that contains the true model. Our goal is to automatically adapt to the complexity measure related to $\cH_{d_\star}$, for an unknown $d_\star$, rather than suffering a complexity measure related to the largest hypothesis class $\cH_D$. 

The model selection problem appears ubiquitously in real-world applications. In fact, cross-validation \citep{stone1974cross, stone1978cross}, a practical method for model selection, appears in almost all successful deployments of machine learning models. The model selection problem was recently introduced to the bandit regret minimization setting by \citet{foster2019model}, and further analyzed by \citet{pacchiano2020model, zhu2022pareto}. \citet{zhu2022pareto} prove that only Pareto optimality can be achieved for regret minimization, which is even weaker than minimax optimality. We introduce the model selection problem in the pure exploration setting and, surprisingly, show that it is possible to design algorithms with \emph{near optimal instance-dependent complexity} for both fixed confidence and fixed budget settings. We further generalize the model selection problem to the regime with misspecified linear models, and show our algorithms are robust to model misspecification.

\subsection{Contribution and Organization}
\label{ms_bai:sec:contribution}

We briefly summarize our contributions as follows:

\begin{itemize}
    \item We introduce the model selection problem for pure exploration in linear bandits in \cref{ms_bai:sec:setting}, and analyze its instance-dependent complexity measure. We provide a general framework to solve the model selection problem for pure exploration linear bandits. Our framework is based on a carefully-designed two-dimensional doubling trick and a new optimization problem that leverages the geometry of the action set to efficiently identify a near-optimal hypothesis class. 

    \item In \cref{ms_bai:sec:fixed_confidence}, we provide an algorithm for the fixed confidence setting with near optimal instance-dependent unverifiable sample complexity. We additionally provide evidence on why one cannot verifiably output recommendations.
    
    \item In \cref{ms_bai:sec:fixed_budget}, we provide an algorithm for the fixed budget setting, which applies a novel selection-validation trick to bandits. Its probability of error matches (up to logarithmic factors) the probability error of an algorithm that chooses its sampling allocation based on knowledge of the true model parameter. In addition, the guarantee of our algorithm is nearly optimal even in the non-model-selection case, and our algorithm also provides a new way to analyze the \emph{understudied} fixed budget setting.
    
    \item We further generalize the model selection problem into the misspecified regime in \cref{ms_bai:sec:misspecification}, and adapt our algorithms to both the fixed confidence and fixed budget settings. Our algorithms reach an instance-dependent sample complexity measure that is relevant to the complexity measure of a closely related perfect linear bandit problem.
\end{itemize}

\section{Problem Setting}
\label{ms_bai:sec:setting}

In the transductive linear bandit pure exploration problem, the learner is given an action set $\cX \subset \R^D$ and a target set $\cZ \subset \R^D$. The expected reward of any arm $x \in \cX \cup \cZ$ is linearly parameterized by an unknown reward vector $\theta_\star \in \Theta \subseteq \R^D$, i.e., $h(x) = \ang*{\theta_\star, x}$. The parameter space $\Theta$ is known to the learner. 
At each round $t$, the learner/algorithm $\alg$ selects an action $X_t \in \cX$, and observes a noisy reward $R_t = h(X_t) + \xi_t$, where $\xi_t$ represents an additive $1$-sub-Gaussian noise.
The action $X_t \in \cX$ can be selected with respect to the history $\cF_{t-1} = \sigma((X_i, R_i)_{i<t})$ up to time $t$.
The goal is to identify the unique optimal arm $z_\star = \argmax_{z \in \cZ} h(z)$ from the target set $\cZ$. We assume $\Theta \subseteq \spn (\cX)$ to obtain unbiased estimators for arms in $\cZ$. Without loss of generality, we assume that $\spn(\cX) = \R^D$ (otherwise one can project actions into a lower dimensional space). We further assume that $\spn(\curly*{z_\star - z}_{z \in \cZ}) = \R^D$ for technical reasons. We consider both fixed confidence and fixed budget settings in this chapter.

\begin{definition}[Fixed confidence]
\label{ms_bai:def:fixed_confidence}
Fix $\cX, \cZ, \Theta \subseteq \R^D$. An algorithm $\alg$ is called $\delta$-PAC for $(\cX, \cZ, \Theta)$ if (1) the algorithm has a stopping time $\tau$ with respect to $\curly*{\cF_t}_{t\in \N}$ and (2) at time $\tau$ it makes a recommendation $\widehat z \in \cZ$ such that $\P_{\theta_\star} \paren*{ \widehat z = z_\star } \geq 1 - \delta$ for all $\theta_\star \in \Theta$.
\end{definition}

\begin{definition}[Fixed budget]
\label{ms_bai:def:fixed_budget}
Fix $\cX, \cZ, \Theta \subseteq \R^D$ and a budget $T$. A fixed budget algorithm $\alg$ returns a recommendation $\widehat z \in \cZ$ after $T$ rounds.
\end{definition}

\paragraph{The model selection problem} 
The learner is given a nested sequence of parameter classes $\Theta_1 \subseteq \Theta_2 \subseteq \dots \subseteq \Theta_D$, where $\Theta_{d} \ldef \curly*{\theta \in \R^D: \theta_i = 0 , \forall i > d}$ is the set of parameters such that for any $\theta \in \Theta_{d}$, it only has non-zero entries on its first $d$ coordinates.\footnote{A nested sequence of linear hypothesis classes $\cH_1 \subseteq \cH_2 \subseteq \dots \subseteq \cH_D$ can be constructed based on the nested sequence of parameter classes $\Theta_1 \subseteq \Theta_2 \subseteq \dots \subseteq \Theta_D$, i.e., $\cH_d \ldef \curly{h(\cdot)=\ang{\theta, \cdot}: \theta \in \Theta_d}$.} 
We assume that $\theta_\star \in \Theta_{d_\star}$ for an \emph{unknown} $d_\star$. We call $d_\star$ the intrinsic dimension of the problem and it is set as the index of the smallest parameter space containing the true reward vector. One interpretation of the intrinsic dimension is that only the first $d_\star$ features (of each arm) play a role in predicting the expected reward. Our goal is to automatically adapt to the sample complexity with respect to the intrinsic dimension $d_\star$, rather than suffering from the sample complexity related to the ambient dimension $D$. In the following, we write $\theta_\star \in \Theta_{d_\star}$ to indicate that the problem instance has intrinsic dimension $d_\star$. Besides dealing with the \emph{well-specified} linear bandit problem as defined in this section, we also extend our framework into the \emph{misspecified} setting in \cref{ms_bai:sec:misspecification}, with additional setups introduced therein.

\paragraph{additional notations} For any $x = [x_1, x_2, \dots, x_D]^\top \in \R^D$ and $d \leq D$, we use $\psi_{d}(x) \ldef  [x_1, x_2, \dots, x_{d}]^\top \in \R^{d}$ to denote the truncated feature representation that only keeps its first $d$ coordinates. We also write $\psi_{d}(\cX) \ldef \curly*{\psi_{d}(x): x \in \cX}$ and $\psi_{d}(\cZ) \ldef \curly*{\psi_{d}(z): z \in \cZ}$ to represent the truncated action set and target set, respectively. Note that we necessarily have $\psi_d (\cZ) \subseteq \spn \paren*{\psi_d(\cX)} = \R^{d}$ as long as $\cZ \subseteq \spn \paren*{\cX} = \R^D$. We use $\cY(\psi_d(\cZ)) \ldef \curly*{ \psi_d(z) - \psi_d(z^\prime): z, z^\prime \in \cZ }$ to denote all possible directions formed by subtracted one item from another in $\psi_d(\cZ)$; and use $\cY^\star(\psi_d(\cZ)) \ldef \curly*{ \psi_d(z_\star) - \psi_d(z): z \in \cZ }$ to denote all possible directions with respect to the optimal arm $z_\star$. For any $z \in \cZ$, we use $\Delta_z \ldef h(z_\star) - h(z)$ to denote its sub-optimality gap; we set $\Delta_{\min} \ldef \min_{z \in \cZ \setminus \curly*{z_\star}} \Delta_z$. As in \citet{fiez2019sequential}, we assume $\max_{z \in \cZ} \Delta_z \leq 2$ when analyzing upper bounds. \linebreak
We denote $\cS_k \ldef \curly*{z \in \cZ: \Delta_z < 4 \cdot 2^{-k}}$ (with $\cS_1 \ldef \cZ$). 
We use 
$\simp_{\cX} = \simp \prn{\cX} \ldef \curly*{ \lambda \in \R^{\abs{\cX}}: \sum_{x \in \cX} \lambda_x = 1, \lambda_x \geq 0  }$ to denote the $(\abs*{\cX}-1)$-dimensional \linebreak
simplex over actions. For any (continuous) design $\lambda \in \simp_{\cX}$, we use $A_{d}(\lambda) \ldef \sum_{x \in \cX} \lambda_x \, \psi_d(x)  \paren*{\psi_d(x)}^\top \in \R^{d \times d}$ to denote the design matrix with respect to $\lambda$. For any set $\cW \subseteq \R^D$, we denote $\iota(\cW) \ldef \inf_{\lambda \in \simp_{\cX}} \sup_{w \in \cW} \norm{w}^2_{A_d(\lambda)^{-1}}$.\footnote{A generalized inversion is used for singular matrices. See \cref{ms_bai:app:inverse} for detailed discussion.}

\section{Towards the True Sample Complexity}
\label{ms_bai:sec:true_complexity}
The instance-dependent sample complexity lower bound for linear bandit is discovered/analyzed in previous papers \citep{soare2014best, fiez2019sequential, degenne2019pure}. We here consider related quantities that take our model selection setting into consideration. For any $d \in [D]$, we define
\begin{align}
\label{ms_bai:eq:rho}
    \rho^\star_{d} \ldef \inf_{\lambda \in \simp_{\cX}}  \sup_{z \in \cZ \setminus \curly*{z_\star }} \frac{\norm*{\psi_d(z_{\star})-\psi_d(z)}^2_{A_{d}(\lambda)^{-1}}}{(h(z_\star) - h(z))^2},
\end{align}
and 
\begin{align}
\label{ms_bai:eq:iota}
    \iota_{d}^\star \ldef \inf_{\lambda \in \simp_{\cX}}  \sup_{z \in \cZ \setminus \curly*{z_\star}} \norm{\psi_d(z_\star) - \psi_d(z)}_{A_{d}(\lambda)^{-1}}^2.
\end{align}

Following analysis in \citet{fiez2019sequential},
we provide a lower bound for the model selection problem $(\cX, \cZ$, $\theta_\star \in \Theta_{d_\star})$ in the fixed confidence setting as follows.

\begin{restatable}{theorem}{thmLowerBoundDeltaPAC}
\label{ms_bai:thm:lower_bound_delta_PAC}
Suppose $\xi_t \sim \cN(0,1)$ for all $t \in \N_+$ and $\delta \in (0, 0.15]$. Any $\delta$-PAC algorithm with respect to $(\cX, \cZ$, $\theta_\star \in \Theta_{d_\star})$ with stopping time $\tau$ satisfies $\E_{\theta_\star} \sq*{\tau} \geq \rho^\star_{d_\star} \log(1/2.4 \delta)$.
\end{restatable}

The above lower bound only works for $\delta$-PAC algorithms, but not for algorithms in the fixed budget setting or with unverifiable sample complexity (see \cref{ms_bai:sec:fixed_confidence}). We now introduce another lower bound for the best possible \emph{non-interactive} algorithm $\alg$. Following the discussion in \citet{katz2020empirical}, we consider any non-interactive algorithm as follows: The algorithm $\alg$ chooses an allocation $\curly*{x_1, x_2, \dots, x_N} \subseteq \cX$ and receive rewards $\curly*{r_1, r_2, \dots, r_N} \subseteq \R$ where $r_i$ is sampled from $\cN(h(x_i), 1)$. The algorithm then recommends $\widehat z = \argmax_{z \in \cZ} \ang*{\widehat \theta_d, z}$ where $\widehat \theta_d = \argmin_{\theta \in \R^d} \sum_{i=1}^N (r_i - \theta^\top \psi_d(x_i))^2$ is the least squares estimator in $\R^d$. The learner is allowed to choose any allocations, \emph{even with the knowledge of $\theta_\star$}, and use any feature mapping such that linearity is preserved, i.e., $d_\star \leq d \leq D$.

\begin{restatable}{theorem}{thmLowerBoundNonInteractive}
\label{ms_bai:thm:lower_bound_non_interactive}
Fix $(\cX, \cZ$, $\theta_\star \in \Theta_{d_\star})$ and $\delta \in (0,0.015]$. Any non-interactive algorithm $\alg$ using a feature mappings of dimension $d \geq d_\star$ makes a mistake with probability at least $\delta$ as long as it uses no more than $\frac{1}{2}  \rho^\star_{d_\star} \log(1/\delta)$ samples.
\end{restatable}

The above lower bound serves as a fairly strong baseline due to the power provided to the non-interactive learner, i.e., the knowledge of $\theta_\star$. 
\cref{ms_bai:thm:lower_bound_non_interactive} indicates (for any non-interactive learner) (1) sample complexity lower bound $\widetilde \Omega(\rho^\star_{d_\star})$ in fixed confidence setting; and (2) error probability lower bound $\Omega(\exp(- T/\rho^\star_{d_\star}))$ in fixed budget setting: Suppose the budget is $T$, one would expect an error probability at least $\Omega(\exp(- T/ \rho^\star_{d_\star}))$ by relating $\frac{1}{2}  \rho^\star_{d_\star} \log(1/\delta)$ to $T$.

Note that all lower bounds are with respect to $\rho^\star_{d_\star}$ rather than $\rho^\star_{d}$ for $d > d_\star$ due to the assumption $\theta_\star \in \Theta_{d_\star}$ for the model selection problem. Our goal is to automatically adapt to the complexity $\rho_{d_\star}^\star$ without knowledge of $d_\star$. The following proposition shows the monotonic relation among $\curly*{\rho_d^\star}_{d = d_\star}^D$.

\begin{restatable}{proposition}{propRhoMonotonic}
\label{ms_bai:prop:rho_monotonic}
The monotonic relation $\rho^\star_{d_1} \leq \rho^\star_{d_2}$ holds true for any $d_\star \leq d_1 \leq d_2 \leq D$.
\end{restatable}
The intuition behind \cref{ms_bai:prop:rho_monotonic} is that the model class $\Theta_{d_2}$ is a superset of $\Theta_{d_1}$ and therefore identifying $z_\star$ in $\Theta_{d_2}$ requires ruling out a larger set of statistical alternatives than in $\Theta_{d_1}$. While \cref{ms_bai:prop:rho_monotonic} is intuitive, its proof is surprisingly technical and involves showing the equivalence of a series of optimization problems.

\subsection{Failure of Standard Approaches}

\begin{restatable}{proposition}{propRhoStarDifferentD}
\label{ms_bai:prop:rho_star_different_d}
For any $\gamma > 0$, there exists an instance $(\cX, \cZ$, $\theta_\star \in \Theta_{d_\star})$ such that $\rho^\star_{d_\star + 1} > \rho^\star_{d_\star} + \gamma$ yet $\iota_{d_\star + 1}^\star \leq 2 \iota_{d_\star}^\star $.
\end{restatable}

One may attempt to solve the model selection problem with a standard doubling trick over dimension, i.e., truncating the feature representations at dimension $d_i = 2^i$ for $i \leq \ceil*{\log_2 D}$ and gradually exploring models with increasing dimension. This approach, however, is directly ruled out by \cref{ms_bai:prop:rho_star_different_d} since such doubling trick could end up with solving a problem with a dimension $d^\prime \leq 2 d_\star$ yet $\rho_{d^\prime}^\star \gg \rho_{d_\star}^\star$. Although doubling trick over dimensions is commonly used to provide \emph{worst-case} guarantees in regret minimization settings \citep{pacchiano2020model, zhu2022pareto}, we emphasize here that matching \emph{instance-dependent} complexities is important in pure exploration setting \citep{soare2014best, fiez2019sequential, katz2020empirical}. Thus, new techniques need to be developed. \cref{ms_bai:prop:rho_star_different_d} also implies that trying to infer the value of $\rho^\star_{d}$ from  $\iota^\star_d$ can be quite misleading. And thus conducting a doubling trick over $\iota^\star_d$ (or an upper bound of it) is likely to fail as well.

\paragraph{Importance of model selection} \cref{ms_bai:prop:rho_star_different_d} also illustrates the importance and necessity of conducting model selection in pure exploration linear bandits. Consider the hard instance used in constructed in \cref{ms_bai:prop:rho_star_different_d} and set $D= d_\star + 1$. All existing algorithms \citep{soare2014best, fiez2019sequential, degenne2019pure, katz2020empirical} that directly work with the \emph{given} feature representation in $\R^D$ end up with a complexity measure scales with $\rho^\star_{D}$, which could be arbitrarily large than the true complexity measure $\rho_{d_\star}^\star$ and even become vacuous (by sending $\gamma \rightarrow \infty$).

\paragraph{Our approaches} In this chapter, we design a more sophisticated doubling scheme over a two-dimensional grid corresponding to the number of elimination steps and the richest hypothesis class considered at each step. We design subroutines for both fixed confidence and fixed budget settings. Our algorithms define a new optimization problem based on experimental design that leverages the geometry of the action set to efficiently identify a near-optimal hypothesis class. Our fixed budget algorithm additionally uses a novel application of a selection-validation trick in bandits. Our guarantees are with respect to the true instance-dependent complexity measure $\rho_{d_\star}^\star$.

\section{Fixed Confidence Setting}
\label{ms_bai:sec:fixed_confidence}

We present our main algorithm (\cref{ms_bai:alg:doubling_fixed_confidence}) for the fixed confidence setting in this section.
\cref{ms_bai:alg:doubling_fixed_confidence} invokes \gemsc (\cref{ms_bai:alg:subroutine_fixed_confidence}) as subroutines and starts to output the optimal arm after $\widetilde O(\rho^\star_{d_\star} + d_\star)$ samples.
Our sample complexity matches, up to an additive $d_\star$ term and logarithmic factors, the strong baseline developed in \cref{ms_bai:thm:lower_bound_non_interactive}.

We first introduce the subroutine \gemsc, which runs for $n$ rounds and takes (roughly) $B$ samples per-round.
\gemsc is built on \rage \citep{fiez2019sequential}, a standard linear bandit pure exploration algorithm works in the ambient space $\R^D$.
The key innovation of \gemsc lies in \emph{adaptive} hypothesis class selection at each round (i.e., selecting $d_k$), which allows us to adapt to the instrinsic dimension $d_\star$.
After selecting the working dimension $d_k$ at round $k$, \gemsc allocates samples based on optimal design (in $\R^{d_k}$); it then eliminate sub-optimal arms based on the estimated rewards constructed using least squares.
Following \citet{fiez2019sequential}, we use a rounding procedure $\round (\lambda,N,d,\zeta)$ to round a continuous experimental design $\lambda \in \simp_{\cX}$ into integer allocations over actions. 
We use $r_d(\zeta)$ to denote the number of samples needed for such rounding in $\R^d$ with approximation factor $\zeta$. 
One can choose $r_d(\zeta) = (d^2+d+2)/\zeta$ \citep{pukelsheim2006optimal, fiez2019sequential} or $r_d(\zeta) = 180d/\zeta^2$ \citep{allen2020near}.
We choose $\zeta$ as a constant throughout this chapter, e.g., $\zeta = 1$.
When $N \geq r_d(\zeta)$, there exist computationally efficient rounding procedures that output an allocation $\curly{x_1, x_2, \dots, x_N}$ satisfying
\begin{align}
    & \max_{y \in \cY(\psi_d(\cZ))}  \norm{y}^2_{\paren{\sum_{i=1}^N \psi_d(x_i) \psi_d(x_i)^{\top}}^{-1}} \leq  \nonumber \\
    &(1+\zeta) \max_{y \in \cY(\psi_d(\cZ))} \norm{y}^2_{\paren{\sum_{x \in \cX} \lambda_x \psi_d(x) \psi_d(x)^\top}^{-1}} / N. \label{ms_bai:eq:rounding}
\end{align}

\begin{algorithm}[]
	\caption{\gemsc Gap Elimination with Model Selection (Fixed Confidence)}
	\label{ms_bai:alg:subroutine_fixed_confidence} 
	\renewcommand{\algorithmicrequire}{\textbf{Input:}}
	\renewcommand{\algorithmicensure}{\textbf{Output:}}
	\begin{algorithmic}[1]
		\REQUIRE Number of iterations $n$, budget for dimension selection $B$ and confidence parameter $\delta$.
		\STATE Set $\widehat \cS_1 = \cZ$.
		\FOR {$k = 1, 2, \dots, n$}
		\STATE Set $\delta_k = \delta/k^2$.
		\STATE Define $g_k(d) \ldef \max \curly{ 2^{2k} \, \iota(\cY(\psi_d(\widehat \cS_k))), r_d(\zeta) }$. 
		\STATE Get $d_k = \text{\OPT}(B, D, g_k(\cdot))$, where $d_k \leq D$ is largest dimension such that $g_k(d_k) \leq B$ (see \cref{ms_bai:eq:opt_d_selection} for the detailed optimization problem); set $\lambda_k$ be the optimal design of the optimization problem\\
		$\inf_{\lambda \in \simp_{\cX}} \sup_{z, z^\prime \in \widehat \cS_k} \norm*{\psi_{d_k}(z)-\psi_{d_k}(z^\prime)}^2_{A_{d_k}(\lambda)^{-1}}$;\\
		set $ N_k = \ceil{g(d_k) 2(1 + \zeta)  \log(\abs{\widehat \cS_k}^2/\delta_k)}.$
		\STATE Get allocation \\
		$\curly*{x_1, \ldots, x_{N_k} } = \text{\round}(\lambda_k,N_k, d_k, \zeta)$.
		\STATE Pull arms $\curly{x_1, \ldots, x_{N_k}} $ and receive rewards $\curly{r_1, \ldots, r_{N_k}}$.
        \STATE Set $\widehat{\theta}_k = A_k^{-1} b_k \in \R^{d_k}$, \\
        where $A_k = \sum_{i=1}^{N_k} \psi_{d_k}(x_i) \psi_{d_k}(x_i)^\top$, \\
        and $b_k =  \sum_{i=1}^{N_k} \psi_{d_k}(x_i) b_i$.
        \STATE Set 
        $\widehat \cS_{k+1} = \widehat \cS_k \setminus \{z \in \widehat \cS_k : 
        \exists z^\prime \text{ s.t. } \ang{\widehat{\theta}_k, \psi_{d_k}(z^\prime) - \psi_{d_k}(z) } \geq \omega(z^\prime, z) \}$, where $\omega(z^\prime, z) \ldef \norm{\psi_{d_k}(z^\prime) - \psi_{d_k}(z)}_{A_k^{-1}} \sqrt{2 \log \paren{ {\abs{\widehat \cS_k}^2}/{\delta_k} }}$.
		\ENDFOR 
		\ENSURE Set of uneliminated arms $\widehat \cS_{n+1}$.
	\end{algorithmic}
\end{algorithm}

We now discuss 
the adaptive selection of hypothesis class, which is achieved through a new optimization problem: 
At round $k$, $d_k \in [D]$ is selected as the largest dimension such that the value of an experimental design is no larger than the fixed selection budget $B$, i.e.,
\begin{align}
    & \max d  \label{ms_bai:eq:opt_d_selection} \\
    & \text{ s.t. } d \in [D], \nonumber \\
    & \qquad \max \curly*{2^{2k} \cdot \inf_{\lambda \in \simp_{\cX}} \sup_{y \in \cY(\psi_d(\widehat \cS_k))} \norm{y}^2_{A_d(\lambda)^{-1}} , r_d(\zeta) }\leq B. \nonumber
\end{align}
The experimental design leverages the geometry of the \emph{uneliminated} set of arms. Intuitively, the algorithm is selecting the \emph{richest} hypothesis class that still allows the learner to improve its estimates of the gaps by a factor of 2 using (roughly) $B$ samples. 
When the budget for dimension selection $B$ is large enough, \gemsc operates on well-specified linear bandits (i.e., using $d_k \geq d_\star$) at all rounds, guaranteeing that the output set of arms are $(2^{1-n})$-optimal. The next lemma provides guarantees for \gemsc.

\begin{restatable}{lemma}{lmSubroutineFixedConfidence}
\label{ms_bai:lm:subroutine_fixed_confidence}
Suppose $B \geq \max \curly*{64 \rho_{d_\star}^\star, r_{d_\star}(\zeta)}$. With probability at least $1-\delta$, \gemsc outputs a set of arms $\widehat \cS_{n+1}$ such that $\Delta_z < 2^{1-n}$ for any $z \in \widehat \cS_{n+1}$.
\end{restatable}

\begin{algorithm}[]
	\caption{Adaptive Strategy for Model Selection (Fixed Confidence)}
	\label{ms_bai:alg:doubling_fixed_confidence} 
	\renewcommand{\algorithmicrequire}{\textbf{Input:}}
	\renewcommand{\algorithmicensure}{\textbf{Output:}}
	\newcommand{\algorithmicbreak}{\textbf{break}}
    \newcommand{\BREAK}{\STATE \algorithmicbreak}
	\begin{algorithmic}[1]
		\REQUIRE Confidence parameter $\delta$.
		\STATE Randomly select a $\widehat z_\star \in \cZ$ as the recommendation for the optimal arm.
		\FOR {$\ell = 1, 2, \dots$}
		\STATE Set $\gamma_\ell = 2^\ell$ and $\delta_\ell = \delta/(2\ell^3)$.
		    \FOR {$i = 1, 2, \dots, \ell$}
		    \STATE Set $n_i  =2^i$, $B_i = \gamma_{\ell}/n_i= 2^{\ell -i}$, and \\
		    get $\widehat \cS_i = \text{\gemsc}(n_i, B_i, \delta_\ell)$.
		    \IF{$\widehat \cS_i = \curly*{\widehat z}$ is a singleton set}
		    \STATE Update the recommendation $\widehat z_\star = \widehat z$. 
		    \BREAK  \, (the inner for loop over $i$)
		    \ENDIF
		    \ENDFOR
		\ENDFOR 
	\end{algorithmic}
\end{algorithm}

We present our main algorithm for model selection in \cref{ms_bai:alg:doubling_fixed_confidence}, which loops over an iterate $\ell$ with roughly geometrically increasing budget $\gamma_\ell = \ell 2^\ell$. Within each iteration $\ell$, \cref{ms_bai:alg:doubling_fixed_confidence} invokes \gemsc $\ell$ times with different configurations $(n_i, B_i)$: $n_i$ is viewed as a guess for the unknown quantity $\log_2(1/\Delta_{\min})$; and $B_i$ is viewed as a guess of $\rho^\star_{d_\star}$, which is then used to determine the adaptive selection hypothesis class. The configurations $\curly*{(n_i, B_i)}_{i=1}^\ell$ are chosen as the diagonal of a two dimensional gird over $n_i$ and $B_i$. 
Within each iteration $\ell$, the recommendation $\widehat z_\star$ is updated as the arm contained in the \emph{first} singleton set returned (if any). Since $B_i$ is chosen in a decreasing order, we are recommending the arm selected from the richest hypothesis class that terminates recommending a single arm. The singleton is guaranteed to contain the optimal arm once a rich enough hypothesis class is considered.
We provide the formal guarantees as follows.\looseness=-1

\begin{restatable}{theorem}{thmDoublingFixedConfidence}
\label{ms_bai:thm:doubling_fixed_confidence}
Let $\tau_\star = \log_2(4/\Delta_{\min}) \max \curly*{\rho^\star_{d_\star}, r_{d_\star}(\zeta)}$.
With probability at least $1-\delta$, \cref{ms_bai:alg:doubling_fixed_confidence} starts to output the optimal arm within iteration $\ell_\star = O( \log_2(\tau_\star))$, and takes at most $N = O \paren{ \tau_\star \log_2(\tau_\star) \log(\abs{\cZ} \log_2(\tau_\star)/\delta) }$ samples.
\end{restatable}

The sample complexity in \cref{ms_bai:thm:doubling_fixed_confidence} is analyzed in an unverifiable way:
\cref{ms_bai:alg:doubling_fixed_confidence} starts to output the optimal arm after $N$ samples, but it does not stop its sampling process.
Nevertheless, up to a rounding-related term and other logarithmic factors,\footnote{We refer readers to \citet{katz2020true} for detailed discussion on unverifiable sample complexity. The rounding term $r_{d_\star}(\zeta) = O(d_\star/\zeta^2)$ commonly appears in the linear bandit pure exploration literature \citep{fiez2019sequential, katz2020empirical}. Although we do not focus on optimizing logarithmic terms in this chapter, e.g., the $\log(\abs{\cZ})$ term, our techniques can be extended to address this by combining techniques developed in \citet{katz2020empirical}.} the unverifiable sample complexity matches the non-interactive lower bound developed in \cref{ms_bai:thm:lower_bound_non_interactive}.
The non-interactive lower bound serves as a fairly strong baseline since the non-interactive learner is allowed to sample \emph{with the knowledge of $\theta_\star$}.
Computationally, \cref{ms_bai:alg:doubling_fixed_confidence} starts to output the optimal arm after iteration $\ell_\star$, with at most $O(\ell_\star^2)$ subroutines (\cref{ms_bai:alg:subroutine_fixed_confidence}) invoked.
At each iteration $\ell \leq \ell_\star$, \cref{ms_bai:alg:subroutine_fixed_confidence} is invoked with configurations $n_i$, $B_i$ such that $n_i B_i = 2^\ell \leq 2^{\ell_\star}$ (note that $\ell_\star$ is of logarithmic order).
Up to a model selection step (i.e., selecting $d_k$),
the per-round computational complexity of \cref{ms_bai:alg:subroutine_fixed_confidence} is similar to the complexity of the standard linear bandit algorithm \rage.

\paragraph{Why not recommend arm verifiably} We provide a simple example to demonstrate that outputting the estimated best arm (using least squares) before examining full vectors in $\R^D$ can lead to incorrect answers, indicating that verifiable sample complexity, i.e., the number of samples required to terminate the game with a recommendation, scales with $D$ ($\rho^\star_D$). We consider a linear bandit problem with action set $\cX = \cZ = \curly*{e_i}_{i=1}^{D}$. We consider two cases: either (1) $\theta_\star  \ldef [1, 0, \dots, 0, 0]^\top \in \R^D$ with $z_\star = e_1$; or (2) $\theta_\star  \ldef [1, 0, \dots, 0, 2]^\top \in \R^D$ with $z_\star = e_D$. We assume \emph{deterministic} feedback in this example. Let $n_x \geq 1$ denote the number of pulls on arm $x \in \cX$. In both cases, for any $d < D$, the design matrix $\sum_{x \in \cX} n_x \psi_d(x) \psi_d(x)^\top$ is diagonal with entries $(n_{e_i})_{i=1}^{d}$, and the least squares estimator is $\widehat \theta_d = e_1 \in \R^d$. As a result, $e_1$ will be recommended as the best arm: the recommendation is correct in the first case but incorrect in the second case. Essentially, one cannot rule out the possibility that $d_\star$ is equal to $D$ without examining full vectors in $\R^D$. 
Verifiably identifying the best arm in $\R^D$ (with noisy feedback) takes $\widetilde \Omega(\rho^\star_D)$ samples \citep{fiez2019sequential}.

\section{Fixed Budget Setting}
\label{ms_bai:sec:fixed_budget}

We study the fixed budget setting with $\cZ \subseteq \cX$, which includes the linear bandit problem $\cZ = \cX$ as a special case. 
Similar to fixed confidence setting, we develop a main algorithm (\cref{ms_bai:alg:doubling_fixed_budget}) that invokes a base algorithm as subroutines (\gemsb, \cref{ms_bai:alg:subroutine_fixed_budget}). 
\cref{ms_bai:alg:doubling_fixed_budget} achieves an error probability $\widetilde O(\exp(-T/\rho^\star_{d_\star}))$, which, again, matches the strong baseline developed in \cref{ms_bai:thm:lower_bound_non_interactive}.

\begin{algorithm}[]
    \caption{\gemsb Gap Elimination with Model Selection (Fixed Budget)}
    \label{ms_bai:alg:subroutine_fixed_budget} 
	\renewcommand{\algorithmicrequire}{\textbf{Input:}}
	\renewcommand{\algorithmicensure}{\textbf{Output:}}
	\begin{algorithmic}[1]
	\REQUIRE Total budget $T$ (allowing non-integer input), number of rounds $n$, budget for dimension selection $B$.
	\STATE Set $T^\prime = \floor*{T/n}$, $\widehat \cS_1 = \cZ$. Set $\widetilde D$ as the largest dimension that ensures rounding with $T^\prime$ samples, i.e., $\widetilde D = \text{\OPT}(T^\prime, D, f(\cdot))$, where $f(d) = r_d(\zeta)$.
	\FOR {$k = 1,  \dots, n$}
	\STATE Define function $g_k(d) \ldef 2^{2k} \, \iota(\cY(\psi_d(\widehat \cS_k)))$. 
	\STATE Get $d_{k} = \text{\OPT}(B, \widetilde D, g_k(\cdot) )$, where where $d_k \leq \widetilde  D$ is largest dimension such that $g_k(d_k) \leq B$ (similar to the optimization problem in \cref{ms_bai:eq:opt_d_selection}). Set $\lambda_k$ be the optimal design of the optimization problem \\
	$\inf_{\lambda \in \simp_{\cX}} \sup_{z, z^\prime \in \widehat \cS_k} \norm*{\psi_{d_k}(z)-\psi_{d_k}(z^\prime)}^2_{A_{d_k}(\lambda)^{-1}}$.
	\STATE Get allocations \\
	$\{x_1, \ldots, x_{T^\prime} \} = \text{\round} (\lambda_{k},T^\prime,d_k, \zeta)$.
	\STATE Pull arms $\curly*{x_1, \ldots, x_{T^\prime}} $ and receive rewards $\curly*{r_1, \ldots, r_{T^\prime}}$.
	\STATE Set $\widehat{\theta}_k = A_k^{-1} b_k \in \R^{d_k}$, \\
	where $A_k = \sum_{i=1}^{N_k} \psi_{d_k}(x_i) \psi_{d_k}(x_i)^\top$,\\
	and $b_k =  \sum_{i=1}^{N_k} \psi_{d_k}(x_i) b_i$.
    \STATE Set $\widehat \cS_{k+1} = \widehat \cS_k \setminus \{z \in \widehat \cS_k : \exists z^\prime \text{ s.t. } \ang{\widehat{\theta}_k, \psi_{d_k}(z^\prime) - \psi_{d_k}(z) } \geq 2^{-k} \}$.
	\ENDFOR
	\ENSURE Any uneliminated arm $\widehat z_\star \in \widehat \cS_{n+1}$.
	\end{algorithmic}
\end{algorithm}

The subroutine \gemsb takes sample budget $T$, number of iterations $n$ and dimension selection budget $B$ as input, and outputs an (arbitrary) uneliminated arm after $n$ iterations. As in the fixed confidence setting, \gemsb performs adaptive selection of the hypothesis class through an optimization problem defined similar to the one in \cref{ms_bai:eq:opt_d_selection}. The main differences from the fixed confidence subroutine is as follows: the selection budget $B$ is only used for dimension selection, and the number of samples allocated per iteration is determined as $\floor*{T/n}$. \gemsb is guaranteed to output the optimal arm with probability $1 - \widetilde O (\exp(- T/\rho^\star_{d_\star}))$ when the selection budget $B$ is selected properly, as detailed in \cref{ms_bai:lm:subroutine_fixed_budget}.

\begin{restatable}{lemma}{lmSubroutineFixedBudget}
\label{ms_bai:lm:subroutine_fixed_budget}
Suppose $64 \rho_{d_\star}^\star \leq B \leq 128 \rho_{d_\star}^\star $ and $T/n \geq r_{d_\star}(\zeta) + 1$. \cref{ms_bai:alg:subroutine_fixed_budget} outputs an arm $\widehat z_\star$ such that $\Delta_{\widehat z_\star} < 2^{1-n}$ with probability at least
\begin{align*}
    1- n \abs*{ \cZ }^2 \exp \paren*{ - { T}/{ 640 \, n \, \rho_{d_\star}^\star } }.
\end{align*}
\end{restatable}

\begin{algorithm}[]
	\caption{Adaptive Strategy for Model Selection (Fixed Budget)}
	\label{ms_bai:alg:doubling_fixed_budget} 
	\renewcommand{\algorithmicrequire}{\textbf{Input:}}
	\renewcommand{\algorithmicensure}{\textbf{Output:}}
	\begin{algorithmic}[1]
		\REQUIRE Total budget $2T$.
		\STATE \textbf{Step 1: Selection.} Initialize an empty selection set $\cA = \emptyset$.
		\STATE Set $p = \floor*{W(T)}$ and $T^\prime = {T/p}$.
		\FOR {$i = 1, \dots, p$}
		\STATE Set $B_i = 2^i$, $q_i = \floor*{W(T^\prime/B_i)}$ and $T^{\prime \prime} = {T^\prime/q_i}$.
		\FOR {$j = 1,\dots, q_i$}
		\STATE Set $n_j = 2^j$. \\
		Get $\widehat z_\star^{ij} = \text{\gemsb}(T^{\prime \prime}, n_j, B_i)$ and insert $\widehat z_\star^{ij}$ into the pre-selection set $\cA$.
		\ENDFOR
		\ENDFOR
		\STATE \textbf{Step 2: Validation.} Pull each arm in the pre-selection set $\cA$ exactly $\floor*{T/\abs*{\cA}}$ times.
		\ENSURE Output arm $\widehat z_\star$ with the highest empirical reward from the validation step.
	\end{algorithmic}
\end{algorithm}

Our main algorithm for the fixed budget setting is introduced in \cref{ms_bai:alg:doubling_fixed_budget}. \cref{ms_bai:alg:doubling_fixed_budget} consists of two phases: a pre-selection phase and a validation phase. The pre-selection phase collects a set of potentially optimal arms, selected by subroutines, and the validation phase examines the optimality of the collected arms. We provide \cref{ms_bai:alg:doubling_fixed_budget} with $2T$ total sample budget, and split the budget equally for each phase. At least one good subroutine is guaranteed to be invoked in the pre-selection phase (for sufficiently large $T$). The validation step focuses on identifying the best arm among the pre-selected $O((\log_2 T)^2)$ candidates (as explained in the next paragraph). Our selection-validation trick can be viewed as a \emph{dimension-reduction} technique: we convert a linear bandit problem in $\R^D$ (with unknown $d_\star$) to another linear bandit problem in $\R^{O((\log_2 T)^2)}$,\footnote{Technically, we treat the problem as a standard multi-armed bandit problem with $O((\log_2 T)^2)$ arms, which is a special case of a linear bandit problem in $\R^{O((\log_2 T)^2)}$.} i.e., a problem whose dimension is only polylogarithmic in the budget $T$. 

For non-negative variable $p$, we use $p = W(T)$ to represent the solution of equation $T = p \cdot 2^p$. One can see that $W(T) \leq \log_2 T$. As a result, at most $(\log_2 T)^2$ subroutines are invoked with different configurations of $\curly*{(T^{\prime \prime}, n_j, B_i)}$. The use of $W(\cdot)$ is to make sure that $T^{\prime \prime} \geq n_j B_i $ for all subroutines invoked. This provides more efficient use of budget since the error probability upper bound guaranteed by \gemsb scales as $\widetilde O (\exp(- T^{\prime \prime} / n_j B_i ))$.

\begin{restatable}{theorem}{thmDoublingFixedBudget}
\label{ms_bai:thm:doubling_fixed_budget}
Suppose $\cZ \subseteq \cX$. If $T = \widetilde \Omega \paren{ \log_2(1/\Delta_{\min}) \max \curly*{\rho_{d_\star}^
    \star, r_{d_\star}(\zeta)} }$, then \cref{ms_bai:alg:doubling_fixed_budget} outputs the optimal arm with error probability at most
    \begin{align*}
    &\log_2 (4/\Delta_{\min}) \abs*{ \cZ }^2 \exp \paren*{ - \frac{ T}{ 1024 \, \log_2 (4/\Delta_{\min}) \, \rho_{d_\star}^\star } } \\
    & \quad + 2 (\log_2 T)^2 \exp \paren*{ - \frac{T}{8 (\log_2 T)^2/ \Delta_{\min}^2} }.
\end{align*}
Furthermore, if there exist universal constants such that $\max_{x \in \cX} \norm{\psi_{d_\star}(x)}^2 \leq c_1$ and $\min_{z \in \cZ} \norm{\psi_{d_\star}(z_\star) - \psi_{d_\star}(z)}^2 \geq c_2$, the error probability is upper bounded by
    \begin{align*}
    O \Bigg(& \max \curly*{ \log_2(1/\Delta_{\min}) \abs*{\cZ}^2,  (\log_2 T)^2 } \\
    & \times  \exp \paren*{ - \frac{c_2 T}{\max \curly*{ \log_2(1/\Delta_{\min}), (\log_2 T)^2 } c_1 \rho_{d_\star}^\star} }  \Bigg).
\end{align*}
\end{restatable}

Under the mild assumption discussed above, the error probability of \cref{ms_bai:alg:doubling_fixed_budget} scales as $\widetilde O(\exp(-T/\rho^\star_{d_\star}))$. Such an error probability not only matches, up to logarithmic factors, the strong baseline developed in \cref{ms_bai:thm:lower_bound_non_interactive}, but also matches the error bound in the non-model-selection setting (with known $d_\star$) \citep{katz2020empirical} (Algorithm 3 therein, which is also analyzed under a mild assumption).  
Computationally, \cref{ms_bai:alg:doubling_fixed_budget} invokes \cref{ms_bai:alg:subroutine_fixed_budget} at most $(\log_2 T)^2$ times, each with budget $T^{\prime \prime} \leq T$ and $n_j, B_i$ such that $n_j B_i \leq T$. 
The per-round computational complexity of \cref{ms_bai:alg:subroutine_fixed_confidence} is similar to the one of \cref{ms_bai:alg:subroutine_fixed_budget} (with similar configurations).

Compared to the fixed confidence setting, the fixed budget setting in linear bandits is relatively less studied \citep{hoffman2014correlation, katz2020empirical, alieva2021robust, yang2021towards}. 
To our knowledge, even without the added challenge of model selection, near \emph{instance optimal} error probability guarantee is only achieved by Algorithm 3 in
\citet{katz2020empirical}.
Our \cref{ms_bai:alg:doubling_fixed_budget} provides an alternative way to tackle the fixed budget setting, through a novel selection-validation procedure.
Our techniques might be of independent interest.

\section{Model Selection with Misspecification}
\label{ms_bai:sec:misspecification}

We generalize the model selection problem into the \emph{misspecified} regime in this section. Our goal here is to identify an $\epsilon$-optimal arm due to misspecification. We aim to provide sample complexity/error probability guarantees with respect to a hypothesis class that is rich enough to allow us to identify an $\epsilon$-optimal arm.
Pure exploration with model misspecification are recently studied in the literature \citep{alieva2021robust, camilleri2021high, zhu2021pure}.
The model selection criterion we consider here further complicates the problem setting and are not covered in previous work.

We consider the case where the expected reward $h(x)$ of any arm $x \in \cX \cup \cZ \subseteq \R^D$ cannot be perfectly represented as a linear model in terms of its feature representation $x$. We use function $\widetilde \gamma (d)$ to capture the misspecification level with respect to truncation the level $d \in [D]$, i.e.,
\begin{align}
   \widetilde \gamma(d) \ldef \min_{\theta \in \R^D} \max_{x \in \cX \cup \cZ} \abs*{h(x) - \ang*{\psi_d(\theta), \psi_d(x)}}. \label{ms_bai:eq:mis_level}
\end{align}
We use $\theta^d_\star \in \argmin_{\theta \in \R^D} \max_{x \in \cX \cup \cZ} \abs*{h(x) - \ang*{\psi_d(\theta), \psi_d(x)}}$ to denote (any) reward parameter that best captures the worst case deviation in $\R^d$, and use $\eta_d(x) \ldef h(x) - \ang*{\psi_d(\theta_\star^d), \psi_d(x)}$ to represent the corresponding misspecification with respect to arm $x \in \cX \cup \cZ$. We have $\max_{x \in \cX \cup \cZ} \abs*{\eta_d(x)} \leq \widetilde \gamma(d)$ by definition. Although the value of $\eta_d(x)$ depends on the selection of the possibly non-unique $\theta^d_\star$, only the worst-case deviation $\widetilde \gamma(d)$ is used in our analysis. Our results in this section are mainly developed in cases when $\cZ \subseteq \cX$, which contains the linear bandit problem $\cZ = \cX$ as a special case. 

\begin{restatable}{proposition}{propNonIncreasingMisspecification}
\label{ms_bai:prop:non_increasing_misspecification}
The misspecification level $\widetilde \gamma(d)$ is non-increasing with respect to $d$.
\end{restatable}

The non-increasing property of $\widetilde \gamma (d)$ reflect the fact that the representation power of the linear component is getting better in higher dimensions. Following \citet{zhu2021pure}, we use $\gamma(d)$ to quantify the sub-optimality gap of the identified arm, i.e.,
\begin{align*}
    \gamma(d) \ldef  \min \Big\{ 2 \cdot 2^{-n}: n \in \N, \forall k \leq n, 
     \paren{2 + \sqrt{(1+\zeta) \iota \paren*{ \cY( \psi_d(\cS_k)) }} } \widetilde \gamma(d) \leq 2^{-k}/2 \Big\}.
\end{align*}

It can be shown that, for any fixed $d \in [D]$, at least a $O(\sqrt{d} \, \widetilde \gamma (d))$-optimal arm can be identified in the existence of misspecification. Such inflation from $\widetilde \gamma(d)$ to $\sqrt{d}\, \widetilde \gamma(d)$ is unavoidable in general: \citet{lattimore2020learning} constructs a hard instance such that identifying a $o(\sqrt{d}\widetilde \gamma(d))$-optimal arm requires sample complexity exponential in $d$, even with \emph{deterministic} feedback. On the other hand, identifying a $\Omega(\sqrt{d} \, \widetilde \gamma(d))$-optimal arm only requires sample complexity polynomial in $d$. Such a sharp tradeoff between sample complexity and achievable optimality motivates our definition of $\gamma(d)$.

We assume $\gamma(d)$ can be made arbitrarily small for $d\in[D]$ large enough, which includes instances with no misspecification in $\R^D$ as special cases.\footnote{We make this assumption in order to identify an $\epsilon$-optimal arm for any pre-defined $\epsilon > 0$. Otherwise, one can adjust the goal and identify arms with appropriate sub-optimality gaps.} For any $\epsilon > 0$, we define 
$d_\star(\epsilon) \ldef \min \curly*{d \in [D]: \forall d^\prime \geq d, \gamma(d^\prime) \leq \epsilon }$.
We aim at identifying an $\epsilon$-optimal arm with sample complexity related to $\rho_{d_\star(\epsilon)}^\star$, which is defined as an $\epsilon$-relaxed version of complexity measure $\rho_{d_\star}^\star$, i.e.,
\begin{align*}
    \rho^\star_{d}(\epsilon) \ldef \inf_{\lambda \in \simp_{\cX}} \sup_{z \in \cZ \setminus \curly*{z_\star }} \frac{\norm*{\psi_d(z_{\star})-\psi_d(z)}^2_{A_{d}(\lambda)^{-1}}}{(\max \curly*{ h(z_\star) - h(z), \epsilon })^2}.
\end{align*}
We consider a closely related complexity measure $\widetilde \rho_{d}^\star(\epsilon)$, which is defined with respect to linear component $\widetilde h(x) \ldef \ang*{\psi_d(\theta_\star^d), \psi_d(x)}$, i.e.,
\begin{align*}
     \widetilde \rho^\star_{d}(\epsilon) \ldef 
    \inf_{\lambda \in \simp_{\cX}} \sup_{z \in \cZ \setminus \curly*{z_\star }} \frac{\norm*{\psi_d(z_{\star})-\psi_d(z)}^2_{A_{d}(\lambda)^{-1}}}{(\max \curly*{\ang*{ \psi_{d}(\theta_\star^d), \psi_{d}(z_\star) - \psi_d(z) }, \epsilon })^2}.
\end{align*}
\begin{restatable}[\citet{zhu2021pure}]{proposition}{propRhoRelation}
\label{ms_bai:prop:rho_relation}
We have $\rho_d^\star (\epsilon) \leq 9 \widetilde \rho_d^\star (\epsilon)$ for any $\epsilon \geq \widetilde \gamma(d)$. Furthermore, if $\widetilde \gamma(d) < \Delta_{\min}/ 2$, $\widetilde \rho_d^\star(0)$ represents the complexity measure for best arm identification with respect to a linear bandit instance with action set $\cX$, target set $\cZ$ and reward function $\widetilde h(x) \ldef \ang*{\psi_d(\theta_\star^d), \psi_d(x)}$.
\end{restatable}

Assuming $\widetilde \gamma(d_\star(\epsilon)) < \min \curly{\epsilon, \Delta_{\min}/2}$, \cref{ms_bai:prop:rho_relation} shows that $\rho_{d_\star(\epsilon)}^\star(\epsilon)$ is at most a constant factor larger than $\widetilde \rho_{d_\star (\epsilon)}^\star(\epsilon)$, which is the $\epsilon$-relaxed complexity measure of a closely related linear bandit problem (without misspecification) in $\R^{d_\star(\epsilon)}$.

\paragraph{Fixed confidence setting} A modified algorithm (and its subroutine, both deferred to \cref{ms_bai:app:alg_misspecification}) is used for the fixed confidence setting with model misspecification. Sample complexity of the modified algorithm is provided as follows.

\begin{restatable}{theorem}{thmDoublingFixedConfidenceMisGen}
\label{ms_bai:thm:doubling_fixed_confidence_mis_gen}
With probability at least $1-\delta$, \cref{ms_bai:alg:doubling_fixed_confidence_mis_gen} starts to output $2 \epsilon$-optimal arms after $N = \widetilde O \paren{ \log_2(1/\epsilon) \max \curly{\rho^\star_{d_\star(\epsilon)} (\epsilon), r_{d_\star(\epsilon)}(\zeta)} + 1/\epsilon^2}$ samples, where we hide logarithmic terms besides $\log_2(1/\epsilon)$ in the $\widetilde O$ notation.
\end{restatable}

\begin{remark}
\label{ms_bai:rm:mis_BAI}
The extra $1/\epsilon^2$ term comes from a validation step in the modified algorithm. If the goal is to identify the optimal arm, then this term can be removed with a slight modification of the algorithm. See \cref{ms_bai:app:BAI_misspecification} for detailed discussion.
\end{remark}

\paragraph{Fixed budget setting} Our algorithms for the fixed budget setting are \emph{robust} to model misspecification, and we provide the following guarantees.

\begin{restatable}{theorem}{thmDoublingFixedBudgetMis}
\label{ms_bai:thm:doubling_fixed_budget_mis}
Suppose $\cZ \subseteq \cX$. If $T = \widetilde \Omega \paren*{ \log_2(1/\epsilon) \max \curly*{\rho_{d_\star (\epsilon)}^
    \star(\epsilon), r_{d_\star (\epsilon)}(\zeta)} }$, then \cref{ms_bai:alg:doubling_fixed_budget} 
    outputs an $2\epsilon$-optimal arm with error probability at most
\begin{align*}
    & \log_2 (4/\epsilon)  \abs*{ \cZ }^2  \exp \paren*{ - \frac{ T}{ 4096 \, \log_2 (4/\epsilon)  \, \rho_{d_\star(\epsilon)}^\star (\epsilon) } } \\
    & \quad + 2 (\log_2 T)^2 \exp \paren*{ - \frac{T}{8 (\log_2 T)^2/ \epsilon^2} }.
\end{align*}
Furthermore, if there exist universal constants such that $\max_{x \in \cX} \norm{\psi_{d_\star(\epsilon)}(x)}^2 \leq c_1$ and $\min_{z \in \cZ} \norm{\psi_{d_\star(\epsilon)}(z_\star) - \psi_{d_\star(\epsilon)}(z)}^2 \geq c_2$, the error probability is upper bounded by
\begin{align*}
    O \Bigg(& \max \curly*{ \log_2(1/\epsilon) \abs*{\cZ}^2, (\log_2 T)^2 }\\
    &\times \exp \paren*{ - \frac{c_2 T}{\max \curly*{ \log_2(1/\epsilon), (\log_2 T)^2 } c_1 \rho_{d_\star(\epsilon)}^\star (\epsilon)} } \Bigg).
\end{align*}
\end{restatable}

\section{Experiments}
\label{ms_bai:sec:experiment}

We empirically compare our \cref{ms_bai:alg:doubling_fixed_confidence} with \rage \citep{fiez2019sequential}, which shares a similar elimination structure to our subroutine (i.e., \cref{ms_bai:alg:subroutine_fixed_confidence}) yet fails to conduct model selection in pure exploration. 
To our knowledge, besides algorithms developed in this chapter, there is no other algorithm that can adapt to the model selection setup for pure exploration linear bandits.\footnote{We defer additional experiment details/results to \cref{ms_bai:app:experiment}.
The purpose of this section is to empirically demonstrate the importance of conducting model selection in pure exploration linear bandits, even on simple problem instances.
We leave large-scale empirical evaluations for future work.}

\paragraph{Problem instances}
We conduct experiments with respect to the problem instance used to construct \cref{ms_bai:prop:rho_star_different_d}, which we detail as follows. 

We consider a problem instance with $\cX =\cZ = \curly*{x_i}_{i=1}^{d_\star + 1} \subseteq \R^{d_\star + 1}$ such that $x_i = e_i, \text{ for } i = 1, 2,\dots, d_\star$ and $x_{d_\star+1} = (1-\epsilon) \cdot e_{d_\star} + e_{d_\star+1}$,
where $e_i$ is the $i$-th canonical basis in $\R^{d_\star + 1}$.
The expected reward of each arm is set as $h(x_i) = \ang*{ e_{d_\star},x_i}$, i.e., $\theta_\star = e_{d_\star}$.
One can see that $d_\star$ is the intrinsic dimension and $D = d_\star +1$ is the ambient dimension.
We also notice that $x_\star = x_{d_\star}$ is the best arm with reward $1$, $x_{d_\star+1}$ is the second best arm with reward $1-\epsilon$ and all other arms have reward $0$. The smallest sub-optimality gap is $\epsilon$.
We choose $d_\star = 9$, $D = 10$, and vary $\epsilon$ to control the instance-dependent complexity. By setting $\epsilon$ to be a small value, we create a problem instance such that $\rho^\star_D \gg \rho^\star_{d_\star}$: we have $\rho^\star_{d_\star} = O(d_\star)$ yet $\rho^\star_D = \Omega(1/\epsilon^2)$ (see \cref{ms_bai:app:rho_star_different_d} for proofs).

\begin{figure}[H]
    \centering
    \includegraphics[width=.8\textwidth]{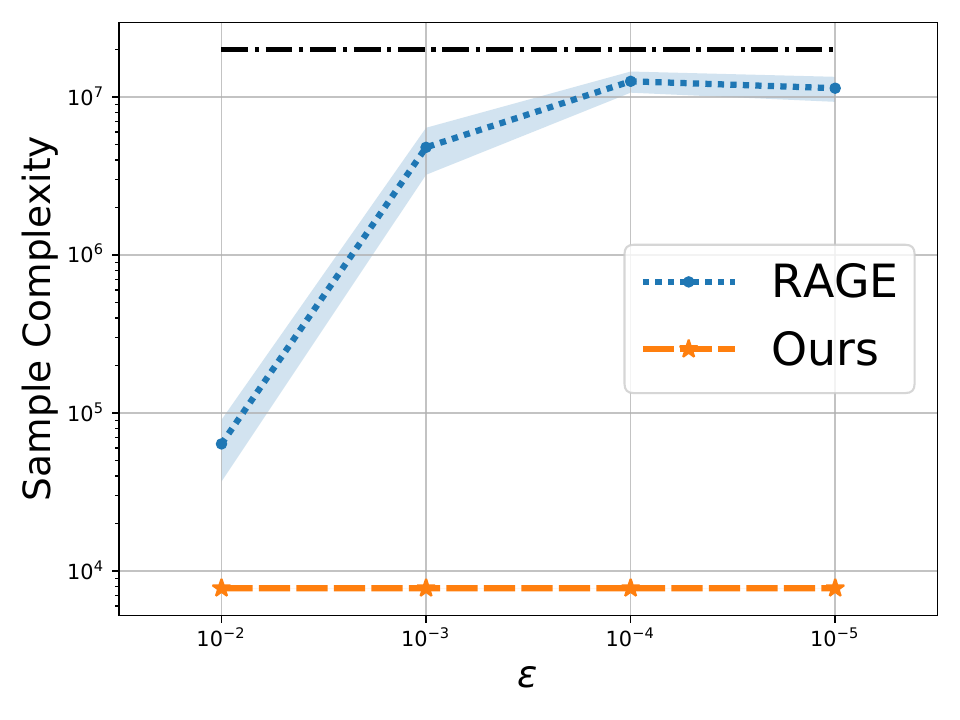}
    \caption{Comparison of sample complexity with varying sub-optimality gap.}
    \label{ms_bai:fig:comparison_prop}
\end{figure}

\paragraph{Empirical evaluations}
We evaluate the performance of each algorithm in terms of success rate, sample complexity and runtime.
We conduct $100$ independent trials for each algorithm. 
Both algorithms are force-stopped after reaching $10$ million samples (denoted as the black line in \cref{ms_bai:fig:comparison_prop}).
We consider an trial as failure if the algorithm fails to identify the best arm within $20$ million samples.
For each algorithm, we calculate the (unverifiable) sample complexity $\tau$ as the smallest integer such that the algorithm (1) empirically identifies the best arm; \emph{and} (2) the algorithm won't change its recommendation for any later rounds $t > \tau$ (up to $20$ million samples).
The (empirical) runtime of the algorithm is calculated as the total time consumed up to round $\tau$.
We average sample complexities and runtimes with respect to succeeded trials.

\begin{table}[H]
  \caption{Comparison of success rate with varying sub-optimality gap.}
  \label{ms_bai:tab:success_rate_1}
  \centering
  \begin{tabular}{lcccc}
    \toprule
          $\epsilon $ & $10^{-2}$   & $10^{-3}$ & $10^{-4}$ & $10^{-5}$\\
    \midrule
    \rage    & $100\%$ & $98\%$ & $56\%$ & $62\%$ \\
    Ours    & $100\%$ & $100\%$ & $100\%$ & $100\%$  \\
    \bottomrule
  \end{tabular}
\end{table}

The success rates of \rage and our algorithm are shown in \cref{ms_bai:tab:success_rate_1}. 
The success rate of \rage drops dramatically as $\epsilon$ (the smallest sub-optimality gap) gets smaller. On the other hand, however, our algorithm is not affected by the change of $\epsilon$ since it automatically adapts to the intrinsic dimension $d_\star$: One can immediately see that $h(x_{d_\star}) \geq h(x_{d_\star+1})$ when working in $\R^{d_\star}$. 
Due to the same reason, our algorithm significantly outperforms \rage in sample complexity as well (see \cref{ms_bai:fig:comparison_prop}): Our algorithm adapts to the true sample complexity $\rho^\star_{d_\star}$ yet \rage suffers from complexity $\rho^\star_D \gg \rho^\star_{d_\star}$, especially when $\epsilon$ is small.

The runtime of both algorithms are shown in \cref{ms_bai:tab:runtime}.
Our algorithm is affected by the computational overhead of conducting model selection (e.g., the two dimensional doubling trick). Thus, \rage shows advantages in runtime when $\epsilon$ is relatively large.
However, our algorithm runs faster than \rage when $\epsilon$ gets smaller.
This observation further shows that the implementation overhead can be small in comparison with the sample complexity gains achieved from model selection.

\begin{table}[H]
  \caption{Comparison of runtime with varying sub-optimality gap.}
  \label{ms_bai:tab:runtime}
  \centering
  \begin{tabular}{lcccc}
    \toprule
          $\epsilon $ & $10^{-2}$   & $10^{-3}$ & $10^{-4}$ & $10^{-5}$\\
    \midrule
    \rage    & $3.46\,$s & $7.87\,$s & $17.33\,$s & $16.81\,$s \\
    Ours    & $12.12\,$s & $11.17\,$s & $12.44\,$s & $12.41\,$s  \\
    \bottomrule
  \end{tabular}
\end{table}

It is worth mentioning that simple variations of the problem instance studied in this section have long been considered as hard instances to examine linear bandit pure exploration algorithms \citep{soare2014best, xu2018fully, tao2018best, fiez2019sequential, degenne2020gamification}. Our results show that, both theoretically and empirically, the problem instance becomes quite easy when viewed from the model selection perspective.

\section{Discussion}
\label{ms_bai:sec:discussion}

We initiate the study of model selection in pure exploration linear bandits, in both fixed confidence and fixed budget settings, and design algorithms with near instance optimal guarantees.
Along the way, we develop a novel selection-validation procedure to deal with the understudied fixed budget setting in linear bandits (even without the added challenge of model selection).
We also adapt our algorithms to problems with model misspecification.

We conclude this chapter with some directions for future work.
An immediate next step is to conduct large-scale evaluations for model selection in pure exploration linear bandits. 
One may need to develop practical version of our algorithms to bypass the computational overheads of conducting model selection.
Another interesting direction is provide guarantees to general transductive linear bandits, i.e., not restricted to cases $\cZ \subseteq \cX$, in fixed budget setting/misspecified regime. We believe one can use a selection-validation procedure similar to the one developed in \cref{ms_bai:alg:subroutine_fixed_budget}, but with the current validation step replaced by another linear bandit pure exploration algorithm. 
Note that the number of arms to be validated is of logarithmic order.

\section{Proofs and Supporting Results}
\label{ms_bai:sec:proofs}

\subsection{Supporting Results}
\label{ms_bai:app:supporting}

\subsubsection{Matrix Inversion and Rounding in Optimal Design}
\label{ms_bai:app:inverse}

Our treatments are similar to the ones discussed in \cite{zhu2021pure}. We provide the details here for completeness.

\textbf{Matrix Inversion.}  The notation $\norm{y}^2_{A_d(\lambda)^{-1}}$ is clear when $A_d(\lambda)$ is invertible. For possibly singular $A_d(\lambda)$, pseudo-inverse is used if $y$ belongs to the range of $A_{d(\lambda)}$; otherwise, we set $\norm{y}_{A_{d}(\lambda)^{-1}}^2 = \infty$. With this (slightly abused) definition of matrix inversion, we discuss how to do rounding next.

\textbf{Rounding in Optimal Design.} For any $\cS \subseteq \cZ$, the following optimal design 
\begin{align*}
    \inf_{\lambda \in \simp_{\cX}} \sup_{y \in \cY(\psi_d(\cS))} \norm{y}^2_{A_{d}(\lambda)^{-1}}
\end{align*}
will select a design $\lambda^\star \in \simp_{\cX}$ such that every $y \in \cY(\psi_d(\cS))$ lies in the range of $A_{d}(\lambda^\star)$.\footnote{If the infimum is not attained, we can simply take a design $\lambda^{\star \star}$ with associated value $\tau^{\star \star} \leq (1+\zeta_0) \inf_{\lambda \in \simp_{\cX}} \sup_{\by \in \cY(\bpsi_d(\cS))} \norm{\by}^2_{\bA_{\bpsi_d}(\lambda)^{-1}}$ for a $\zeta_0 > 0$ arbitrarily small. This modification is used in our algorithms as well, and our results (bounds on sample complexity and error probability) goes through with changes only in constant terms. } If $\spn(\cY(\psi_d(\cS))) = \R^d$, then $A_{_d}(\lambda^\star)$ is positive definite (recall that $A_{d}(\lambda^\star) = \sum_{x \in \cX} \lambda_{x} \psi_d(x) \psi_d(x)^\top$ and $\spn(\psi_d(\cX))= \R^d$ comes from the assumption that $\spn(\psi(\cX))= \R^D$). Thus the rounding guarantees in \cite{allen2020near} goes through (Theorem 2.1 therein, which requires a positive definite design; with additional simple modifications dealt as in Appendix B of \cite{fiez2019sequential}).

We now consider the case when $A_{d}(\lambda^\star)$ is singular. Since $\spn(\psi_d(\cX)) = \R^d$, we can always find another $\lambda^\prime$ such that $A_{d}(\lambda^\prime)$ is invertible. For any $\zeta_1 >0$, let $\widetilde \lambda^\star = (1-\zeta_1) \lambda^\star + \zeta_1 \lambda^\prime $. We know that $\widetilde \lambda^\star$ leads to a positive definite design. With respect to $\zeta_1$, we can find another $\zeta_2 > 0$ small enough (e.g., smaller than the smallest eigenvalue of $\zeta_1 A_{d}(\lambda^\prime)$) such that $A_{d}(\widetilde \lambda^\star) \succeq A_{d}((1-\zeta_1) \lambda^\star) + \zeta_2 I$. Since $A_{d}((1-\zeta_1) \lambda^\star) + \zeta_2 I$ is positive definite, for any $y \in \cY(\psi_d(\cS))$, we have 
\begin{align*}
    \norm{y}^2_{A_{d}(\widetilde \lambda^\star)^{-1}} \leq \norm{y}^2_{(A_{d}((1-\zeta_1) \lambda^\star) + \zeta_2 I)^{-1}}.
\end{align*}
Fix any $y \in \cY(\psi_d(\cS))$. Since $y$ lies in the range of $A_{d}(\lambda^\star)$ (by definition of the objective and matrix inversion), we clearly have 
\begin{align*}
    \norm{y}^2_{(A_{d}((1-\zeta_1) \lambda^\star) + \zeta_2 I)^{-1}} 
    \leq \norm{y}^2_{(A_{d}((1-\zeta_1) \lambda^\star))^{-1}}
    \leq \frac{1}{1-\zeta_1} \norm{y}^2_{A_{d}(\lambda^\star)^{-1}}.
\end{align*}
To summarize, we have 
\begin{align*}
    \norm{y}^2_{A_{d}(\widetilde \lambda^\star)^{-1}} \leq \frac{1}{1-\zeta_1} \norm{y}^2_{A_{d}(\lambda^\star)^{-1}},
\end{align*}
where $\zeta_1$ can be chosen arbitrarily small. We can thus send the positive definite design $\widetilde \lambda^\star$ to the rounding procedure in \cite{allen2020near}. We can incorporate the additional $1/(1-\zeta_1)$ overhead, for $\zeta_1 >0$ chosen sufficiently small, into the sample complexity requirement $r_d(\zeta)$ of the rounding procedure.

\subsubsection{Supporting Theorems and Lemmas}
\label{ms_bai:app:supporting_thm_lm}

\begin{lemma}[\citep{kaufmann2016complexity}]
\label{ms_bai:lm:change_of_measure}
Fixed any pure exploration algorithm $\pi$. Let $\nu$ and $\nu^\prime$ be two bandit instances with $K$ arms such that the distribution $\nu_i$ and $\nu_i^\prime$ are mutually absolutely continuous for all $i \in [K]$. For any almost-surely finite stopping time $\tau$ with respect to the filtration $\curly{\cF_t}_{t \geq 0}$, let $N_i(\tau)$ be the number of pulls on arm $i$ at time $\tau$. We then have 
\begin{align*}
    \sum_{i=1}^K \E_{\nu} [N_i(\tau)] \kl \paren{\nu_i, \nu_i^\prime} \geq \sup_{\cE \in \cF_{\tau}} d \paren{ \P_\nu (\cE), \P_{\nu^\prime} (\cE)},
\end{align*}
where $d(x, y) = x \log(x/y) + (1-x) \log((1-x)/(1-y))$ for $x, y \in [0,1]$ and with the convention that $d(0,0) = d(1,1) = 0$.
\end{lemma}

The following two lemmas largely follow the analysis in \cite{fiez2019sequential}.
\begin{lemma}
\label{ms_bai:lm:rho_stratified_eps}
Let $\cS_k = \curly*{z \in \cZ: \Delta_z < 4 \cdot 2^{-k}}$. We then have 
\begin{align}
    \sup_{k \in [\floor{\log_2 \paren{4/\epsilon}}]} \curly*{  2^{2k}  \iota \paren{ \cY(\psi_d(\cS_k)) } }  \leq 64 \rho_d^\star(\epsilon), \label{ms_bai:eq:psi_stratified_eps_0}
\end{align}
and
\begin{align}
    \sup_{k \in [\floor{\log_2 \paren{4/\epsilon}}]} \curly*{ \max \curly*{ 2^{2k}  \iota \paren{ \cY(\psi_d(\cS_k)) }, r_{d}(\zeta) } } \leq \max \curly{ 64 \rho_d^\star(\epsilon), r_d(\zeta) }, \label{ms_bai:eq:psi_stratified_eps_1}
\end{align}
where $\zeta$ is the rounding parameter.
\end{lemma}
\begin{proof}
For $y = \psi_d(z_\star) - \psi_d(z)$, we define $\Delta_y = \Delta_z = h(z_\star) - h(z)$. We have that 
\begin{align}
    \rho_d^\star (\epsilon) & = \inf_{\lambda \in \simp_{\cX}} \sup_{ y \in \cY^\star(\psi_d(\cZ))} \frac{\norm{y}^2_{A_d(\lambda)^{-1}}}{\max \curly*{\Delta_y, \epsilon}^2} \nonumber\\
    & = \inf_{\lambda \in \simp_{\cX}} \sup_{k \in [\floor{\log_2 \paren{4/\epsilon}}]} \sup_{ y \in \cY^\star(\psi_d(\cS_k))} \frac{\norm{y}^2_{A_d(\lambda)^{-1}}}{\max \curly*{\Delta_y, \epsilon}^2} \nonumber\\
    & \geq \sup_{k \in [\floor{\log_2 \paren{4/\epsilon}}]} \inf_{\lambda \in \simp_{\cX}}  \sup_{ y \in \cY^\star(\psi_d(\cS_k))} \frac{\norm{y}^2_{A_d(\lambda)^{-1}}}{\max \curly*{\Delta_y, \epsilon}^2} \nonumber\\
    & >  \sup_{k \in [\floor{\log_2 \paren{4/\epsilon}}]} \inf_{\lambda \in \simp_{\cX}} \sup_{ y \in \cY^\star(\psi_d(\cS_k))} \frac{\norm{y}^2_{A_d(\lambda)^{-1}}}{\paren{4 \cdot 2^{-k}}^2} \label{ms_bai:eq:psi_stratified_eps_2}\\
    & \geq \sup_{k \in [\floor{\log_2 \paren{4/\epsilon}}]} \inf_{\lambda \in \simp_{\cX}} \sup_{ y \in \cY(\psi_d(\cS_k))} \frac{\norm{y}^2_{A_d(\lambda)^{-1}} / 4}{\paren{4 \cdot 2^{-k}}^2} \label{ms_bai:eq:psi_stratified_eps_3}\\
    & \geq  \sup_{k \in [\floor{\log_2 \paren{4/\epsilon}}]} 2^{2k} \iota (\cY(\psi_d(\cS_k))) / 64, \nonumber
\end{align}
where \cref{ms_bai:eq:psi_stratified_eps_2} comes from the fact that $4 \cdot 2^{-k} \geq \epsilon$ when $k \leq \floor*{\log_2(4/\epsilon)}$; \cref{ms_bai:eq:psi_stratified_eps_3} comes from the fact that $\psi_d(z) - \psi_d(z^\prime) = (\psi_d(z) - \psi_d(z_\star)) + (\psi_d(z_\star) - \psi_d(z^\prime))$. This implies that, for any $k \in [\floor{\log_2 \paren{4/\epsilon}}]$,
\begin{align*}
    \max \curly{2^{2k} \rho(\cY(\psi_d(\cS_k))), r_d(\zeta)} \leq \max \curly{64 \rho_d^\star(\epsilon), r_d(\zeta)}.
\end{align*}
And the desired \cref{ms_bai:eq:psi_stratified_eps_1} immediately follows.
\end{proof}

\begin{lemma}
\label{ms_bai:lm:rho_stratified}
Let $\cS_k = \curly*{z \in \cZ: \Delta_z < 4 \cdot 2^{-k}}$. We then have 
\begin{align}
    \sup_{k \in [\ceil{\log_2 \paren{4/\Delta_{\min}}}]} \curly*{  2^{2k}  \iota \paren{ \cY(\psi_d(\cS_k)) } }  \leq 64 \rho_d^\star, \label{ms_bai:eq:psi_stratified_0}
\end{align}
and
\begin{align}
    \sup_{k \in [\ceil{\log_2 \paren{4/\Delta_{\min}}}]} \curly*{ \max \curly*{ 2^{2k}  \iota \paren{ \cY(\psi_d(\cS_k)) }, r_{d}(\zeta) } } \leq \max \curly{ 64 \rho_d^\star, r_d(\zeta) }, \label{ms_bai:eq:psi_stratified_1}
\end{align}
where $\zeta$ is the rounding parameter.
\end{lemma}
\begin{proof}
Take $\epsilon = \Delta_{\min}$ in \cref{ms_bai:lm:rho_stratified_eps}.
\end{proof}

The following lemma largely follows the analysis in \cite{soare2014best}, with generalization to the transductive setting and more careful analysis in terms of matrix inversion. 
\begin{lemma}
\label{ms_bai:lm:psi_ub_lb}
Fix $\cZ \subseteq \cX \subseteq \R^D$. Suppose $\max_{x \in \cX} \norm{x}^2 \leq c_1$ and $\min_{z \in \cZ \setminus \curly*{z_\star}}\norm{z_\star - z}^2 \geq c_2$ with some absolute constant $c_1$ and $c_2$. We have 
\begin{align*}
    \frac{c_2}{c_1 \Delta_{\min}^2} \leq \rho^\star \ldef \inf_{\lambda \in \simp_{\cX}} \sup_{z \in \cZ \setminus \{z_\star \}} \frac{\norm*{z_{\star}-z}^2_{A(\lambda)^{-1}}}{\Delta_z^2},
\end{align*}
where $\Delta_{\min} = \min_{z \in \cZ \setminus \curly{z_\star}}\curly{\Delta_z}$.
\end{lemma}
\begin{proof}
Let $\lambda^\star$ be the optimal design that attains $\rho^\star$;\footnote{If the infimum is not attained, one can apply the argument that follows with a limit sequence. See footnote in \cref{ms_bai:app:inverse} for more details on how to construct an approximating design.} and let $z^\prime \in \cZ$ be any arm with the smallest sub-optimality gap $\Delta_{\min}$. We then have 
\begin{align}
    \rho^\star & = \max_{z \in \cZ \setminus \{z_\star \}} \frac{\norm*{z_{\star}-z}^2_{A(\lambda^\star)^{-1}}}{\Delta_z^2} \nonumber\\
    & \geq \frac{\norm*{z_{\star}-z^\prime}^2_{A(\lambda^\star)^{-1}}}{\Delta_{z^\prime}^2} \nonumber \\
    & = \frac{\norm*{z_{\star}-z^\prime}^2_{A(\lambda^\star)^{-1}}}{\Delta_{\min}^2}, \label{ms_bai:eq:psi_ub_lb_Delta_min}
\end{align}
where $z_\star - z^\prime$ necessarily lie in the range of $A(\lambda^\star)$ according to the definition of matrix inversion in \cref{ms_bai:app:inverse}. 

We now lower bound $\norm*{z_{\star}-z^\prime}^2_{A(\lambda^\star)^{-1}}$. Note that $A(\lambda^\star)$ is positive semi-definite. We write $A(\lambda^\star) = Q \Sigma Q^\top$ where $Q$ is an orthogonal matrix and $\Sigma$ is a diagonal matrix storing eigenvalues. We assume that the last $k$ eigenvalues of $\Sigma$ are zero. Let $\gamma_{\max} = \norm{A(\lambda^\star)}_2 = \norm{\Sigma}_2$ be the largest eigenvalue, we have $\gamma_{\max} \leq \max_{x \in \cX} \norm{x}^2 \leq c_1$ since $A(\lambda^\star) = \sum_{x \in \cX} \lambda^\star(x) x x^\top$ and $\sum_{x \in \cX} \lambda^\star (x) = 1$. Let $w = Q^\top (z_\star - z^\prime)$. Since $z_\star - z^\prime$ is in the range of $A(\lambda^\star)$, we know that the last $k$ entries of $w$ must be zero. We then have 
\begin{align}
    \norm*{z_{\star}-z^\prime}^2_{A(\lambda^\star)^{-1}} & = (z_\star - z)^\top A(\lambda^\star)^{-1} (z_\star - z) \nonumber \\
    & = w^\top \Sigma^{-1} w  \nonumber \\
    & \geq{\norm{w}^2}/{c_1} \nonumber \\
    & \geq c_2/c_1, \label{ms_bai:eq:psi_ub_lb_dist}
\end{align}
where \cref{ms_bai:eq:psi_ub_lb_dist} comes from fact that $\norm{w}^2 = \norm{z_\star - z^\prime}^2$ and the assumption $\norm{z_\star - z}^2 \geq c_2$ for all $z \in \cZ$.
\end{proof}

\begin{lemma}
\label{ms_bai:lm:relation_log}
The following statements hold.
\begin{enumerate}
    \item $T \geq 4a \log 2a \implies T \geq a \log_2 T$ for $T,a > 0$.
    \item $T \geq 16 a \, (\log 16 a)^2 \implies T \geq a \, (\log_2 T)^2$ for $T,a > 1$.
\end{enumerate}
\end{lemma}

\begin{proof}
We first recall that $T \geq 2a \log a \implies T \geq a \log T$ for $T,a > 0$ \citep{shalev2014understanding}. Since $\log_2 T = \log T / \log 2 < 2 \log T$, the first statement immediately follows.

To prove the second statement, we only need to find conditions on $T$ such that $T \geq 4a \, (\log T)^2$. Note that we have $\sqrt{T} \geq 8 \sqrt{a}  \log 4 \sqrt{a} = 4 \sqrt{a} \log 16 a \implies \sqrt{T} \geq 4 \sqrt{a} \log \sqrt{T} = 2 \sqrt{a} \log T$. For $T,a > 1$, this is equivalent to 
$T \geq 16 a \, (\log 16 a)^2 \implies T \geq 4 a \, (\log T)^2 \geq a \, (\log_2 T)^2$, and thus the second statement follows.
\end{proof}

\subsubsection{Supporting Algorithms}
\label{ms_bai:app:supporting_alg}

\begin{algorithm}[H]
    \caption{\OPT}
    \label{ms_bai:alg:opt}
    \renewcommand{\algorithmicrequire}{\textbf{Input:}}
	\renewcommand{\algorithmicensure}{\textbf{Output:}}
	\begin{algorithmic}[1]
	    \REQUIRE Selection budget $B$, dimension upper bound $D$ and selection function $g(\cdot)$ (which is a function of the dimension $d \in [D]$).
	    \STATE Get $d_k$ such that
	        \begin{align*}
                d_k   = & \max d \\
                & \text{ s.t. } g(d) \leq B, \text{ and } d \in [D] \nonumber .
            \end{align*}
        \ENSURE The selected dimension $d_k$.
	\end{algorithmic}
\end{algorithm}

\subsection{Proofs and Supporting Results for \cref{ms_bai:sec:true_complexity}}

\subsubsection{Proof of \cref{ms_bai:thm:lower_bound_delta_PAC}}
\label{ms_bai:app:lower_bound_PAC}
\thmLowerBoundDeltaPAC*
\begin{proof}
The proof of the theorem mostly follows the proof of lower bound in \cite{fiez2019sequential}. We additionally consider the model selection problem $(\cX, \cZ$, $\theta_\star \in \Theta_{d_\star})$ and carefully deal with the matrix inversion. 

Consider the instance $(\cX, \cZ$, $\theta_\star \in \Theta_{d_\star})$, where $\cX = \{x_1,\ldots, x_n\}$ and $\spn(\cX) = \R^D$, $\cZ = \{z_1,\ldots, z_m\}$. Suppose that $z_1 = \argmax_{z \in \cZ} \ang*{\theta_\star, z}$. We consider the alternative set $\cC_{d_\star} \ldef \curly*{ \theta \in \Theta_{d_\star}: \exists i \in [m] \text{ s.t. } \ang*{\theta, z_1 - z_i} < 0 }$, where $z_1$ is not the best arm for any $\theta \in \cC_{d_\star}$. Following the ``change of measure'' argument in \cref{ms_bai:lm:change_of_measure}, we know that $\E_{\theta_\star}[\tau] \geq \tau^\star$, where $\tau^\star$ is the solution of the following constrained optimization
\begin{align}
    \tau^\star  & \ldef \min_{t_1, \dots, t_n \in \R_{+}} \sum_{i=1}^n t_i \label{ms_bai:eq:lower_bound_PAC_opt} \\
    & \qquad \text{ s.t. } \inf_{\theta \in \cC_{d_\star}} \sum_{i=1}^n t_i \kl(\nu_{\theta_\star,i}, \nu_{\theta,i}) \geq \log(1/2.4 \delta)\nonumber ,
\end{align}
where we use the notation $\nu_{\theta,i} = \cN ( \ang*{ \theta, x_i }, 1) = \cN (\ang*{\psi_{d_\star}(\theta), \psi_{d_\star}(x_i)}, 1)$ (due to the fact that $\theta \in \cC_{d_\star}$). We also have $\kl(\nu_{\theta_\star,i}, \nu_{\theta,i}) = \frac{1}{2} \ang*{\psi_{d_\star}(\theta_\star)-\psi_{d_\star}(\theta), \psi_{d_\star}(x_i)}^2$.

We next show that for any $t = (t_1, \dots, t_n)^\top \in \R_+^n$ satisfies the constraint of \cref{ms_bai:eq:lower_bound_PAC_opt}, we must have $\psi_{d_\star}(z_1) -\psi_{d_\star}(z_i) \in \spn(\{ \psi_{d_\star}(x_i) : t_i > 0\}), \forall \, 2\leq i \leq m$. Suppose not, there must exists a $\psi_{d_\star}(u) \in \R^{d_\star}$ such that (1) $\ang*{\psi_{d_\star}(u), \psi_{d_\star}(x_i)} = 0$ for all $i \in [n]$ such that $t_i > 0$; and (2) there exists a $2 \leq j \leq m$ such that $\ang*{\psi_{d_\star}(z_1) - \psi_{d_\star}(z_j), \psi_{d_\star}(u) } \neq 0$. Suppose $\ang*{\psi_{d_\star}(z_1 ) - \psi_{d_\star}(z_j), \psi_{d_\star}(u) } > 0$ (the other direction is similar), we can choose a $\theta^\prime \in \Theta_{d_\star}$ such that the first $d_\star$ coordinates of $\theta^\prime$ equals to $\psi_{d_\star}(\theta_\star) - \alpha \, \psi_{d_\star}(u)$ for a $\alpha > 0$ large enough (so that $\theta^\prime \in \cC_{d_\star}$). With such $\theta^\prime$, however, we have 
\begin{align*}
    \sum_{i=1}^n t_i\kl(\nu_{\theta_\star,i}, \nu_{\theta^\prime,i}) = \sum_{i=1}^n t_i \frac{1}{2} \ang*{\alpha \, \psi_{d_\star}(u), \psi_{d_\star}(x_i)}^2 = 0 < \log(1/2.4 \delta),
\end{align*}
which leads to a contradiction. As a result, we can safely calculate \linebreak
$\norm{\psi_{d_\star}(z_1) -\psi_{d_\star}(z_i)}^2_{A_{d_\star}(t)^{-1}}$ or $A_{d_\star}(t)^{-1} (\psi_{d_\star}(z_1) -\psi_{d_\star}(z_i))$ where \linebreak
$A_{d_\star}(t) \ldef \sum_{i=1}^n t_i \psi_{d_\star}(x_i) \psi_{d_\star}(x_i)^\top / \bar t$ and $\bar t \ldef \sum_{i=1}^n t_i$. The rest of the proof follows from the proof of theorem 1 in \cite{fiez2019sequential}.
\end{proof}

\subsubsection{Proof of \cref{ms_bai:thm:lower_bound_non_interactive}}
\label{ms_bai:app:non_interactive}
\thmLowerBoundNonInteractive*
\begin{proof}
The proof largely follows from the proof of Theorem 3 in \cite{katz2020empirical} (but ignore the $\gamma^\star$ term therein. We are effectively using a weaker lower bound, yet it suffices for our purpose. ). The non-interactive MLE uses at least $\frac{1}{2} \rho^\star_d \log(1/\delta)$ with respect to any feature mapping $\psi_d(\cdot)$ for $d_\star \leq d \leq D$. The statement then follows from the monotonicity of $\curly*{\rho^\star_d}_{d=d_\star}^D$ as shown in \cref{ms_bai:prop:rho_monotonic}.
\end{proof}

\subsubsection{Proof of \cref{ms_bai:prop:rho_monotonic}}

\propRhoMonotonic*

\begin{proof}

We first prove equivalence results in the general setting in Step 1, 2 and 3; and then apply the results to the model selection problem in Step 4 to prove monotonicity over $\curly*{\rho_d^\star}_{d = d_\star}^D$. 

We consider instance $(\cX, \cZ$, $\theta_\star)$ in the general setting, where $\cX = \{x_1,\ldots, x_n\} \subseteq \R^d$, $\spn(\cX) = \R^d$, $\cZ = \{z_1,\ldots, z_m\}$ and $\theta_\star \in \R^d$. We suppose that \linebreak
$z_1  = \argmax_{z \in \cZ} \ang*{\theta_\star, z}$ is the unique optimal arm and $\spn(\curly*{z_1 - z}_{z \in \cZ \setminus \curly*{z_1}}) = \R^d$. We use the notations $y_j \ldef z_1-z_j$ for $j=2,\ldots, m$, and $\nu_{\theta,i} \ldef \mathcal{N}(x_i^\top \theta, 1)$. For any $t = (t_1, \ldots, t_n)^\top \in \R^n_+$, we also use the notation $A(t) = \sum_{i=1}^n t_i x_i x_i^\top \in \R^{d \times d}$ to denote a design matrix with respect to $t$ ($t$ doesn't need to be inside the simplex $\simp_{\cX}$). We consider any fixed $\delta \in (0,0.15]$.

\textbf{Step 1: Closure of constraints.} Let $\cC$ denote the set of parameters where $z_1$ is no longer the best arm anymore, i.e.,
\begin{align*}
    \cC & \ldef \{\theta \in \R^d : \exists i \in [m] \text{ s.t. } \theta^\top (z_1-z_i) < 0 \}.
\end{align*}
Using the ``change of measure'' argument from \cite{kaufmann2016complexity}, the lower bound is given by the following optimization problem \citep{audibert2010best, fiez2019sequential}
\begin{align*}
    \tau^\star & := \min_{t_1,\ldots, t_n \in \R_+} \sum_{i=1}^n t_i \\
    & \qquad \text{ s.t. } \inf_{\theta \in \cC} \sum_{i=1}^n t_i \kl( \nu_{\theta_\star,i}, \nu_{\theta,i}) \geq \log(1/2.4 \delta). 
\end{align*}
First, we show that the value $\tau^\star$ equals to the value of another optimization problem, i.e.,
\begin{align*}
    \tau^\star & = \min_{t_1,\ldots, t_n \in \R_+} \sum_{i=1}^n t_i \\
    & \qquad \text{ s.t. } \min_{\theta \in \bar{\cC}} \sum_{i=1}^n t_i \kl( \nu_{\theta_\star,i}, \nu_{\theta,i}) \geq \log(1/2.4 \delta),
\end{align*}
where $\bar{\cC} =  \{\theta \in \R^d : \exists i \in [m] \text{ s.t. } \theta^\top (z_1-z_i) \leq 0 \}$. Note that that we must show that the minimum in the constraint is attained, i.e., the $\min_{\theta \in \bar \cC}$ part. We first show the equivalence between the original problem and the problem with respect to $\inf_{\theta \in \bar \cC}$; and then show the equivalence between problems with respect to $\inf_{\theta \in \bar \cC}$ and $\min_{\theta \in \bar \cC}$. We fix any $t=(t_1,\ldots,t_n)^\top \in \R^n_+$.

\textbf{Step 1.1:} We claim that $\inf_{\theta \in \cC} \sum_{i=1}^n t_i \kl( \nu_{\theta_\star,i}, \nu_{\theta,i}) \geq \log(1/2.4 \delta)$ if and only if $\inf_{\theta \in \bar{\cC}} \sum_{i=1}^n t_i \kl( \nu_{\theta_\star,i},  \nu_{\theta,i}) \geq \log(1/2.4 \delta)$. 

Since $\bar{\cC} \supset \cC$, the $\Longleftarrow$ direction is obvious. 

Now, suppose $\inf_{\theta \in \bar{\cC}} \sum_{i=1}^n t_i \kl( \nu_{\theta_\star,i}, \nu_{\theta,i}) < \log(1/2.4 \delta)$. By definition of $\inf$, there exists $\theta_0 \in \bar{\cC}$ such that
\begin{align*}
     \sum_{i=1}^n t_i \kl( \nu_{\theta_\star,i}, \nu_{\theta_0,i}) < \log(1/2.4 \delta).
\end{align*}
Since $\bar \cC$ is the closure of an open set $\cC$, there exists a sequence $\curly*{\theta_j}$ in $\cC$ approaching $\theta_0$. 
Note that 
\begin{align*}
    \sum_{i=1}^n t_i \kl( \nu_{\theta_\star,i}, \nu_{\theta,i}) = \sum_{i=1}^n t_i  \frac{1}{2} (x_i^\top (\theta_\star - \theta))^2 = \frac{1}{2} \norm{\theta_\star-\theta}_{A(t)}^2.
\end{align*}
Then, by the continuity of $\frac{1}{2} \norm{\theta_\star -\theta}_{A(t)}^2$ in $\theta$, there exists a $\theta \in \cC$ such that $\sum_{i=1}^n t_i \kl( \nu_{\theta_\star,i}, \nu_{\theta,i}) < \log(1/2.4 \delta)$. This gives a contradiction and thus proves the $\Longrightarrow$ direction.

\textbf{Step 1.2:} Now, we must show that the infimum is attained whenever 
$$\inf_{\theta \in \bar \cC} \sum_{i=1}^n t_i \kl( \nu_{\theta_\star,i} || \nu_{\theta,i}) \geq \log(1/2.4 \delta),$$ 
that is, there exists $\theta_0 \in \bar \cC $ such that
\begin{align*}
  \sum_{i=1}^n t_i \kl( \nu_{\theta_\star,i}, \nu_{\theta_0,i}) & =   \inf_{\theta \in \bar{\cC}} \sum_{i=1}^n t_i \kl( \nu_{\theta_\star,i}, \nu_{\theta,i}).
\end{align*}

\textbf{Claim:} Fix $t = (t_1,\ldots, t_n)^\top \in \R_+^n$. If $\spn(\{x_i : t_i > 0\}) \neq \R^d$, then 
$$\inf_{\theta \in \bar{\cC}} \sum_{i=1}^n t_i \kl( \nu_{\theta_\star,i} , \nu_{\theta,i}) < \log(1/2.4 \delta).$$ 

First, we show the claim. Fix $t = (t_1,\ldots, t_n)^\top \in \R_+^n$ and suppose $\spn(\{x_i : t_i > 0\}) \neq \R^d$. Since $\spn(\{x_i : t_i > 0\}) \neq \R^d$, there exists $u \in \R^d $ such that $u^\top x_i = 0$ for all $i$ such that $t_i > 0$. Since $\{z_1-z_i : i \in [m]\}$ spans $\R^d$ by assumption, there exists $i \in [m]$ such that $u^\top (z_1-z_i) \neq 0$. Suppose that $u^\top (z_1-z_i) < 0$ (the other case is similar). Then, there exists a sufficiently large $\alpha > 0$ such that $(\theta_\star + \alpha u)^\top (z_1 -z_i) < 0$, implying that $\theta_\star + \alpha u \in \cC$. Moreover, by construction of $u$, we have
\begin{align*}
    \sum_{i=1}^n t_i \kl( \nu_{\theta_\star,i}, \nu_{\theta_\star + \alpha u,i}) &  = \sum_{i=1}^n t_i  \frac{1}{2} (x_i^\top (\alpha u))^2 =  \sum_{i : t_i > 0} t_i  \frac{1}{2} (x_i^\top (\alpha u))^2 = 0 < \log(1/2.4 \delta),
\end{align*}
and thus leads to the claim.

Now, suppose $\inf_{\theta \in \bar{\cC}} \sum_{i=1}^n t_i \kl( \nu_{\theta_\star,i}, \nu_{\theta,i}) \geq \log(1/2.4 \delta)$. Then, $\spn(\{x_i : t_i > 0\}) = \R^d$. Then, $\norm{\cdot}_{A(t)}^2$ is a norm, and the set
\begin{align*}
    \curly*{ \theta \in \R^d : \frac{1}{2} \norm{\theta-\theta_\star}_{A(t)}^2 \leq \epsilon }
\end{align*}
is compact for every $\epsilon$.  Then, since $\bar{\cC}$ is closed and $\frac{1}{2} \norm{\theta-\theta_\star}_{A(t)}^2$ has compact sublevel sets, there exists a $\theta_0 \in \bar{\cC}$ such that
\begin{align*}
    \sum_{i=1}^n t_i \kl( \nu_{\theta_\star,i} , \nu_{\theta_0,i}) = \inf_{\theta \in \bar{\cC}} \sum_{i=1}^n t_i \kl( \nu_{\theta_\star,i} , \nu_{\theta,i}).
\end{align*}
This shows the equivalence between problems with respect to $\inf_{\theta \in \bar \cC}$ and $\min_{\theta \in \bar \cC}$.

\textbf{Step 2: Rewrite the optimization problem.} Define
\begin{align*}
    \bar{\cC}_i & = \{\theta \in \R^d :  \theta^\top (z_1-z_i) \leq 0 \},
\end{align*}
and note that $\bar{\cC} = \cup_{i=1}^m \bar{\cC}_i$. Observe that
\begin{align*}
        \tau^\star & := \min_{t_1,\ldots, t_n \in \R_+} \sum_{i=1}^n t_i \\
    & \qquad \text{ s.t. } \min_{\theta \in \bar{\cC}} \sum_{i=1}^n t_i \kl( \nu_{\theta_\star,i}, \nu_{\theta,i}) \geq \log(1/2.4 \delta) \\
    & = \min_{t_1,\ldots, t_n \in \R_+} \sum_{i=1}^n t_i \\
    & \qquad \text{ s.t. } \min_{i \in [m]} \min_{\theta \in \bar{\cC}_i} \sum_{i=1}^n t_i \kl( \nu_{\theta_\star,i} , \nu_{\theta,i}) \geq \log(1/2.4 \delta).
\end{align*}

Consider the optimization problem:
\begin{align*}
    \min_{\theta \in \bar{\cC}_i} & \frac{1}{2} \sum_{i=1}^n t_i (x_i^\top (\theta_\star - \theta))^2 = \min_{\theta \in \bar{\cC}_i} \frac{1}{2} \norm{\theta_\star-\theta}_{A(t)}^2 
\end{align*}
Note that since the objective is convex and there exists $\theta \in \R^d$ such that $\theta^\top(z_1-z_i) < 0$, Slater's condition holds and, therefore, strong duality holds. We form the Lagrangian with lagrange multiplier $\gamma \in \R_+$ to obtain
\begin{align*}
    \L(\theta, \gamma ) &= \frac{1}{2} \norm{\theta_\star-\theta}_{A(t)}^2 + \gamma \cdot y_i^\top \theta  \\
\end{align*}
Differentiating with respect to $\theta$ and $\gamma$, we have that (note that $A(t)$ is invertible from the claim in Step 1)
\begin{align*}
\begin{cases}
        \theta  &= \theta_\star - \gamma A(t)^{-1} y_i, \\
     y_i^\top \theta  &= 0.
\end{cases}
\end{align*}
These imply that $\theta_0 \ldef  \theta_\star - \frac{y_i^\top \theta_\star A(t)^{-1} y_i}{y_i^\top A(t)^{-1} y_i}$ and $\gamma_0 \ldef \frac{y_i\top \theta_\star}{y_i^\top A(t)^{-1} y_i} \in \R_+$ satisfy the K.K.T. conditions, and $\theta = \theta_0$ is the minimizer (primal optimal solution) of the constrained optimization problem (note that it's a convex program). Therefore, we have
\begin{align*}
   \min_{\theta \in \bar{\cC}_{i}} & \frac{1}{2} \sum_{i=1}^n t_i (x_i^\top (\theta_\star - \theta))^2 = \frac{(y_i^\top \theta_\star )^2}{\norm{y_i}^2_{A(t)^{-1}}} 
\end{align*}
In conclusion, we have 
\begin{align*}
            \tau^\star & = \min_{t_1,\ldots, t_n \in \R_+} \sum_{i=1}^n t_i \\
    & \qquad \text{ s.t. } \frac{(y_j^\top \theta_\star )^2}{\norm{y_j}^2_{A(t)^{-1}}}  \geq \log(1/2.4 \delta) , \forall \, 2 \leq j \leq m.
\end{align*}

\textbf{Step 3: Re-express the optimization problem.} Furthermore, we have that
\begin{align}
    \tau^\star & = \min_{s, t_1,\ldots, t_n \in \R_+} s \label{ms_bai:eq:monotonic_original_opt}\\
    & \qquad \text{ s.t. }  (y_j^\top \theta_\star )^2 \geq \log(1/2.4 \delta) \norm{y_j}^2_{A(t)^{-1}}  , \forall \, 2 \leq j \leq m \nonumber\\
    & \qquad \qquad s \geq \sum_{i=1}^n t_i. \nonumber
\end{align}
Rearranging these constraints, we have that 
\begin{align*}
    s \geq \sum_{i=1}^n t_i \geq \log(1/2.4 \delta) \sum_{i=1}^n t_i \frac{\norm{y_j}^2_{A(t)^{-1}}}{(y_j^\top \theta_\star )^2} = \log(1/2.4 \delta)  \frac{\norm{y_j}^2_{A(\lambda)^{-1}}}{(y_j^\top \theta_\star )^2} , \forall \, 2 \leq j \leq m.
\end{align*}
We do a change of variables $\lambda \in \simp_{\cX}$ and $\lambda_i = \frac{t_i}{\sum_{i=1}^n t_i}$, and the optimization problem is equivalent to
\begin{align*}
     \tau^\star & = \min_{s \in \R_+, \lambda \in \simp_{\cX}} s \\
    & \qquad \text{ s.t. }   s \geq \max_{j =2,\ldots, m} \log(1/2.4 \delta)  \frac{\norm{y_j}^2_{A(\lambda)^{-1}}}{(y_j^\top \theta_\star )^2}.
\end{align*}
Thus, we have that
\begin{align*}
    \tau^\star \geq \inf_{\lambda \in \simp_{\cX}} \max_{j=2,\ldots, m} \frac{\norm{y_j}^2_{A(\lambda)^{-1}}}{(y_j^\top \theta_\star )^2} \log(1/2.4 \delta).
\end{align*}
Now let 
\begin{align*}
    \widetilde \tau^\star \ldef  \inf_{\lambda \in \simp_{\cX}} \max_{j=2,\ldots, m} \frac{\norm{y_j}^2_{A(\lambda)^{-1}}}{(y_j^\top \theta_\star )^2} \log(1/2.4 \delta) = \max_{j=2,\ldots, m} \frac{\norm{y_j}^2_{A(\lambda^\star)^{-1}}}{(y_j^\top \theta_\star )^2} \log(1/2.4 \delta),
\end{align*}
where $\lambda^\star$ is the optimal design of the above optimization problem.\footnote{Again, if the infimum is not attained, one can apply the argument that follows with a limit sequence. See footnote in \cref{ms_bai:app:inverse} for more details on how to construct an approximating design.} Set $\widetilde t = \widetilde \tau^\star \lambda^\star \in \R_+^n$ with $\widetilde t_i = \widetilde \tau^\star \lambda^\star_i \in \R_+$, we can then see that 
\begin{align*}
    \sum_{i=1}^n \widetilde t_i = \widetilde \tau^\star  = \max_{j=2,\ldots, m} \sum_{i=1}^n \widetilde t_i \frac{\norm{y_j}^2_{A(\widetilde t)^{-1}}}{(y_j^\top \theta_\star )^2} \log(1/2.4 \delta)  , \forall \, 2 \leq j \leq m.
\end{align*}
and such $\curly*{\widetilde{t}_i}$ satisfies the constraints in the original optimization problem described in \cref{ms_bai:eq:monotonic_original_opt}. 
As a result, we have $\tau^\star \leq \widetilde \tau^\star$.

We now can write 
\begin{align}
    \tau^\star = \inf_{\lambda \in \simp_{\cX}} \max_{j=2,\ldots, m} \frac{\norm{y_j}^2_{A(\lambda)^{-1}}}{(y_j^\top \theta_\star )^2} \log(1/2.4 \delta) = \rho^\star \log(1/2.4 \delta). \label{ms_bai:eq:monotonic_equivalence}
\end{align}

\textbf{Step 4: Monotonicity.} We now apply the established equivalence to the model selection problem and prove monotonicity over $\curly*{\rho_d^\star}_{d= d_\star}^D$.

Now, define
\begin{align*}
    \tau^\star_{d_\ell} & = \min_{t_1,\ldots, t_n \in \R_+} \sum_{i=1}^n t_i \\
    & \qquad \text{ s.t. } \inf_{\theta \in \cC_{d_\ell}} \sum_{i=1}^n t_i \kl( \nu_{\theta_\star,i}, \nu_{\theta,i}) \geq \log(1/2.4 \delta),
\end{align*}
where $\cC_{d_\ell} = \curly*{ \theta \in \R^D : \forall j > d_\ell : \theta_j = 0 \wedge \exists i \in [m] \text{ s.t. } \theta^\top (z_1-z_i) < 0 }$. Let $d_\star \leq d_1 \leq d_2 \leq D$. Then, since the optimization problem in $\tau^\star_{d_1}$ has fewer constraints than the optimization problem in $\tau^\star_{d_2}$, we have that $\tau^\star_{d_1} \leq \tau^\star_{d_2}$. The established equivalence in \cref{ms_bai:eq:monotonic_equivalence} can be applied with respect to feature mappings $\psi_d(\cdot)$ for $d_\star \leq d \leq D$ (note that we necessarily have $\spn(\curly*{\psi_d(z_\star) - \psi_d(z)}_{z \in \cZ \setminus \curly*{z_\star}}) = \R^d$ as long as $\spn(\curly*{z_\star - z}_{z \in \cZ \setminus \curly*{z_\star}}) = \R^D$). Therefore, we have
\begin{align*}
 \rho^\star_{d_1} \log(1/2.4 \delta)  =  \tau^\star_{d_1}  \leq \tau^\star_{d_2} = \rho^\star_{d_2} \log(1/2.4 \delta),
\end{align*}
leading to the desired result.
\end{proof}

\subsubsection{Proof of \cref{ms_bai:prop:rho_star_different_d}}
\label{ms_bai:app:rho_star_different_d}

\propRhoStarDifferentD*

\begin{proof}
For any $\lambda \in \simp_{\cX}$, we define
\begin{align*}
    \rho_{d}(\lambda) \ldef \max_{z \in \cZ \setminus \curly*{z_\star }} \frac{\norm*{\psi_d(z_{\star})-\psi_d(z)}^2_{A_{d}(\lambda)^{-1}}}{(h(z_\star) - h(z))^2},
\end{align*}
and 
\begin{align*}
    \iota_{d}(\lambda) \ldef \max_{z \in \cZ \setminus \curly*{z_\star }} {\norm*{\psi_d(z_{\star})-\psi_d(z)}^2_{A_{d}(\lambda)^{-1}}}.
\end{align*}

We consider an instance $\cX =\cZ = \curly*{x_i}_{i=1}^{d_\star + 1} \subseteq \R^{d_\star + 1}$ and expected reward function $h(\cdot)$. The action set is constructed as follows:
\begin{align*}
    x_i = e_i, \text{ for } i = 1, 2,\dots, d_\star, \quad x_{d_\star+1} = (1-\epsilon) \cdot e_{d_\star} + e_{d_\star+1},
\end{align*}
where $e_i$ is the $i$-th canonical basis in $\R^{d_\star + 1}$. The expected reward of each action is set as 
\begin{align*}
    h(x_i) \ldef \ang*{x_i, e_{d_\star}}.
\end{align*}
One can easily see that $d_\star$ is the intrinsic dimension of the problem (in fact, it is the smallest dimension such that linearity in rewards is preserved).

We notice that $\theta_\star \in \R^{d_\star}$; $x_\star = x_{d_\star}$ is the best arm with reward $1$, $x_{d_\star+1}$ is the second best arm with reward $1-\epsilon$ and all other arms have reward $0$. The smallest sub-optimality gap is $\Delta_{\min} = \epsilon$. $\epsilon \in (0,1/2]$ is selected such that $1/4\epsilon^2 > 2d_\star + \gamma$ for any given $\gamma > 0$.\footnote{One can also add an additional arm $x_{0} = e_D/2$ so that $\spn(\curly*{x_\star - x}_{x \in \cX}) = \R^{d_\star + 1}$ (the lower bound on $\rho_{d_\star + 1}^\star$ will be changed to $1/16\epsilon^2$).}

We first consider truncating arms into $\R^{d_\star}$. For any $\lambda \in \simp_{\cX}$, we notice that $A_{d_\star}(\lambda) = \sum_{x \in \cX} \lambda_x \psi_{d_\star}(x) \psi_{d_\star}(x)^\top$ is a diagonal matrix with the $d_\star$-th entry being $\lambda_{x_{d_\star}} + (1-\epsilon)^2 \lambda_{x_{d_\star+1}}$ and the rest entries being $\lambda_{x_i}$. We first show that $\iota_{d_\star}^\star \geq d_{\star}-1$ by contradiction as follows. Suppose $\iota^\star_{d_\star} < d_\star-1$. Since $\norm*{\psi_{d_\star}(x_\star) - \psi_{d_\star}(x_i)}^2_{A_{d_\star}(\lambda)^{-1}} \geq 1/\lambda_{x_i}$ for $i = 1, 2, \dots, d_\star-1$, we must have $\lambda_{x_i} > 1/(d_\star-1)$ for $i = 1, 2, \dots, d_\star-1$. Thus, $\sum_{i=1}^{d_\star-1} \lambda_{x_i} >  1$, which leads to a contradiction for $\lambda \in \simp_{\cX}$. We next analyze $\rho^\star_d$. Let $\lambda^\prime \in \simp_{\cX}$ be the design such that $\lambda^\prime_{x_i} = 1/d_{\star}$ for $i = 1,\dots, d_\star$. With design $\lambda^\prime$, we have $\norm*{\psi_{d_\star}(x_\star) - \psi_{d_\star}(x_i)}^2_{A_{d_\star}(\lambda^\prime)^{-1}} = 2d_\star$ for $i = 1, 2, \dots, d_\star-1$ and $\norm*{\psi_{d_\star}(x_\star) - \psi_{d_\star}(x_{d_\star + 1})}^2_{A_{d_\star}(\lambda^\prime)^{-1}} = \epsilon^2 d_\star$. As a result, we have $\rho_{d_\star}(\lambda^\prime) \leq 2d_\star$, and thus $\rho^\star_{d_\star} \leq \rho_{d_\star}(\lambda^\prime) \leq 2d_\star$.

We now consider arms in the original space, i.e., $\R^{d_\star+1}$. We first upper bound $\iota^\star_{d_\star + 1}$. With an uniform design $\lambda^{\prime \prime}$ such that $\lambda_{x_i}^{\prime \prime} = 1/(d_\star+1), \forall i \in [d_\star + 1]$, we have $\iota^\star_{d_\star+1} \leq \iota_{d_\star + 1}(\lambda^{\prime \prime}) \leq \max \curly*{(3-\epsilon)/(2-\epsilon), \epsilon^2/(2-\epsilon) + 1} \cdot (d_\star + 1) \leq 5(d_\star+1)/3$ when $\epsilon \in (0,1/2]$. In fact, with the same design, we can also upper bound $\iota(\cY(\psi_{d_\star+1}(\cX))) \leq 3(d_\star + 1)$. We analyze $\rho_{d_\star + 1}^\star$ now. Since $\max_{x \in \cX} \norm*{x}^2 \leq 4$ and $\min_{x \in \cX \setminus \curly*{x_\star}} \norm*{x_\star - x}^2 \geq 1$, \cref{ms_bai:lm:psi_ub_lb} leads to the fact that $\rho^\star_{d_\star+1} \geq 1/4 \epsilon^2$. Note that we only have $\min_{x \in \cX \setminus \curly*{x_\star}} \norm*{\psi_{d_\star}(x_\star) - \psi_{d_\star }(x)}^2 \geq \epsilon^2$ when truncating arms into $\R^{d_\star}$.

To summarize, for any given $\gamma > 0$, we have $\rho^\star_{d_\star+1} > \rho^\star_{d_\star} + \gamma$ yet $\iota^\star_{d_\star+1} \leq 2\iota^\star_{d_\star}$ (when $d_\star \geq 11$). Further more, we also have $\iota(\cY(\psi_{d_\star+1}(\cX))) \leq 4 \iota(\cY(\psi_{d_\star}(\cX)))$ (when $d_\star \geq 7$) since $\iota(\cY(\psi_{d_\star}(\cX))) \leq \iota_{d_\star}^\star$.
\end{proof}

\subsection{Proofs and Supporting Results for \cref{ms_bai:sec:fixed_confidence}}
\label{ms_bai:app:fixed_confidence}

\subsubsection{Proof of \cref{ms_bai:lm:subroutine_fixed_confidence}}

\lmSubroutineFixedConfidence*

\begin{proof}
We consider event 
\begin{align*}
    \cE_k = \curly{z_\star \in \widehat \cS_k \subseteq \cS_k},
\end{align*}
and prove through induction that 
\begin{align*}
    \P \paren{ \cE_{k+1} \mid \cap_{i \leq k} \cE_i } \geq 1 - \delta_{k},
\end{align*}
where $\delta_0 \ldef 0$. Recall that $\cS_k = \curly*{z \in \cZ: \Delta_z < 4 \cdot 2^{-k}}$ (with $\cS_1 = \cZ$).

\textbf{Step 1: The induction.} We have $\curly*{ z_\star \in \widehat \cS_1 \subseteq \cS_1 }$ since $\widehat \cS_1 = \cS_1 = \cZ$ by definition for the base case (recall that we assume $\max_{z \in \cZ} \Delta_z \leq 2$). We now assume that $\cap_{i \leq k}\cE_{i}$ holds true and we prove for iteration $k+1$. We only need to consider the case when $\abs*{\widehat \cS_k} > 1$, which implies $\abs*{\cS_k} > 1$ and thus $k \leq \floor*{ \log_2 (4 / \Delta_{\min}) }$.

\textbf{Step 1.1: $d_k \geq d_\star$ (Linearity is preserved).} Since $\widehat \cS_{k} \subseteq \cS_k$, we have 
\begin{align}
    g_k(d_\star) & = \max \curly{ {2^{2k} \iota(\cY(\psi_{d_\star}(\widehat \cS_k)))}, r_{d_\star}(\zeta) } \nonumber \\
    & \leq \max \curly{ {2^{2k} \iota(\cY(\psi_{d_\star}(\cS_k)))}, r_{d_\star}(\zeta) } \nonumber\\
    & \leq \max \curly{64 \rho_{d_\star}^\star, r_{d_\star}(\zeta)} \label{ms_bai:eq:subroutine_fixed_confidence_rho_stratified}\\
    & \leq B \label{ms_bai:eq:subroutine_fixed_confidence_B},
\end{align}
where \cref{ms_bai:eq:subroutine_fixed_confidence_rho_stratified} comes from \cref{ms_bai:lm:rho_stratified} and \cref{ms_bai:eq:subroutine_fixed_confidence_B} comes from the assumption. As a result, we know that $d_k \geq d_\star$ since $d_k$ is selected as the largest integer such that $g_k(d_k) \leq B$. 

\textbf{Step 1.2: Concentration.} Let $\curly*{x_1, \ldots, x_{N_k}} $ be the arms pulled at iteration $k$ and $\curly*{r_1, \ldots, r_{N_k}}$ be the corresponding rewards. Let $\widehat{\theta}_k = A_k^{-1} b_k \in \R^{d_k}$ where $A_k = \sum_{i=1}^{N_k} \psi_{d_k}(x_i) \psi_{d_k}(x_i)^\top$, and $b_k =  \sum_{i=1}^{N_k} \psi_{d_k}(x_i) b_i$. Since $d_k \geq d_\star$ and the model is well-specified, we can write $r_i = \ang*{\theta_\star, x_i} + \xi_i = \ang*{\psi_{d_k}(\theta_\star), \psi_{d_k}(x_i)} + \xi_i$, where $\xi_i$ is i.i.d. generated $1$-sub-Gaussian noise. For any $y \in \cY(\psi_{d_k}(\widehat \cS_k))$, we have
\begin{align*}
    \ang*{y, \widehat \theta_k - \psi_{d_k}(\theta_\star)} & = y^\top A_k^{-1} \sum_{i=1}^{N_k} \psi_{d_k}(x_i) r_i - y^\top \psi_{d_k}(\theta_\star) \nonumber \\
    & = y^\top A_k^{-1} \sum_{i=1}^{N_k} \psi_{d_k}(x_i) \paren{ \psi_{d_k}(x_i) ^\top \psi_{d_k}(\theta_\star) + \xi_i} - y^\top \psi_{d_k}(\theta_\star) \nonumber \\
    & = y^\top A_k^{-1} \sum_{i=1}^{N_k} \psi_{d_k}(x_i) \xi_i.
\end{align*}
Since $\xi_i$s are independent 1-sub-Gaussian random variables, we know that the random variable $ y^\top A_k^{-1} \sum_{i=1}^{N_k} \psi_{d_k}(x_i) \xi_i $ has variance proxy \linebreak
$\sqrt{\sum_{i=1}^{N_k} \paren{y^\top A_k^{-1} \sum_{i=1}^{N_k} \psi_{d_k}(x_i)}^2 } = \norm{y}_{A_k^{-1}}$. Combining the standard Hoeffding's inequality with a union bound leads to
\begin{align}
    \P \paren*{ \forall y \in \cY(\psi_{d_k}(\widehat \cS_k)) , \abs*{ \ang*{ y, \widehat \theta_k - \psi_{d_k}(\theta_\star)  } } \leq \norm{y}_{A_k^{-1}} \sqrt{2 \log \paren*{ {\abs{\widehat \cS_k}^2}/{\delta_k} }}  } \geq 1-\delta_k, \label{ms_bai:eq:subroutine_fixed_confidence_event}
\end{align}
where we use the fact that $\abs{ \cY(\psi_{d_k}(\widehat \cS_k)) } \leq \abs{\widehat \cS_k}^2/2$ in the union bound. 

\textbf{Step 1.3: Correctness.} We prove $z_\star \in \widehat \cS_{k+1} \subseteq \cS_{k+1}$ under the good event analyzed in \cref{ms_bai:eq:subroutine_fixed_confidence_event}.

\textbf{Step 1.3.1: $z_\star \in \widehat \cS_{k+1}$.} For any $\widehat z \in \widehat \cS_k$ such that $\widehat z \neq z_\star$, we have 
\begin{align*}
    & \ang{ \psi_{d_k}(\widehat z) - \psi_{d_k}(z_\star) , \widehat \theta_k } \\
    & \leq \ang{ \psi_{d_k}(\widehat z) - \psi_{d_k}(z_\star) , \psi_{d_k} (\theta_\star) } + \norm{\psi_{d_k}(\widehat z) - \psi_{d_k}(z_\star)}_{A_k^{-1}} \sqrt{2 \log \paren*{ {\abs{\widehat \cS_k}^2}/{\delta_k} }} \\
    & = h(\widehat z ) - h(z_\star) + \norm{\psi_{d_k}(\widehat z) - \psi_{d_k}(z_\star)}_{A_k^{-1}} \sqrt{2 \log \paren*{ {\abs{\widehat \cS_k}^2}/{\delta_k} }} \\
    & < \norm{\psi_{d_k}(\widehat z) - \psi_{d_k}(z_\star)}_{A_k^{-1}} \sqrt{2 \log \paren*{ {\abs{\widehat \cS_k}^2}/{\delta_k} }}.
\end{align*}
As a result, $z_\star$ remains in $\widehat \cS_{k+1}$ according to the elimination criteria.

\textbf{Step 1.3.2: $\widehat \cS_{k+1} \subseteq \cS_{k+1}$.} Consider any $z \in \widehat \cS_k \cap \cS_{k+1}^c$, we know that $\Delta_z \geq 2\cdot 2^{-k}$ by definition. Since $z_\star \in \widehat \cS_k$, we then have 
\begin{align}
    & \ang{ \psi_{d_k}(z_\star) - \psi_{d_k}(z) , \widehat \theta_k } \nonumber \\
    & \geq \ang{ \psi_{d_k}(z_\star) - \psi_{d_k}(z) , \psi_{d_k} (\theta_\star) } - \norm{\psi_{d_k}(z_\star) - \psi_{d_k}(z)}_{A_k^{-1}} \sqrt{2 \log \paren*{ {\abs{\widehat \cS_k}^2}/{\delta_k} }} \nonumber \\
    & = h(z_\star) - h(z) - \norm{\psi_{d_k}(z_\star) - \psi_{d_k}(z)}_{A_k^{-1}} \sqrt{2 \log \paren*{ {\abs{\widehat \cS_k}^2}/{\delta_k} }} \nonumber \\ 
    & \geq 2 \cdot 2^{-k} - \norm{\psi_{d_k}(z_\star) - \psi_{d_k}(z)}_{A_k^{-1}} \sqrt{2 \log \paren*{ {\abs{\widehat \cS_k}^2}/{\delta_k} }} \nonumber \\
    & \geq \norm{\psi_{d_k}(z_\star) - \psi_{d_k}(z)}_{A_k^{-1}} \sqrt{2 \log \paren*{ {\abs{\widehat \cS_k}^2}/{\delta_k} }} \label{ms_bai:eq:subroutine_fixed_confidence_rounding},
\end{align}
where \cref{ms_bai:eq:subroutine_fixed_confidence_rounding} comes from the fact that $\norm{\psi_{d_k}(z_\star) - \psi_{d_k}(z)}_{A_k^{-1}} \sqrt{2 \log \paren*{ {\abs{\widehat \cS_k}^2}/{\delta_k} }} \leq 2^{-k}$, which is resulted from the choice of $N_k$ and the guarantee in \cref{ms_bai:eq:rounding} from the rounding procedure. As a result, we have $z \notin \widehat \cS_{k+1}$ and $\widehat \cS_{k+1} \subseteq \cS_{k+1}$.

To summarize, we prove the induction at iteration $k+1$, i.e.,
\begin{align*}
    \P \paren{ \cE_{k+1} \mid \cap_{i< k+1} \cE_{i} } \geq 1 - \delta_k.
\end{align*}

\textbf{Step 2: The error probability.} Let $\cE = \cap_{i=1}^{n+1} \cE_{i}$ denote the good event, we then have 
\begin{align}
    \P \paren*{ \cE } & = \prod_{k=1}^{n} \P \paren*{ \cE_k \mid \cE_{k-1} \cap \dots \cap \cE_1 } \nonumber \\
    & = \prod_{k=1}^{n} \paren*{1 - \delta_k} \nonumber \\
    & \geq \prod_{k=1}^\infty \paren*{1 - \delta/k^2} \nonumber \\
    & = \frac{\sin(\pi \delta)} {\pi \delta} \nonumber \\
    & \geq 1- \delta \label{ms_bai:eq:subroutine_fixed_confidence_delta},
\end{align}
where we use the fact that ${\sin(\pi \delta)}/{\pi \delta} \geq 1-\delta$ for any $\delta \in (0,1)$ in \cref{ms_bai:eq:subroutine_fixed_confidence_delta}. 
\end{proof}

\subsubsection{Proof of \cref{ms_bai:thm:doubling_fixed_confidence}}

\thmDoublingFixedConfidence*

\begin{proof}
The proof is decomposed into three steps: (1) locating good subroutines; (2) bounding error probability and (3) bounding unverifiable sample complexity.

\textbf{Step 1: Locating good subroutines.} Consider $B_\star = \max \curly{64 \rho^\star_{d_\star}, r_{d_\star}(\zeta)}$ and $n_\star = \ceil{ \log_2 (2/\Delta_{\min}) }$. For any subroutines invoked with $B_i \geq B_\star$ and $n_i \geq n_\star$, we know that, from \cref{ms_bai:lm:subroutine_fixed_confidence}, the output set of arms are those with sub-optimality gap $< \Delta_{\min}$, which is a singleton set containing the optimal arm, i.e., $\curly{z_\star}$. Let $i_\star = \ceil{\log_2 (B_\star)}$, $j_\star = \ceil{ \log_2 (n_\star) }$ and $\ell_\star = i_\star + j_\star$. We know that in outer loops $\ell \geq \ell_\star$, there must exists at least one subroutine invoked with $B_i = 2^{i_\star} \geq B_\star$ and $n_i = 2^{j_\star} \geq n_\star$. Once a subroutine, invoked with $B_i \geq B_\star$, outputs a singleton set, it must be the optimal arm $z_\star$ according to \cref{ms_bai:lm:subroutine_fixed_confidence} (up to small error probability, analyzed as below). Since, within each outer loop $\ell$, the value of $B_i = 2^{\ell - i}$ is chosen in a decreasing order, updating the recommendation and breaking the inner loop once a singleton set is identified will not miss the chance of recommending the optimal arm in later subroutines within outer loop $\ell$.

\textbf{Step 2: Error probability.} We consider the good event where all subroutines invoked in \cref{ms_bai:alg:doubling_fixed_confidence} with $B_i \geq B_\star$ and (any) $n_i$ correctly output a set of arms with sub-optimality gap $<2^{1-n_i}$ with probability at least $1 - \delta_\ell$, as shown in \cref{ms_bai:lm:subroutine_fixed_confidence}. This good event clearly happens with probability at least $1 - \sum_{\ell = 1}^\infty \sum_{i=1}^\ell \delta_\ell = 1- \sum_{\ell = 1}^\infty \delta/(2 \ell^2) > 1 - \delta$, after applying a union bound argument. We upper bound the unverifiable sample complexity under this event in the following.

\textbf{Step 3: Unverifiable sample complexity.} 
For any subroutine invoked within outer loop $\ell \leq \ell_\star$, we know, from \cref{ms_bai:alg:subroutine_fixed_budget}, that its sample complexity is upper bounded by (note that $\abs*{\cZ}^2 \geq 4$ trivially holds true)
\begin{align*}
    N_\ell & \leq n_i \paren*{ B_i \cdot \paren*{2.5 \, \log(\abs{\cZ}^2/\delta_{\ell_\star})} + 1 } \\
    & \leq \gamma_\ell \, 3.5 \, \log \paren*{ 2 \abs{\cZ}^2 \ell_\star^3/\delta }.
\end{align*}
Thus, the total sample complexity up to the end of outer loop $\ell_\star$ is upper bounded by 
\begin{align*}
    N & \leq \sum_{\ell = 1}^{\ell_\star} \ell N_\ell \\
    & \leq 3.5\, \log \paren*{ 2 \abs{\cZ}^2 \ell_\star^3/\delta } \sum_{\ell = 1}^{\ell_\star} \ell 2^\ell \\
    & \leq 7 \, \log \paren*{ 2 \abs{\cZ}^2 \ell_\star^3/\delta } \ell_\star 2^{\ell_\star}.
\end{align*}

Recall that $\tau_\star = \log_2(4/\Delta_{\min}) \max \curly*{\rho^\star_{d_\star}, r_{d_\star}(\zeta)}$. By definition of $\ell_\star$, we have
\begin{align*}
    \ell_\star \leq \log_2 \paren*{ 4  \log_2(4/\Delta_{\min})  \max \curly{64 \rho^\star_{d_\star}, r_{d_\star}(\zeta)} } = O(\log_2 (\tau_\star)),
\end{align*}
and
\begin{align*}
    2^{\ell_\star} & = 2^{(i_\star + j_\star)} \\
    & \leq 4  \paren*{ \log_2(2/\Delta_{\min})+1}  \max \curly{64 \rho^\star_{d_\star}, r_{d_\star}(\zeta)}, \\
    & = 4  \log_2(4/\Delta_{\min})  \max \curly{64 \rho^\star_{d_\star}, r_{d_\star}(\zeta)}, \\
    & = O(\tau_\star).
\end{align*}

 The unverifiable sample complexity is thus upper bounded by
\begin{align*}
    N & \leq 1792 \, \tau_\star \cdot \paren*{\log_2(\tau_\star) + 8} \cdot \log \paren*{ {2 \abs{\cZ}^2 \paren{\log_2(\tau_\star) + 8}^3}/{\delta} } \\
    & = O \paren*{ \tau_\star \log_2(\tau_\star) \log(\abs{\cZ} \log_2(\tau_\star)/\delta) }.
\end{align*}
\end{proof}

\subsection{Proofs and Supporting Results for \cref{ms_bai:sec:fixed_budget}}

\subsubsection{Proof of \cref{ms_bai:lm:subroutine_fixed_budget}}

\lmSubroutineFixedBudget*

\begin{proof}
We consider event 
\begin{align*}
    \cE_k = \curly{z_\star \in \widehat \cS_k \subseteq \cS_k},
\end{align*}
and prove through induction that 
\begin{align*}
    \P \paren*{ \cE_{k+1} \mid \cap_{i \leq k} \cE_i } \geq 1 - \delta_{k},
\end{align*}
where the value of $\curly{\delta_k}_{k=0}^{n}$ will be specified in the proof. 

\textbf{Step 1: The induction.} The base case $\curly{z_\star \in \widehat \cS_1 \subseteq \cS_1}$ holds with probability $1$ by construction (thus, we have $\delta_0 = 0$). Conditioned on events $\cap_{i=1}^k \cE_i$, we next analyze the event $\cE_{k+1}$. We only need to consider the case when $\abs{\widehat \cS_k} > 1$, which implies $\abs{\cS_k} > 1$ and thus $k \leq \floor{ \log_2 (4 / \Delta_{\min}) }$.

\textbf{Step 1.1: $d_k \geq d_\star$ (Linearity is preserved).} We first notice that $\widetilde D$ is selected as the largest integer such that $r_{\widetilde D}(\zeta) \leq T^\prime$, where $r_d(\zeta)$ represents the number of samples needed for the rounding procedure in $\R^d$ (with parameter $\zeta$). When $T/ n \geq r_{d_\star}(\zeta) + 1$, we have $\widetilde D \geq d_\star$ since $T^\prime \geq T/ n -1 \geq r_{d_\star}(\zeta)$. 
Thus, for whatever $d_k \in [\widetilde D]$ selected, we always have $r_{d_k}(\zeta) \leq r_{\widetilde D}(\zeta) \leq T^\prime$ and can thus safely apply the rounding procedure described in \cref{ms_bai:eq:rounding}.

Since $\widehat \cS_{k} \subseteq \cS_k$, we also have 
\begin{align}
    g_k(d_\star) & =  {2^{2k} \iota(\cY(\psi_{d_\star}(\widehat \cS_k)))} \nonumber \\
    & \leq {2^{2k} \iota(\cY(\psi_{d_\star}(\cS_k)))} \nonumber\\
    & \leq 64 \rho_{d_\star}^\star \label{ms_bai:eq:subroutine_fixed_budget_rho_stratified}\\
    & \leq B \label{ms_bai:eq:subroutine_fixed_budget_B},
\end{align}
where \cref{ms_bai:eq:subroutine_fixed_budget_rho_stratified} comes from \cref{ms_bai:lm:rho_stratified} and \cref{ms_bai:eq:subroutine_fixed_budget_B} comes from the assumption. As a result, we know that $d_{k} \geq d_\star$ since $d_{k} \in [\widetilde D]$ is selected as the largest integer such that $g_k(d_{k}) \leq B$. 

\textbf{Step 1.2: Concentration and error probability.} Let $\curly*{x_1, \ldots, x_{T^\prime}} $ be the arms pulled at iteration $k$ and $\curly*{r_1, \ldots, r_{T^\prime}}$ be the corresponding rewards. Let $\widehat{\theta}_k = A_k^{-1} b_k \in \R^{d_k}$ where $A_k = \sum_{i=1}^{T^\prime} \psi_{d_k}(x_i) \psi_{d_k}(x_i)^\top$, and $b_k =  \sum_{i=1}^{T^\prime} \psi_{d_k}(x_i) b_i$. Since $d_k \geq d_\star$ and the model is well-specified, we can write $r_i = \ang*{\theta_\star, x_i} + \xi_i = \ang*{\psi_{d_k}(\theta_\star), \psi_{d_k}(x_i)} + \xi_i$, where $\xi_i$ is i.i.d. generated zero-mean Gaussian noise with variance $1$. Similarly as analyzed in \cref{ms_bai:eq:subroutine_fixed_confidence_event}, we have 
\begin{align}
    \P \paren*{ \forall y \in \cY(\psi_{d_k}(\widehat \cS_k)) , \abs*{ \ang*{ y, \widehat \theta_k - \psi_{d_k}(\theta_\star)  } } \leq \norm{y}_{A_k^{-1}} \sqrt{2 \log \paren*{ {\abs{\widehat \cS_k}^2}/{\delta_k} }}  } \geq 1-\delta_k. \label{ms_bai:eq:subroutine_fixed_budget_event}
\end{align}
By setting $\max_{y \in \psi_{d_k}(\widehat \cS_k)} \norm{y}_{A_k^{-1}} \sqrt{2 \log \paren*{ {\abs{\widehat \cS_k}^2}/{\delta_k} }} = 2^{-k}$, we have 
\begin{align}
    \delta_k & = \abs{\widehat \cS_k}^2 \exp \paren*{ - \frac{1}{ 2 \cdot 2^{2k} \, \max_{y \in \psi_{d_k}(\widehat \cS_k)} \norm{y}_{A_k^{-1}}^2 } } \nonumber \\
    & \leq \abs{\widehat \cS_k}^2 \exp \paren*{ - \frac{T^\prime}{ 2 \cdot 2^{2k} \, (1 + \zeta) \, \iota(\cY(\psi_{d_k} (\widehat \cS_k))) } } \label{ms_bai:eq:sub_fixed_budget_rounding} \\
    & \leq \abs{ \cZ }^2 \exp \paren*{ - \frac{ T}{ 1024 \, n \, \rho_{d_\star}^\star } } \label{ms_bai:eq:sub_fixed_budget_error_bound},
\end{align}
where \cref{ms_bai:eq:sub_fixed_budget_rounding} comes from the guarantee of the rounding procedure \cref{ms_bai:eq:rounding}; and \cref{ms_bai:eq:sub_fixed_budget_error_bound} comes from combining the following facts: (1) $2^{2k} \, \iota(\cY(\psi_{d_k} (\widehat \cS_k))) \leq B \leq 128 \rho_{d_\star}^\star$; (2) $T^\prime \geq T/n - 1 \geq T/2n$ (note that $T/n \geq r_{d_\star}(\zeta) + 1 \implies T/n \geq 2$ since $r_{d_\star}(\zeta) \geq 1$); (3) $\widehat \cS_k \subseteq \cZ$ and (4) consider some $\zeta \leq 1$ ($\zeta$ only affects constant terms).

\textbf{Step 1.3: Correctness.} We prove $z_\star \in \widehat \cS_{k+1} \subseteq \cS_{k+1}$ under the good event analyzed in \cref{ms_bai:eq:subroutine_fixed_budget_event}.

\textbf{Step 1.3.1: $z_\star \in \widehat \cS_{k+1}$.} For any $\widehat z \in \widehat \cS_k$ such that $\widehat z \neq z_\star$, we have 
\begin{align*}
    \ang{ \psi_{d_k}(\widehat z) - \psi_{d_k}(z_\star) , \widehat \theta_k } & \leq \ang{ \psi_{d_k}(\widehat z) - \psi_{d_k}(z_\star) , \psi_{d_k} (\theta_\star) } + 2^{-k} \\
    & = h(\widehat z ) - h(z_\star) + 2^{-k} \\
    & < 2^{-k}.
\end{align*}
As a result, $z_\star$ remains in $\widehat \cS_{k+1}$ according to the elimination criteria.

\textbf{Step 1.3.2: $\widehat \cS_{k+1} \subseteq \cS_{k+1}$.} Consider any $z \in \widehat \cS_k \cap \cS_{k+1}^c$, we know that $\Delta_z \geq 2\cdot 2^{-k}$ by definition. Since $z_\star \in \widehat \cS_k$, we then have 
\begin{align}
    \ang{ \psi_{d_k}(z_\star) - \psi_{d_k}(z) , \widehat \theta_k } & \geq \ang{ \psi_{d_k}(z_\star) - \psi_{d_k}(z) , \psi_{d_k} (\theta_\star) } - 2^{-k} \nonumber \\
    & = h(z_\star) - h(z) - 2^{-k} \nonumber \\ 
    & \geq 2 \cdot 2^{-k} - 2^{-k} \nonumber \\
    & = 2^{-k}.
\end{align}
As a result, we have $z \notin \widehat \cS_{k+1}$ and $\widehat \cS_{k+1} \subseteq \cS_{k+1}$.

To summarize, we prove the induction at iteration $k+1$, i.e.,
\begin{align*}
    \P \paren*{ \cE_{k+1} \mid \cap_{i< k+1} \cE_{i} } & \geq 1 - \delta_k. \nonumber
\end{align*}

\textbf{Step 2: The error probability.} Let $\cE = \cap_{i=1}^{n+1} \cE_{i}$ denote the good event, we then have 
\begin{align}
    \P \paren*{ \cE } & = \prod_{k=1}^{n+1} \P \paren*{ \cE_k \mid \cE_{k-1} \cap \dots \cap \cE_1 } \nonumber \\
    & = \prod_{k=1}^{n+1} \paren*{1 - \delta_k} \nonumber \\
    & \geq 1 - \sum_{i=1}^{n+1} \delta_k \label{ms_bai:eq:subroutine_fixed_budget_induction} \\
    & \geq 1- n \abs*{ \cZ }^2 \exp \paren*{ - \frac{ T}{ 640 \, n \, \rho_{d_\star}^\star } } \nonumber,
\end{align}
where \cref{ms_bai:eq:subroutine_fixed_budget_induction} can be proved using a simple induction. 
\end{proof}

\subsubsection{Proof of \cref{ms_bai:thm:doubling_fixed_budget}}

\thmDoublingFixedBudget*

\begin{proof}
The proof is decomposed into three steps: (1) locate a good subroutine in the pre-selection step; (2) bound error probability in the validation step; and (3) analyze the total error probability. Some preliminaries are analyzed as follows.

We note that both pre-selection and validation steps use budget less than $T$: in the pre-selection phase, each outer loop indexed by $i$ uses budget less than $T/p$ and there are $p$ such outer loops; it's also clear that the validation steps uses at most $T$ budget. We notice that $p \leq \log_2 T$ since $p \cdot 2^p \leq T$; and $q_i \leq \log_2 T$ since $q_i \cdot 2^{q_i} \leq T/p B_i \leq T$. As a result, at most $(\log_2 T)^2$ subroutines are invoked in \cref{ms_bai:alg:doubling_fixed_budget}, and each subroutine is invoked with budget $T^{\prime \prime} \geq T/(\log_2 T)^2$.

\textbf{Step 1: The good subroutines.} Consider
\begin{align*}
    i_\star \ldef \ceil{ \log_2 \paren{ 64 \rho_{d_\star}^\star } } \quad \text{and} \quad 
    j_\star \ldef \ceil{ \log_2 \paren{ \log_2 \paren{ 2/\Delta_{\min} } } }.
\end{align*}
One can easily see that $64 \rho_{d_\star}^\star \leq B_{i_\star} \leq 128 \rho_{d_\star}^\star$ and $n_{j_\star} \geq \log_2 (2/\Delta_{\min})$. Thus, once a subroutine is invoked with $(i_\star, j_\star)$ and $T^{\prime \prime}/ n_{j_\star} \geq r_{d_\star}(\zeta) + 1$, \cref{ms_bai:lm:subroutine_fixed_budget} guarantees to output the optimal arm with error probability at most 
\begin{align}
     \log_2 (4/\Delta_{\min}) \abs*{ \cZ }^2 \exp \paren*{ - \frac{ T}{1024 \, \log_2 (4/\Delta_{\min}) \, \rho_{d_\star}^\star } } \label{ms_bai:eq:sub_fixed_budget_good_sub_error_prob}.
\end{align}
We next show that for sufficiently large $T$, one can invoke the subroutine with $(i_\star, j_\star)$ and $T^{\prime \prime}/ n_{j_\star} \geq r_{d_\star}(\zeta) + 1$. 

We clearly have $p \geq i_\star$ as long as $T \geq \log_2(128 \rho_{d_\star}^\star) \, 128 \rho_{d_\star}^\star$. Focusing on the outer loop with index $i_\star$, we have $q_{i_\star} \geq j_\star$ as long as 
\begin{align*}
    \log_2 \paren{2 \log_2 (2 /\Delta_{\min})} \cdot \paren{2 \log_2 (2/\Delta_{\min})} \leq T^\prime / B_{i_\star},
\end{align*}
Since $T^\prime/B_{i_\star} \geq T/ (128 \rho_{d_\star}^\star \log_2 T )$, we have $q_{i_\star} \geq j_\star$ as long as $T$ is such that
\begin{align}
    T \geq 256 \, \log_2 \paren{2 \log_2 (2 /\Delta_{\min})} \cdot \log_2 (2/\Delta_{\min}) \cdot \rho_{d_\star}^\star \cdot \log_2 T. \label{ms_bai:eq:budget_requirement_1}
\end{align}
Since $T^{\prime \prime} \geq T/(\log_2 T)^2$, we have $T^{\prime \prime}/ n_{j_\star} \geq r_{d_\star}(\zeta) + 1$ as long as $T$ is such that
\begin{align}
    T \geq (r_{d_\star}(\zeta) + 1) \cdot \log_2(4/\Delta_{\min}) \cdot (\log_2 T)^2. \label{ms_bai:eq:budget_requirement_2}
\end{align}
According to \cref{ms_bai:lm:relation_log}, \cref{ms_bai:eq:budget_requirement_1} and \cref{ms_bai:eq:budget_requirement_2} can be satisfied when 
\begin{align*}
    T = \widetilde \Omega \paren*{ \log_2(1/\Delta_{\min}) \max \curly*{\rho_{d_\star}^
    \star, r_{d_\star}(\zeta)} },
\end{align*}
where lower order terms with respect to $\log_2(1/\Delta_{\min})$, $\rho_{d_\star}^\star$ and $r_{d_\star}(\zeta)$ are hidden in the $\widetilde \Omega$ notation.

\textbf{Step 2: The validation step.} We have $\abs*{\cA} \leq (\log_2 T)^2$ since there are at most $(\log_2 T)^2$ subroutines and each subroutine outputs one arm. We view each $x \in \cA$ as individual arm and pull it $\floor*{T/\abs*{\cA}} \geq T/(\log_2 T)^2 -1 \geq T/2(\log_2 T)^2$ (as long as $T \geq 2 (\log_2 T)^2$) times. We use $\widehat h(x)$ to denote the empirical mean of $h(x)$. Applying Hoeffding's inequality with a union bound leads to the following concentration result
\begin{align*}
    \P \paren*{ \forall x \in \cA: \abs{\widehat h(x) - h(x)} \geq \Delta_{\min} /2  } \leq 2 (\log_2 T)^2 \exp \paren*{ - \frac{T}{8 (\log_2 T)^2/ \Delta_{\min}^2} }
\end{align*}
Thus, as long as $z_\star \in \cA$ is selected in $\cA$ from the pre-selection step, the validation step correctly output $z_\star$ with error probability at most 
\begin{align}
    2 (\log_2 T)^2 \exp \paren*{ - \frac{T}{8 (\log_2 T)^2/ \Delta_{\min}^2} }
    \label{ms_bai:eq:sub_fixed_budget_validation_error_prob}.
\end{align}

\textbf{Step 3: Total error probability.} Combining \cref{ms_bai:eq:sub_fixed_budget_good_sub_error_prob} with \cref{ms_bai:eq:sub_fixed_budget_validation_error_prob}, we know that
\begin{align*}
    \P \paren{\widehat z_\star \neq z_\star} \leq & \log_2 (4/\Delta_{\min}) \abs*{ \cZ }^2 \exp \paren*{ - \frac{ T}{ 1024 \, \log_2 (4/\Delta_{\min}) \, \rho_{d_\star}^\star } } \\
    &+ 2 (\log_2 T)^2 \exp \paren*{ - \frac{T}{8 (\log_2 T)^2/ \Delta_{\min}^2} }.
\end{align*}
Furthermore, if there exists universial constants such that $\max_{x \in \cX} \norm{\psi_{d_\star}(x)}^2 \leq c_1$ and $\min_{z \in \cZ} \norm{\psi_{d_\star}(z_\star) - \psi_{d_\star}(z)}^2 \geq c_2$, \cref{ms_bai:lm:psi_ub_lb} implies that $1/\Delta_{\min}^2 \leq c_1 \rho_{d_\star}^\star/ c_2$. We thus have 
\begin{align*}
    & \P  \paren{\widehat z_\star \neq z_\star} = \\
    &   O \paren*{ \max \curly*{ \log_2(1/\Delta_{\min}) \abs*{\cZ}^2, (\log_2 T)^2 } \cdot \exp \paren*{ - \frac{c_2 T}{\max \curly*{ \log_2(1/\Delta_{\min}), (\log_2 T)^2 } c_1 \rho_{d_\star}^\star} } }.
\end{align*}
\end{proof}

\subsection{Proofs and Supporting Results for \cref{ms_bai:sec:misspecification}}

\subsubsection{Proofs for Propositions}

Some of the propositions are borrowed from \citet{zhu2021pure}, we present detailed proofs here for completeness.

\propNonIncreasingMisspecification*

\begin{proof}
Consider any $1 \leq d < d^\prime \leq D$. Suppose
\begin{align*}
    \theta^d \in \argmin_{\theta \in \R^D} \max_{x \in \cX \cup \cZ} \abs*{h(x) - \ang*{\psi_d(\theta), \psi_d(x)}}.
\end{align*}
Since $\psi_d(\theta^d)$ only keeps the first $d$ component of $\theta^d$, we can choose $\theta^d$ such that it only has non-zero values on its first $d$ entries. As a result, we have $\ang*{\psi_d(\theta^d), \psi_d(x)} = \ang*{\psi_{d^\prime}(\theta^d), \psi_{d^\prime}(x)}$, which implies that $\widetilde \gamma(d^\prime) \leq \widetilde \gamma (d)$.
\end{proof}

\propRhoRelation*

\begin{proof}
To relate $\rho_d^\star(\epsilon)$ with $\widetilde \rho_d^\star(\epsilon)$, we only need to relate $\max \curly{ h(z_\star)- h(z), \epsilon}$ with $\max \curly{ \ang{\psi_d(z_\star) - \psi_d(z), \theta_\star^d}, \epsilon}$. From \cref{ms_bai:eq:mis_level} and the fact that $\epsilon \geq \widetilde \gamma (d)$, we know that 
\begin{align*}
	\ang{\psi_d(z_\star) - \psi_d(z), \theta_\star^d} & \leq h(z_\star)- h(z) + 2 \widetilde \gamma (d) \\
							  &\leq h(z_\star)- h(z) + 2 \epsilon \\
							  & \leq 3 \max \curly{ h(z_\star)- h(z), \epsilon},
\end{align*}
and thus 
\begin{align*}
    \max \curly*{ \ang{\psi_d(z_\star) - \psi_d(z), \theta_\star^d}, \epsilon} \leq 3 \max \curly{ h(z_\star)- h(z), \epsilon}.
\end{align*}
As a result, we have $\rho_d^\star(\epsilon) \leq 9 \widetilde \rho_d^\star (\epsilon)$.

When $\widetilde \gamma (d) < \Delta_{\min}/2$, we know that $z_\star$ is still the best arm in the perfect linear bandit model (without misspecification) $\widetilde h(x) = \ang*{\psi_d(x), \psi_d(\theta_\star^d)}$. Thus, $\widetilde \rho_d^\star(0)$ represents the complexity measure, in the corresponding linear model, for best arm identification.
\end{proof}

\begin{restatable}[\citet{zhu2021pure}]{proposition}{propGammaUpperBound}
\label{ms_bai:prop:gamma_upper_bound}
The following inequalities hold:
\begin{align*}
    \gamma(d) \leq \paren*{16 + 16 \sqrt{(1+\zeta) d}} \widetilde \gamma(d) = O(\sqrt{d} \, \widetilde \gamma (d)).
\end{align*}
\end{restatable}

\begin{proof}
We first notice that 
\begin{align}
    \iota \paren*{ \cY( \psi_d(\cS_k)) } & = \inf_{\lambda \in \simp_{\cX}} \sup_{y \in \cY( \psi_d(\cS_k))} \norm{y}_{A_{d}(\lambda)^{-1}}^2 \nonumber \\
    & \leq \inf_{\lambda \in \simp_{\cX}} \sup_{y \in \cY( \psi_d(\cX))} \norm{y}_{A_{d}(\lambda)^{-1}}^2 \nonumber \\
    & \leq \inf_{\lambda \in \simp_{\cX}} \sup_{ x \in \cX} 4 \norm{\psi_d(x)}_{A_{d}(\lambda)^{-1}}^2 \nonumber \\
    & = 4d \label{ms_bai:eq:prop_gamma_upper_bound_kw},
\end{align}
where \cref{ms_bai:eq:prop_gamma_upper_bound_kw} comes from Kiefer-Wolfowitz theorem \citep{kiefer1960equivalence}. We then have 
\begin{align*}
    \paren{2 + \sqrt{(1+\zeta) \iota \paren*{ \cY( \psi_d(\cS_k)) }} } \widetilde \gamma(d) \leq \paren{2 + \sqrt{(1+\zeta) 4 d} } \widetilde \gamma(d).
\end{align*}
As a result, we can always find a $n \in \N$ such that 
\begin{align*}
    2^{-n}/2 \leq 2 \, \paren{2 + \sqrt{(1+\zeta) 4 d} } \widetilde \gamma(d),
\end{align*}
and 
\begin{align*}
    \paren{2 + \sqrt{(1+\zeta) \iota \paren*{ \cY( \psi_d(\cS_k)) }} } \widetilde \gamma(d) \leq \paren{2 + \sqrt{(1+\zeta) 4 d} } \widetilde \gamma(d) \leq 2^{-k}/2, \forall k \leq n.
\end{align*}
This leads to the fact that
\begin{align*}
    \gamma(d) \leq 8 \, \paren{2 + \sqrt{(1+\zeta) 4 d} } \widetilde \gamma(d),
\end{align*}
which implies the desired result.
\end{proof}

\begin{restatable}{proposition}{propRoundNumber}
\label{ms_bai:prop:round_number}
If $\gamma(d) \leq \epsilon$, we have 
\begin{align*}
    \paren{2 + \sqrt{(1+\zeta) \iota \paren*{ \cY( \psi_d(\cS_k)) }} } \widetilde \gamma(d) \leq 2^{-k}/2, \forall k \leq \ceil*{\log_2(2/\epsilon)}.
\end{align*}
\end{restatable}

\begin{proof}
Suppose $\gamma(d) = 2 \cdot 2^{-\widetilde n}$ for a $\widetilde n \in \N$. Since $\gamma(d) \leq \epsilon$, we have $\widetilde n \geq \log_2(2/\epsilon)$. Since $\widetilde n \in \N$, we know that $\widetilde n \geq \ceil*{\log_2(2/\epsilon)}$. The desired result follows from the definition of $\gamma(d)$.
\end{proof}

\subsubsection{Omitted Details for the Fixed Confidence Setting with Misspecification}
\label{ms_bai:app:alg_misspecification}

\paragraph{Omitted Algorithms}

\noindent
\begin{algorithm}[H]
	\caption{\gemsm Gap Elimination with Model Selection with Misspecification (Fixed Confidence)}
	\label{ms_bai:alg:subroutine_fixed_confidence_mis_gen} 
	\renewcommand{\algorithmicrequire}{\textbf{Input:}}
	\renewcommand{\algorithmicensure}{\textbf{Output:}}
	\begin{algorithmic}[1]
		\REQUIRE  Number of iterations $n$, budget for dimension selection $B$ and confidence parameter $\delta$.
		\STATE Set $\widehat \cS_1 = \cZ$.
		\FOR {$k = 1, 2, \dots, n$}
		\STATE Set $\delta_k = \delta/k^2$.
		\STATE Define function $g_k(d) \ldef \max \curly{ 2^{2k} \, \iota_{k,d}, r_d(\zeta) }$, where $\iota_{k,d} \ldef \iota(\cY(\psi_d(\widehat \cS_k)))$. 
		\STATE Get $d_k = \OPT(B, D, g_k(\cdot))$, where $d_k \leq D$ is largest dimension such that $g_k(d_k) \leq B$ (see \cref{ms_bai:eq:opt_d_selection} for the detailed optimization problem). Set $\lambda_k$ be the optimal design of the optimization problem
		$\inf_{\lambda \in \simp_{\cX}} \sup_{z, z^\prime \in \widehat \cS_k} \norm*{\psi_{d_k}(z)-\psi_{d_k}(z^\prime)}^2_{A_{d_k}(\lambda)^{-1}}$;
		set   $N_k = \ceil{g(d_k) 8(1 + \zeta)  \log(\abs{\widehat \cS_k}^2/\delta_k)}$.
		\STATE Get allocation $\curly*{x_1, \ldots, x_{N_k} } = \round (\lambda_k,N_k, d_k, \zeta)$.
		\STATE Pull arms $\curly*{x_1, \ldots, x_{N_k}} $ and receive rewards $\curly*{r_1, \ldots, r_{N_k}}$.
        \STATE Set $\widehat{\theta}_k = A_k^{-1} b_k \in \R^{d_k}$ where $A_k = \sum_{i=1}^{N_k} \psi_{d_k}(x_i) \psi_{d_k}(x_i)^\top$, and $b_k =  \sum_{i=1}^{N_k} \psi_{d_k}(x_i) b_i$.
        \STATE Set $\widehat \cS_{k+1} = \widehat \cS_k \setminus \{z \in \widehat \cS_k : \exists z^\prime \text{ s.t. } \ang{\widehat{\theta}_k, \psi_{d_k}(z^\prime) - \psi_{d_k}(z) } \geq 2^{-k} \}$.
		\ENDFOR 
		\ENSURE Any $\widehat z_\star \in \widehat \cS_{n+1}$ (or the whole set $\widehat \cS_{n+1}$ when aiming at identifying the optimal arm).
	\end{algorithmic}
\end{algorithm}

\begin{algorithm}[H]
	\caption{Adaptive Strategy for Model Selection with misspecification (Fixed Confidence)}
	\label{ms_bai:alg:doubling_fixed_confidence_mis_gen} 
	\renewcommand{\algorithmicrequire}{\textbf{Input:}}
	\renewcommand{\algorithmicensure}{\textbf{Output:}}
	\newcommand{\algorithmicbreak}{\textbf{break}}
    \newcommand{\BREAK}{\STATE \algorithmicbreak}
	\begin{algorithmic}[1]
		\REQUIRE Confidence parameter $\delta$.
		\STATE Randomly select a $\widehat z_\star \in \cX$ as the recommendation for the $\epsilon$-optimal arm.
		\FOR {$\ell = 1, 2, \dots$}
		\STATE Set $\gamma_\ell = 2^\ell$ and $\delta_\ell = \delta/(4\ell^3)$. Initialize an empty pre-selection set $\cA_{\ell} = \curly{}$.
		    \FOR {$i = 1, 2, \dots, \ell$}
		    \STATE Set $n_i = 2^i$, $B_i = 2^{\ell -i}$ and get $\widehat z_{\star}^i = \text{\gemsm}(n_i, B_i, \delta_\ell)$. Insert $\widehat z_{\star}^i$ into $\cA_{\ell}$.
		    \ENDFOR
		    \STATE \textbf{Validation.} Pull each arm in $\cA$ exactly $\ceil{8 \log(2/\delta_{\ell})/\epsilon^2}$ times. Update $\widehat z_\star$ as the arm with the highest empirical mean (break ties arbitrarily).
		\ENDFOR 
	\end{algorithmic}
\end{algorithm}

\paragraph{\cref{ms_bai:lm:subroutine_fixed_confidence_mis_gen} and Its Proof}
\paranewline

\noindent
We introduce function $f:\N_+ \rightarrow \R_+$ as follows, which is also used in \cref{ms_bai:app:fix_budget_mis}.
\begin{align*}
    f(k) \ldef \begin{cases} 4 \cdot 2^{-k} & \text{ if } k \leq \ceil{\log_2 (2/\epsilon)} + 1, \\
    4 \cdot \epsilon^{- \ceil{\log_2 (4/\epsilon)}} & \text{ if } k > \ceil{\log_2 (2/\epsilon)} + 1.\\
    \end{cases}
\end{align*}
$f(k)$ is used to quantify the optimality of the identified arm, and one can clearly see that $f(k)$ is non-increasing in $k$.

\begin{restatable}{lemma}{lmSubroutineFixedConfidenceMisGen}
\label{ms_bai:lm:subroutine_fixed_confidence_mis_gen}
Suppose $B \geq \max \curly{64 \rho_{d_\star(\epsilon)}^\star( \epsilon), r_{d_\star(\epsilon)}(\zeta)}$. With probability at least $1-\delta$, \cref{ms_bai:alg:subroutine_fixed_confidence_mis_gen} outputs an arm $\widehat z_\star$ such that $\Delta_{\widehat z_\star} < f(n+1)$. Furthermore, an $\epsilon$-optimal arm is output as long as $n \geq {\log_2 (2/\epsilon)}$.
\end{restatable}

\begin{proof}
The logic of this proof is similar to the proof of \cref{ms_bai:lm:subroutine_fixed_confidence}. We additionally deal with misspecification in the proof. For fixed $\epsilon$, we use the notation $d_\star = d_\star(\epsilon)$ throughout the proof.

We consider event 
\begin{align*}
    \cE_k = \curly{z_\star \in \widehat \cS_k \subseteq \cS_k},
\end{align*}
and prove through induction that, for $k \leq \ceil{\log_2 (2/\epsilon)}$,
\begin{align*}
    \P \paren*{ \cE_{k+1} \mid \cap_{i \leq k} \cE_i } \geq 1 - \delta_{k},
\end{align*}
where $\delta_0 \ldef 0$. Recall that $\cS_k = \curly{z \in \cZ: \Delta_z < 4 \cdot 2^{-k}}$ (with $\cS_1 = \cZ$). For $n \geq k+1$, we have $\widehat \cS_n \subseteq \widehat \cS_{k+1}$ due to the nature of the elimination-styled algorithm, which guarantees outputting an arm such that $\Delta_{z} < f(n+1)$.

\textbf{Step 1: The induction.} We have $\curly{ z_\star \in \widehat \cS_1 \subseteq \cS_1 }$ since $\widehat \cS_1 = \cS_1 = \cZ$ by definition for the base case (recall we assume that $\max_{z \in \cZ} \Delta_z \leq 2$). We now assume that $\cap_{i < k+1}\cE_{i}$ holds true and we prove for iteration $k+1$.

\textbf{Step 1.1: $d_k \geq d_\star$.} Since $\widehat \cS_{k} \subseteq \cS_k$, we have 
\begin{align}
    g_k(d_\star) & = \max \curly{ {2^{2k} \iota(\cY(\psi_{d_\star}(\widehat \cS_k)))}, r_{d_\star}(\zeta) } \nonumber \\
    & \leq \max \curly{ {2^{2k} \iota(\cY(\psi_{d_\star}(\cS_k)))}, r_{d_\star}(\zeta) } \nonumber\\
    & \leq \max \curly{64 \rho_{d_\star}^\star (\epsilon), r_{d_\star}(\zeta)} \label{ms_bai:eq:subroutine_fixed_confidence_rho_stratified_mis_gen}\\
    & \leq B \label{ms_bai:eq:subroutine_fixed_confidence_B_mis_gen},
\end{align}
where \cref{ms_bai:eq:subroutine_fixed_confidence_rho_stratified_mis_gen} comes from \cref{ms_bai:lm:rho_stratified_eps} and \cref{ms_bai:eq:subroutine_fixed_confidence_B_mis_gen} comes from the assumption. As a result, we know that $d_k \geq d_\star$ since $d_k$ is selected as the largest integer such that $g_k(d_k) \leq B$. 

\textbf{Step 1.2: Concentration.} Let $\curly*{x_1, \ldots, x_{N_k}} $ be the arms pulled at iteration $k$ and $\curly*{r_1, \ldots, r_{N_k}}$ be the corresponding rewards. Let $\widehat{\theta}_k = A_k^{-1} b_k \in \R^{d_k}$ where $A_k = \sum_{i=1}^{N_k} \psi_{d_k}(x_i) \psi_{d_k}(x_i)^\top$, and $b_k =  \sum_{i=1}^{N_k} \psi_{d_k}(x_i) b_i$. Based on the definition of $\theta_\star^d \in \R^D$ and $\eta_d(\cdot)$, we can write $r_i = h(x_i) + \xi_i = \ang*{\psi_{d_k}(\theta_\star^{d_k}), \psi_{d_k}(x_i)} + \eta_{d_k}(x_i) + \xi_i$, where $\xi_i$ is i.i.d. generated zero-mean Gaussian noise with variance $1$; we also have $\abs*{\eta_{d_k}(x_i)} \leq \widetilde \gamma (d_k)$ by definition of $\widetilde \gamma (\cdot)$. For any $y \in \cY(\psi_{d_k}(\widehat \cS_k))$, we have
\begin{align}
    &\abs*{\ang{y, \widehat \theta_k - \psi_{d_k}(\theta_\star^{d_k})}} \nonumber \\
    & = \abs*{y^\top A_k^{-1} \sum_{i=1}^{N_k} \psi_{d_k}(x_i) r_i - y^\top \psi_{d_k}(\theta_\star^{d_k})} \nonumber \\
    & = \abs*{y^\top A_k^{-1} \sum_{i=1}^{N_k} \psi_{d_k}(x_i) \paren{ \psi_{d_k}(x_i) ^\top \psi_{d_k}(\theta_\star^{d_k}) + \eta_{d_k}(x_i) + \xi_i} - y^\top \psi_{d_k}(\theta_\star)} \nonumber \\
    & = \abs*{ y^\top A_k^{-1} \sum_{i=1}^{N_k} \psi_{d_k}(x_i) \paren{ \eta_{d_k}(x_i) + \xi_i}} \nonumber \\
    & \leq \abs*{ y^\top A_k^{-1} \sum_{i=1}^{N_k} \psi_{d_k}(x_i) \eta_{d_k}(x_i) } + \abs*{ y^\top A_k^{-1} \sum_{i=1}^{N_k} \psi_{d_k}(x_i) \xi_i }. \label{ms_bai:eq:mis_gen_sub_fixed_conf_two_terms}
\end{align}
We next bound the two terms in \cref{ms_bai:eq:mis_gen_sub_fixed_conf_two_terms} separately. For the first term, we have
\begin{align}
    \abs*{ y^\top A_k^{-1} \sum_{i=1}^{N_k} \psi_{d_k}(x_i) \eta_{d_k}(x_i) } & \leq \widetilde \gamma(d_k) \sum_{i=1}^{N_k} \abs*{ y^\top A_k^{-1} \psi_{d_k}(x_i) } \nonumber \\
    & = \widetilde \gamma (d_k) \sum_{i=1}^{N_k} \sqrt{ \paren*{ y^\top A_k^{-1} \psi_{d_k}(x_i) }^2 } \nonumber \\
    & \leq \widetilde \gamma(d_k) \sqrt{N_k \sum_{i=1}^{N_k} \paren*{ y^\top A_k^{-1} \psi_{d_k}(x_i) }^2 } \label{ms_bai:eq:mis_gen_sub_fixed_conf_jensen} \\
    & = \widetilde \gamma(d_k) \sqrt{N_k \sum_{i=1}^{N_k}  y^\top A_k^{-1} \psi_{d_k}(x_i)  \psi_{d_k}(x_i)^\top A_k^{-1} y } \nonumber \\
    & = \widetilde \gamma(d_k) \sqrt{N_k \norm{y}_{A_k^{-1}}^2 } \nonumber \\
    & \leq \widetilde \gamma(d_k) \sqrt{(1+\zeta) \iota(\cY(\psi_{d_k}(\widehat \cS_k))) } \label{ms_bai:eq:mis_gen_sub_fixed_conf_round} \\
    & \leq \widetilde \gamma(d_k) \sqrt{(1+\zeta) \iota(\cY(\psi_{d_k}(\cS_k))) } \label{ms_bai:eq:mis_gen_sub_fixed_conf_Sk}
  \end{align}
where \cref{ms_bai:eq:mis_gen_sub_fixed_conf_jensen} comes from Jensen's inequality; \cref{ms_bai:eq:mis_gen_sub_fixed_conf_round} comes from the guarantee of rounding in \cref{ms_bai:eq:rounding}; and \cref{ms_bai:eq:mis_gen_sub_fixed_conf_Sk} comes from the fact that $\widehat \cS_k \subseteq \cS_k$.

For the second term in \cref{ms_bai:eq:mis_gen_sub_fixed_conf_two_terms}, since $\xi_i$s are independent 1-sub-Gaussian random variables, we know that the random variable $ y^\top A_k^{-1} \sum_{i=1}^{N_k} \psi_{d_k}(x_i) \xi_i $ has variance proxy $\sqrt{\sum_{i=1}^{N_k} \paren*{y^\top A_k^{-1} \sum_{i=1}^{N_k} \psi_{d_k}(x_i)}^2 } = \norm{y}_{A_k^{-1}}$. Combining the standard Hoeffding's inequality with a union bound leads to
\begin{align}
    \P \paren*{ \forall y \in \cY(\psi_{d_k}(\widehat \cS_k)) , \abs*{ y^\top A_k^{-1} \sum_{i=1}^{N_k} \psi_{d_k}(x_i) \xi_i } \leq \norm{y}_{A_k^{-1}} \sqrt{2 \log \paren*{ {\abs{\widehat \cS_k}^2}/{\delta_k} }}  } \geq 1-\delta_k, \label{ms_bai:eq:mis_gen_sub_fixed_conf_second}
\end{align}
where we use the fact that $\abs{ \cY(\psi_{d_k}(\widehat \cS_k)) } \leq \abs{\widehat \cS_k}^2/2$ in the union bound. 

Putting \cref{ms_bai:eq:mis_gen_sub_fixed_conf_round} and \cref{ms_bai:eq:mis_gen_sub_fixed_conf_second} together, we have 
\begin{align}
    \P \paren{ \forall y \in \cY(\psi_{d_k}(\widehat \cS_k)) , \abs{\ang{y, \widehat \theta_k - \psi_{d_k}(\theta_\star^{d_k})}} \leq \widetilde \gamma(d_k) \iota_k +  \omega_k(y)   } \geq 1-\delta_k, \label{ms_bai:eq:mis_gen_sub_fixed_conf_concentration}
\end{align}
where $\iota_k \ldef  \sqrt{(1+\zeta) \iota(\cY(\psi_{d_k}(\cS_k))) }$ and $ \omega_k(y) \ldef  \norm{y}_{A_k^{-1}} \sqrt{2 \log \paren*{ {\abs{\widehat \cS_k}^2}/{\delta_k} }}$.

\textbf{Step 1.3: Correctness.} We prove $z_\star \in \widehat \cS_{k+1} \subseteq \cS_{k+1}$ under the good event analyzed in \cref{ms_bai:eq:mis_gen_sub_fixed_conf_concentration}. 

\textbf{Step 1.3.1: $z_\star \in \widehat \cS_{k+1}$.} For any $\widehat z \in \widehat \cS_k$ such that $\widehat z \neq z_\star$, we have 
\begin{align}
	&\ang{ \psi_{d_k}(\widehat z)  - \psi_{d_k}(z_\star) , \widehat \theta_k } \nonumber \\
    & \leq \ang{ \psi_{d_k}(\widehat z) - \psi_{d_k}(z_\star) , \psi_{d_k} (\theta_\star^{d_k}) } + \gamma(d_k) \iota_k + \omega_k(\psi_{d_k}(\widehat z) - \psi_{d_k}(z_\star)) \nonumber \\
    & = h(\widehat z ) -\eta_{d_k}(\widehat z) - h(z_\star) + \eta_{d_k}(z_\star) + \gamma(d_k) \iota_k + \omega_k(\psi_{d_k}(\widehat z) - \psi_{d_k}(z_\star)) \nonumber \\
    & < (2 + \iota_k) \widetilde \gamma(d_k)+ \omega_k(\psi_{d_k}(\widehat z) - \psi_{d_k}(z_\star)) \nonumber \\
    & \leq 2^{-k}/2 + 2^{-k}/2 \label{ms_bai:eq:mis_gen_sub_fixed_conf_optimal_arm} \\
    & = 2^{-k}, \nonumber
\end{align}
where \cref{ms_bai:eq:mis_gen_sub_fixed_conf_optimal_arm} comes from \cref{ms_bai:prop:round_number} combined with the fact that $d_k \geq d_\star$ (as shown in Step 1.1), and the selection of $N_k$ together with the guarantees in the rounding procedure \cref{ms_bai:eq:rounding}.

\textbf{Step 1.3.2: $\widehat \cS_{k+1} \subseteq \cS_{k+1}$.} Consider any $z \in \widehat \cS_k \cap \cS_{k+1}^c$, we know that $\Delta_z \geq 2\cdot 2^{-k}$ by definition. Since $z_\star \in \widehat \cS_k$, we then have 
\begin{align}
	&\ang{ \psi_{d_k}(z_\star)  - \psi_{d_k}(z) , \widehat \theta_k } \nonumber \\
    & \geq \ang{ \psi_{d_k}(\widehat z) - \psi_{d_k}(z_\star) , \psi_{d_k} (\theta_\star^{d_k}) } - \gamma(d_k) \iota_k - \omega_k(\psi_{d_k}(\widehat z) - \psi_{d_k}(z_\star)) \nonumber \\
    & = h(z_\star) - \eta_{d_k}(z_\star) - h(z) + \eta_{d_k}(z) - \gamma(d_k) \iota_k - \omega_k(\psi_{d_k}(\widehat z) - \psi_{d_k}(z_\star)) \nonumber \\ 
    & \geq 2 \cdot 2^{-k} - (2 + \iota_k)\widetilde \gamma(d_k)  - \omega_k(\psi_{d_k}(\widehat z) - \psi_{d_k}(z_\star)) \nonumber \\
    & \geq 2 \cdot 2^{-k} - 2^{-k}/2  - 2^{-k}/2 \label{ms_bai:eq:mis_gen_sub_fixed_conf_bad_arms} \\
    & = 2^{-k} \nonumber,
\end{align}
where \cref{ms_bai:eq:mis_gen_sub_fixed_conf_bad_arms} comes from a similar reasoning as appearing in \cref{ms_bai:eq:mis_gen_sub_fixed_conf_optimal_arm}. As a result, we have $z \notin \widehat \cS_{k+1}$ and $\widehat \cS_{k+1} \subseteq \cS_{k+1}$.

To summarize, we prove the induction at iteration $k+1$, i.e.,
\begin{align*}
    \P \paren*{ \cE_{k+1} \mid \cap_{i< k+1} \cE_{i} } \geq 1 - \delta_k.
\end{align*}

\textbf{Step 2: The error probability.} The analysis on the error probability is the same as in the Step 2 in the proof of \cref{ms_bai:lm:subroutine_fixed_confidence}. Let $\cE = \cap_{i=1}^{n+1} \cE_{i}$ denote the good event, we then have 
\begin{align}
    \P \paren*{ \cE } &  \geq 1- \delta \nonumber.
\end{align}
\end{proof}

\paragraph{Proof of \cref{ms_bai:thm:doubling_fixed_confidence_mis_gen}}

\noindent
\thmDoublingFixedConfidenceMisGen*

\begin{proof}
The proof is decomposed into four steps: (1) locating good subroutines; (2) guarantees for the validation step; (3) bounding error probability and (4) bounding unverifiable sample complexity. For fixed $\epsilon$, we use shorthand $d_\star = d_\star(\epsilon)$ throughout the proof.

\textbf{Step 1: The good subroutines.} Consider $B_\star = \max \curly{64 \rho^\star_{d_\star}, r_{d_\star}(\zeta)}$ and $n_\star = \ceil{ \log_2 (2/\epsilon) }$. For any subroutines invoked with $B_i \geq B_\star$ and $n_i \geq n_\star$, we know that, from \cref{ms_bai:lm:subroutine_fixed_confidence_mis_gen}, the output set of arms are those with sub-optimality gap $< \epsilon$. Let $i_\star = \ceil{\log_2 (B_\star)}$, $j_\star = \ceil{ \log_2 (n_\star) }$ and $\ell_\star = i_\star + j_\star$. We know that in outer loops $\ell \geq \ell_\star$, there must exists at least one subroutine invoked with $B_i = 2^{i_\star} \geq B_\star$ and $n_i = 2^{j_\star} \geq n_\star$. As a result, $\cA_{\ell}$ contains at least one $\epsilon$-optimal arm for $\ell \geq \ell_\star$. 

\textbf{Step 2: The validation step.} For any $x \in \cA_\ell$, we use $\widehat h(x)$ to denote its sample mean after  $\ceil*{8 \log(2/\delta_{\ell})/\epsilon^2}$ samples. With $1$-sub-Gaussian noise, a standard Hoeffding's inequality shows that and a union bound gives
\begin{align}
    \P \paren*{\forall x \in \cA_\ell: \abs{\widehat h(x) - h(x)} \geq \epsilon/2} \leq \ell \delta_\ell. \label{ms_bai:eq:mis_gen_validation}
\end{align}
As a result, a $2\epsilon$-optimal arm will be selected with probability at least $1 - \ell \delta_\ell$, as long as at least one $\epsilon$-optimal arm is contained in $\cA_\ell$.

\textbf{Step 3: Error probability.} We consider the good event where all subroutines invoked in \cref{ms_bai:alg:doubling_fixed_confidence} with $B_i \geq B_\star$ and (any) $n_i$ correctly output a set of arms with sub-optimality gap $<f(n_i+1)$, as shown in \cref{ms_bai:lm:subroutine_fixed_confidence_mis_gen}, together with the confidence bound described in \cref{ms_bai:eq:mis_gen_validation} in the validation step. This good event clearly happens with probability at least $1 - \sum_{\ell = 1}^\infty \sum_{i=1}^\ell 2 \delta_\ell = 1- \sum_{\ell = 1}^\infty \delta/(2 \ell^2) > 1 - \delta$, after applying a union bound argument. We upper bound the unverifiable sample complexity under this good event in the following.

\textbf{Step 4: Unverifiable sample complexity.} 
For any subroutine invoked within outer loop $\ell \leq \ell_\star$, we know, from \cref{ms_bai:alg:subroutine_fixed_confidence_mis_gen}, that its sample complexity is upper bounded by (note that $\abs*{\cZ}^2 \geq 4$ trivially holds true)
\begin{align*}
    N_\ell & \leq n_i \paren*{ B_i \cdot \paren*{10 \, \log(\abs{\cZ}^2/\delta_{\ell_\star})} + 1 } \\
    & \leq \gamma_\ell \, 11 \, \log \paren*{ 4 \abs{\cZ}^2 \ell_\star^3/\delta }.
\end{align*}
The validation step within any outer loop $\ell \leq \ell_\star$ takes at most $\ell \cdot \ceil{8 \log(2/\delta_{\ell})/\epsilon^2} \leq 9 \log(8 \ell_\star^3/\delta) \ell_\star /\epsilon^2$ samples. Thus, the total sample complexity up to the end of outer loops $\ell \leq \ell_\star$ is upper bounded by 
\begin{align*}
    N & \leq \sum_{\ell = 1}^{\ell_\star} \paren*{ \ell N_\ell + \ell \cdot \ceil*{8 \log(2/\delta_{\ell})/\epsilon^2}}\\
    & \leq 11\, \log \paren*{ 4 \abs{\cZ}^2 \ell_\star^3/\delta } \sum_{\ell = 1}^{\ell_\star} \ell 2^\ell +  9 \log\paren*{8 \ell_\star^3/\delta} \ell_\star^2 /\epsilon^2\\
    & \leq 22 \, \log \paren*{ 4 \abs{\cZ}^2 \ell_\star^3/\delta } \ell_\star 2^{\ell_\star} + 9 \log \paren*{8 \ell_\star^3/\delta} \ell_\star^2 /\epsilon^2.
\end{align*}

By definition of $\ell_\star$, we have
\begin{align*}
    \ell_\star \leq \log_2 \paren*{ 4  \log_2(4/\epsilon)  \max \curly{64 \rho^\star_{d_\star}, r_{d_\star}(\zeta)} },
\end{align*}
and
\begin{align*}
    2^{\ell_\star} & = 2^{(i_\star + j_\star)} \\
    & \leq 4  \paren*{ \log_2(2/\epsilon)+1}  \max \curly{64 \rho^\star_{d_\star}, r_{d_\star}(\zeta)}, \\
    & =  4   \log_2(4/\epsilon)  \max \curly{64 \rho^\star_{d_\star}, r_{d_\star}(\zeta)}. 
\end{align*}

Set $\tau_\star = \log_2(4/\epsilon) \max \curly*{\rho^\star_{d_\star}, r_{d_\star}(\zeta)}$. The unverifiable sample complexity is upper bounded by (we only consider the case when $\epsilon \leq 1$ in simplifying the bound: otherwise there is no need to prove anything since $\max_{x \in \cX} \Delta_x \leq 2$) 
\begin{align*}
    N & \leq 5632 \, \tau_\star \cdot \paren*{\log_2(\tau_\star) + 8} \cdot \log \paren*{ {4 \abs{\cZ}^2 \paren{\log_2(\tau_\star) + 8}^3}/{\delta} }\\
      & \quad + 9/\epsilon^2 \cdot \paren*{\log_2(\tau_\star) + 8}^2 \cdot \log \paren*{ {8 \paren{\log_2(\tau_\star) + 8}^3}/{\delta} } \\
    & = \widetilde O \paren*{ \log_2(1/\epsilon) \max \curly{\rho^\star_{d_\star}, r_{d_\star}(\zeta)} + 1/\epsilon^2 },
\end{align*}
where we hide logarithmic terms besides $\log(1/\epsilon)$ in the $\widetilde O$ notation.
\end{proof}

\subsubsection{Identifying the Optimal Arm under misspecification}
\label{ms_bai:app:BAI_misspecification}
When the goal is to identify the optimal arm under misspecification, i.e., by choosing $\epsilon = \Delta_{\min}$, one can apply \cref{ms_bai:alg:doubling_fixed_confidence} together with \cref{ms_bai:alg:subroutine_fixed_confidence_mis_gen} as the subroutine (thus removing the $1/\epsilon^2$ term in sample complexity). This combination works since, with appropriate choice of $B$, \cref{ms_bai:alg:subroutine_fixed_confidence_mis_gen} is guaranteed to output a subset of arms $\widehat \cS_{n+1}$ with optimality gap $<\Delta_{\min}$ when $n \geq \log_2(2/\Delta_{\min})$. This implies that $\widehat \cS = \curly*{z_\star}$ and thus the one can reuse the selection rule of \cref{ms_bai:alg:doubling_fixed_confidence} by recommending arms contained in the singleton set. Note that we can work with the general transductive linear bandit setting in this case, i.e., we don't require $\cZ \subseteq \cX$ anymore.

\subsubsection{Omitted Proofs for the Fixed Budget Setting with Misspecification}
\label{ms_bai:app:fix_budget_mis}

\paragraph{\cref{ms_bai:lm:subroutine_fixed_budget_mis} and Its Proof}

\noindent
\begin{restatable}{lemma}{lmSubroutineFixedBudgetMis}
\label{ms_bai:lm:subroutine_fixed_budget_mis}
Suppose $64 \rho_{d_\star (\epsilon)}^\star(\epsilon) \leq B \leq 128 \rho_{d_\star(\epsilon)}^\star (\epsilon) $ and $T/n \geq r_{d_\star(\epsilon)}(\zeta) + 1$. \cref{ms_bai:alg:subroutine_fixed_budget} outputs an arm $\widehat z_\star$ such that $\Delta_{\widehat z_\star} < f(n+1)$ with probability at least
\begin{align*}
    1- n \abs*{ \cZ }^2 \exp \paren*{ - \frac{ T}{ 2560 \, n \, \rho_{d_\star (\epsilon)}^\star (\epsilon)} }.
\end{align*}
Furthermore, an $\epsilon$-optimal arm is output as long as $n \geq {\log_2 (2/\epsilon)}$.
\end{restatable}

\begin{proof}
The proof is similar to the proof of \cref{ms_bai:lm:subroutine_fixed_budget}, with main differences in dealing with misspecification. We provide the proof here for completeness. We consider event 
\begin{align*}
    \cE_k = \curly{z_\star \in \widehat \cS_k \subseteq \cS_k},
\end{align*}
and prove through induction that, for $k \leq \ceil{\log_2 (2/\epsilon)}$,
\begin{align*}
    \P \paren*{ \cE_{k+1} \mid \cap_{i \leq k} \cE_i } \geq 1 - \delta_{k},
\end{align*}
where the value of $\curly{\delta_k}_{k=0}^{\ceil*{\log_2 (2/\epsilon)}}$ will be specified in the proof. For $n \geq k+1$, we have $\widehat \cS_n \subseteq \widehat \cS_{k+1}$ due to the nature of the elimination-styled algorithm, which guarantees outputting an arm such that $\Delta_{z} < f(n+1)$. We use the notation $d_\star = d_\star(\epsilon)$ throughout the rest of the proof.

\textbf{Step 1: The induction.} The base case $\curly{z_\star \in \widehat \cS_1 \subseteq \cS_1}$ holds with probability $1$ by construction (thus, we have $\delta_0 = 0$). Conditioned on events $\cap_{i=1}^k \cE_i$, we next analyze the event $\cE_{k+1}$. 

\textbf{Step 1.1: $d_k \geq d_\star$.} We first notice that $\widetilde D$ is selected as the largest integer such that $r_{\widetilde D}(\zeta) \leq T^\prime$. When $T/ n \geq r_{d_\star}(\zeta) + 1$, we have $\widetilde D \geq d_\star$ since $T^\prime \geq T/ n -1 \geq r_{d_\star}(\zeta)$. We remark here that for whatever $d_k \in [\widetilde D]$ selected, we always have $r_{d_\star}(\zeta) \leq r_{\widetilde D}(\zeta) \leq T^\prime$ and can thus safely apply the rounding procedure described in \cref{ms_bai:eq:rounding}.

Since $\widehat \cS_{k} \subseteq \cS_k$, we also have 
\begin{align}
    g_k(d_\star) & =  {2^{2k} \iota(\cY(\psi_{d_\star}(\widehat \cS_k)))} \nonumber \\
    & \leq {2^{2k} \iota(\cY(\psi_{d_\star}(\cS_k)))} \nonumber\\
    & \leq 64 \rho_{d_\star}^\star (\epsilon) \label{ms_bai:eq:subroutine_fixed_budget_rho_stratified_mis}\\
    & \leq B \label{ms_bai:eq:subroutine_fixed_budget_B_mis},
\end{align}
where \cref{ms_bai:eq:subroutine_fixed_budget_rho_stratified_mis} comes from \cref{ms_bai:lm:rho_stratified_eps} and \cref{ms_bai:eq:subroutine_fixed_budget_B_mis} comes from the assumption. As a result, we know that $d_{k} \geq d_\star$ since $d_{k} \in [\widetilde D]$ is selected as the largest integer such that $g_k(d_{k}) \leq B$. 

\textbf{Step 1.2: Concentration and error probability.} Let $\curly*{x_1, \ldots, x_{T^\prime}} $ be the arms pulled at iteration $k$ and $\curly*{r_1, \ldots, r_{T^\prime}}$ be the corresponding rewards. Let $\widehat{\theta}_k = A_k^{-1} b_k \in \R^{d_k}$ where $A_k = \sum_{i=1}^{T^\prime} \psi_{d_k}(x_i) \psi_{d_k}(x_i)^\top$, and $b_k =  \sum_{i=1}^{T^\prime} \psi_{d_k}(x_i) b_i$. Since $d_k \geq d_\star$ and the model is well-specified, we can write $r_i = \ang*{\theta_\star, x_i} + \xi_i = \ang*{\psi_{d_k}(\theta_\star), \psi_{d_k}(x_i)} + \xi_i$, where $\xi_i$ is i.i.d. generated zero-mean Gaussian noise with variance $1$. Similarly as analyzed in \cref{ms_bai:eq:mis_gen_sub_fixed_conf_concentration}, we have 
\begin{align}
    \P \paren*{ \forall y \in \cY(\psi_{d_k}(\widehat \cS_k)) , \abs{ \ang{ y, \widehat \theta_k - \psi_{d_k}(\theta_\star)  } } \leq  \widetilde \gamma(d_k) \iota_k +  \omega_k(y)   } \geq 1-\delta_k, \label{ms_bai:eq:subroutine_fixed_budget_event_mis}
\end{align}
where $\iota_k \ldef  \sqrt{(1+\zeta) \iota(\cY(\psi_{d_k}(\cS_k))) }$ and $ \omega_k(y) \ldef  \norm{y}_{A_k^{-1}} \sqrt{2 \log \paren*{ {\abs{\widehat \cS_k}^2}/{\delta_k} }}$.

By setting $\max_{y \in \psi_{d_k}(\widehat \cS_k)}\norm{y}_{A_k^{-1}} \sqrt{2 \log \paren*{ {\abs{\widehat \cS_k}^2}/{\delta_k} }} = 2^{-k}/2$, we have 
\begin{align}
    \delta_k & = \abs{\widehat \cS_k}^2 \exp \paren*{ - \frac{1}{ 8 \cdot 2^{2k} \, \max_{y \in \psi_{d_k}(\widehat \cS_k)} \norm{y}_{A_k^{-1}}^2 } } \nonumber \\
    & \leq \abs{\widehat \cS_k}^2 \exp \paren*{ - \frac{T^\prime}{ 8 \cdot 2^{2k} \, (1 + \zeta) \, \iota(\cY(\psi_{d_k} (\widehat \cS_k))) } } \label{ms_bai:eq:sub_fixed_budget_rounding_mis} \\
    & \leq \abs*{ \cZ }^2 \exp \paren*{ - \frac{ T}{ 4096 \, n \, \rho_{d_\star}^\star (\epsilon)} } \label{ms_bai:eq:sub_fixed_budget_error_bound_mis},
\end{align}
where \cref{ms_bai:eq:sub_fixed_budget_rounding_mis} comes from the guarantee of the rounding procedure \cref{ms_bai:eq:rounding}; and \cref{ms_bai:eq:sub_fixed_budget_error_bound_mis} comes from combining the following facts: (1) $2^{2k} \, \iota(\cY(\psi_{d_k} (\widehat \cS_k))) \leq B \leq 128 \rho_{d_\star}^\star (\epsilon)$; (2) $T^\prime \geq T/n - 1 \geq T/2n$ (note that $T/n \geq r_{d_\star}(\zeta) + 1 \implies T/n \geq 2$ since $r_{d_\star}(\zeta) \geq 1$);
(3) $\widehat \cS_k \subseteq \cZ$ and (4) consider some $\zeta \leq 1$ ($\zeta$ only affects constant terms).

\textbf{Step 1.3: Correctness.} We prove $z_\star \in \widehat \cS_{k+1} \subseteq \cS_{k+1}$ under the good event analyzed in \cref{ms_bai:eq:subroutine_fixed_budget_event_mis}.

\textbf{Step 1.3.1: $z_\star \in \widehat \cS_{k+1}$.} For any $\widehat z \in \widehat \cS_k$ such that $\widehat z \neq z_\star$, we have 
\begin{align}
    & \ang{ \psi_{d_k}(\widehat z) - \psi_{d_k}(z_\star) , \widehat \theta_k } \nonumber \\
    & \leq \ang*{ \psi_{d_k}(\widehat z) - \psi_{d_k}(z_\star) , \psi_{d_k} (\theta_\star^{d_k}) } + \widetilde \gamma(d_k) \iota_k + 2^{-k}/2  \nonumber \\
    & = h(\widehat z ) -\eta_{d_k}(\widehat z) - h(z_\star) + \eta_{d_k}(z_\star) + \widetilde \gamma(d_k) \iota_k + 2^{-k}/2  \nonumber \\
    & < (2 + \iota_k) \, \widetilde \gamma(d_k)+ 2^{-k}/2  \nonumber \\
    & \leq 2^{-k}/2 + 2^{-k}/2 \label{ms_bai:eq:mis_sub_fixed_budget_optimal_arm} \\
    & = 2^{-k}, \nonumber
\end{align}
where \cref{ms_bai:eq:mis_sub_fixed_budget_optimal_arm} comes from comes from \cref{ms_bai:prop:round_number} combined with the fact that $d_k \geq d_\star$ (as shown in Step 1.1).
As a result, $z_\star$ remains in $\widehat \cS_{k+1}$ according to the elimination criteria.

\textbf{Step 1.3.2: $\widehat \cS_{k+1} \subseteq \cS_{k+1}$.} Consider any $z \in \widehat \cS_k \cap \cS_{k+1}^c$, we know that $\Delta_z \geq 2\cdot 2^{-k}$ by definition. Since $z_\star \in \widehat \cS_k$, we then have 
\begin{align}
    & \ang{ \psi_{d_k}(z_\star) - \psi_{d_k}(z) , \widehat \theta_k } \nonumber \\
    & \geq \ang{ \psi_{d_k}(\widehat z) - \psi_{d_k}(z_\star) , \psi_{d_k} (\theta_\star^{d_k}) } - \widetilde \gamma(d_k) \iota_k - 2^{-k}/2  \nonumber \\
    & = h(z_\star) - \eta_{d_k}(z_\star) - h(z) + \eta_{d_k}(z) - \widetilde \gamma(d_k) \iota_k - 2^{-k}/2  \nonumber \\ 
    & \geq 2 \cdot 2^{-k} - (2 +  \iota_k)\widetilde \gamma(d_k)  - 2^{-k}/2  \nonumber \\
    & = 2 \cdot 2^{-k} - \gamma(d_k)  - 2^{-k}/2  \nonumber \\
    & \geq 2^{-k} \label{ms_bai:eq:mis_sub_fixed_budget_bad_arms},
\end{align}
where \cref{ms_bai:eq:mis_sub_fixed_budget_bad_arms} comes from a similar reasoning as appearing in \cref{ms_bai:eq:mis_sub_fixed_budget_optimal_arm}. As a result, we have $z \notin \widehat \cS_{k+1}$ and $\widehat \cS_{k+1} \subseteq \cS_{k+1}$.

To summarize, we prove the induction at iteration $k+1$, i.e.,
\begin{align*}
    \P \paren*{ \cE_{k+1} \mid \cap_{i< k+1} \cE_{i} } \geq 1 - \delta_k.
\end{align*}

\textbf{Step 2: The error probability.} This step is exactly the same as the Step 2 in the proof of \cref{ms_bai:lm:subroutine_fixed_budget}. Let $\cE = \cap_{i=1}^{n+1} \cE_{i}$ denote the good event, we then have 
\begin{align}
    \P \paren*{ \cE } &  \geq 1- n \abs*{ \cZ }^2 \exp \paren*{ - \frac{ T}{ 4096 \, n \, \rho_{d_\star}^\star (\epsilon) } } \nonumber.
\end{align}
\end{proof}

\paragraph{Proof of \cref{ms_bai:thm:doubling_fixed_budget_mis}}

\noindent

\thmDoublingFixedBudgetMis*

\begin{proof}
The proof follows similar steps as the proof of \cref{ms_bai:thm:doubling_fixed_budget}. Although we are dealing with a misspecified model, guarantees derived in \cref{ms_bai:lm:subroutine_fixed_budget_mis} is similar to the ones in \cref{ms_bai:lm:subroutine_fixed_budget}. 
When $\epsilon \leq \Delta_{\min}$, the proof goes almost exactly the same as the proof of \cref{ms_bai:thm:doubling_fixed_budget} (with $\rho_{d_\star}^\star$ replaced by $\rho_{d_\star(\epsilon)}^\star (\epsilon)$), and \cref{ms_bai:alg:doubling_fixed_budget} identifies the optimal arm. When $\epsilon > \Delta_{\min}$, we additionally replace $\Delta_{\min}$ by $\epsilon$ and equally split the $2\epsilon$ slackness between selection and validation steps. We also slightly modify \cref{ms_bai:lm:psi_ub_lb} to an $\epsilon$-relaxed version (e.g., in the derivation of \cref{ms_bai:eq:psi_ub_lb_Delta_min}, select a $z^\prime \in \cZ$ with sub-optimality gap $\leq \epsilon$ and then replace $\Delta_{\min}$ by $\epsilon$). 
\end{proof}

\subsection{Other Details for Experiments}
\label{ms_bai:app:experiment}

We set confidence parameter $\delta = 0.05$ in our experiments, and generate rewards with Gaussian noise $\xi_t \sim \cN(0,1)$.
We parallelize our simulations on a cluster consists of two Intel® Xeon® Gold 6254 Processors.

Similar to \citet{fiez2019sequential}, we use a Frank-Wolfe type of algorithm \citep{jaggi2013revisiting} with constant step-size $\frac{2}{k+2}$ (we use $k$ to denote the iteration counter in the Frank-Wolfe algorithm) to approximately solve optimal designs. We terminate the Frank-Wolfe algorithm when the relative change of the design value is smaller than $0.01$ or when $1000$ iterations are reached. We use the rounding procedure developed in \citet{pukelsheim2006optimal} to round continuous designs to discrete allocations (with $\zeta=1$, also see \citet{fiez2019sequential} for a detailed discussion on the rounding procedure). In the implementation of \cref{ms_bai:alg:doubling_fixed_confidence}, we set $\gamma_\ell = 4^\ell$, $n_i = 4^i$ and $B_i = 4^{\ell - i}$, which only affect constant terms in our theoretical guarantees. We use a binary search procedure to select $d_k$ in \cref{ms_bai:alg:subroutine_fixed_confidence}. 

\paragraph{Other Experiment Results}
We consider a problem instance with $\cX = \cZ$ being $100$ randomly selected arms from the $D$ dimensional unit sphere.
We set reward function $h(x) = \ang{\theta_\star, x}$ with $\theta_\star = [\frac{1}{1^2}, \frac{1}{2^2}, \dots, \frac{1}{d_\star^2}, 0 \dots, 0]^\top \in \R^D$. We filter out instances whose smallest sub-optimality gap is smaller than $0.08$. We set $d_\star = 5$ and vary the ambient dimension $D\in\curly{25,50,75,100}$. As in \cref{ms_bai:sec:experiment}, we evaluate each algorithm with success rate, (unverifiable) sample complexity and runtime. We run $100$ independent random trials for each algorithm. Due to computational burdens, we force-stop both algorithms after $50,000$ samples; we also force-stop the Frank-Wolfe algorithm when $500$ iterations are reached.

\begin{table}[H]
  \caption{Comparison of success rate with varying ambient dimension.}
  \label{ms_bai:tab:success_rate_2}
  \centering
  \begin{tabular}{lcccc}
    \toprule
          $D $ & $25$   & $50$ & $75$ & $100$ \\
    \midrule
    \rage    & $100\%$ & $100\%$ & $98\%$ & $95\%$  \\
    Ours    & $91\%$ & $98\%$ & $97\%$ & $98\%$   \\
    \bottomrule
  \end{tabular}
\end{table}

\begin{figure}[H]
    \centering
    \includegraphics[width=.8\textwidth]{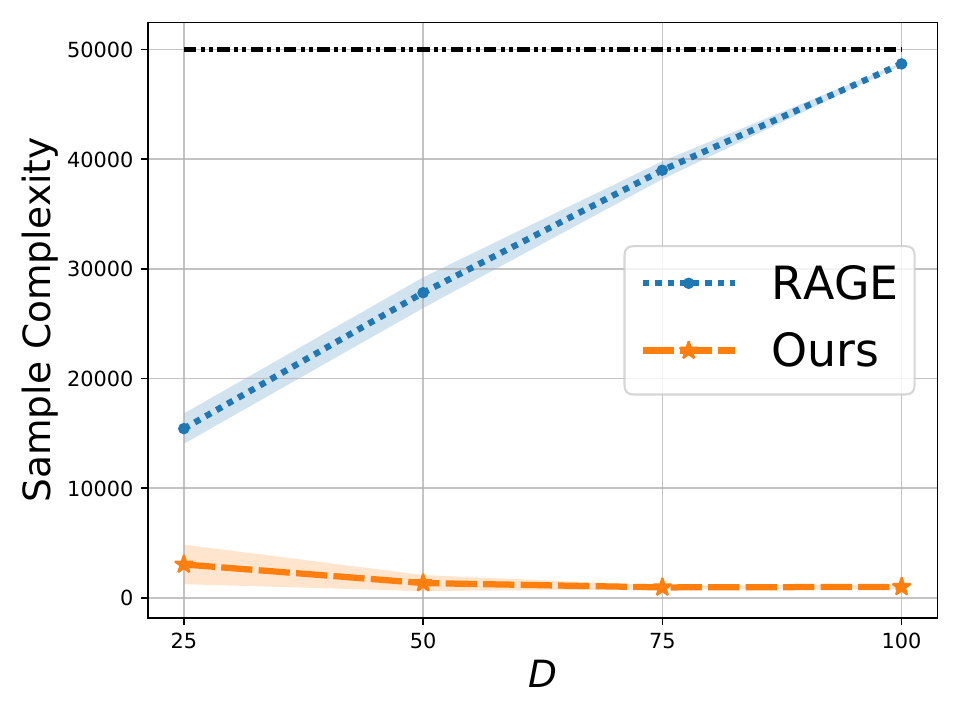}
    \caption{Comparison of sample complexity with varying ambient dimension.}
    \label{ms_bai:fig:comparison_unif}
\end{figure}

Success rates of both algorithms are shown in \cref{ms_bai:tab:success_rate_2}, and \rage shows advantages over our algorithm when $D$ is small. \cref{ms_bai:fig:comparison_unif} shows the sample complexity of both algorithms: Our algorithm adapts to the true dimension $d_\star$ yet \rage is heavily affected by the increasing ambient dimension $D$.

The runtime of both algorithms are shown in \cref{ms_bai:tab:runtime_unif}. \rage shows clear advantage in runtime and our algorithm suffers from computational overheads of conducting model selection.

\begin{table}[H]
  \caption{Comparison of runtime with varying ambient dimension.}
  \label{ms_bai:tab:runtime_unif}
  \centering
  \begin{tabular}{lcccc}
    \toprule
          $D$ & $25$   & $50$ & $75$ & $100$\\
    \midrule
    \rage    & $85.99\,$s & $144.78\,$s & $249.79\,$s & $357.98\,$s \\
    Ours    & $287.09\,$s & $339.67\,$s & $489.50\,$s & $678.93\,$s  \\
    \bottomrule
  \end{tabular}
\end{table}

We remark that, for the current experiment setups with $d_\star$ and $D\in\curly{25,50,75,100}$, our algorithm does not perform well if $\theta_\star$ is chosen to be flat, e.g., \linebreak 
$\theta_\star = [\frac{1}{\sqrt{d_\star}},\dots,\frac{1}{\sqrt{d_\star}},0,\dots,0]^\top \in \R^D$. However, we believe that one will eventually see model selection gains if $D$ is chosen to be large enough (and allowing each algorithm takes more samples before force-stopped).
One may need to overcome the computational burdens, e.g., developing practical (or heuristic-based) implementations of our algorithm and \rage, before running experiments in higher dimensional spaces. We leave large-scale evaluations for future work.

\bibliographystyle{mcbride}
\bibliography{refs}

\end{document}